\documentclass[sigconf]{acmart}

\usepackage{color}
\usepackage{soul}
\usepackage{graphicx, animate}
\usepackage{tabularx}
\usepackage{caption}
\usepackage{subcaption}
\usepackage{wrapfig}

\usepackage{amssymb}
\usepackage{amsmath}
\usepackage[table,xcdraw]{xcolor}
\usepackage{tikz}
\usetikzlibrary{spy}
\usepackage[utf8]{inputenc}
\usepackage{pgfplots}
\pgfplotsset{compat=1.18}
\usepackage{pifont}
\usepackage{makecell}
\usepackage{float}
\usepackage{array}
\usepackage{longtable}
\usepackage{booktabs}
\usepackage{multicol}
\usepackage{multirow}
\usepackage{svg}
\usepackage{amsmath}
\usepackage{dblfloatfix}

\usepackage{soul}
\sethlcolor{blue!15}
\soulregister{\cite}{7}     
\soulregister{\ref}{1}      
\soulregister{\pageref}{1}  
\soulregister{\eqref}{1}    

\makeatletter
\@addtoreset{equation}{section}
\makeatother

\usepackage[ruled]{algorithm2e} 

\SetAlFnt{\small}
\SetAlCapFnt{\small}
\SetAlCapNameFnt{\small}
\SetAlCapHSkip{0pt}

\newcommand{\cmark}{\color{darkgreen}{\ding{51}}}
\newcommand{\xmark}{\color{darkred}{\ding{55}}}

\newcommand{\PaperName}{ASIA}

\definecolor{title_purple}{rgb}{0.65,0.1,0.65}
\definecolor{title_col_1}{rgb}{0.827703, 0.147925, 0.127676}
\definecolor{title_col_2}{rgb}{0.842658, 0.229482, 0.179246}
\definecolor{title_col_3}{rgb}{0.856618, 0.293874, 0.225622}
\definecolor{title_col_4}{rgb}{0.869563, 0.350689, 0.269494}

\definecolor{darkred}{rgb}{0.7,0.1,0.1}
\definecolor{darkgreen}{rgb}{0.1,0.6,0.1}
\definecolor{green}{rgb}{0, 0.5, 0}
\definecolor{orange}{rgb}{0.8, 0.6, 0.2}
\definecolor{red}{rgb}{1.0, 0.0, 0.0}
\definecolor{teal}{rgb}{0.0, 0.4, 0.4}
\definecolor{purple}{rgb}{0.65,0.0,0.65}
\definecolor{saffron}{rgb}{0.95,0.75,0.2}
\definecolor{turquoise}{rgb}{0.0,0.5,0.5}
\definecolor{black}{rgb}{0.0, 0.0, 0.0}
\definecolor{gray}{rgb}{0.5, 0.5, 0.5}

\newcommand{\am}[1]{{\color{black}#1}}

\newcommand{\rzz}[1]{{\color{black}#1}}
\newcommand{\sr}[1]{{\color{black}#1}}

\makeatletter
\@ifundefined{@titlefont}
  {\newcommand{\SMtitlefont}{\LARGE\bfseries}}
  {\newcommand{\SMtitlefont}{\@titlefont}}
\@ifundefined{@subtitlefont}
  {\newcommand{\SMsubtitlefont}{\large\itshape}}
  {\newcommand{\SMsubtitlefont}{\@subtitlefont}}
\makeatother

\copyrightyear{2025}
\acmYear{2025}
\setcopyright{cc}
\setcctype{by}
\acmConference[SA Conference Papers '25]{SIGGRAPH Asia 2025 Conference Papers}{December 15--18, 2025}{Hong Kong, Hong Kong}
\acmBooktitle{SIGGRAPH Asia 2025 Conference Papers (SA Conference Papers '25), December 15--18, 2025, Hong Kong, Hong Kong}
\acmDOI{10.1145/3757377.3763821}
\acmISBN{979-8-4007-2137-3/2025/12}

\begin{document}
\title{\textcolor{title_col_1}{A}\textcolor{title_col_2}{S}\textcolor{title_col_3}{I}\textcolor{title_col_4}{A}: \textcolor{title_col_1}{A}daptive 3D \textcolor{title_col_2}{S}egmentation 
 using Few \textcolor{title_col_3}{I}mage \textcolor{title_col_4}{A}nnotations}

\author{Sai Raj Kishore Perla}
\affiliation{%
 \institution{Simon Fraser University}
 \country{Canada}
}
\email{srp7@sfu.ca}

\author{Aditya Vora}
\affiliation{%
 \institution{Simon Fraser University}
 \country{Canada}
}
\email{ava40@sfu.ca}

\author{Sauradip Nag}
\affiliation{%
 \institution{Simon Fraser University}
 \country{Canada}
}
\email{snag@sfu.ca}

\author{Ali Mahdavi-Amiri}
\affiliation{%
 \institution{Simon Fraser University}
 \country{Canada}
}
\email{amahdavi@sfu.ca}

\author{Hao Zhang}
\affiliation{%
 \institution{Simon Fraser University}
 \country{Canada}
}
\email{haoz@sfu.ca}

\begin{abstract}
We introduce ASIA (\textit{\underline{A}daptive 3D \underline{S}egmentation using few \underline{I}mage \underline{A}nnotations}), a novel framework that enables segmentation of possibly non-semantic and non-text describable "parts" in 3D. Our segmentation is controllable through a few user-annotated in-the-wild images, which are easier to collect than multi-view images, less demanding to annotate than 3D models, and more precise than potentially ambiguous text descriptions. Our method leverages the rich priors of text-to-image diffusion models, such as Stable Diffusion, to transfer segmentations from image space to 3D, even when the annotated and target objects differ significantly in geometry or structure.
\sr{
During training, we optimize a text token for each segment and fine-tune our model with a novel cross-view part correspondence loss. At inference, we segment multi-view renderings of the 3D mesh, fuse the labels in UV-space via voting, refine them with our novel \textit{Noise Optimization} technique, and finally map the UV-labels back onto the mesh.
}
ASIA provides a practical and generalizable solution for both semantic and non-semantic 3D segmentation tasks, outperforming existing methods by a noticeable margin in both quantitative and qualitative evaluations.

\vspace{3pt}
\noindent
\textbf{Project Page:} \href{https://sairajk.github.io/asia/}{\texttt{https://sairajk.github.io/asia}}

\end{abstract}

%
%


%
%


\begin{teaserfigure}
\centering
  \includegraphics[width=\textwidth]{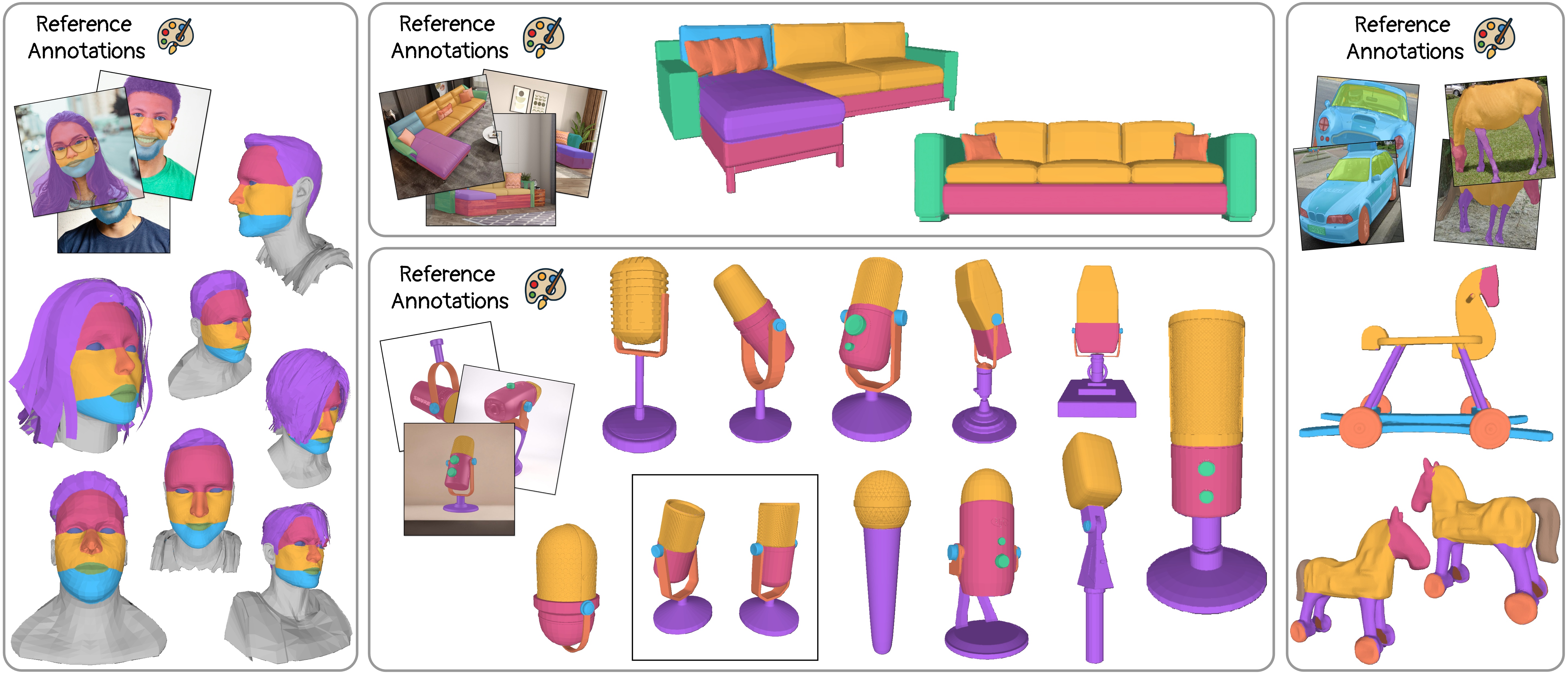}
  \caption{
  Our method ASIA (\underline{A}daptive 3D \underline{S}egmentation using few \underline{I}mage \underline{A}nnotations) is trained on a few user-annotated \emph{images}, \textit{i.e.}, the "Reference Annotations", and produces a part segmentation for a given 3D shape that adheres to the references. See Fig.~\ref{fig:asia_examples} for usage scenarios involving adaptive shoe, face, and sofa bed segmentations. Note how our method generates accurate segmentations despite geometric (\textit{e.g.}, the hair) and structural (\textit{e.g.}, the microphones and sofa beds) variations in the 3D shapes. The two microphones in the box were treated as a {\em single shape\/}. The rightmost example shows that our method can adapt reference annotations of cars and horses to properly segment their "hybrids", namely, 3D ponycycles. These results also highlight the model’s ability to generalize effectively to objects that differ substantially from the training images.
  }
  \Description{Teaser Figure.}
  \label{fig:teaser}
\end{teaserfigure}

\maketitle

\section{Introduction}
\label{sec:introduction}

3D segmentation \rzz{remains one of the prominent problems} in computer graphics and vision. It allows for precise control \rzz{and attribute transfer} over individual components of 3D models, facilitating tasks such as texture mapping, animation, and realistic rendering. It also helps in 3D object understanding, \rzz{manipulation}, and scene reconstruction, supporting applications such as tracking and robotic interaction.  Accurately segmenting objects into meaningful parts enhances both the functionality \rzz{and customizability} of digital 3D models, benefiting industries ranging from \rzz{design, automation, manufacturing, to eCommerce and entertainment.}

\begin{figure*}[!t]
  \includegraphics[width=\textwidth]{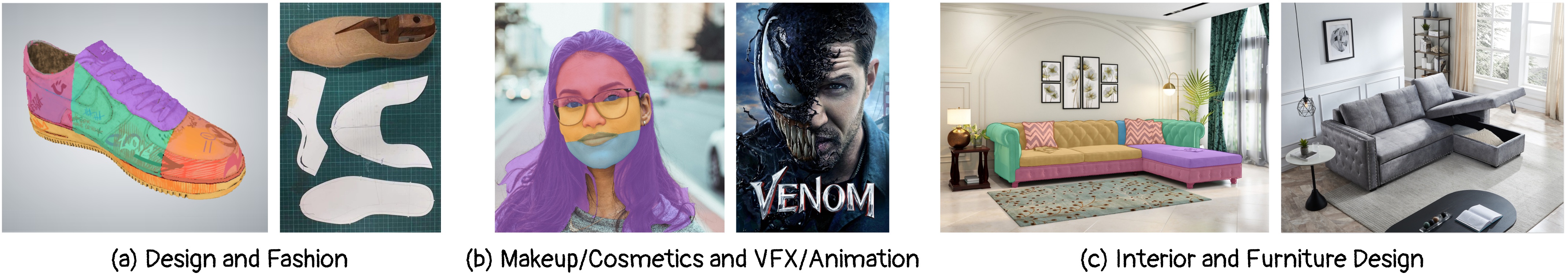}
    \caption{
\rzz{Potential applications of \textit{adaptive} 3D Part Segmentation}. (a) \rzz{Design and fashion}, \textit{e.g.}, pattern cutting for shoes and garments. (b) \rzz{Makeup/cosmetics (left) and VFX/animation (right)}, enabling localized editing or simulation. (c) Interior and furniture design, for assigning different colors or materials to specific \rzz{functional parts}, or for splitting furniture to ease assembly or packing. Despite its broad range of applications, \rzz{such adaptive segmentations remain underexplored}.
}
\label{fig:asia_examples}
\end{figure*}

\begin{table}[]
\centering
\caption{
A comparative summary of related works, highlighting their ability to segment \emph{adaptive} parts, the types of data required to guide segmentation, and the foundation model(s) employed. See Sec.~\ref{sec:related_works} for more details.
}
\label{tab:rel_works_summary}

\resizebox{0.93\columnwidth}{!}{%
\begin{tabular}{ccccc}
\toprule
 & Methods & \begin{tabular}[c]{@{}c@{}}Adaptive Part\\ Segmentation\end{tabular} & \begin{tabular}[c]{@{}c@{}}Segmentation\\ Guidance\end{tabular} & \begin{tabular}[c]{@{}c@{}}Foundation\\ Model(s)\end{tabular} \\ \midrule
 & 3D Highlighter & \xmark & Text & CLIP \\
 \arrayrulecolor{lightgray}\cmidrule(lr){2-5}
 \arrayrulecolor{black}
 & SATR & \xmark & Text & GLIP \\
 \arrayrulecolor{lightgray}\cmidrule(lr){2-5}
 \arrayrulecolor{black}
\multirow{-4.05}{*}{\rotatebox[origin=c]{90}{Zero-Shot}} & MeshSegmenter & \xmark & Text & GLIP, SAM \\  

\arrayrulecolor{gray}\cmidrule(lr){1-5}
\arrayrulecolor{black}
 & PartSLIP & \xmark & Text, 3D Shapes & GLIP \\
 \arrayrulecolor{lightgray}\cmidrule(lr){2-5}
 \arrayrulecolor{black}
 & PartSTAD & \xmark & Text, 3D Shapes & GLIP, SAM \\
 \arrayrulecolor{lightgray}\cmidrule(lr){2-5}
 \arrayrulecolor{black}
 & PartDistill & \xmark & Text, 3D Shapes & GLIP, CLIP \\
 \arrayrulecolor{lightgray}\cmidrule(lr){2-5}
 \arrayrulecolor{black}
 & 3$\times$2 & \xmark & Images & SD, SAM \\
 \arrayrulecolor{lightgray}\cmidrule(lr){2-5}
 \arrayrulecolor{black}
\multirow{-7.05}{*}{\rotatebox[origin=c]{90}{Few-Shot}} & Ours & \cmark & Images & SD \\
\bottomrule
\end{tabular}%
}
\end{table}


\rzz{The key challenge to object segmentation, whether in 2D or 3D, is that there is \emph{no} universal "truth" on what a "part" is. To date, {\em semantic\/} segmentation has played a dominant role in the research literature, where the most advanced methods have been data-driven~\cite{he2024deeplearningbased3d}, resorting to supervised or semi-supervised learning which necessitates human annotations to set up ground-truth segmentations. In the 3D case, however, manually segmenting 3D objects is a demanding task. With the emergence of large foundational models (LFMs), zero-shot and text-driven segmentations have gained significant popularity in the image/video domains, as exemplified by the Segment Anything Model (SAM) \cite{SAM_SegmentAnything} and its follow-ups. Unfortunately, 3D LFMs are still elusive. Most critically, regardless of the training setup, the resulting model is always optimized over a large and often highly diverse data collection.}

\rzz{We stipulate that, in practice and ultimately, the most desirable object segmentation should be one that is \emph{adaptive}, catering to {\em specific\/} application or usage scenarios, which may be uncommon. Accordingly, the associated segmentation criteria may go beyond semantics and relate to material, function, assembly, or design intents. Furthermore, some of the parts may not be easily or naturally describable by texts, \textit{e.g.}, the middle part on one's face between the eye and lip lines. Fig.~\ref{fig:asia_examples} illustrates several examples.}

%


\rzz{In this paper, we introduce {\em \underline{A}daptive 3D \underline{S}egmentation} using few {\em \underline{I}mage \underline{A}nnotations}, a novel task, coined ASIA, which allows segmentation of possibly non-semantic and non-text describable "parts" in 3D. 
Our segmentation is {\em controllable\/} through a few user-annotated on-line or in-the-wild images, which are easier to gather than multi-view images, less demanding to annotate than 3D models, and more precise than potentially ambiguous text descriptions. Aiming to develop a method that generalizes well across a wide range of object categories and adapts effectively to diverse shapes within the same category,
}
we utilize the 
rich priors of text-to-image diffusion models to transfer object segmentations from images to 3D. 
Importantly, the annotated objects are not the same as the target 3D object; they may vary both in geometry and structure; see Fig.~\ref{fig:teaser}.

While methods do exist for transferring segmentations between images~\cite{khani2023slime}, directly applying them to 3D objects is not straightforward, as segmentations from different views may be inconsistent. To address this issue and effectively solve our generalizable 2D-to-3D segmentation transfer problem, we first leverage Stable Diffusion (SD)'s~\cite{rombach2021highresolution} emergent features for segmentation tasks. By training the text tokens of the diffusion model to represent segmentation parts, and rendering the 3D object from multiple views, we first segment each view independently. To enhance segmentation, we use ControlNet that is conditioned on edges to obtain structural priors and more hints about the potential boundaries of segments. We also fine-tune SD using LoRA~\cite{hu2021lora} as optimizing the text embedding alone may not be sufficient for all segments.

To enhance multi-view consistency during training, we introduce a consistency loss between the features of cross-attentions in SD that correspond to the same part. During inference, after segmenting each view, we project the segmentations onto a UV map of the 3D object. Since the same point on the UV map may receive conflicting labels after projection from different views, we resolve this by applying a voting mechanism to ensure label consistency. To further refine the results and better account for the segmentation of each view, we optimize for a per-view noise that balances the individual view segmentations with the voting results.

\rzz{Our approach offers a practical, controllable, and data-efficient solution for few-shot 3D object segmentation from image annotations --- the first of its kind. It taps into the power of text-image LFMs, with light annotation efforts, for generalizable 3D part segmentation, where semantic or text-describable segmentation can be seen as a special case.
Through extensive experiments, we demonstrate significant improvements of our method in segmentation accuracy, especially for adaptive and application-/usage-specific segmentation tasks. At the same time, our method also outperforms existing approaches on semantic part segmentation by a significant margin. We show the efficacy and generalizablity of our method through several applications, quantitative and qualitative results, as well a series of ablation
studies to justify our design choices.}

\section{Related Works}
\label{sec:related_works}

3D object segmentation is a popular topic that has been extensively explored~\cite{shamir2006segmentation, wang_projective}. Traditionally, segmentation approaches rely on either annotated datasets for training~\cite{qi2017pointnet, qi2017pointnet++} or algorithms (\textit{e.g.}, graph cuts or skeletonization) that utilize geometric cues to distinguish between segments~\cite{podolak2006planar,katz2003hierarchical}. However, the emergence of foundation models, like the vision language models (VLMs) and diffusion-based methods, has enabled more flexible, fine-grained segmentation. This section discusses related methods, and Tab.~\ref{tab:rel_works_summary} provides a summary comparing key approaches with our work.




\subsubsection*{3D Shape Segmentation with VLMs}
Compared to image segmentation~\cite{zhou2024image}, progress in 3D part segmentation~\cite{yu2023hal3d, jones2022SHRED} has been relatively limited, primarily due to the lack of high quality large scale datasets with part annotations~\cite{mo2019partnet, Xiang_2020_SAPIEN}. 
However, recent advancements in large-scale vision models~\cite{SAM_SegmentAnything, ravi2024sam2} have motivated efforts to leverage their rich semantic knowledge in multi-view settings~\cite{yang2024sampart3d}. This enables 3D part segmentation while minimizing annotation requirements. Some of these models, namely Vision-Language Models (VLMs) such as GLIP~\cite{li2022grounded}, support open-vocabulary localization, enabling users to specify regions of interest through text prompts.
3D Highlighter~\cite{decatur20233d} leverages CLIP~\cite{radford2021learning} to optimize the part location by minimizing the difference between features of the query text prompt and the rendered images of the part-localized 3D mesh.
Likewise, 3D PaintBrush~\cite{decatur2024paintbrush} uses SD~\cite{rombach2021highresolution} to perform text-guided localized texturing through optimization. However, similar to 3D Highlighter~\cite{decatur20233d}, it is also limited to text-describable parts for localization.
iSeg~\cite{lang2024iseg} uses SAM~\cite{SAM_SegmentAnything} to perform interactive segmentation; however, it relies on costly per-shape optimization and is limited to user-click control.
In contrast, methods such as SATR~\cite{abdelreheem2023satr}, MeshSegmenter~\cite{zhong2024meshsegmenter}, PartSLIP~\cite{liu2023partslip}, and PartSTAD~\cite{kim2024partstad} first extract 2D bounding boxes and segmentation masks of parts from multi-view renderings of the 3D shape using foundational VLMs~\cite{li2022grounded, radford2021learning} and then lift the predictions to 3D through a voting scheme.
On the other hand, PartDistill transfers the knowledge of text-describable parts from a pre-trained VLM~\cite{radford2021learning, li2022grounded} to a 3D point cloud segmentation network~\cite{zhang2022point}.


\subsubsection*{Segmentation with Diffusion Models}
Diffusion models~\cite{sohl2015deep, song2019generative, ho2020denoising} have demonstrated remarkable generative capabilities~\cite{rombach2021highresolution, baldridge2024imagen, blattmann2023stable, alliegro2023polydiff, jun2023shap}. 
A key feature of these models is their rich semantic understanding, obtained as a by-product of self-supervised training of the diffusion process~\cite{tang2023emergent, hedlin2024unsupervised, el2024probing}. Building on these insights, the follow-up works~\cite{khani2023slime, alimohammadi2024smitesegmenttime, namekata2024emerdiff} leveraged intermediate features of a diffusion model for few-shot generative image \am{and video} segmentation through personalization~\cite{gal2022image}.
3$\times$2~\cite{3by2} uses diffusion features and SAM~\cite{SAM_SegmentAnything} segments to find correspondence between annotated references and rendered images of the shape to generate 3D segmentation labels.

\subsubsection*{Adaptive Part Segmentation}
All these recent 3D shape segmentation methods~\cite{3by2, kim2024partstad, liu2023partslip} somehow rely on "part" priors from 2D foundational localization models, like GLIP~\cite{li2022grounded} and SAM~\cite{SAM_SegmentAnything}, and unsupervised 3D clustering techniques, like Super Points~\cite{landrieu2018large}, for segmentation.
Vision language models (\textit{e.g.}, CLIP, GLIP) are trained to associate visual content with text descriptions that are interpretable by humans. Consequently, 3D shape segmentation methods~\cite{abdelreheem2023satr, decatur20233d} built upon such models struggle to detect parts that lack clear semantic meaning or are hard to describe accurately by text.
%
In contrast, \PaperName{} is the first method that \textit{only} utilizes a 2D generative foundation model, like SD, for 3D shape segmentation. It allows users to control segmentation granularity and can even segment parts that may not be easily described with text. 
Moreover, this is achieved using only a few in-the-wild reference images during training, and the model demonstrates strong generalization to diverse 3D shapes during inference (see Fig.~\ref{fig:teaser}).
\section{Method}
\label{sec:method}

We propose \PaperName{}, a method to partition 3D shapes into user-defined segments, using only a \textit{few segmented images} as reference. Unlike previous works~\cite{liu2023partslip, kim2024partstad, umam2024partdistill} that rely on a \textit{few segmented meshes}, which are often difficult to obtain due to the expertise needed with 3D modeling software, our method \am{only requires} 2D segmentation references.
In the following, we first describe the problem setting in Sec.~\ref{subsec:setting} and then provide the concepts relevant to this work in Sec.~\ref{subsec:preliminary}, followed by the details on training our model in Sec.~\ref{subsec:training}, and finally multiview-consistent inference for segmenting 3D shapes in Sec.~\ref{subsec:inference}.

\subsection{Problem Setting}
\label{subsec:setting}
Given a set of $N$ \am{training} images, $\{\mathbf{I}_j\}_{j=1}^N$, possibly from different viewpoints and objects within the same category (\textit{e.g.}, microphones), along with their corresponding part annotations, $\{\mathbf{M}_j\}_{j=1}^N$, our goal is to learn a multi-view consistent model for segmenting parts of any 3D shape, $\mathcal{S}$, from the same category. Each $\mathbf{M}_j^r$, $r \in R$, is a binary mask indicating the region of part $r$ (\textit{e.g.}, base) in image $\mathbf{I}_j$, and $R$ represents the total number of parts for segmentation for that category.
We make no assumptions about the training data, except for images having good coverage of the object, as this helps achieve better 3D segmentation and generalization during inference. Consequently, each image in our training data may depict a different object in any pose and one can easily collect such images from multiple online sources.

\subsubsection*{Model overview}
We leverage the structural and contextual information embedded in the latent space of a pre-trained text-to-image diffusion model~\cite{rombach2021highresolution} to effectively learn and propagate part annotations from a limited set of segmented images, acquired from varying viewpoints, to a novel 3D shape, $\mathcal{S}$, at inference time.
To segment a 3D shape, we first render a few images from different views, segment these images using our trained model, and then lift these multi-view 2D labels to 3D for final segmentation.

However, achieving high-quality 3D segmentation requires consistent multi-view 2D segmentations, which depends on the model’s ability to share information across different views. To enforce this, we introduce a cross-view part correspondence loss during training. Additionally, we ensure multi-view consistency during inference by projecting the model's predictions into a common \textit{UV}-space for aggregation.
We also propose \textit{Noise Optimization}, a novel test-time optimization technique to further refine the predicted labels.
Our training and inference pipelines are shown in Fig.~\ref{fig:training_pipeline} and Fig.~\ref{fig:inference_pipeline}, respectively.

\subsection{Preliminary}
\label{subsec:preliminary}
\subsubsection*{Latent Diffusion Models (LDMs)}
LDMs are a class of generative models that learn the data distribution by applying the diffusion process~\cite{sohl2015deep, ho2020denoising} in the latent space of an autoencoder~\cite{kingma2014vae}.
To train a text-conditioned LDM~\cite{rombach2021highresolution}, a tokenized language prompt $\mathcal{P}$ is processed by a text encoder, $\mathcal{E}_{txt}$, to produce $\mathbf{f_{txt}} = \mathcal{E}_{txt}(\texttt{tokenize(}\mathcal{P}\texttt{)})$.
The image $I$ corresponding to the prompt $\mathcal{P}$ is encoded to $\mathbf{z_0}$ by the autoencoder. It is then perturbed with Gaussian noise $\epsilon \sim \mathcal{N}(0, \mathcal{I})$, determined by a timestep $t \in [1,T]$ and a predefined noise schedule, to produce $\mathbf{z_t} = \sqrt{\alpha_t}\,\mathbf{z_0} + \sqrt{1 - \alpha_t}\,\epsilon$, where $\alpha_t = \prod_{s=1}^t (1-\beta_s)$ with $\{\beta_t\}_{t=1}^T$ denoting the variance schedule.
The LDM, $\epsilon_{\theta}(\cdot)$, is then trained to minimize:
\begin{equation}
\label{eq:sd_loss}
    \mathcal{L}_{\text{\textit{LDM}}} = \mathbb{E}_{\mathbf{z_t}, \epsilon, t} \left[ \|\epsilon - \epsilon_{\theta}(\mathbf{z_t}, t,\mathbf{f_{txt}})\|^2 \right]
\end{equation}


\subsubsection*{Segmentation with LDMs}
SD employs a UNet-based~\cite{ronneberger2015u} model for denoising, using residual convolution layers for feature extraction followed by attention layers. Self-attention layers capture the global context, while cross-attention layers incorporate the conditioning information from text embeddings $\mathbf{f_{txt}}$ to the intermediate latents $\mathbf{z_t}$.

SLiMe~\cite{khani2023slime} demonstrated SD’s ability to segment unseen images in a one-shot (or few-shot) manner by optimizing a small set of text embeddings to align the attention maps with the given segmented reference(s). Each segment is represented by a separate embedding of $\mathbf{f_{txt}}$, enabling multi-part segmentation. 
It uses Weighted Accumulated Self-attention (WAS) map, which combines cross- and self-attention features for image segmentation.

To compute WAS, we first define the normalized attention map extracted from an intermediate layer $l$ of SD as $F^l_{xa} = \texttt{Softmax}\!\left(\frac{Q^l (K^l)^\top}{\sqrt{d^l}}\right)$, where $Q^l$ and $K^l$ denote the query and key projections, and $d^l$ is their feature dimension. For self-attention, $F^l_{sa}$, both $Q^l$ and $K^l$ are derived from the intermediate latents $\mathbf{z_t}$. For cross-attention, $F^l_{ca}$, the queries $Q^l$ are still computed from $\mathbf{z_t}$, while the keys $K^l$ are obtained from the text embeddings $\mathbf{f_{txt}}$. Given this, we then compute the averaged attention maps as:
\begin{equation}
\label{eq:avg_attn_feats}
\begin{gathered}
    F_{ca} = \frac{1}{|l_{ca}|}\sum_{l \in l_{ca}}F_{ca}^l, \quad
    F_{sa} = \frac{1}{|l_{sa}|}\sum_{l \in l_{sa}}F_{sa}^l
\end{gathered}
\end{equation}
where $l_{ca}$ and $l_{sa}$ denote the layers used to compute $F_{ca}$ and $F_{sa}$, respectively.
We represent $F_{sa}$ and $F_{ca}$ as having shapes $[hw, hw]$ and $[R, h, w]$, respectively, where $h$ and $w$ denote the spatial dimensions of the image latents $\mathbf{z_0}$, and the channel dimensions are omitted for simplicity. Also, since the attention maps from different layers may vary in resolution, they are resized to a common spatial size before averaging.
We then flatten $F_{ca}$ to shape $[R, hw]$ and multiply it with $F_{sa}$ to obtain the WAS map $\mathbf{W}$. The map $\mathbf{W}$ is then reshaped from $[R, hw]$ to $[R, h, w]$, resized to produce image-shaped segmentation masks, and finally softmax-ed along the $R$ dimension to assign per-pixel part probabilities. Mathematically,
\begin{equation}
\label{eq:was}
\begin{gathered}
    \mathbf{W} = \texttt{Sum(Flatten(}F_{ca}\texttt{)} \odot F_{sa}\texttt{)} \\
    \mathbf{W} = \texttt{Softmax(Resize(Reshape(}\mathbf{W}\texttt{)))}
\end{gathered}
\end{equation}

SLiMe~\cite{khani2023slime} uses the following multi-term loss function to optimize the text embeddings:
\begin{equation}
    \mathcal{L}_{SLiMe} = \mathcal{L}_{CE} + \mu\mathcal{L}_{MSE} + \rho\mathcal{L}_{LDM}
    \label{eq:slime_loss}
\end{equation}
where $\mu$ and $\rho$ are weights for the losses. $\mathcal{L}_{CE}$ computes the cross-entropy between $F_{ca}$ and the resized GT masks, this enforces an intermediate alignment between the cross-attention features and coarse low-resolution segmentation masks. $\mathcal{L}_{MSE}$ is the MSE between the resized WAS map $\mathbf{W}$ and the GT segmentations, enforcing $\mathbf{W}$ to match the per-class masks at higher resolution. Finally, we have $\mathcal{L}_{LDM}$ (Eq.~\ref{eq:sd_loss}), the standard denoising objective, which ensures the optimized embeddings remain within the model distribution.
We refer readers to SLiMe~\cite{khani2023slime} for more details.

\begin{figure}[t]
  \includegraphics[width=\linewidth]{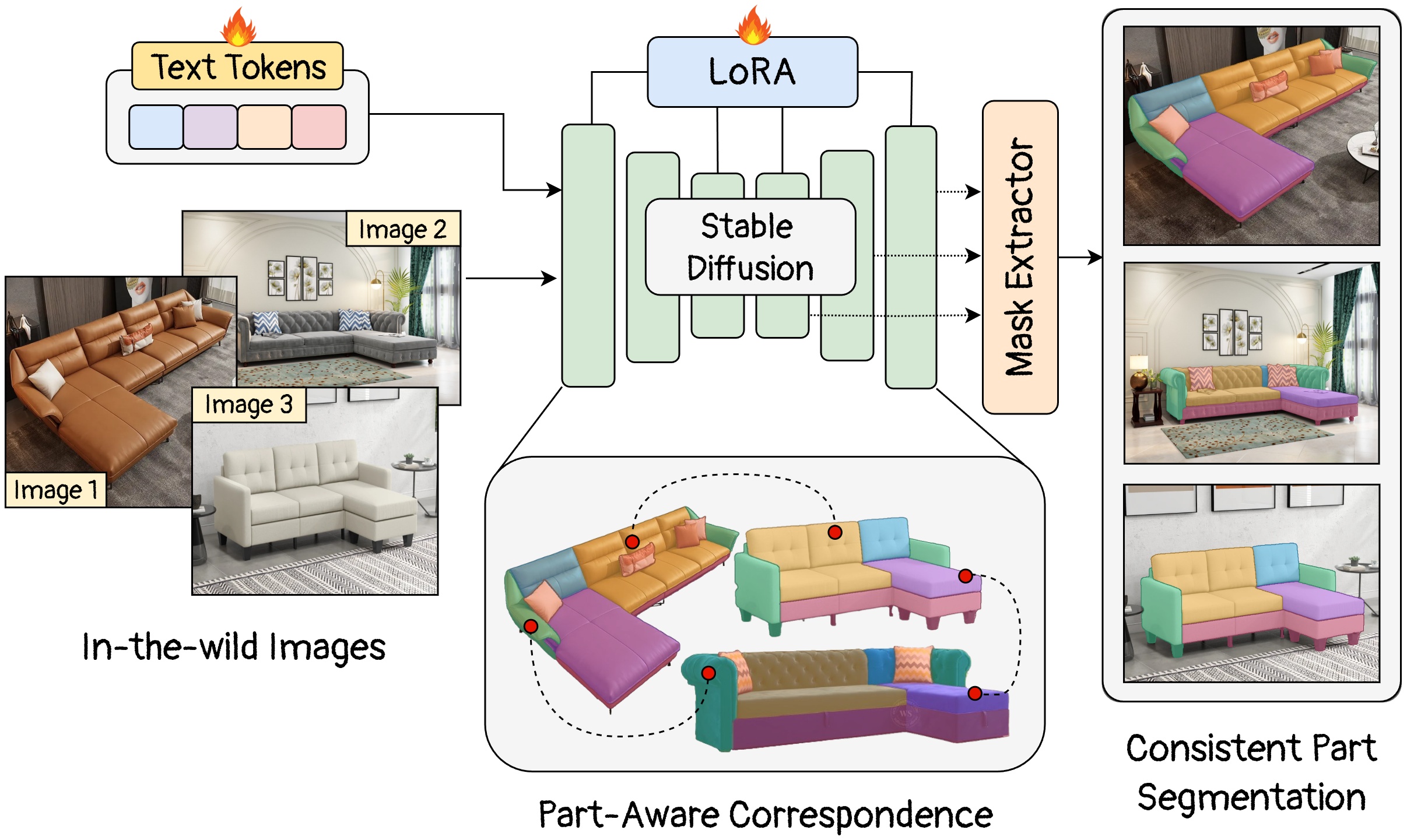}
\caption{
\emph{Training Pipeline.} Given a few in-the-wild images and their segmentations, \PaperName{} learns a set of tokens, one for each \emph{part}, which align the SD attention maps to localize the segments.
It also finetunes the UNet through LoRA with our \emph{part-aware correspondence} loss ($\mathcal{L}_{corr}$ of Eq.~\ref{eq:corr_loss}) to make sure the predicted segments are consistent across views.
\emph{Mask Extractor} computes the segmentation masks $\mathbf{W}$ using features from multiple SD layers (as in Eq.~\ref{eq:avg_attn_feats} and Eq.~\ref{eq:was}).
}
\label{fig:training_pipeline}
\end{figure}

\subsection{Correspondence-aware Segmentation}
\label{subsec:training}
In \PaperName{}, we first optimize a text embedding for each segment across different views, which is then used to segment a given 3D mesh at inference. To further refine segmentation quality, we fine-tune the entire model using LoRA~\cite{hu2021lora}, as text embeddings alone may not produce accurate segmentations (see Tab. \ref{tab:ablation_combined} and Fig. \ref{fig:model_ablation}).

\subsubsection*{Per-view Part Segmentation}
We input the reference images $\{\mathbf{I}_j\}_{j=1}^N$ and text embeddings $\mathbf{f_{txt}}$ to \PaperName{}, denoted as $\psi(\cdot)$, to obtain the segmentation masks $\{\mathbf{W}_j\}_{j=1}^N$. This is formulated as $\mathbf{W}_j = \psi( \mathbf{I}_j,\mathbf{f_{txt}})$, where each $\mathbf{W}_j$ is given by Eq.~\ref{eq:was}.
The text embeddings $\mathbf{f_{txt}}$, initialized randomly or using part names, correspond to the reference masks $\{\mathbf{M}_j^r\}_{r=1}^R$.
The ground-truth masks $\{\mathbf{M}_j\}_{j=1}^N$ and the predicted masks $\{\mathbf{W}_j\}_{j=1}^N$ are then used to optimize the text embeddings.

We use the same losses as SLiMe~\cite{khani2023slime} ($\mathcal{L}_{CE}$, $\mathcal{L}_{MSE}$ and $\mathcal{L}_{LDM}$ in Eq.~\ref{eq:slime_loss}) to optimize the text embeddings. However, the masks generated by \PaperName{} after this optimization remain independent across views, meaning there is no direct correspondence between segmentations from different views. 
As a result, variations in appearance across views can lead to inconsistent part segmentation. To mitigate this, we leverage part-level correspondences in the reference images and use them to further fine-tune \PaperName{}.
\subsubsection*{Cross-view Part Correspondence}
To ensure multi-view consistency in the predicted segmentation masks across different views, we establish correspondences among the attention features, as they directly influence the WAS maps $\mathbf{W}$ (Eq.~\ref{eq:was}).
As mentioned earlier, each token in the text embedding $\mathbf{f_{txt}}$ corresponds to the same part across all reference images (see Fig.~\ref{fig:training_pipeline}). 
Since text tokens are involved in cross-attention formation in SD, the cross-attention features, $F_{ca}$, serve as a crucial component for consistent segmentation across different views.
Building on this intuition, we enforce part-level correspondences on the cross-attention features, $F_{ca}$, using \emph{cosine similarity}. 
First, we select all pixels $\mathbf{P}_i^r$ corresponding to a part $r$ in a reference segmentation $\mathbf{M}_i$. Next, we identify all pixels $\mathbf{P}_j^r$ corresponding to the same part $r$ in another randomly chosen reference segmentation $\mathbf{M}_j$, and align their features using a correspondence loss defined as:
\begin{equation}
\begin{gathered}
    \mathcal{L}_{corr} = \frac{1}{R} \sum_{r=1}^R\left[\frac{1}{|\mathbf{P}_i^r|} \sum_{p_i \in \mathbf{P}_i^r}(1 - \mathrm{cos}(F^i_{ca}(p_i), F^j_{ca}(p_j)) \right], \\
    \text{where, }\,p_j = \underset{p \in \mathbf{P}_j^r}{\mathrm{\bf arg\,min}} \, cos(F^i_{ca}(p_i),F^j_{ca}(p))
\end{gathered}
\label{eq:corr_loss}
\end{equation}
where $cos(\cdot)$ denotes the cosine similarity score, $F^j_{ca}(x)$ denotes the cross-attention feature corresponding to pixel $x$ of image $\mathbf{I}_j$, and ${\mathrm{\bf arg\,min}}$ indicates that we select the feature location, $p_j$, with the lowest cosine similarity to $F^i_{ca}(p_i)$ for computing the loss.
This loss enforces cross-view consistency by ensuring cross-attention features, $F_{ca}$, for part $r$ remain consistent across different views.


\begin{figure*}[t]
  \includegraphics[width=\textwidth]{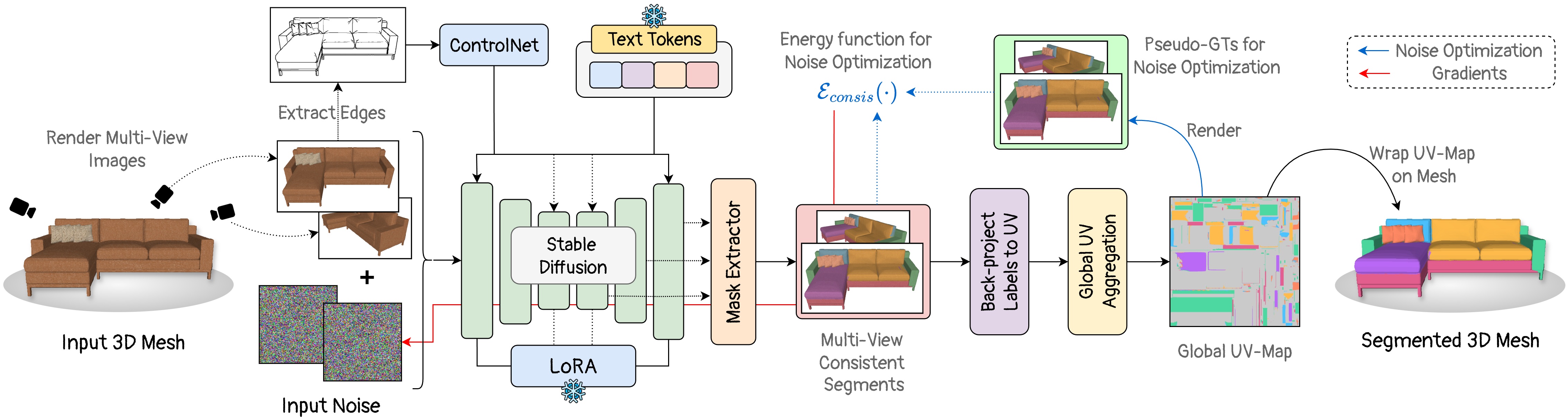}
  \caption{
  \emph{Inference Pipeline.} Given an input 3D mesh, we first render multi-view RGB images and noise them to prepare the input to our model. We also extract geometric edges of the mesh from the same views as the RGB images to be passed as input through a pre-trained ControlNet. We then pass these, along with our trained set of text embeddings, to our model with the trained LoRA layers.
  We then extract the segmentation masks $\mathbf{W}$ from the \textit{Mask Extractor}, back-project the labels to UV, and aggregate all the partial UV-maps, $\mathbf{A}$, to a single \textit{globally consistent atlas}, $\mathcal{A}$. This can then be wrapped onto the input 3D mesh to get the segmented output.
  For Noise Optimization, we render the \textit{atlas} $\mathcal{A}$ into the same views as the input RGB images and optimize $\mathcal{E}_{consis}(\cdot)$ (of Eq.~\ref{eq:noise_opt}) to update the per-view input noise and further enhance the multi-view consistency.
  }
  \Description{Inference Pipeline.}
  \label{fig:inference_pipeline}
\end{figure*}

\subsubsection*{Optimization Objectives}
We employ a two-phase optimization strategy~\cite{avrahami2023bas} to train our model. In the first phase, the model $\psi$ remains frozen while only the text embeddings for the part segments, $\mathbf{f_{txt}}$, are optimized with a high learning rate. This quickly produces well-optimized embeddings without compromising model generality, providing a robust starting point for the next phase.
In the second phase, we fine-tune the SD weights using LoRA~\cite{hu2021lora}, optimizing them together with the text embeddings at a significantly lower learning rate. This step enhances the cross-view segmentation consistency of \PaperName{}. 
Our network is trained using the following optimization objective:
\begin{equation}
    \mathcal{L}_{ASIA} = \alpha \mathcal{L}_{\text{CE}} + \beta \mathcal{L}_{\text{MSE}} + \gamma \mathcal{L}_{\text{LDM}} + \delta \mathcal{L}_{\text{corr}}
     \label{eq:l2}
\end{equation}
where $\alpha$, $\beta$, $\gamma$, and $\delta$ are weights for the losses. $\delta$ is 0 in the first phase and assigned a non-zero value in the second phase.
This two-phase optimization results in an optimized \PaperName{} model, $\psi^{*}$, and text embeddings, $\mathbf{f^*_{txt}}$, which together enable the generation of multi-view consistent segmentation masks and effective generalization to diverse 3D shapes during inference.

\subsection{Multi-View Consistent 3D Segmentation}
\label{subsec:inference}
To segment a 3D shape $\mathcal{S}$ during inference, it is first rendered from multiple views by positioning it within a virtual setup of cameras. Each view is then segmented independently using our learned segmentation model $\psi^{*}$, and the results are merged and lifted to produce the final segmentation of the 3D shape $\mathcal{S}$.
However, as each view is processed independently, variations may arise due to the initial sampled noise ($\epsilon$) for Stable Diffusion, leading to inconsistent part segmentations across views.
To address this issue and improve consistency across views during inference, the segmented outputs are projected into a canonical \textit{UV}-space, which is then used to jointly model the noise distribution such that the segmentations are consistent across all the views (Fig.~\ref{fig:inference_pipeline}).

\subsubsection*{Lifting 2D Segmentations to 3D}
At inference, given a 3D shape $\mathcal{S}$, and a set of virtual cameras, we generate a set of $m$ renderings $\mathcal{I}_{r} = \{v_{1}, v_{2}, ..., v_{m}\}$ that provide full coverage of the object. 
We also extract structural priors, such as edges, from the object geometry to provide the network with stronger cues about the input shape \cite{perla2024easitex}.
These priors are integrated into our model through a pre-trained ControlNet~\cite{zhang2023adding} to enhance the segmentation quality. Using the edge-conditioned \PaperName{} model, $\psi^{*}$, we perform part segmentation on each view $v_{i}$ as: $\mathbf{W} = \psi^{*}(\mathcal{I}_{r}, \mathbf{f^*_{txt}}, \varepsilon),$ where $\varepsilon = \{ \epsilon_{1}, \dots, \epsilon_{m} \}$ is the set of Gaussian noise samples used for denoising, generated independently for each view.

To lift these predicted 2D labels $\mathbf{W}$ to 3D we propose using \textit{UV}-map of the input shape which assigns each point $(x, y, z)$ on its surface to a corresponding 2D coordinate $(u,v)$ in the \textit{UV}-space. 
We project per-view segmentations into this common \textit{UV}-space, generating a set of $m$ partial \textit{UV}-atlases $\mathbf{A} = \{a_{1}, a_{2}, ..., a_{m}\}$, one for each view $v_i$, where $a_i = \mathbf{UV}(\mathbf{W}_i)$ and $\mathbf{UV}(.)$ is a projection function which maps information, in our case the predicted labels $\mathbf{W}$ for a view, from the image-space to the \textit{UV}-space. 
Note that each \textit{UV}-atlas, $a_i$, is partial since it only contains the labels for the region of the 3D mesh visible in the rendered view $v_i$. 
To extract a single, complete, globally consistent segmentation mask $\mathcal{A}$, we aggregate the projected per-view \textit{UV}-maps $\mathbf{A}$ as:
\begin{equation}
    \mathcal{A} = \phi(\mathbf{A}) = \phi(\mathbf{UV}(\mathbf{W}))
\end{equation}
where $\phi(.)=\texttt{mode()}$ is the majority voting operator that aggregates and selects a single label for each pixel in $\mathcal{A}$ from the corresponding pixels in the partial \textit{UV}-maps $\mathbf{A}$.

Using \textit{UV}-maps for lifting 2D segmentation labels to 3D allows us to eliminate the dependency on the viewing direction as well as lift fine-grained 2D labels to 3D irrespective of the underlying object geometry, efficiently. 
However, the overall accuracy of the final 3D segmentation remains heavily reliant on the consistency of the individual 2D predictions $\mathbf{W}$ across all views, which continues to present a challenge. 
Directly mapping segmentations, generated with independently sampled noises $\varepsilon$, often leads to fragmentation or sparsity artifacts in the final atlas of labels.
To address this, we propose optimizing $\varepsilon$ to determine a noise sample for each image $v_i$ that produces multiview-consistent 2D predictions.
This ensures a coherent atlas of segmentation labels, $\mathcal{A}$, that resolves sparsity issues within the \textit{UV}-space and accurately maps object parts onto the 3D shape $\mathcal{S}$ when projected onto its surface.

\begin{table*}[h]
\caption{\emph{Quantitative Results.} Comparison of methods 
for semantic segmentation on the PartNetE~\cite{liu2023partslip} dataset. We report the average per-part mIoU ($\uparrow$) in percentage for each category and the average mIoU across 40 categories of the dataset in the last column. Please refer to the supplementary for more results.}
\small
\begin{tabular}{c|ccccccccccc}
\toprule
Method & Chair   & Door   & Lamp   & Lighter   & Phone   & Printer   & Remote   & Scissors   & Toaster   & \multicolumn{1}{c|}{Wash. Mach.}                           & Overall (40)         \\ \midrule
PartSLIP~\cite{liu2023partslip}                        & 85.3          & 40.8          & 66.1          & 64.7          & 48.4          & 4.3         & 38.3          & 60.3          & 60.0          & \multicolumn{1}{c|}{53.4}                                  & 60.5          \\

PartSTAD~\cite{kim2024partstad}                        & 85.3          & 61.4          &  68.4          & 65.9          & 63.2          & 7.9       & 53.4          & 68.5          & 58.6          & \multicolumn{1}{c|}{ 48.2}                                  & 66.1         \\

PartDistill~\cite{umam2024partdistill}                        & 88.4          & 55.5          &  69.2          & 64.9          & 50.8          & 6.3          & 40.7          & 68.8          & 58.7          & \multicolumn{1}{c|}{55.1}                                  & 67.1         \\

3$\times$2~\cite{3by2}                        & 84.4          & 54.4          & 59.5          & 65.0          &  41.0   &  8.5         & 54.1        & 65.7         & 56.5                & \multicolumn{1}{c|}{52.6}                                  & 65.9          \\

\arrayrulecolor{gray}\cmidrule(lr){1-12}
\arrayrulecolor{black}

Ours                     & \textbf{90.1} & \textbf{77.5} & \textbf{76.9} & \textbf{73.3} & \textbf{69.3} & \textbf{38.4} & \textbf{74.7} & \textbf{83.7} & \textbf{73.0} & \multicolumn{1}{c|}{\textbf{76.0}} & \textbf{75.8} \\ 
\bottomrule
\end{tabular}
\label{Tab:quant-comp}
\end{table*}

\begin{table}[]
\centering
\caption{\emph{User Study.} Comparison of methods for \textit{adaptive} segmentation on meshes from the Objaverse~\cite{deitke2023objaverse} dataset. We report the \textit{Average Rank} and \textit{Win Rate} (in \%) averaged over the entire study.}
\small
\label{tab:custom_quant}
\begin{tabular}{ccc}
\toprule
Method & Avg. Rank ($\downarrow$) & Win Rate ($\uparrow$) \\ \midrule
3D Highlighter~\cite{decatur20233d}      & 2.92    & 1.25  \\
SATR~\cite{abdelreheem2023satr}          & 2.01    & 5.21  \\
\arrayrulecolor{gray}\cmidrule(lr){1-3}
\arrayrulecolor{black}
Ours          & \textbf{1.08}  & \textbf{93.54} \\ \bottomrule
\end{tabular}%
\end{table}



\subsubsection*{Noise Guided Multi-View Consistency}
Although projecting labels to \textit{UV}-space and applying voting improves 3D segmentation accuracy, the individual per-view segmentation masks $\mathbf{W}$ may still exhibit inconsistencies in corresponding parts across different views.
One of the reasons for this is that the noise samples, $\varepsilon$, are drawn independently, and each view is processed separately and in isolation.
To tackle this, we propose updating the sampled noises, $\epsilon_{i} \in \varepsilon$, in the latent space, which are responsible for the inconsistency, in a way that aligns the part predictions in the image space.
For this, we first project the globally aggregated \textit{UV}-map of labels, $\mathcal{A}$, to the image space in the same poses as $\mathcal{I}_{r}$, as: $\widehat{\mathbf{W}} = \textbf{UV}^{-1}(\mathcal{A}),$
where $\textbf{UV}^{-1}$ maps the aggregated \textit{UV}-map of labels $\mathcal{A}$ to individual views as $v_i$ through rendering. The segments $\widehat{\mathbf{W}}$ are now multiview consistent since they all are generated by a single \textit{UV}-map of labels, $\mathcal{A}$. 
To improve consistency in our model, we leverage test-time optimization~\cite{chen2024training}, where the latent noises $\varepsilon$ of the Stable Diffusion framework are optimized using the multiview-consistent segments $\widehat{\mathbf{W}}$ as pseudo-ground truth, using an energy $\mathcal{E}_{consis}$, as: 
\begin{equation}
\label{eq:noise_opt}
\begin{gathered}
    \mathcal{E}_{consis}  = \frac{1}{m}\sum^{m}_{i=1}|| \mathbf{W}_{i} - \widehat{\mathbf{W}}_{i}||_{2} + \lambda \mathcal{L}_{reg}, \\
   \,\text{where, } \mathcal{L}_{reg} = \frac{1}{2|\varepsilon|}\left( 1 + \log(\sigma^2_{\varepsilon}) - \mu^2_{\varepsilon} - \sigma^2_{\varepsilon} \right)
\end{gathered}
\end{equation}
$\mathcal{L}_{reg}$ is a regularization term which ensures $\varepsilon$ remains close to the normal distribution $\mathcal{N}(0, \mathcal{I})$ during optimization and $\lambda$ controls the strength of this regularization.
Optimizing this function with respect to $\varepsilon$ encourages multi-view consistency in the WAS attention maps $\mathbf{W}$. Specifically, we update the latent noise $\varepsilon$ using gradient descent on $\mathcal{E}_{consis}$ as: $\varepsilon^{*} \leftarrow \varepsilon - \eta.\nabla\mathcal{E}_{consis}$, where $\eta$ is the learning rate and $\nabla$ is the gradient operator. After a few optimization steps, the optimized set of noises $\varepsilon^{*} =\{\epsilon^{*}_{1}, \epsilon^{*}_{2},..., \epsilon^{*}_{m}\}$ achieve a state which has: i) an optimized Gaussian noise $\epsilon^{*}_{i}$ for each camera view $v_i$, tailored to offer consistency;
ii) utilized 3D information from the mesh to optimize the noise at different views implicitly through \textit{UV}-mapping. After this process, we obtain a new set of partial atlases $\mathbf{A}^*$ obtained by projecting multiview consistent part segments $\mathcal{W}$ corresponding to the optimized noise $\varepsilon^{*}$ into the \textit{UV}-space. This \textit{UV}-maps of part labels, $\mathbf{A}^*$, are then aggregated and projected onto the object $\mathcal{S}$ to obtain the final part segmentations in 3D.

\section{Experiments and Results}
\label{sec:experiments}

In this section, we present both qualitative and quantitative results demonstrating our method's ability to segment adaptive as well as semantic parts covered by open vocabulary models.

\subsection{Dataset and Metrics} 
\label{subsec:dataset_n_metrics}
\subsubsection*{Dataset Details}
Existing part segmentation datasets such as PartNet-Ensembled (PartNetE) \cite{liu2023partslip} primarily focus on simple object categories with part annotations that are easily described using text prompts. This limits their utility for adaptive segmentation tasks involving more complex or unconventional segmentation. To enable a rigorous evaluation of such task, we select 10 challenging object categories—including chairs, human faces, gladiator helmets, microphones, scooters, shoes, pistons, \textit{etc.}—as listed in Tab.~\ref{tab:custom_full_comp_1}. To train our model, we collect a small set of reference images from the internet, gathering around 25 to 30 images per category. These images are manually annotated according to the desired adaptive segmentation. Some annotated reference images can be found in Tab.~\ref{tab:custom_full_comp_1}. We source the corresponding 3D meshes from the Objaverse~\cite{deitke2023objaverse, deitke2023objaverse_xl} dataset.
To further validate our approach, we also evaluate on the standard open-vocabulary segmentation task using the PartNetE~\cite{liu2023partslip} dataset, which includes 1,906 objects across 45 categories. This dataset combines objects from both PartNet~\cite{mo2019partnet} and PartNet Mobility~\cite{Xiang_2020_SAPIEN} datasets. We follow the same train-val-test splits used by prior works~\cite{liu2023partslip, kim2024partstad, 3by2, umam2024partdistill} for training, validation, and inference. We train a separate model for each category using shapes with ground-truth segmentations from the train split. We render 10 views per object to train our model, similar to prior works.

\subsubsection*{Metric Details}
For quantitative evaluation of open-vocabulary segmentation, we use \emph{mean Intersection over Union} (\textbf{mIoU}), which measures the average overlap between predicted and ground-truth parts across all classes.
For adaptive segmentation, we conduct a user study and report the \emph{average rank}, \textit{i.e.}, the mean ranking of each method across all comparisons (lower is better), and the \emph{win percentage}, \textit{i.e.}, the fraction of pairwise comparisons where a method is preferred over others.

\subsection{Baselines and Results Discussion}
\label{subsec:baselines}
\subsubsection*{Baselines}
In the absence of direct baselines---open-source methods which use images for guidance (see Tab.~\ref{tab:rel_works_summary})---for adaptive 3D segmentation, we use SATR~\cite{abdelreheem2023satr} and 3D Highlighter~\cite{decatur20233d} for text-based 3D localization.
We provide manually curated text prompts that best describe the target adaptive parts for localization.
Since 3D Highlighter can handle only one part per query, we apply it iteratively to segment the entire mesh.
We also evaluate \PaperName{} on open-vocabulary, text-describable part segmentation and compare it with recent baselines like PartSLIP~\cite{liu2023partslip}, PartSTAD~\cite{kim2024partstad}, PartDistill~\cite{umam2024partdistill}, and 3$\times$2~\cite{3by2}.

\subsubsection*{Quantitative and Qualitative Analysis}
We present quantitative and qualitative results for text-describable, semantic part segmentation on PartNetE dataset in Tab.~\ref{Tab:quant-comp} and Tab.~\ref{tab:PN_full_comp_1}, respectively. As shown in both tables, our method consistently outperforms existing baselines on a wide range of categories. These results highlight the effectiveness of our approach in segmenting complex 3D shapes in an open-vocabulary setting.
We further evaluate our method on adaptive segmentation through a user study on $15$ meshes spanning a diverse set of novel object categories from the Objaverse dataset (Sec.~\ref{subsec:dataset_n_metrics}). The study involved $32$ graduate students with a background in computer science, who were shown both the reference annotations and the outputs from all methods. Each mesh was presented from four views (front, back, left, and right), and the participants were asked to rank the methods from best to worst based on how well the generated segmentations align with the reference. As shown in Tab.~\ref{tab:custom_quant}, our method achieves the best \textit{average rank}, 1.08, with a \textit{win rate} of $93.54\%$, demonstrating a strong and consistent performance for adaptive segmentation.

Our approach leverages strong object-centric diffusion priors together with the correspondence loss $\mathcal{L}_{corr}$, enabling fast and robust finetuning of the diffusion model for accurate part prediction. Visual comparisons for adaptive segmentation in Tab.\ref{tab:custom_full_comp_1} further illustrate the advantages of our method over the baselines. The baselines we evaluate are zero-shot methods that only rely on text descriptions of parts for guidance (Sec.\ref{subsec:baselines}) and are therefore not specifically designed for adaptive segmentation. In addition, many object parts in our evaluation are difficult to describe precisely using text, which further limits these baselines—yet this very challenge underscores the motivation for this work, ASIA, an approach for few-shot \textit{adaptive} part segmentation.

\begin{table}[t]
  \centering
  \caption{Ablation Study of design choices for training (left) and inference (right). See Sec.~\ref{subsec:ablation} for a detailed analysis.}
  \label{tab:ablation_combined}
  \begin{minipage}{0.50\linewidth}
    \centering
    \footnotesize
    \begin{tabular}{l|c}
        \toprule
        \multicolumn{1}{c|}{Method}                                                       & mIoU ($\uparrow$) \\ \midrule
        BL (SLiMe)                                                                             & 55.94           \\
        BL + LoRA                                                                       & 60.65           \\ \arrayrulecolor{gray}\cmidrule(lr){1-2}
        \arrayrulecolor{black}
        
        \multicolumn{1}{c|}{BL + LoRA + $\mathcal{L}_{corr}$ (Ours)} & \textbf{63.69}  \\ 
        \bottomrule
    \end{tabular}
  \end{minipage}
  \hspace{0.05\linewidth}
  \begin{minipage}{0.40\linewidth}
    \centering
    \footnotesize
    \begin{tabular}{cc|c}
        \toprule
        \begin{tabular}[c]{@{}c@{}}Edge\\ Cond.\end{tabular} & \begin{tabular}[c]{@{}c@{}}Noise\\ Optim.\end{tabular} & mIoU ($\uparrow$) \\ 
        \midrule
        \xmark           & \xmark            & 59.33           \\ 
        \cmark            & \xmark              & 62.42           \\
        \arrayrulecolor{gray}\cmidrule(lr){1-3}
        \arrayrulecolor{black}
        \cmark    & \cmark     & \textbf{63.69}  \\ 
        \bottomrule
    \end{tabular}
  \end{minipage}
\end{table}

\begin{figure}
    \centering
    \includegraphics[width=0.95\linewidth]{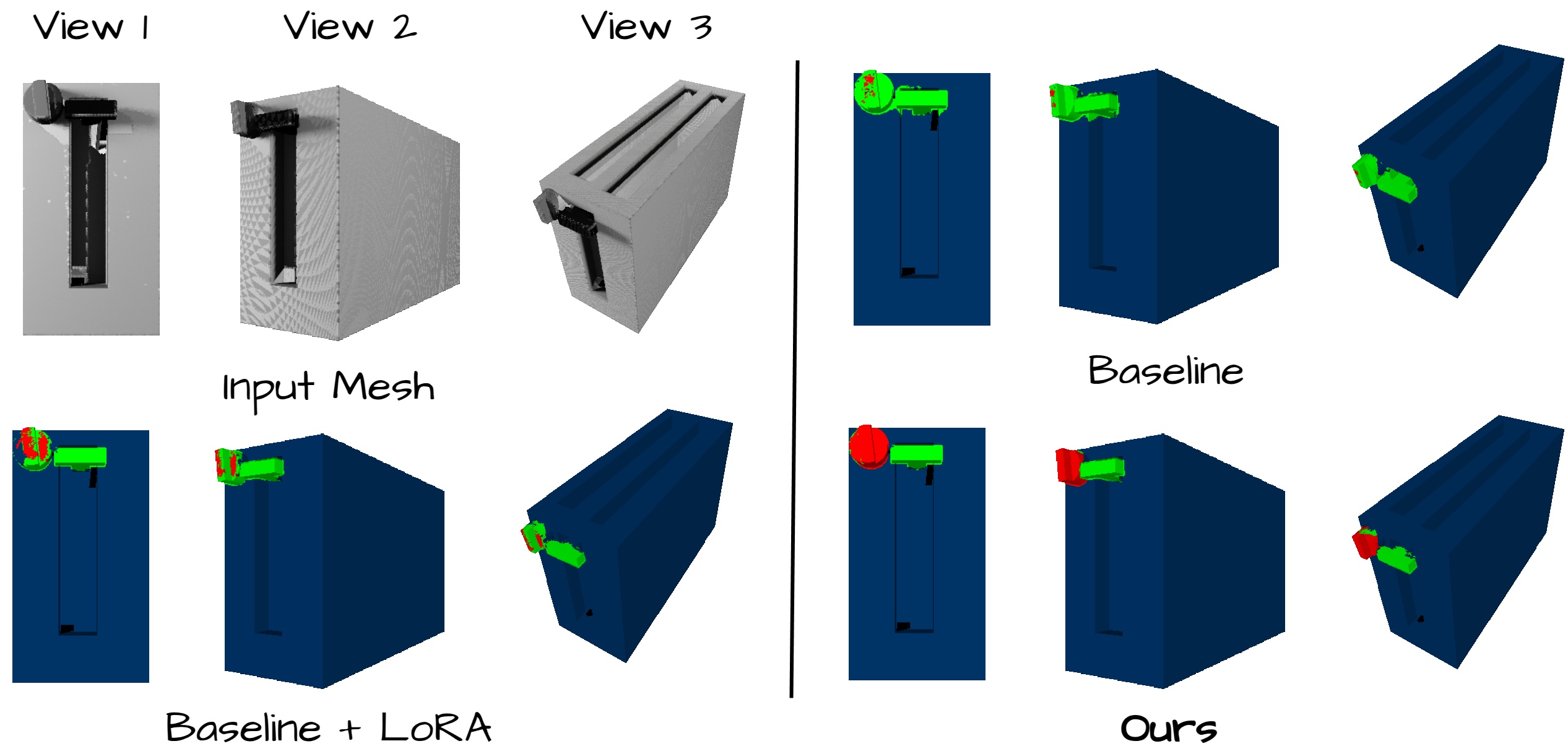}
    \caption{Visualization of results from different design choices while training. The figure shows a \textit{Toaster} from the PartNetE~\cite{liu2023partslip} dataset with parts: Button (red), Slider (green), and Background (blue).}
    \label{fig:model_ablation}
\end{figure}

\subsection{Ablation Studies}
\label{subsec:ablation}

To assess the contribution of each component in our framework, we conduct a series of ablation studies. Throughout these experiments, we report mIoU on four PartNetE categories: Printer, Lighter, Toaster, and Washing Machine.
The results in Tab.~\ref{tab:ablation_combined} (left) and Fig.~\ref{fig:model_ablation} demonstrate the importance of different design components in our training pipeline. Starting from the baseline method (vanilla SLiMe), incorporating LoRA fine-tuning yields a substantial improvement of approximately 5\% in mIoU. This gain arises because LoRA preserves the semantic priors of the pre-trained Stable Diffusion model while adapting it to the provided annotations, making the model less prone to overfitting and more generalizable. The inclusion of the correspondence loss $\mathcal{L}_{corr}$ further improves results, yielding the best performance and demonstrating its effectiveness in enhancing multi-view consistency.
We also evaluate the effect of different design components in our inference pipeline, as reported in Tab.~\ref{tab:ablation_combined} (right). Edge conditioning through ControlNet provides additional structural information about the shape and encourages the segmentations to align with geometric details. In addition, our per-view noise optimization enhances cross-view consistency by refining segmentations across viewpoints and thereby improving overall results.



\section{Conclusion}
\label{conclusion}

We proposed \PaperName{}, a novel approach for \textit{adaptive} segmentation, consisting of possibly non-semantic and non-text describable “parts", of 3D shapes using a few annotated in-the-wild images as reference.
We leverage the rich priors of a text-to-image diffusion model to quickly learn parts and transfer them to a 3D shape through its 2D projections.
\PaperName{} offers a practical, generalizable solution for both semantic and non-semantic 3D segmentation, outperforming existing methods by a significant margin in both quantitative and qualitative evaluations.

\begin{figure}[t]
    \includegraphics[width=0.9\linewidth]{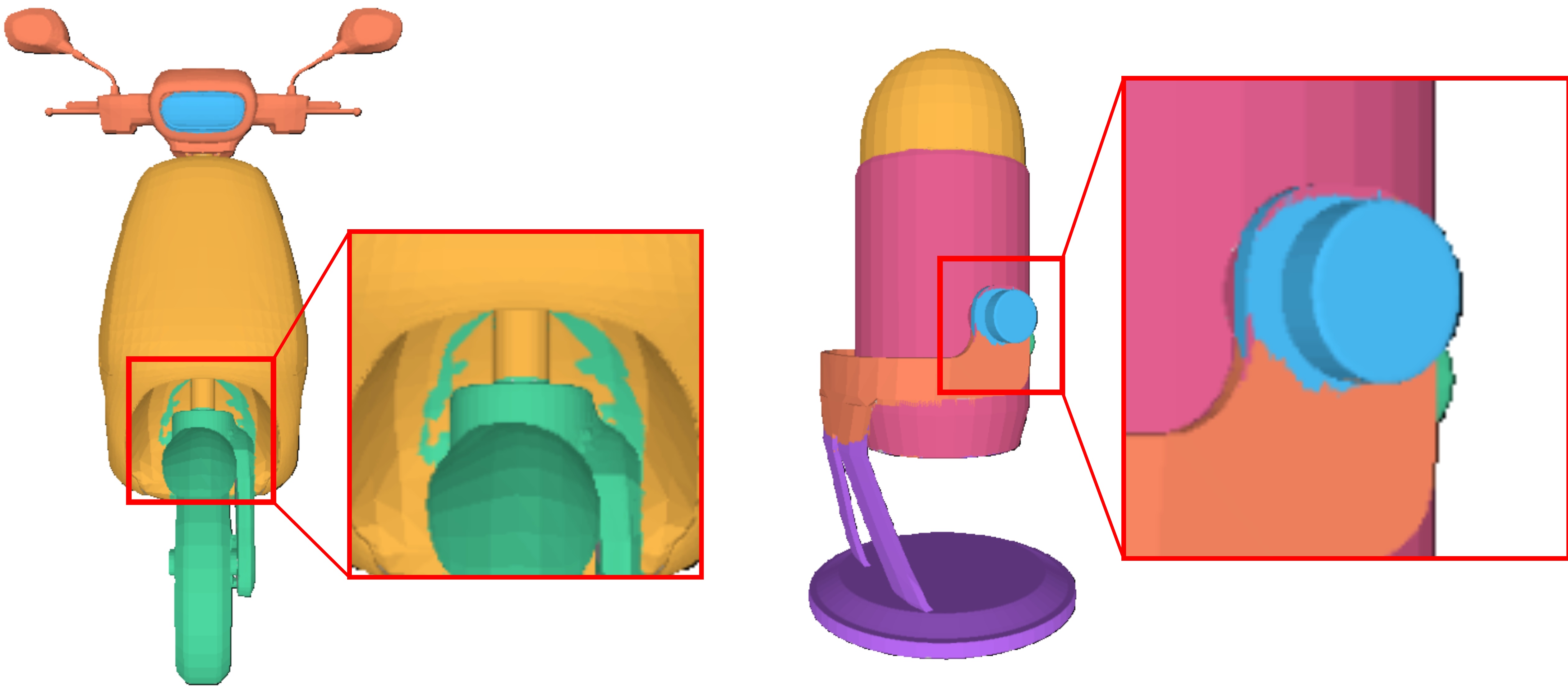}
  \caption{\emph{Failure Cases.} Segmentation spillover can occasionally occur for small/thin parts or at low-confidence regions.}
  \label{fig:fail_case}
  \Description{Failure Cases}
\end{figure}

\subsubsection*{Limitations}
Despite its strengths, our method has certain limitations—particularly in scenarios involving small or thin parts, since we extract the masks from low-resolution attention maps, or low-confidence regions, where the multi-view 2D predictions may fail to accurately localize the region of interest.
This issue becomes more pronounced when such regions overlap with other parts across different views.
Consequently, as illustrated in Fig.~\ref{fig:fail_case}, these overlaps can lead to segmentation spillover, where the predicted mask for one part incorrectly extends into adjacent or overlapping parts, causing occasional inaccuracies in the final output.

\subsubsection*{Future Works}
For future works, we aim to extend adaptive segmentation to 3D and 4D scenes, handling multiple objects at once and adding motion to the problem will be an interesting direction to explore. Lastly, since we use multi-view images for segmentation, our approach suffers from self-occlusions. Consequently, we also hope to explore adaptive camera sampling techniques or pivot to processing the 3D shape directly, without any intermediate 2D projections, to improve the coverage.




\section{Acknowledgements}
\label{sec:acks}

We thank the anonymous reviewers for their insightful comments and constructive feedback.
This work was supported in part by NSERC Discovery Grants \textit{RGPIN-2022-03111} and \textit{611370}, and a NSERC Discovery Launch Supplement \textit{DGECR-2022-00359}.

\clearpage
\newpage
\begin{table*}[ht]


\centering
\caption{
Qualitative comparison of results from 3D Highlighter~\cite{decatur20233d}, SATR~\cite{abdelreheem2023satr}, and Ours for adaptive 3D segmentation. Each row shows the reference annotation followed by two views of the same object.
}
\label{tab:custom_full_comp_1}

\begin{tabular}{@{}p{0.1171\textwidth}@{} | @{}p{0.44\textwidth}@{} | @{}p{0.44\textwidth}@{}}

\multicolumn{1}{c}{\textbf{}} &
\multicolumn{1}{c}{\textbf{View 1}} & \multicolumn{1}{c}{\textbf{View 2}} \\

\vtop{\vskip0pt
\resizebox{0.1171\textwidth}{!}{
\begin{tabular}{@{}c@{}}
\toprule
\small{Reference} \\ \midrule
\includegraphics[width=0.1\textwidth]{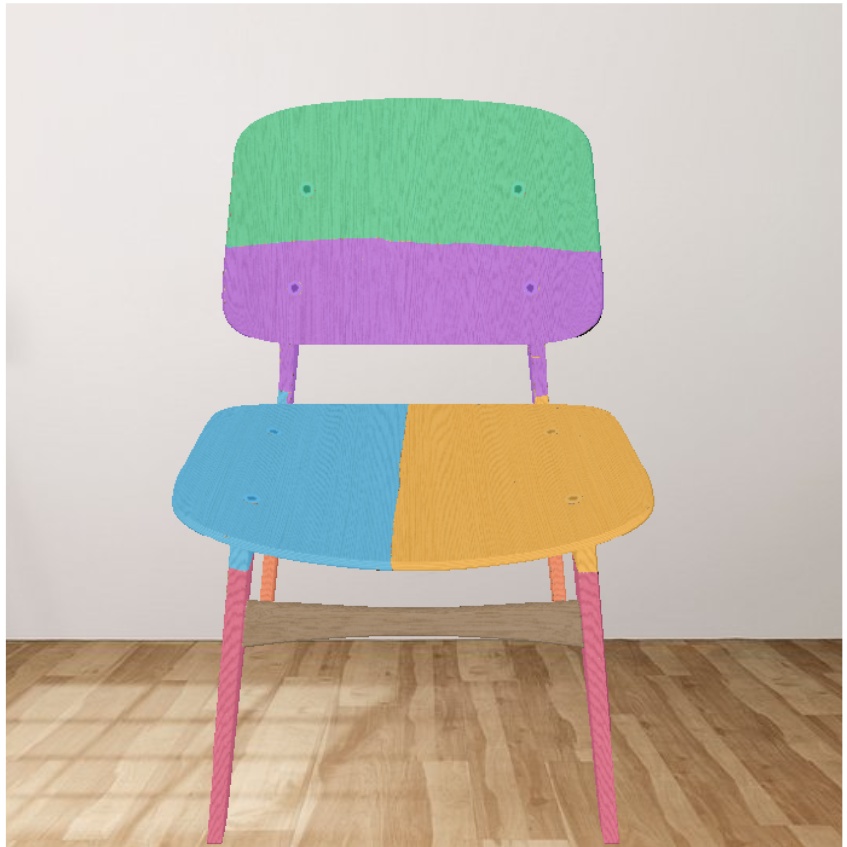} \\

\addlinespace[-2pt]
\arrayrulecolor{gray}\cmidrule(lr){1-1}
\arrayrulecolor{black}

\includegraphics[width=0.1\textwidth]{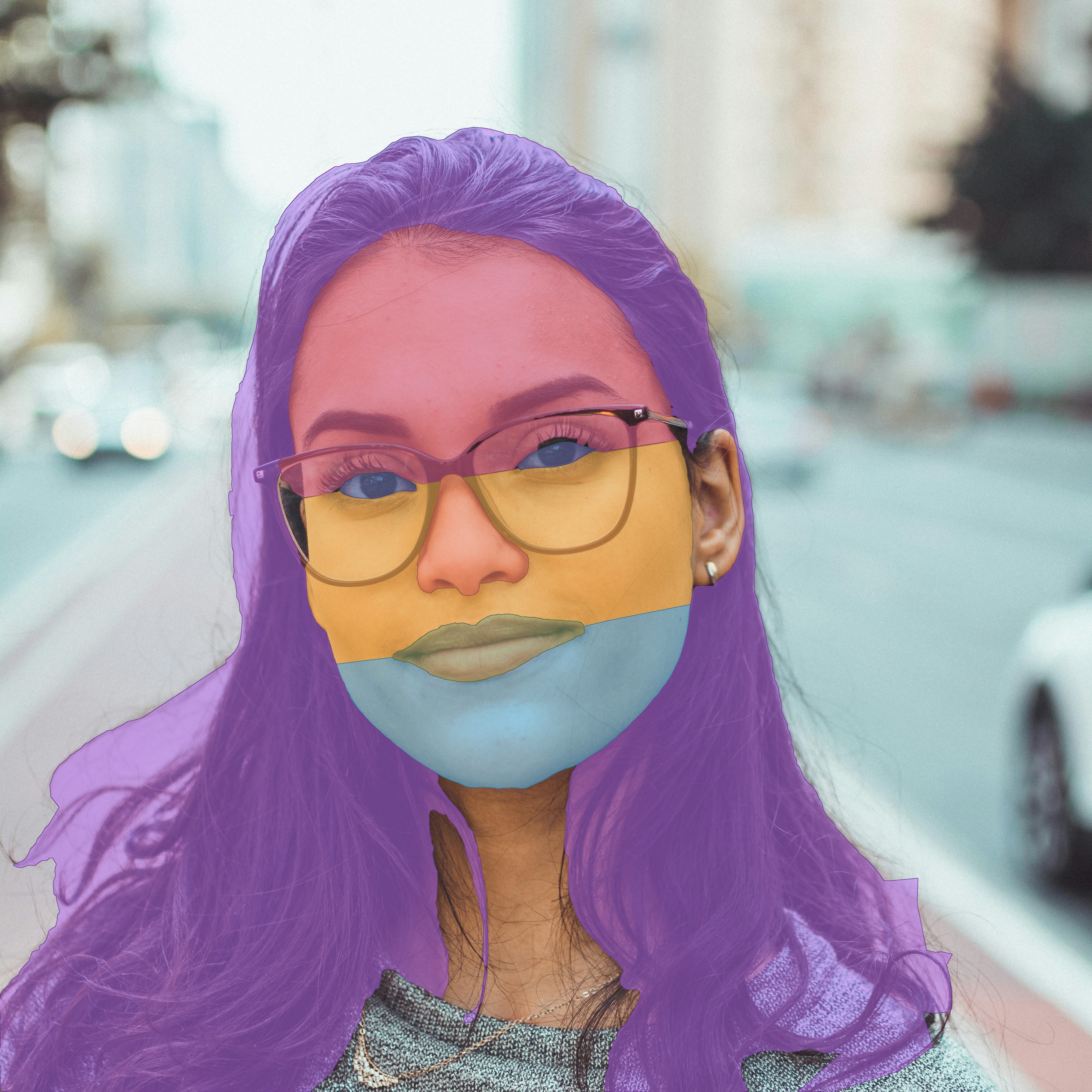} \\

\addlinespace[-2pt]
\arrayrulecolor{gray}\cmidrule(lr){1-1}
\arrayrulecolor{black}

\includegraphics[width=0.1\textwidth]{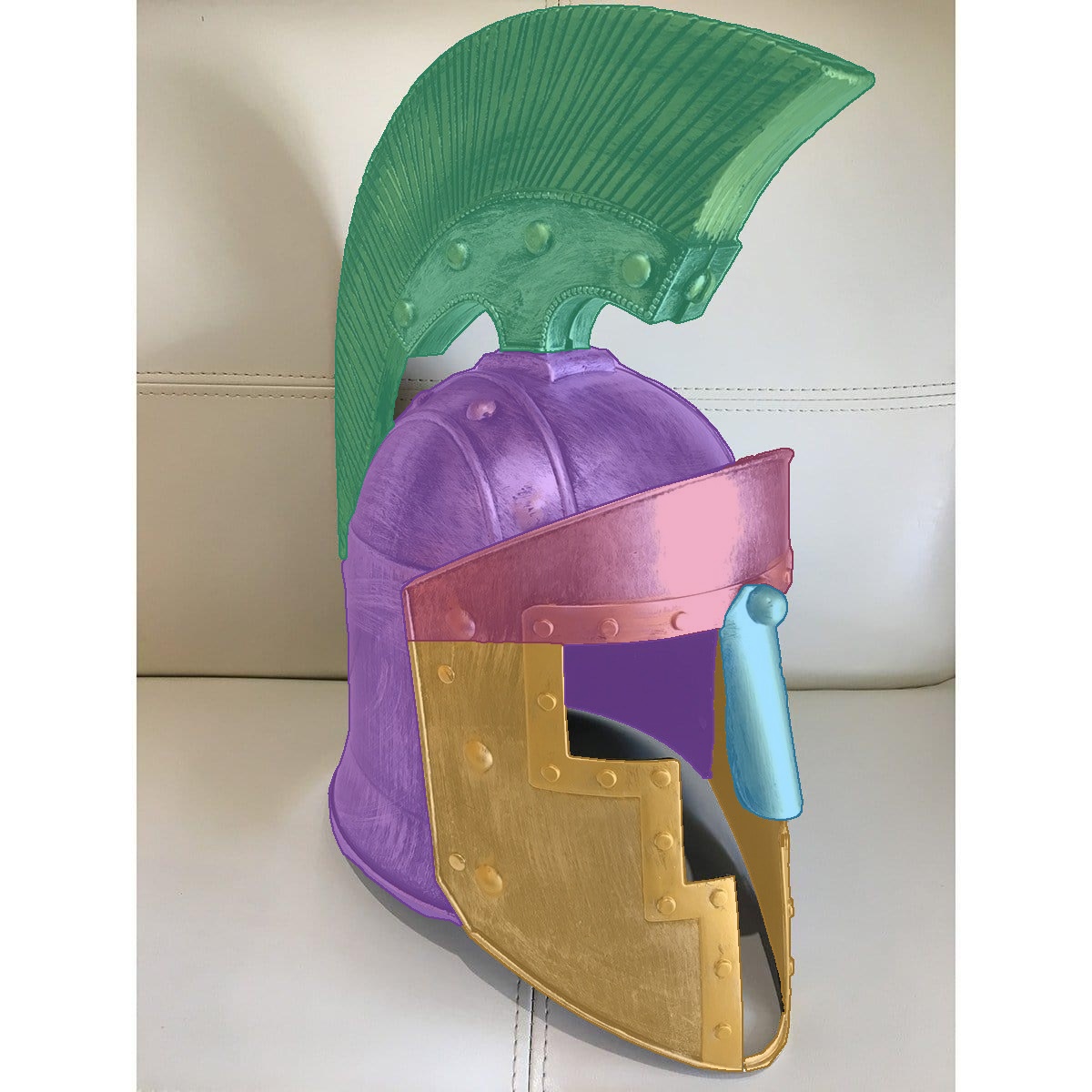} \\

\addlinespace[-2pt]
\arrayrulecolor{gray}\cmidrule(lr){1-1}
\arrayrulecolor{black}

\includegraphics[width=0.1\textwidth]{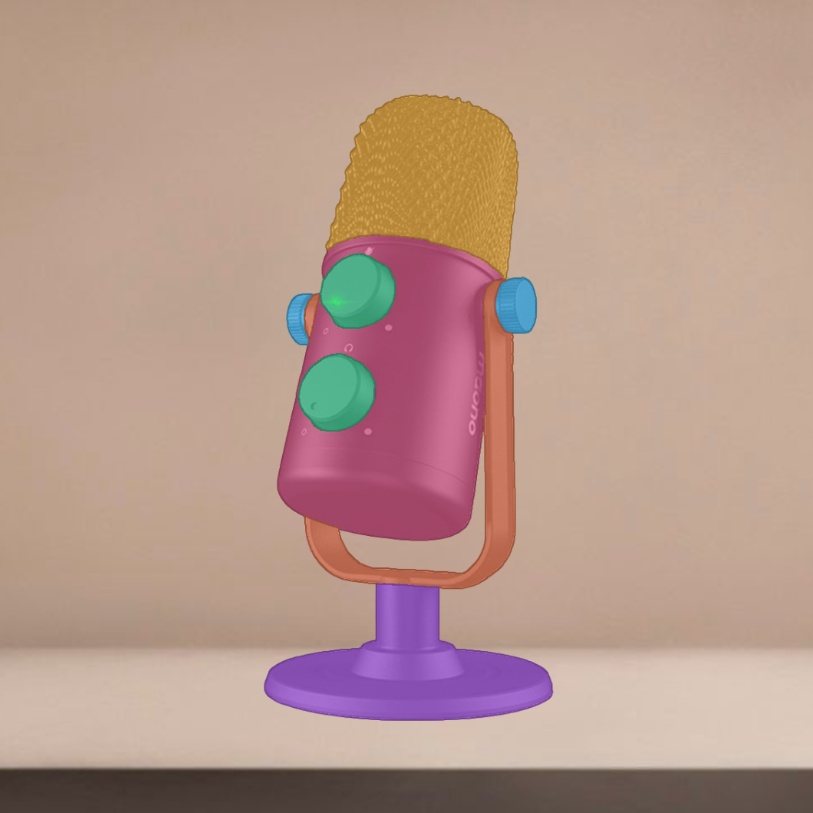} \\

\addlinespace[-2pt]
\arrayrulecolor{gray}\cmidrule(lr){1-1}
\arrayrulecolor{black}

\includegraphics[width=0.1\textwidth]{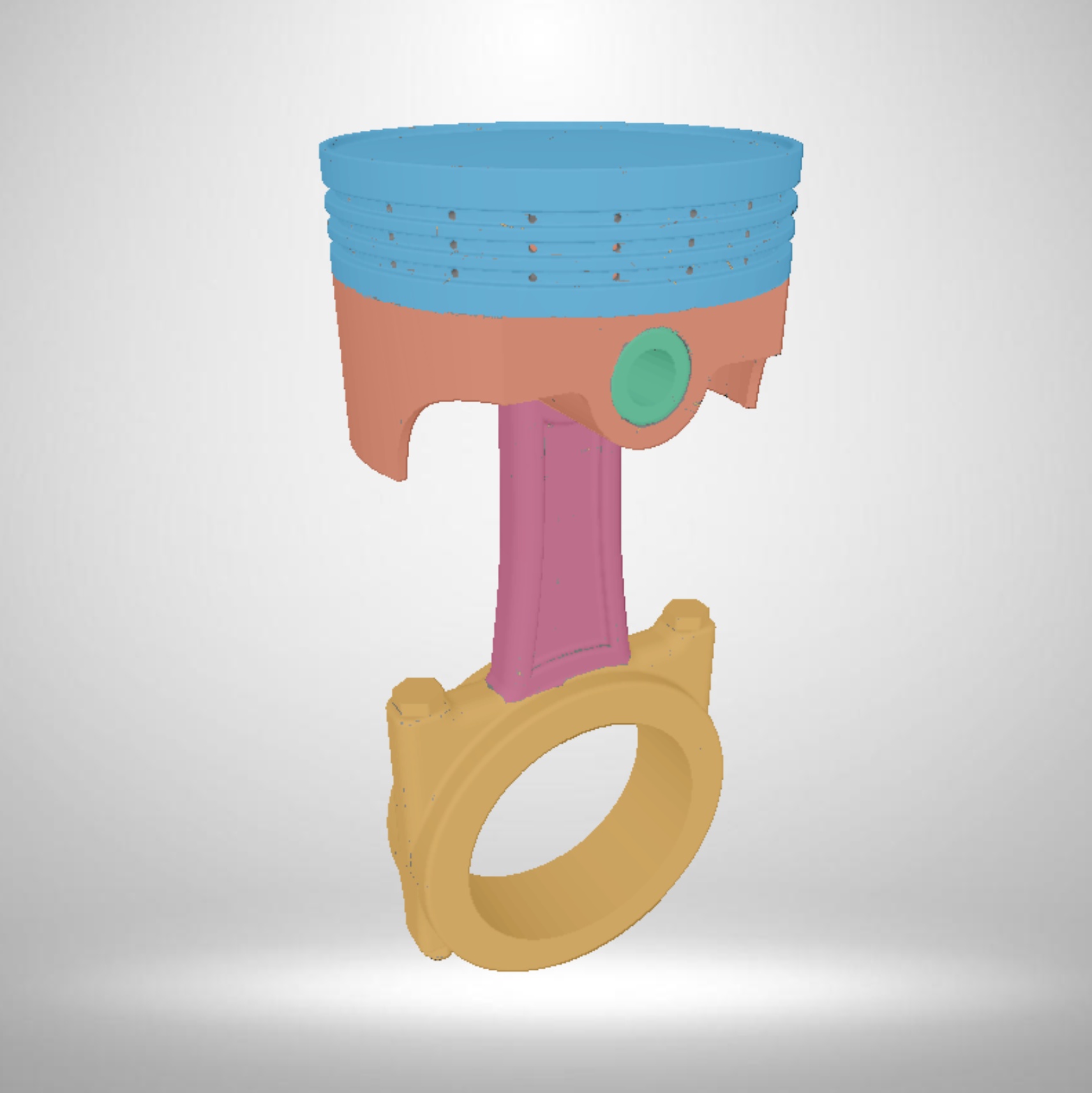} \\

\addlinespace[-2pt]
\arrayrulecolor{gray}\cmidrule(lr){1-1}
\arrayrulecolor{black}

\includegraphics[width=0.1\textwidth]{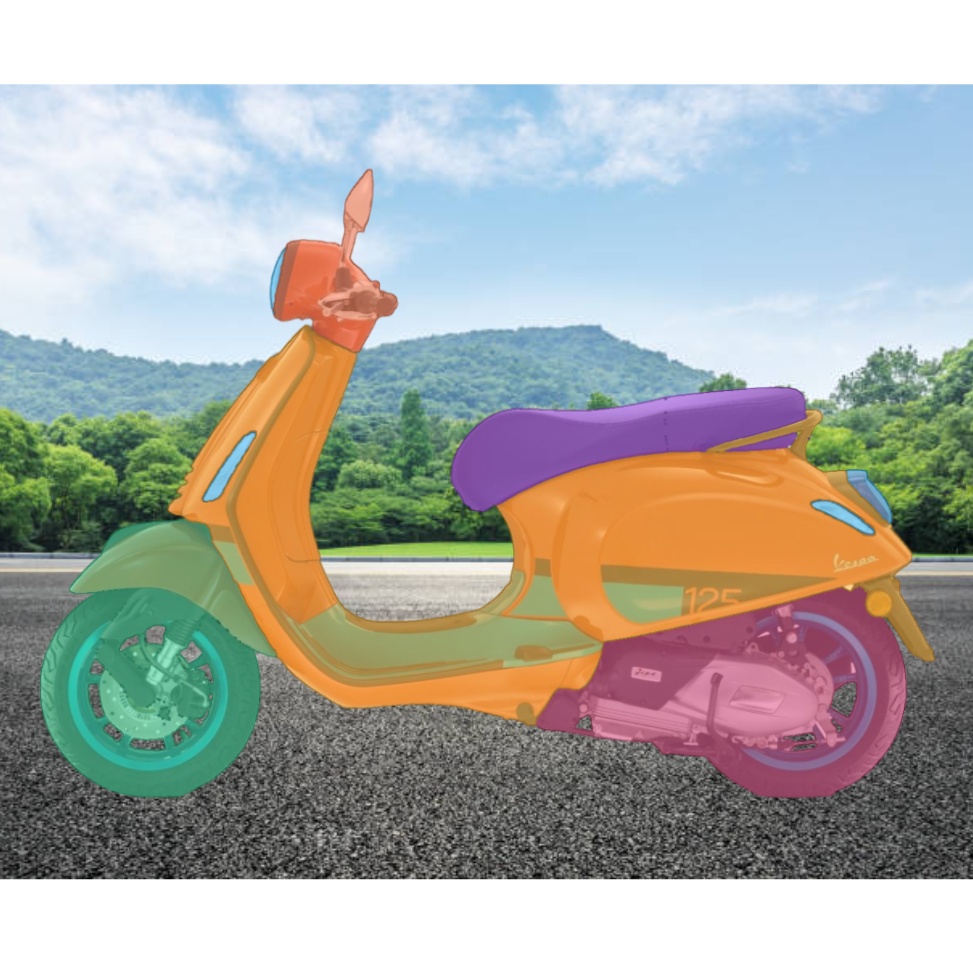} \\

\addlinespace[-2pt]
\arrayrulecolor{gray}\cmidrule(lr){1-1}
\arrayrulecolor{black}

\includegraphics[width=0.1\textwidth]{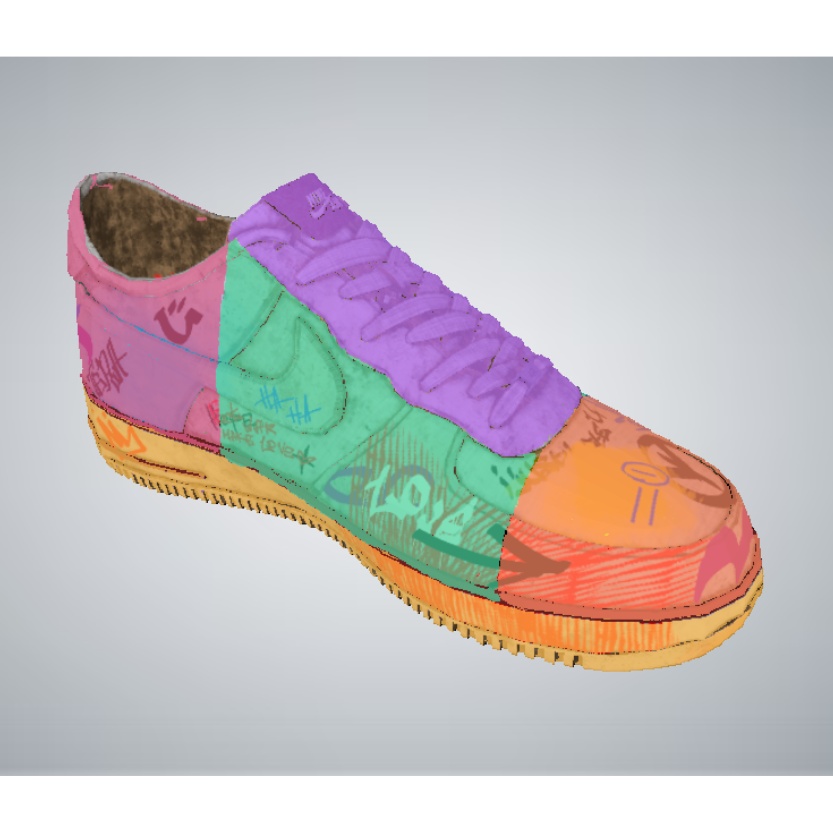} \\

\addlinespace[-2pt]
\arrayrulecolor{gray}\cmidrule(lr){1-1}
\arrayrulecolor{black}

\includegraphics[width=0.1\textwidth]{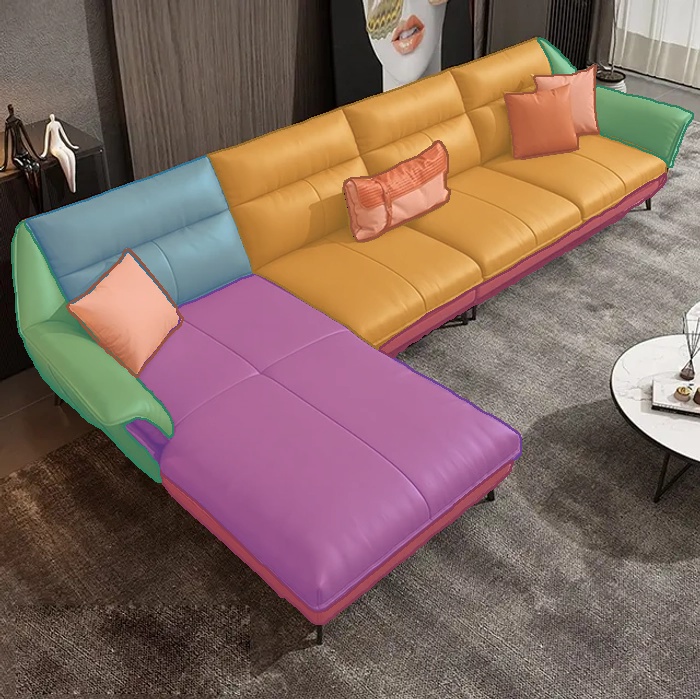} \\

\addlinespace[-2pt]
\arrayrulecolor{gray}\cmidrule(lr){1-1}
\arrayrulecolor{black}

\includegraphics[width=0.1\textwidth]{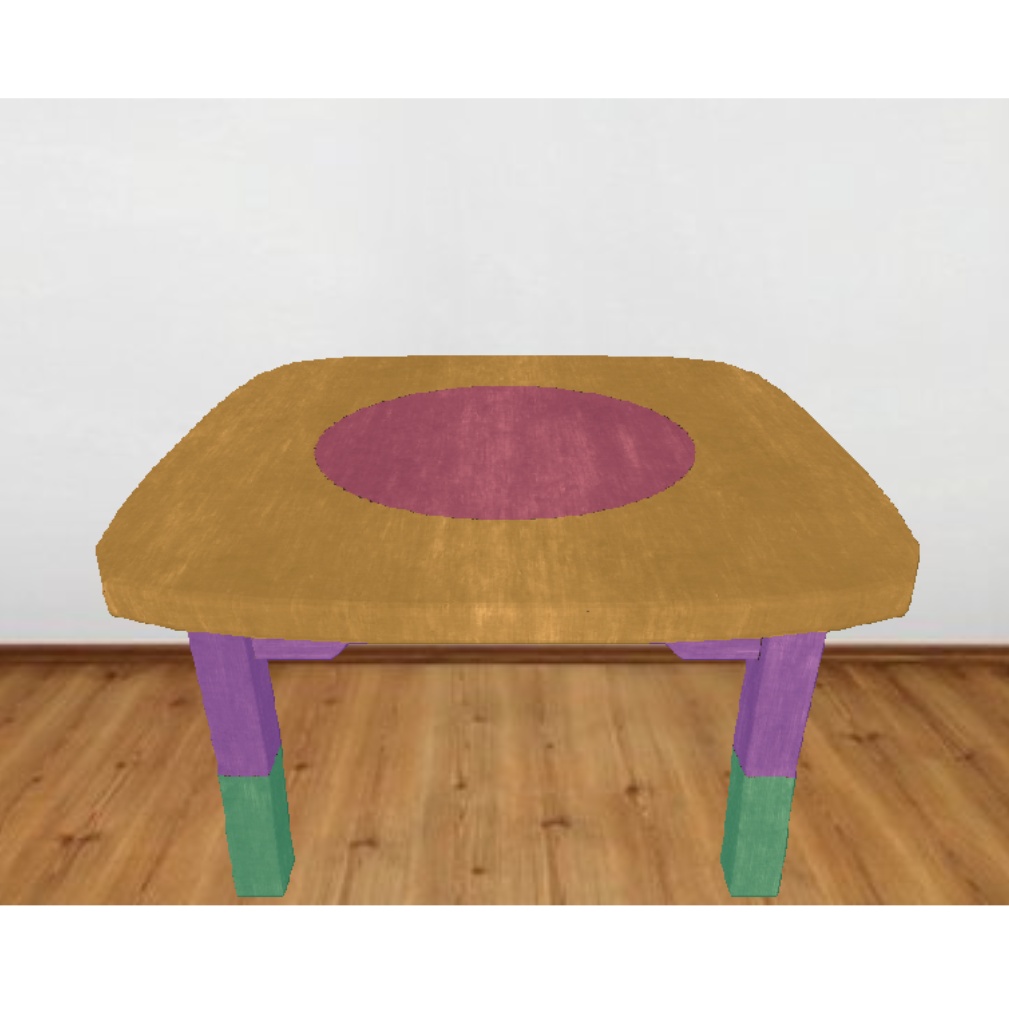} \\

\addlinespace[-2pt]
\bottomrule
\end{tabular}
}}
&
\vtop{\vskip0pt
\resizebox{0.44\textwidth}{!}{
\begin{tabular}{@{}c@{}c@{}c@{}c@{}c@{}}
\toprule
\small{Input} & \small{3D Highlighter} & \small{SATR} & \small{Ours} \\ \midrule
\includegraphics[width=0.1\textwidth]{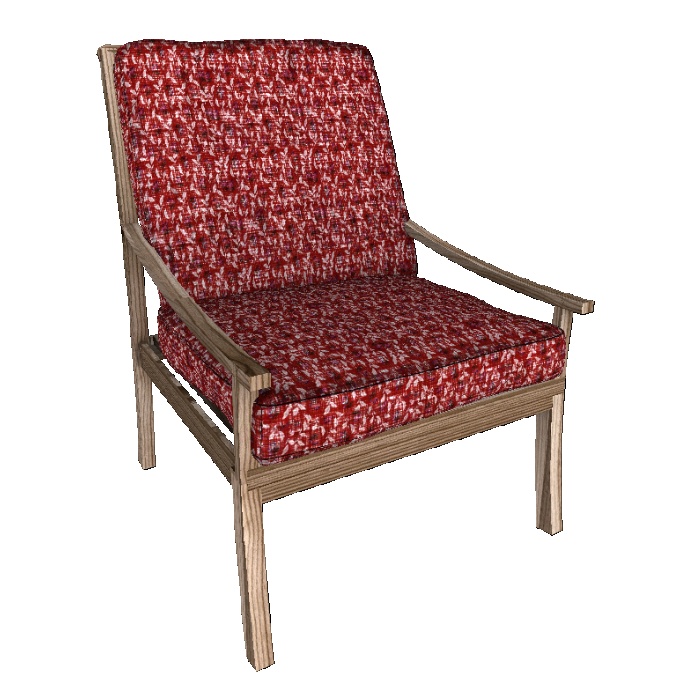} &
\includegraphics[width=0.1\textwidth]{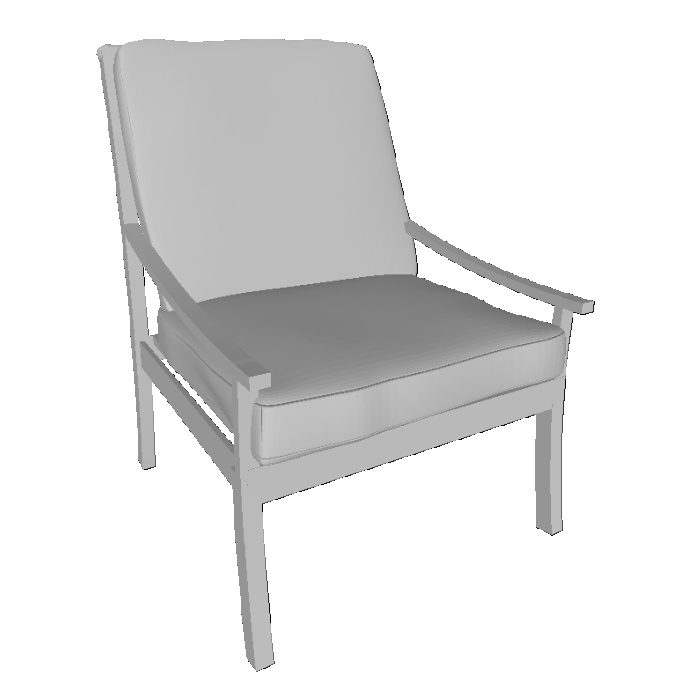} &
\includegraphics[width=0.1\textwidth]{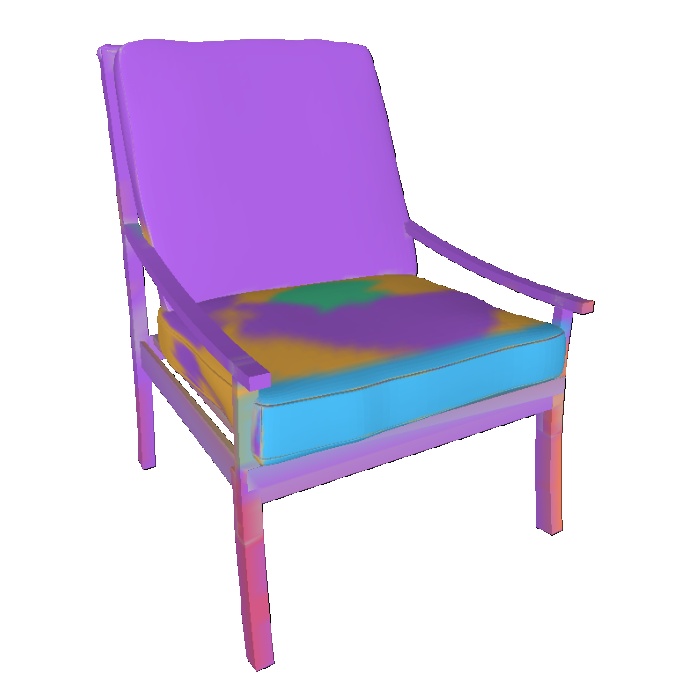} &
\includegraphics[width=0.1\textwidth]{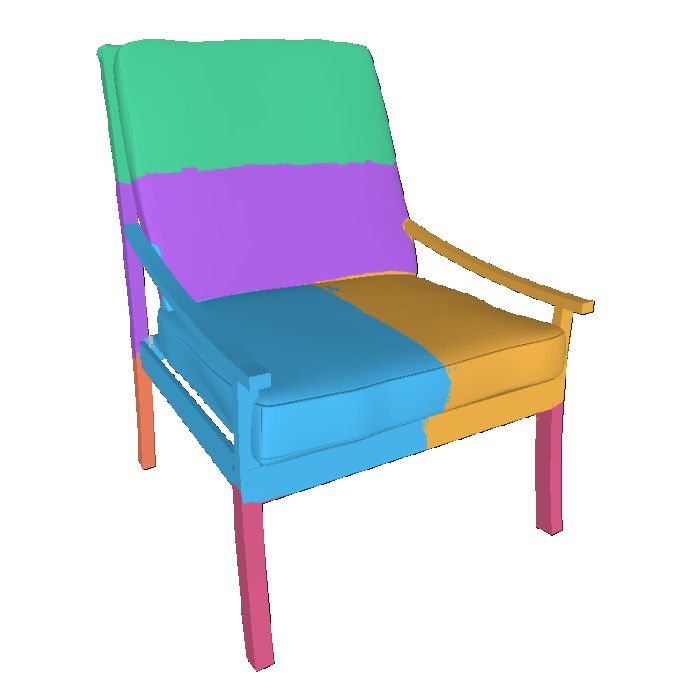} \\

\addlinespace[-2pt]
\arrayrulecolor{gray}\cmidrule(lr){1-4}
\arrayrulecolor{black}

\includegraphics[width=0.1\textwidth]{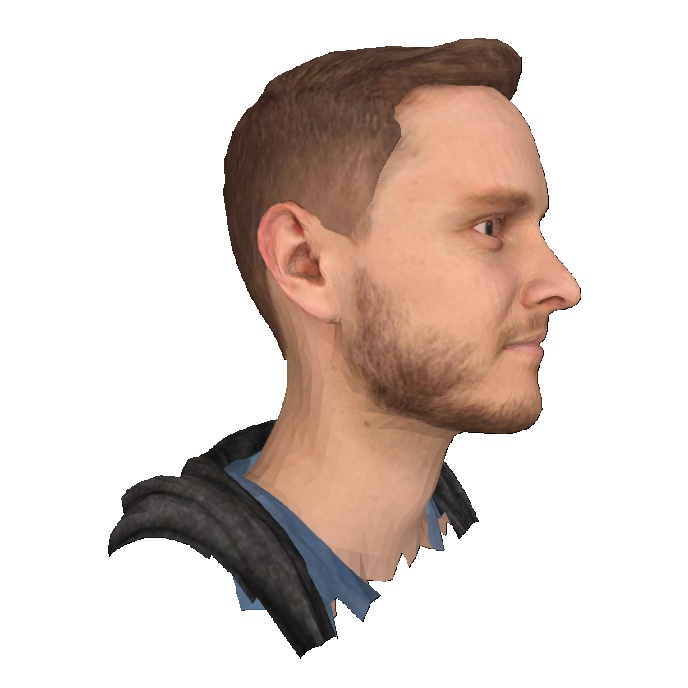} &
\includegraphics[width=0.1\textwidth]{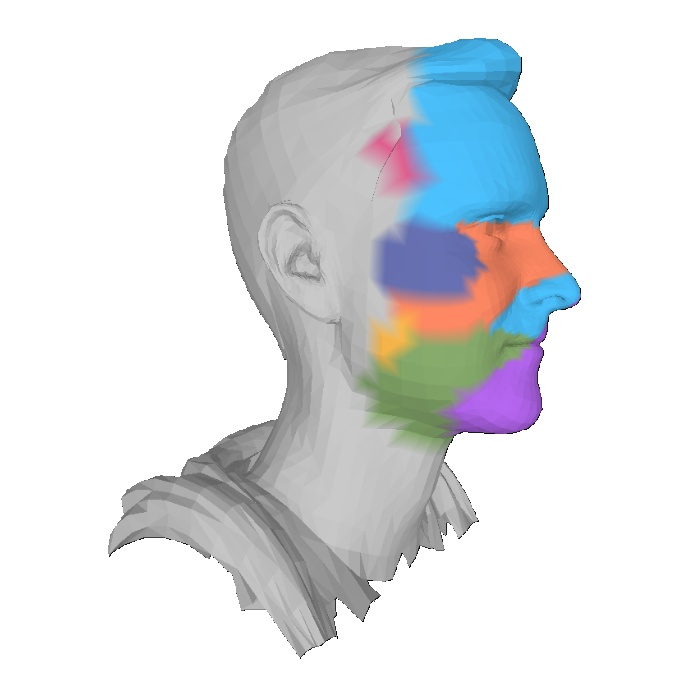} &
\includegraphics[width=0.1\textwidth]{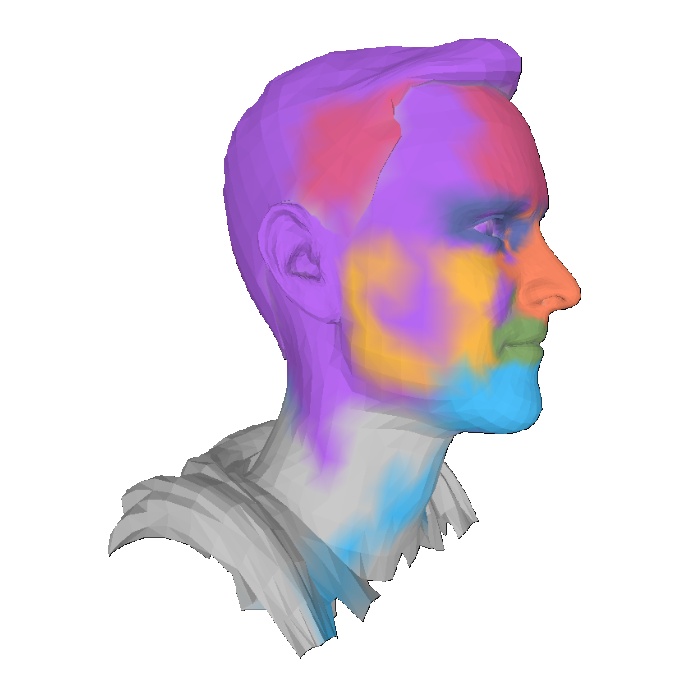} &
\includegraphics[width=0.1\textwidth]{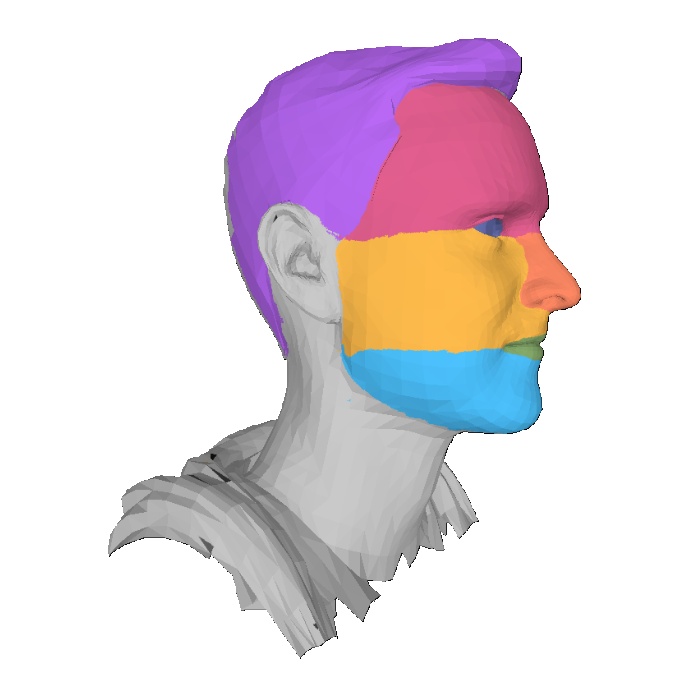} \\

\addlinespace[-2pt]
\arrayrulecolor{gray}\cmidrule(lr){1-4}
\arrayrulecolor{black}

\includegraphics[width=0.1\textwidth]{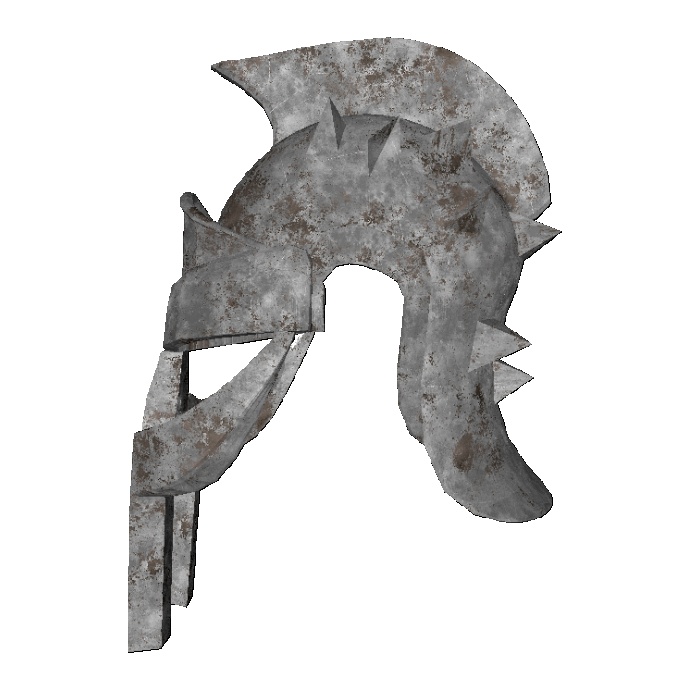} &
\includegraphics[width=0.1\textwidth]{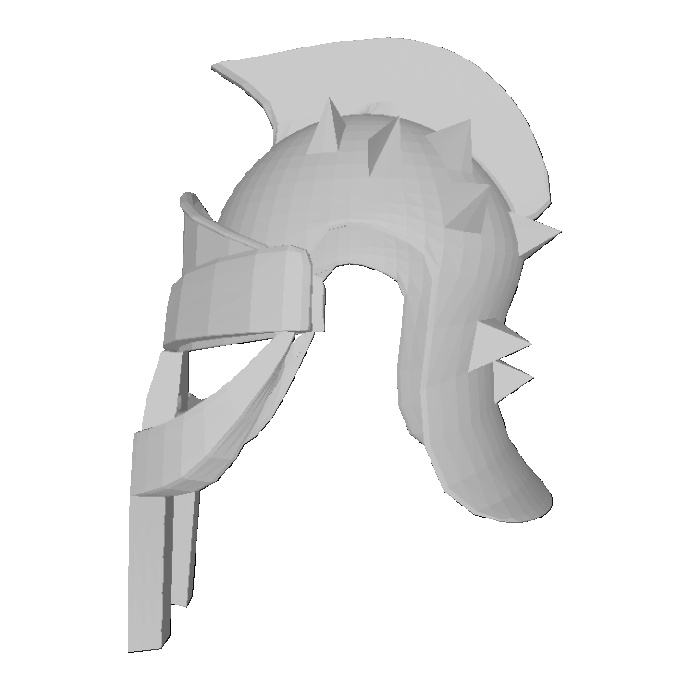} &
\includegraphics[width=0.1\textwidth]{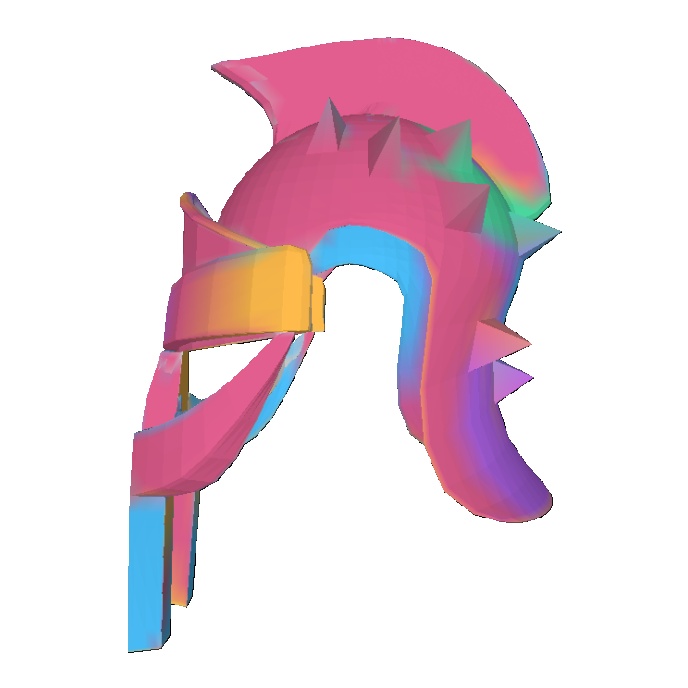} &
\includegraphics[width=0.1\textwidth]{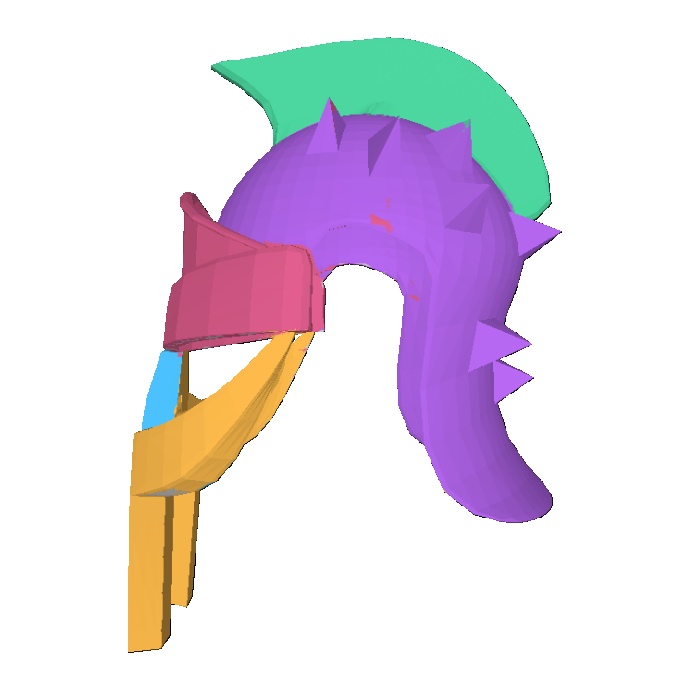} \\

\addlinespace[-2pt]
\arrayrulecolor{gray}\cmidrule(lr){1-4}
\arrayrulecolor{black}

\includegraphics[width=0.1\textwidth]{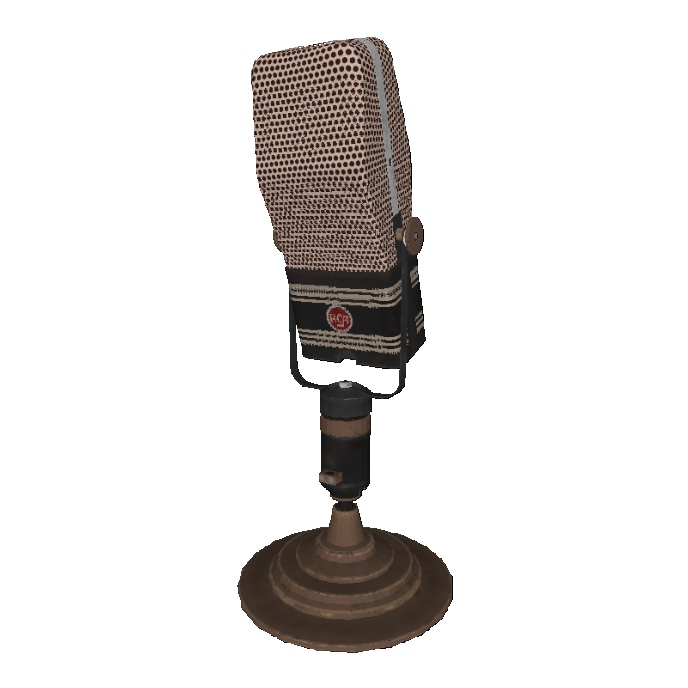} &
\includegraphics[width=0.1\textwidth]{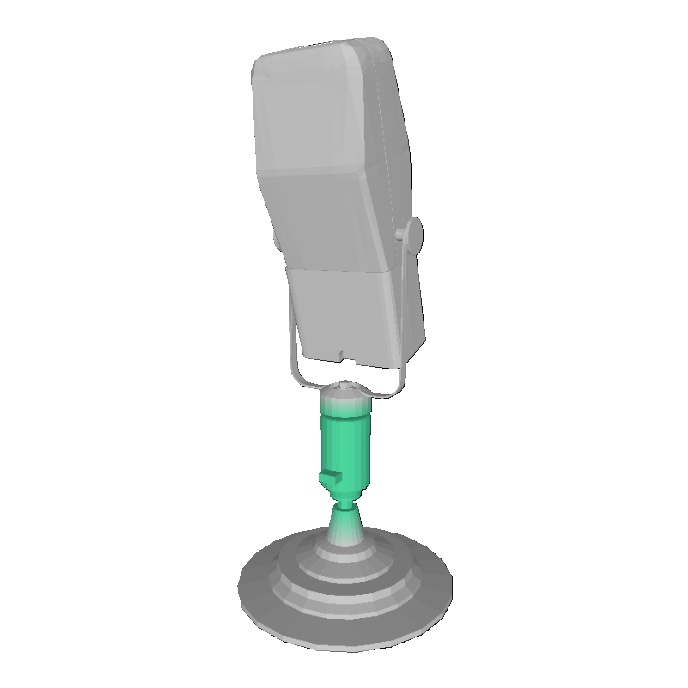} &
\includegraphics[width=0.1\textwidth]{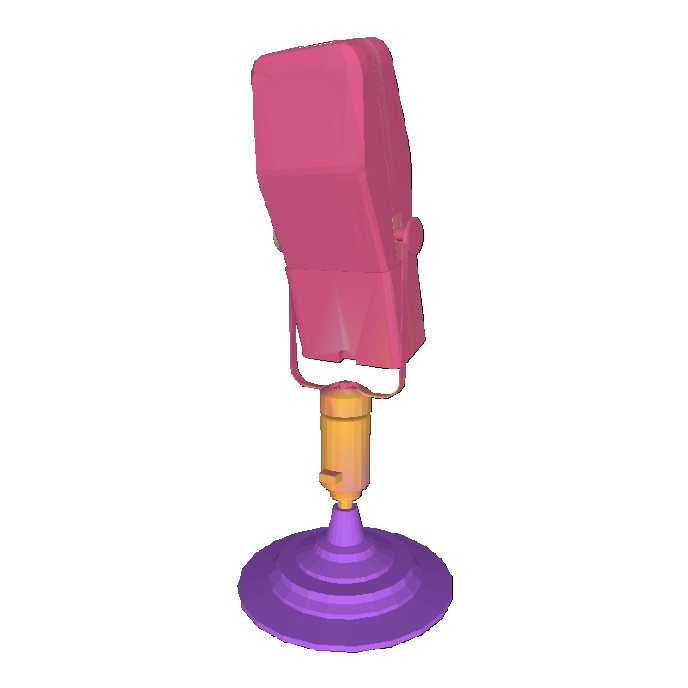} &
\includegraphics[width=0.1\textwidth]{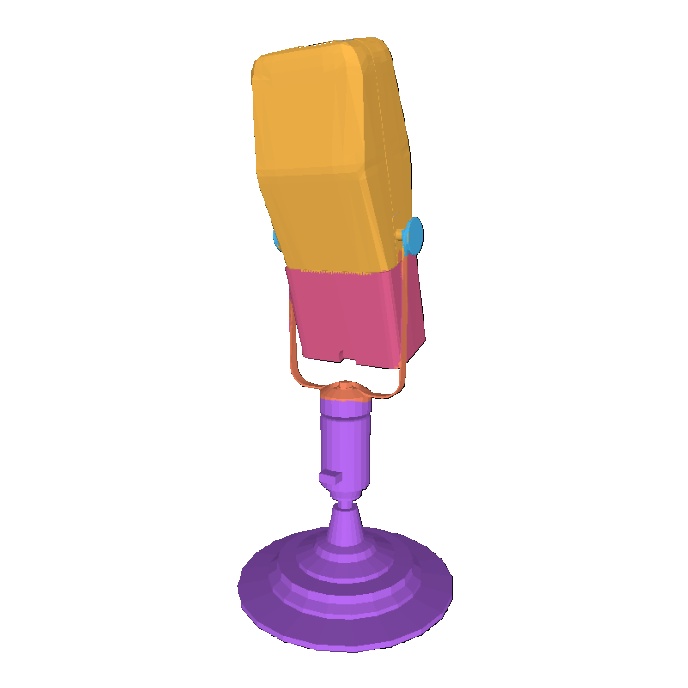} \\

\addlinespace[-2pt]
\arrayrulecolor{gray}\cmidrule(lr){1-4}
\arrayrulecolor{black}

\includegraphics[width=0.1\textwidth]{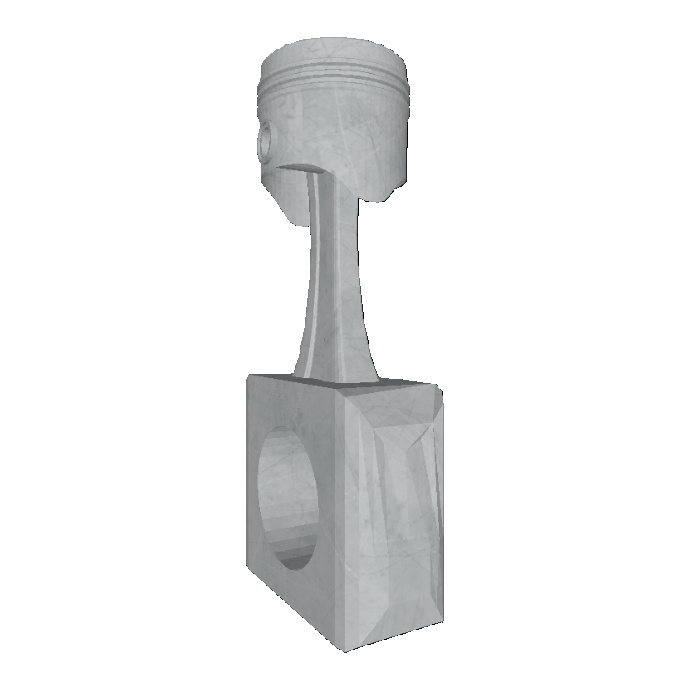} &
\includegraphics[width=0.1\textwidth]{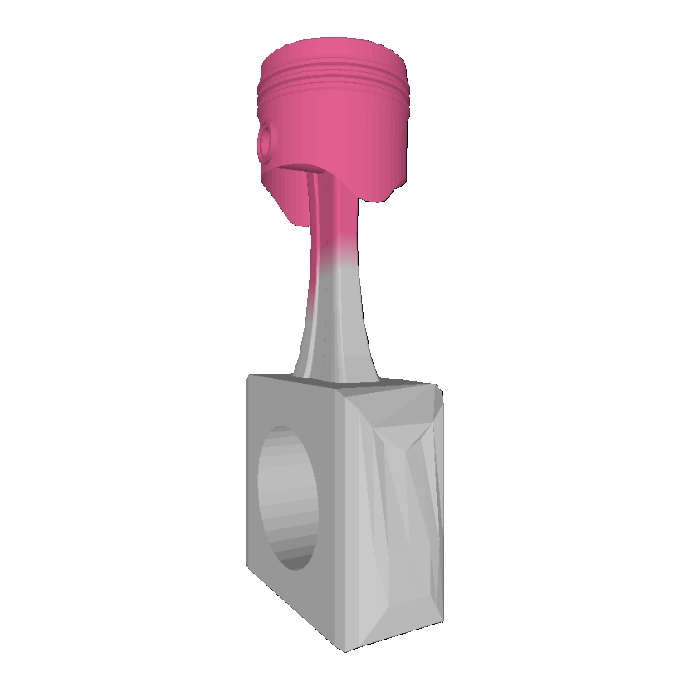} &
\includegraphics[width=0.1\textwidth]{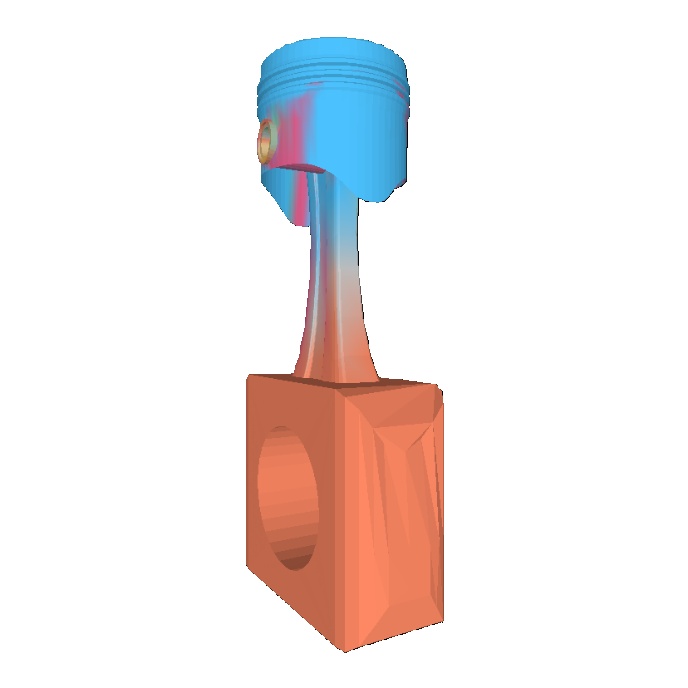} &
\includegraphics[width=0.1\textwidth]{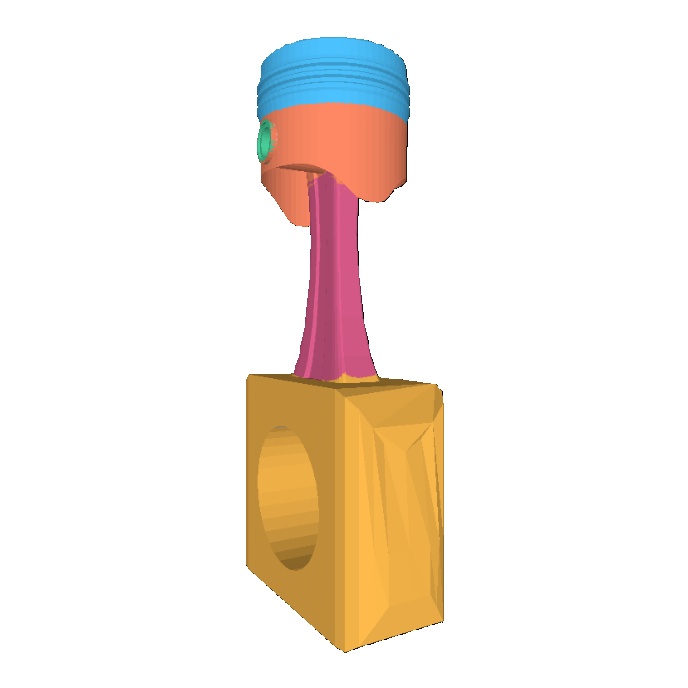} \\

\addlinespace[-2pt]
\arrayrulecolor{gray}\cmidrule(lr){1-4}
\arrayrulecolor{black}

\includegraphics[width=0.1\textwidth]{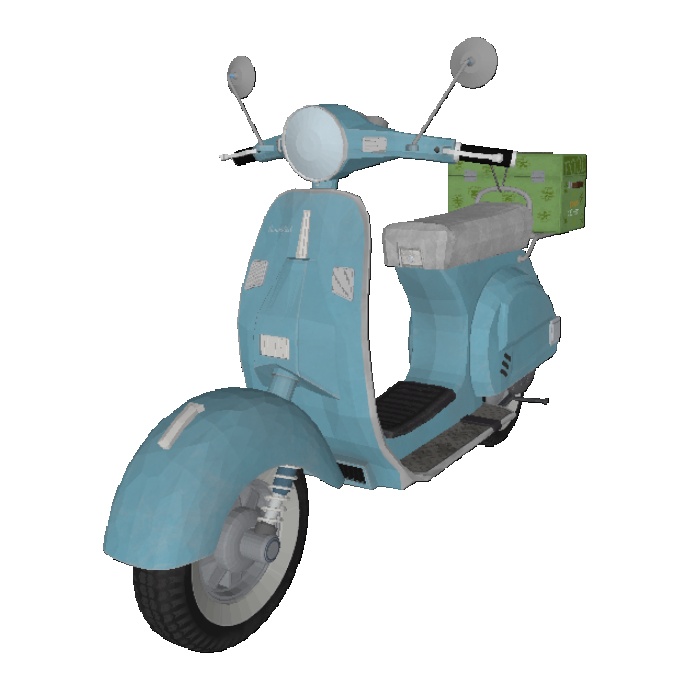} &
\includegraphics[width=0.1\textwidth]{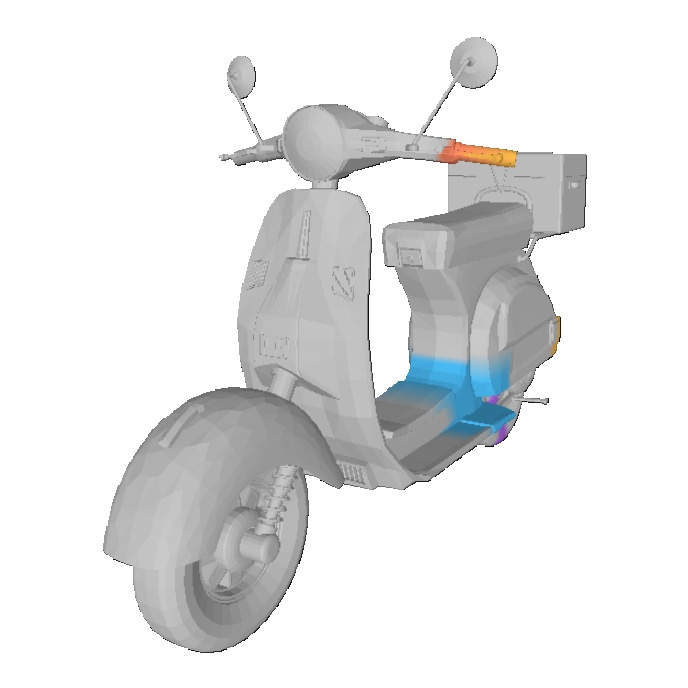} &
\includegraphics[width=0.1\textwidth]{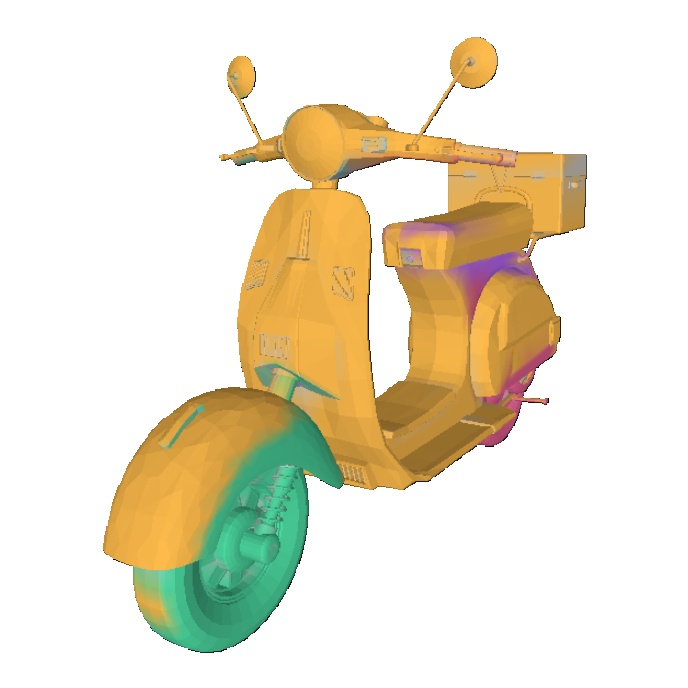} &
\includegraphics[width=0.1\textwidth]{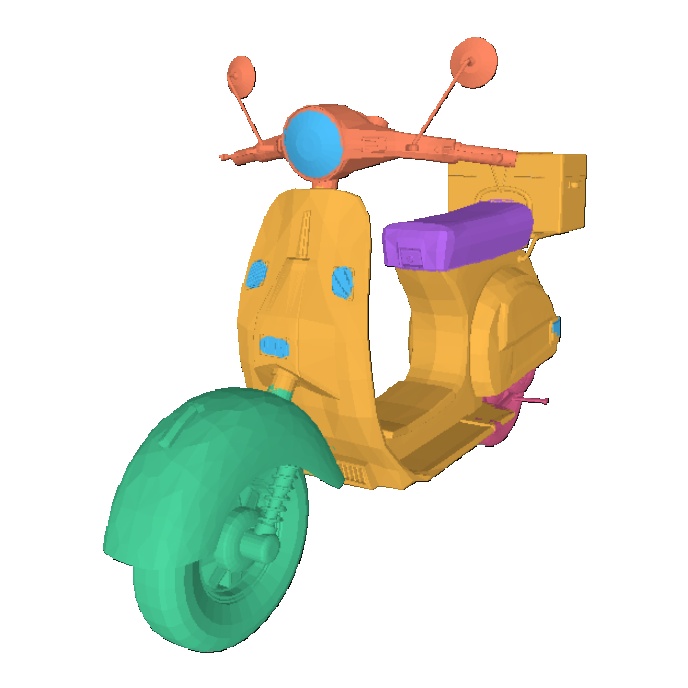} \\

\addlinespace[-2pt]
\arrayrulecolor{gray}\cmidrule(lr){1-4}
\arrayrulecolor{black}

\includegraphics[width=0.1\textwidth]{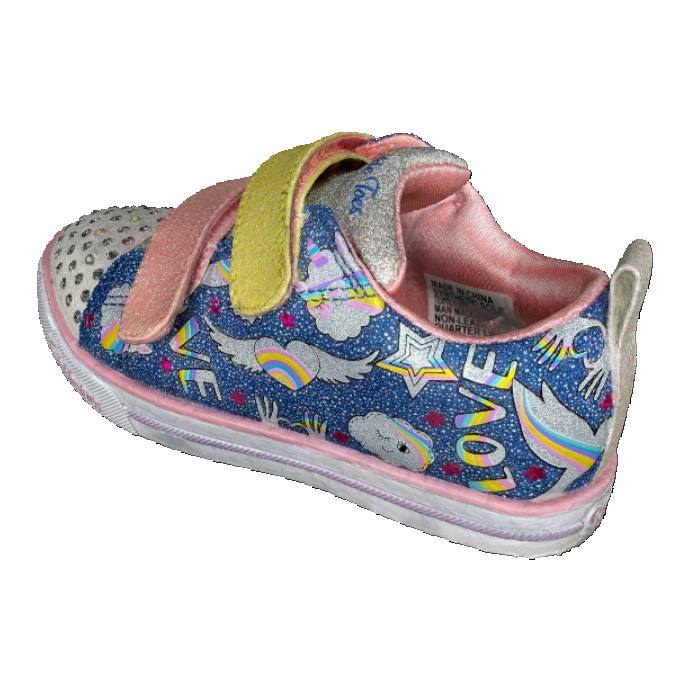} &
\includegraphics[width=0.1\textwidth]{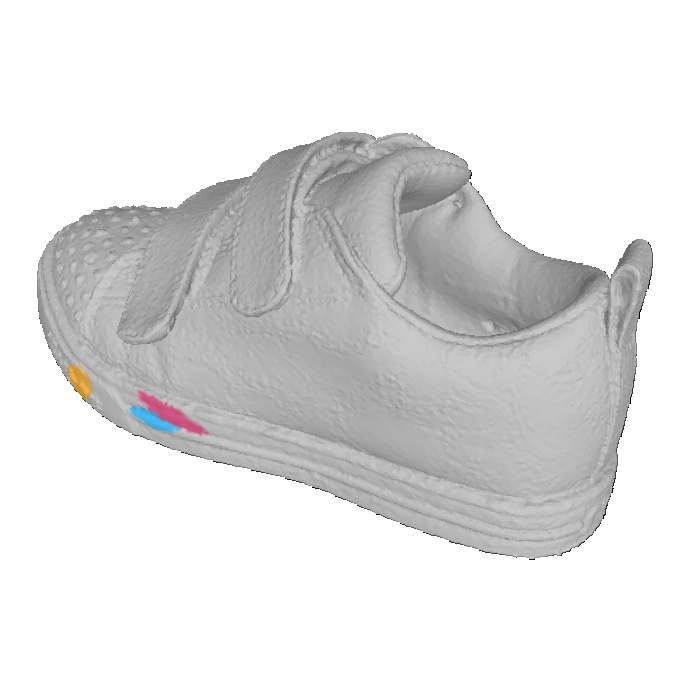} &
\includegraphics[width=0.1\textwidth]{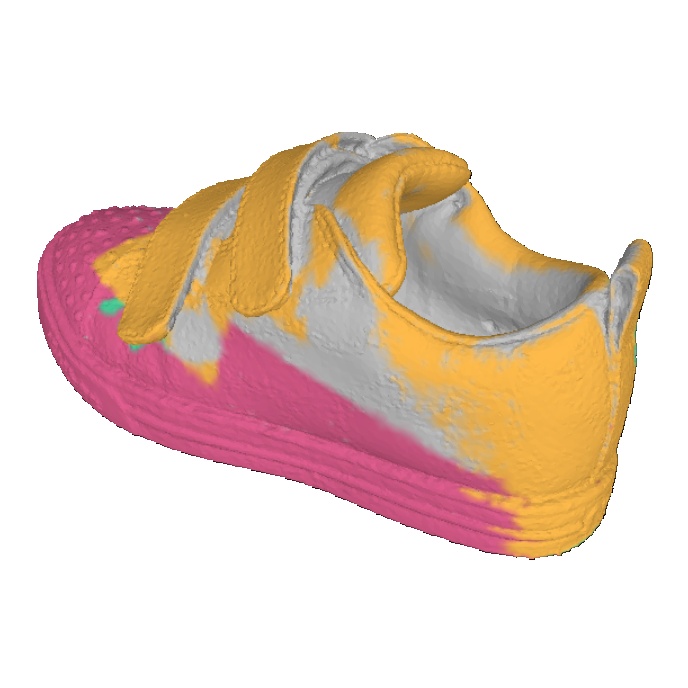} &
\includegraphics[width=0.1\textwidth]{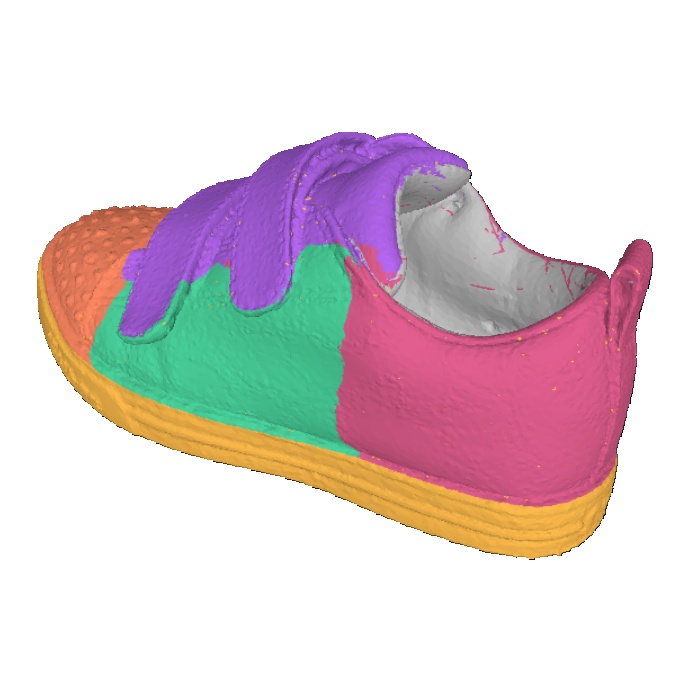} \\

\addlinespace[-2pt]
\arrayrulecolor{gray}\cmidrule(lr){1-4}
\arrayrulecolor{black}

\includegraphics[width=0.1\textwidth]{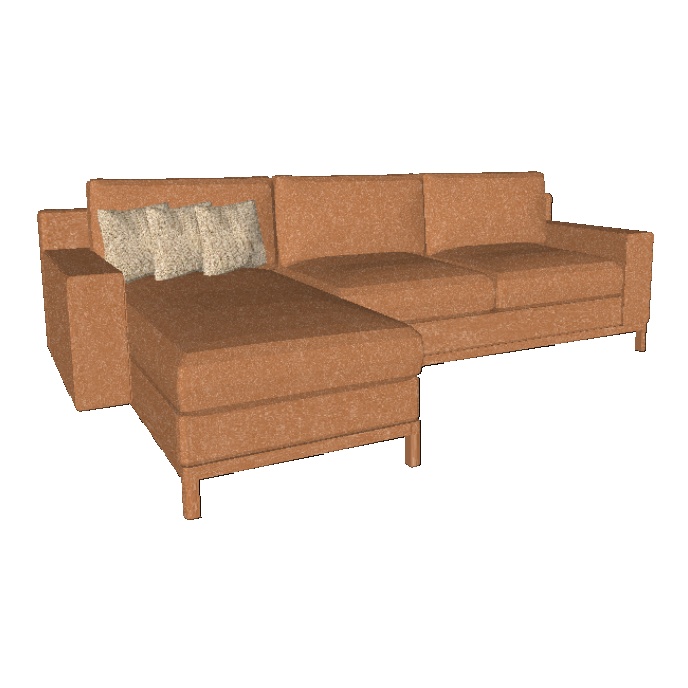} &
\includegraphics[width=0.1\textwidth]{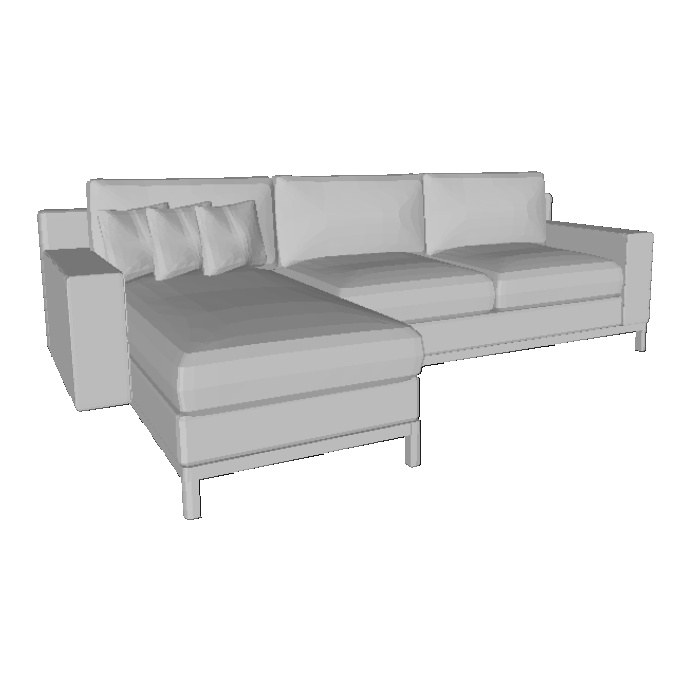} &
\includegraphics[width=0.1\textwidth]{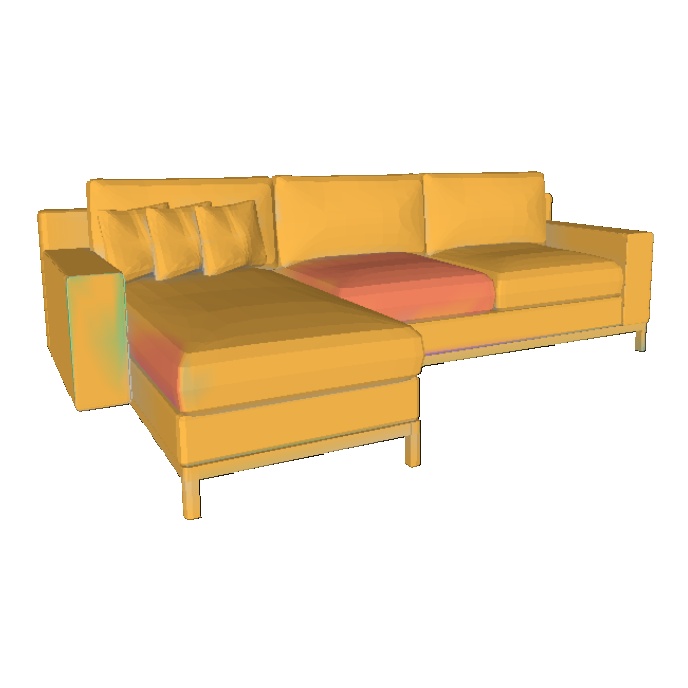} &
\includegraphics[width=0.1\textwidth]{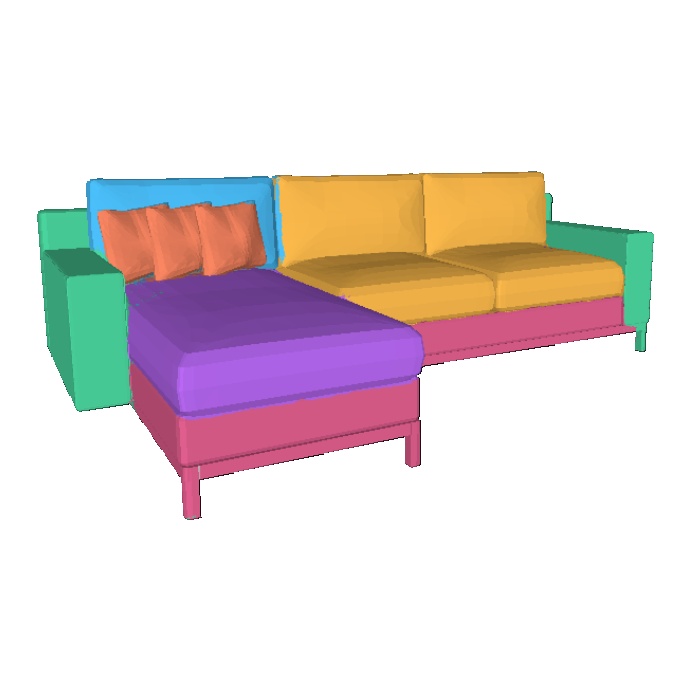} \\

\addlinespace[-2pt]
\arrayrulecolor{gray}\cmidrule(lr){1-4}
\arrayrulecolor{black}

\includegraphics[width=0.1\textwidth]{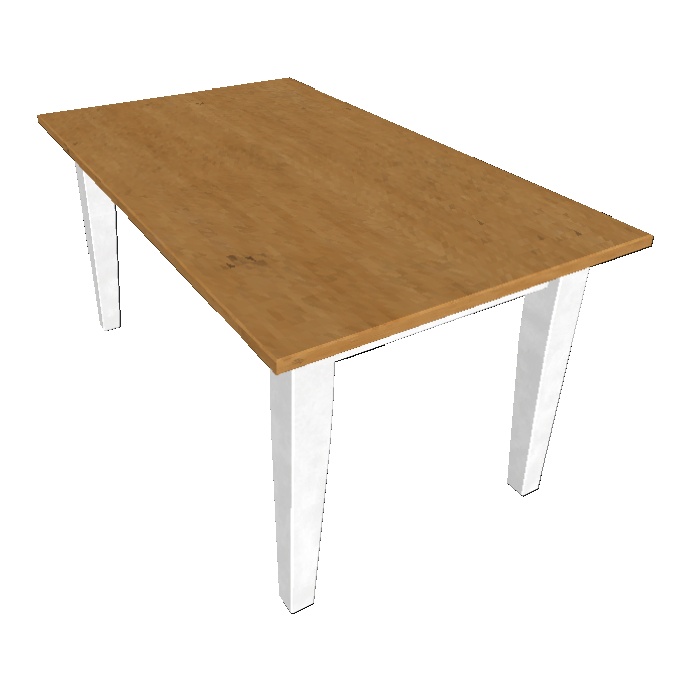} &
\includegraphics[width=0.1\textwidth]{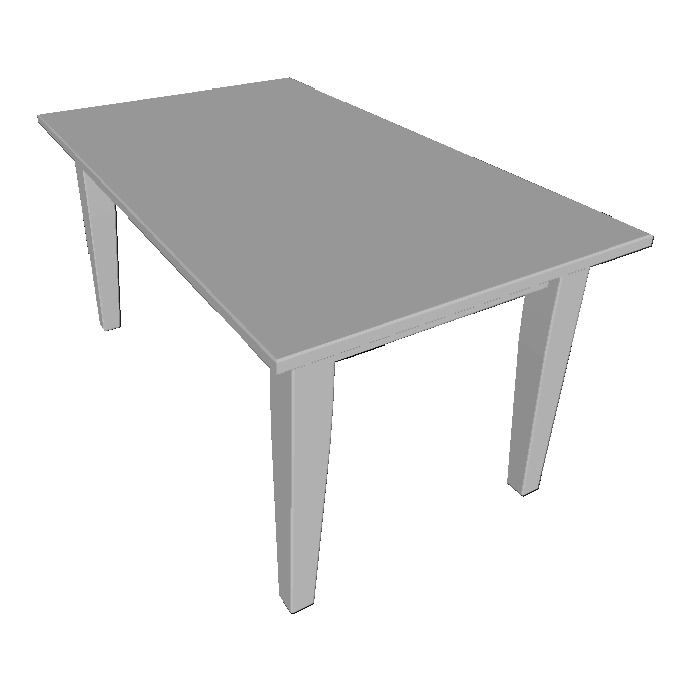} &
\includegraphics[width=0.1\textwidth]{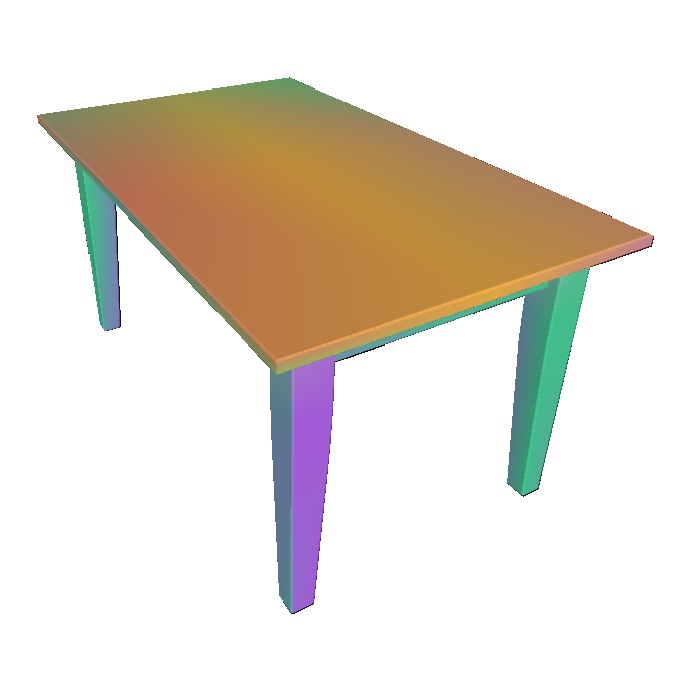} &
\includegraphics[width=0.1\textwidth]{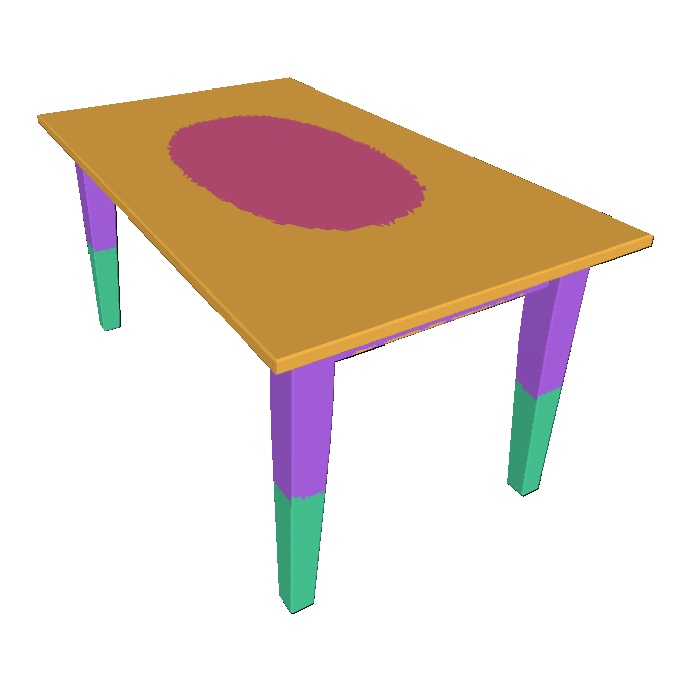} \\

\addlinespace[-2pt]
\bottomrule
\end{tabular}
}}
&
\vtop{\vskip0pt
\resizebox{0.44\textwidth}{!}{
\begin{tabular}{@{}c@{}c@{}c@{}c@{}c@{}}
\toprule
\small{Input} & \small{3D Highlighter} & \small{SATR} & \small{Ours} \\ \midrule
\includegraphics[width=0.1\textwidth]{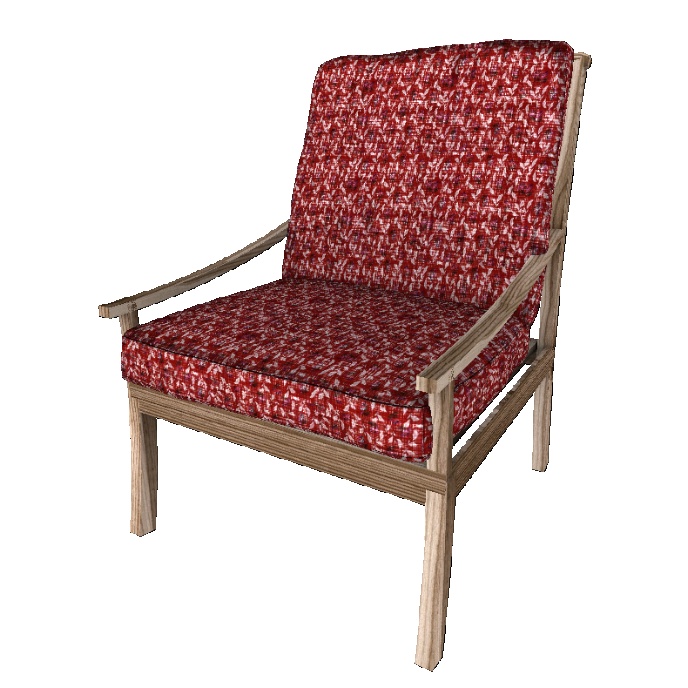} &
\includegraphics[width=0.1\textwidth]{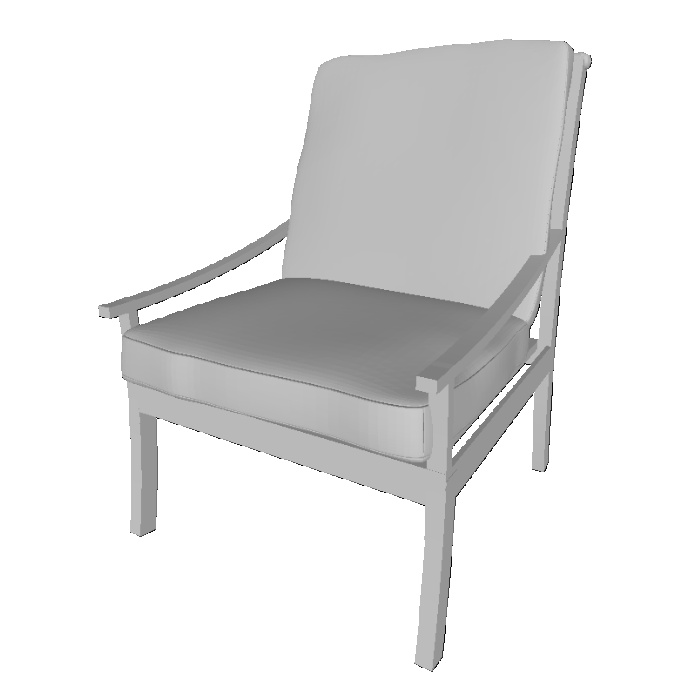} &
\includegraphics[width=0.1\textwidth]{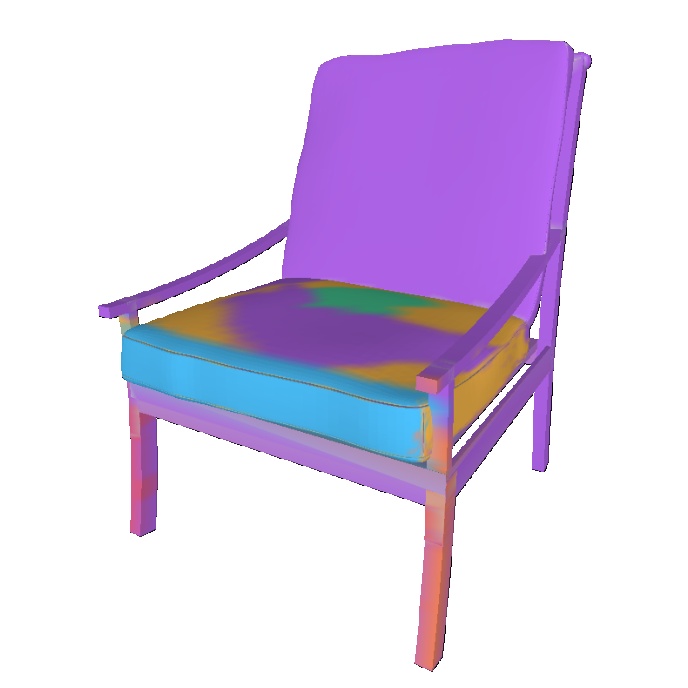} &
\includegraphics[width=0.1\textwidth]{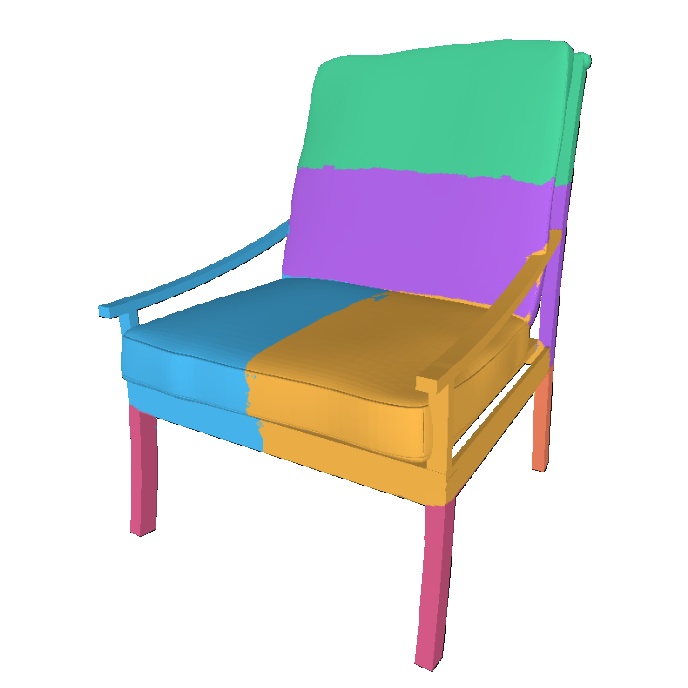} \\

\addlinespace[-2pt]
\arrayrulecolor{gray}\cmidrule(lr){1-4}
\arrayrulecolor{black}

\includegraphics[width=0.1\textwidth]{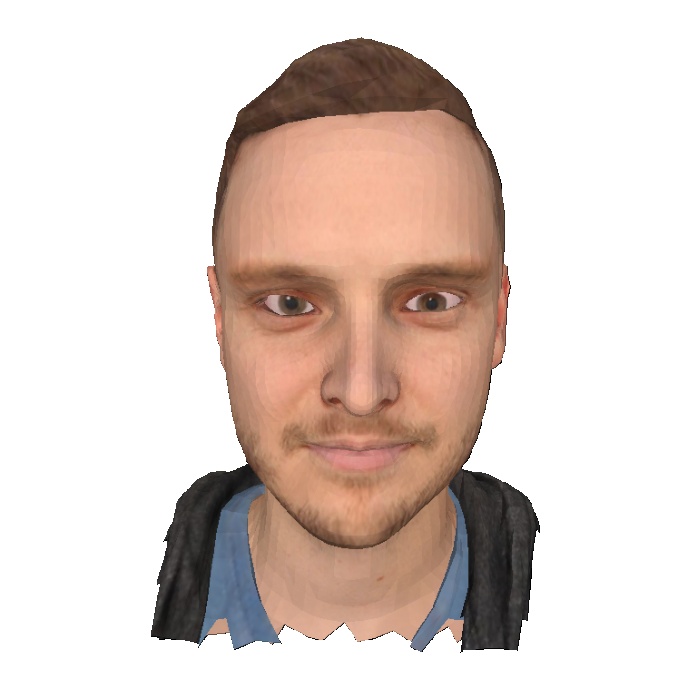} &
\includegraphics[width=0.1\textwidth]{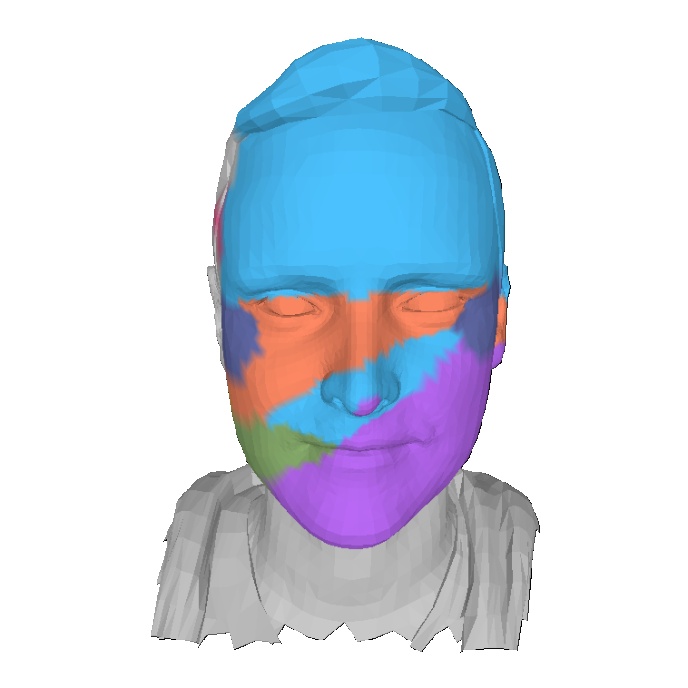} &
\includegraphics[width=0.1\textwidth]{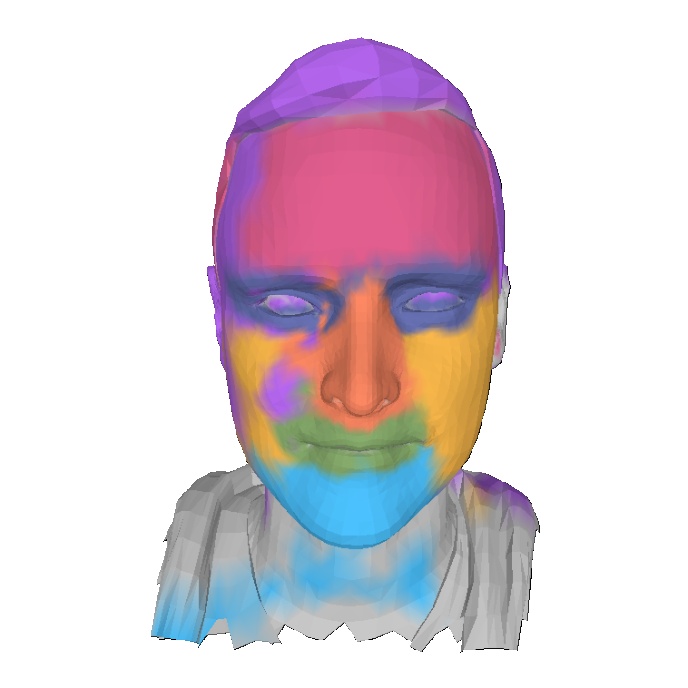} &
\includegraphics[width=0.1\textwidth]{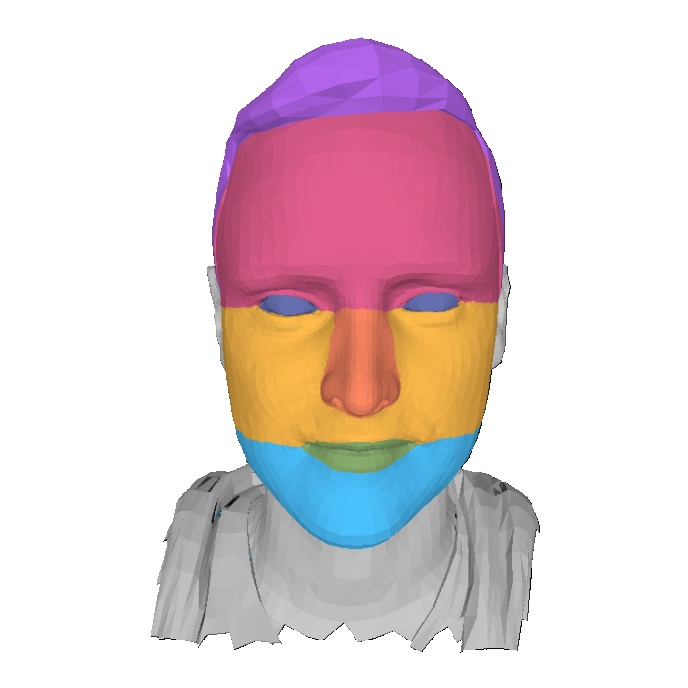} \\

\addlinespace[-2pt]
\arrayrulecolor{gray}\cmidrule(lr){1-4}
\arrayrulecolor{black}

\includegraphics[width=0.1\textwidth]{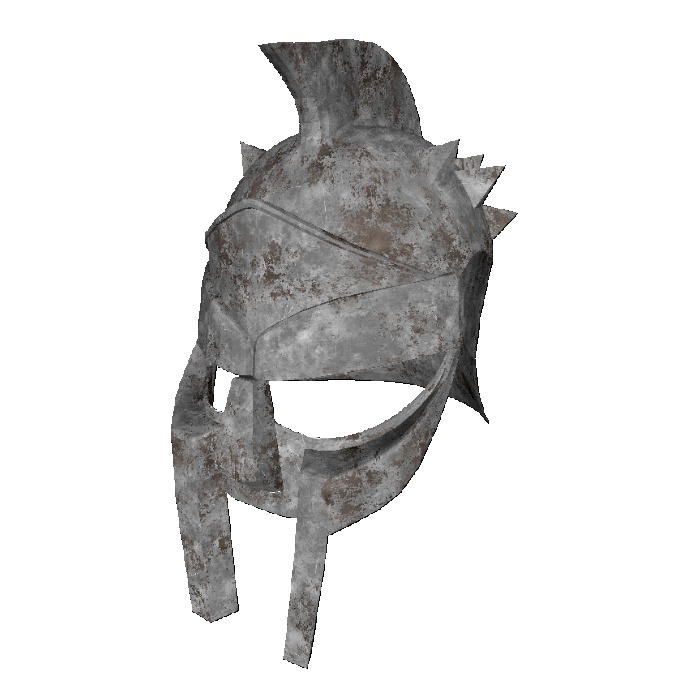} &
\includegraphics[width=0.1\textwidth]{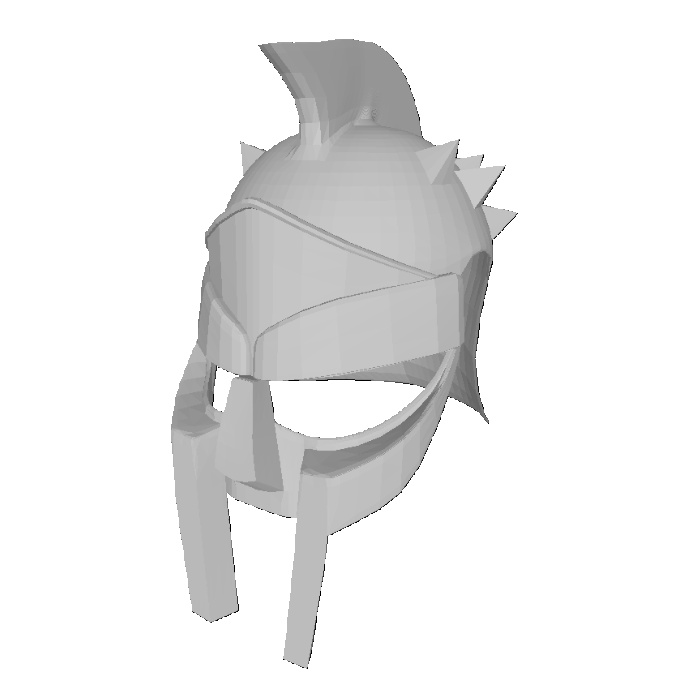} &
\includegraphics[width=0.1\textwidth]{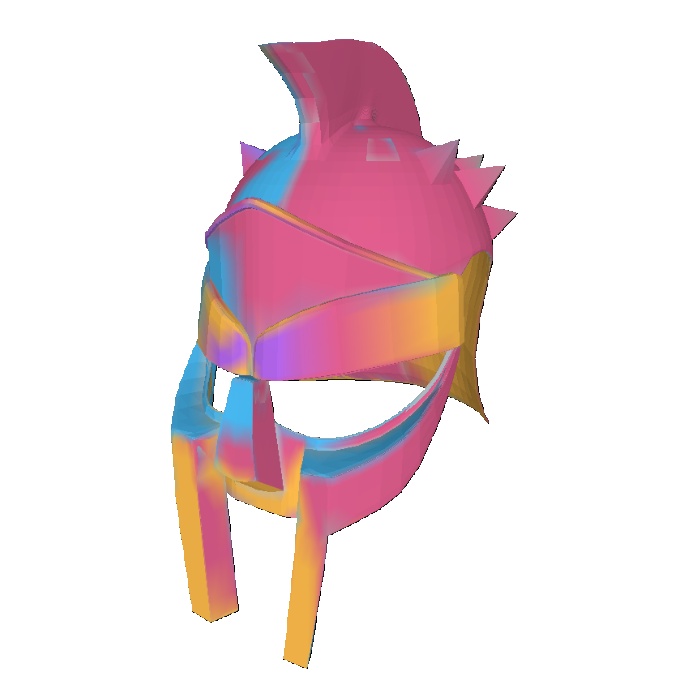} &
\includegraphics[width=0.1\textwidth]{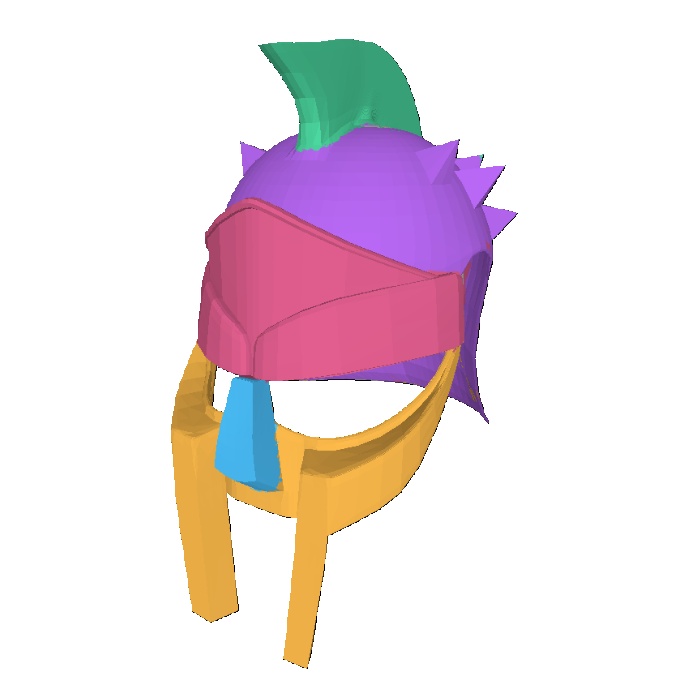} \\

\addlinespace[-2pt]
\arrayrulecolor{gray}\cmidrule(lr){1-4}
\arrayrulecolor{black}

\includegraphics[width=0.1\textwidth]{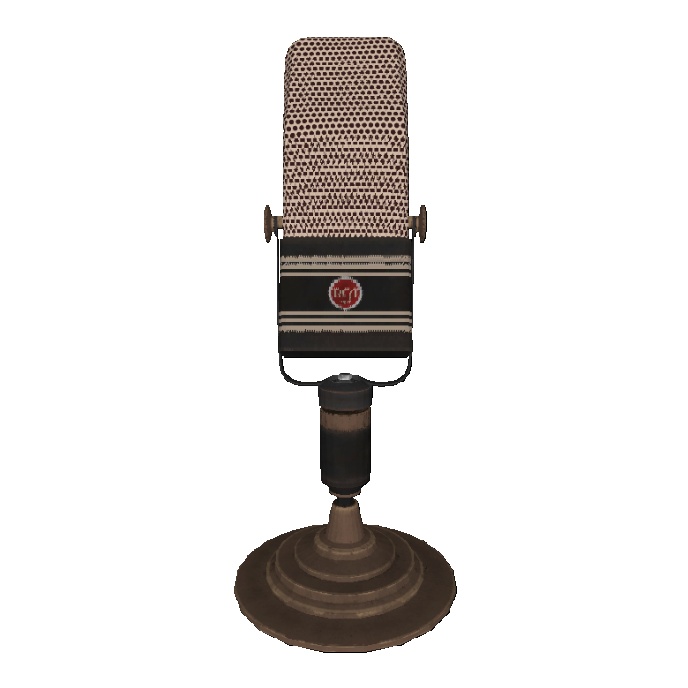} &
\includegraphics[width=0.1\textwidth]{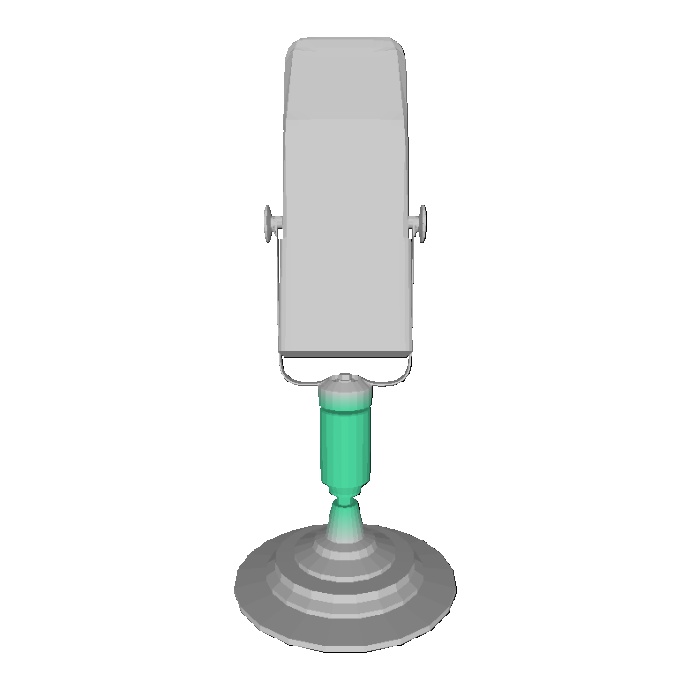} &
\includegraphics[width=0.1\textwidth]{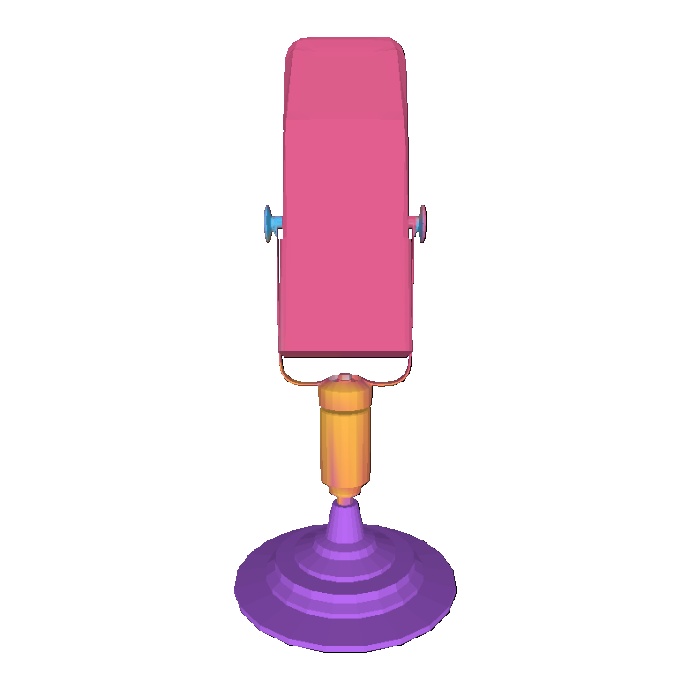} &
\includegraphics[width=0.1\textwidth]{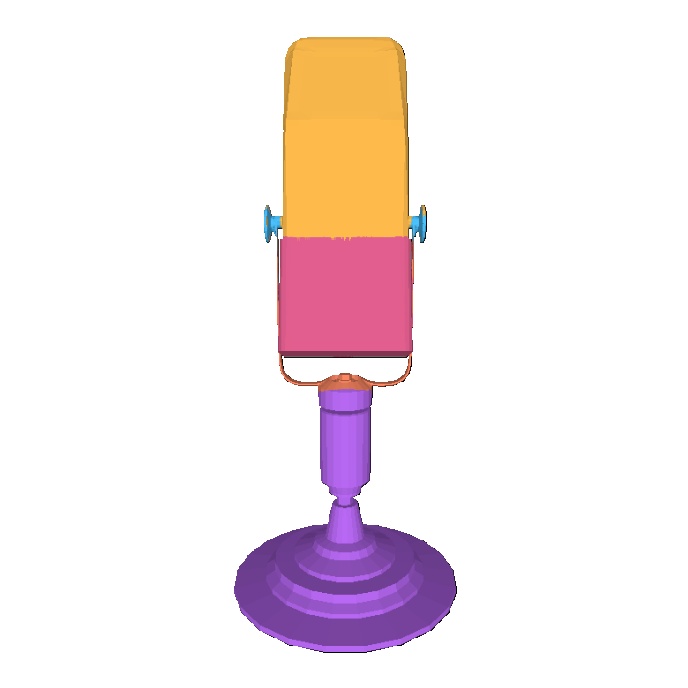} \\

\addlinespace[-2pt]
\arrayrulecolor{gray}\cmidrule(lr){1-4}
\arrayrulecolor{black}

\includegraphics[width=0.1\textwidth]{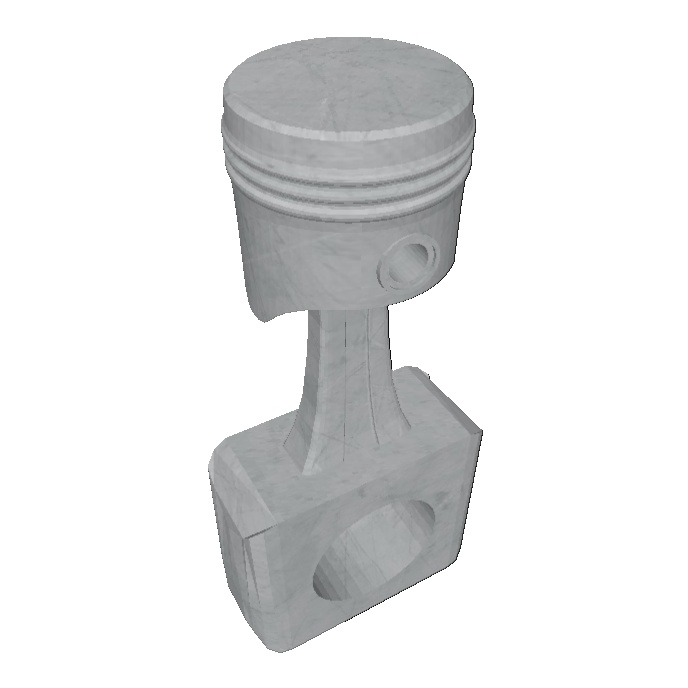} &
\includegraphics[width=0.1\textwidth]{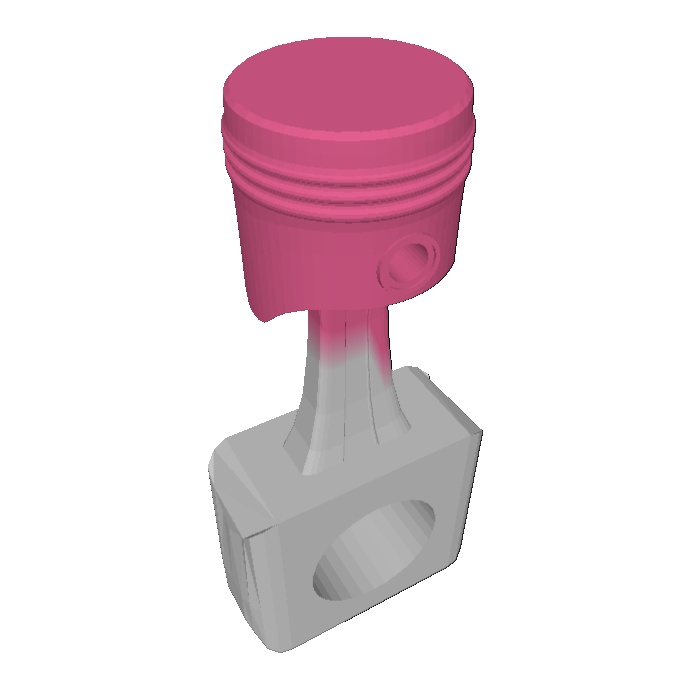} &
\includegraphics[width=0.1\textwidth]{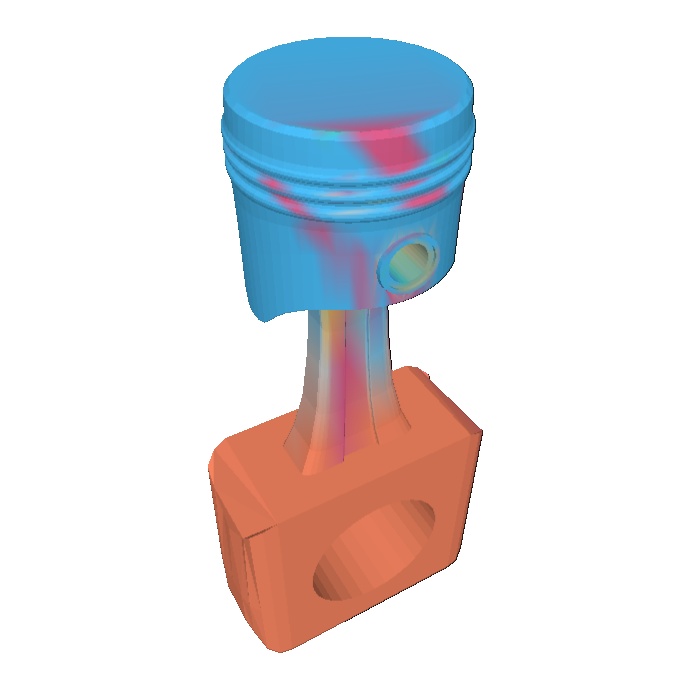} &
\includegraphics[width=0.1\textwidth]{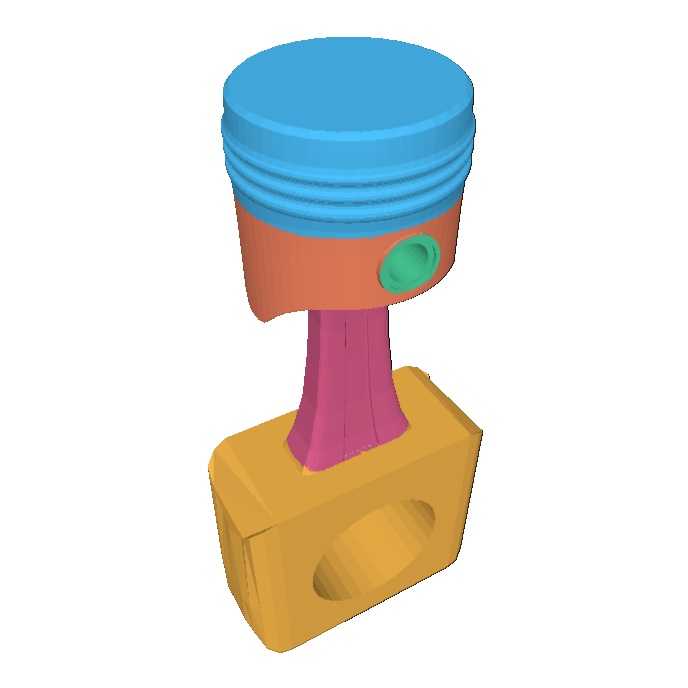} \\

\addlinespace[-2pt]
\arrayrulecolor{gray}\cmidrule(lr){1-4}
\arrayrulecolor{black}

\includegraphics[width=0.1\textwidth]{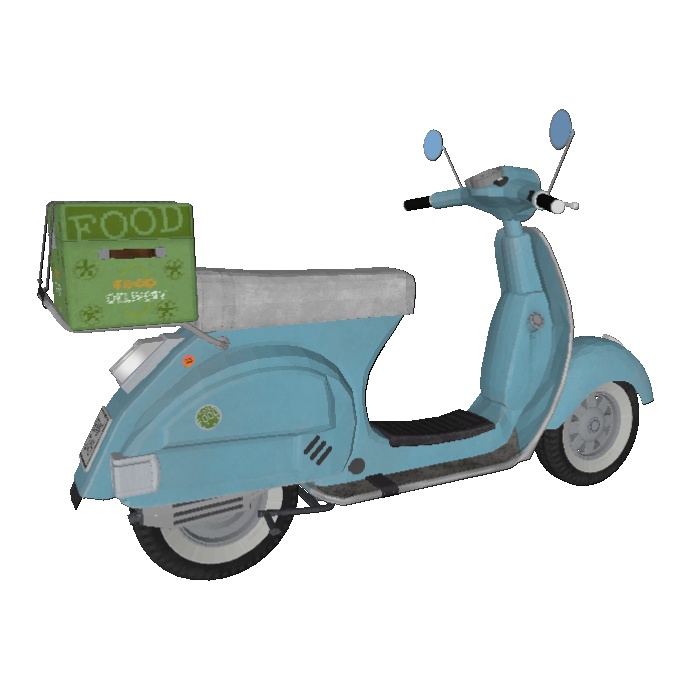} &
\includegraphics[width=0.1\textwidth]{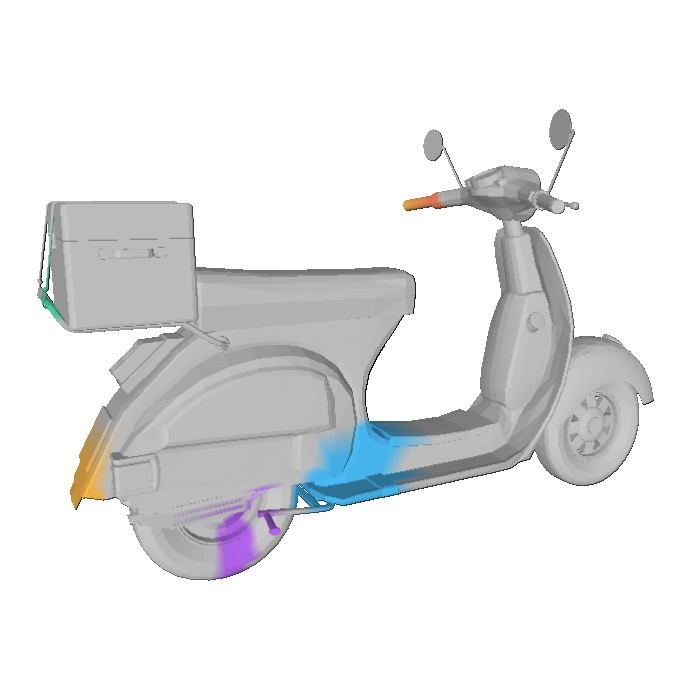} &
\includegraphics[width=0.1\textwidth]{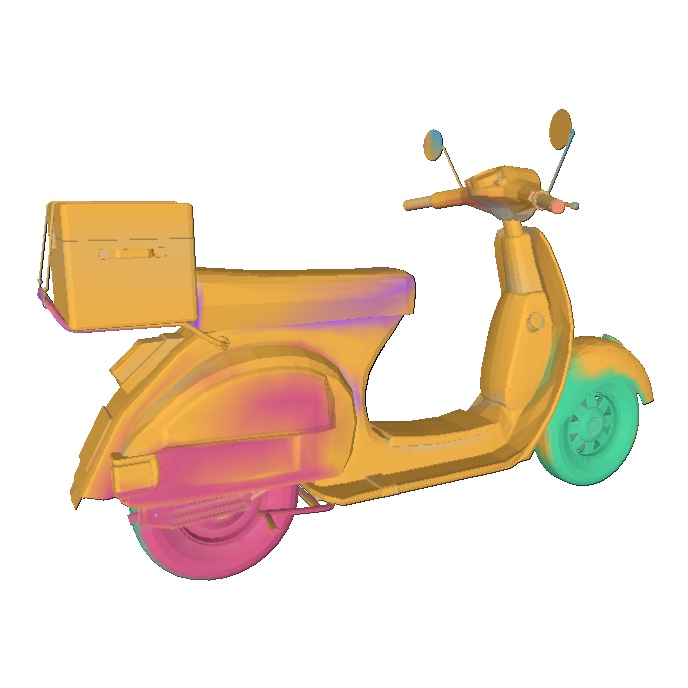} &
\includegraphics[width=0.1\textwidth]{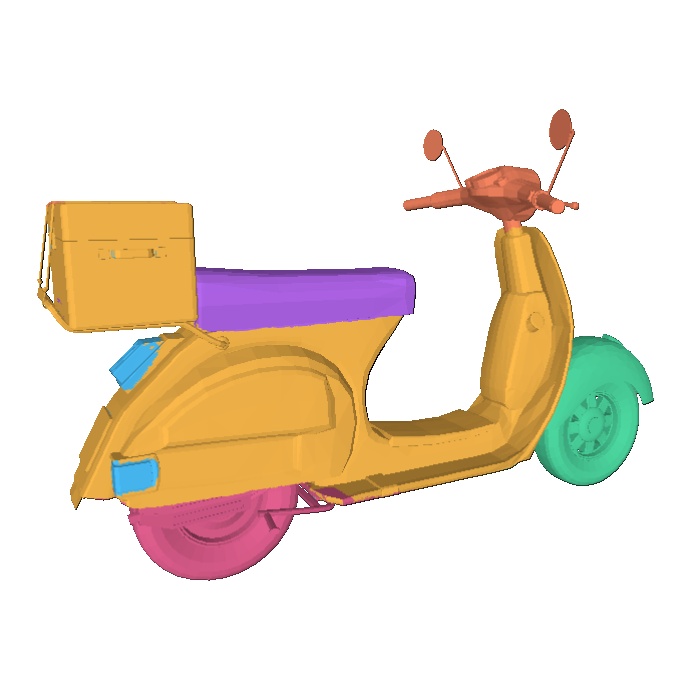} \\

\addlinespace[-2pt]
\arrayrulecolor{gray}\cmidrule(lr){1-4}
\arrayrulecolor{black}

\includegraphics[width=0.1\textwidth]{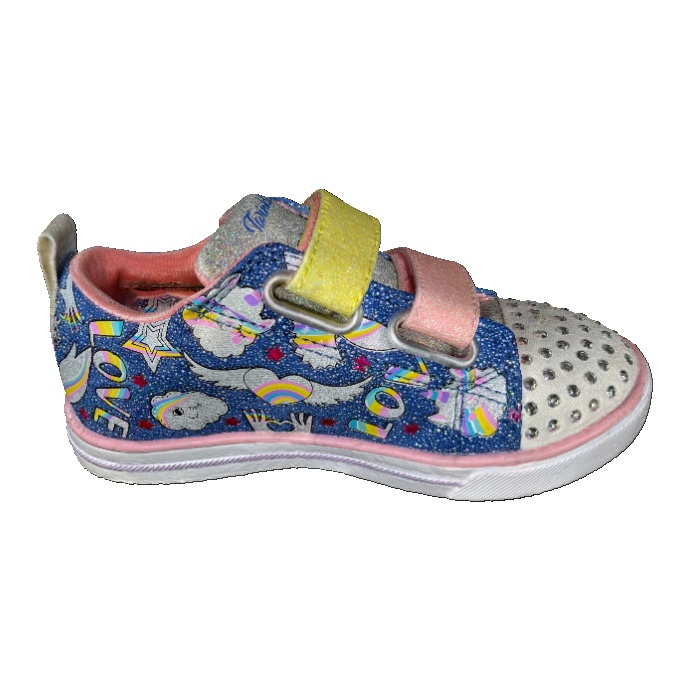} &
\includegraphics[width=0.1\textwidth]{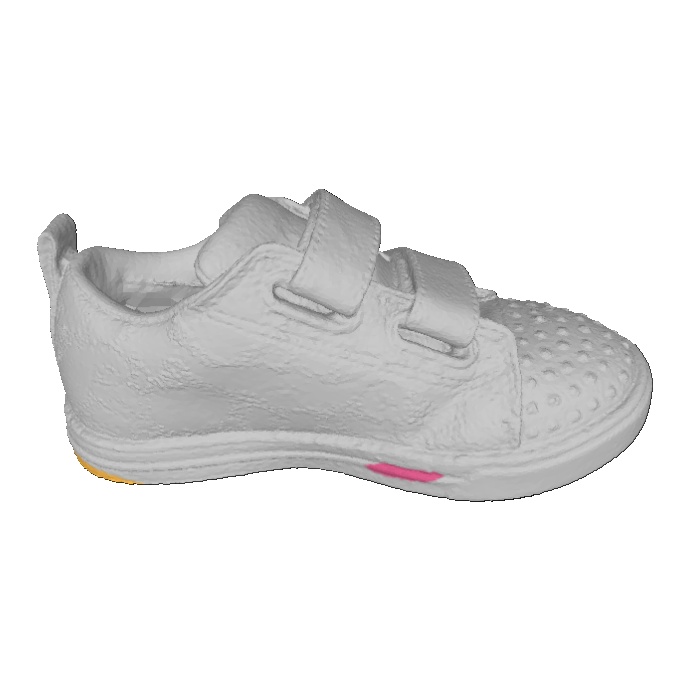} &
\includegraphics[width=0.1\textwidth]{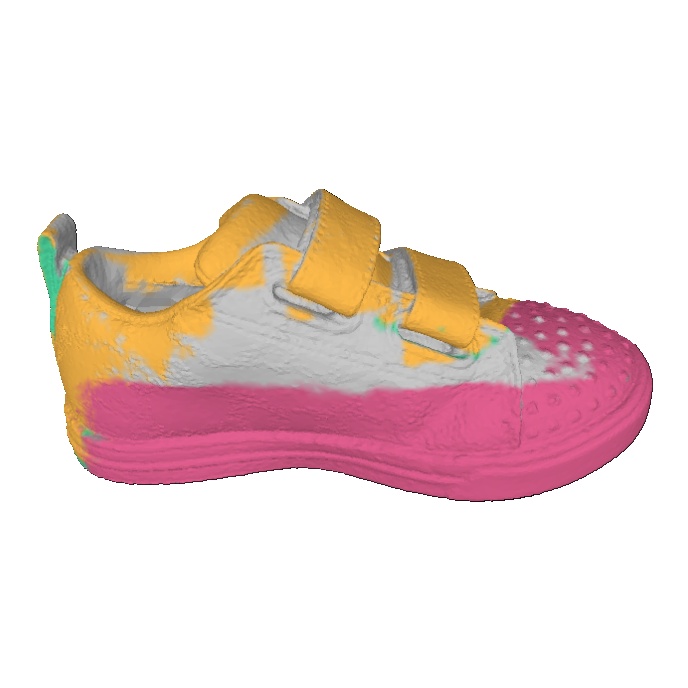} &
\includegraphics[width=0.1\textwidth]{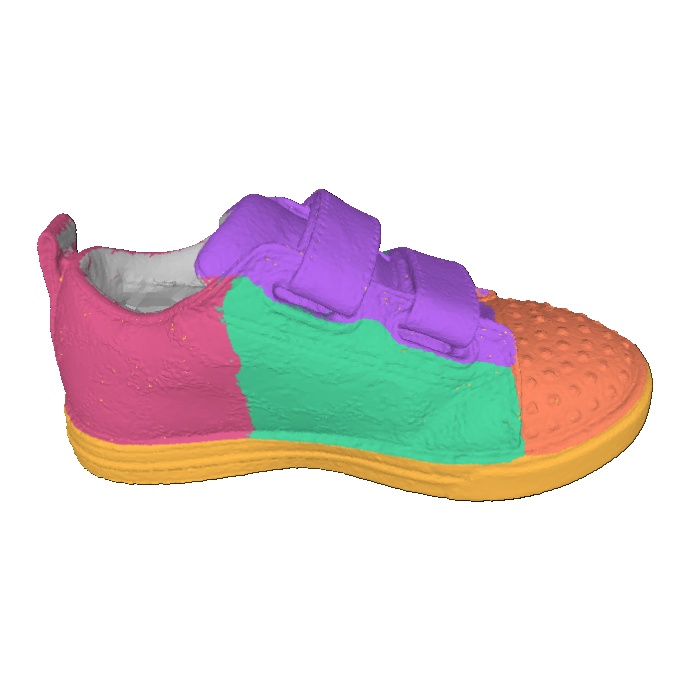} \\

\addlinespace[-2pt]
\arrayrulecolor{gray}\cmidrule(lr){1-4}
\arrayrulecolor{black}

\includegraphics[width=0.1\textwidth]{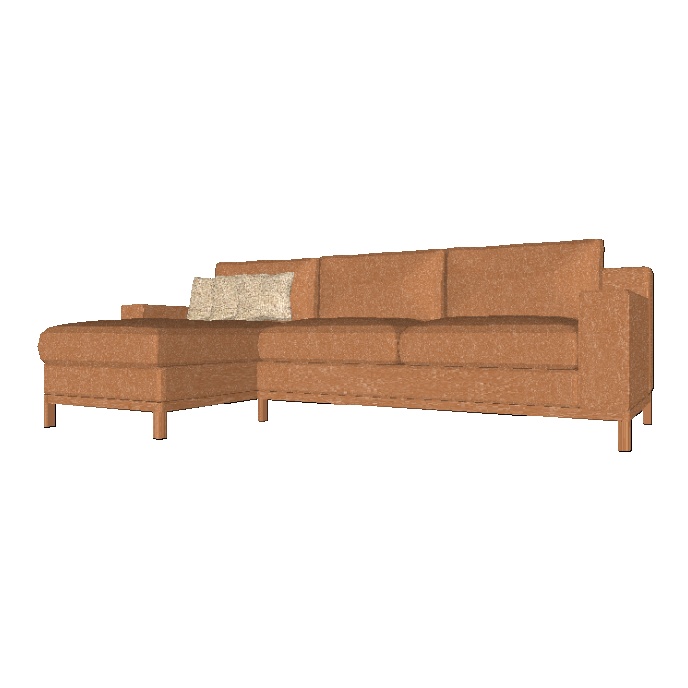} &
\includegraphics[width=0.1\textwidth]{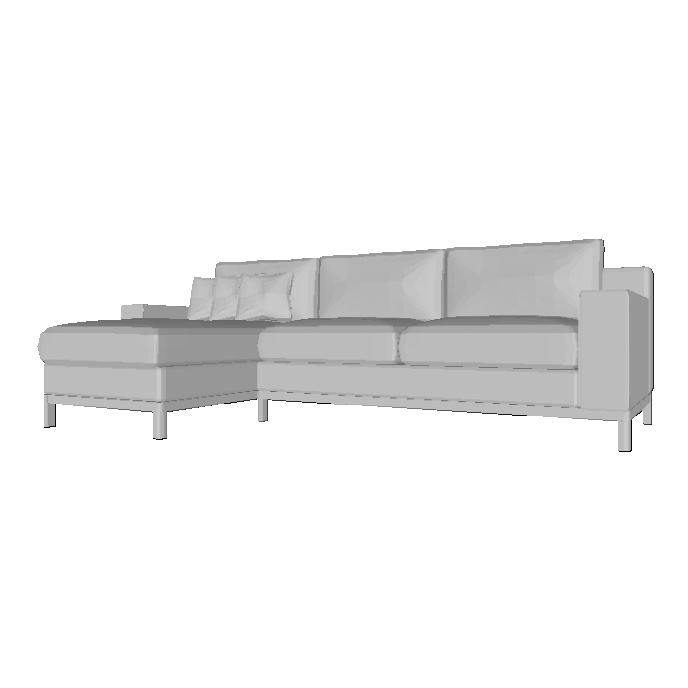} &
\includegraphics[width=0.1\textwidth]{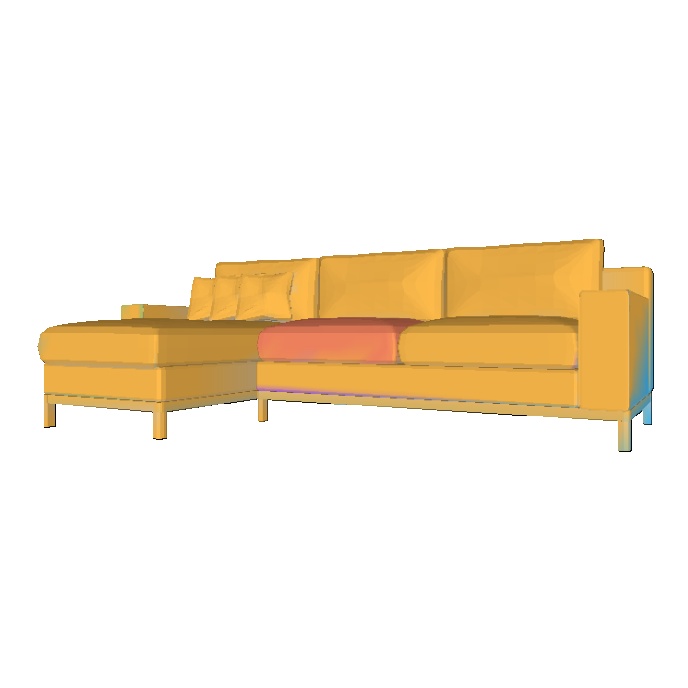} &
\includegraphics[width=0.1\textwidth]{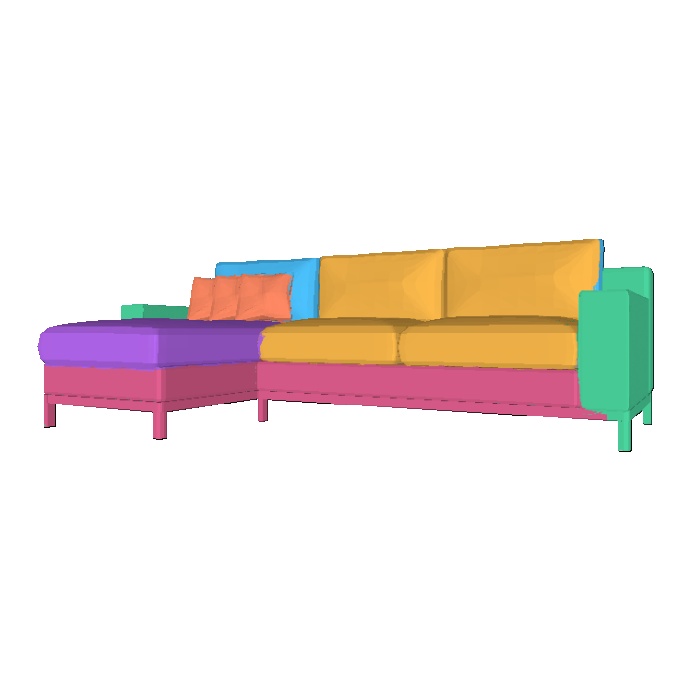} \\

\addlinespace[-2pt]
\arrayrulecolor{gray}\cmidrule(lr){1-4}
\arrayrulecolor{black}

\includegraphics[width=0.1\textwidth]{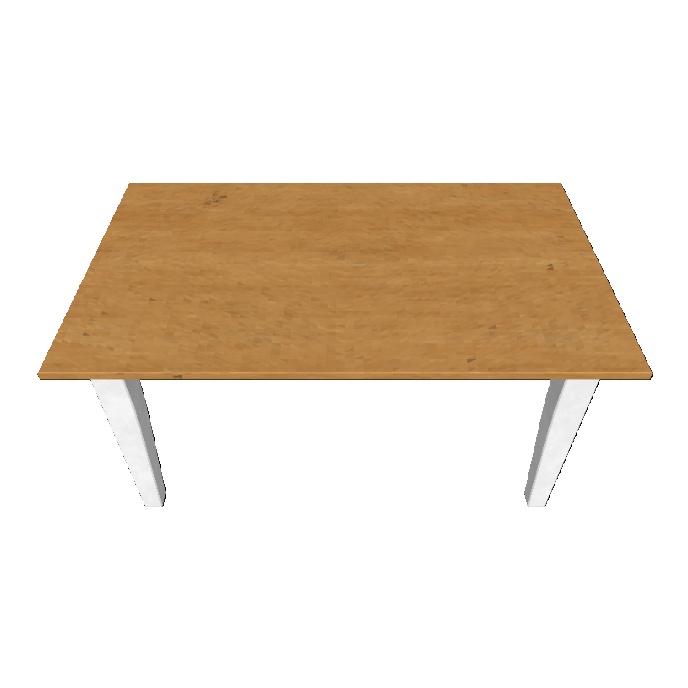} &
\includegraphics[width=0.1\textwidth]{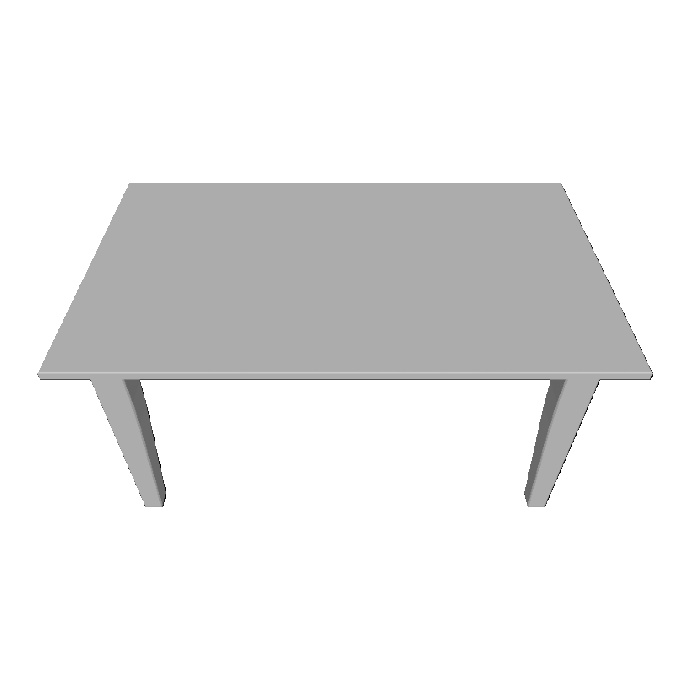} &
\includegraphics[width=0.1\textwidth]{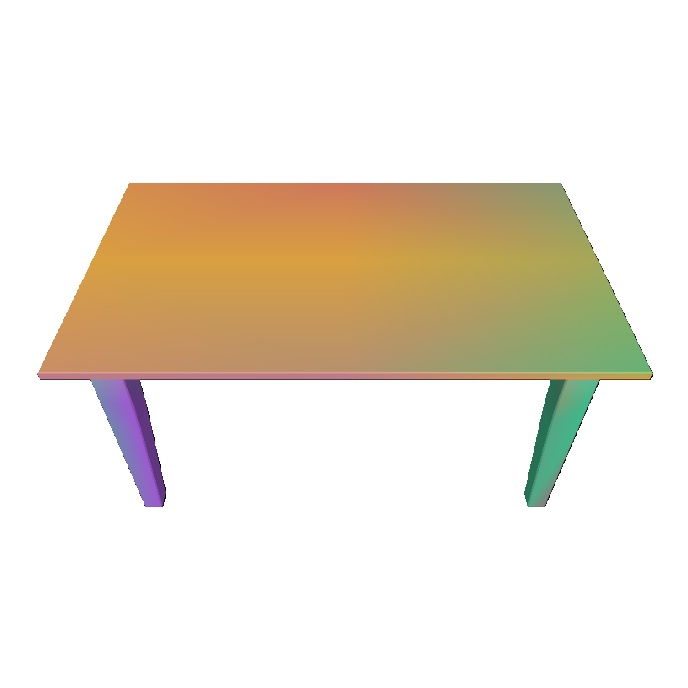} &
\includegraphics[width=0.1\textwidth]{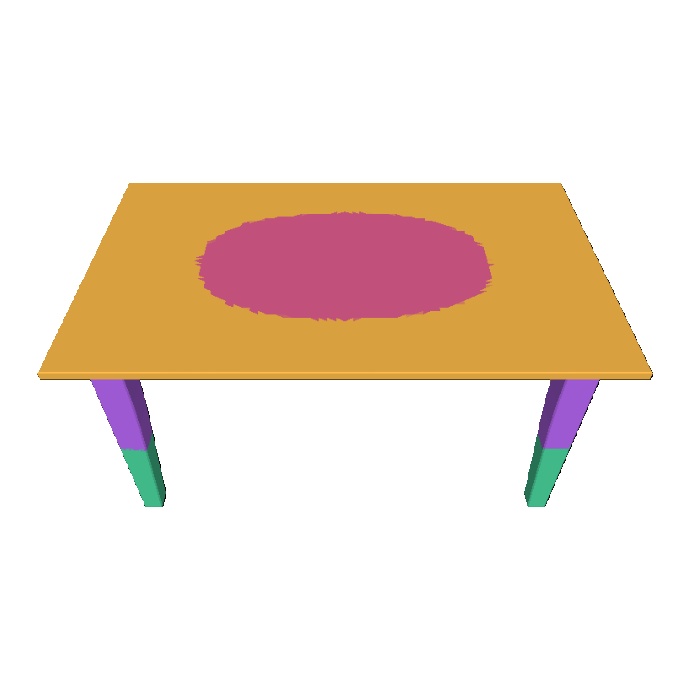} \\

\addlinespace[-2pt]
\bottomrule
\end{tabular}
}}
\end{tabular}
\end{table*}

\begin{table*}[ht]


\centering
\caption{
Qualitative comparison of results from PartSLIP~\cite{liu2023partslip}, PartSTAD~\cite{kim2024partstad}, and Ours for semantic segmentation on the PartNetE~\cite{liu2023partslip} dataset. Each row shows the same object from two different views.
}
\label{tab:PN_full_comp_1}

\begin{tabular}{@{}p{0.5\textwidth}@{} | @{}p{0.5\textwidth}@{}}

\multicolumn{1}{c}{\textbf{View 1}} & \multicolumn{1}{c}{\textbf{View 2}} \\

\vtop{\vskip0pt
\resizebox{0.5\textwidth}{!}{
\begin{tabular}{@{}c@{}c@{}c@{}c@{}c@{}}
\toprule
Input & GT  & PartSLIP & PartSTAD & Ours \\ \midrule
\includegraphics[width=0.1\textwidth]{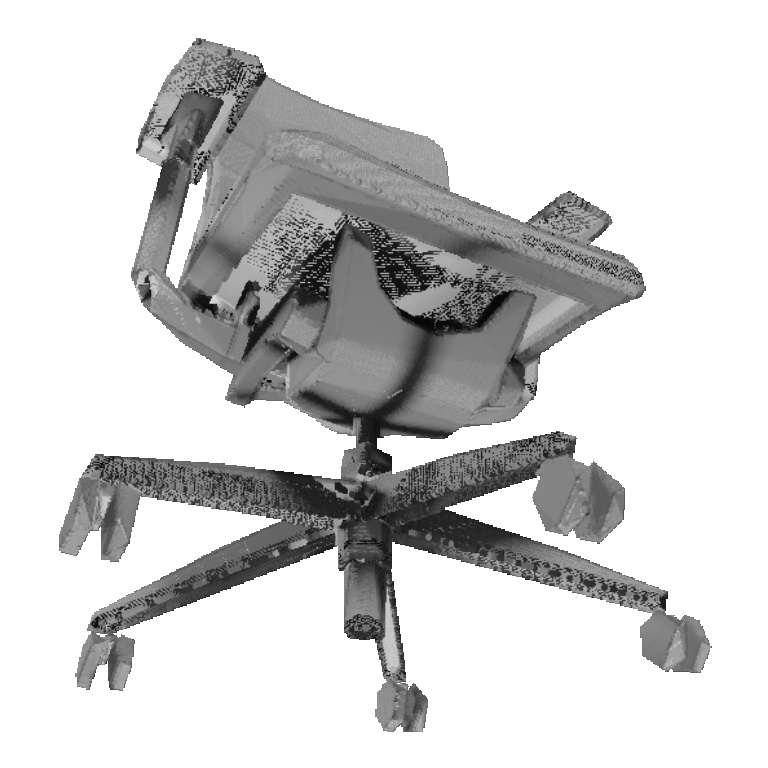} &
\includegraphics[width=0.1\textwidth]{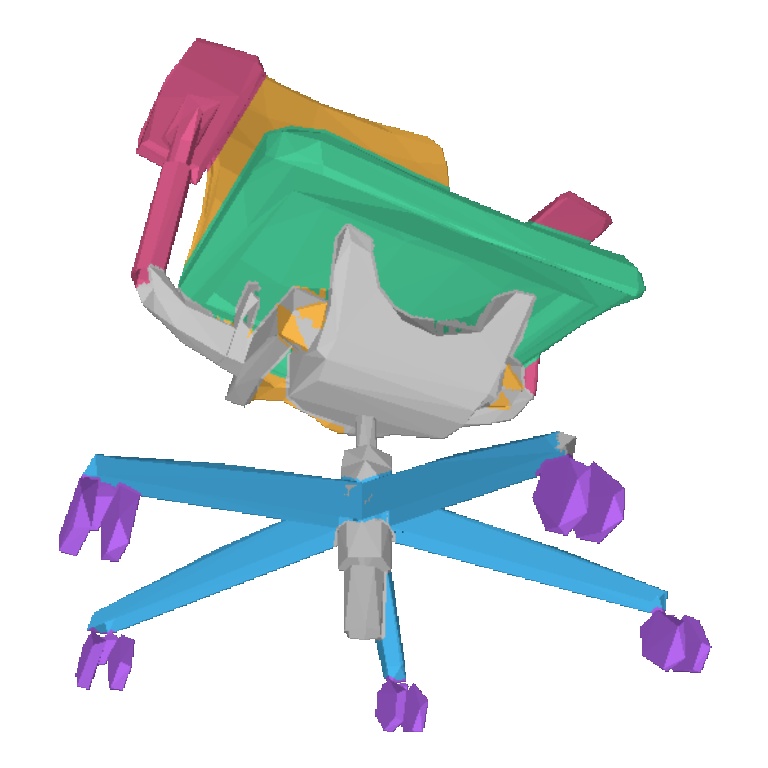} &
\includegraphics[width=0.1\textwidth]{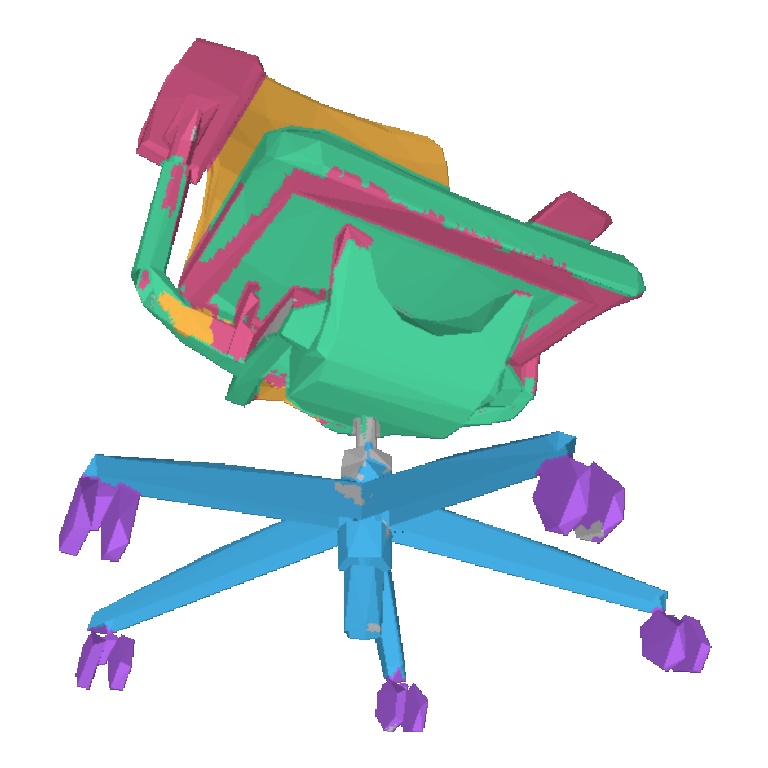} &
\includegraphics[width=0.1\textwidth]{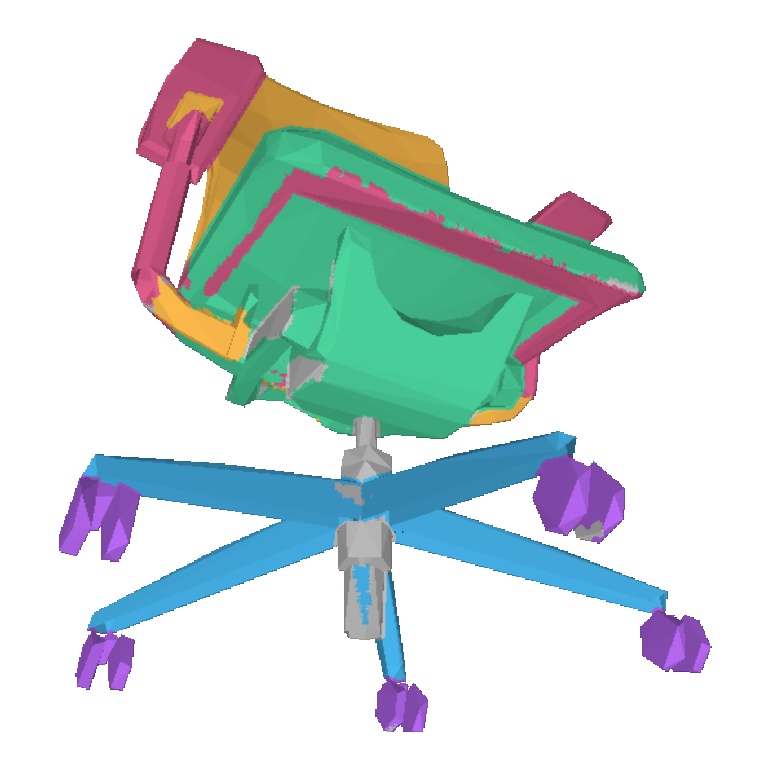} &
\includegraphics[width=0.1\textwidth]{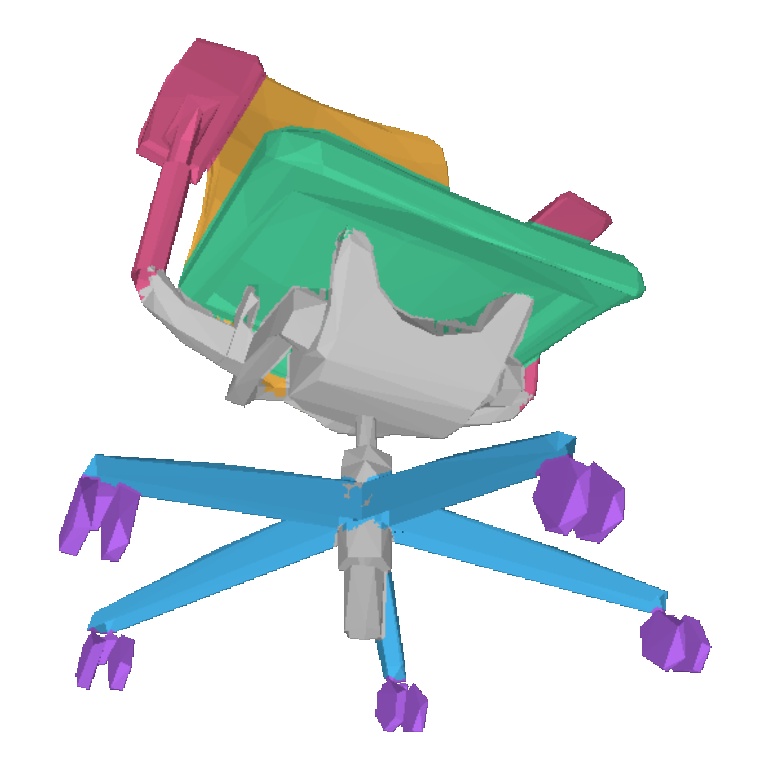} \\

\addlinespace[-2pt]
\arrayrulecolor{gray}\cmidrule(lr){1-5}
\arrayrulecolor{black}

\includegraphics[width=0.1\textwidth]{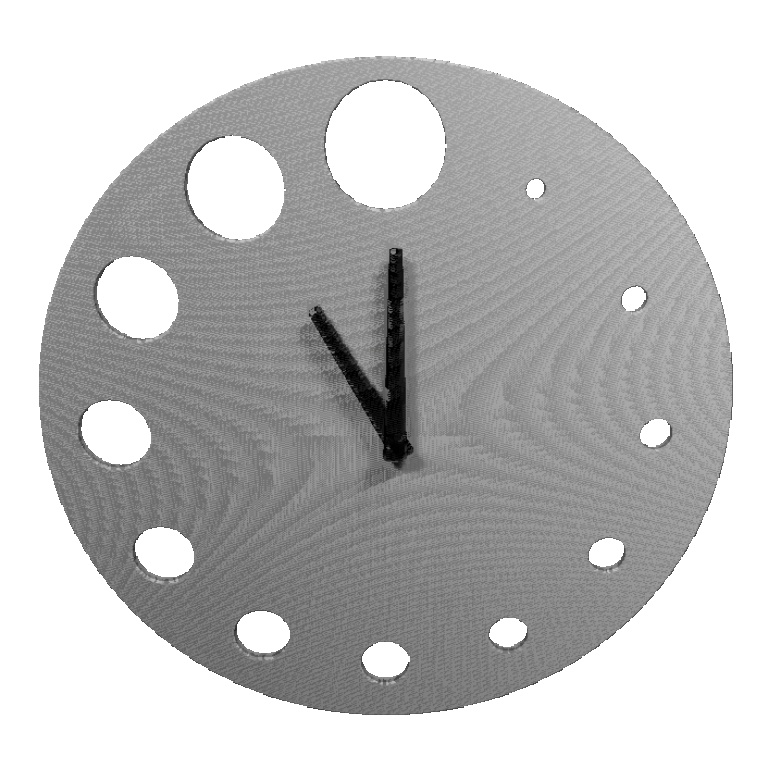} &
\includegraphics[width=0.1\textwidth]{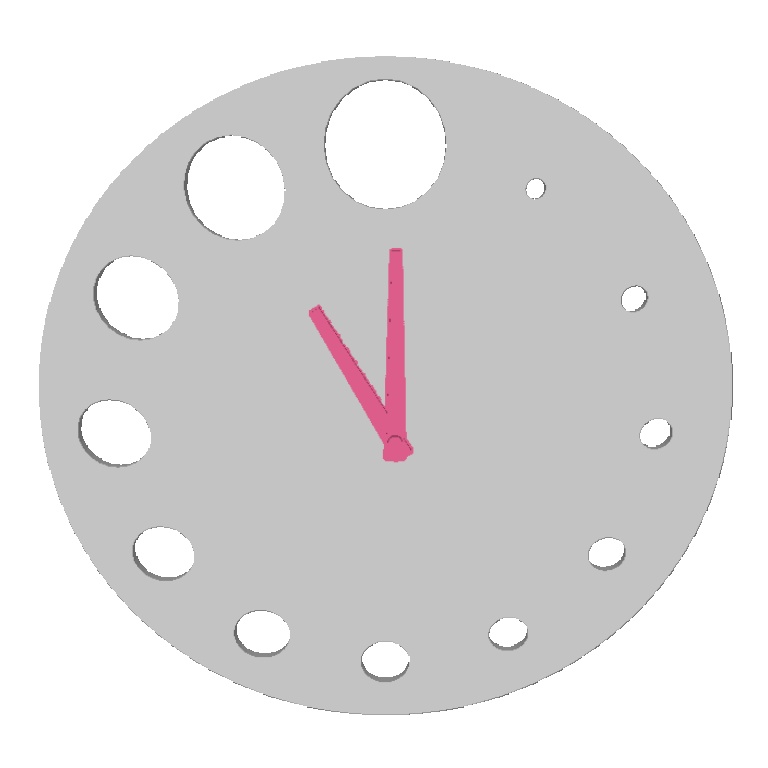} &
\includegraphics[width=0.1\textwidth]{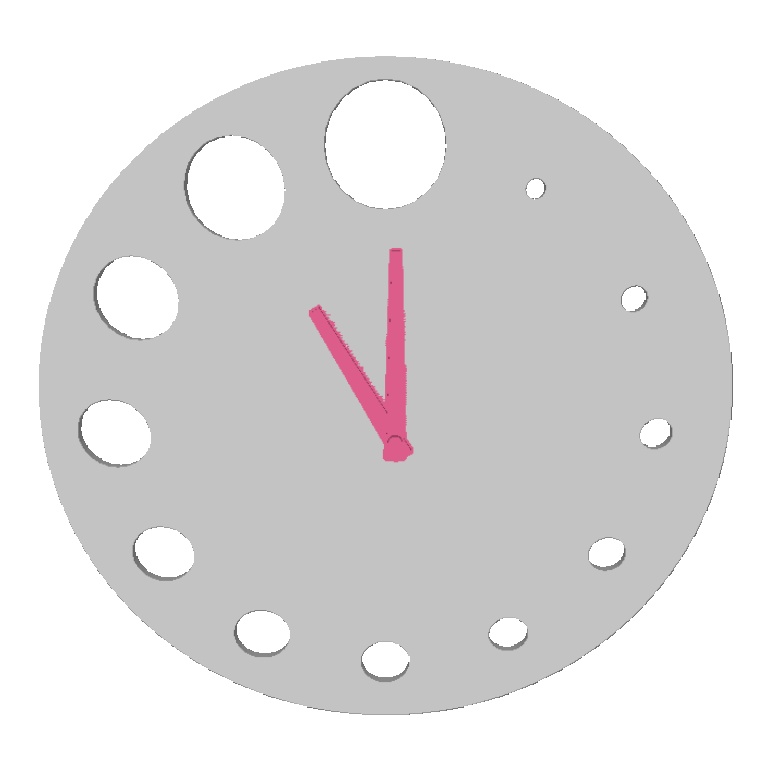} &
\includegraphics[width=0.1\textwidth]{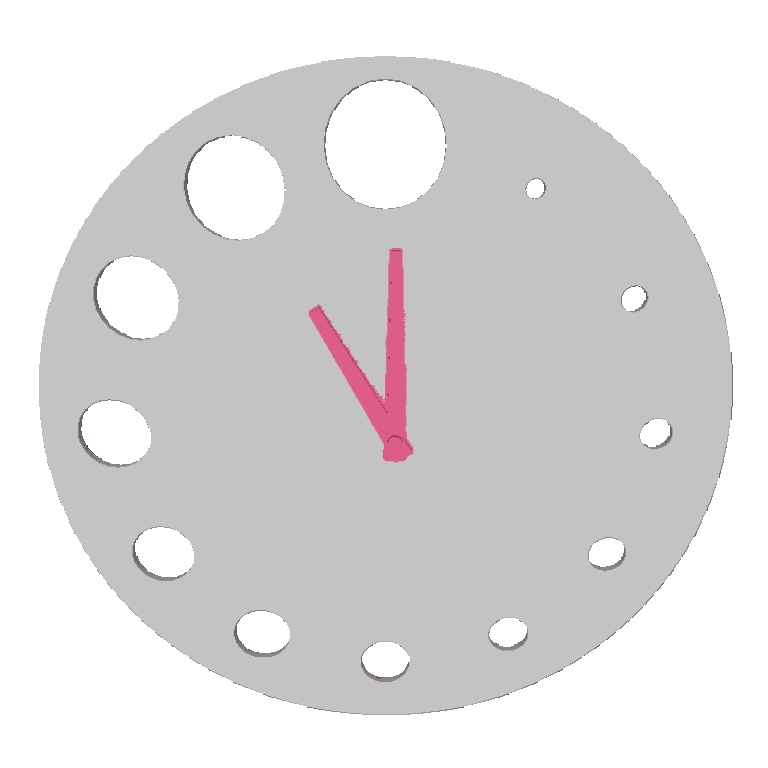} &
\includegraphics[width=0.1\textwidth]{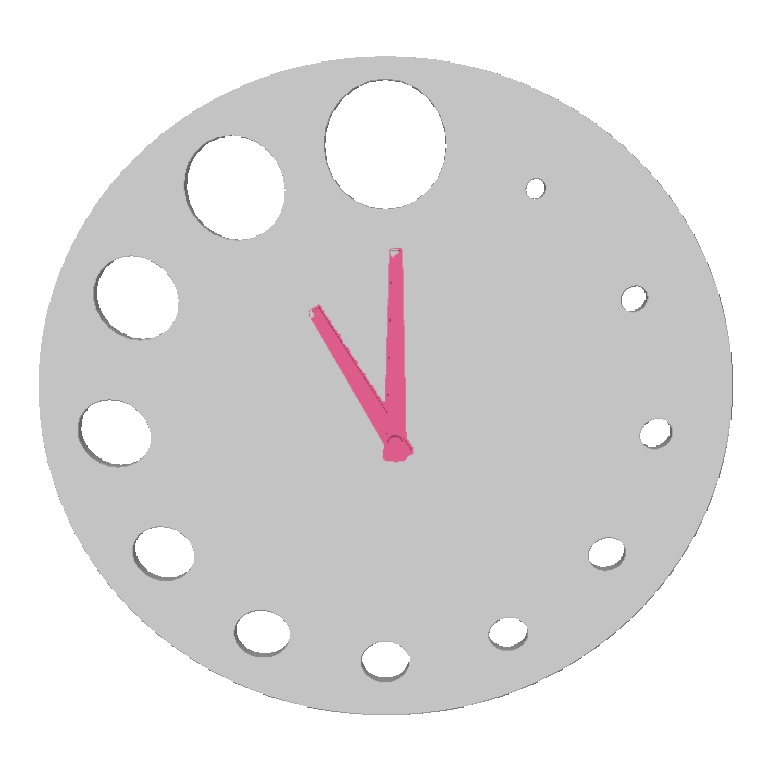} \\

\addlinespace[-2pt]
\arrayrulecolor{gray}\cmidrule(lr){1-5}
\arrayrulecolor{black}

\includegraphics[width=0.1\textwidth]{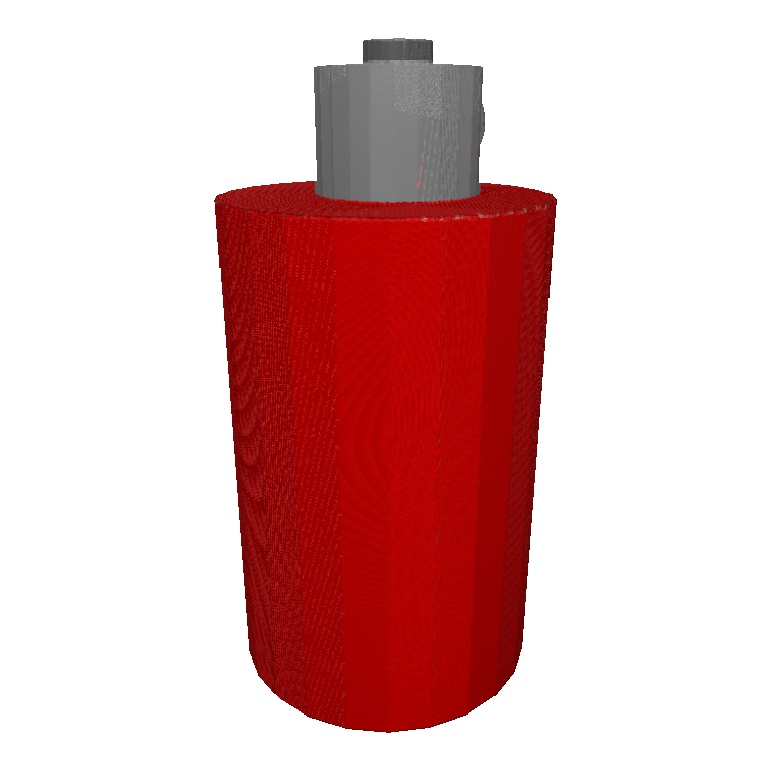} &
\includegraphics[width=0.1\textwidth]{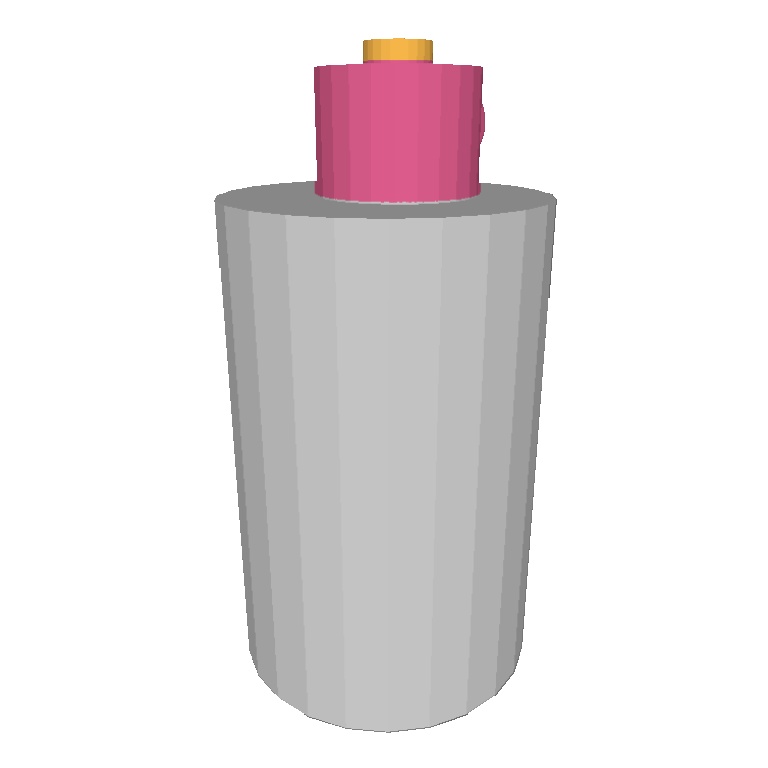} &
\includegraphics[width=0.1\textwidth]{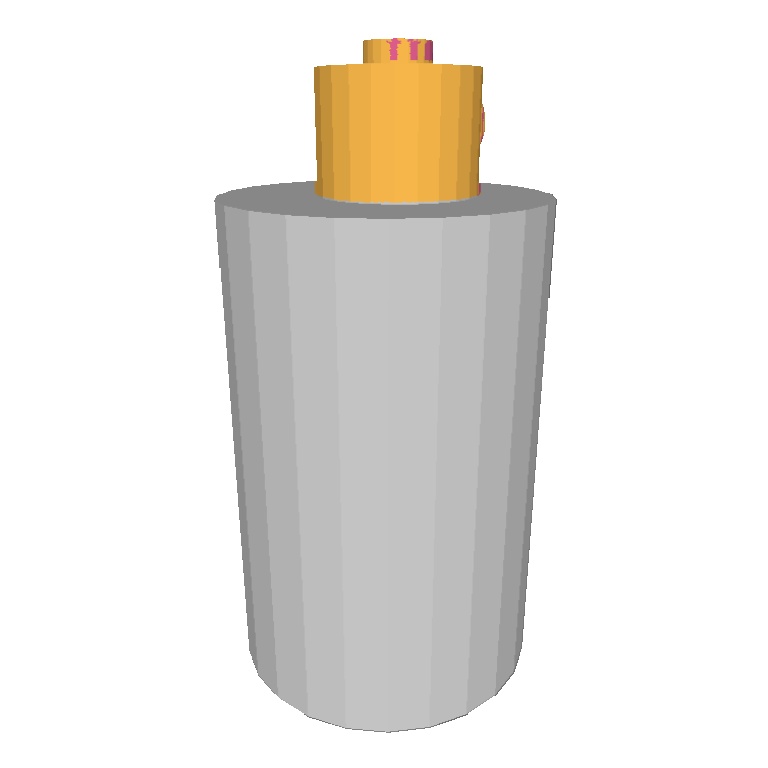} &
\includegraphics[width=0.1\textwidth]{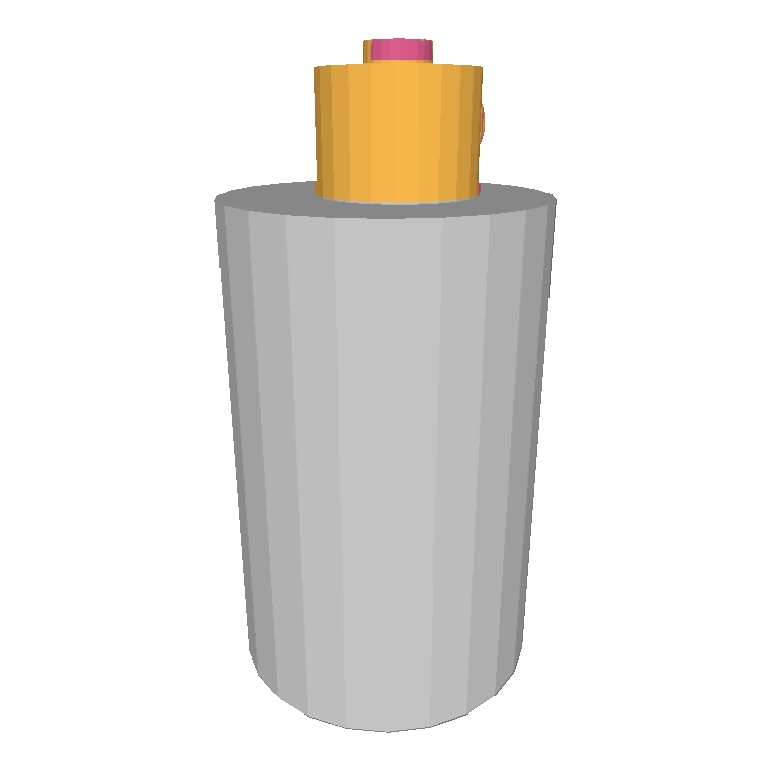} &
\includegraphics[width=0.1\textwidth]{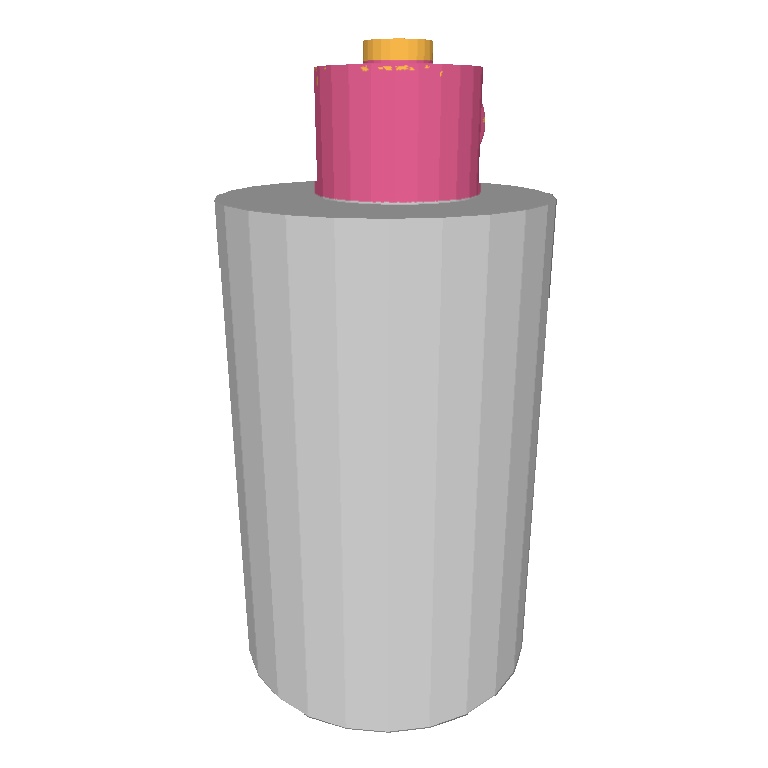} \\

\addlinespace[-2pt]
\arrayrulecolor{gray}\cmidrule(lr){1-5}
\arrayrulecolor{black}

\includegraphics[width=0.1\textwidth]{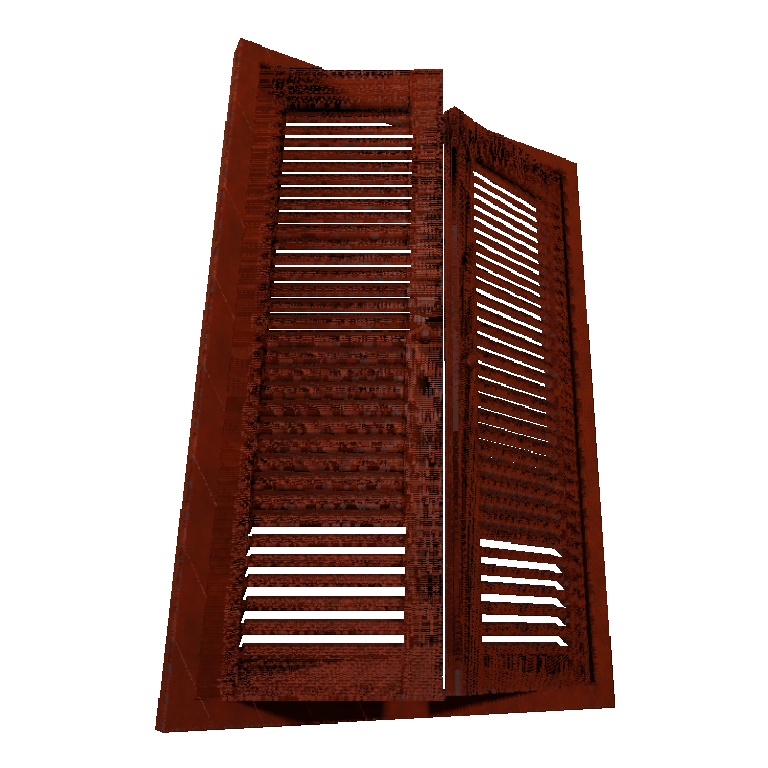} &
\includegraphics[width=0.1\textwidth]{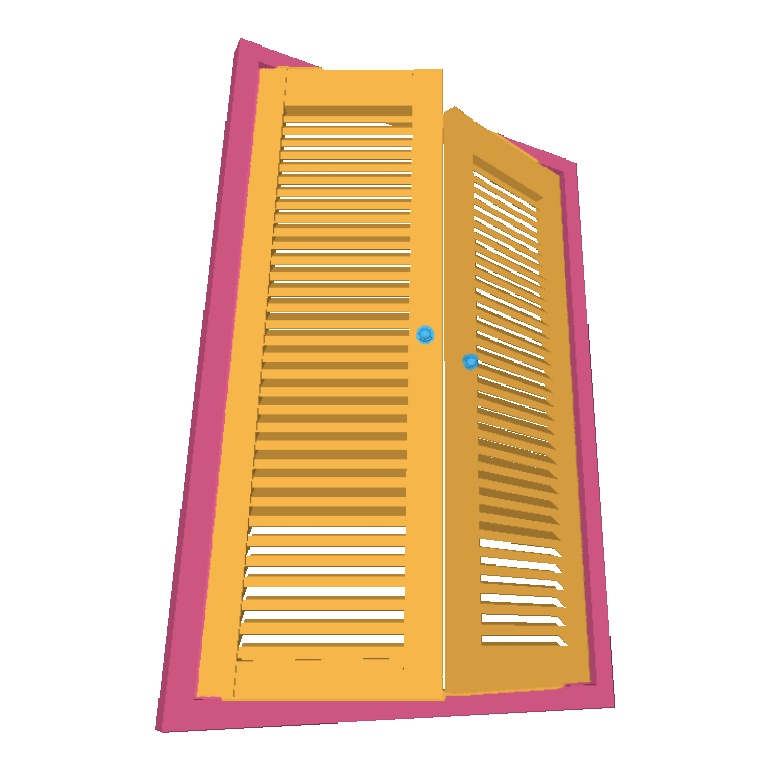} &
\includegraphics[width=0.1\textwidth]{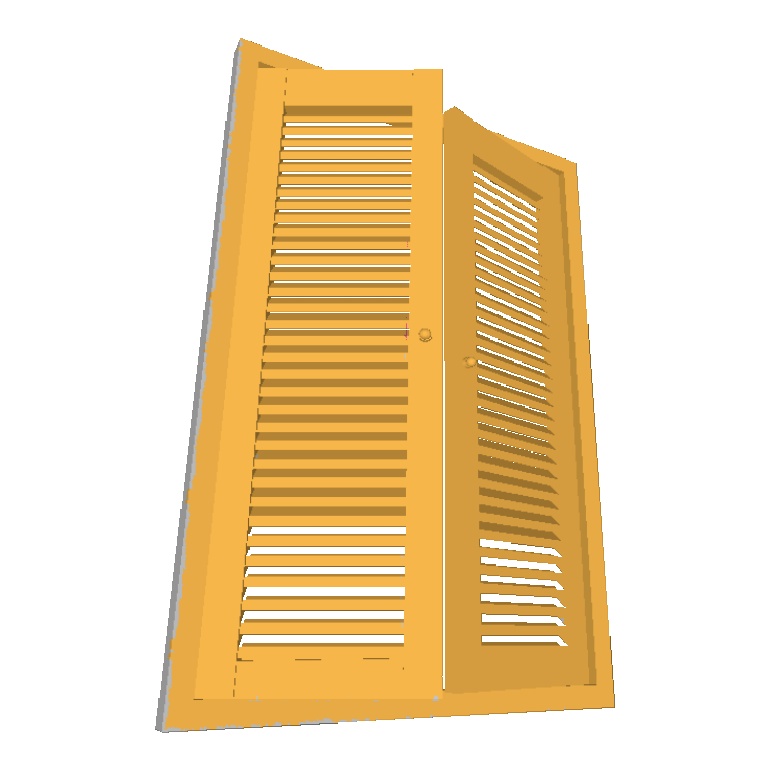} &
\includegraphics[width=0.1\textwidth]{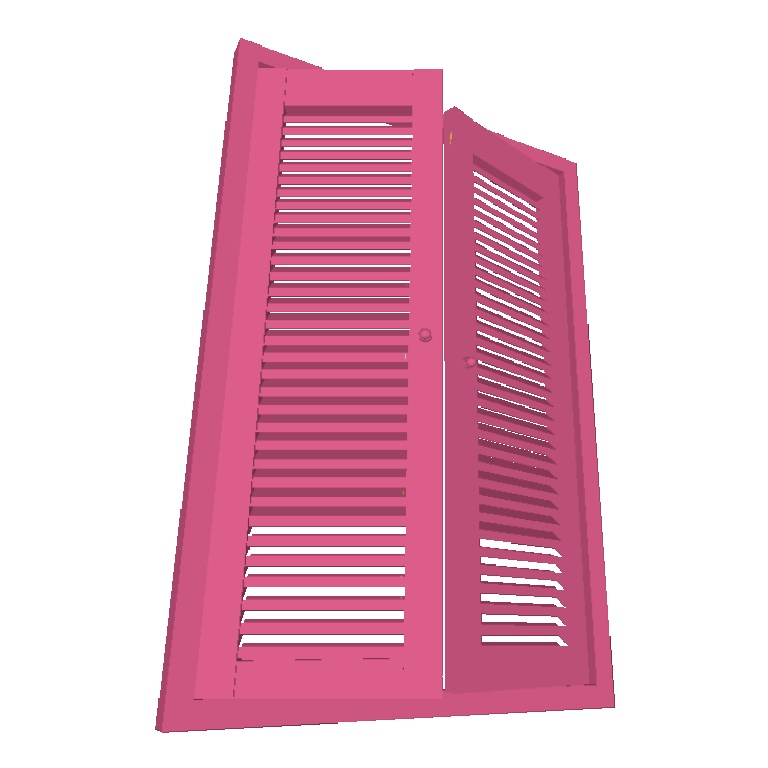} &
\includegraphics[width=0.1\textwidth]{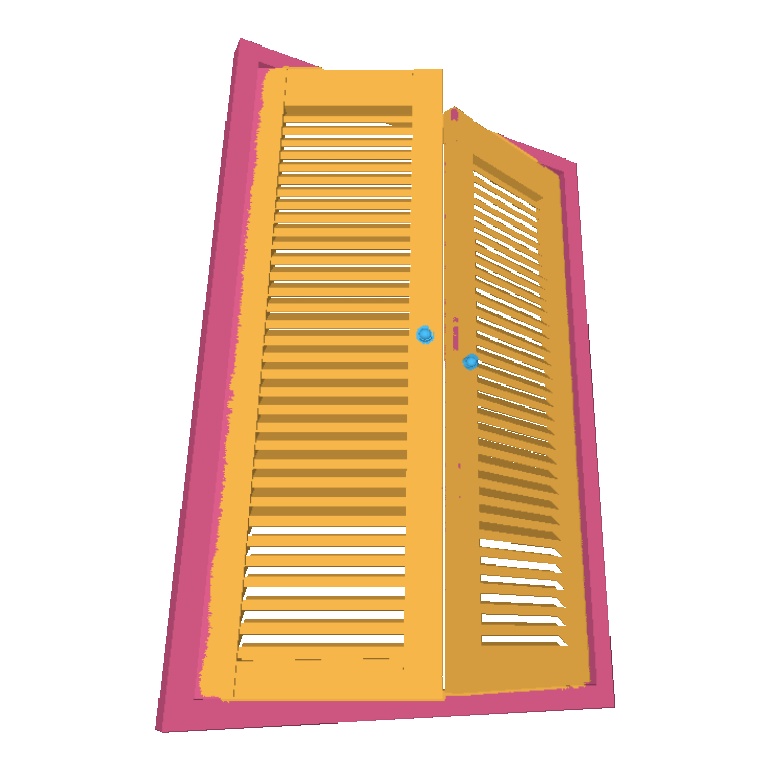} \\

\addlinespace[-2pt]
\arrayrulecolor{gray}\cmidrule(lr){1-5}
\arrayrulecolor{black}

\includegraphics[width=0.1\textwidth]{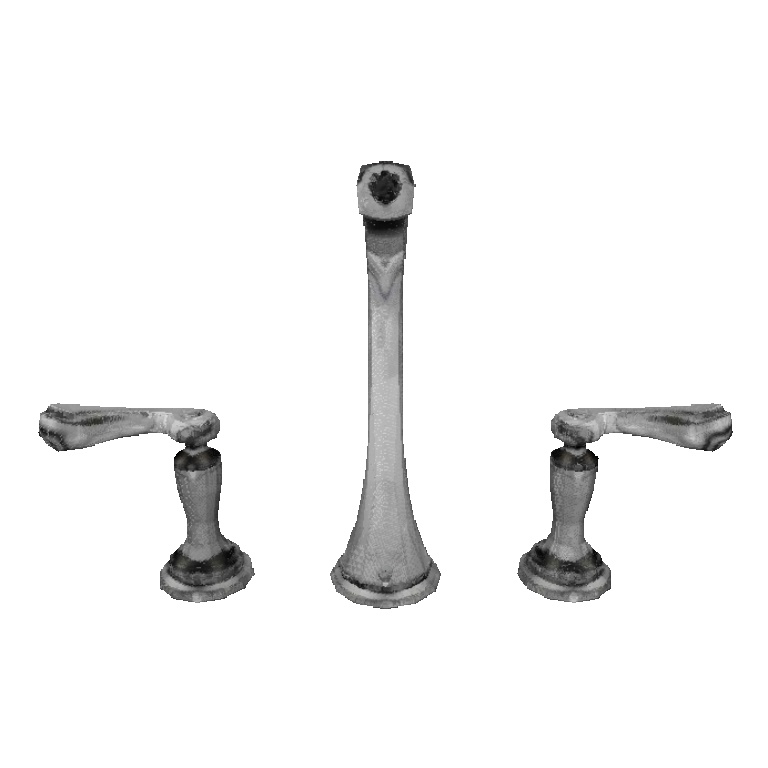} &
\includegraphics[width=0.1\textwidth]{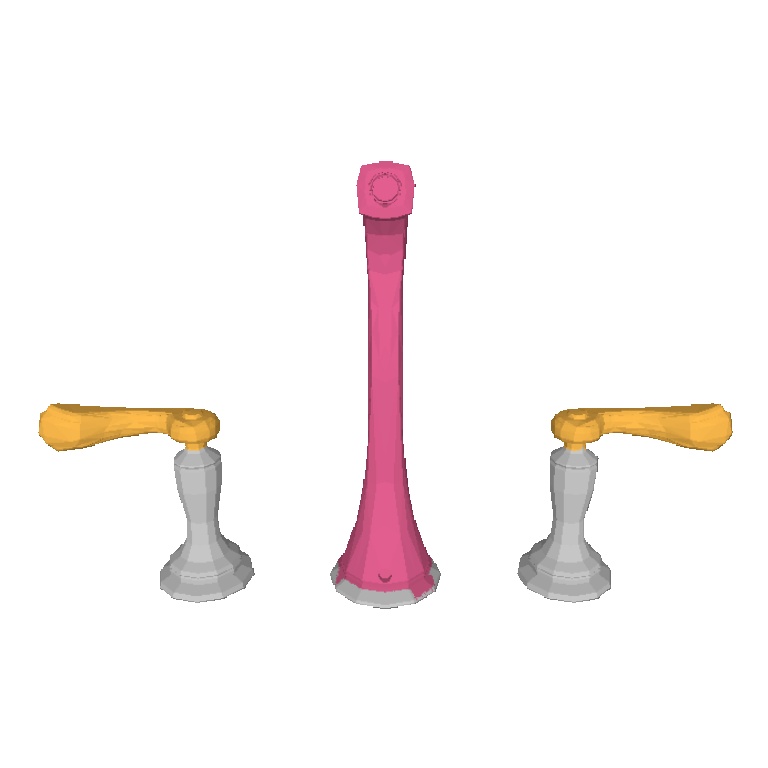} &
\includegraphics[width=0.1\textwidth]{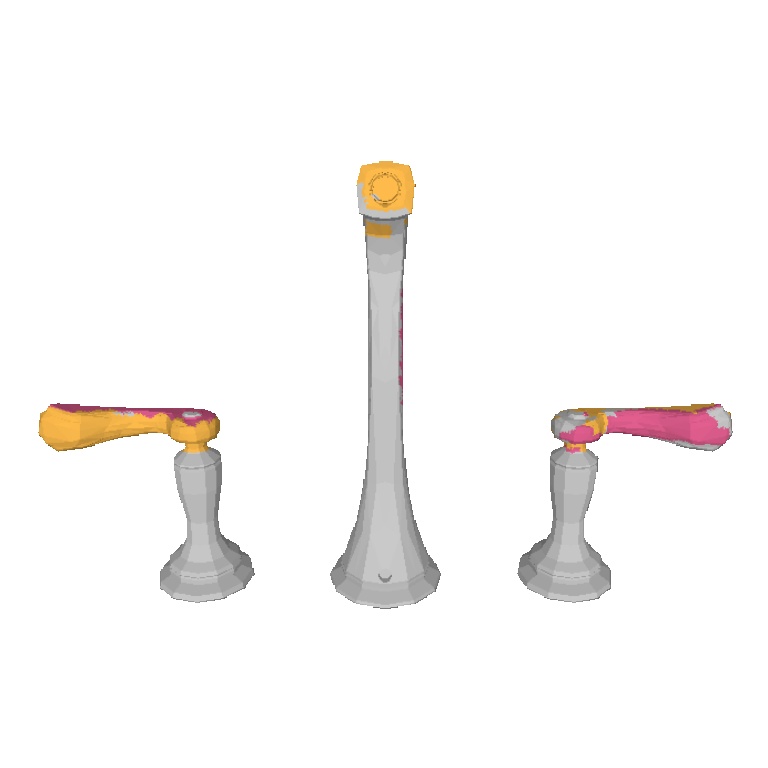} &
\includegraphics[width=0.1\textwidth]{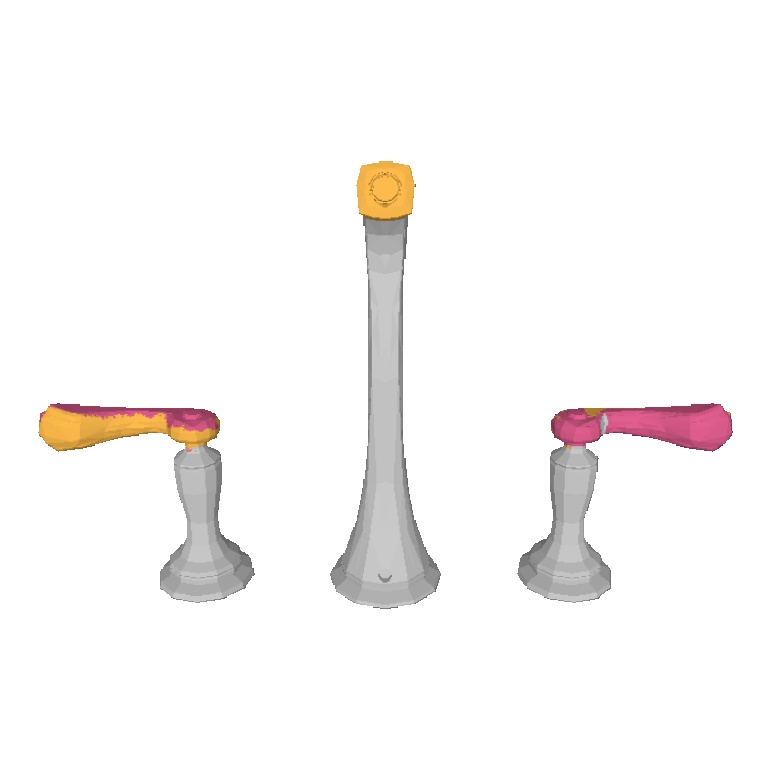} &
\includegraphics[width=0.1\textwidth]{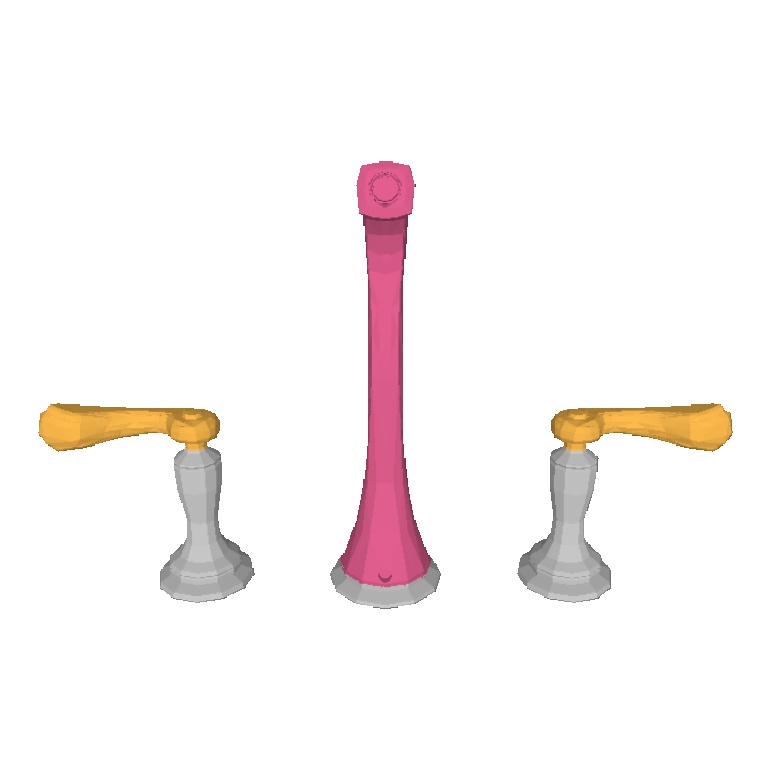} \\

\addlinespace[-2pt]
\arrayrulecolor{gray}\cmidrule(lr){1-5}
\arrayrulecolor{black}

\includegraphics[width=0.1\textwidth]{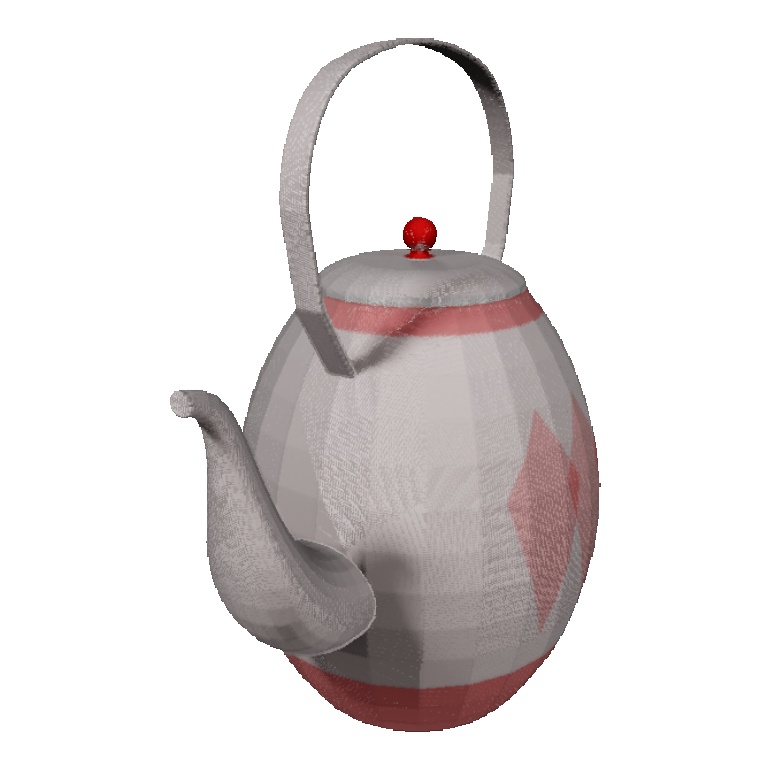} &
\includegraphics[width=0.1\textwidth]{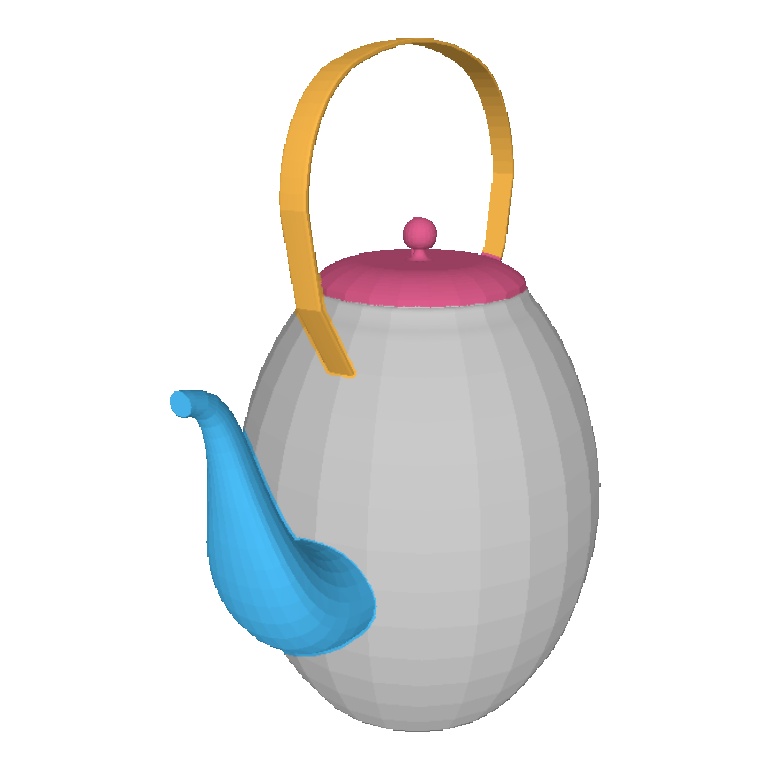} &
\includegraphics[width=0.1\textwidth]{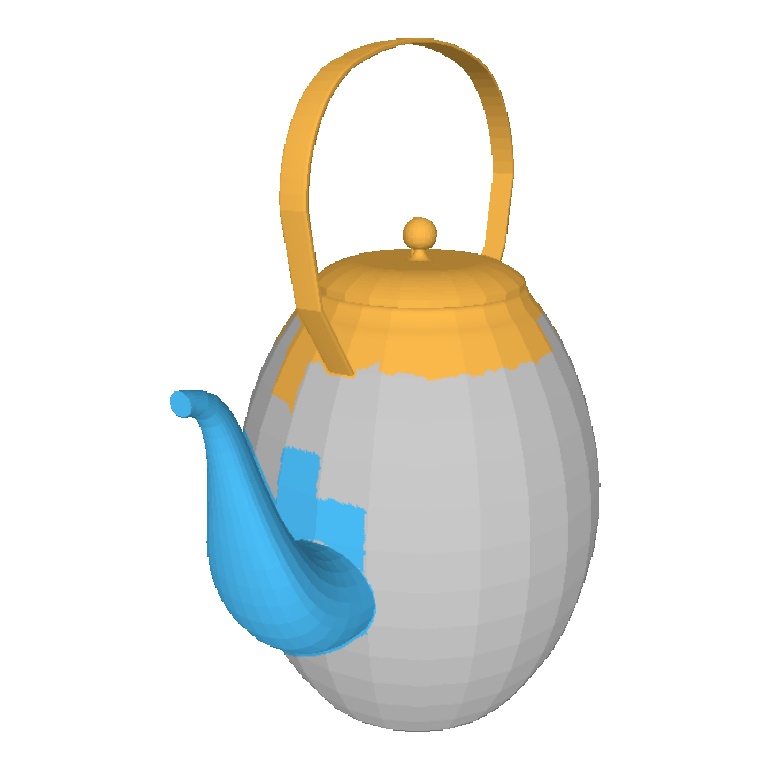} &
\includegraphics[width=0.1\textwidth]{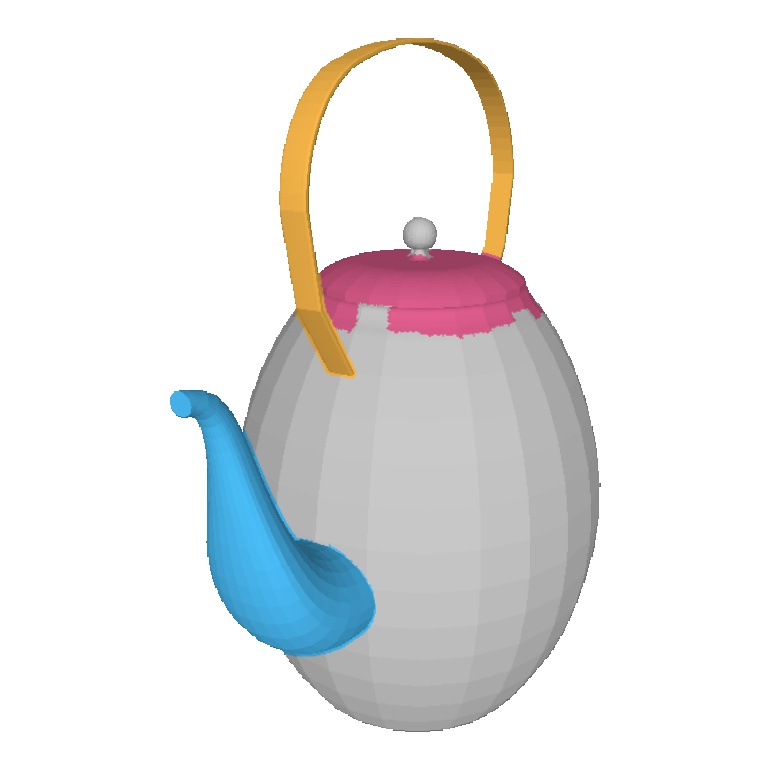} &
\includegraphics[width=0.1\textwidth]{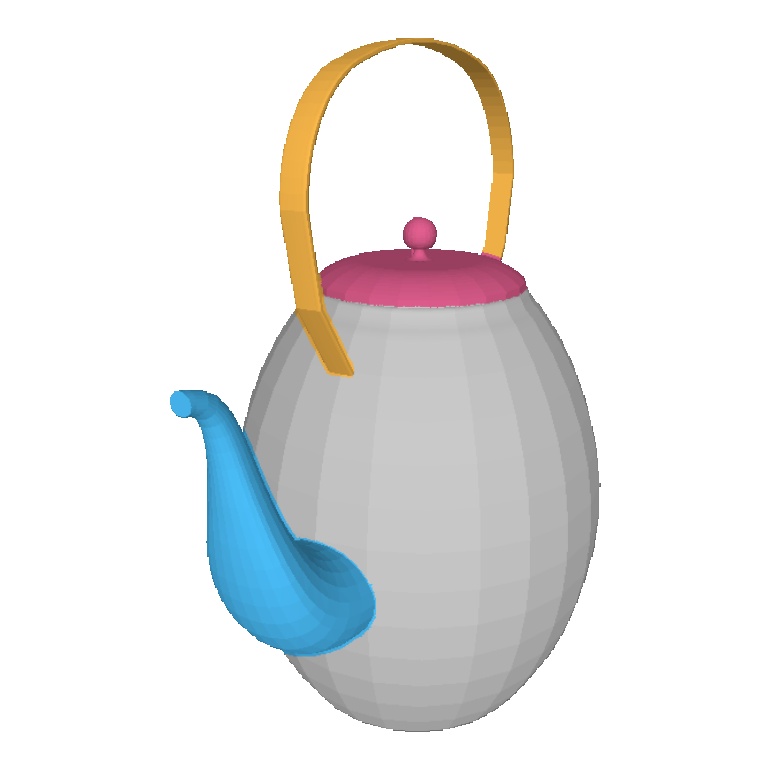} \\

\addlinespace[-2pt]
\arrayrulecolor{gray}\cmidrule(lr){1-5}
\arrayrulecolor{black}

\includegraphics[width=0.1\textwidth]{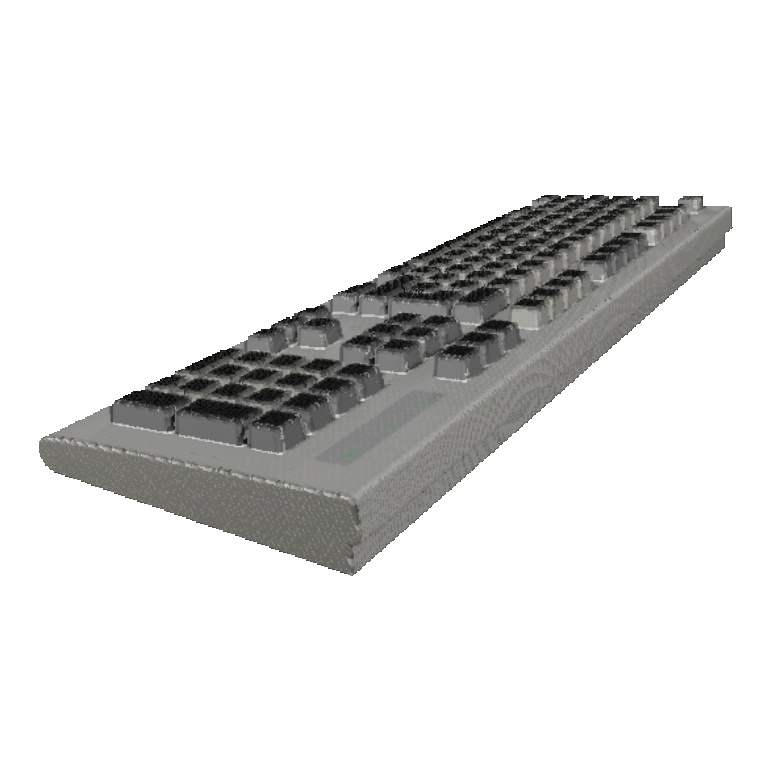} &
\includegraphics[width=0.1\textwidth]{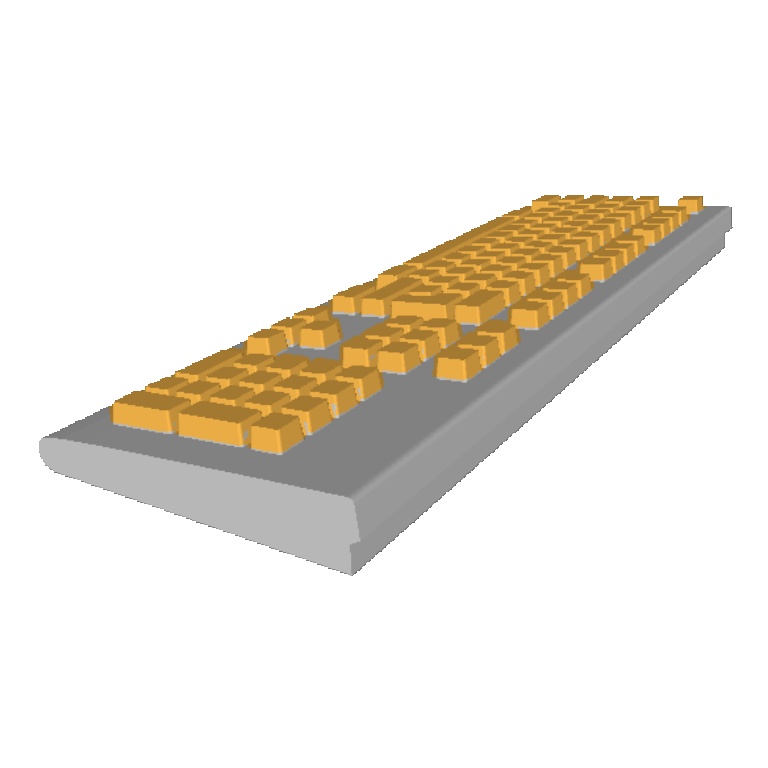} &
\includegraphics[width=0.1\textwidth]{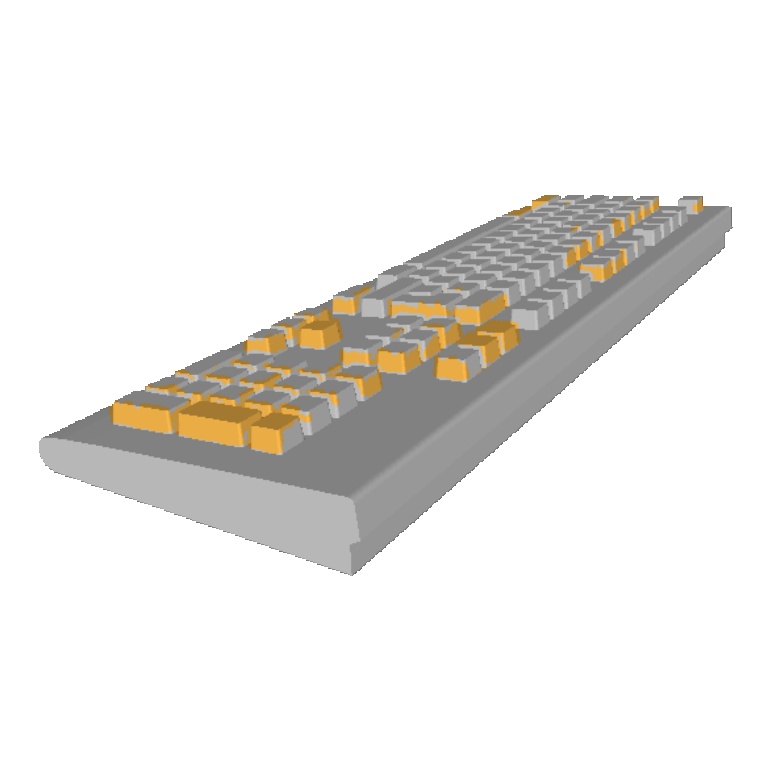} &
\includegraphics[width=0.1\textwidth]{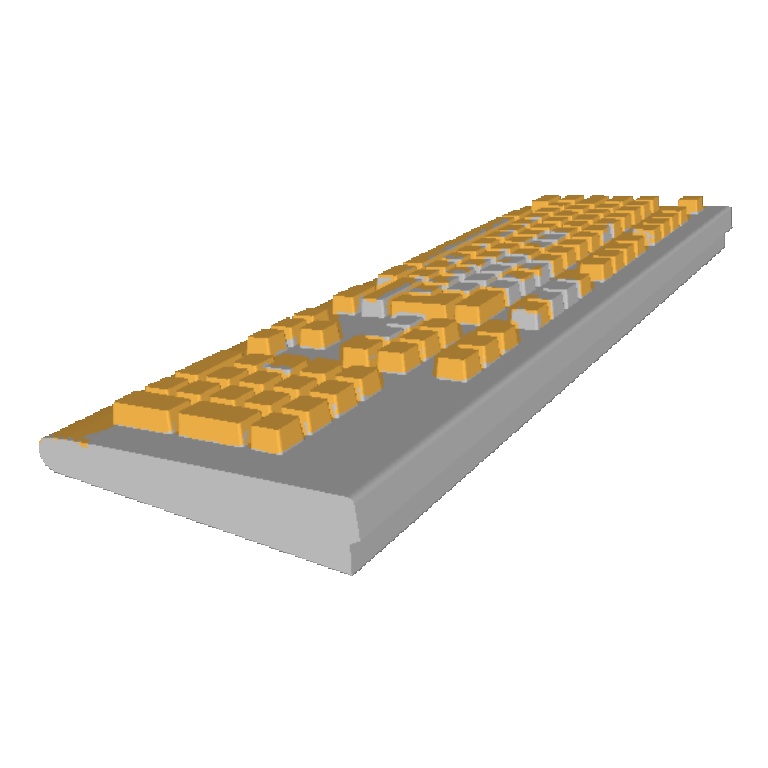} &
\includegraphics[width=0.1\textwidth]{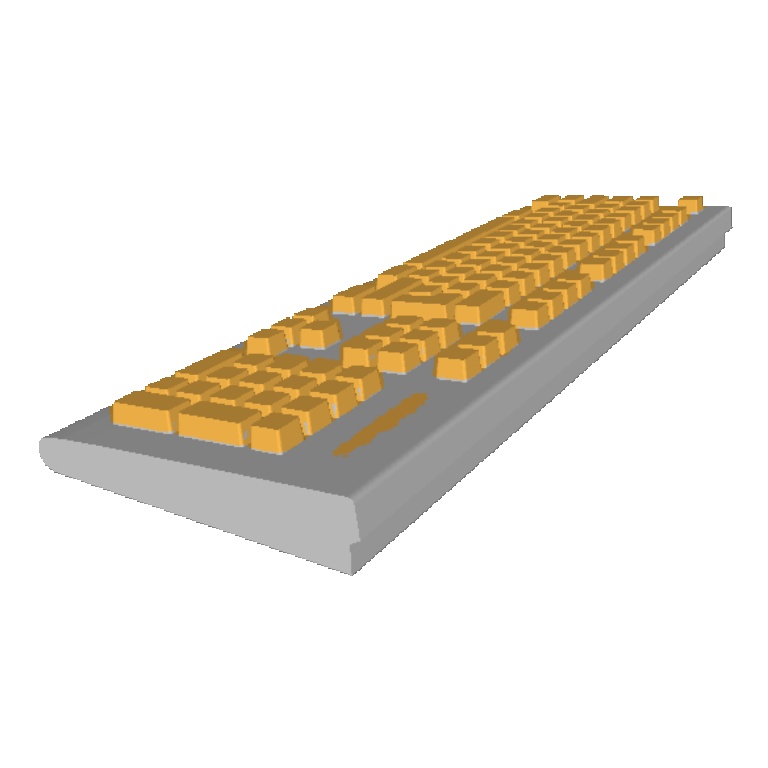} \\

\addlinespace[-2pt]
\arrayrulecolor{gray}\cmidrule(lr){1-5}
\arrayrulecolor{black}

\includegraphics[width=0.1\textwidth]{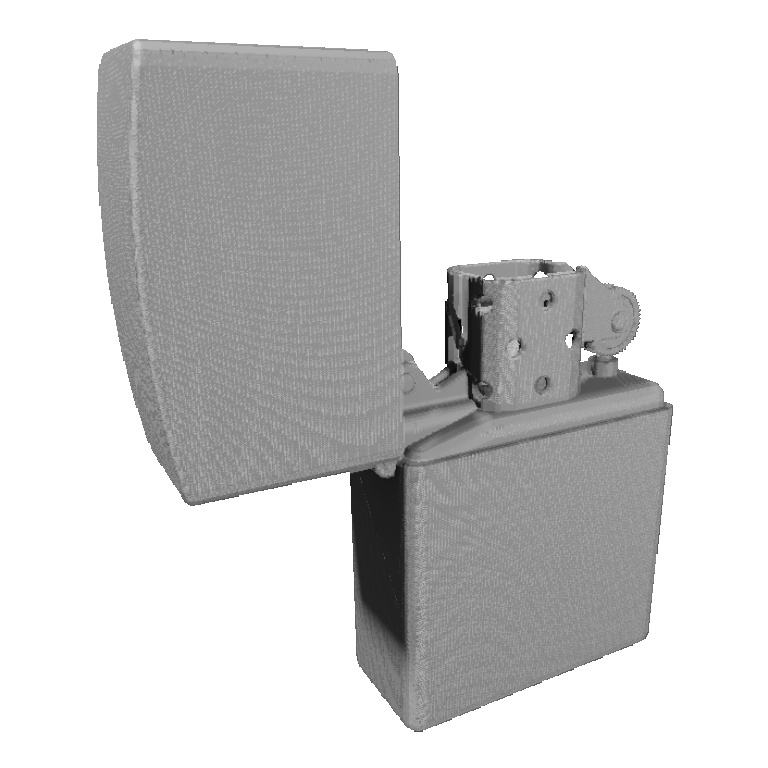} &
\includegraphics[width=0.1\textwidth]{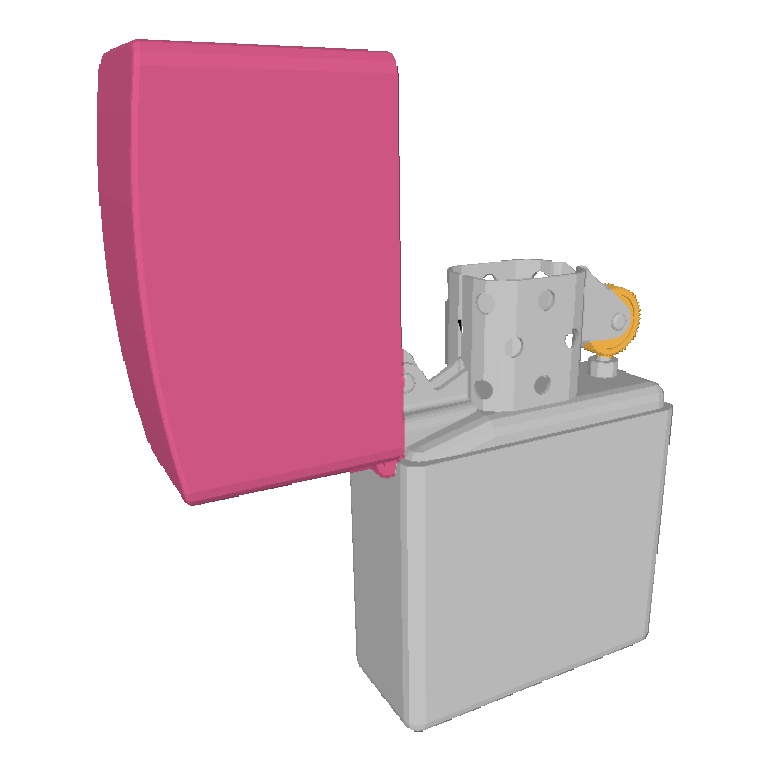} &
\includegraphics[width=0.1\textwidth]{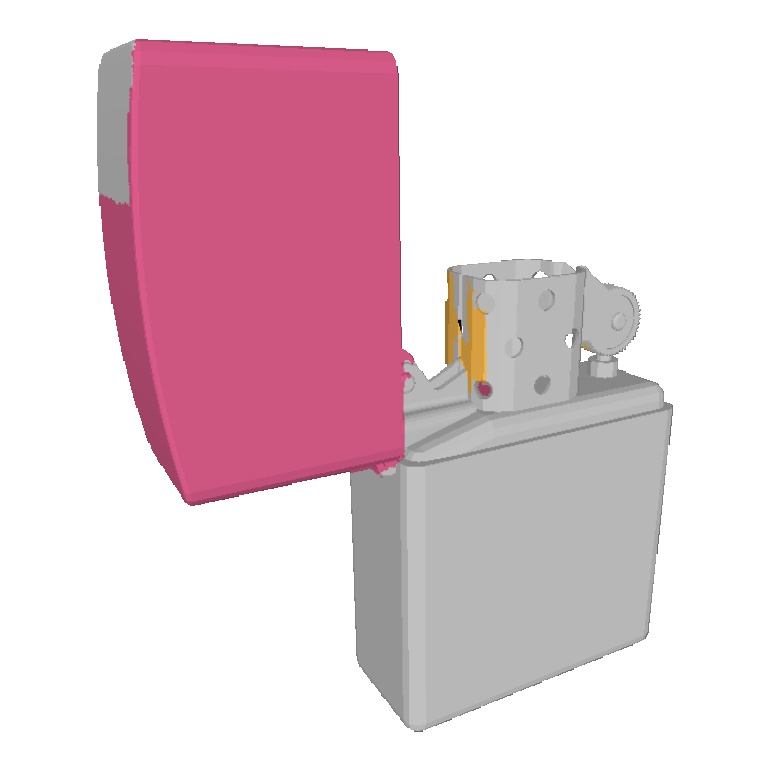} &
\includegraphics[width=0.1\textwidth]{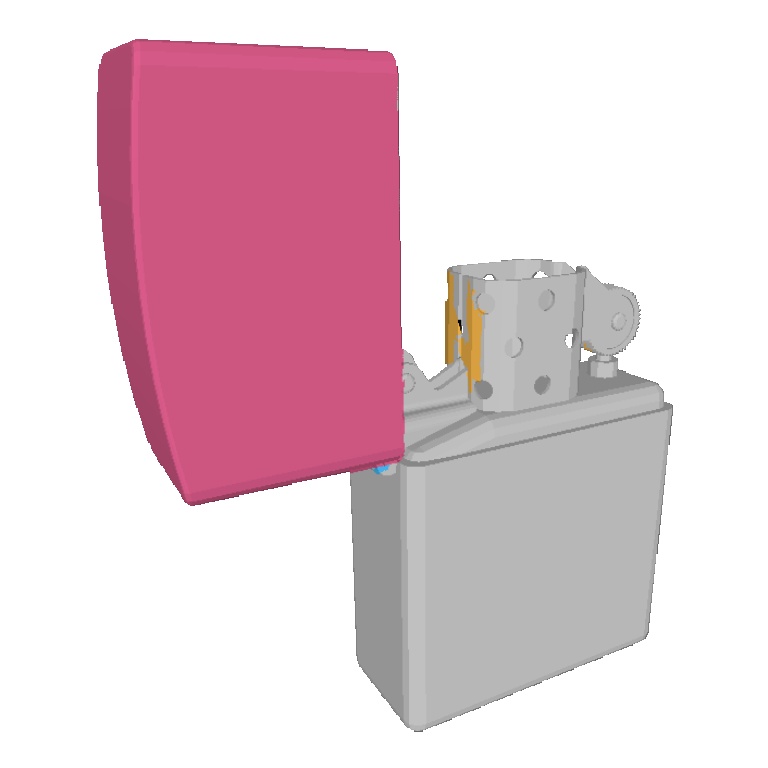} &
\includegraphics[width=0.1\textwidth]{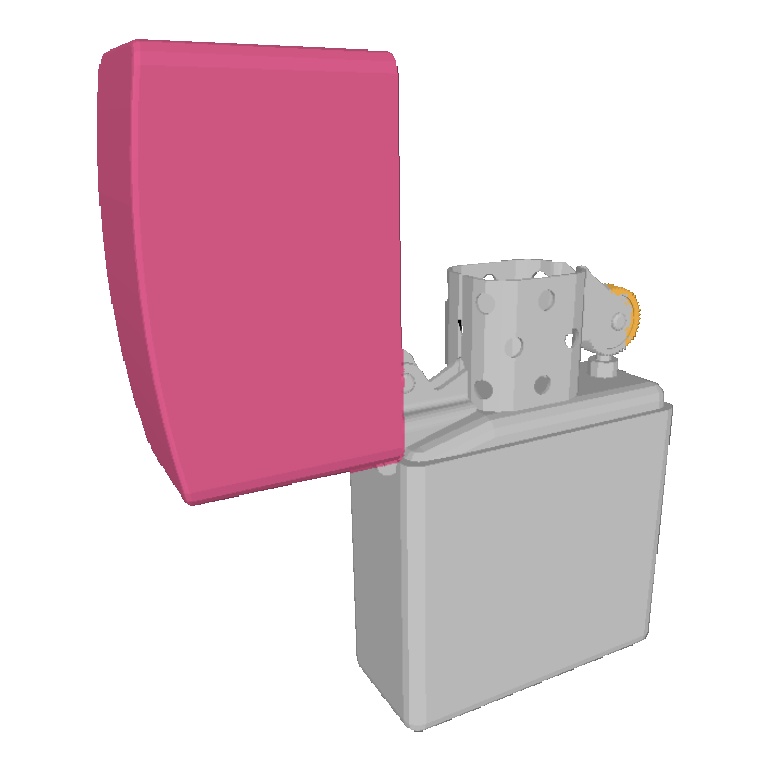} \\

\addlinespace[-2pt]
\arrayrulecolor{gray}\cmidrule(lr){1-5}
\arrayrulecolor{black}

\includegraphics[width=0.1\textwidth]{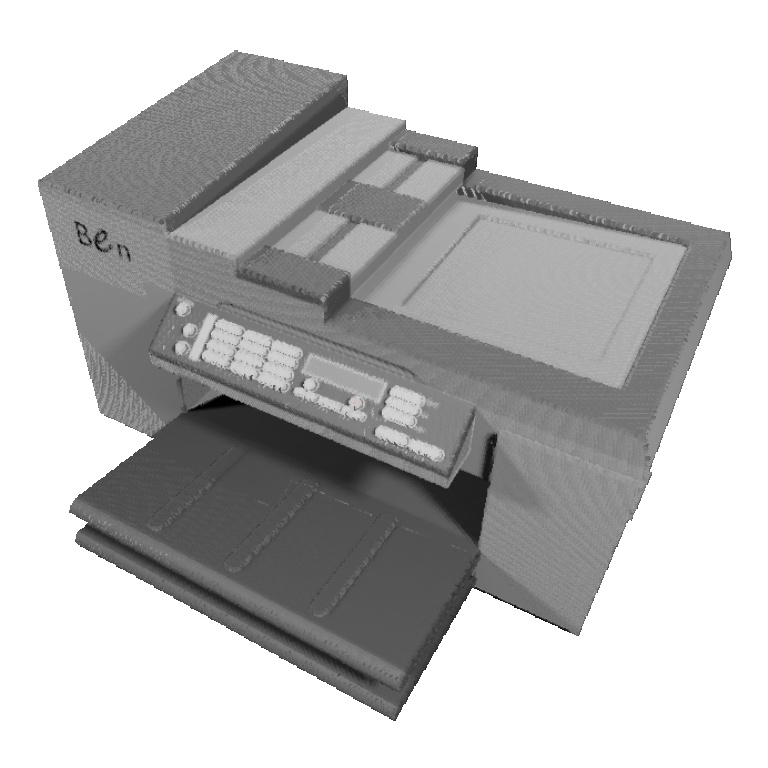} &
\includegraphics[width=0.1\textwidth]{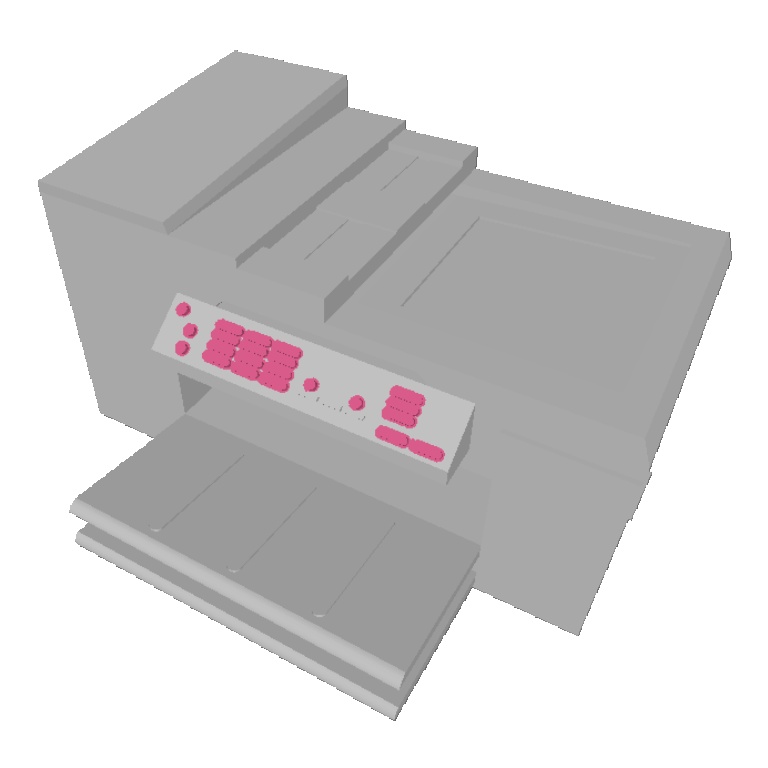} &
\includegraphics[width=0.1\textwidth]{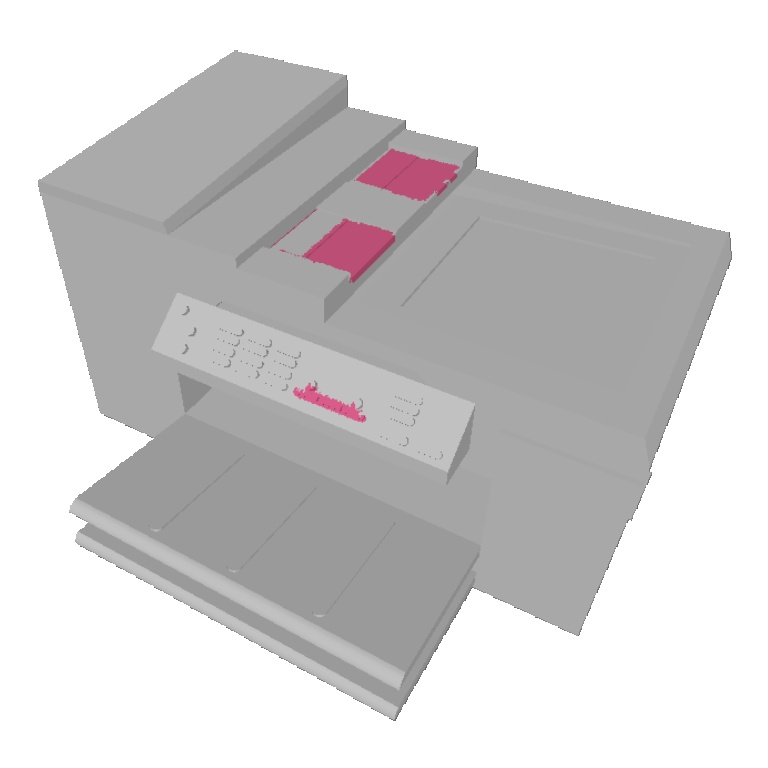} &
\includegraphics[width=0.1\textwidth]{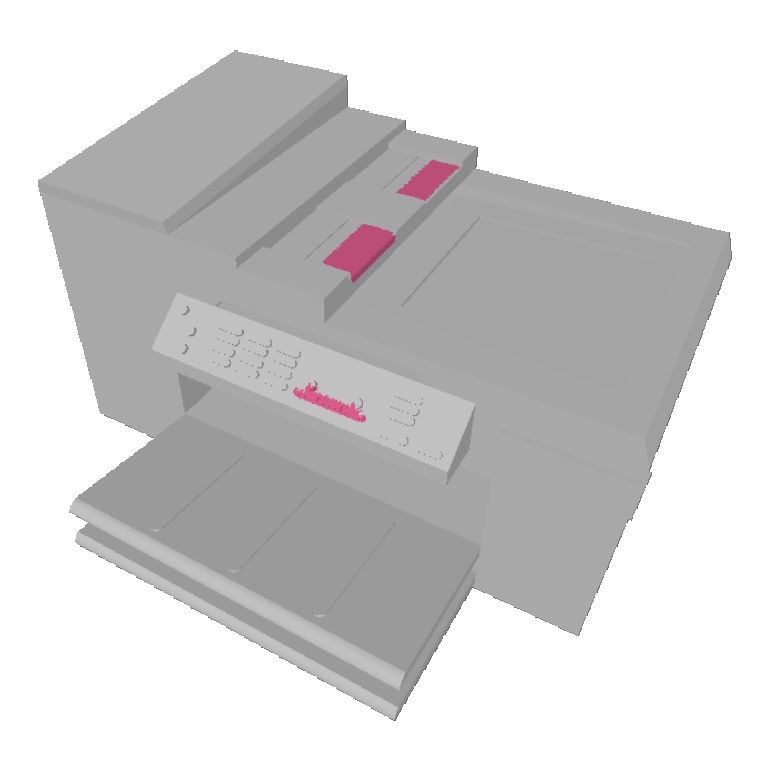} &
\includegraphics[width=0.1\textwidth]{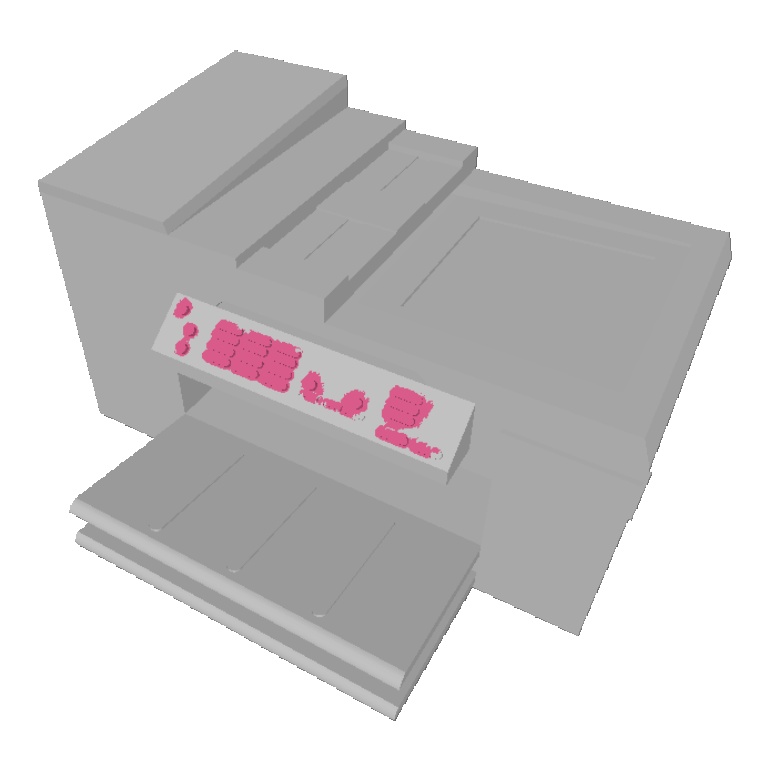} \\

\addlinespace[-2pt]
\arrayrulecolor{gray}\cmidrule(lr){1-5}
\arrayrulecolor{black}

\includegraphics[width=0.1\textwidth]{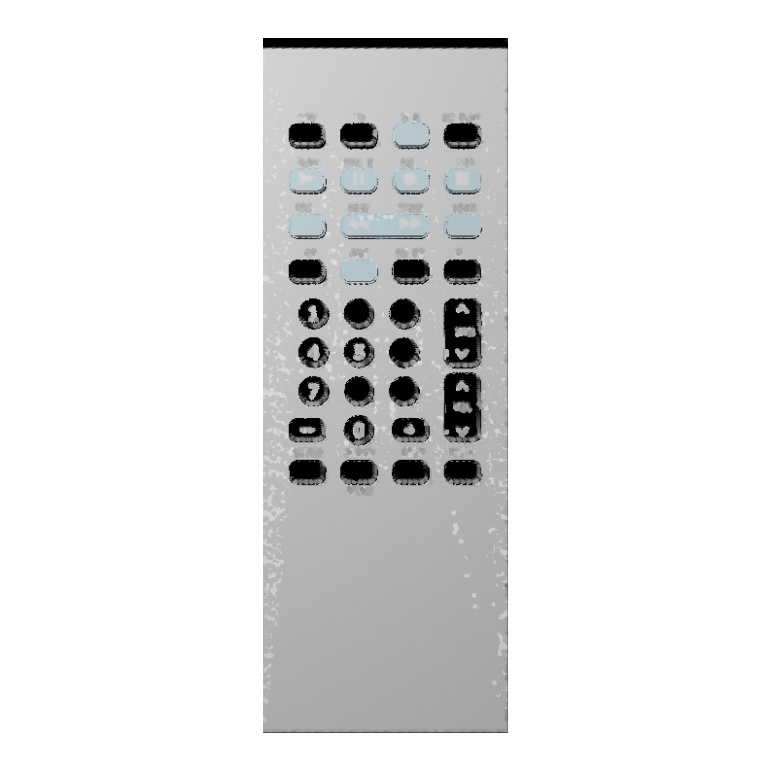} &
\includegraphics[width=0.1\textwidth]{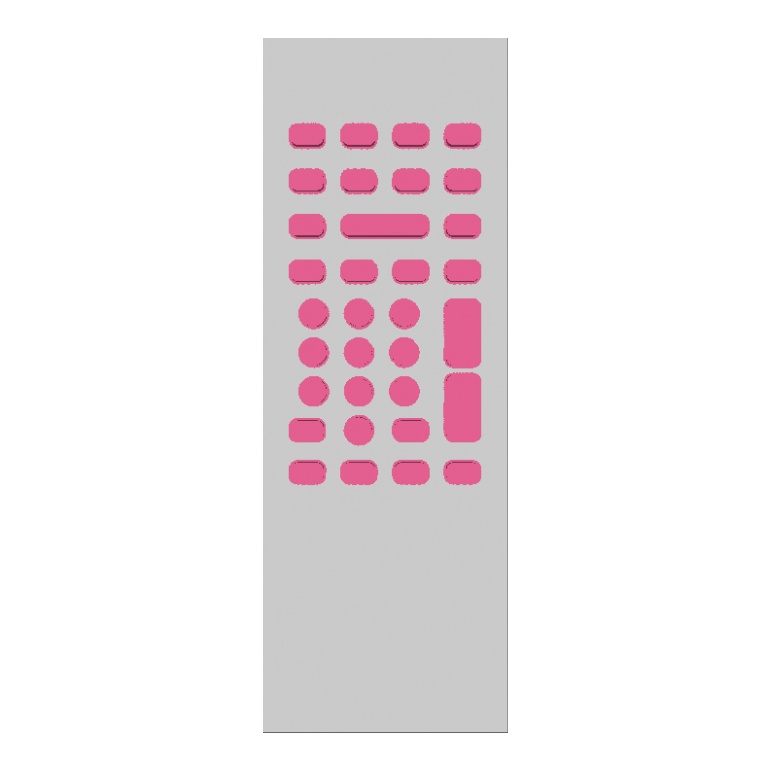} &
\includegraphics[width=0.1\textwidth]{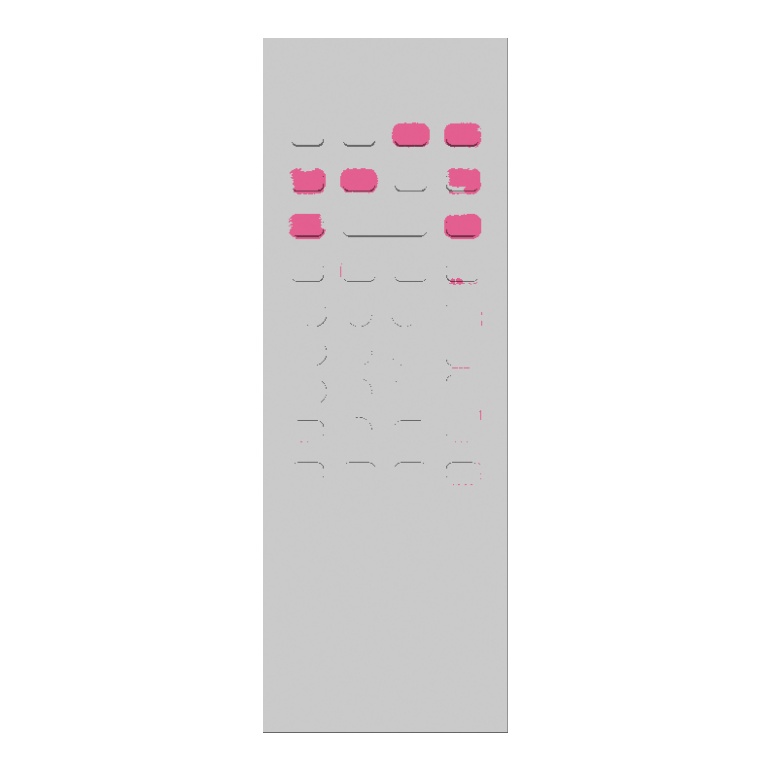} &
\includegraphics[width=0.1\textwidth]{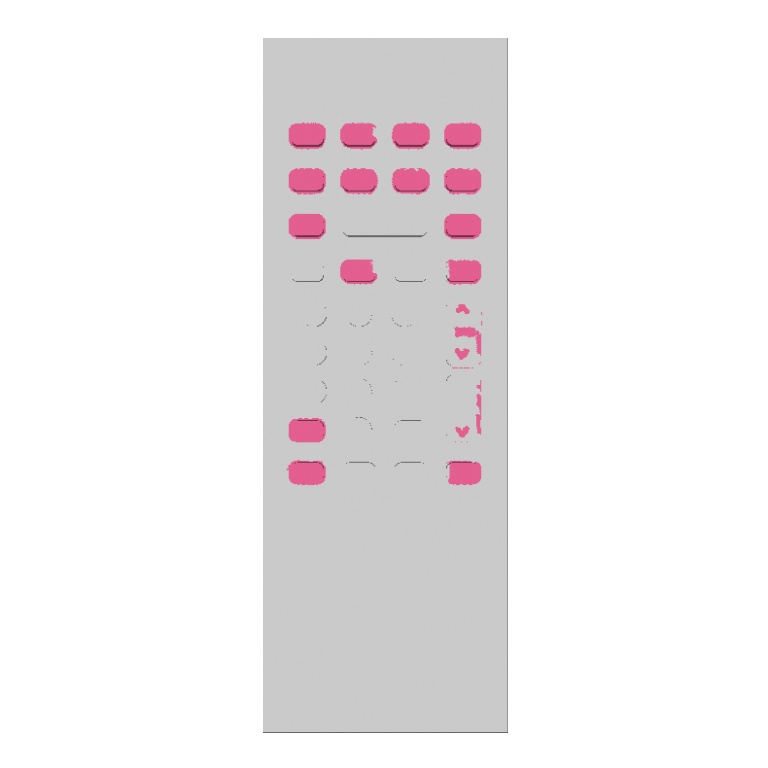} &
\includegraphics[width=0.1\textwidth]{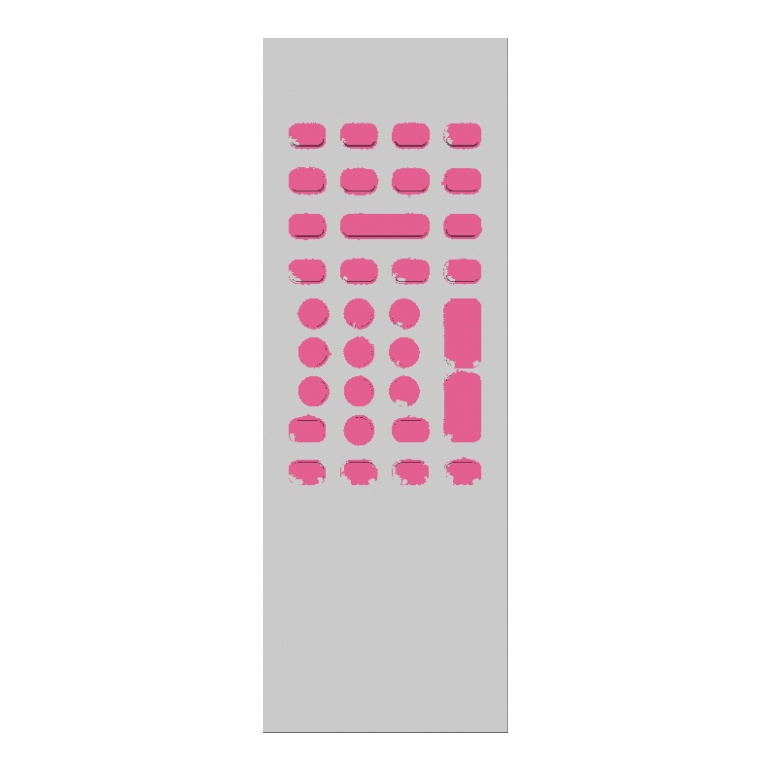} \\

\addlinespace[-2pt]
\bottomrule
\end{tabular}
}}
&
\vtop{\vskip0pt
\resizebox{0.5\textwidth}{!}{
\begin{tabular}{@{}c@{}c@{}c@{}c@{}c@{}}
\toprule
Input & GT  & PartSLIP & PartSTAD & Ours \\ \midrule
\includegraphics[width=0.1\textwidth]{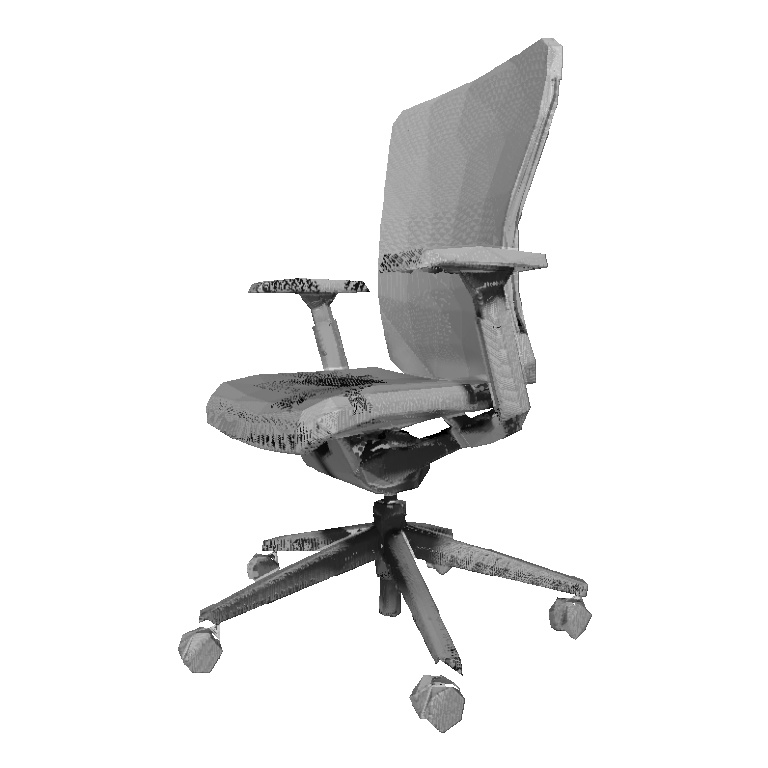} &
\includegraphics[width=0.1\textwidth]{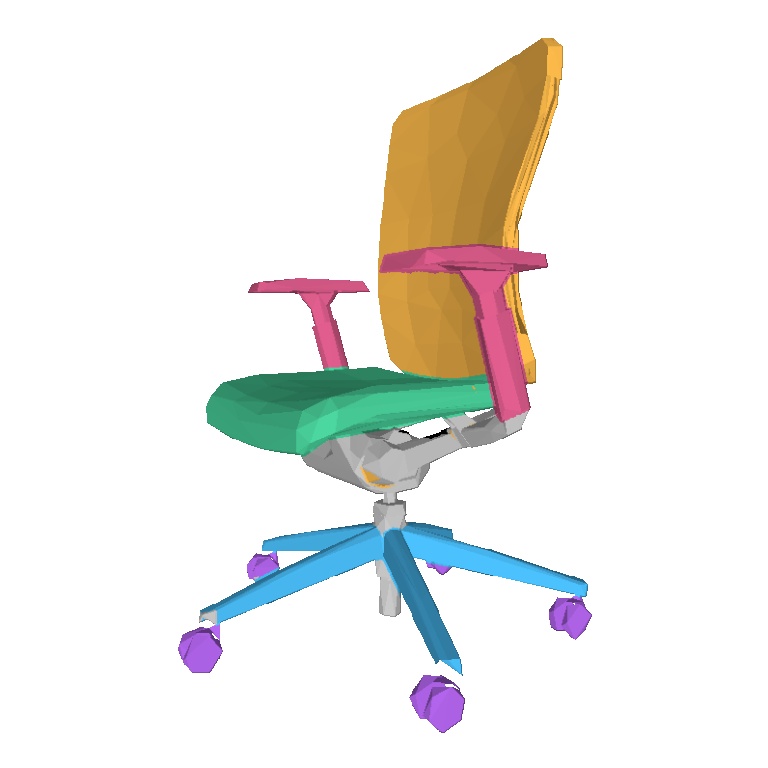} &
\includegraphics[width=0.1\textwidth]{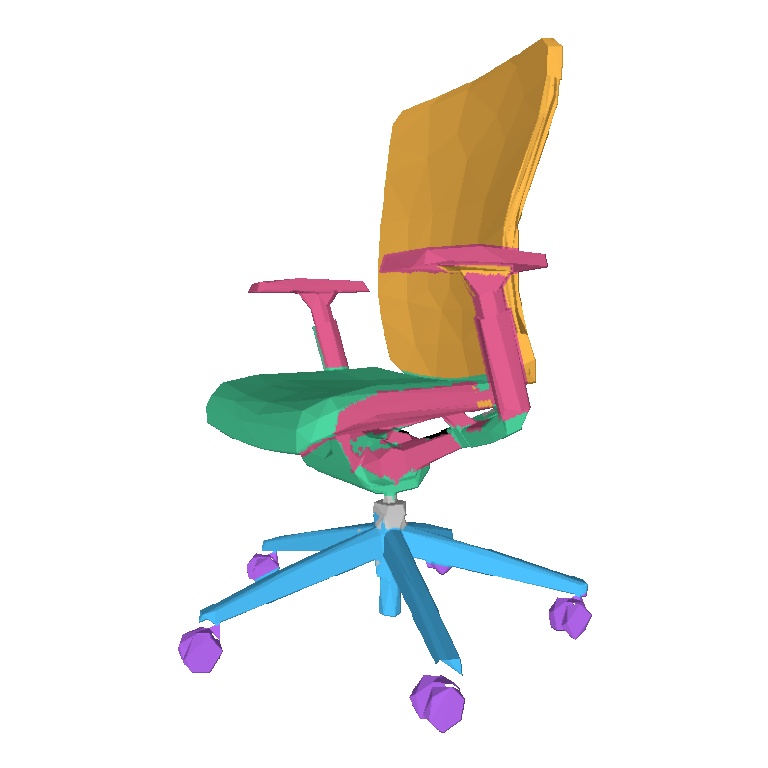} &
\includegraphics[width=0.1\textwidth]{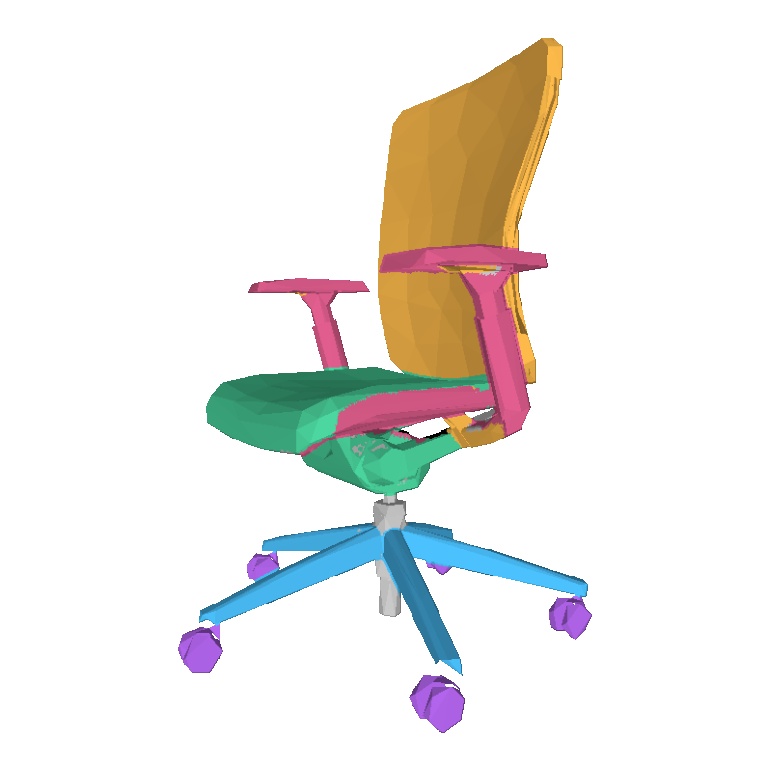} &
\includegraphics[width=0.1\textwidth]{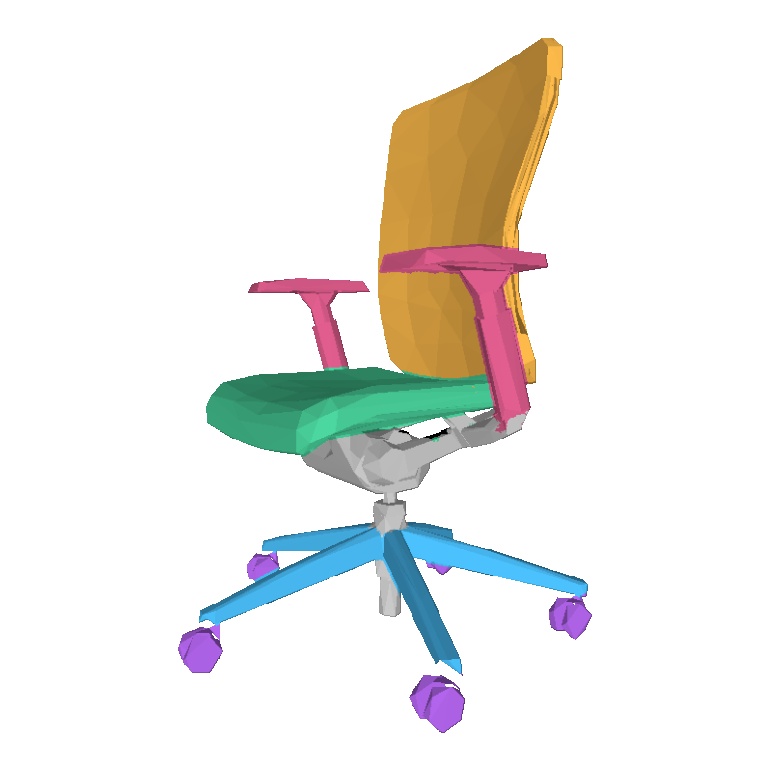} \\

\addlinespace[-2pt]
\arrayrulecolor{gray}\cmidrule(lr){1-5}
\arrayrulecolor{black}

\includegraphics[width=0.1\textwidth]{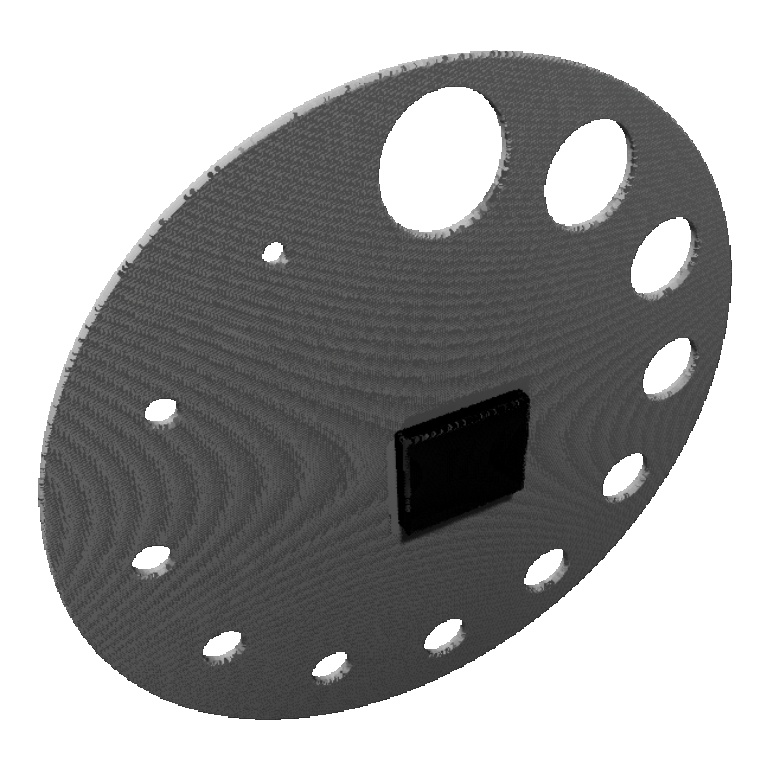} &
\includegraphics[width=0.1\textwidth]{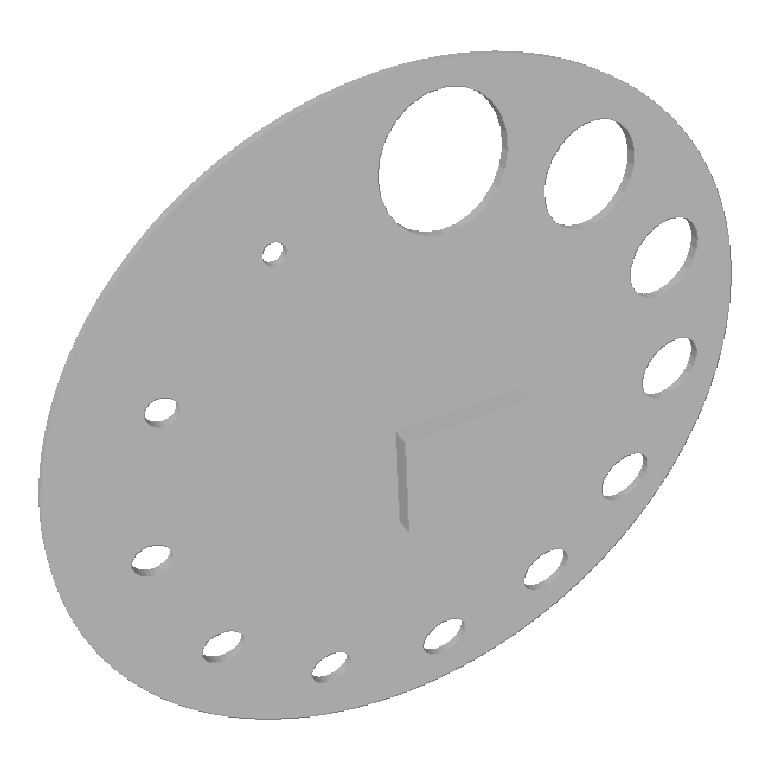} &
\includegraphics[width=0.1\textwidth]{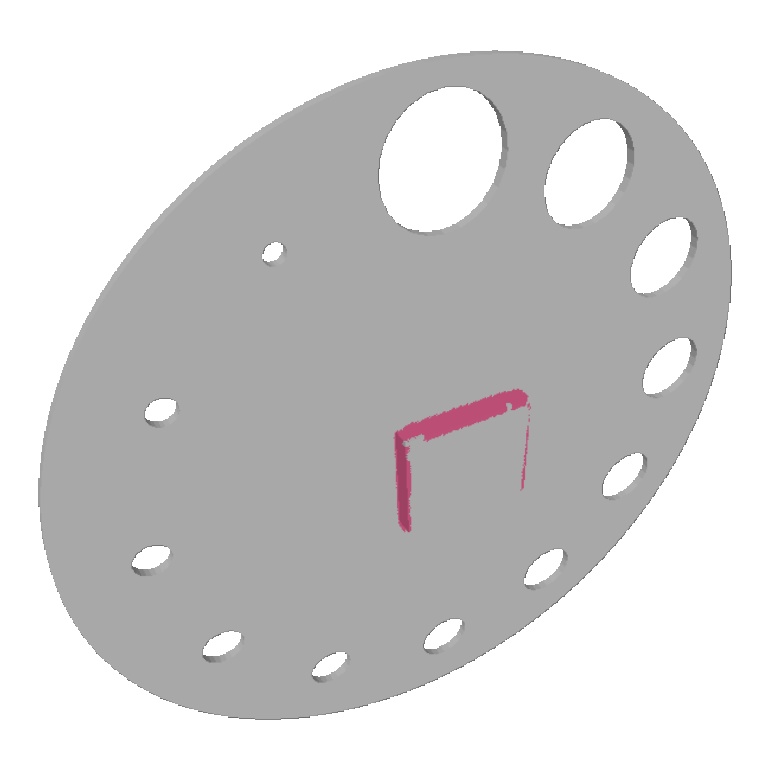} &
\includegraphics[width=0.1\textwidth]{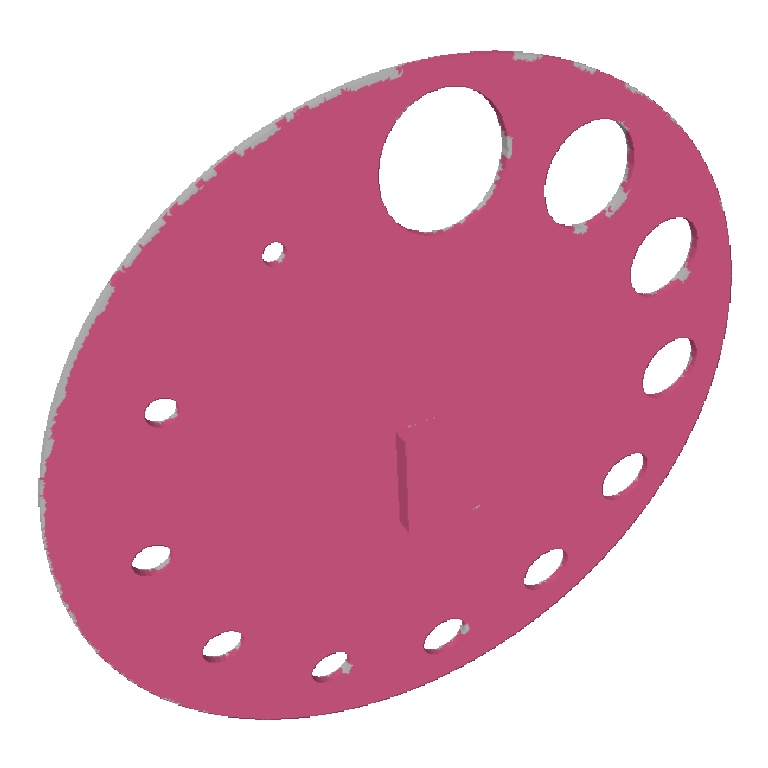} &
\includegraphics[width=0.1\textwidth]{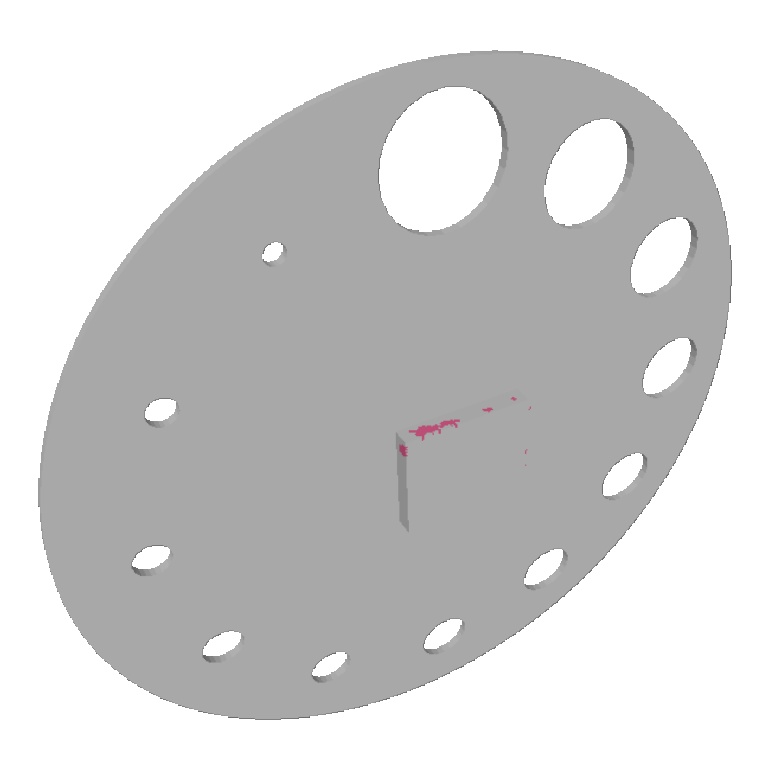} \\

\addlinespace[-2pt]
\arrayrulecolor{gray}\cmidrule(lr){1-5}
\arrayrulecolor{black}

\includegraphics[width=0.1\textwidth]{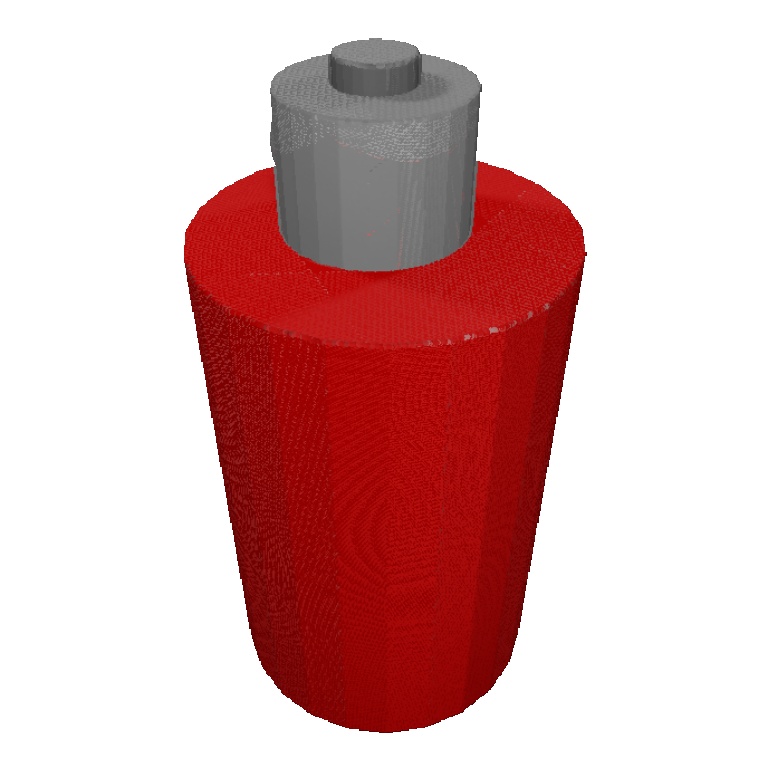} &
\includegraphics[width=0.1\textwidth]{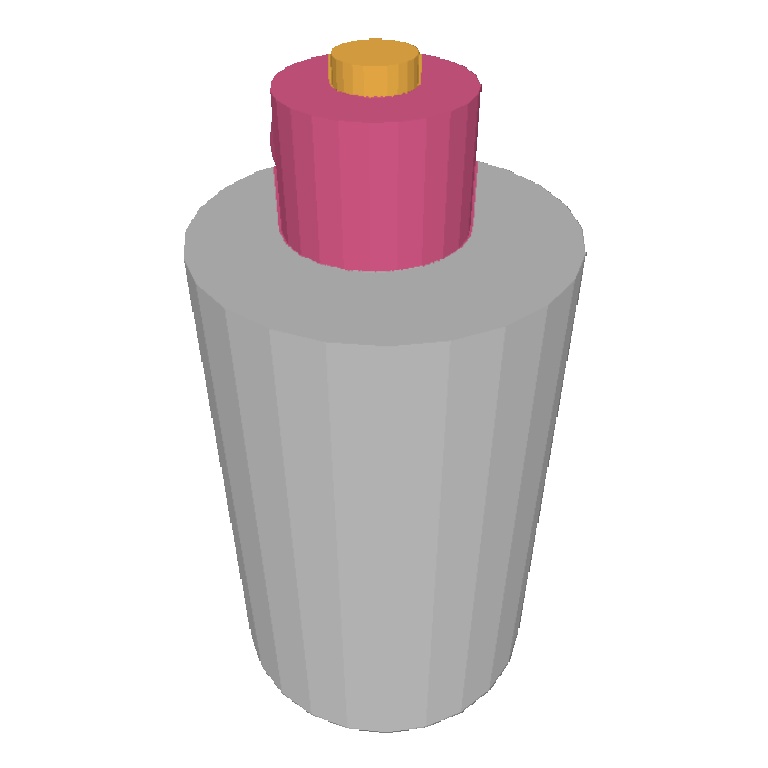} &
\includegraphics[width=0.1\textwidth]{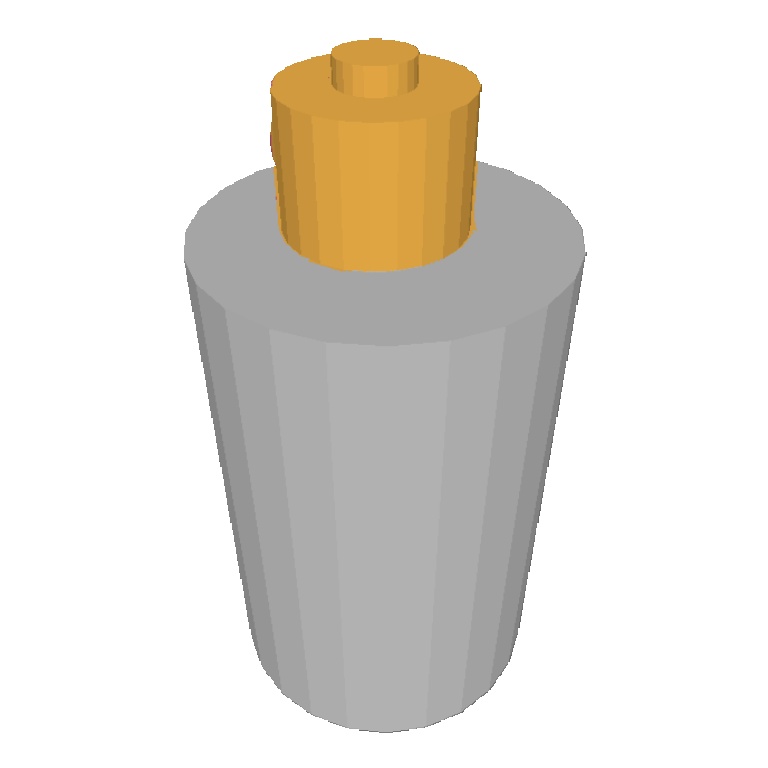} &
\includegraphics[width=0.1\textwidth]{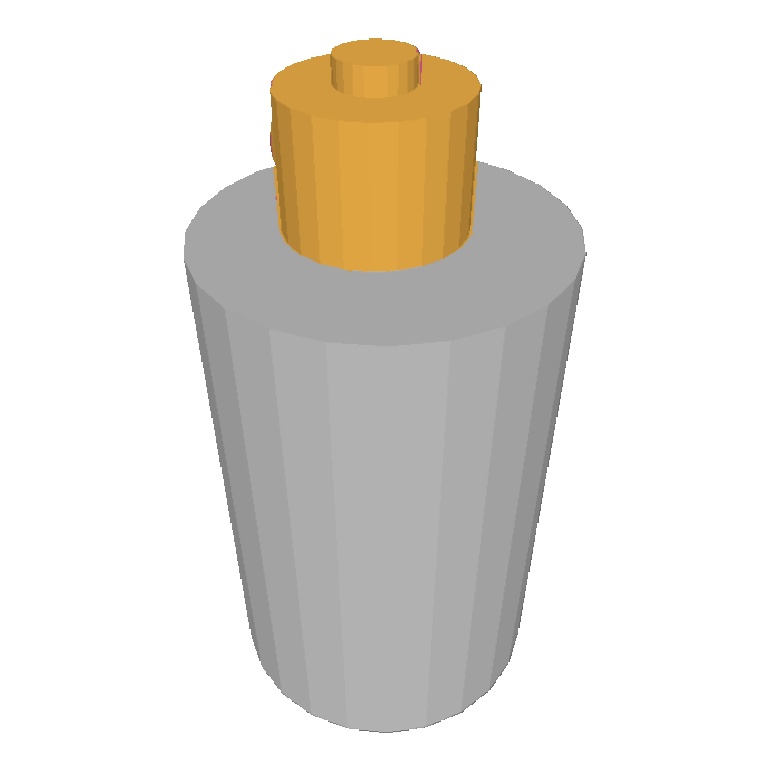} &
\includegraphics[width=0.1\textwidth]{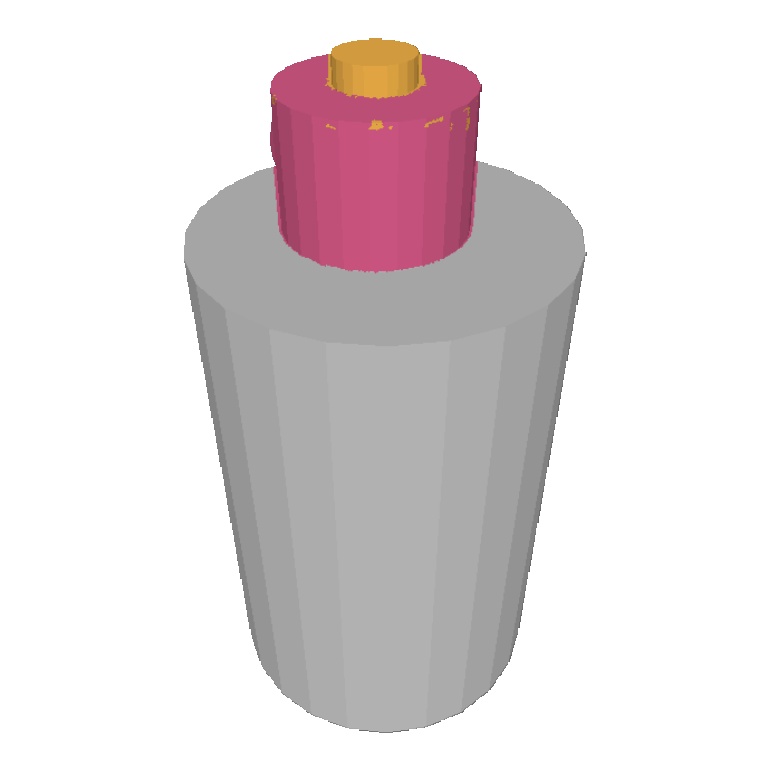} \\

\addlinespace[-2pt]
\arrayrulecolor{gray}\cmidrule(lr){1-5}
\arrayrulecolor{black}

\includegraphics[width=0.1\textwidth]{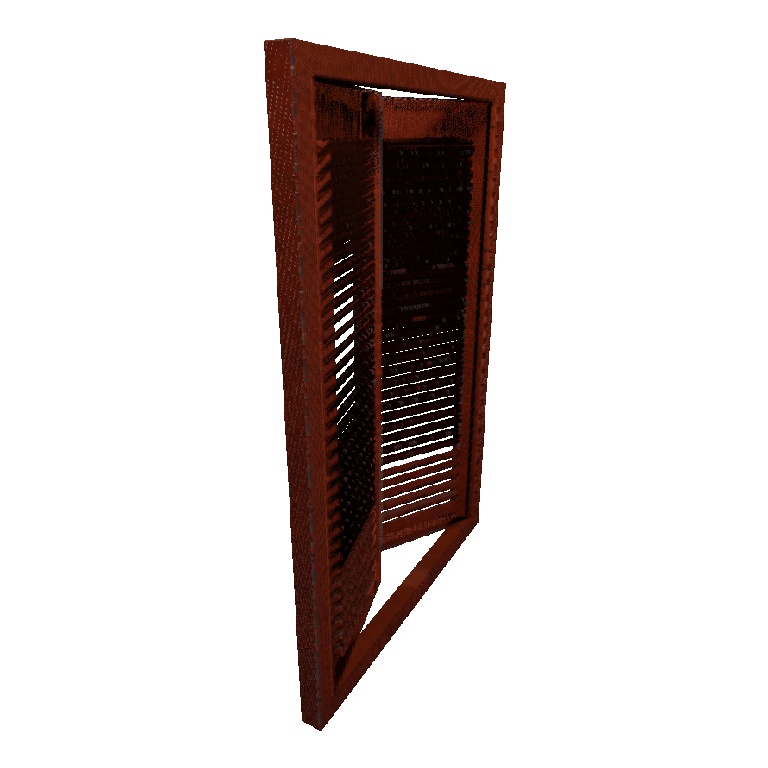} &
\includegraphics[width=0.1\textwidth]{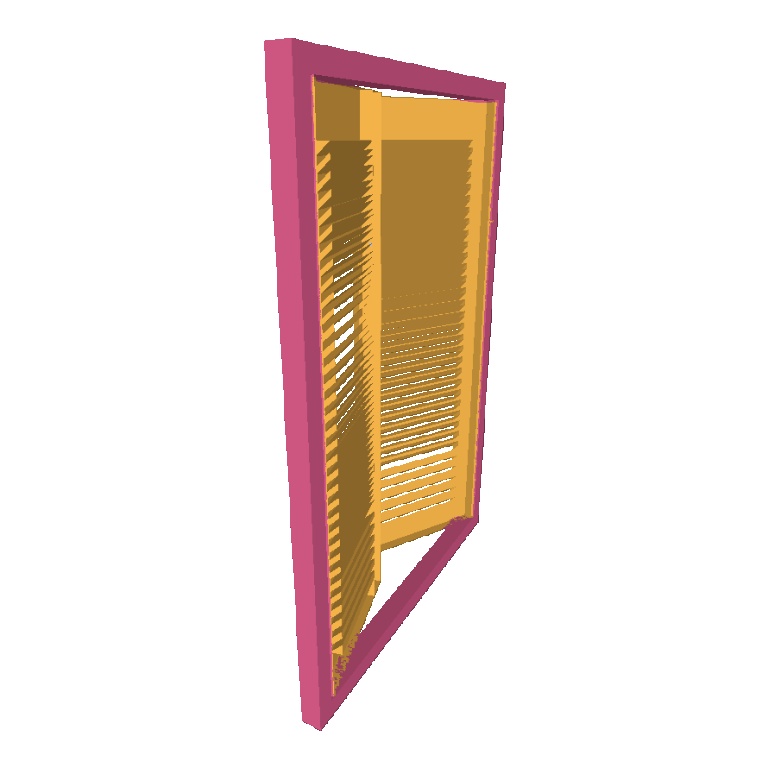} &
\includegraphics[width=0.1\textwidth]{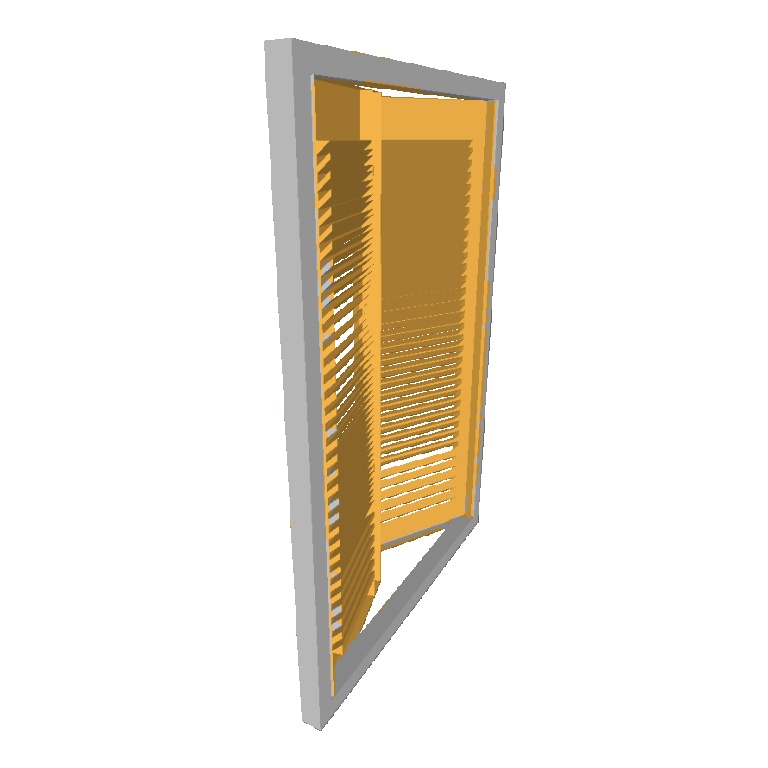} &
\includegraphics[width=0.1\textwidth]{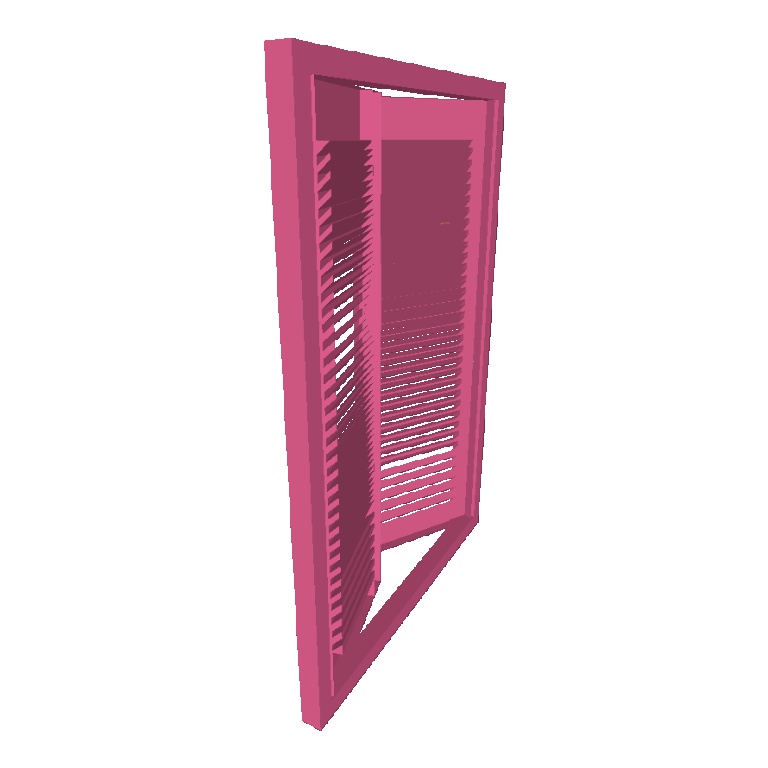} &
\includegraphics[width=0.1\textwidth]{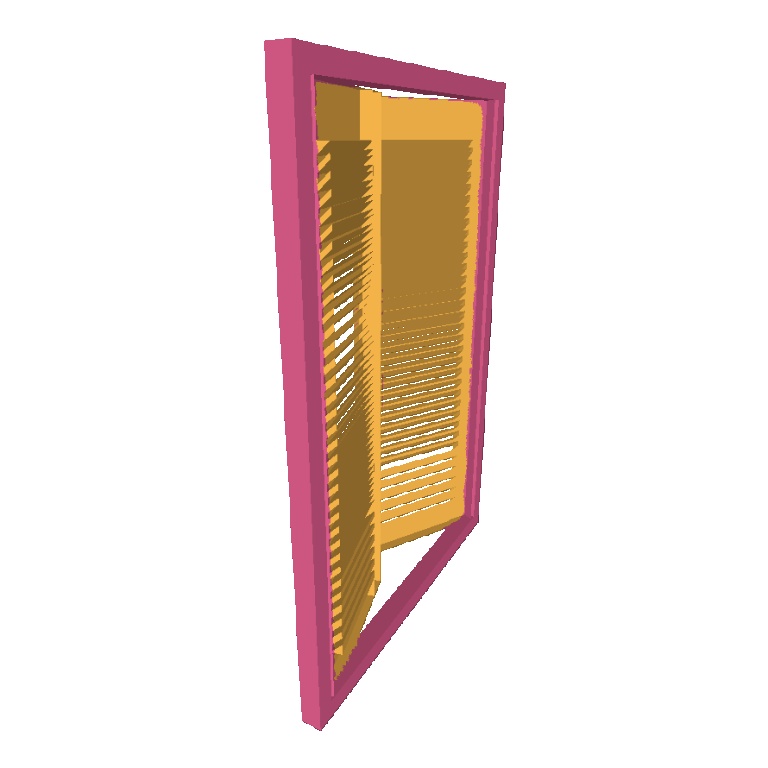} \\

\addlinespace[-2pt]
\arrayrulecolor{gray}\cmidrule(lr){1-5}
\arrayrulecolor{black}

\includegraphics[width=0.1\textwidth]{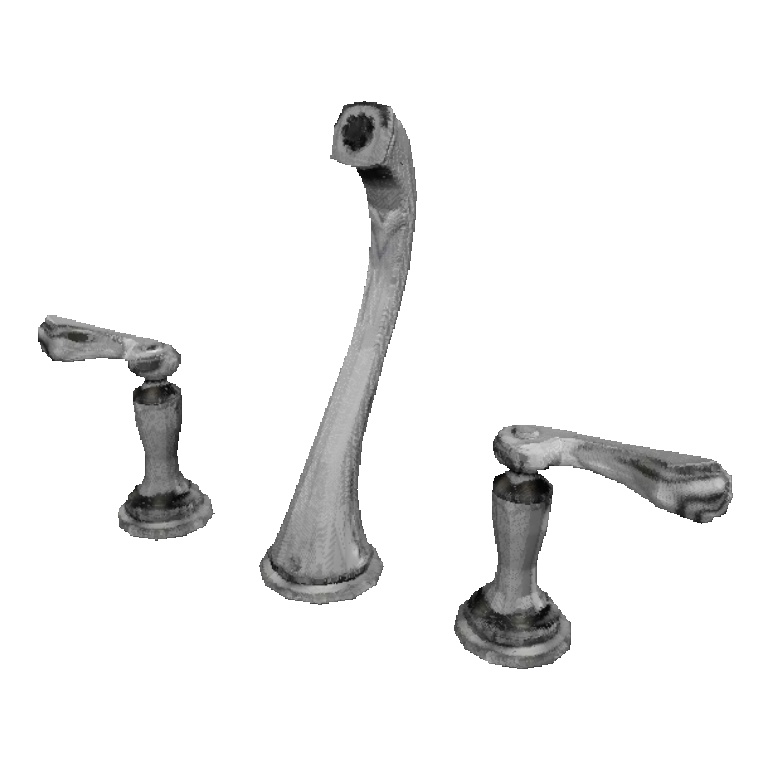} &
\includegraphics[width=0.1\textwidth]{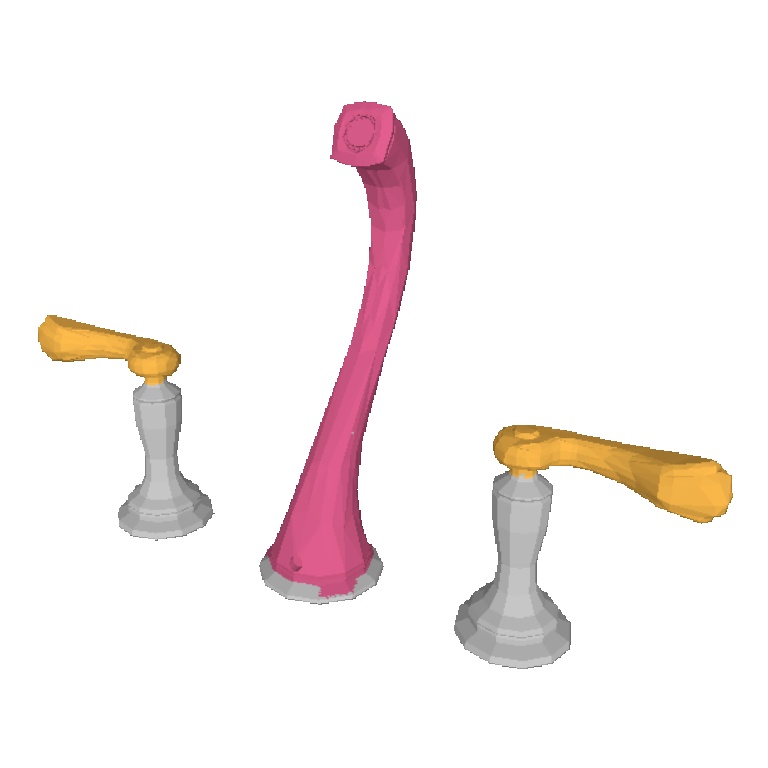} &
\includegraphics[width=0.1\textwidth]{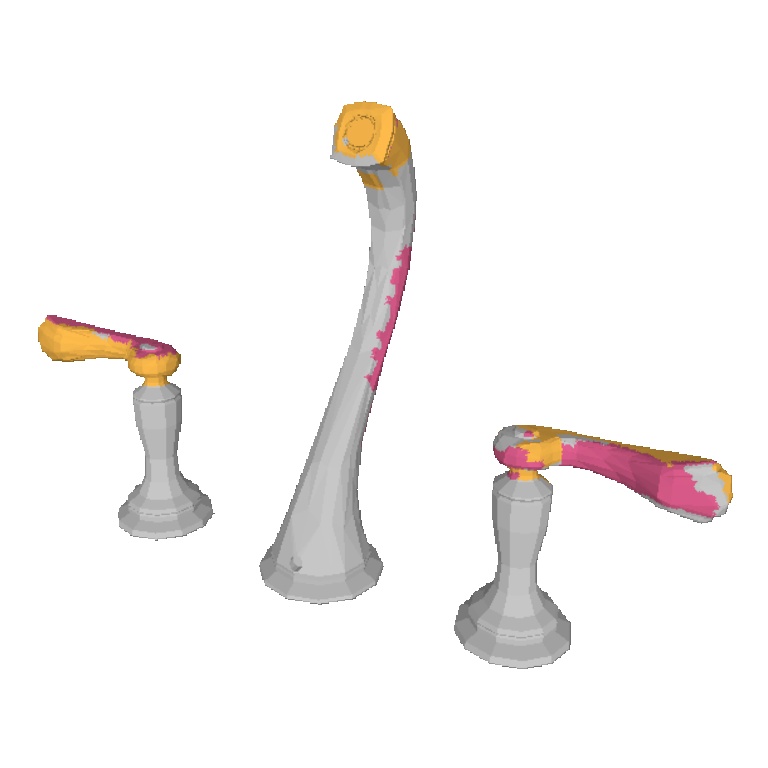} &
\includegraphics[width=0.1\textwidth]{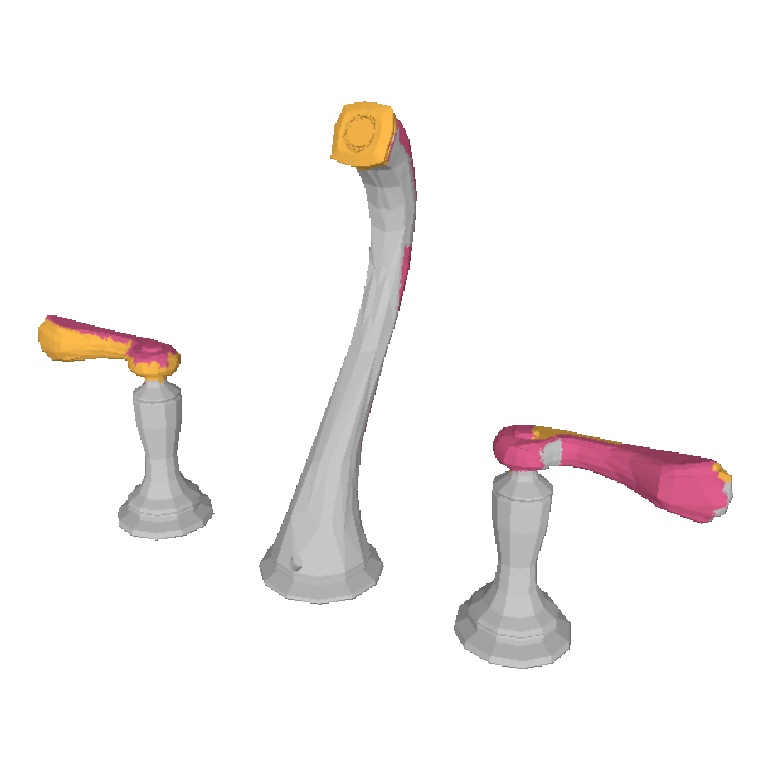} &
\includegraphics[width=0.1\textwidth]{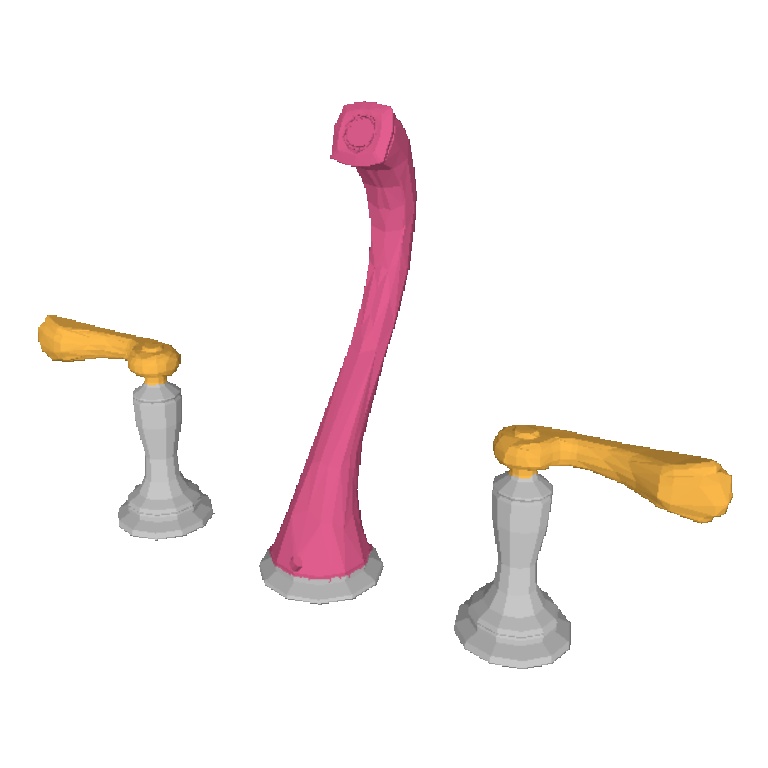} \\

\addlinespace[-2pt]
\arrayrulecolor{gray}\cmidrule(lr){1-5}
\arrayrulecolor{black}

\includegraphics[width=0.1\textwidth]{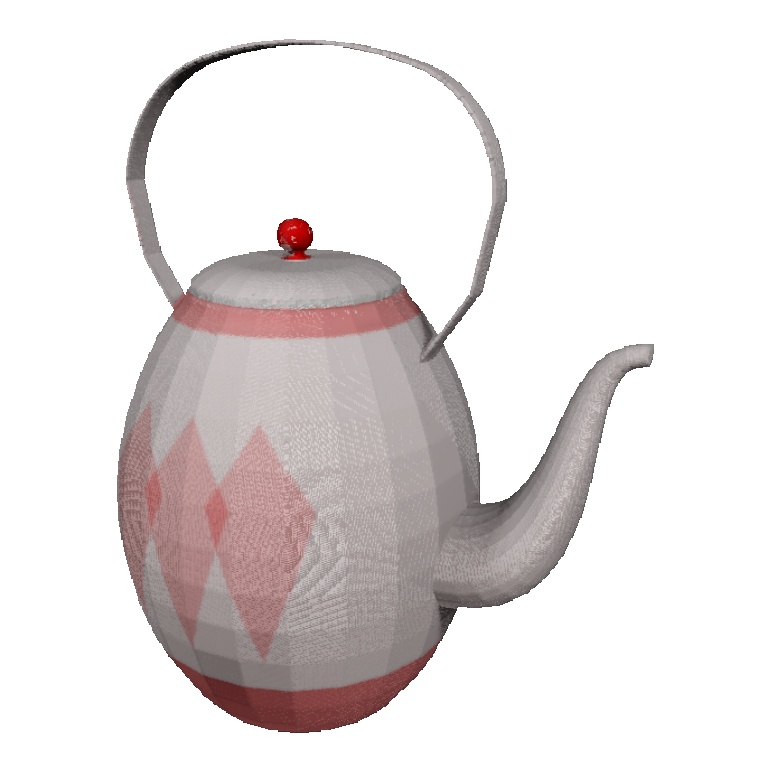} &
\includegraphics[width=0.1\textwidth]{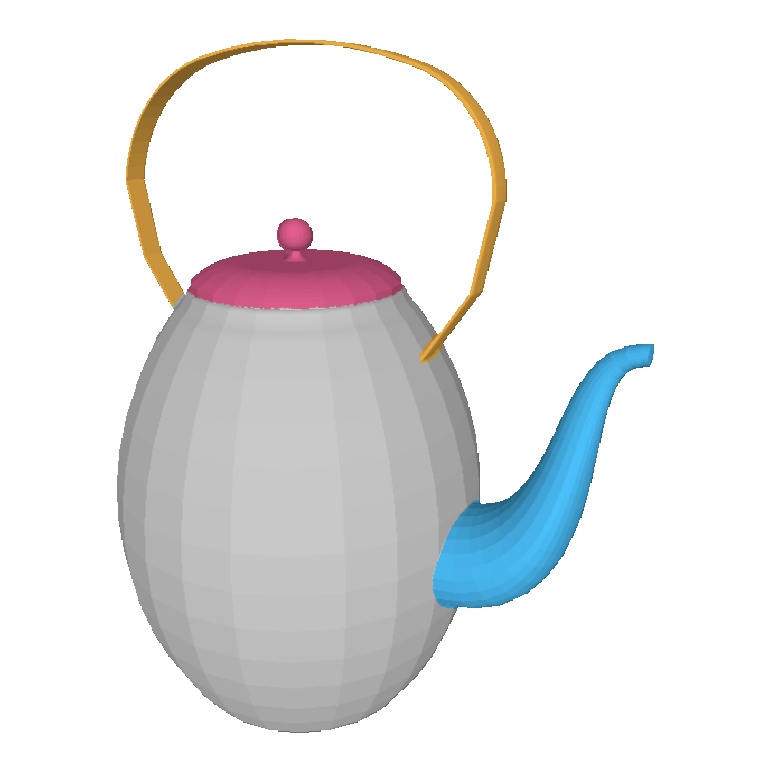} &
\includegraphics[width=0.1\textwidth]{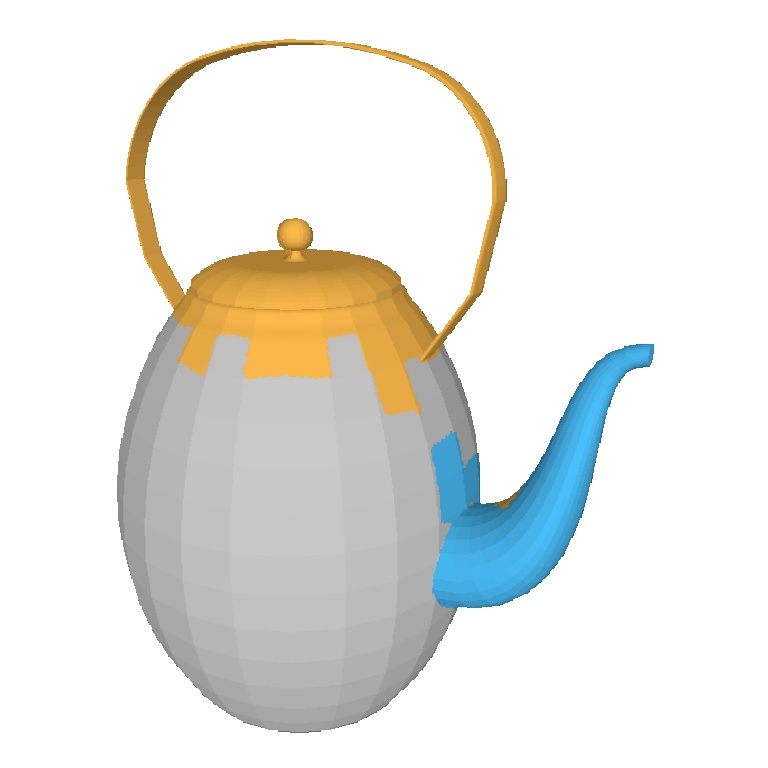} &
\includegraphics[width=0.1\textwidth]{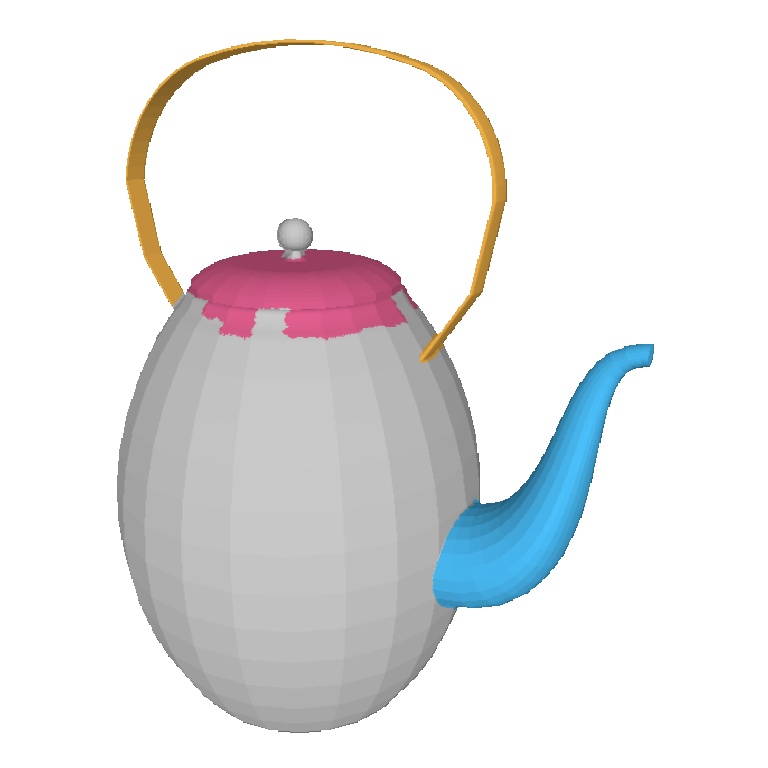} &
\includegraphics[width=0.1\textwidth]{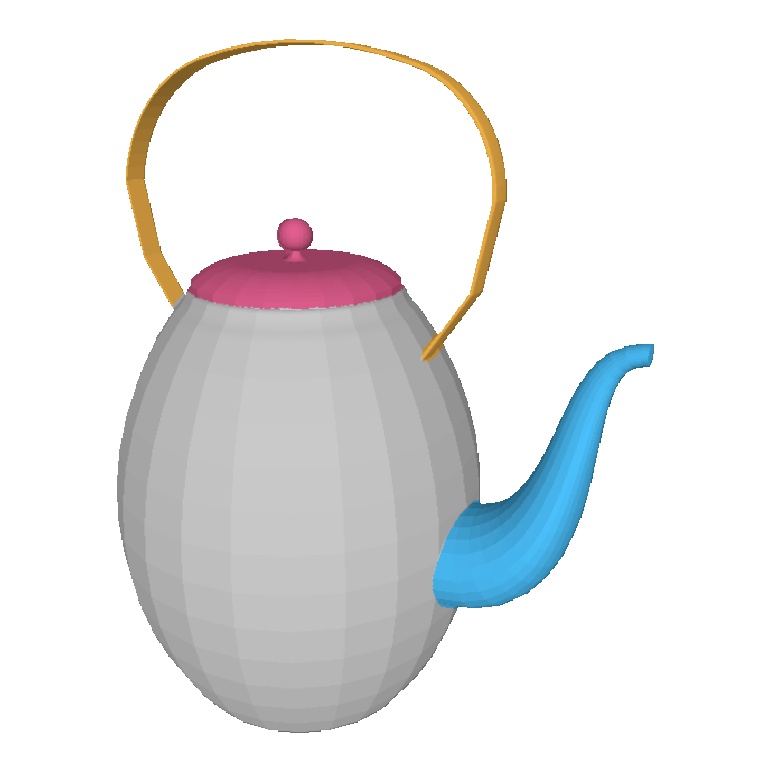} \\

\addlinespace[-2pt]
\arrayrulecolor{gray}\cmidrule(lr){1-5}
\arrayrulecolor{black}

\includegraphics[width=0.1\textwidth]{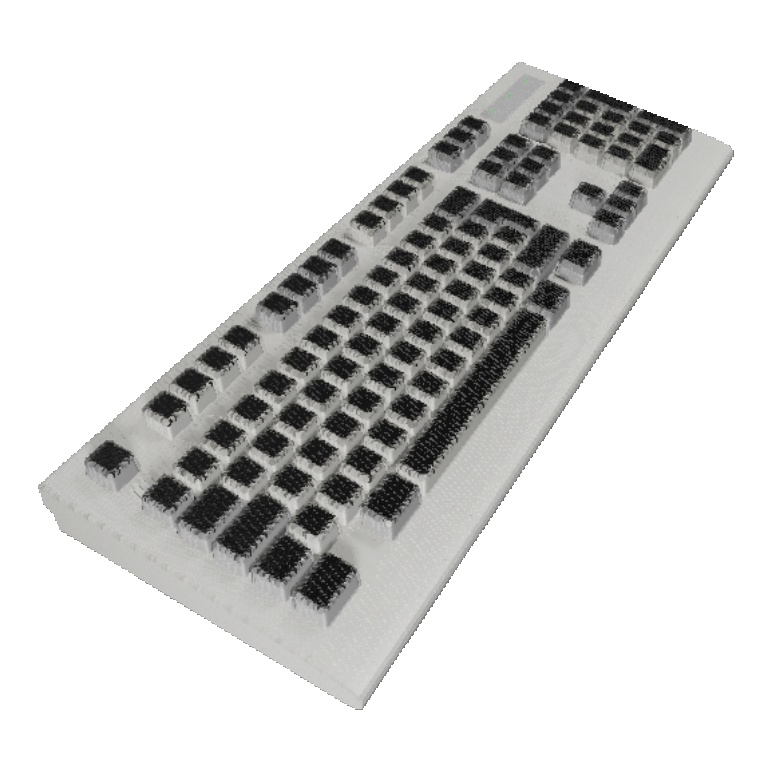} &
\includegraphics[width=0.1\textwidth]{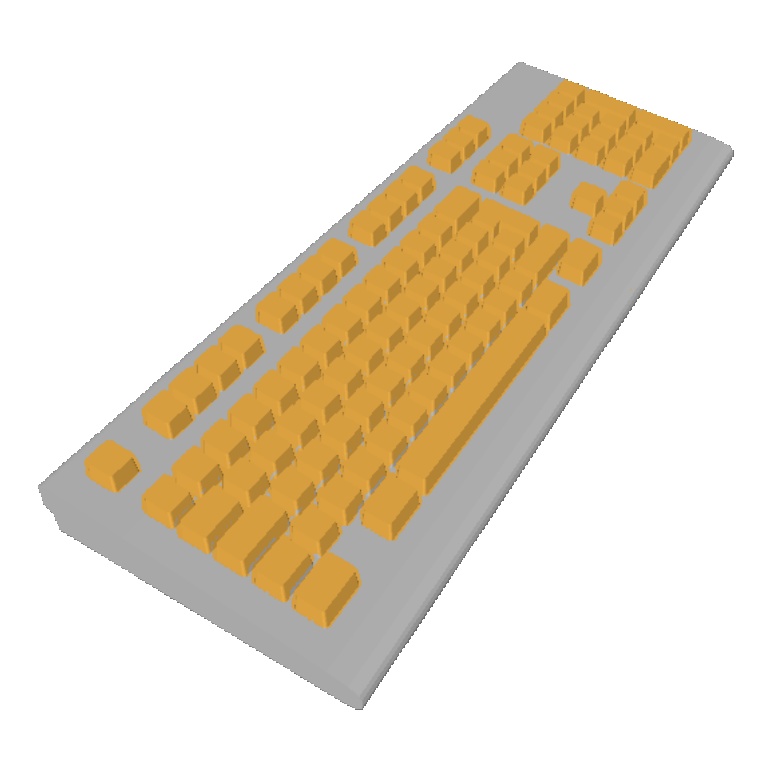} &
\includegraphics[width=0.1\textwidth]{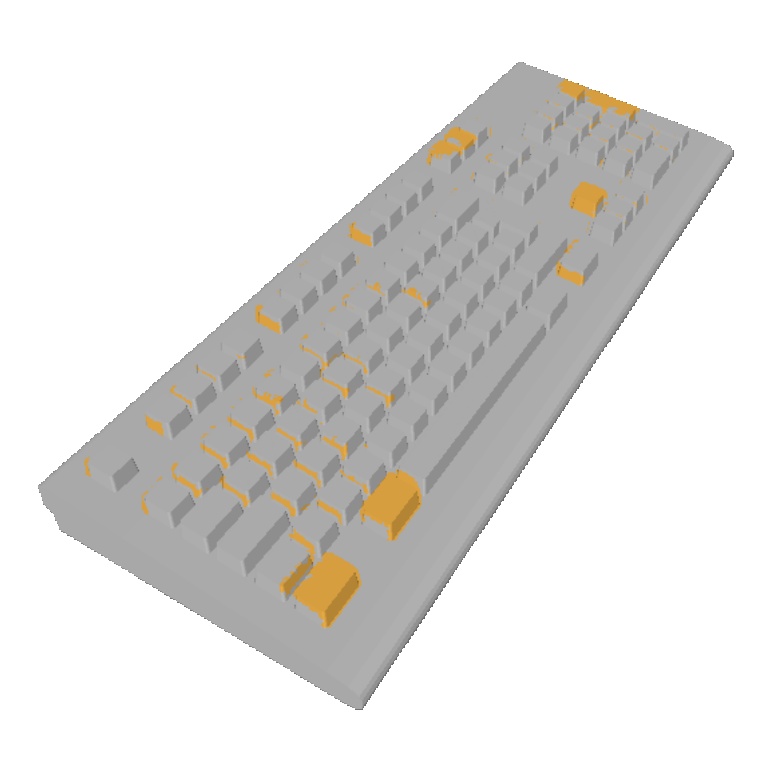} &
\includegraphics[width=0.1\textwidth]{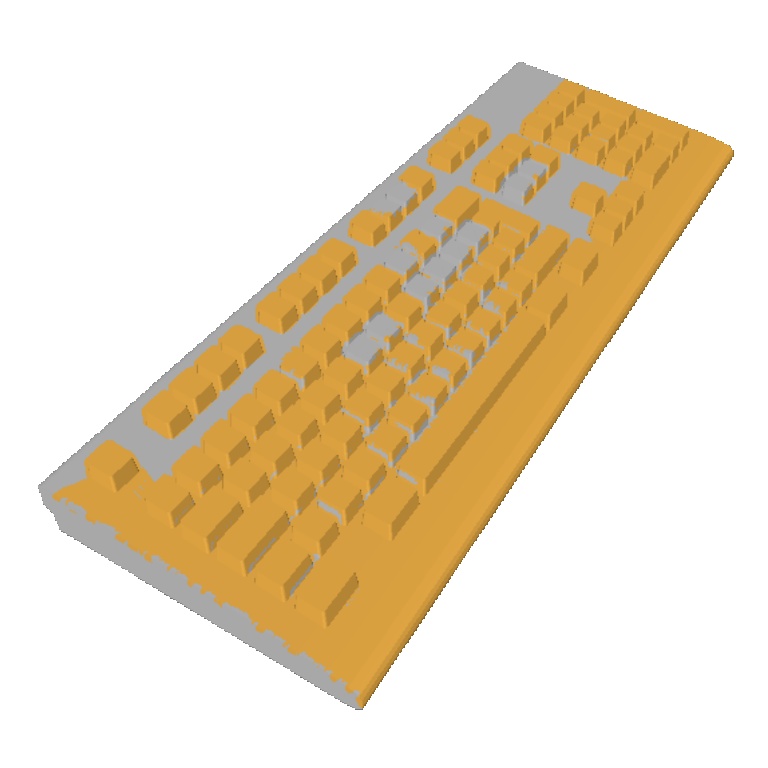} &
\includegraphics[width=0.1\textwidth]{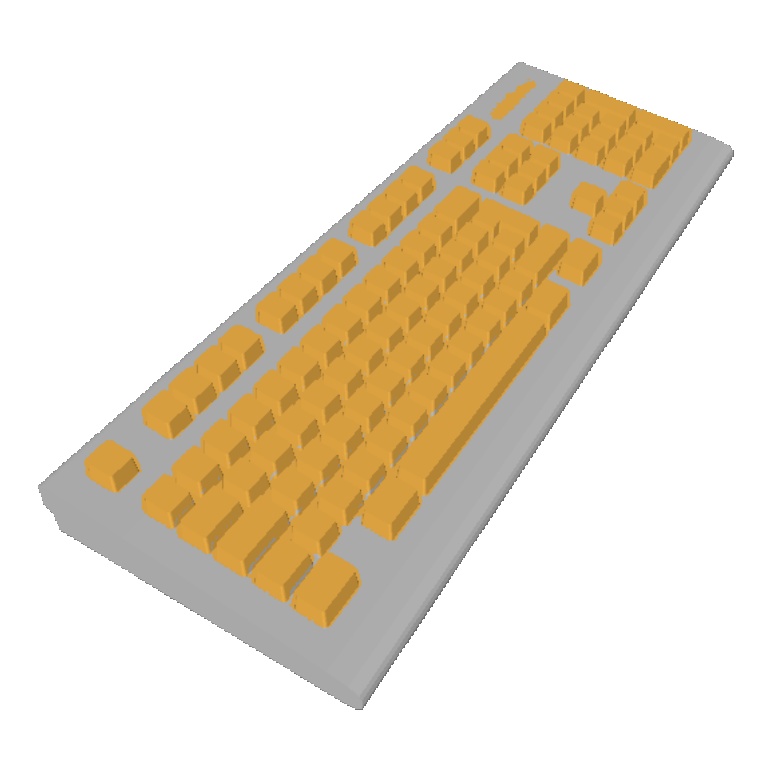} \\

\addlinespace[-2pt]
\arrayrulecolor{gray}\cmidrule(lr){1-5}
\arrayrulecolor{black}

\includegraphics[width=0.1\textwidth]{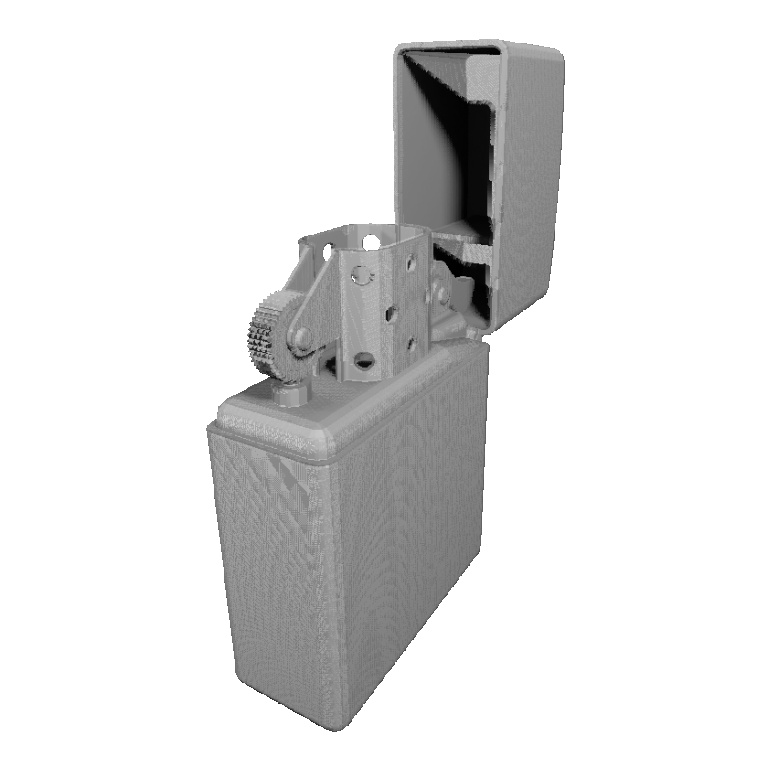} &
\includegraphics[width=0.1\textwidth]{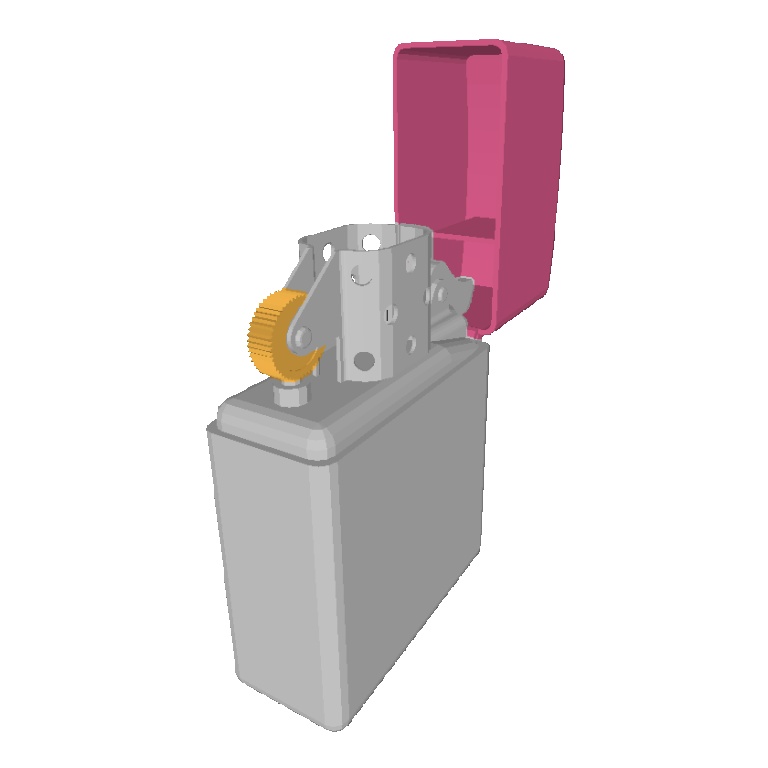} &
\includegraphics[width=0.1\textwidth]{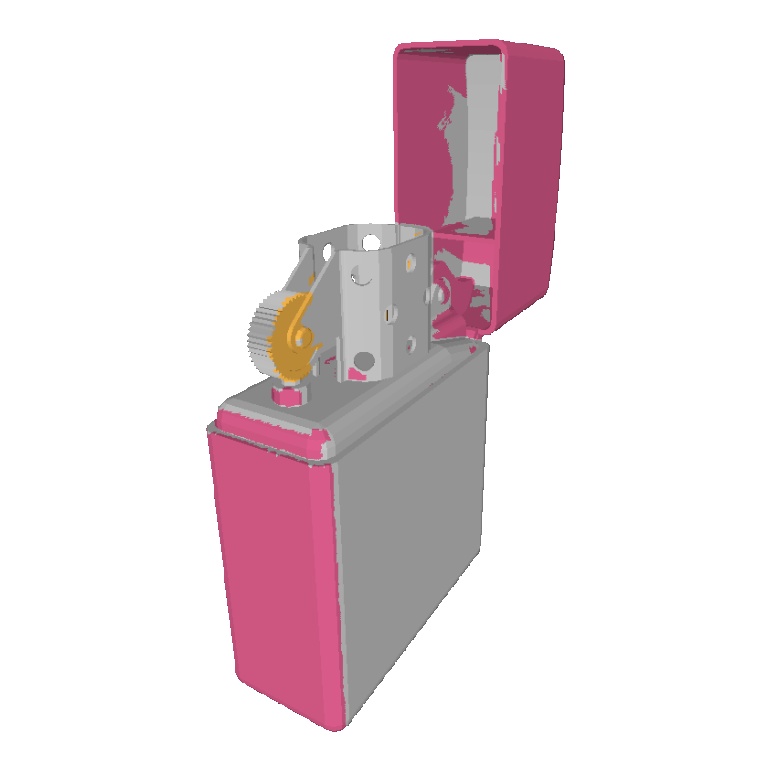} &
\includegraphics[width=0.1\textwidth]{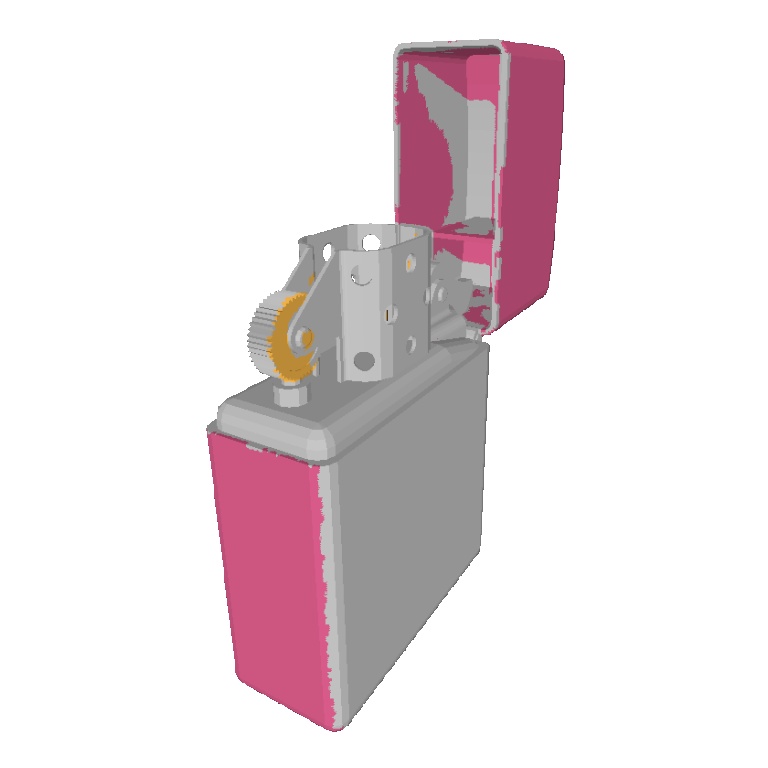} &
\includegraphics[width=0.1\textwidth]{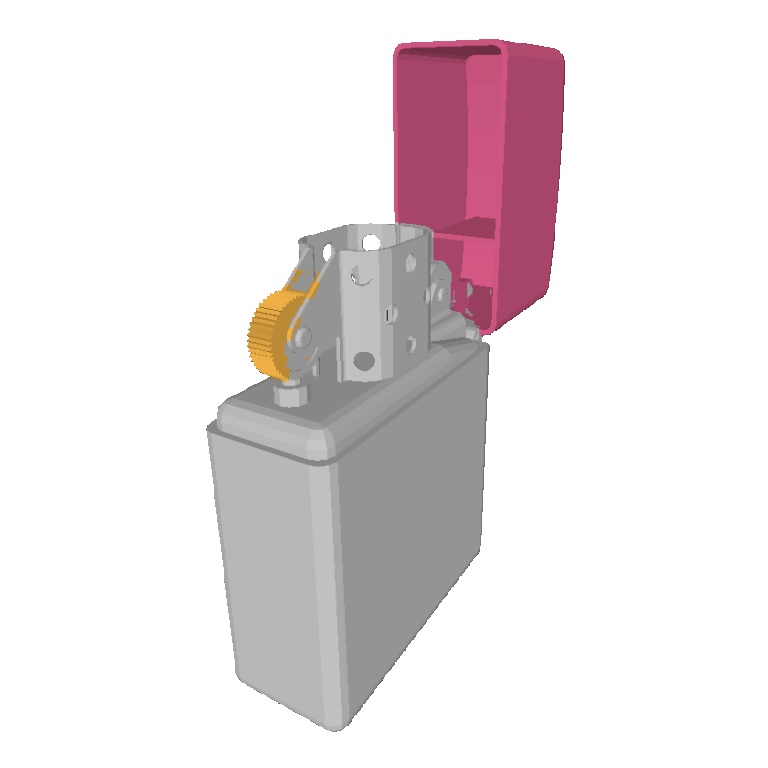} \\

\addlinespace[-2pt]
\arrayrulecolor{gray}\cmidrule(lr){1-5}
\arrayrulecolor{black}

\includegraphics[width=0.1\textwidth]{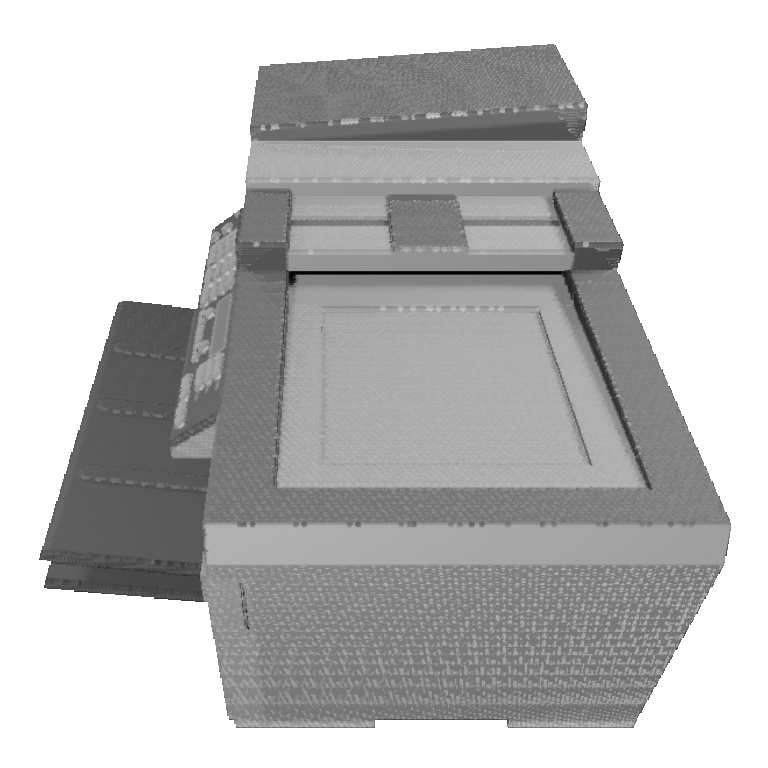} &
\includegraphics[width=0.1\textwidth]{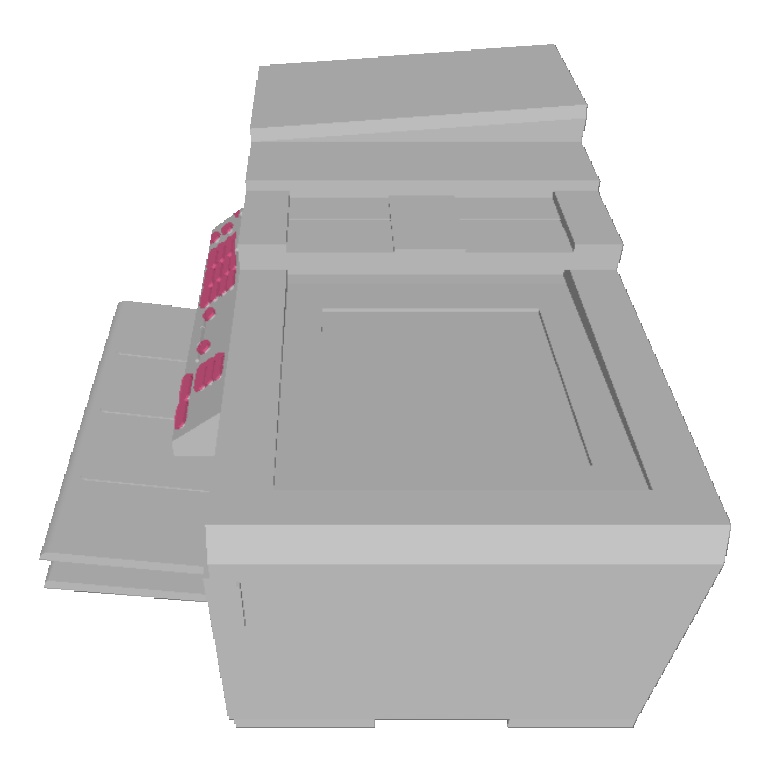} &
\includegraphics[width=0.1\textwidth]{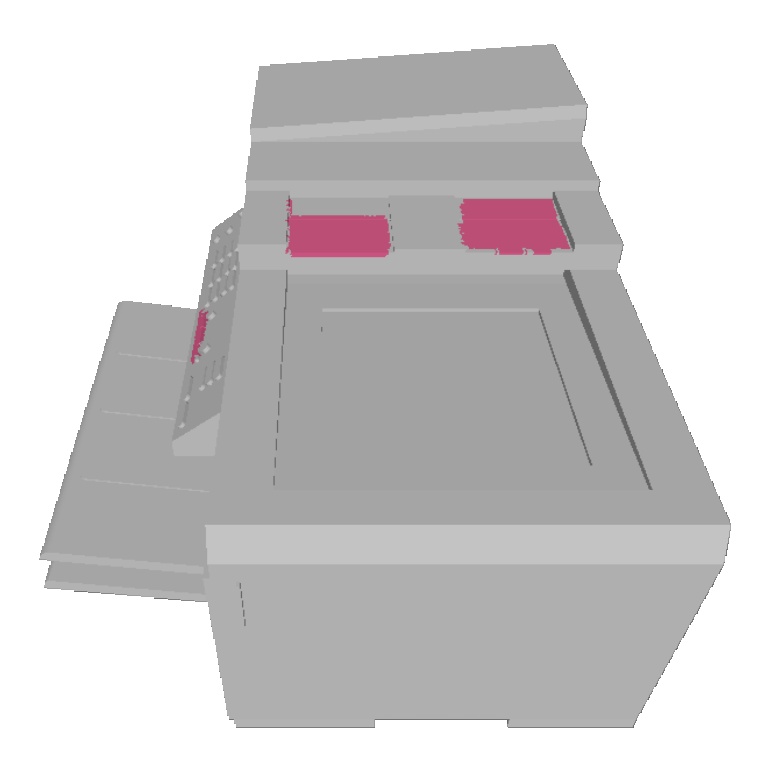} &
\includegraphics[width=0.1\textwidth]{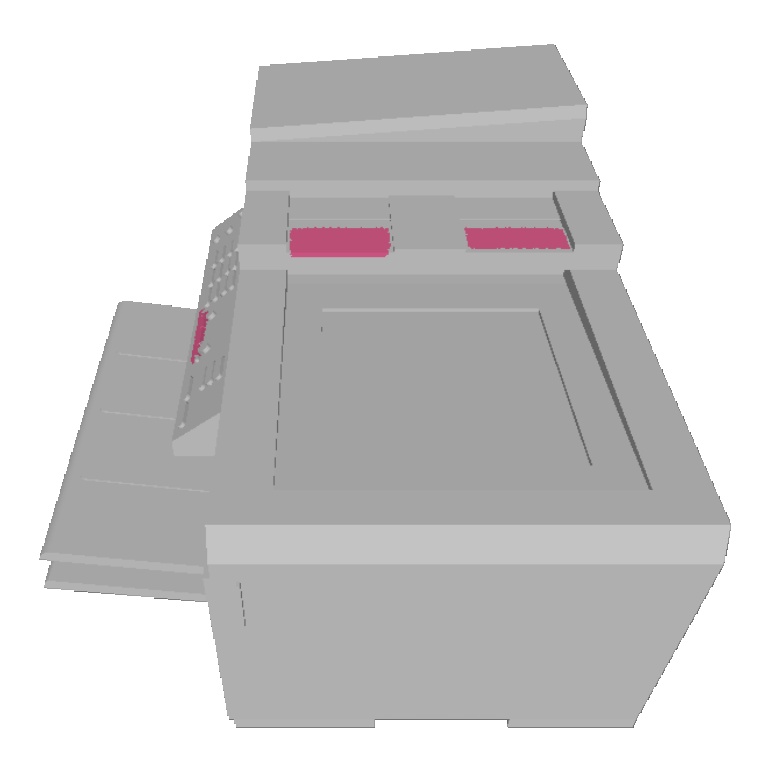} &
\includegraphics[width=0.1\textwidth]{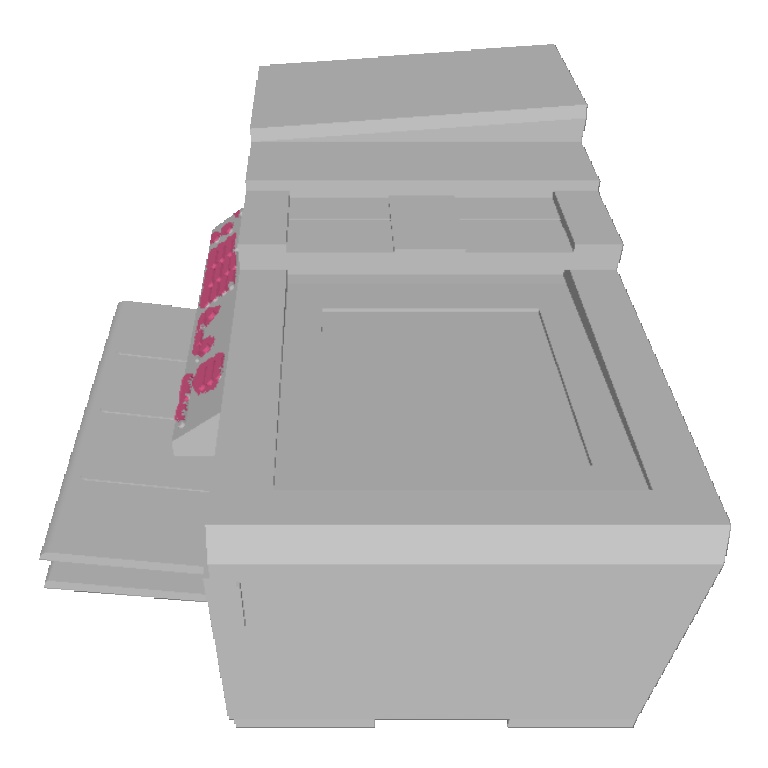} \\

\addlinespace[-2pt]
\arrayrulecolor{gray}\cmidrule(lr){1-5}
\arrayrulecolor{black}

\includegraphics[width=0.1\textwidth]{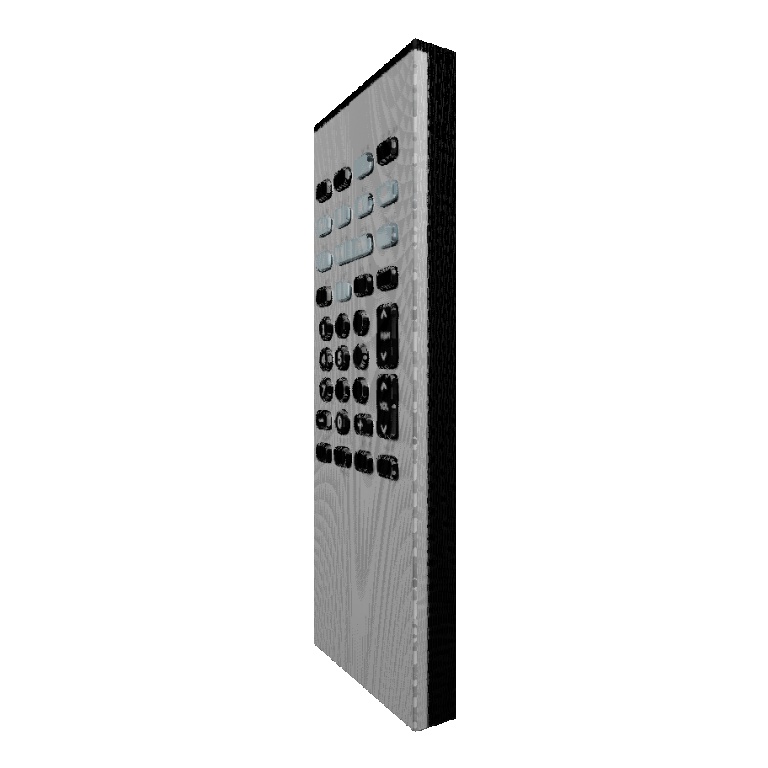} &
\includegraphics[width=0.1\textwidth]{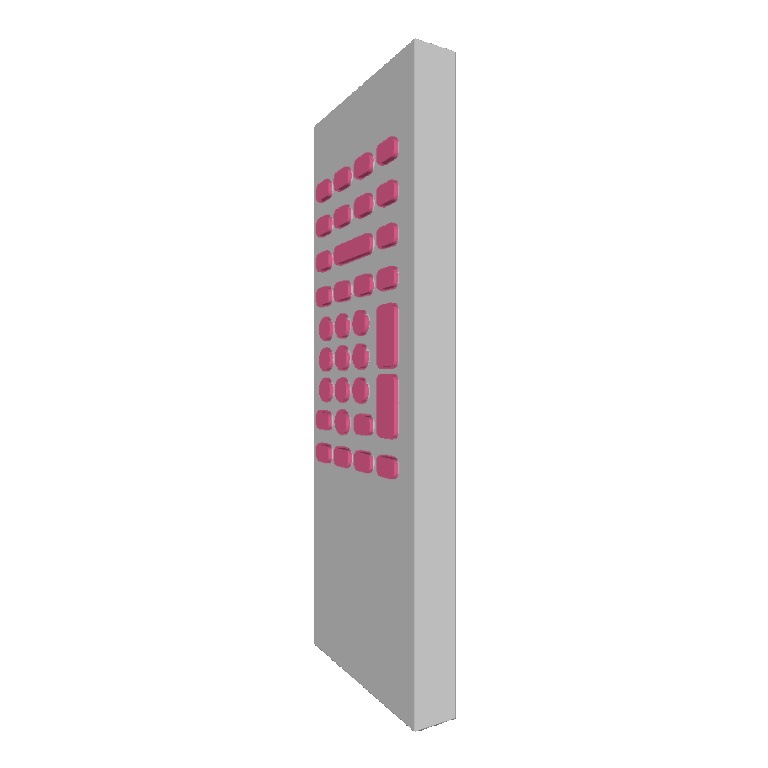} &
\includegraphics[width=0.1\textwidth]{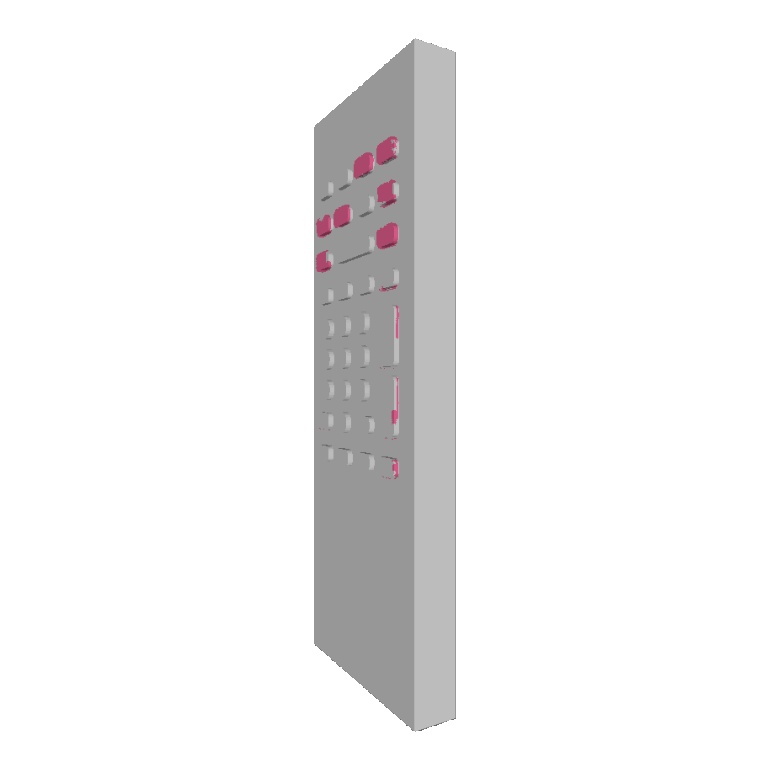} &
\includegraphics[width=0.1\textwidth]{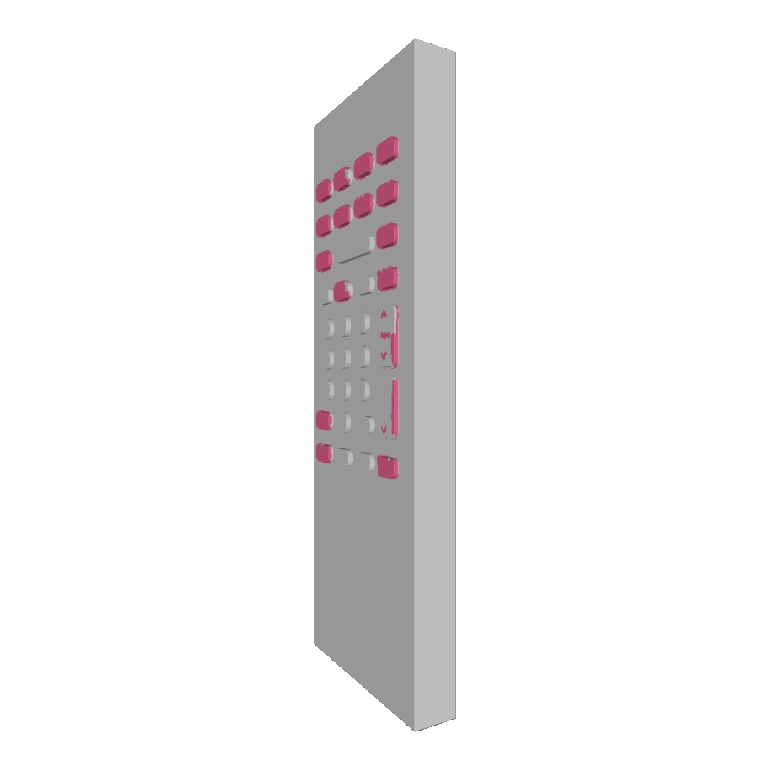} &
\includegraphics[width=0.1\textwidth]{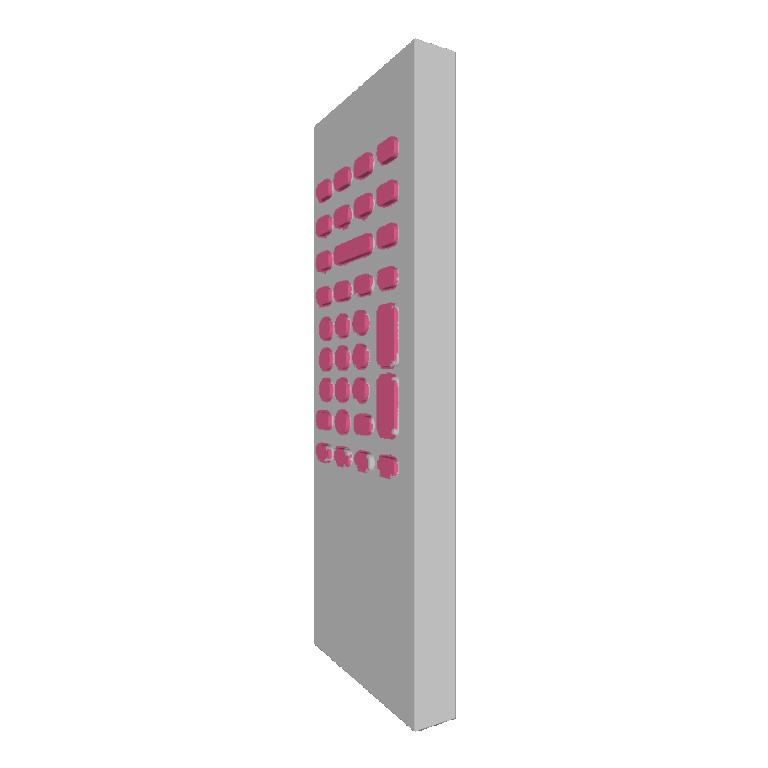} \\

\addlinespace[-2pt]
\bottomrule
\end{tabular}
}}
\end{tabular}
\end{table*}

\clearpage
\newpage
\thispagestyle{empty} 
\twocolumn[
\begin{center}
    {\SMtitlefont \textcolor{title_col_1}{A}\textcolor{title_col_2}{S}\textcolor{title_col_3}{I}\textcolor{title_col_4}{A}: \textcolor{title_col_1}{A}daptive 3D \textcolor{title_col_2}{S}egmentation 
 using Few \textcolor{title_col_3}{I}mage \textcolor{title_col_4}{A}nnotations\par}
    \vspace{0.75em}
    {\SMsubtitlefont\bfseries (Supplementary Material)\par}
    \vspace{1em}
\end{center}
]
\appendix


This supplementary material provides a detailed overview of how our approach, \PaperName, differs from existing methods for 3D shape segmentation, outlines the motivation behind our method and the rationale for key design choices in Sec.~\ref{sup_sec:rel_works}, provides implementation details in Sec.~\ref{sup_sec:impl_details}, presents additional ablation studies in Sec.~\ref{sup_sec:ablation_exps}, provides additional results for both adaptive and semantic part segmentation in Sec.~\ref{sup_sec:result_discussion}, and showcases our model's ability to perform \textit{Part Composition} in Sec.~\ref{sup_sec:part composition}.

\section{\PaperName{} vs Traditional 3D Segmentation}
\label{sup_sec:rel_works}

3D shape segmentation is a well-studied problem~\cite{shamir2006segmentation, qi2017pointnet, decatur20233d}, with a wide range of applications including localized editing, texture mapping, and simulation. 
However, nearly all existing methods focus on segmenting semantic or text-describable parts of 3D shapes. 
Earlier approaches~\cite{qi2017pointnet, qi2017pointnet++} typically relied on traditional supervised learning, requiring large amounts of annotated data. 
More recent methods~\cite{liu2023partslip, abdelreheem2023satr, zhong2024meshsegmenter} leverage pre-trained foundational vision-language models (VLMs) to enable zero- or few-shot segmentation. 
Nevertheless, these efforts continue to be limited to segmenting only semantic and text-describable components of 3D shapes.

In this work, we introduce \PaperName, a \textit{controllable} \textit{few-shot} approach for \textit{Adaptive 3D Part Segmentation}—the first of its kind.
It refers to segmenting 3D shapes into non-semantic and non-text describable parts (\textit{adaptive}), as the user desires (\textit{controllability}) using only a few annotated in-the-wild images as references (\textit{few-shot}).
Semantic or text-describable segmentation can be seen as a special case of adaptive part segmentation. In fact, our method outperforms existing approaches on semantic part segmentation by a significant margin (see Tab.~\ref{tab:supp_sem_seg_full_table}).
Furthermore, we demonstrate that our approach generalizes well across a wide range of object categories and adapts effectively to diverse shapes within the same category (see Fig.~\ref{fig:custom_seg_1}, Fig.~\ref{fig:custom_seg_2}, Fig.~\ref{fig:custom_seg_3}, and Fig.~\ref{fig:custom_seg_4}).
To better contextualize our contributions, we address a few key questions surrounding the motivation, challenges, and design choices behind \textit{\PaperName}, as well as the broader task of adaptive 3D part segmentation in what follows.

\subsubsection*{Why do we need Adaptive 3D Part Segmentation?}
Adaptive 3D part segmentation represents the most general form of 3D shape segmentation, with semantic segmentation being a special case of it. Unlike traditional approaches~\cite{liu2023partslip, kim2024partstad} that rely solely on predefined, text-describable parts, adaptive segmentation enables users to define parts in a flexible and task-specific manner.
This capability has broad applicability across various domains:
\begin{itemize}
    \item \textit{Fashion Industry:} Enables automated pattern cutting for garments and shoes by identifying adaptive regions aligned with design or manufacturing constraints.

    \item \textit{Visual Effects and Animation:} Facilitates localized editing and simulation by isolating meaningful adaptive regions.

    \item \textit{Interior Design and Manufacturing:} Assists in assigning materials or colors to user-specified parts, or breaking down furniture into assembly-friendly segments.

    \item \textit{Robotics and Grasp Planning:} Supports identifying functional regions for effective manipulation.

    \item \textit{Medical Applications:} Supports personalized segmentation of anatomical structures which may not align with standard semantic labels—enabling applications such as surgical planning, implant design, and patient-specific analysis.
\end{itemize}

The ability to segment 3D shapes in a user-defined manner, using only a few annotated in-the-wild reference images, provides a powerful and practical tool for real-world tasks that demand flexibility beyond standard semantic part definitions.

\subsubsection*{Why is few-shot Adaptive 3D Part Segmentation challenging?}
3D part segmentation can be done using traditional supervised methods that rely on large amounts of annotated training data. However, the challenge becomes significantly greater under a few-shot constraint—especially when only a handful of annotated \textit{in-the-wild} or \textit{online} reference images are available.

Existing works~\cite{liu2023partslip, 3by2, decatur20233d} leverage part priors from 2D foundational localization models such as GLIP~\cite{li2022grounded}, SAM~\cite{SAM_SegmentAnything}, and CLIP~\cite{radford2021learning} for 3D shape segmentation. These models are designed to associate visual content with human-interpretable, text-based semantic labels. As a result, 3D segmentation methods built on top of them often struggle to detect parts that lack clear semantic meaning or cannot be easily described in words.

To address this, we propose using a generative model, Stable Diffusion~\cite{rombach2021highresolution}, and leverage its emergent understanding of object structures for few-shot \textit{adaptive} part segmentation.

\subsubsection*{Why use images instead of text or 3D shapes for reference?}
While textual descriptions are inexpensive to obtain, they often fall short when it comes to precisely defining non-semantic regions, leading to ambiguity or inaccuracy. 
Annotated 3D shapes, though more precise, are challenging to create as they require expertise in 3D modeling software.
Images strike a balance — they can provide references nearly as precise as 3D shapes, but at a significantly lower annotation cost. Moreover, our approach does not require multi-view images of the same object while training, making it highly compatible with \textit{in-the-wild} or \textit{online} images and thereby enhancing its practical applicability.

\subsubsection*{Why is lifting 2D adaptive part labels to 3D challenging?}
Existing methods~\cite{zhong2024meshsegmenter, abdelreheem2023satr, hanocka2019meshcnn} for 3D mesh segmentation typically assign class labels to either vertices or faces, and there are two main reasons for this. First, assigning labels to vertices or faces provides a straightforward and practical way to segment meshes, whether lifting labels from multi-view 2D images or predicting them directly via a 3D shape segmentation network. Second, since most prior works focus on segmenting semantic parts—and meshes are often constructed with semantic structures in mind (unless generated automatically)—this labeling approach aligns well with their goals and works effectively.

In contrast, our work targets user-defined, adaptive part segmentation, where the parts may not correspond to semantic regions on the mesh. This requires a representation that can faithfully and efficiently lift such adaptive 2D labels into 3D. The UV-map of the mesh plays a crucial role in this context, as it assigns each 3D point $(x, y, z)$ on the mesh surface to a 2D coordinate $(u, v)$ in UV-space, effectively decoupling the label assignment from the underlying mesh geometry. This allows us to project adaptive labels from multi-view 2D images onto 3D surfaces with high fidelity and efficiency.

One might argue that vertex or face labels on finely subdivided meshes could yield higher-resolution segmentations. However, achieving this would require significantly subdividing coarse input meshes, something that users may not desire. Moreover, such an approach is also impractical due to the substantial increase in computational cost and processing time. Even with these efforts, the resulting meshes may still lack the spatial granularity that a high-resolution UV-map provides with ease.

However, this fine-grained representational power comes at a cost—a higher susceptibility to errors. This is where multi-view consistency becomes critical. If the predicted 2D labels across different views are not consistent, the aggregated 3D label after voting can be noisy, particularly around part boundaries and in low-confidence regions.
Previous methods~\cite{liu2023partslip, kim2024partstad, 3by2} that segment dense point clouds using multi-view 2D images also face similar issues. To mitigate this, they leverage part priors derived from clustering and segmentation models like Super-Points~\cite{landrieu2018large} and SAM~\cite{SAM_SegmentAnything}. However, such predefined groupings, of points or pixels, can restrict the flexibility needed for adaptive part segmentation.
Consequently, we deliberately avoid relying on any external "grouping" mechanisms and instead focus on enhancing multi-view consistency within our model. Specifically, we introduce \textit{part correspondence} loss, $\mathcal{L}_{corr}$, while training and per-view \textit{Noise Optimization} during inference. These components help reinforce multi-view alignment and lead to cleaner, more reliable 3D masks.


\subsubsection*{Comparisons}
Tab.~\ref{tab:rel_works_summary} of the main paper provides a comparative summary of relevant prior works, highlighting key attributes such as the ability to segment adaptive 3D parts, the modality of the query or annotated references, and the foundational models used.
We extensively evaluate our method on both adaptive and semantic 3D part segmentation tasks, and it establishes a new state-of-the-art in few-shot 3D shape segmentation. Please refer to Sec.~\ref{sup_sec:result_discussion} of this supplementary and Sec.~\ref{sec:experiments} of the main paper for detailed qualitative and quantitative results.


\section{Implementation Details}
\label{sup_sec:impl_details}

We use Stable Diffusion 2.1~\cite{rombach2021highresolution} in all our experiments and adopt the same set of self- and cross-attention layers as SLiMe~\cite{khani2023slime} to generate the segmentations $\mathbf{W}$.
We render 10 images per object at a resolution of $768 \times 768$ from fixed viewpoints for training.
All 2D renderings were generated using PyTorch3D~\cite{ravi2020pytorch3d}, with camera-aligned lighting and either a black or white background.
All the meshes we selected had textures, but any untextured mesh can also be easily textured using recent techniques like TEXTure~\cite{richardson2023texture} or EASI-Tex~\cite{perla2024easitex}.
Also, the input 3D meshes may lack a UV-map or contain multiple ones for textures, complicating the back-projection of predicted labels.
Consequently, we re-parameterize every mesh using \texttt{xatlas}\footnote{https://github.com/jpcy/xatlas} to have a single UV-map.
All our experiments were conducted on a single NVIDIA A100 GPU.

The weights for the losses $\alpha$, $\beta$, and $\gamma$ in Eq.~\ref{eq:l2} (of the main paper) are set to $1$ in both training phases, while $\delta$ is set to $0$ in the first phase and $0.005$ in the second phase.
We use a learning rate of $0.06$ in the first phase and $0.00008$ in the second phase.
The model is trained for $100$ epochs in each phase, totaling $200$ epochs.
We use LoRA~\cite{hu2021lora} with rank~$8$ and a scaling factor of~$1.0$ to fine-tune Stable Diffusion in the second phase.
We use a batch size of 2 while training and a batch size of 1 during inference.
The regularization weight $\lambda$ (Eq.~\ref{eq:noise_opt} of the main paper) is set to $0.1$, and the learning rate $\eta$ is set to $0.005$.
\sr{As evident from our ablation study (Tab.~\ref{tab:ablation_combined} of the main paper), the results are already strong even without applying \textit{Noise Optimization}. Nonetheless, when enabled, optimization helps extract the last bit of performance and further improves the results. In such cases, we limit \textit{Noise Optimization} to at most 5 epochs, as longer runs may occasionally degrade performance.}

Training a model takes approximately 3-4 hours, which is comparable to baseline methods. Inference without Noise Optimization requires less than a minute, while enabling optimization increases runtime to about five minutes.


\section{Additional Ablation Experiments}
\label{sup_sec:ablation_exps}

We present additional ablation experiments to further validate our design choices, conducted on the same object categories used for the ablations in the main paper.

\subsection{Varying the Number of Training Samples}
Tab.~\ref{tab:num_meshes} reports the results of an ablation study that varies the number of training samples. As expected, the model’s generalizability—and consequently its \textit{mIoU}—improves as more training data are provided. However, the gains become less pronounced as the number of samples continues to increase. This illustrates the diminishing returns of additional data and suggests that a relatively small number of examples may already provide sufficient supervision for effective~generalization.

\begin{table}[!ht]
\caption{
Ablation experiment studying the effect of varying the number of training samples.
}
\begin{tabular}{c|c}
\toprule
\#Shapes per category & mIoU ($\uparrow$) \\ 
\midrule
1                                 & 28.58           \\
4                                & 54.73           \\
8                                & \textbf{63.69}           \\ 
\bottomrule
\end{tabular}
\label{tab:num_meshes}
\end{table}






\section{Additional Results and Discussions}
\label{sup_sec:result_discussion}


\subsection{Results on Adaptive 3D Segmentation}
\label{sup_subsec:custom_seg_results}

Tab.~\ref{tab:custom_full_comp_2} provides additional comparisons between our approach and baseline methods~\cite{decatur20233d, abdelreheem2023satr} for adaptive 3D segmentation. As shown, our method consistently outperforms the baselines across all object categories.

The relatively lower performance of 3D Highlighter~\cite{decatur20233d} and SATR~\cite{abdelreheem2023satr}, compared to ASIA, is to be expected. Both are zero-shot methods that \textit{only} rely on text descriptions of parts for localization, and are not specifically designed for this task. Moreover, the object parts in our evaluation were deliberately chosen to be difficult to describe precisely with text, making the task particularly challenging for these baselines—
though this difficulty is precisely what motivated this work, \textit{ASIA}.

We were unable to evaluate against PartSLIP~\cite{liu2023partslip}, PartSTAD~\cite{kim2024partstad}, and PartDistill~\cite{umam2024partdistill}, as these methods require annotated 3D meshes for training, whereas our approach uses annotated in-the-wild images as reference. While $3\times2$~\cite{3by2} represents a promising baseline, its code was not open-sourced at the time of publication. The same limitation applies to MeshSegmenter~\cite{zhong2024meshsegmenter}.

\begin{figure*}[t]
  \includegraphics[width=0.99\textwidth]{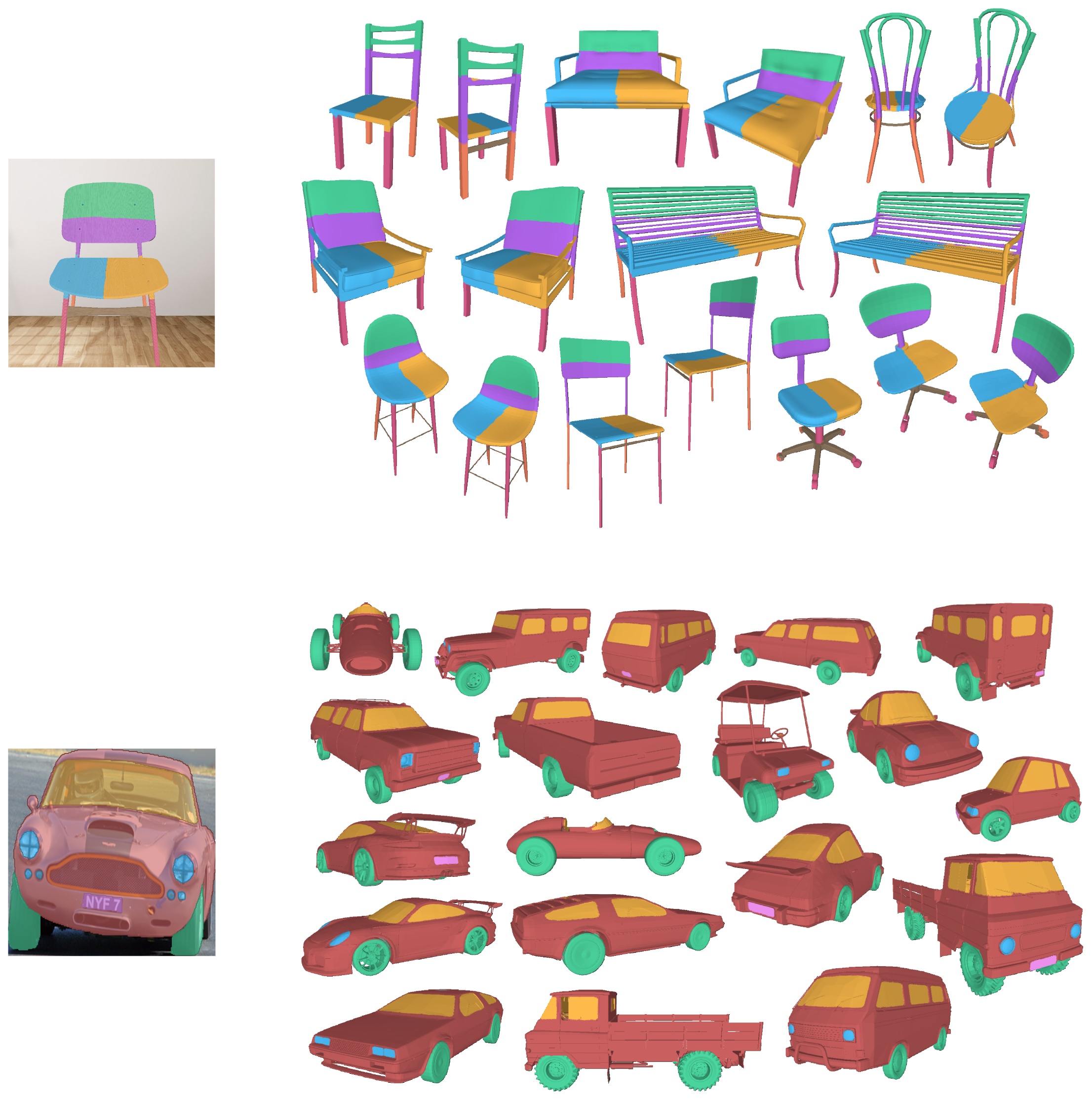}
  \caption{
  Qualitative results of our approach for adaptive 3D segmentation, with reference annotations on the left and generated results on the right.
  }
  \label{fig:custom_seg_1}
\end{figure*}

\begin{figure*}[t]
  \includegraphics[width=0.99\textwidth]{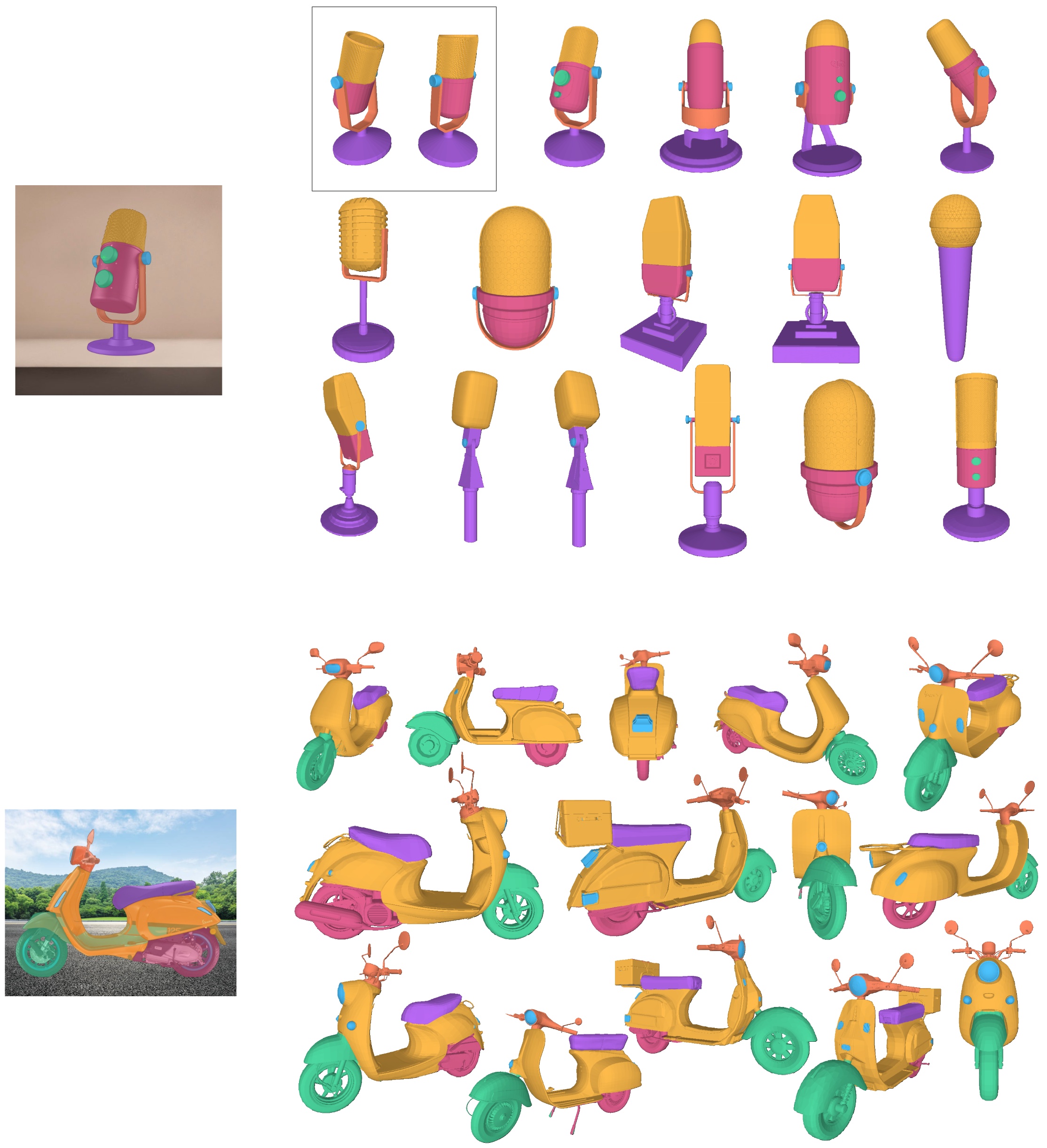}
  \caption{
  Qualitative results of our approach for adaptive 3D segmentation, with reference annotations on the left and generated results on the right.
  }
  \label{fig:custom_seg_2}
\end{figure*}

\begin{figure*}[t]
  \includegraphics[width=0.99\textwidth]{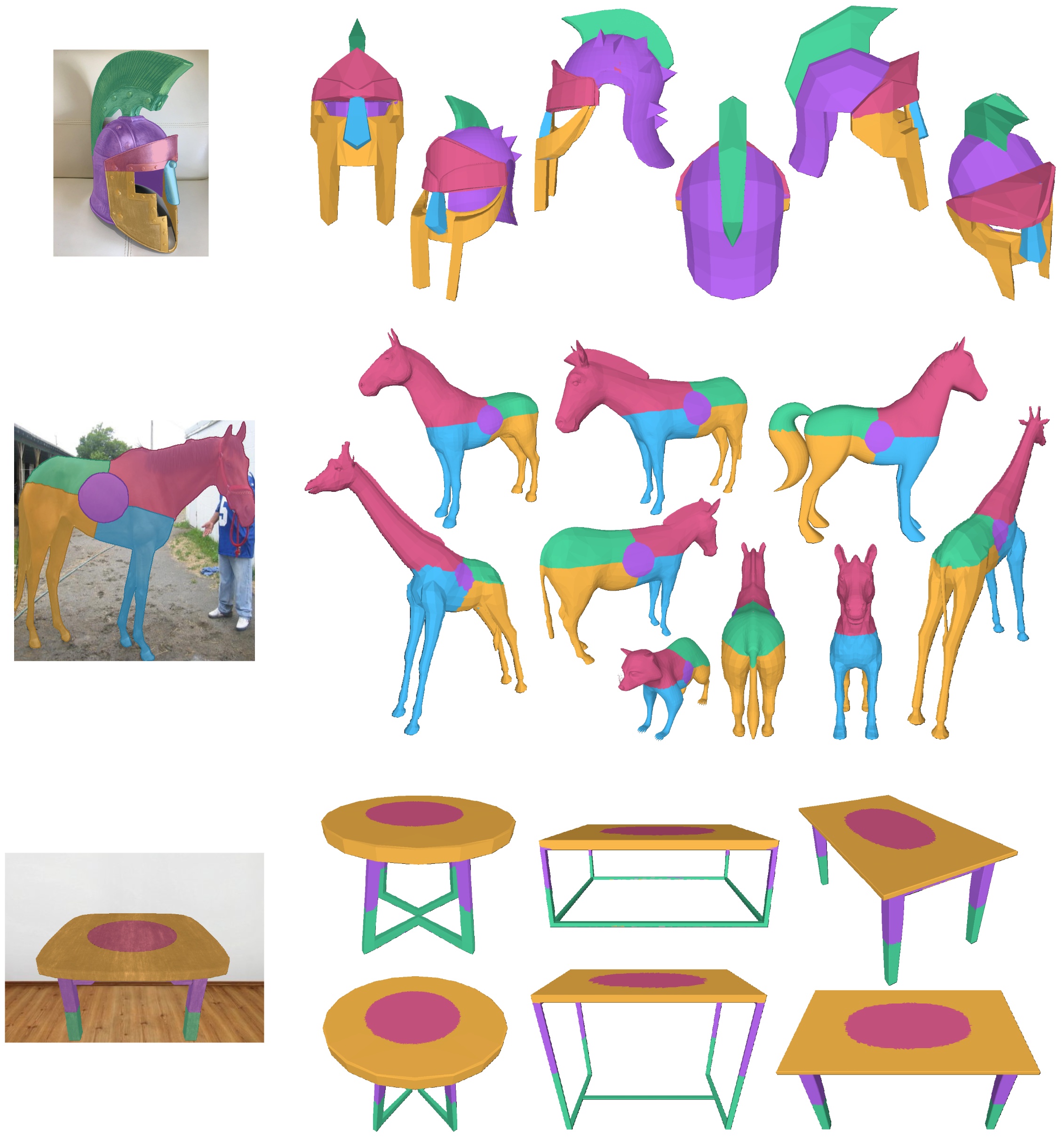}
  \caption{
  Qualitative results of our approach for adaptive 3D segmentation, with reference annotations on the left and generated results on the right.
  }
  \label{fig:custom_seg_3}
\end{figure*}

\begin{figure*}[t]
  \includegraphics[width=0.99\textwidth]{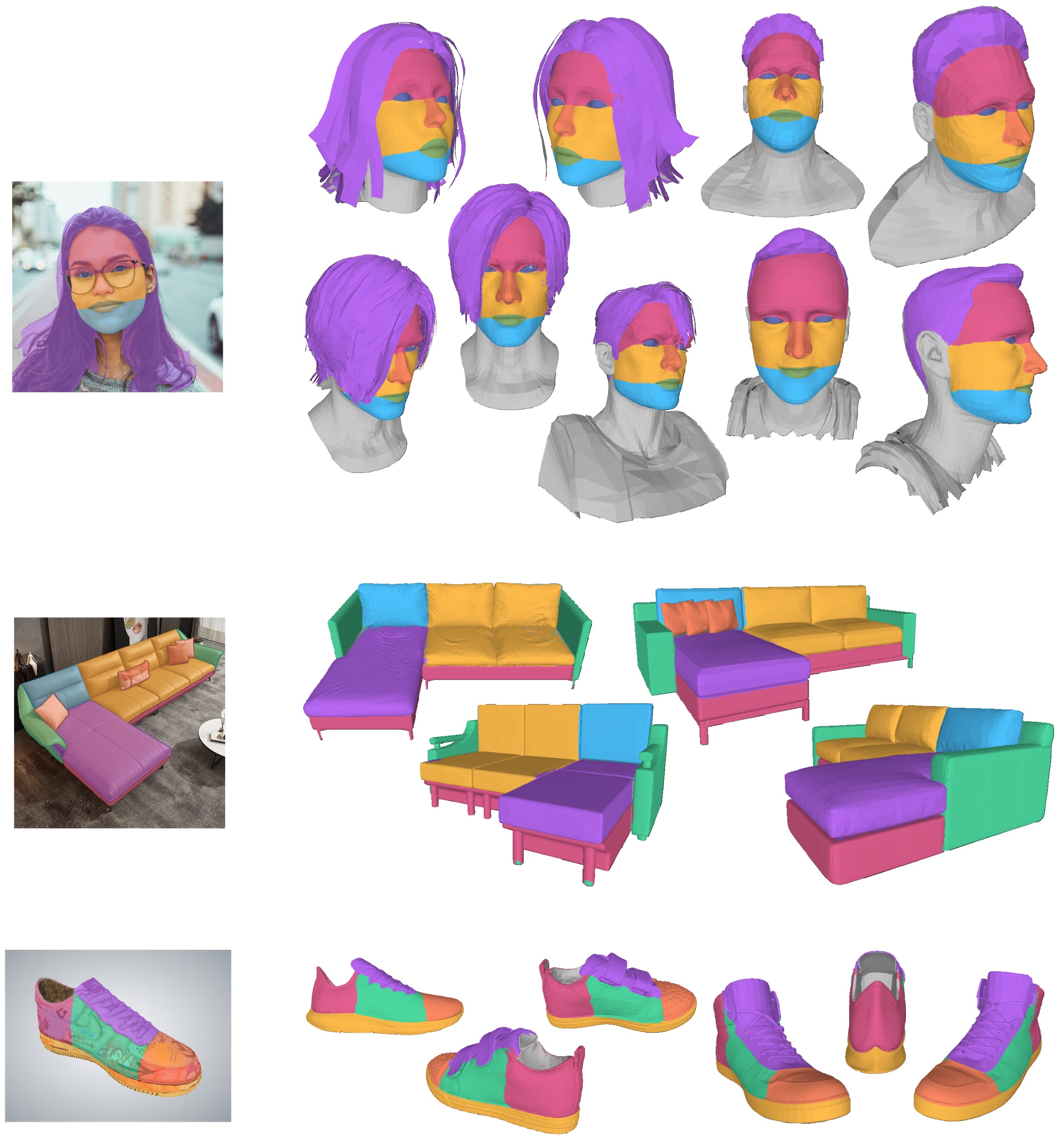}
  \caption{
  Qualitative results of our approach for adaptive 3D segmentation, with reference annotations on the left and generated results on the right.
  }
  \label{fig:custom_seg_4}
\end{figure*}

Finally, we present additional qualitative results from our approach in Fig.~\ref{fig:custom_seg_1}, Fig.~\ref{fig:custom_seg_2}, Fig.~\ref{fig:custom_seg_3}, and Fig.~\ref{fig:custom_seg_4}. These examples span a broad range of object categories, cover varying levels of segmentation difficulty, and include diverse samples within each category—highlighting the strong generalization capabilities of our model in a few-shot setting.

\subsection{Results on PartNetE Dataset}
\label{sup_subsec:pm_results}

Tab.~\ref{tab:supp_sem_seg_full_table} presents quantitative results for semantic segmentation across 40 categories of the PartNetE~\cite{liu2023partslip} dataset. Our approach outperforms the baselines in 30 out of 40 categories ($75\%$) by a significant margin, and performs competitively on the remaining 10 categories. Overall, it achieves the best performance across all categories, with a mean mIoU of \textbf{75.77$\%$}, marking a substantial improvement of \textbf{8.7} points (or $13\%$) over the best-performing baseline, PartDistill~\cite{umam2024partdistill}. Evaluation metrics for the baselines~\cite{liu2023partslip, kim2024partstad, umam2024partdistill, 3by2} are reported as provided in their respective~papers.

We also provide additional qualitative results for semantic segmentation in Tables~\ref{tab:PN_full_comp_2} and \ref{tab:PN_full_comp_3}. Our method consistently performs well even on small parts—such as buttons—where other approaches often fail. These visual results further support and reinforce our quantitative~findings.

We were unable to include qualitative comparisons with $3\times2$~\cite{3by2}, as its code was not open-sourced at the time of publication. For PartDistill~\cite{umam2024partdistill}, we could not run the provided code due to insufficient documentation and instructions.

\section{Part Composition}
\label{sup_sec:part composition}
\begin{figure}[h]
    \includegraphics[width=0.95\linewidth]{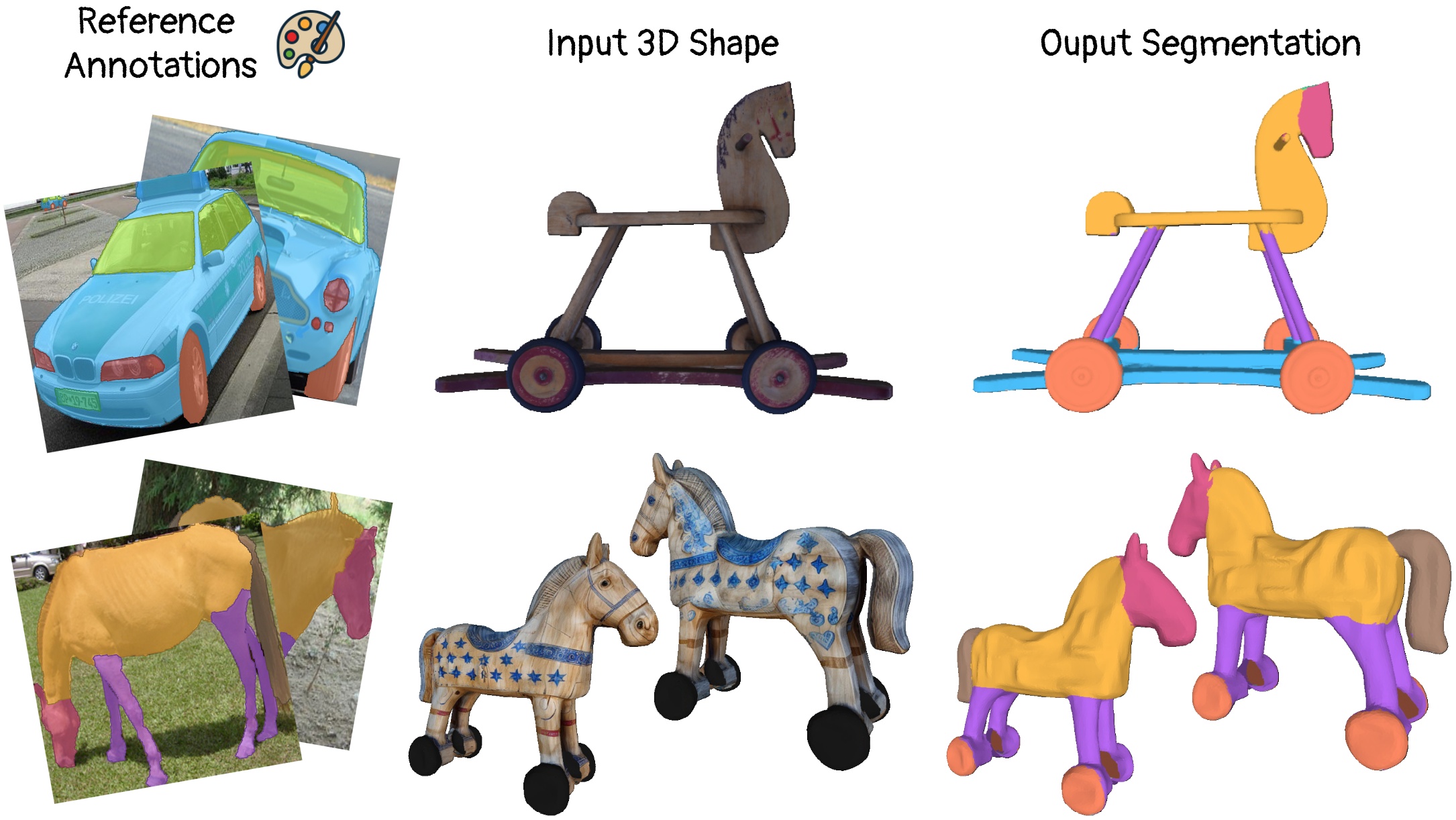}
  \caption{
  \textit{Part Composition.} During training (left), we learn disjoint sets of parts from images of \textit{cars} and \textit{horses}. At inference (center and right), the model segments \emph{3D ponycycles} that combines parts from both.
  }
  \label{fig:part_comp}
\end{figure}

In this section, we showcase additional capabilities of \PaperName{} on tasks such as \textit{Part Composition}, which makes our approach practical and convenient for a variety of applications. Our approach learns object parts from different image sets, each containing disjoint subsets of parts. At inference, however, it can segment objects that combine parts it has never observed co-occurring during training. For instance, in Fig.~\ref{fig:part_comp}, the model learns object parts from images of \emph{cars} and \emph{horses}, and at inference segments \emph{3D ponycycles} containing parts from both. These results further highlight the model’s ability to generalize effectively to objects that differ substantially from the training images.

\begin{table*}[!ht]


\centering
\caption{
Qualitative comparison of results from 3D Highlighter~\cite{decatur20233d}, SATR~\cite{abdelreheem2023satr}, and Ours for adaptive 3D segmentation. Each row shows the reference annotation followed by two views of the same object.
}
\label{tab:custom_full_comp_2}

\begin{tabular}{@{}p{0.1171\textwidth}@{} | @{}p{0.44\textwidth}@{} | @{}p{0.44\textwidth}@{}}

\multicolumn{1}{c}{\textbf{}} &
\multicolumn{1}{c}{\textbf{View 1}} & \multicolumn{1}{c}{\textbf{View 2}} \\

\vtop{\vskip0pt
\resizebox{0.1171\textwidth}{!}{
\begin{tabular}{@{}c@{}}
\toprule
\small{Reference} \\ \midrule
\includegraphics[width=0.1\textwidth]{images/viz_custom/annotated/chair_annotated.jpg} \\

\addlinespace[-2pt]
\arrayrulecolor{gray}\cmidrule(lr){1-1}
\arrayrulecolor{black}

\includegraphics[width=0.1\textwidth]{images/viz_custom/annotated/faces_2_annotated.jpg} \\

\addlinespace[-2pt]
\arrayrulecolor{gray}\cmidrule(lr){1-1}
\arrayrulecolor{black}

\includegraphics[width=0.1\textwidth]{images/viz_custom/annotated/gladiator_helmet_annotated.jpg} \\

\addlinespace[-2pt]
\arrayrulecolor{gray}\cmidrule(lr){1-1}
\arrayrulecolor{black}

\includegraphics[width=0.1\textwidth]{images/viz_custom/annotated/microphone_annotated.jpg} \\

\addlinespace[-2pt]
\arrayrulecolor{gray}\cmidrule(lr){1-1}
\arrayrulecolor{black}

\includegraphics[width=0.1\textwidth]{images/viz_custom/annotated/piston_annotated.jpg} \\

\addlinespace[-2pt]
\arrayrulecolor{gray}\cmidrule(lr){1-1}
\arrayrulecolor{black}

\includegraphics[width=0.1\textwidth]{images/viz_custom/annotated/scooter_annotated.jpg} \\

\addlinespace[-2pt]
\arrayrulecolor{gray}\cmidrule(lr){1-1}
\arrayrulecolor{black}

\includegraphics[width=0.1\textwidth]{images/viz_custom/annotated/shoe_annotated.jpg} \\

\addlinespace[-2pt]
\arrayrulecolor{gray}\cmidrule(lr){1-1}
\arrayrulecolor{black}

\includegraphics[width=0.1\textwidth]{images/viz_custom/annotated/sofa_annotated.jpg} \\

\addlinespace[-2pt]
\arrayrulecolor{gray}\cmidrule(lr){1-1}
\arrayrulecolor{black}

\includegraphics[width=0.1\textwidth]{images/viz_custom/annotated/table_annotated.jpg} \\

\addlinespace[-2pt]
\bottomrule
\end{tabular}
}}
&
\vtop{\vskip0pt
\resizebox{0.44\textwidth}{!}{
\begin{tabular}{@{}c@{}c@{}c@{}c@{}c@{}}
\toprule
\small{Input} & \small{3D Highlighter} & \small{SATR} & \small{Ours} \\ \midrule
\includegraphics[width=0.1\textwidth]{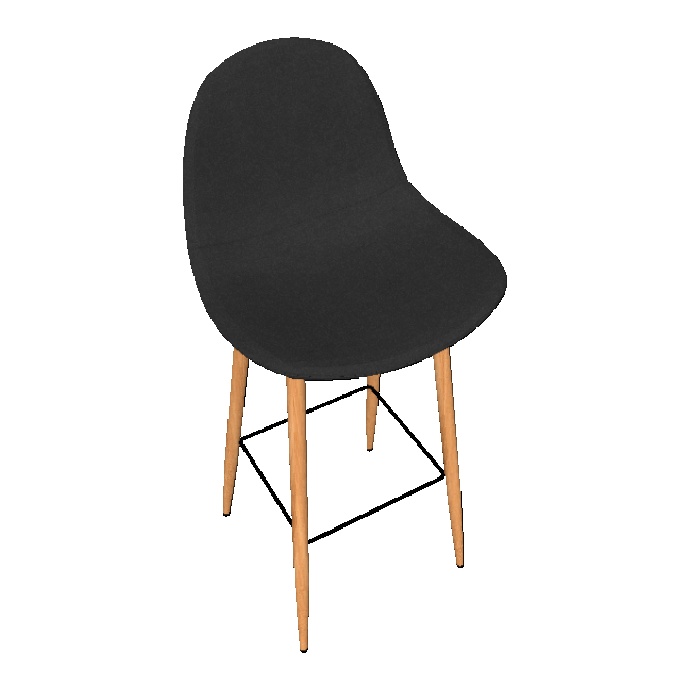} &
\includegraphics[width=0.1\textwidth]{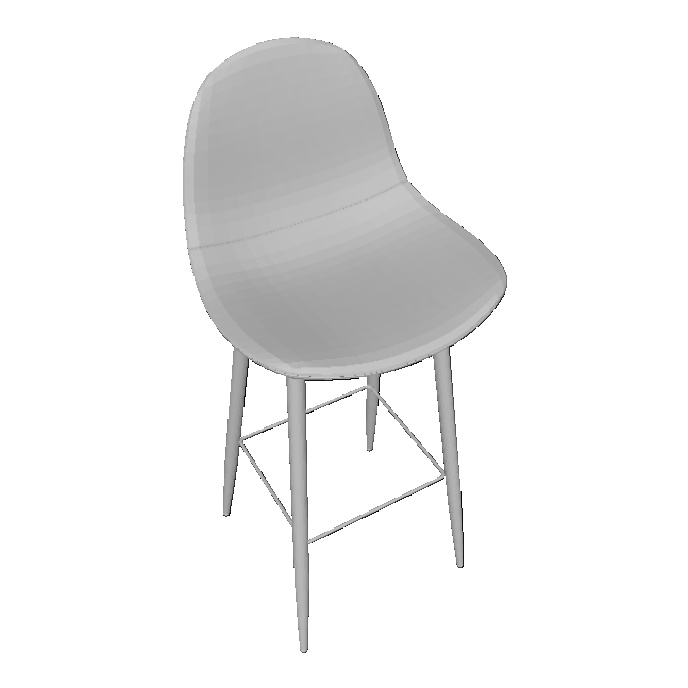} &
\includegraphics[width=0.1\textwidth]{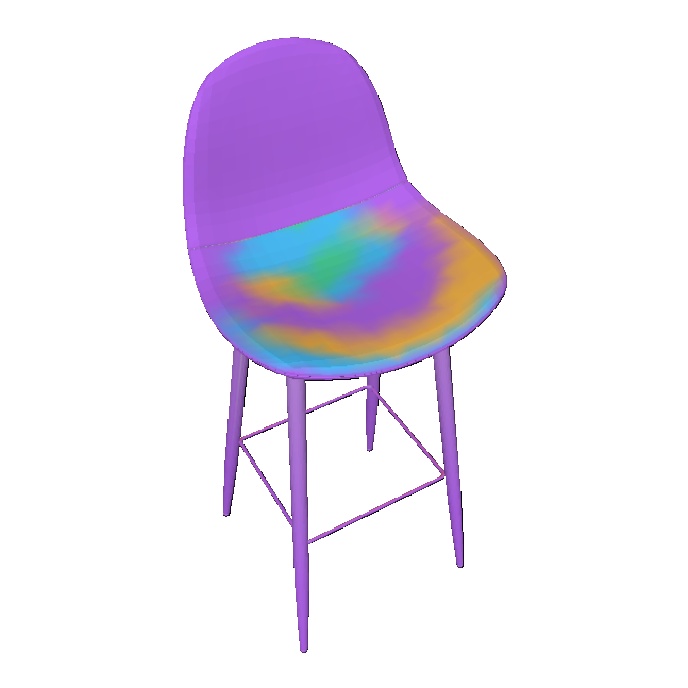} &
\includegraphics[width=0.1\textwidth]{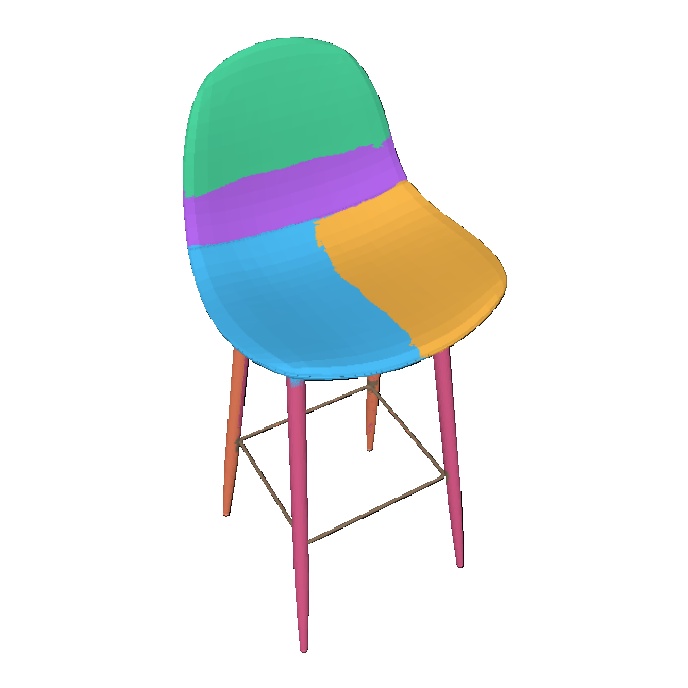} \\

\addlinespace[-2pt]
\arrayrulecolor{gray}\cmidrule(lr){1-4}
\arrayrulecolor{black}

\includegraphics[width=0.1\textwidth]{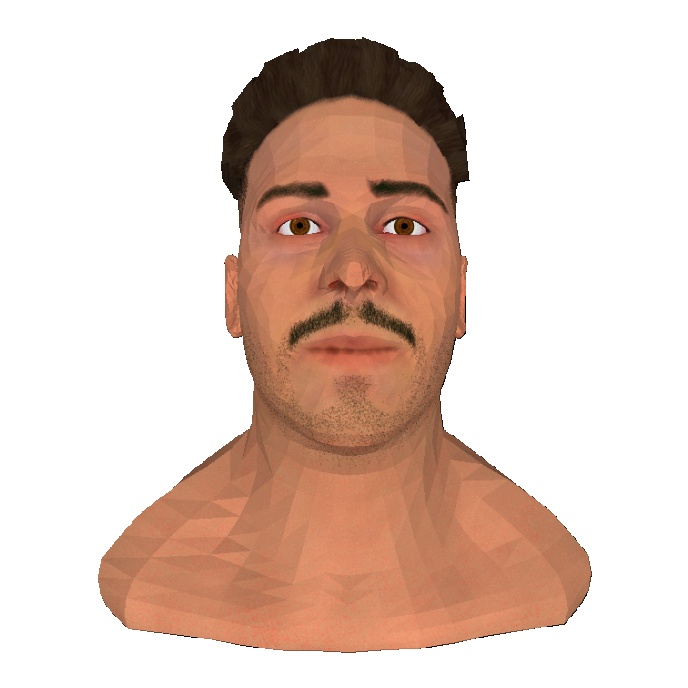} &
\includegraphics[width=0.1\textwidth]{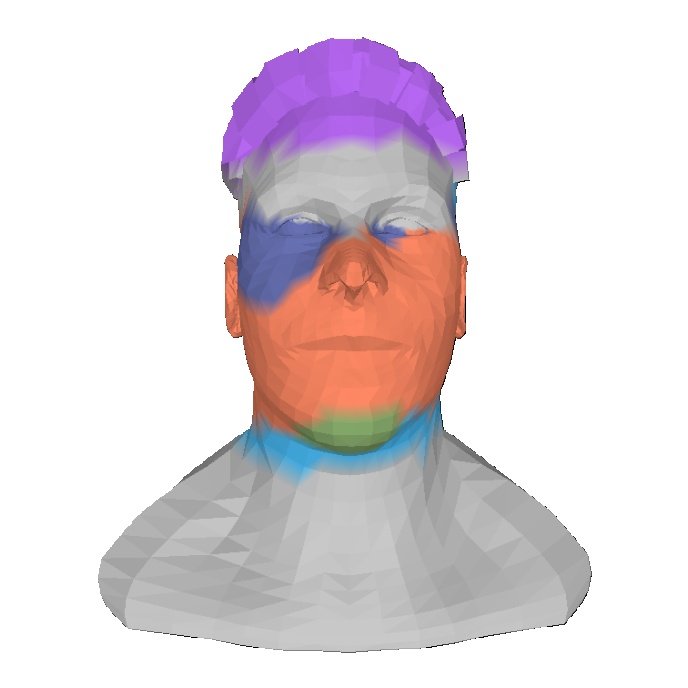} &
\includegraphics[width=0.1\textwidth]{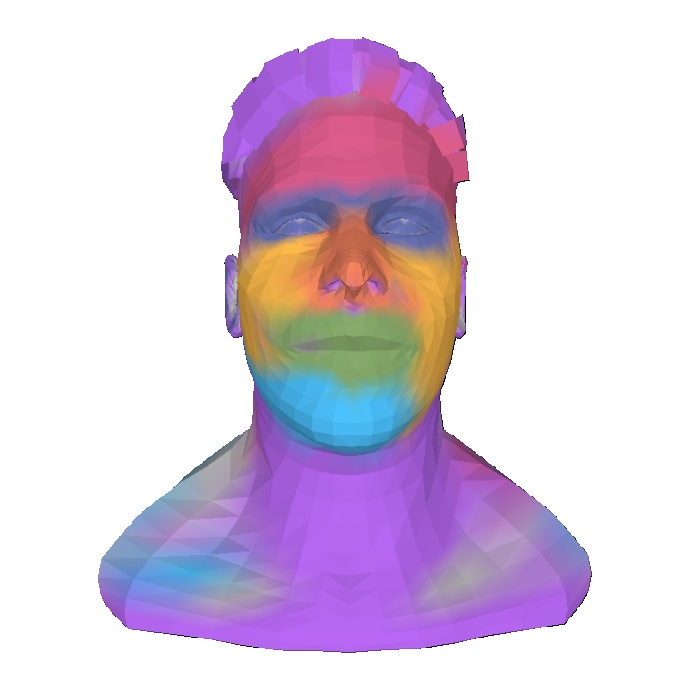} &
\includegraphics[width=0.1\textwidth]{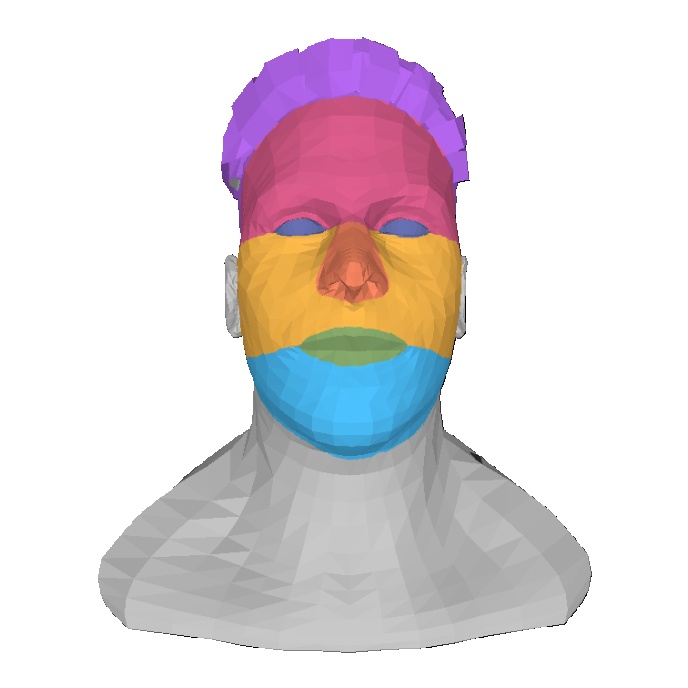} \\

\addlinespace[-2pt]
\arrayrulecolor{gray}\cmidrule(lr){1-4}
\arrayrulecolor{black}

\includegraphics[width=0.1\textwidth]{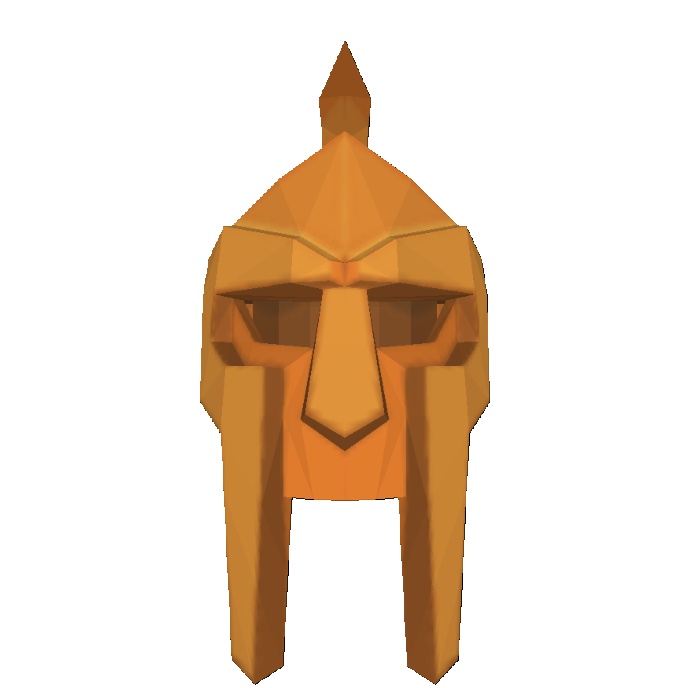} &
\includegraphics[width=0.1\textwidth]{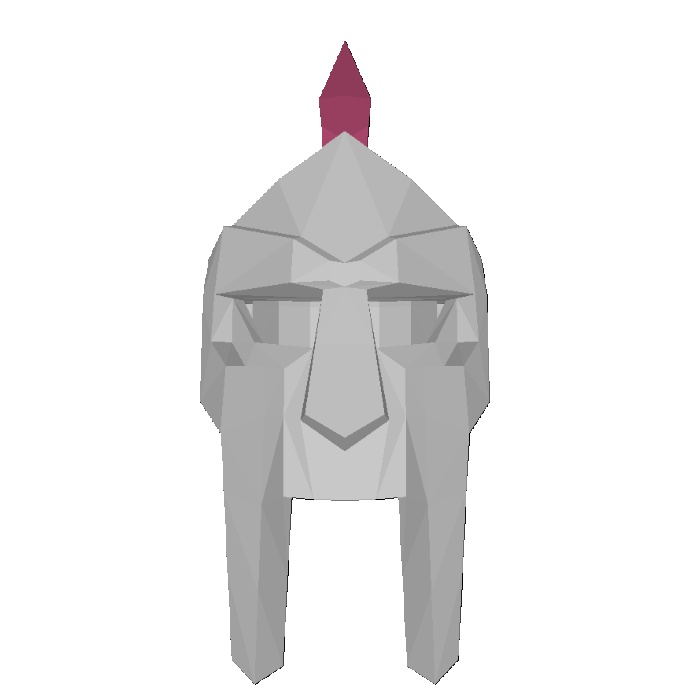} &
\includegraphics[width=0.1\textwidth]{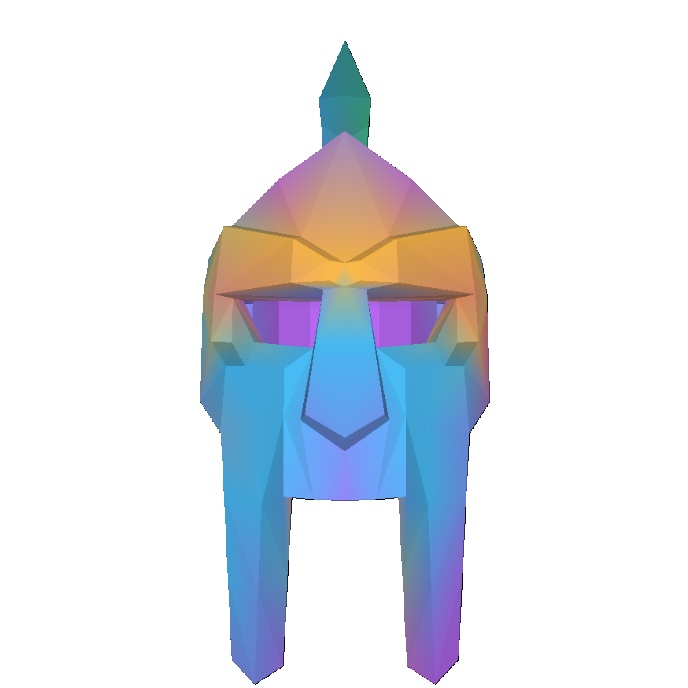} &
\includegraphics[width=0.1\textwidth]{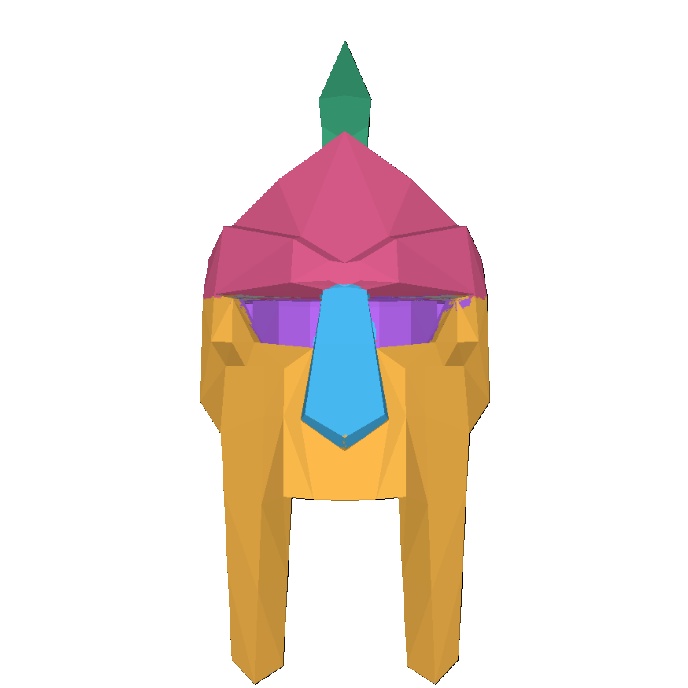} \\

\addlinespace[-2pt]
\arrayrulecolor{gray}\cmidrule(lr){1-4}
\arrayrulecolor{black}

\includegraphics[width=0.1\textwidth]{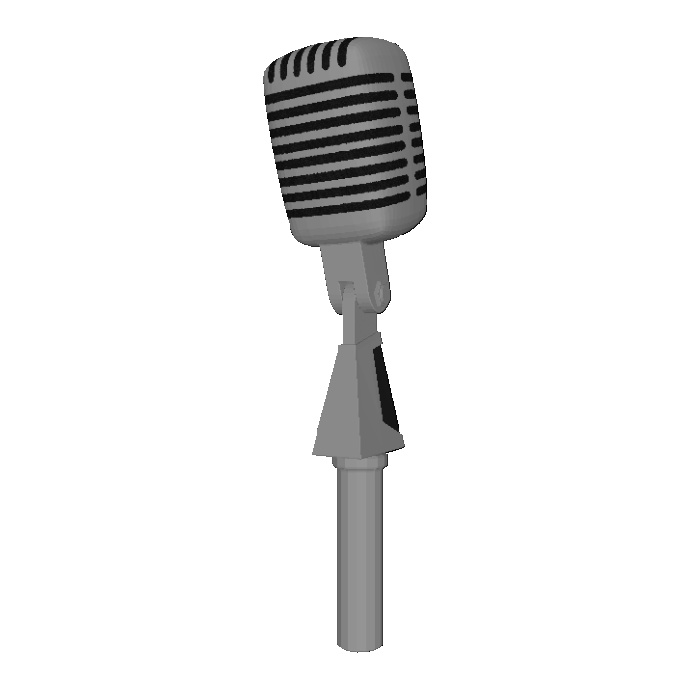} &
\includegraphics[width=0.1\textwidth]{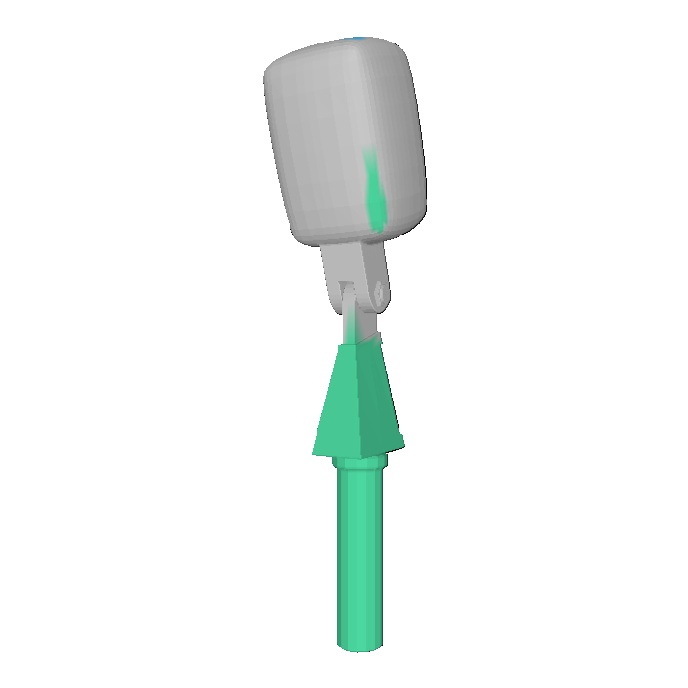} &
\includegraphics[width=0.1\textwidth]{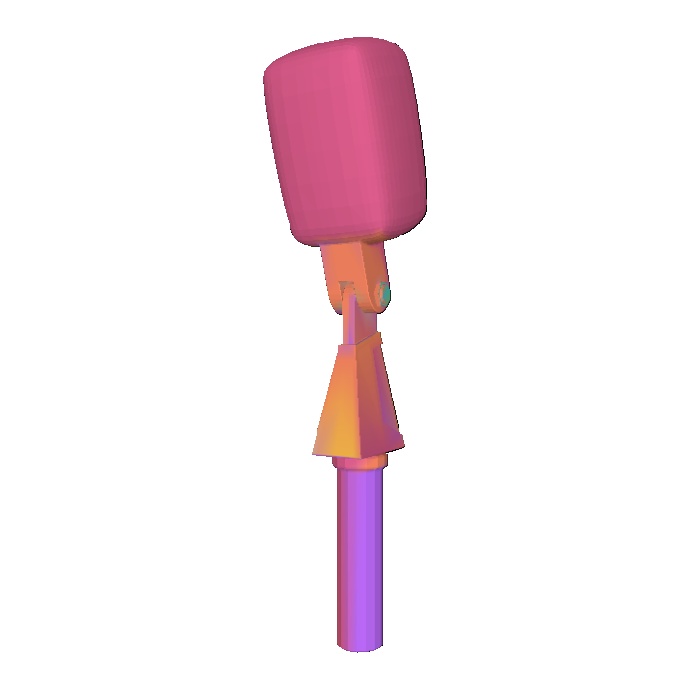} &
\includegraphics[width=0.1\textwidth]{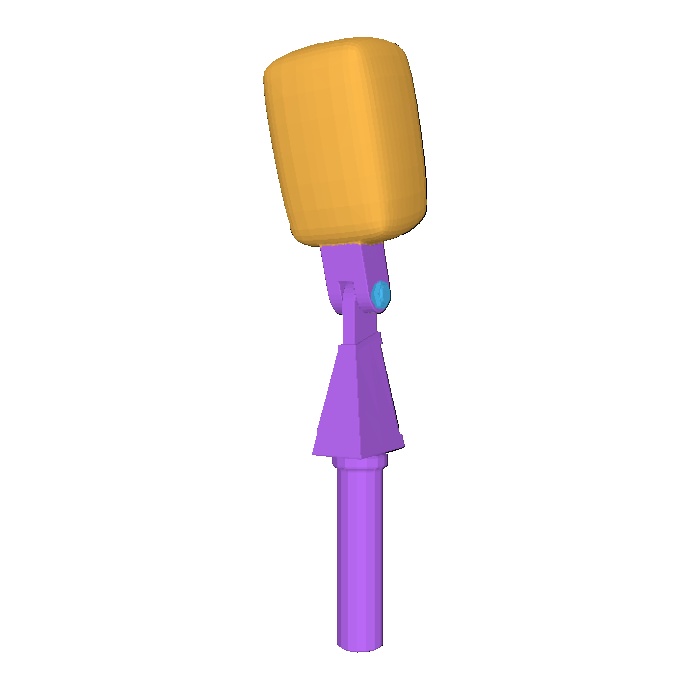} \\

\addlinespace[-2pt]
\arrayrulecolor{gray}\cmidrule(lr){1-4}
\arrayrulecolor{black}

\includegraphics[width=0.1\textwidth]{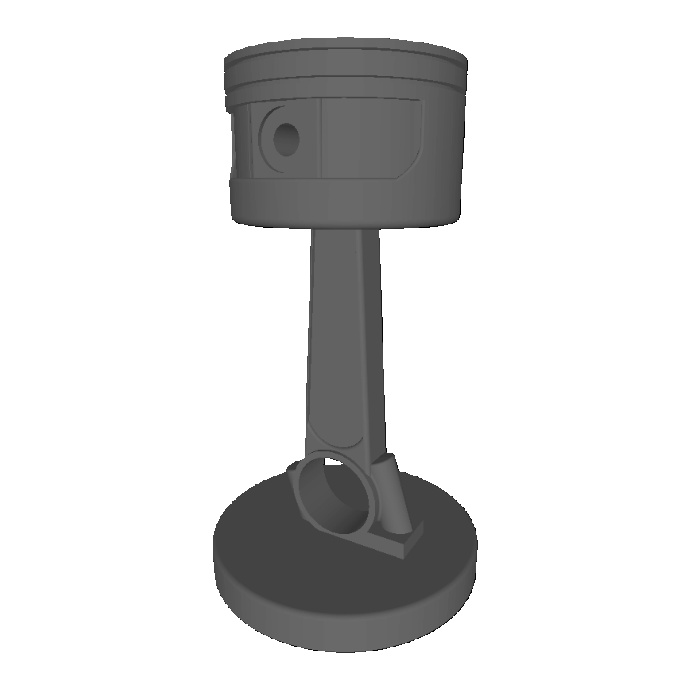} &
\includegraphics[width=0.1\textwidth]{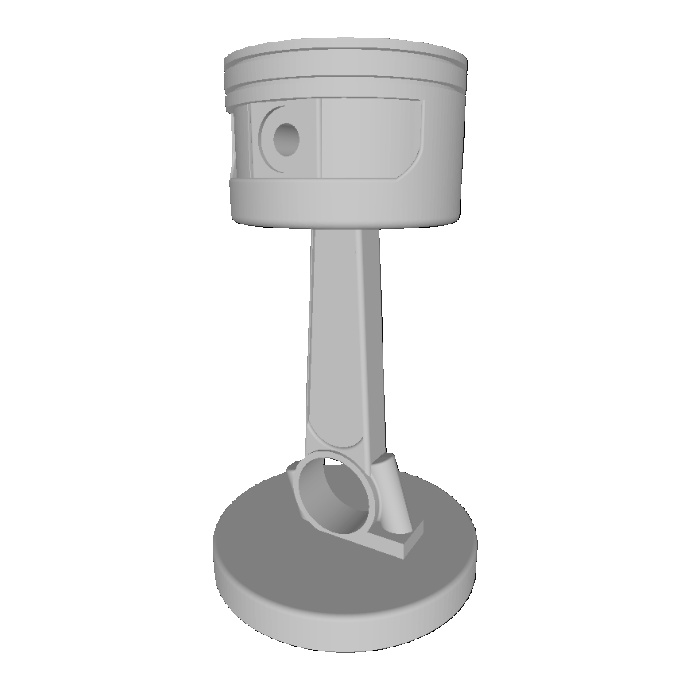} &
\includegraphics[width=0.1\textwidth]{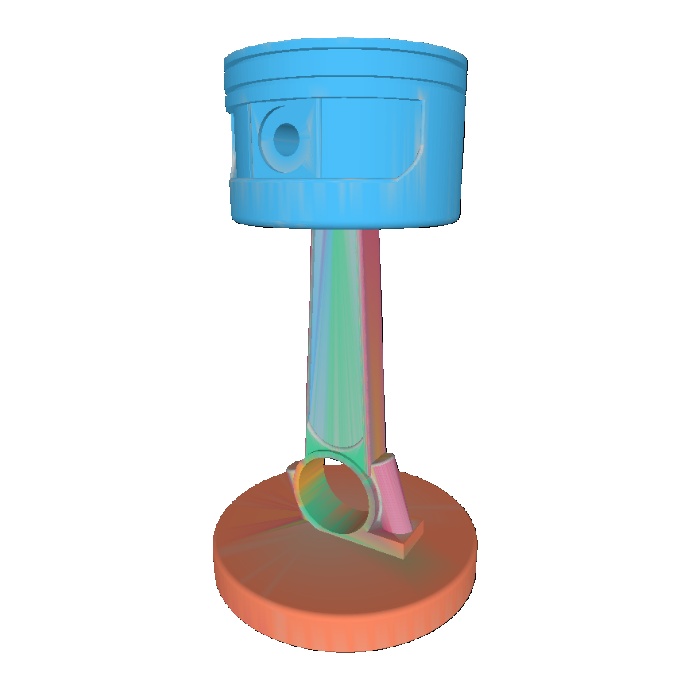} &
\includegraphics[width=0.1\textwidth]{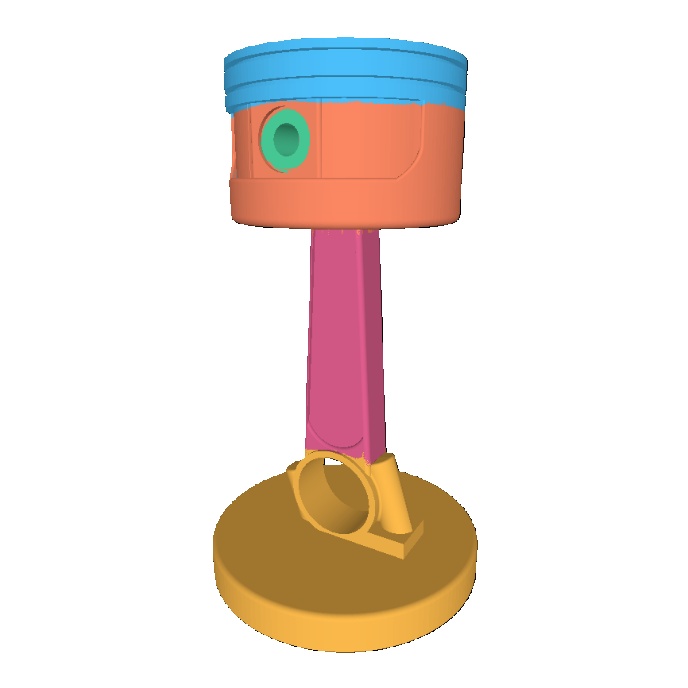} \\

\addlinespace[-2pt]
\arrayrulecolor{gray}\cmidrule(lr){1-4}
\arrayrulecolor{black}

\includegraphics[width=0.1\textwidth]{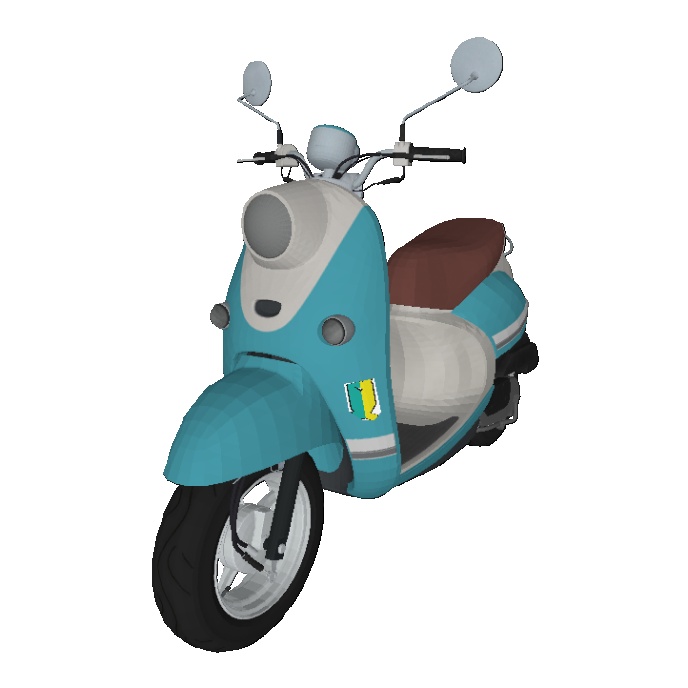} &
\includegraphics[width=0.1\textwidth]{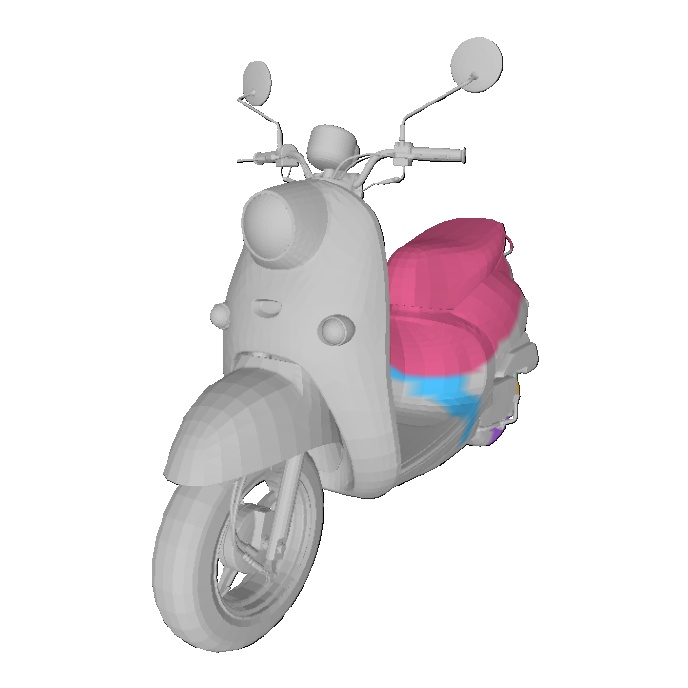} &
\includegraphics[width=0.1\textwidth]{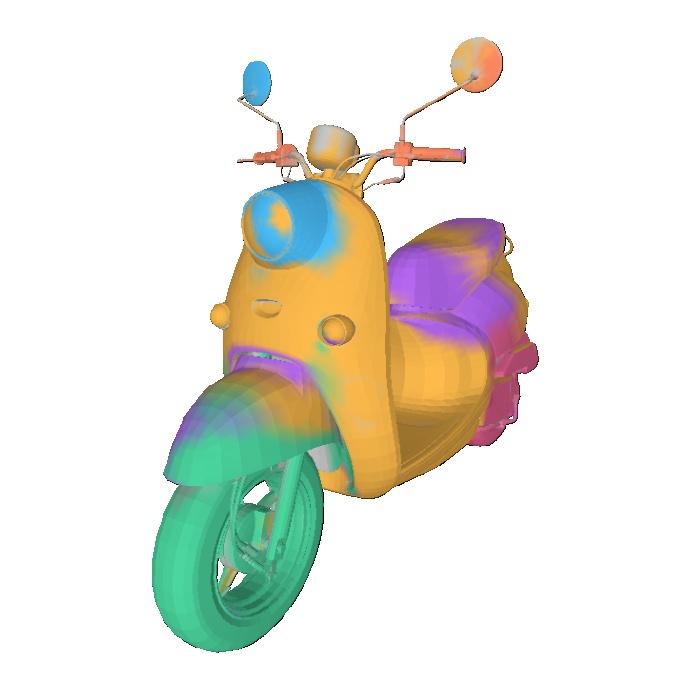} &
\includegraphics[width=0.1\textwidth]{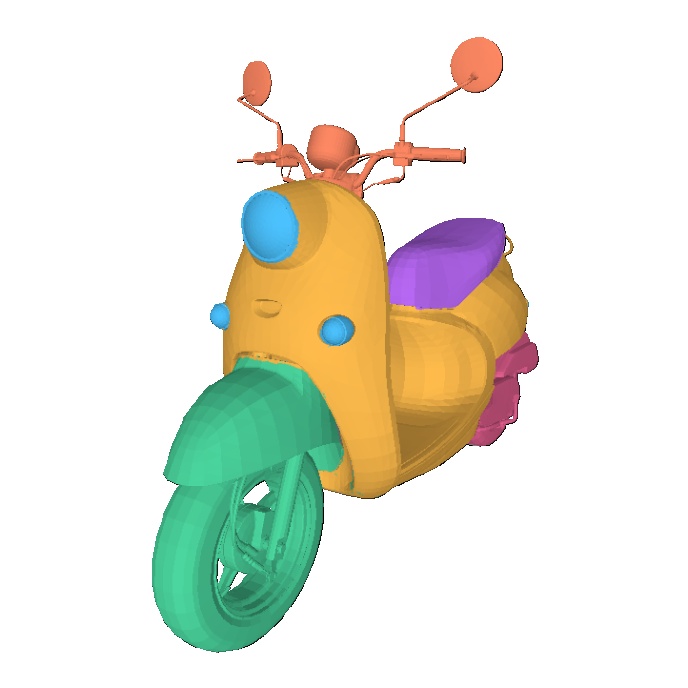} \\

\addlinespace[-2pt]
\arrayrulecolor{gray}\cmidrule(lr){1-4}
\arrayrulecolor{black}

\includegraphics[width=0.1\textwidth]{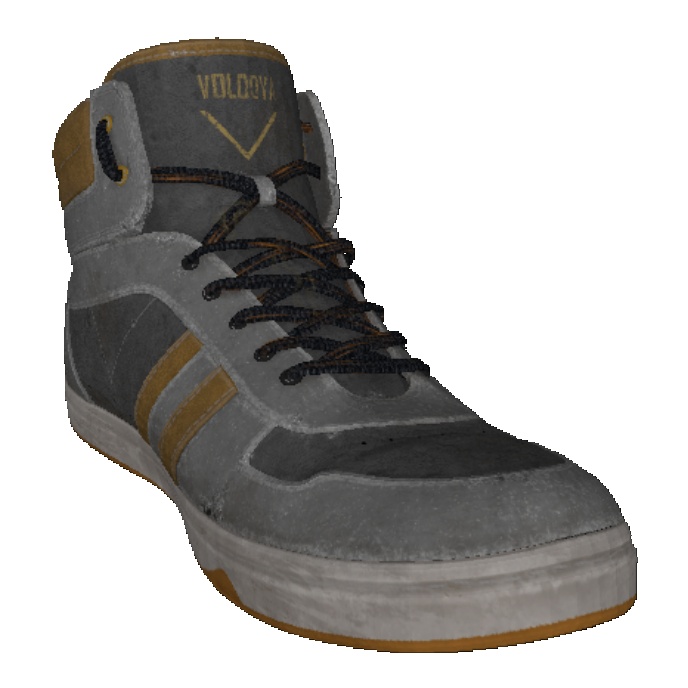} &
\includegraphics[width=0.1\textwidth]{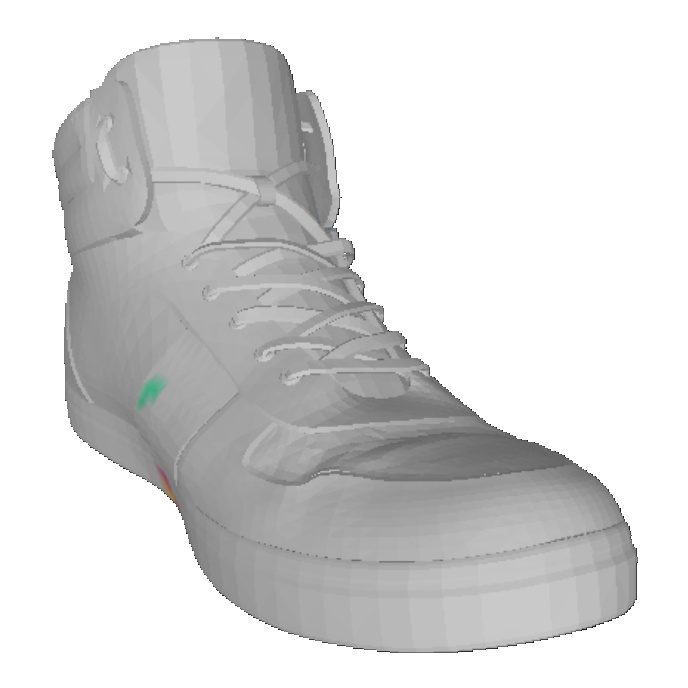} &
\includegraphics[width=0.1\textwidth]{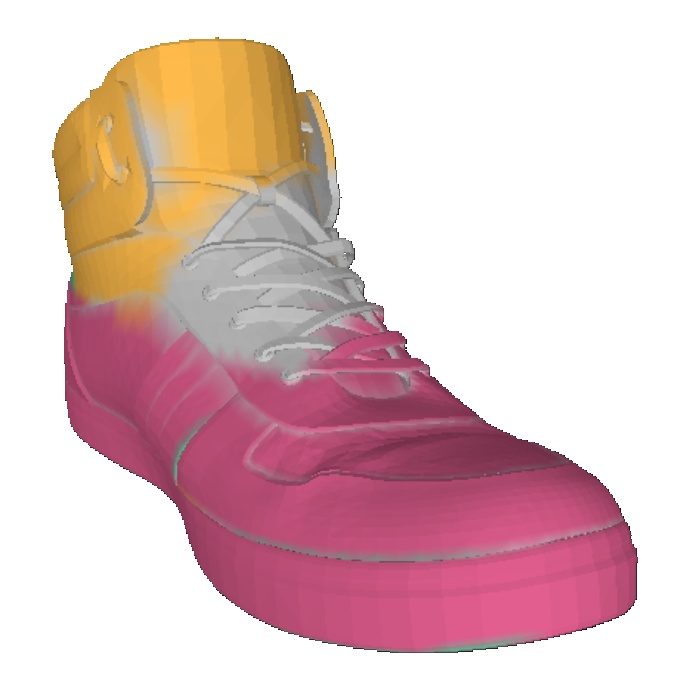} &
\includegraphics[width=0.1\textwidth]{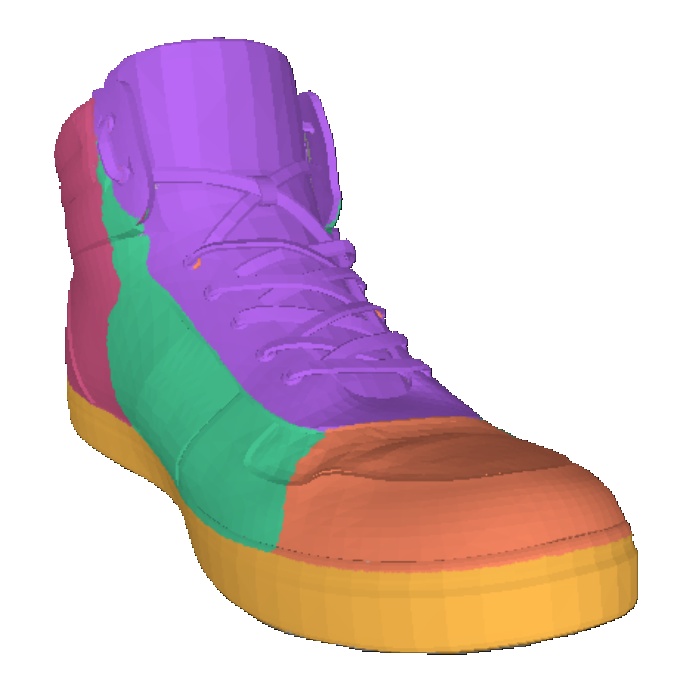} \\

\addlinespace[-2pt]
\arrayrulecolor{gray}\cmidrule(lr){1-4}
\arrayrulecolor{black}

\includegraphics[width=0.1\textwidth]{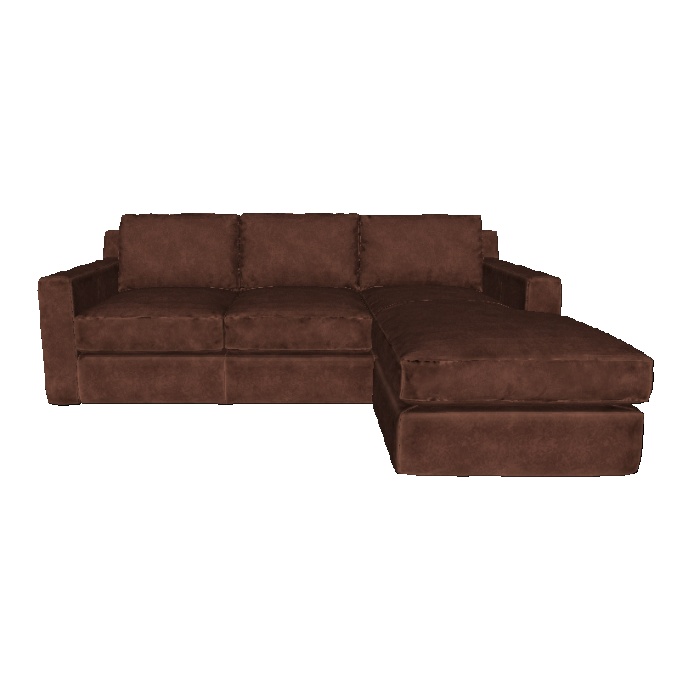} &
\includegraphics[width=0.1\textwidth]{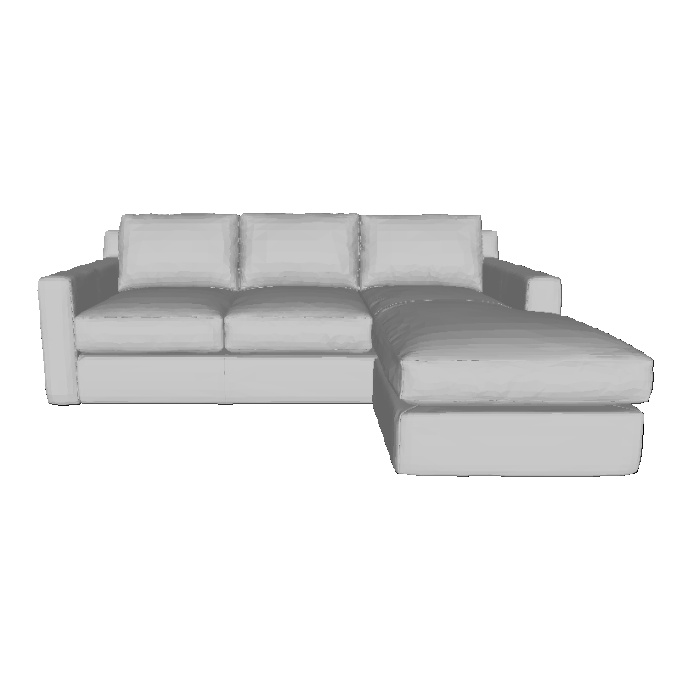} &
\includegraphics[width=0.1\textwidth]{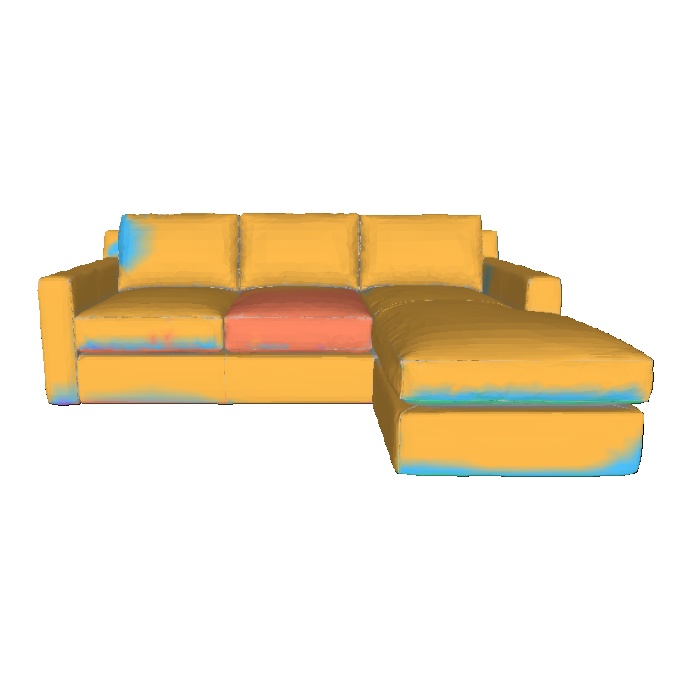} &
\includegraphics[width=0.1\textwidth]{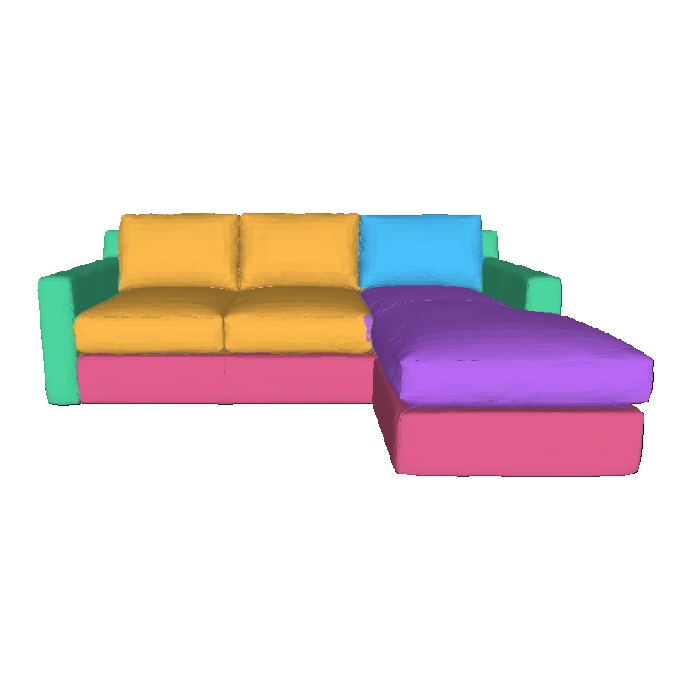} \\

\addlinespace[-2pt]
\arrayrulecolor{gray}\cmidrule(lr){1-4}
\arrayrulecolor{black}

\includegraphics[width=0.1\textwidth]{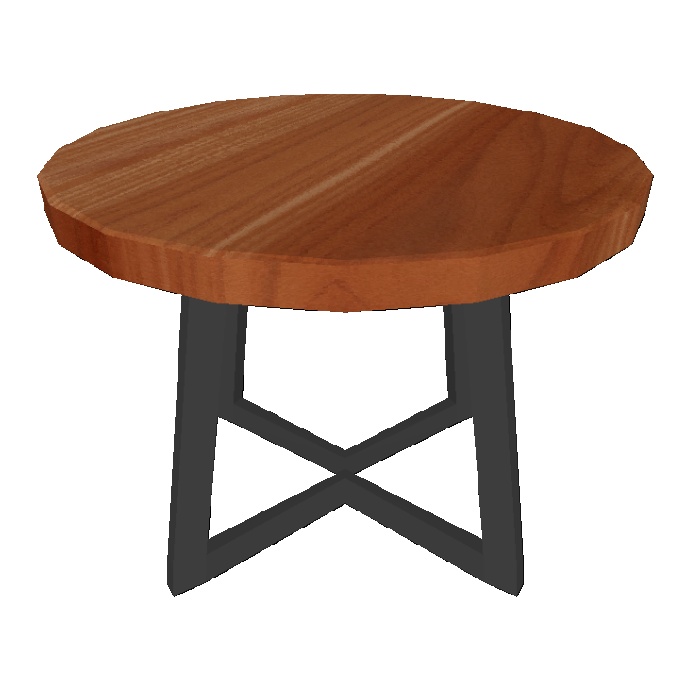} &
\includegraphics[width=0.1\textwidth]{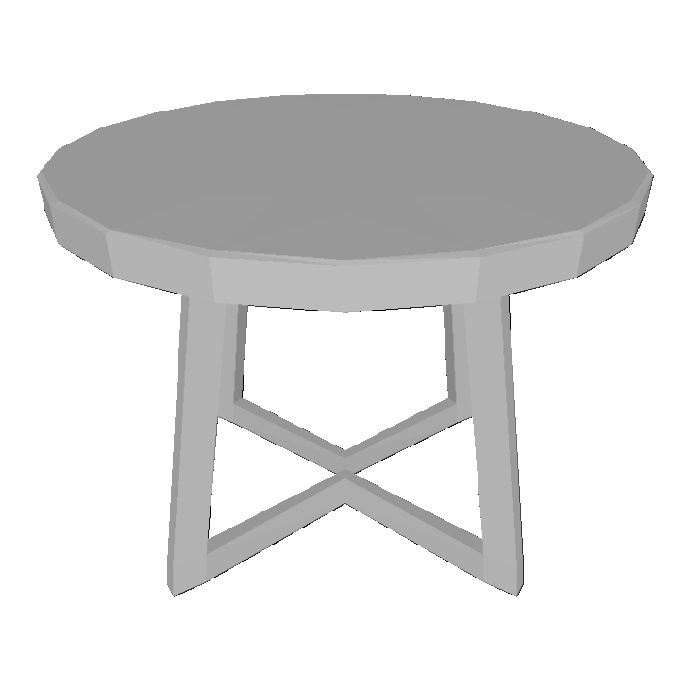} &
\includegraphics[width=0.1\textwidth]{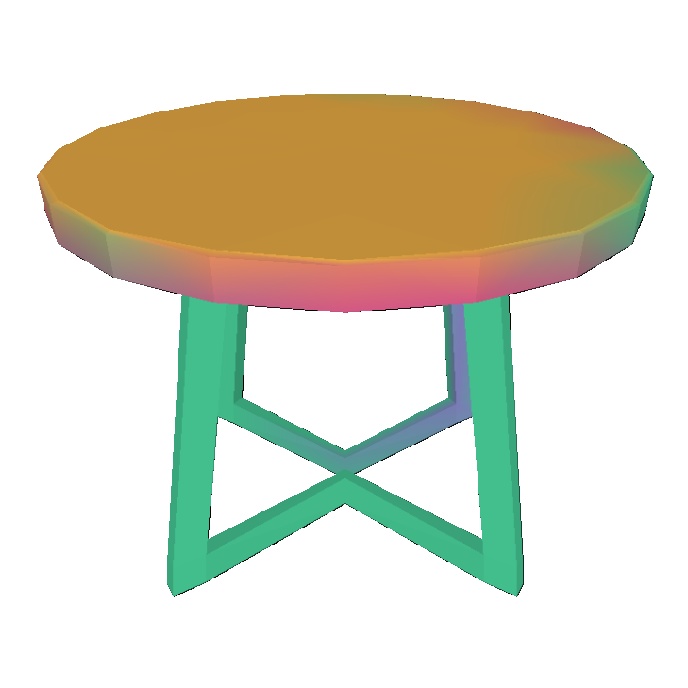} &
\includegraphics[width=0.1\textwidth]{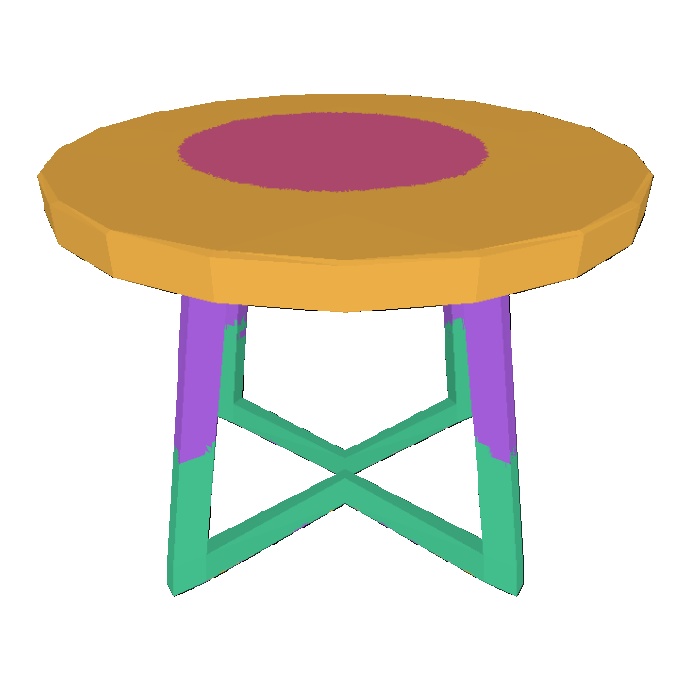} \\

\addlinespace[-2pt]
\bottomrule
\end{tabular}
}}
&
\vtop{\vskip0pt
\resizebox{0.44\textwidth}{!}{
\begin{tabular}{@{}c@{}c@{}c@{}c@{}c@{}}
\toprule
\small{Input} & \small{3D Highlighter} & \small{SATR} & \small{Ours} \\ \midrule
\includegraphics[width=0.1\textwidth]{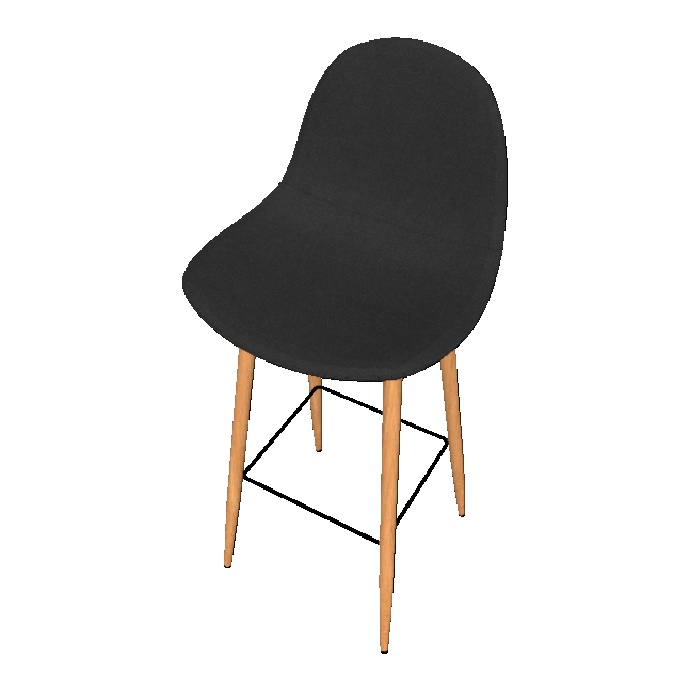} &
\includegraphics[width=0.1\textwidth]{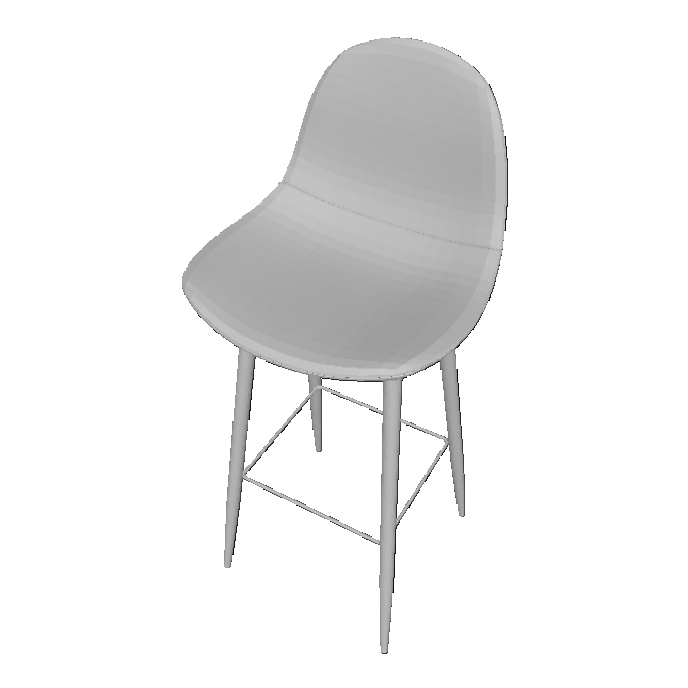} &
\includegraphics[width=0.1\textwidth]{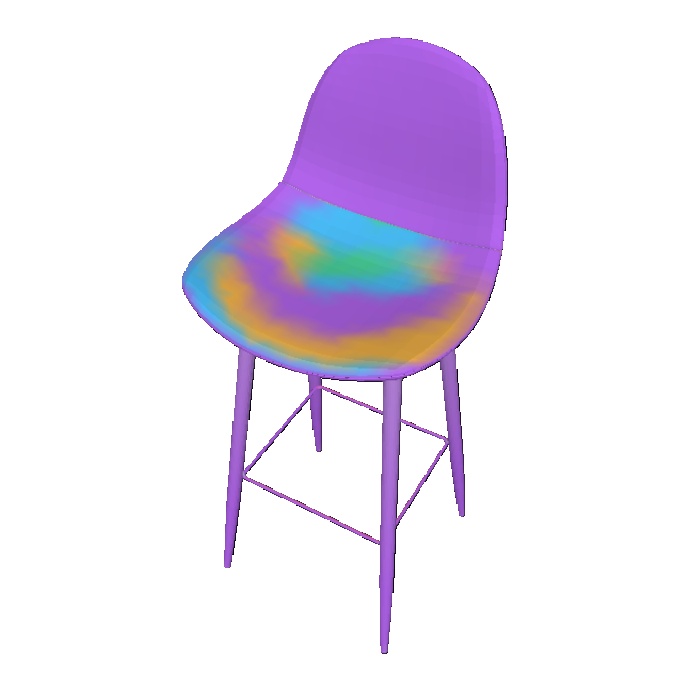} &
\includegraphics[width=0.1\textwidth]{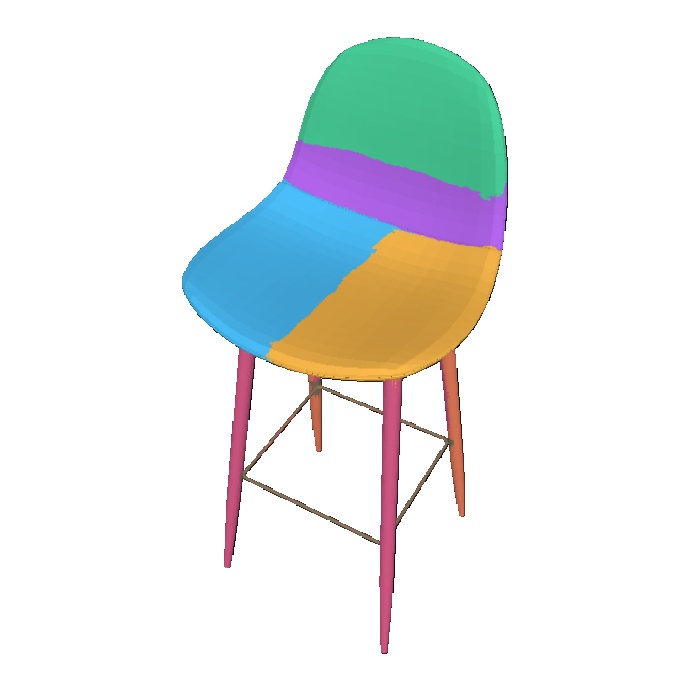} \\

\addlinespace[-2pt]
\arrayrulecolor{gray}\cmidrule(lr){1-4}
\arrayrulecolor{black}

\includegraphics[width=0.1\textwidth]{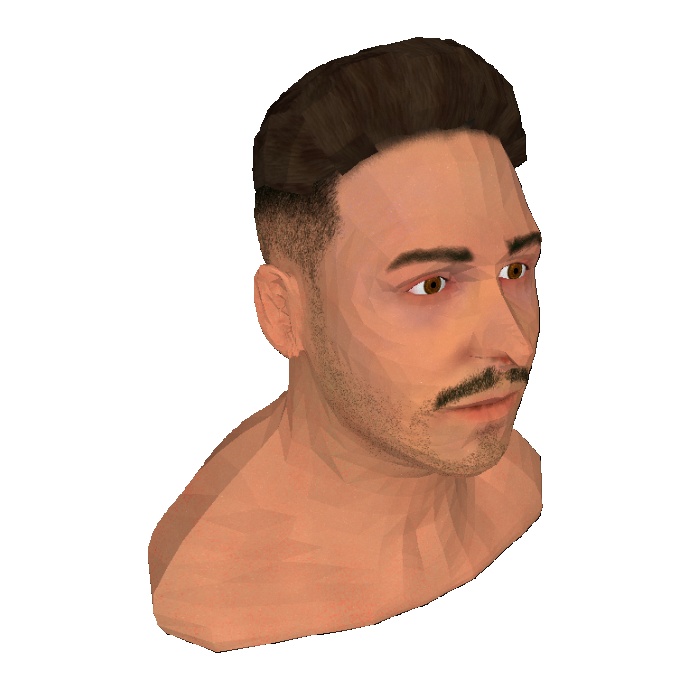} &
\includegraphics[width=0.1\textwidth]{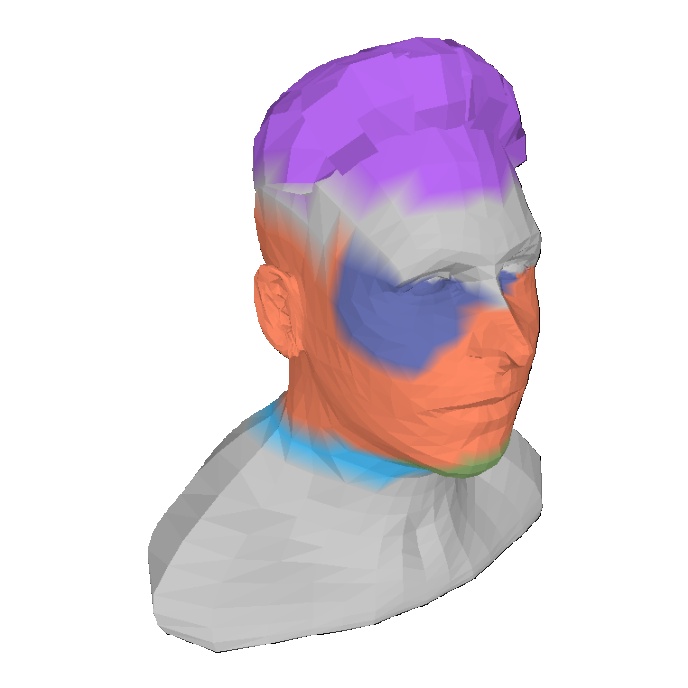} &
\includegraphics[width=0.1\textwidth]{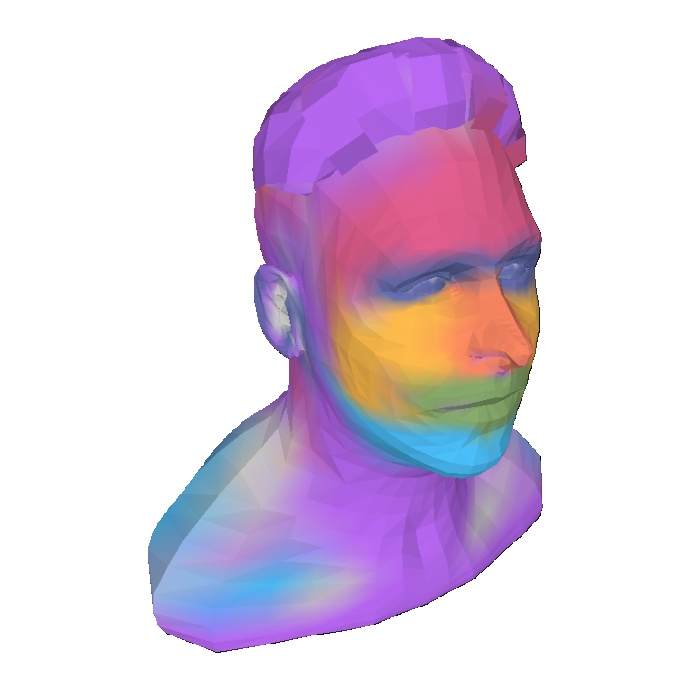} &
\includegraphics[width=0.1\textwidth]{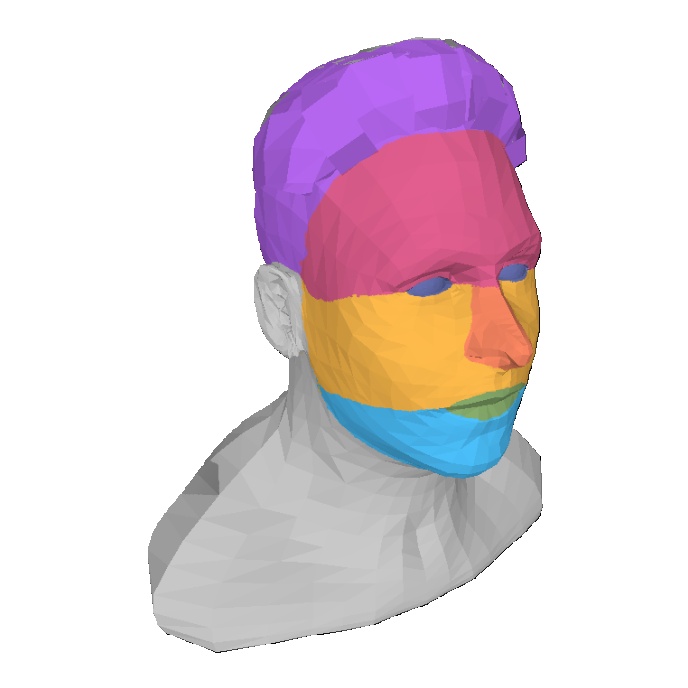} \\

\addlinespace[-2pt]
\arrayrulecolor{gray}\cmidrule(lr){1-4}
\arrayrulecolor{black}

\includegraphics[width=0.1\textwidth]{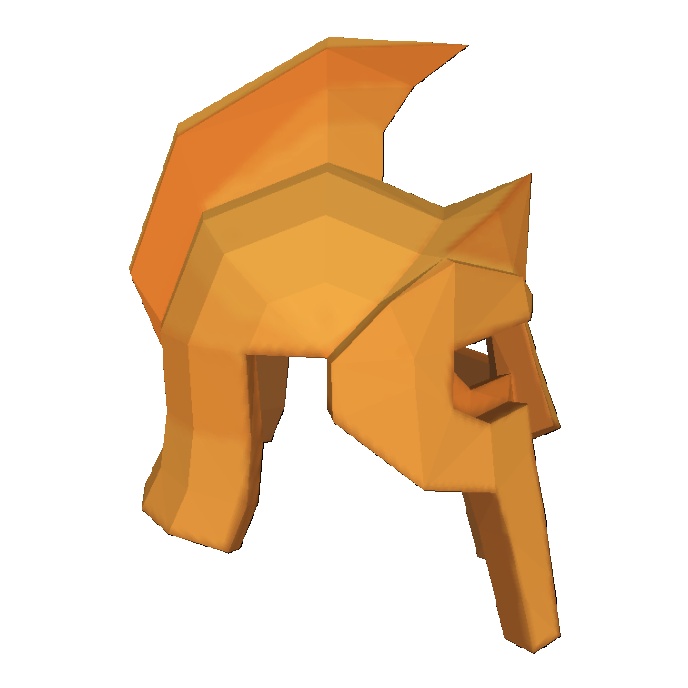} &
\includegraphics[width=0.1\textwidth]{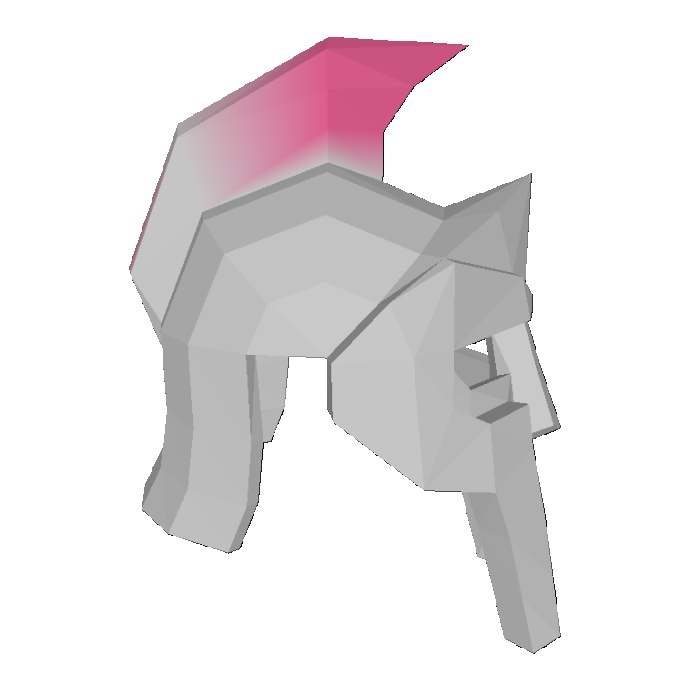} &
\includegraphics[width=0.1\textwidth]{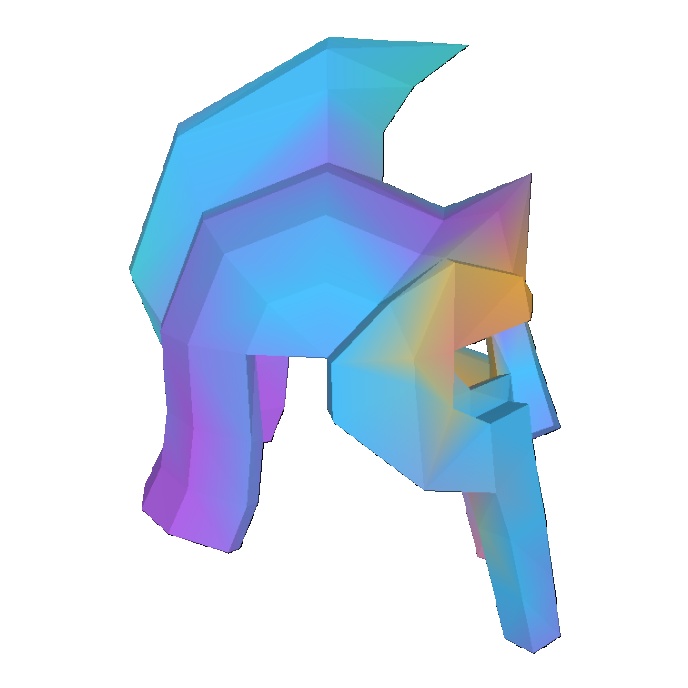} &
\includegraphics[width=0.1\textwidth]{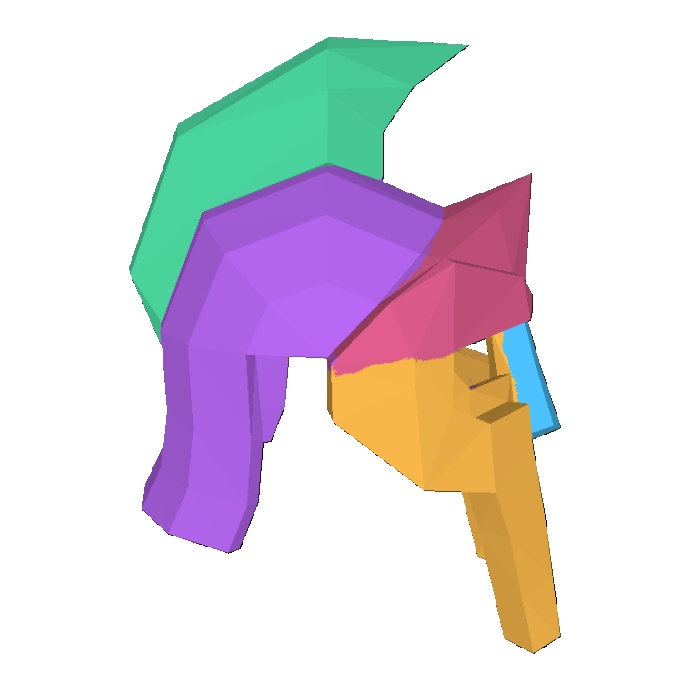} \\

\addlinespace[-2pt]
\arrayrulecolor{gray}\cmidrule(lr){1-4}
\arrayrulecolor{black}

\includegraphics[width=0.1\textwidth]{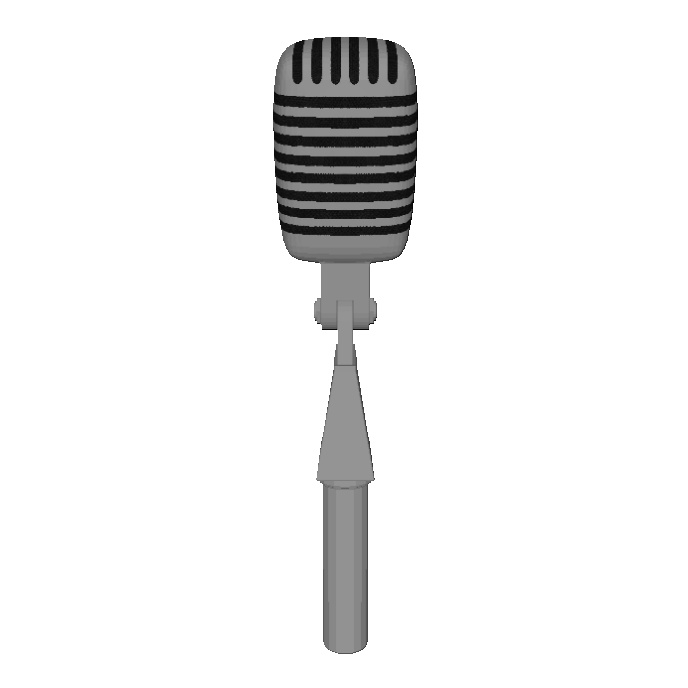} &
\includegraphics[width=0.1\textwidth]{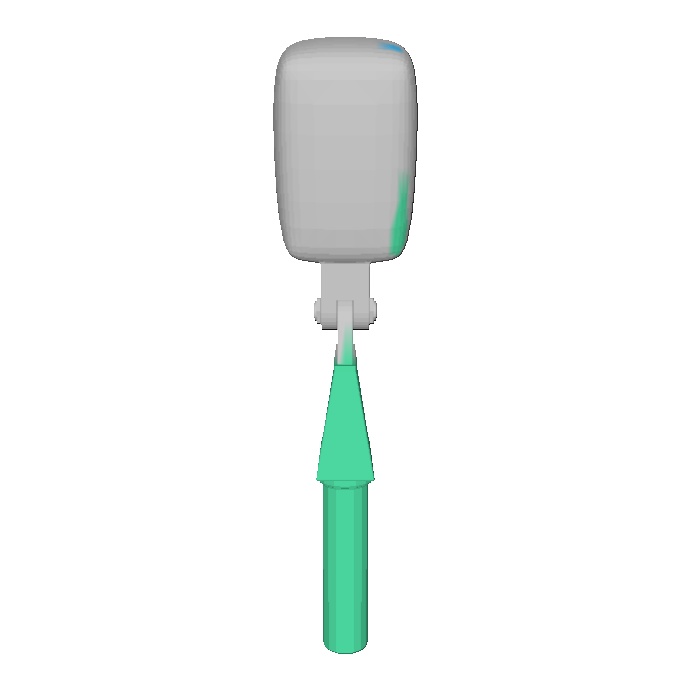} &
\includegraphics[width=0.1\textwidth]{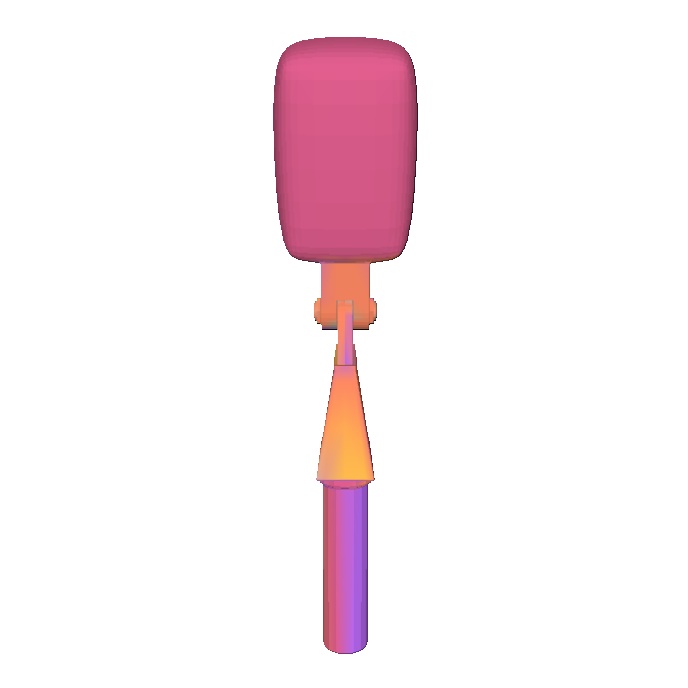} &
\includegraphics[width=0.1\textwidth]{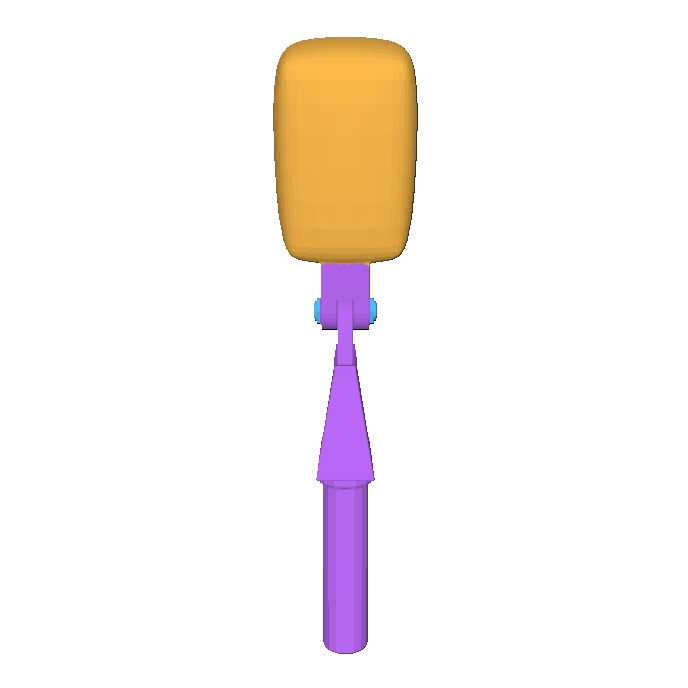} \\

\addlinespace[-2pt]
\arrayrulecolor{gray}\cmidrule(lr){1-4}
\arrayrulecolor{black}

\includegraphics[width=0.1\textwidth]{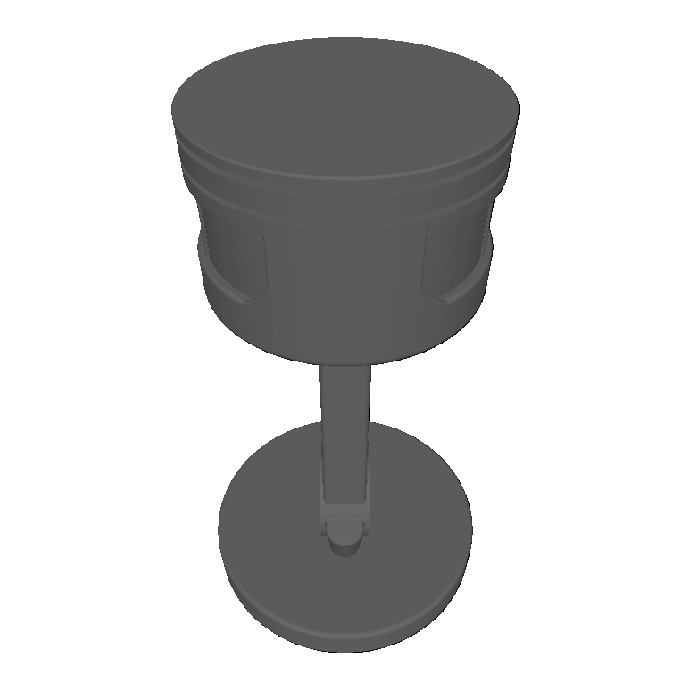} &
\includegraphics[width=0.1\textwidth]{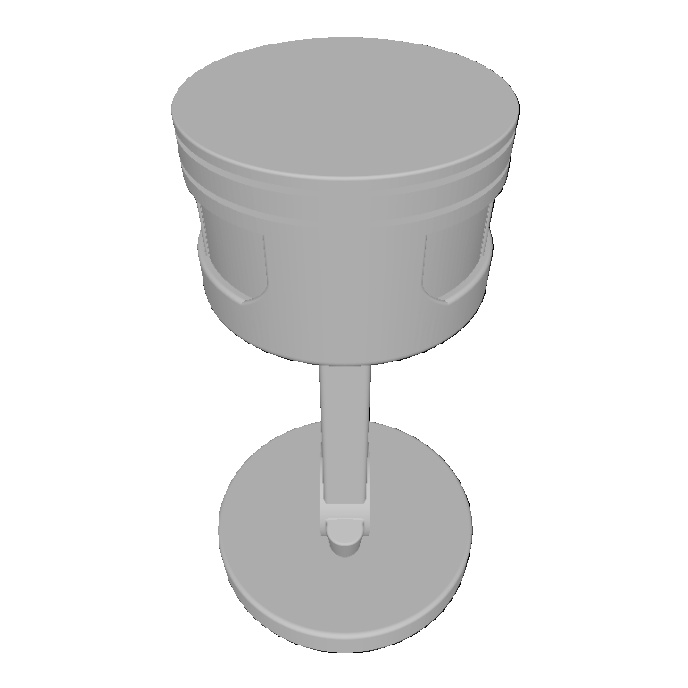} &
\includegraphics[width=0.1\textwidth]{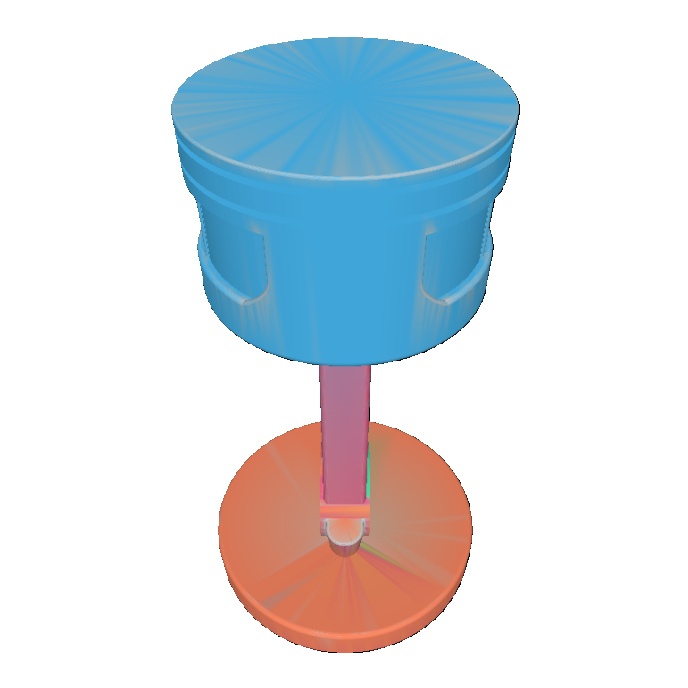} &
\includegraphics[width=0.1\textwidth]{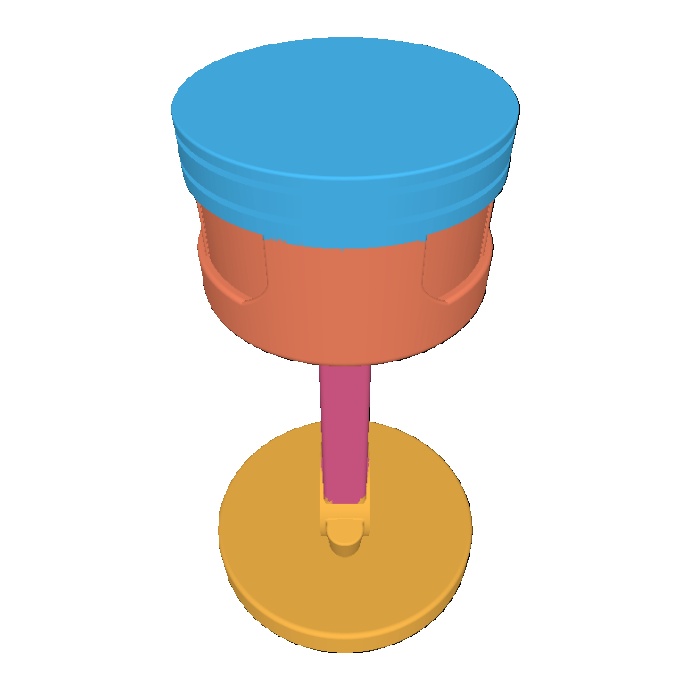} \\

\addlinespace[-2pt]
\arrayrulecolor{gray}\cmidrule(lr){1-4}
\arrayrulecolor{black}

\includegraphics[width=0.1\textwidth]{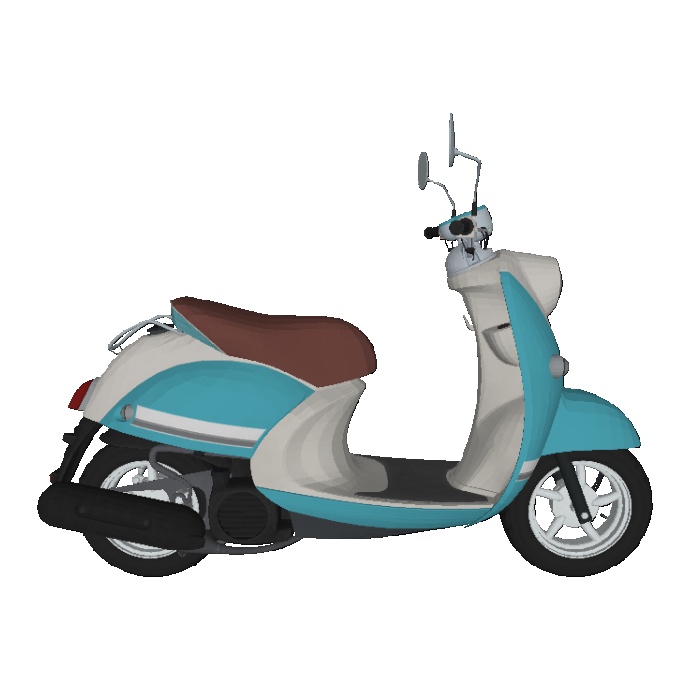} &
\includegraphics[width=0.1\textwidth]{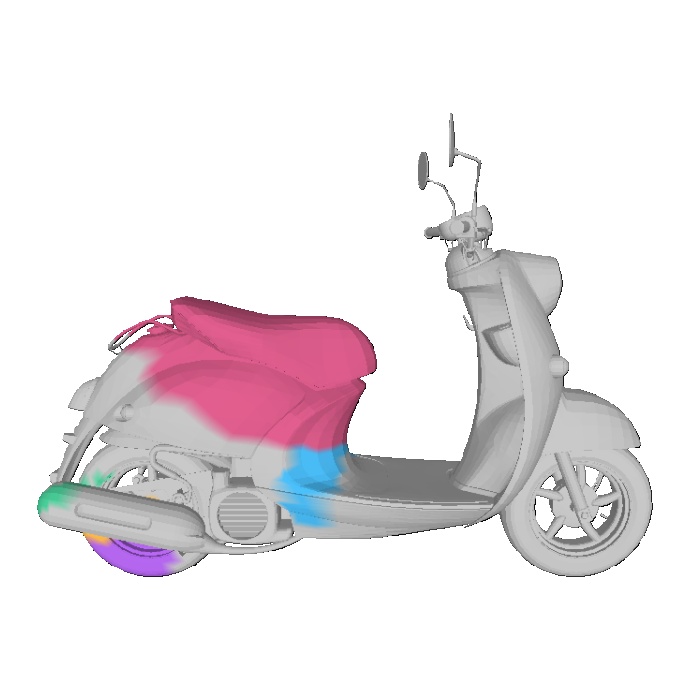} &
\includegraphics[width=0.1\textwidth]{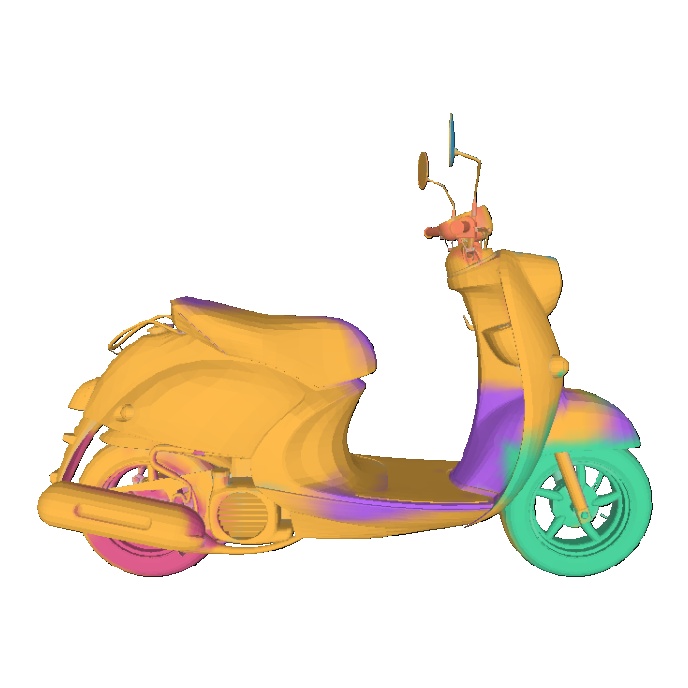} &
\includegraphics[width=0.1\textwidth]{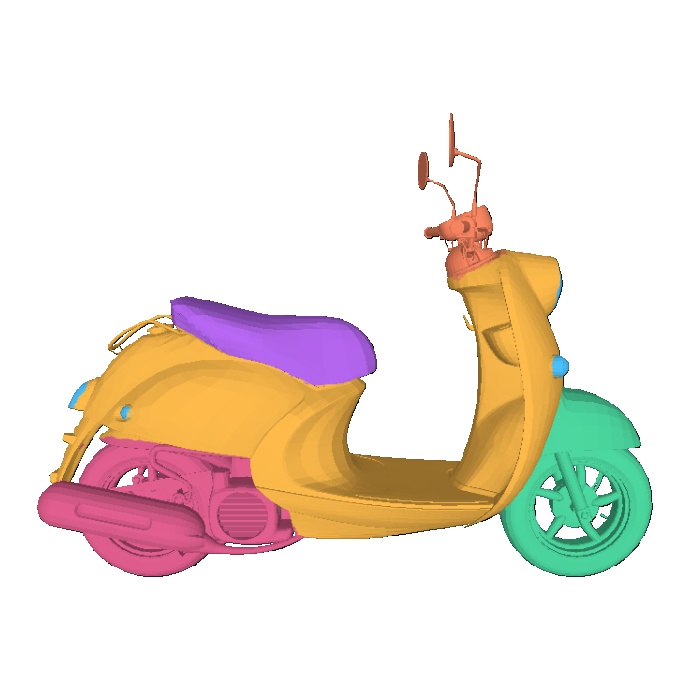} \\

\addlinespace[-2pt]
\arrayrulecolor{gray}\cmidrule(lr){1-4}
\arrayrulecolor{black}

\includegraphics[width=0.1\textwidth]{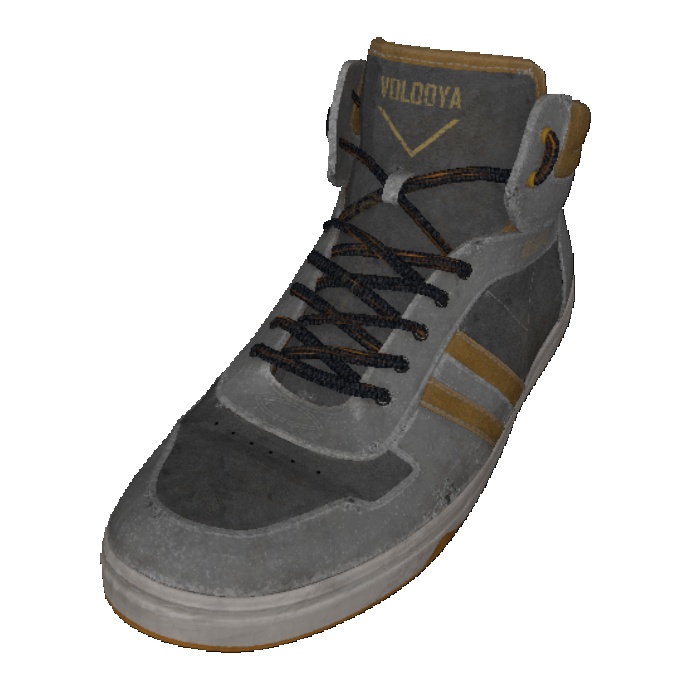} &
\includegraphics[width=0.1\textwidth]{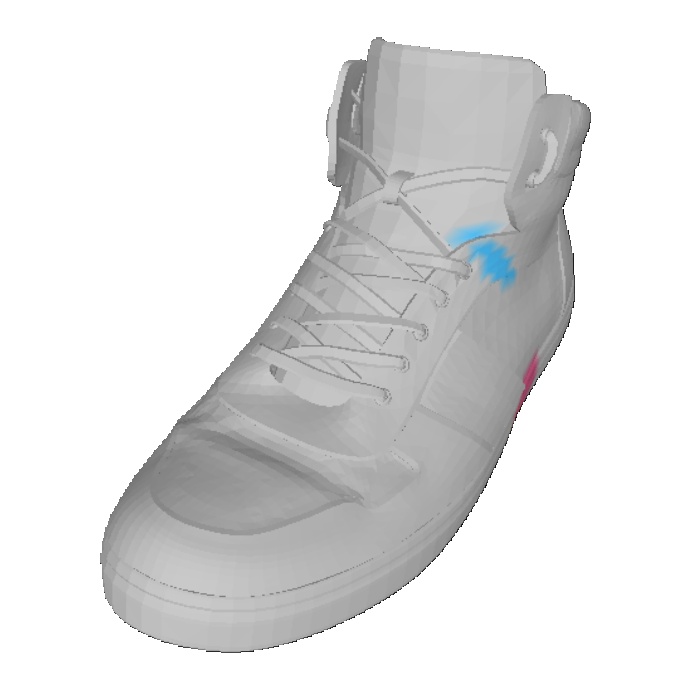} &
\includegraphics[width=0.1\textwidth]{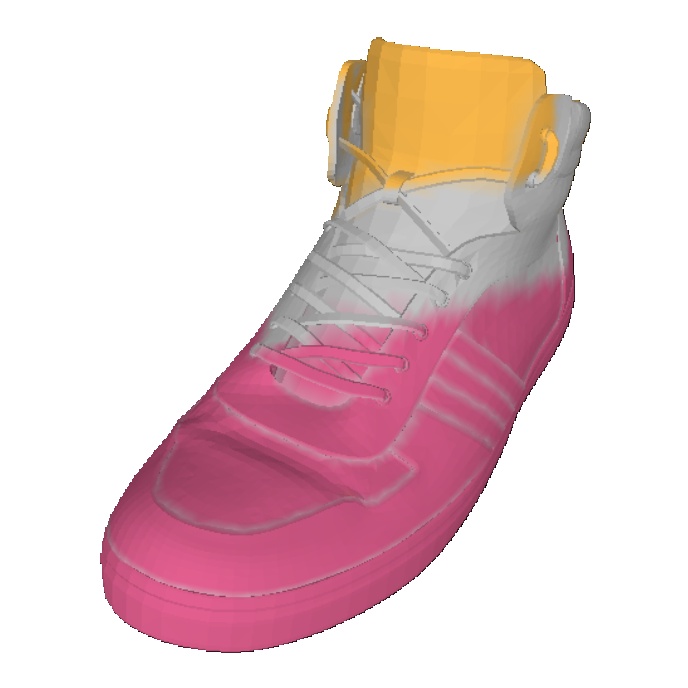} &
\includegraphics[width=0.1\textwidth]{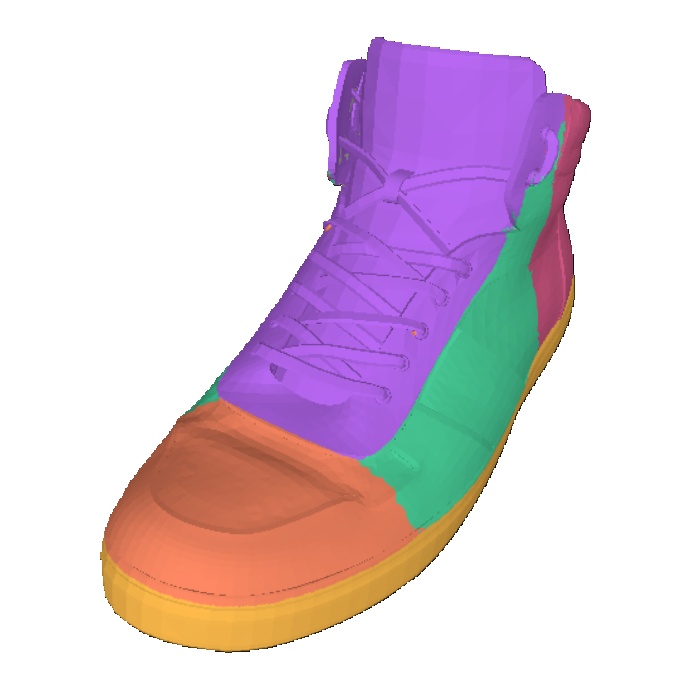} \\

\addlinespace[-2pt]
\arrayrulecolor{gray}\cmidrule(lr){1-4}
\arrayrulecolor{black}

\includegraphics[width=0.1\textwidth]{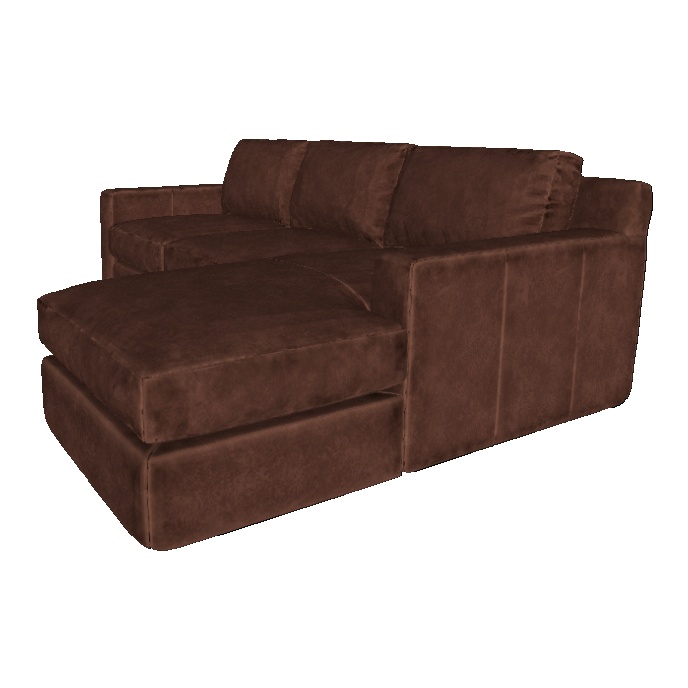} &
\includegraphics[width=0.1\textwidth]{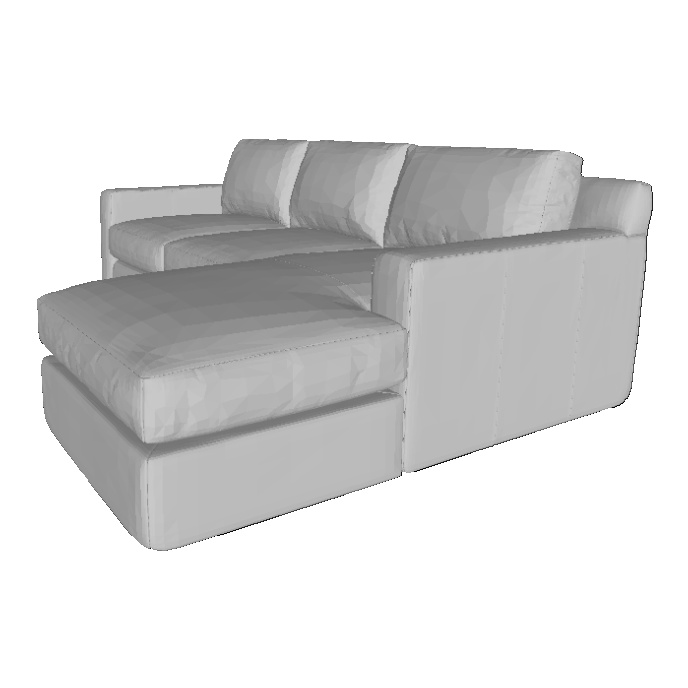} &
\includegraphics[width=0.1\textwidth]{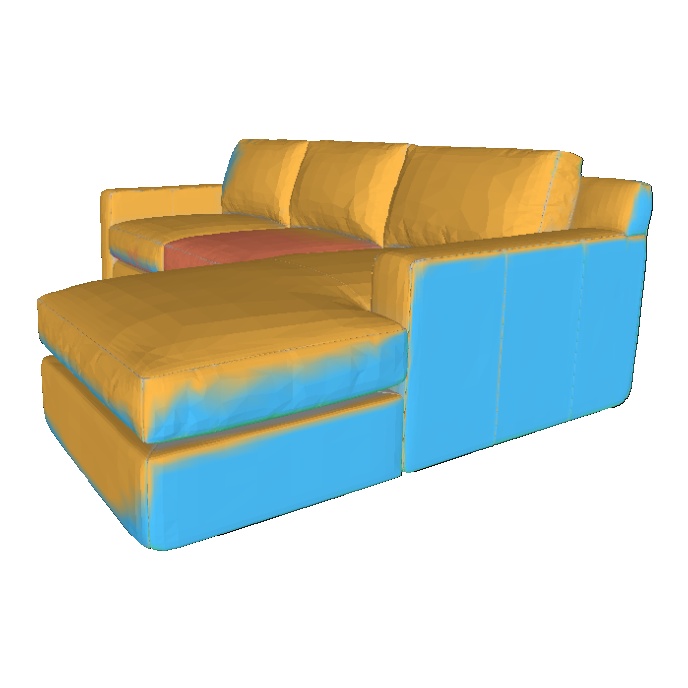} &
\includegraphics[width=0.1\textwidth]{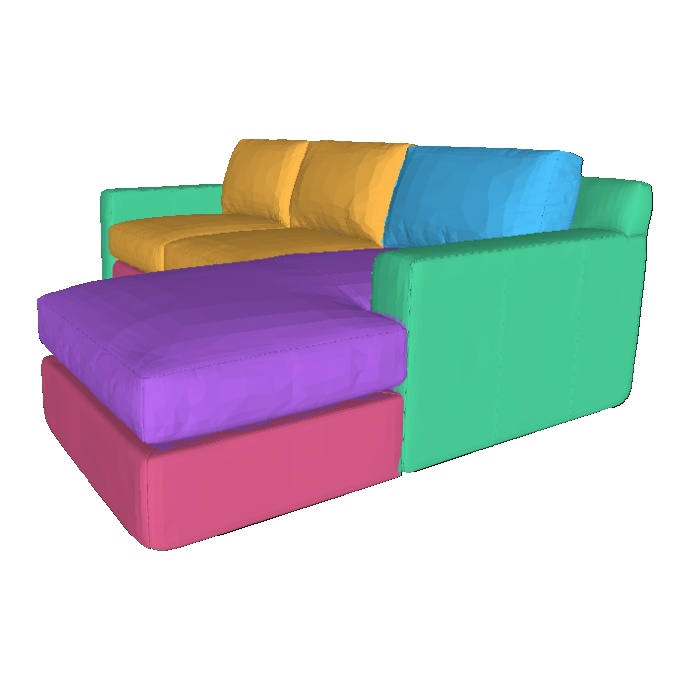} \\

\addlinespace[-2pt]
\arrayrulecolor{gray}\cmidrule(lr){1-4}
\arrayrulecolor{black}

\includegraphics[width=0.1\textwidth]{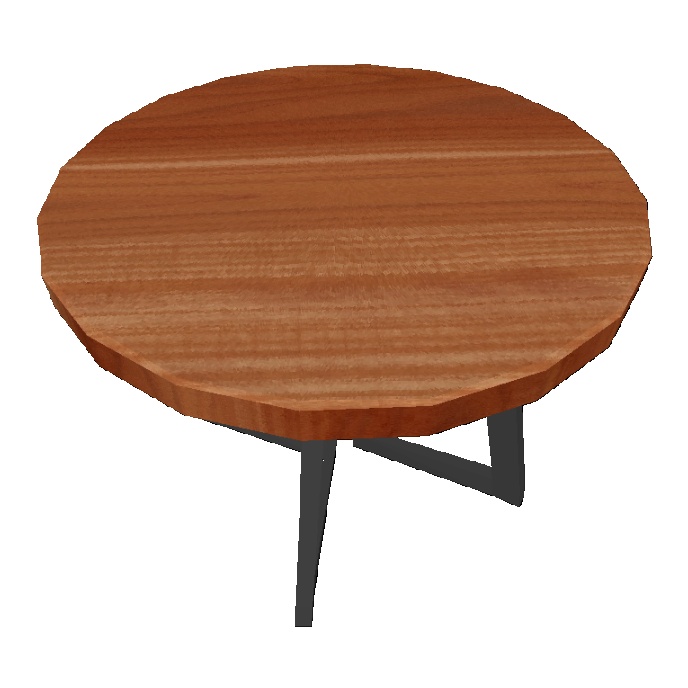} &
\includegraphics[width=0.1\textwidth]{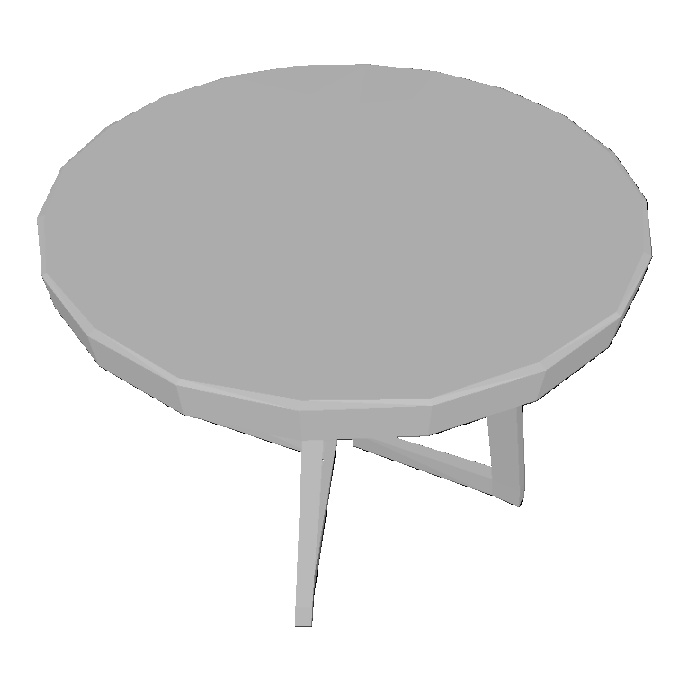} &
\includegraphics[width=0.1\textwidth]{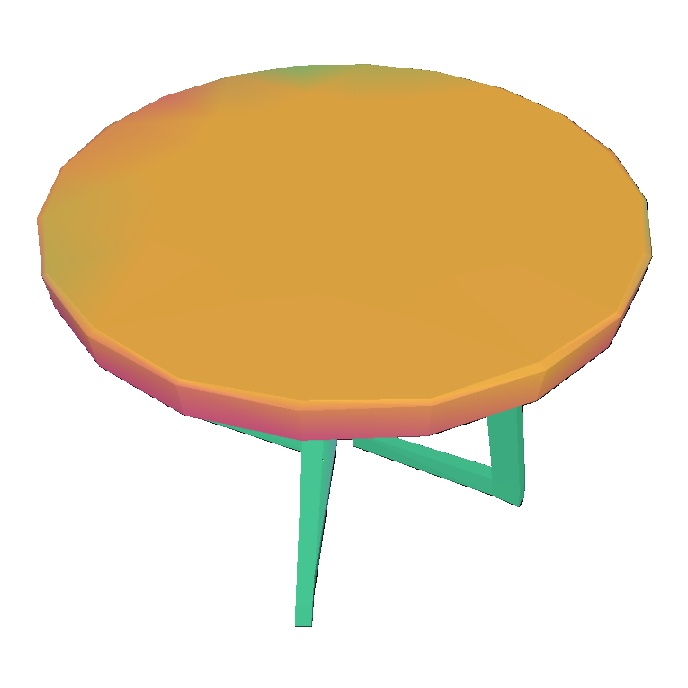} &
\includegraphics[width=0.1\textwidth]{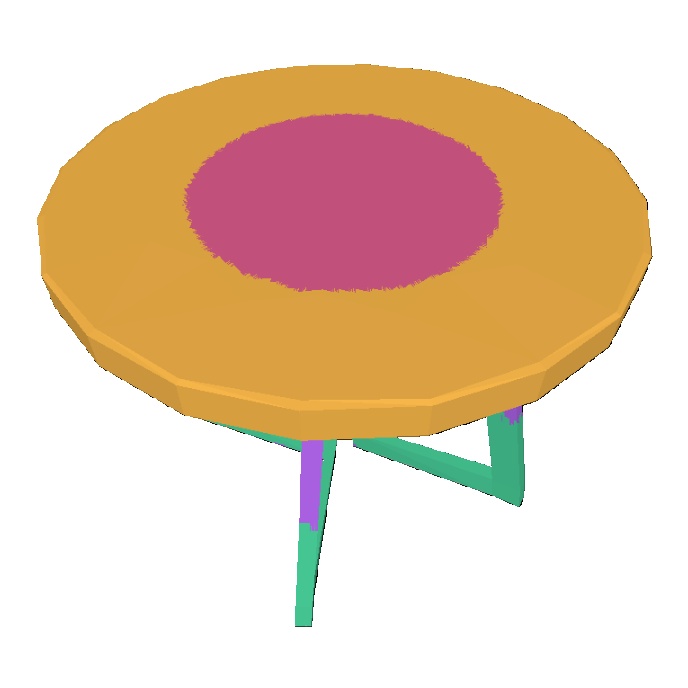} \\

\addlinespace[-2pt]
\bottomrule
\end{tabular}
}}
\end{tabular}
\end{table*}

\begin{table*}[]

\centering
\caption{
Quantitative comparison of results from PartSLIP~\cite{liu2023partslip}, PartSTAD~\cite{kim2024partstad}, PartDistill~\cite{umam2024partdistill}, 3$\times$2~\cite{3by2}, and Ours for semantic segmentation on 40 categories of the PartNetE~\cite{liu2023partslip} dataset. We report the average per-part mIoU ($\uparrow$) in percentage for each category and the mean mIoU for all 40 categories at the end.
}
\label{tab:supp_sem_seg_full_table}

\resizebox{0.95\textwidth}{!}{%
\centering
\begin{tabular}
{p{0.5\textwidth} | p{0.5\textwidth}}
\vtop{\vskip0pt
\resizebox{0.5\textwidth}{!}{
\begin{tabular}{c|ccccc}
\toprule
Category & PartSLIP & PartSTAD & PartDistill & 3$\times$2 & Ours \\ \midrule
Bottle  & 83.4     & 83.6     & 84.6        & 80.9 & \textbf{88.8} \\ 
\arrayrulecolor{gray}\cmidrule(lr){1-6}
\arrayrulecolor{black}
Box  & 84.5     & 81.1     & \textbf{87.9}        & 76.1 & \textbf{87.9} \\
\arrayrulecolor{gray}\cmidrule(lr){1-6}
\arrayrulecolor{black}
Bucket  & 36.5     & \textbf{83.6}     & 50.7        & 78.4 & 76.6 \\
\arrayrulecolor{gray}\cmidrule(lr){1-6}
\arrayrulecolor{black}
Camera  & 58.3     & 64.4     & 60.1        & 62.6 & \textbf{69.8} \\
\arrayrulecolor{gray}\cmidrule(lr){1-6}
\arrayrulecolor{black}
Cart  & 88.1     & 85.0     & \textbf{90.1}        & 81.2 & 89.9 \\
\arrayrulecolor{gray}\cmidrule(lr){1-6}
\arrayrulecolor{black}
Chair  & 85.3     & 85.32     & 88.4        & 84.4 & \textbf{90.1} \\
\arrayrulecolor{gray}\cmidrule(lr){1-6}
\arrayrulecolor{black}
Clock  & 37.6     & 47.4     & 37.2        & 45.8 & \textbf{56.0} \\
\arrayrulecolor{gray}\cmidrule(lr){1-6}
\arrayrulecolor{black}
\begin{tabular}[c]{@{}c@{}}Coffee\\ Machine\end{tabular}  & 37.8     & 35.8     & \textbf{40.2}        & 34.2 & 39.0 \\
\arrayrulecolor{gray}\cmidrule(lr){1-6}
\arrayrulecolor{black}
\begin{tabular}[c]{@{}c@{}}Dis-\\ penser\end{tabular}  & 73.8     & 73.7     & 74.7        & 78.2 & \textbf{81.6} \\
\arrayrulecolor{gray}\cmidrule(lr){1-6}
\arrayrulecolor{black}
Display  & 84.8     & 82.3     & \textbf{87.4}        & 72.6 & 82.9 \\
\arrayrulecolor{gray}\cmidrule(lr){1-6}
\arrayrulecolor{black}
Door  & 40.8     & 61.4     & 55.5        & 54.4 & \textbf{77.5} \\
\arrayrulecolor{gray}\cmidrule(lr){1-6}
\arrayrulecolor{black}
\begin{tabular}[c]{@{}c@{}}Eye-\\ glasses\end{tabular}  & 88.3     & 92.4     & 91.1        & 92.8 & \textbf{95.9} \\
\arrayrulecolor{gray}\cmidrule(lr){1-6}
\arrayrulecolor{black}
Faucet  & 71.4     & 65.3     & 73.5        & 66.9 & \textbf{78.1} \\
\arrayrulecolor{gray}\cmidrule(lr){1-6}
\arrayrulecolor{black}
\begin{tabular}[c]{@{}c@{}}Folding\\ Chair\end{tabular}  & 86.3     & 91.6     & 90.7        & 93.6 & \textbf{94.2} \\
\arrayrulecolor{gray}\cmidrule(lr){1-6}
\arrayrulecolor{black}
Globe  & 95.7     & 93.5     & 97.4        & 95.2 & \textbf{98.3} \\
\arrayrulecolor{gray}\cmidrule(lr){1-6}
\arrayrulecolor{black}
Kettle  & 77.0     & 84.2     & 78.6        & 81.5 & \textbf{91.0} \\
\arrayrulecolor{gray}\cmidrule(lr){1-6}
\arrayrulecolor{black}
\begin{tabular}[c]{@{}c@{}}Key-\\ board\end{tabular}  & 53.6     & 82.4     & 70.8        & 89.6 & \textbf{92.6} \\
\arrayrulecolor{gray}\cmidrule(lr){1-6}
\arrayrulecolor{black}
\begin{tabular}[c]{@{}c@{}}Kitchen\\ Pot\end{tabular}  & 69.6     & 73.5     & 69.7        & 65.0 & \textbf{82.5} \\
\arrayrulecolor{gray}\cmidrule(lr){1-6}
\arrayrulecolor{black}
Knife  & 65.2     & 63.8     & 71.4        & 75.1 & \textbf{78.4} \\
\bottomrule
\end{tabular}
}}
&
\vtop{\vskip0pt
\resizebox{0.5\textwidth}{!}{
\begin{tabular}{c|ccccc}
\toprule
Category & PartSLIP & PartSTAD & PartDistill & 3$\times$2 & Ours \\ \midrule

Lamp  & 66.1     & 68.4     & 69.2        & 59.5 & \textbf{76.9} \\
\arrayrulecolor{gray}\cmidrule(lr){1-6}
\arrayrulecolor{black}
Laptop  & 29.7     & 34.56     & 40.0        & 45.3 & \textbf{58.5} \\
\arrayrulecolor{gray}\cmidrule(lr){1-6}
\arrayrulecolor{black}
Lighter  & 64.7     & 65.9     & 64.9        & 65.0 & \textbf{73.3} \\
\arrayrulecolor{gray}\cmidrule(lr){1-6}
\arrayrulecolor{black}
Mouse  & 44.0     & 48.9     & 46.9        & \textbf{68.4} & 51.0 \\
\arrayrulecolor{gray}\cmidrule(lr){1-6}
\arrayrulecolor{black}
Oven  & 73.5     & 71.8     & 72.8        & 60.0 & \textbf{77.2} \\
\arrayrulecolor{gray}\cmidrule(lr){1-6}
\arrayrulecolor{black}
Pen  & 71.5     & 60.6     & \textbf{74.4}        & 62.8 & 71.5 \\
\arrayrulecolor{gray}\cmidrule(lr){1-6}
\arrayrulecolor{black}
Phone  & 48.4     & 63.2     & 50.8        & 41.0 & \textbf{69.3} \\
\arrayrulecolor{gray}\cmidrule(lr){1-6}
\arrayrulecolor{black}
Pliers  & 33.2     & 99.3     & 90.4        & 99.4 & \textbf{99.5} \\
\arrayrulecolor{gray}\cmidrule(lr){1-6}
\arrayrulecolor{black}
Printer  & 4.3     & 7.9     & 6.3        & 0.85 & \textbf{38.4} \\
\arrayrulecolor{gray}\cmidrule(lr){1-6}
\arrayrulecolor{black}
Remote  & 38.3     & 53.4     & 40.7        & 54.1 & \textbf{74.7} \\
\arrayrulecolor{gray}\cmidrule(lr){1-6}
\arrayrulecolor{black}
Safe  & 32.2     & 36.8     & \textbf{58.6}        & 34.7 & 54.5 \\
\arrayrulecolor{gray}\cmidrule(lr){1-6}
\arrayrulecolor{black}
Scissors  & 60.3     & 68.5     & 68.8        & 65.7 & \textbf{83.7} \\
\arrayrulecolor{gray}\cmidrule(lr){1-6}
\arrayrulecolor{black}
Stapler  & 84.8     & 85.8     & 86.3        & 90.1 & \textbf{95.3} \\
\arrayrulecolor{gray}\cmidrule(lr){1-6}
\arrayrulecolor{black}
Suitcase  & 70.4     & 68.2     & \textbf{73.4}        & 65.2 & 70.3 \\
\arrayrulecolor{gray}\cmidrule(lr){1-6}
\arrayrulecolor{black}
Switch  & 59.4     & 57.9     & 60.7        & 54.6 & \textbf{75.7} \\
\arrayrulecolor{gray}\cmidrule(lr){1-6}
\arrayrulecolor{black}
Toaster  & 60.0     & 58.6     & 58.7        & 56.5 & \textbf{73.0} \\
\arrayrulecolor{gray}\cmidrule(lr){1-6}
\arrayrulecolor{black}
Toilet  & 53.8     & 57.5     & 55.0        & 56.9 & \textbf{60.6} \\
\arrayrulecolor{gray}\cmidrule(lr){1-6}
\arrayrulecolor{black}
\begin{tabular}[c]{@{}c@{}}Trash-\\ can\end{tabular}  & 22.3     & 21.2     & \textbf{70.0}        & 33.1 & 36.6 \\
\arrayrulecolor{gray}\cmidrule(lr){1-6}
\arrayrulecolor{black}
USB  & 54.4     & 59.8     & 64.3        & \textbf{79.0} & 77.8 \\
\arrayrulecolor{gray}\cmidrule(lr){1-6}
\arrayrulecolor{black}
\begin{tabular}[c]{@{}c@{}}Washing\\ Machine\end{tabular}  & 53.4     & 48.2     & 55.1        & 52.6 & \textbf{76.0} \\
\arrayrulecolor{gray}\cmidrule(lr){1-6}
\arrayrulecolor{black}
Window  & 75.4     & 76.1     & 78.1        & 73.9 & \textbf{85.9} \\
\midrule
\textbf{Mean}  & 60.5     & 66.1     & 67.1        & 65.9 & \textbf{75.8} \\
\bottomrule
\end{tabular}
}}
\end{tabular}
}
\end{table*}


\begin{table*}[ht]


\centering
\caption{
Qualitative comparison of results from PartSLIP~\cite{liu2023partslip}, PartSTAD~\cite{kim2024partstad}, and Ours for semantic segmentation on the PartNetE~\cite{liu2023partslip} dataset. Each row shows the same object from two different views.
}
\label{tab:PN_full_comp_2}

\begin{tabular}{@{}p{0.5\textwidth}@{} | @{}p{0.5\textwidth}@{}}

\multicolumn{1}{c}{\textbf{View 1}} & \multicolumn{1}{c}{\textbf{View 2}} \\

\vtop{\vskip0pt
\resizebox{0.5\textwidth}{!}{
\begin{tabular}{@{}c@{}c@{}c@{}c@{}c@{}}
\toprule
Input & GT  & PartSLIP & PartSTAD & Ours \\ \midrule
\includegraphics[width=0.1\textwidth]{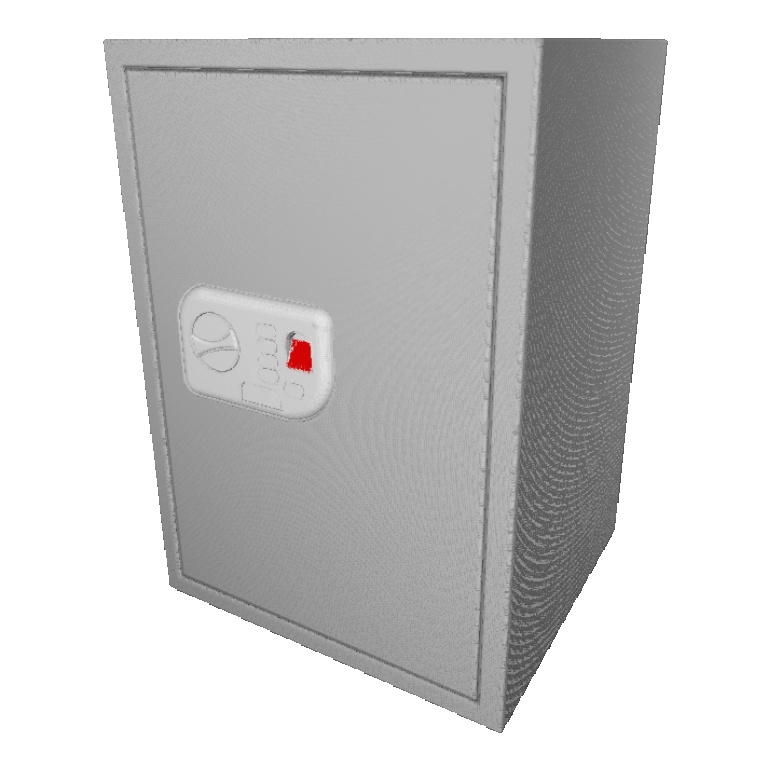} &
\includegraphics[width=0.1\textwidth]{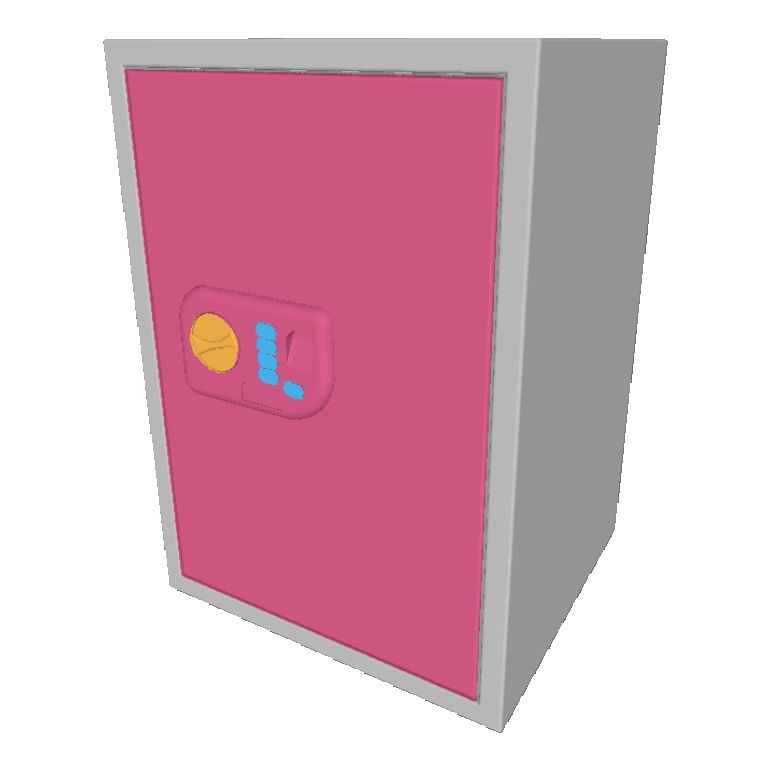} &
\includegraphics[width=0.1\textwidth]{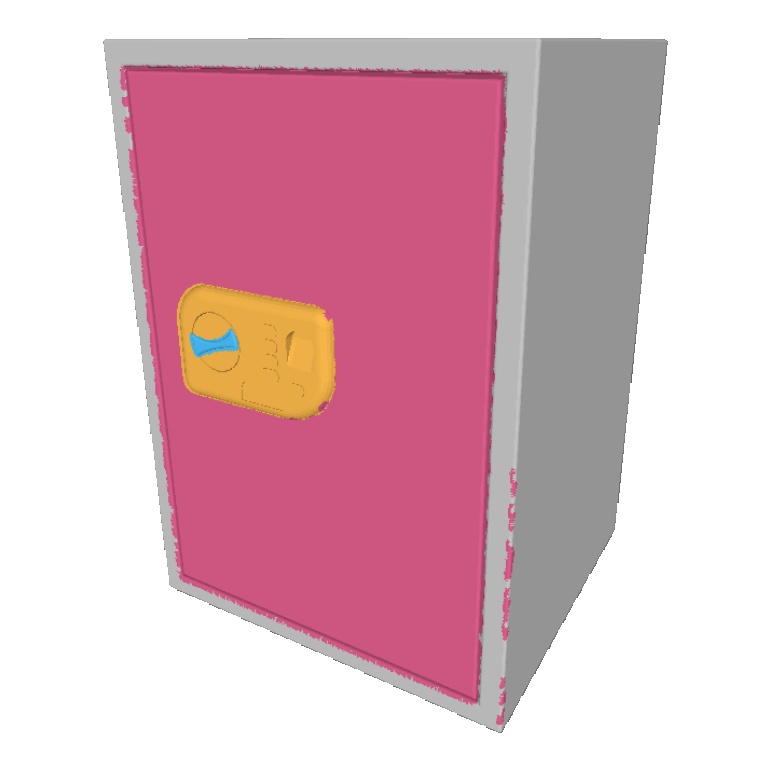} &
\includegraphics[width=0.1\textwidth]{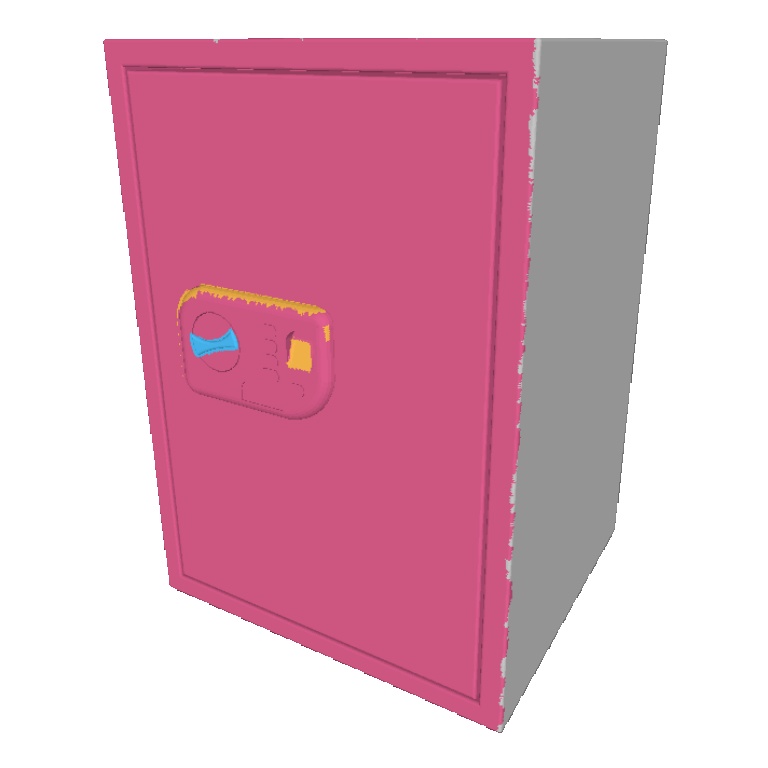} &
\includegraphics[width=0.1\textwidth]{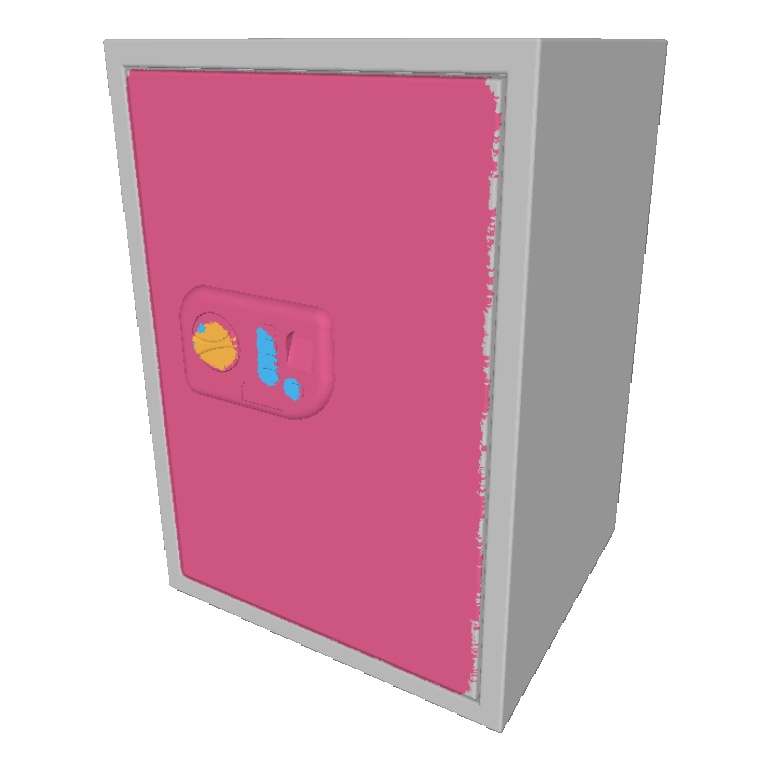} \\

\addlinespace[-2pt]
\arrayrulecolor{gray}\cmidrule(lr){1-5}
\arrayrulecolor{black}

\includegraphics[width=0.1\textwidth]{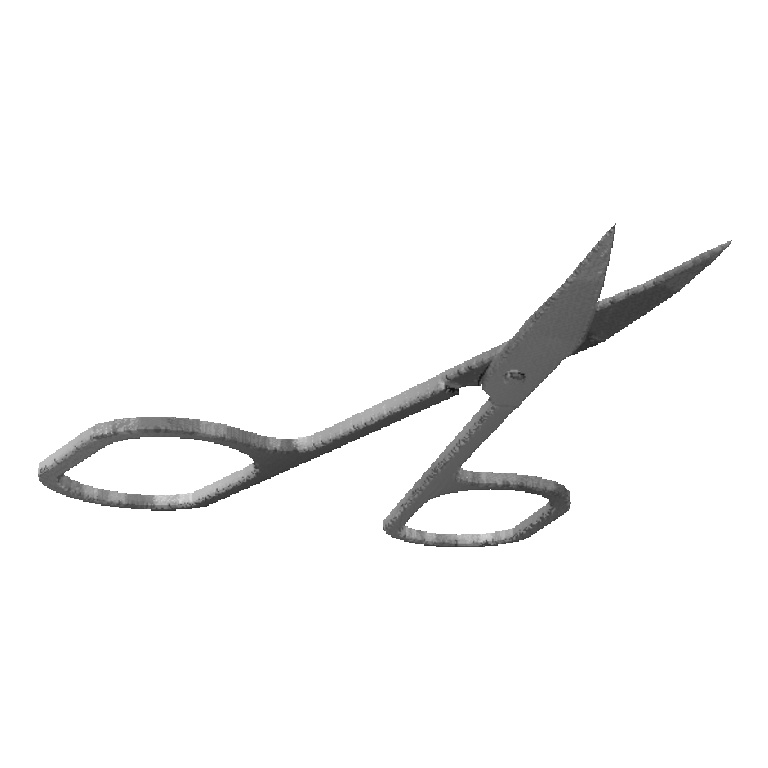} &
\includegraphics[width=0.1\textwidth]{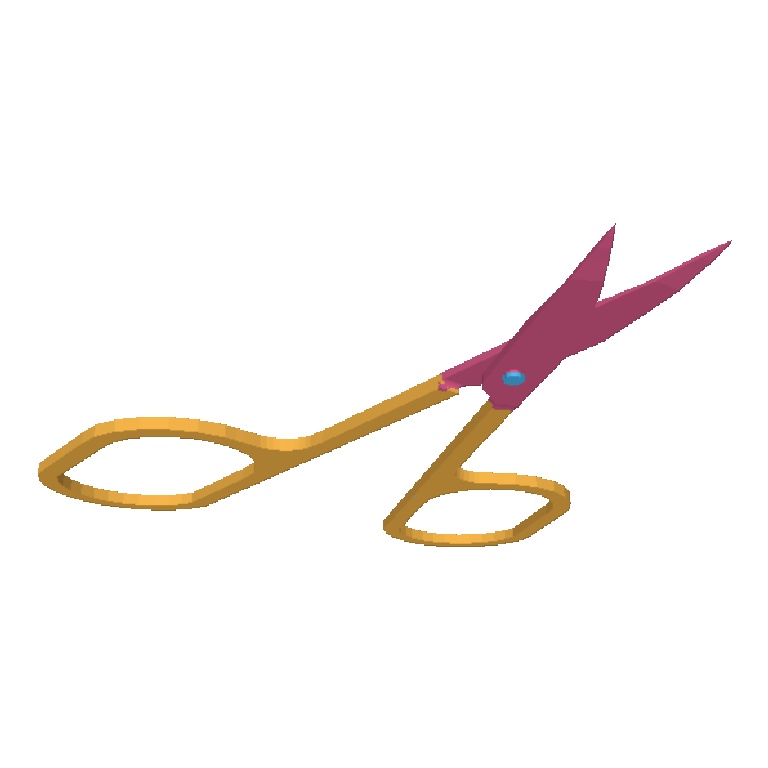} &
\includegraphics[width=0.1\textwidth]{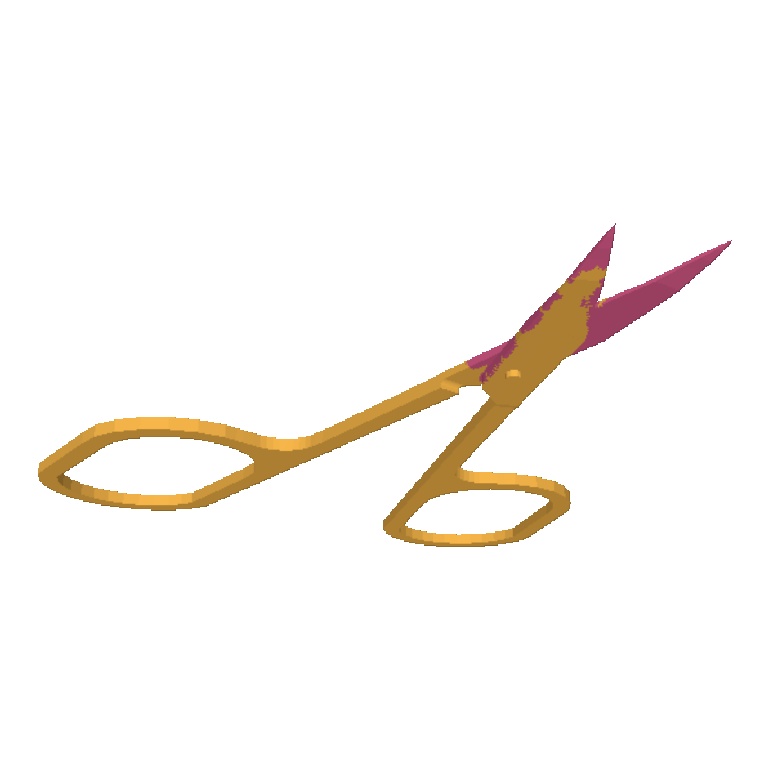} &
\includegraphics[width=0.1\textwidth]{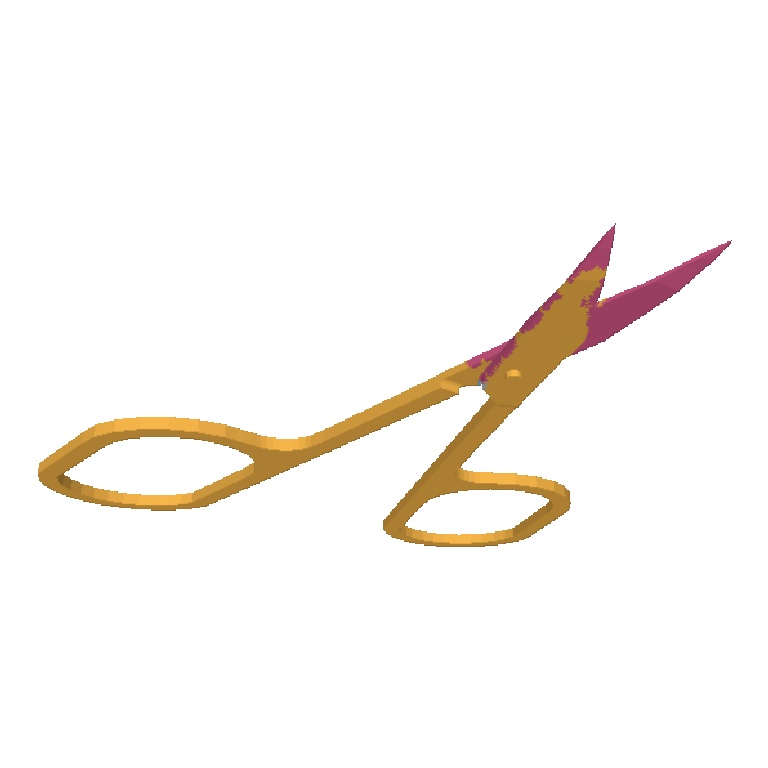} &
\includegraphics[width=0.1\textwidth]{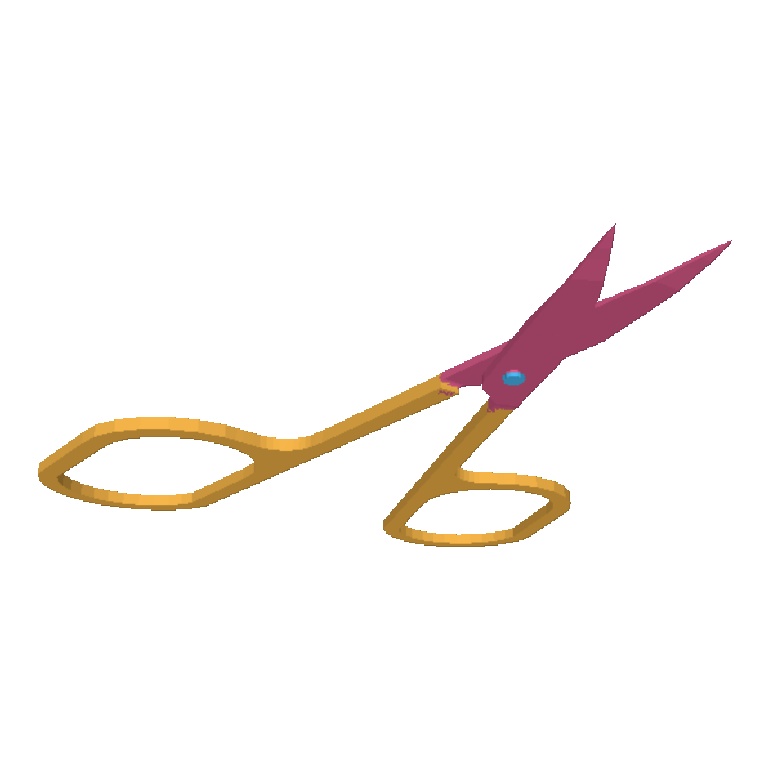} \\

\addlinespace[-2pt]
\arrayrulecolor{gray}\cmidrule(lr){1-5}
\arrayrulecolor{black}

\includegraphics[width=0.1\textwidth]{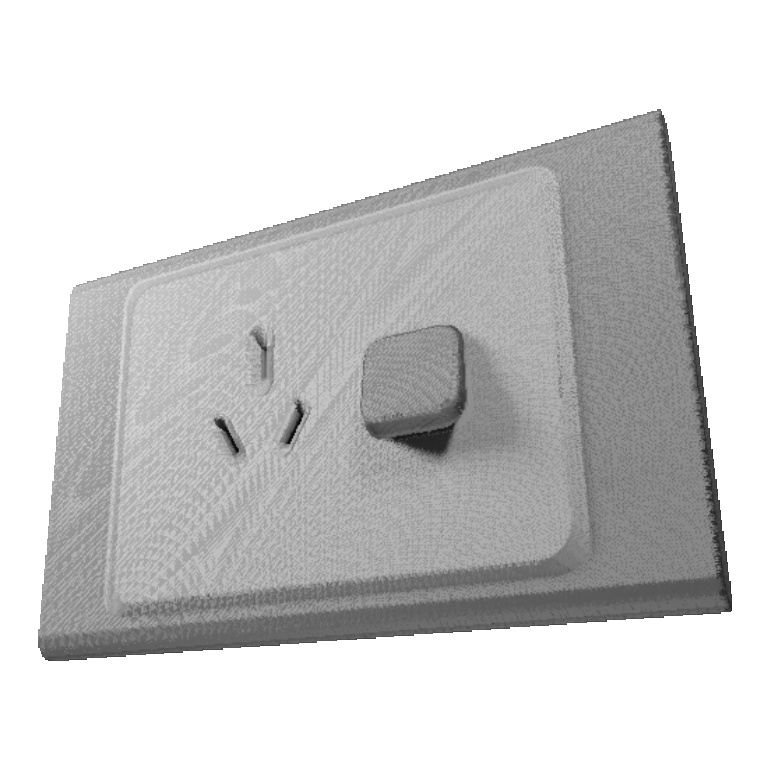} &
\includegraphics[width=0.1\textwidth]{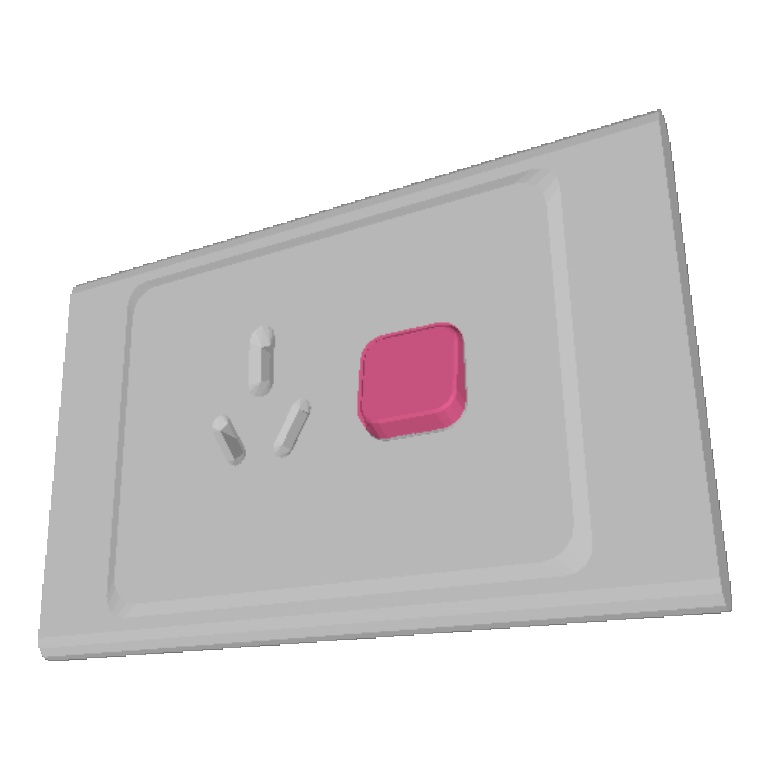} &
\includegraphics[width=0.1\textwidth]{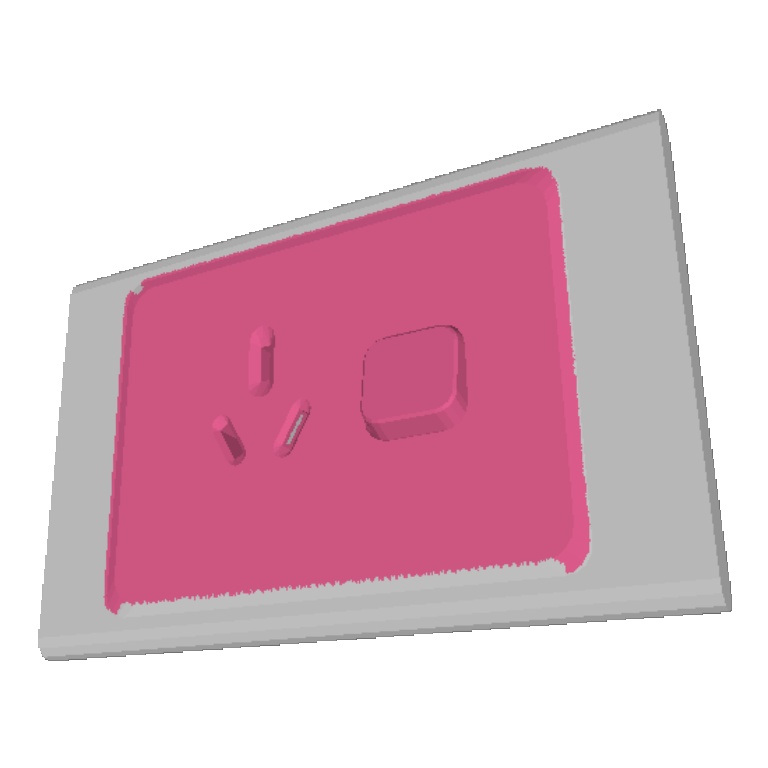} &
\includegraphics[width=0.1\textwidth]{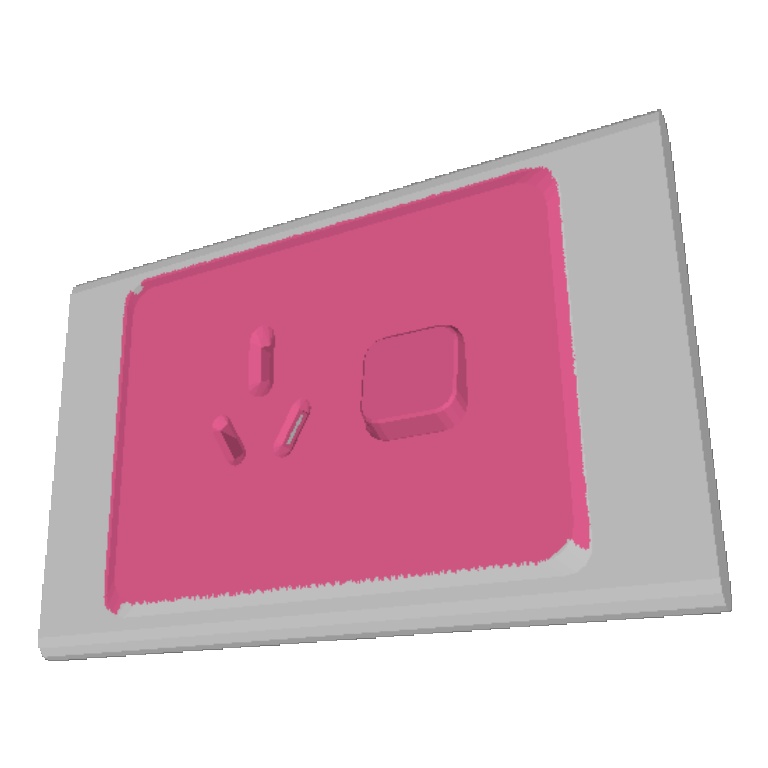} &
\includegraphics[width=0.1\textwidth]{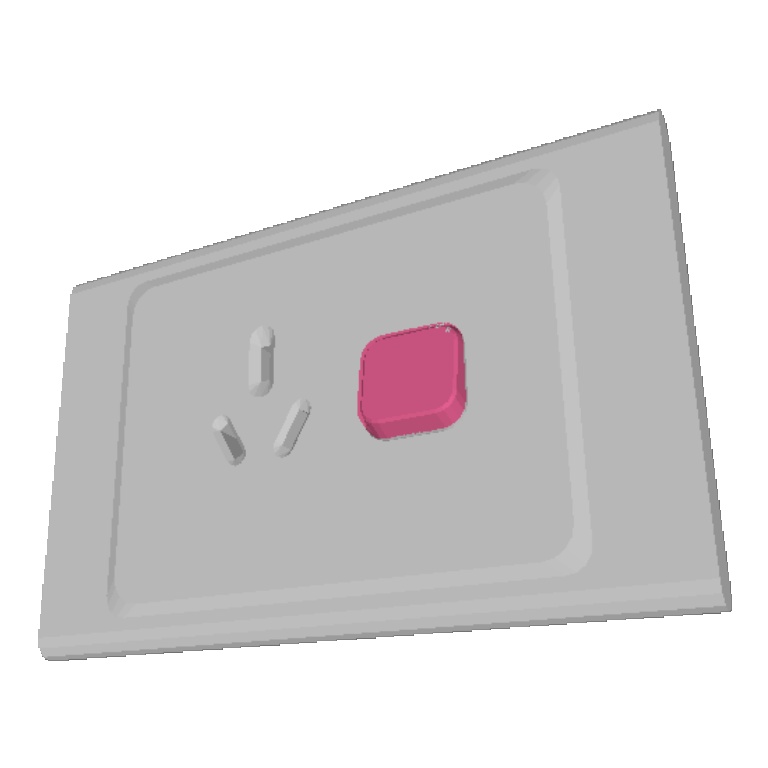} \\

\addlinespace[-2pt]
\arrayrulecolor{gray}\cmidrule(lr){1-5}
\arrayrulecolor{black}

\includegraphics[width=0.1\textwidth]{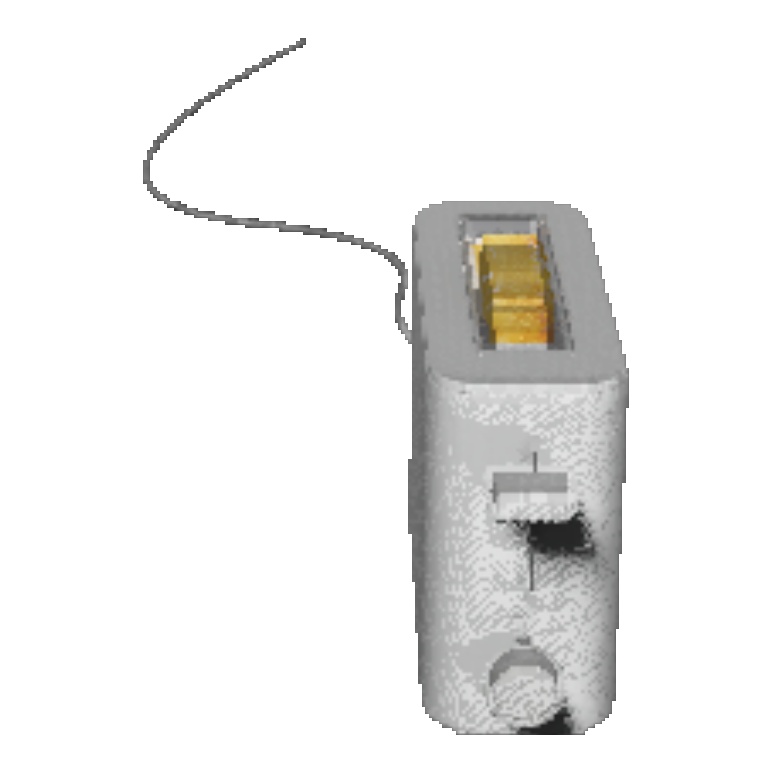} &
\includegraphics[width=0.1\textwidth]{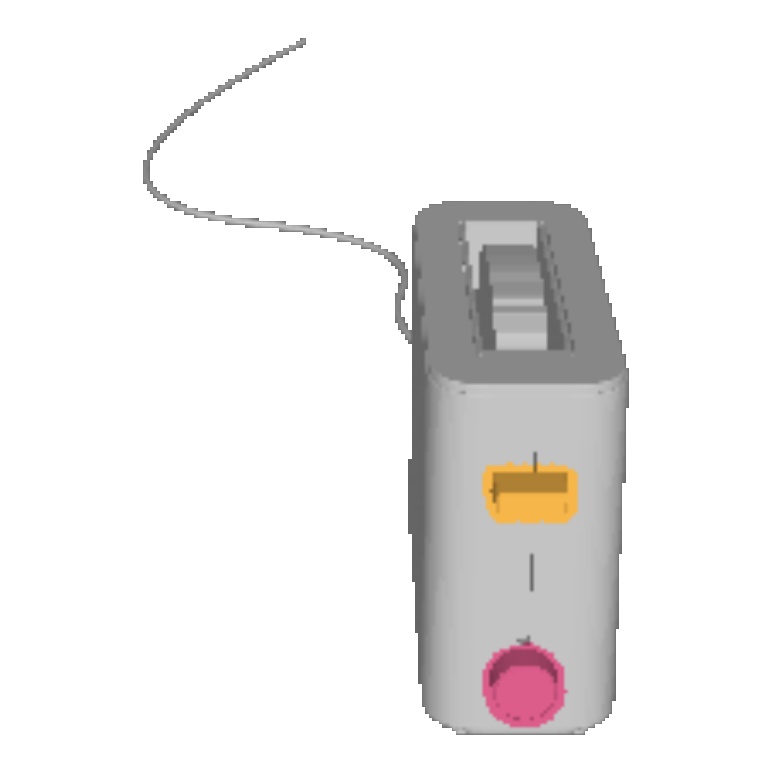} &
\includegraphics[width=0.1\textwidth]{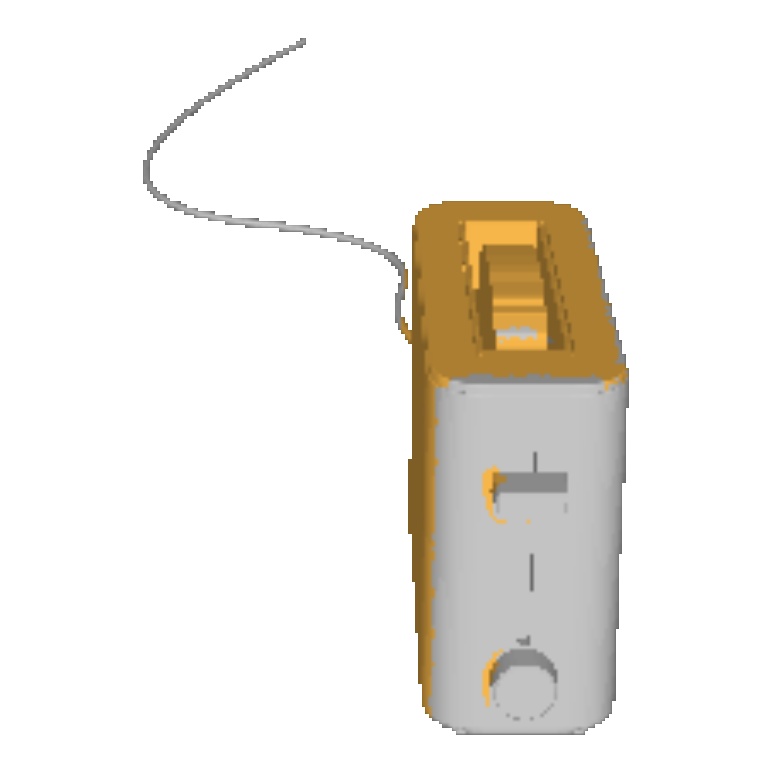} &
\includegraphics[width=0.1\textwidth]{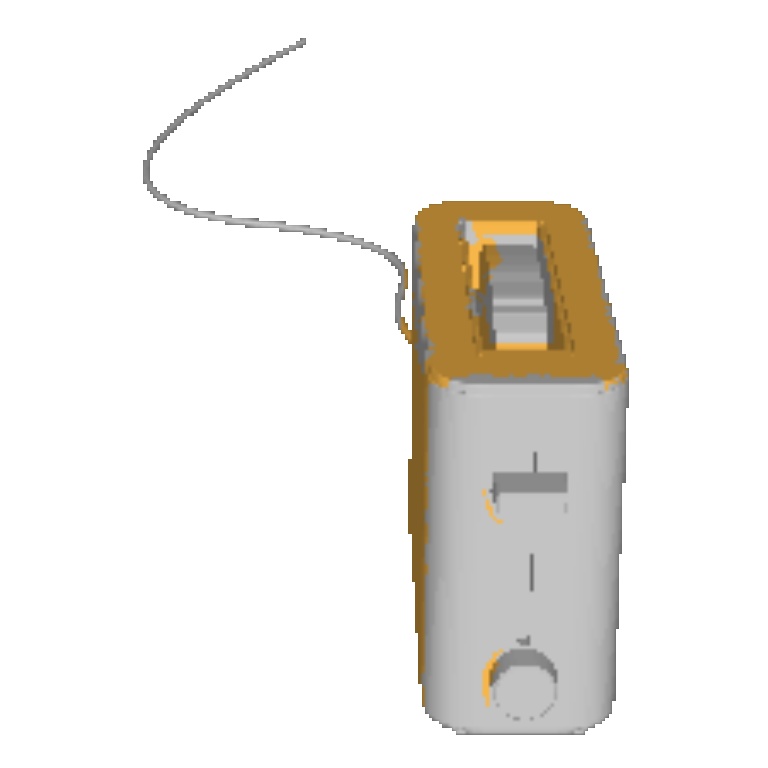} &
\includegraphics[width=0.1\textwidth]{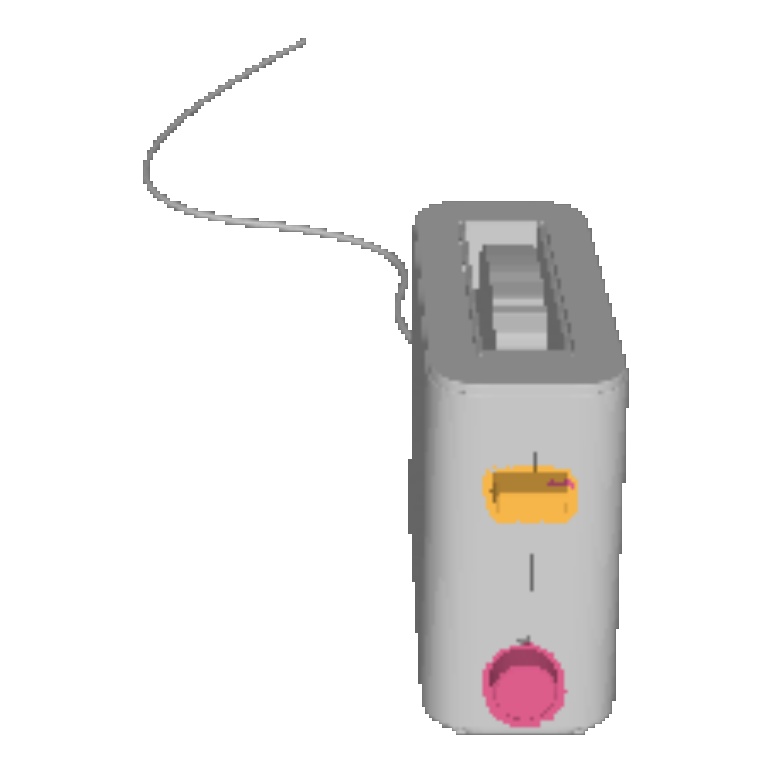} \\

\addlinespace[-2pt]
\arrayrulecolor{gray}\cmidrule(lr){1-5}
\arrayrulecolor{black}

\includegraphics[width=0.1\textwidth]{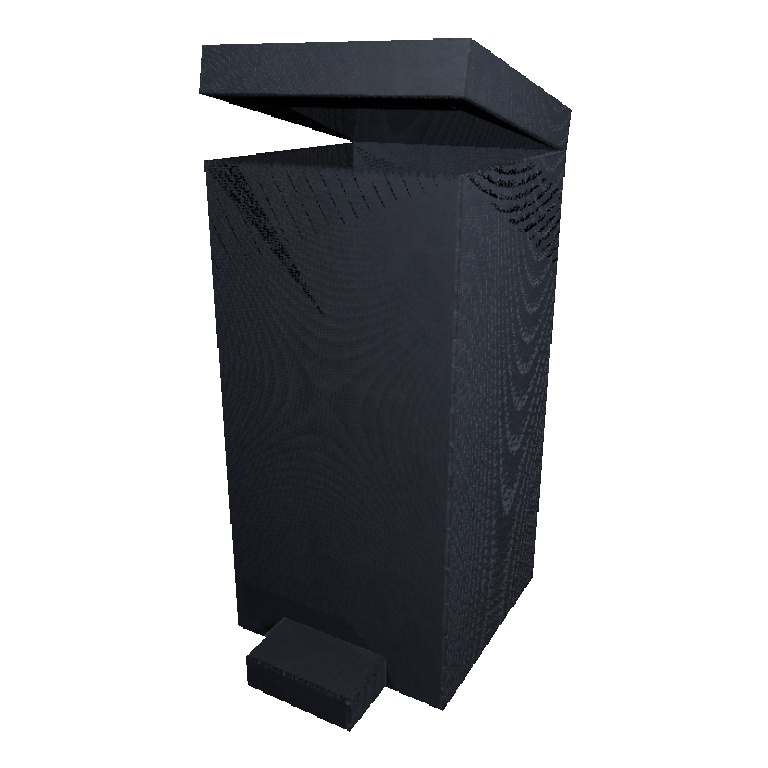} &
\includegraphics[width=0.1\textwidth]{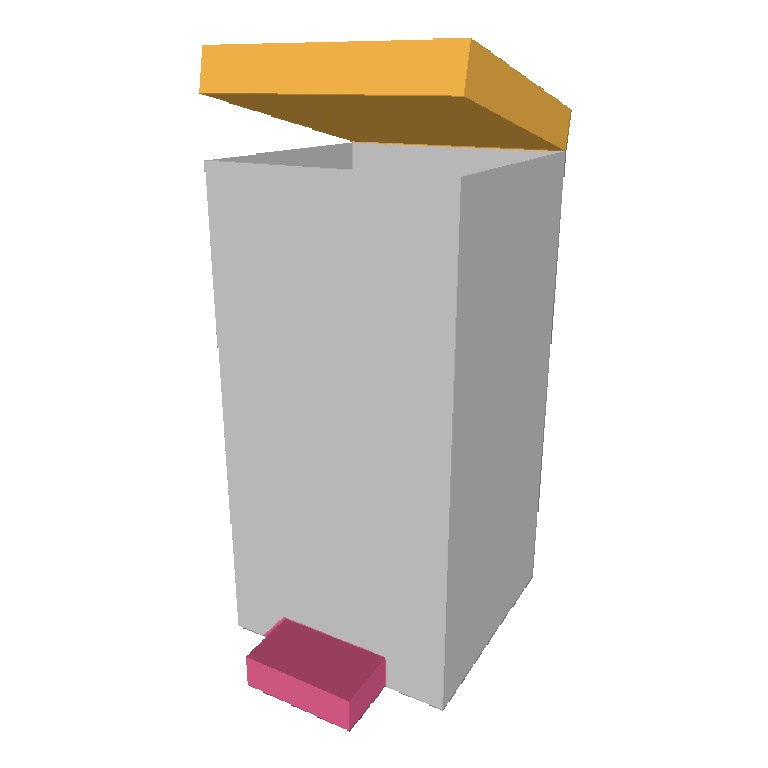} &
\includegraphics[width=0.1\textwidth]{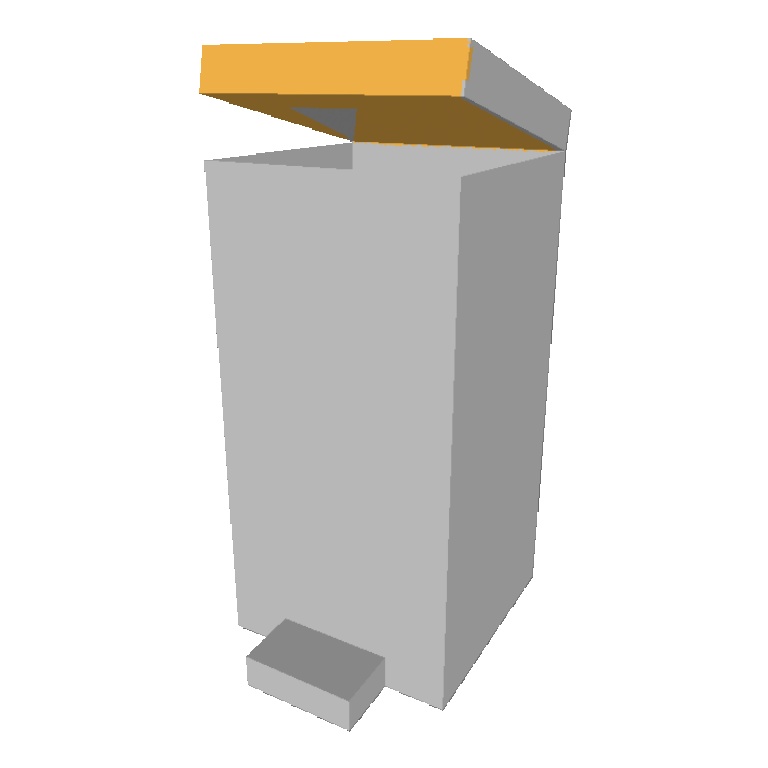} &
\includegraphics[width=0.1\textwidth]{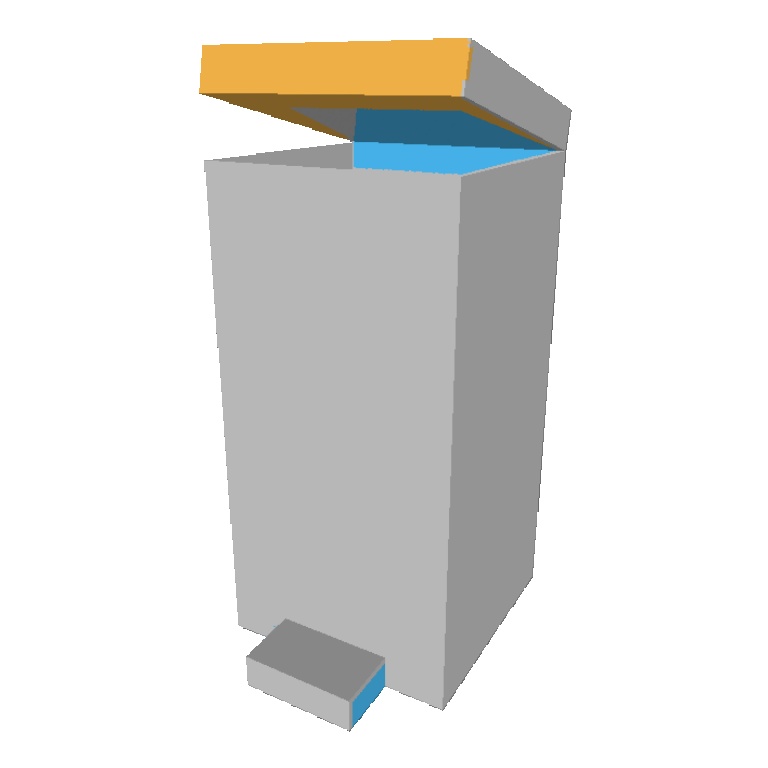} &
\includegraphics[width=0.1\textwidth]{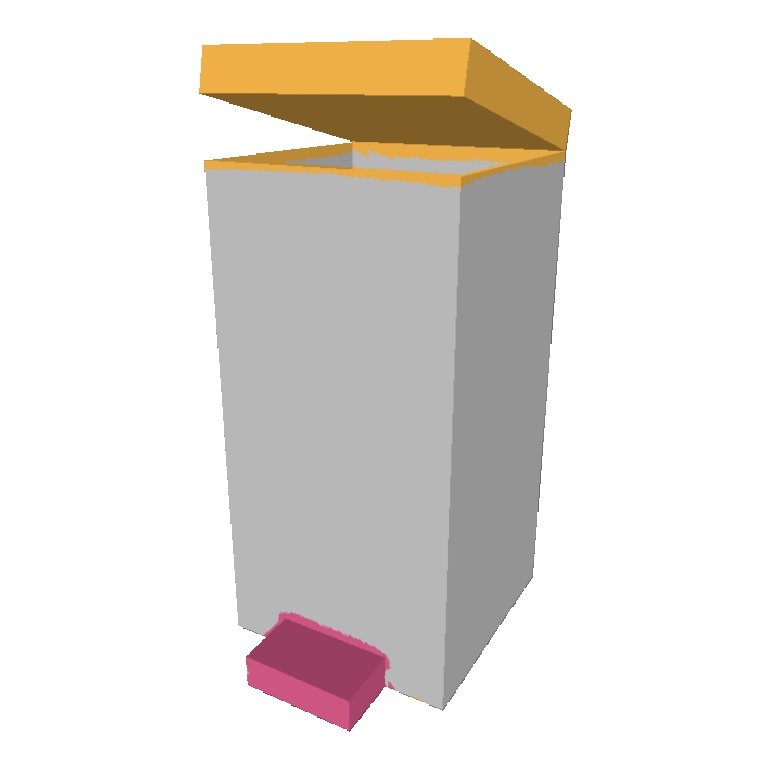} \\

\addlinespace[-2pt]
\arrayrulecolor{gray}\cmidrule(lr){1-5}
\arrayrulecolor{black}

\includegraphics[width=0.1\textwidth]{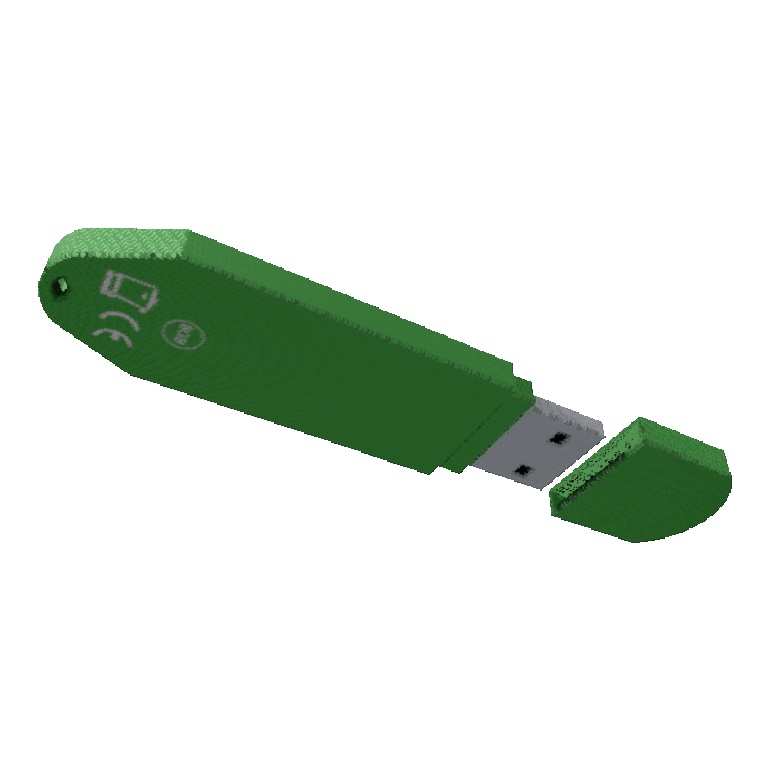} &
\includegraphics[width=0.1\textwidth]{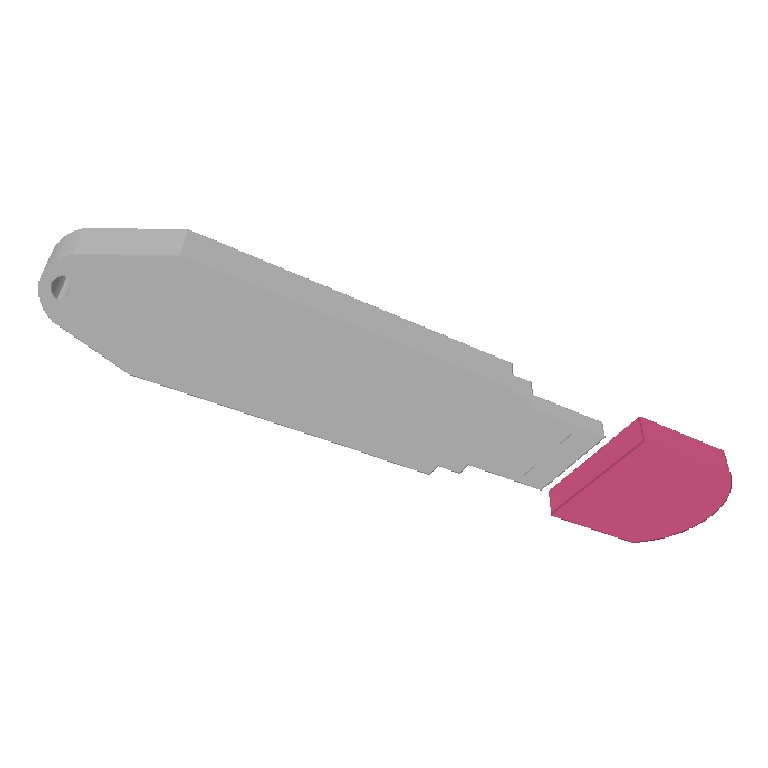} &
\includegraphics[width=0.1\textwidth]{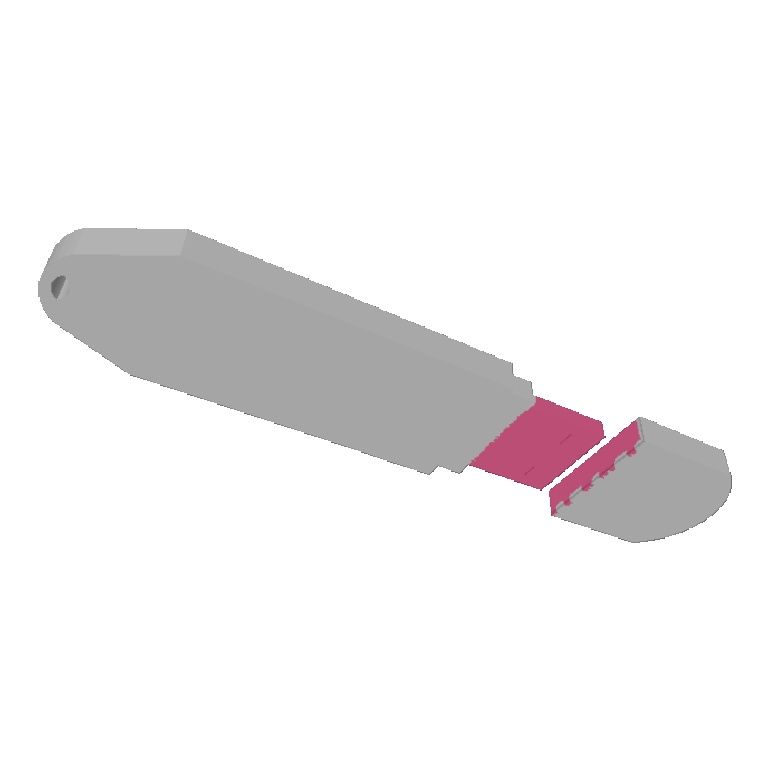} &
\includegraphics[width=0.1\textwidth]{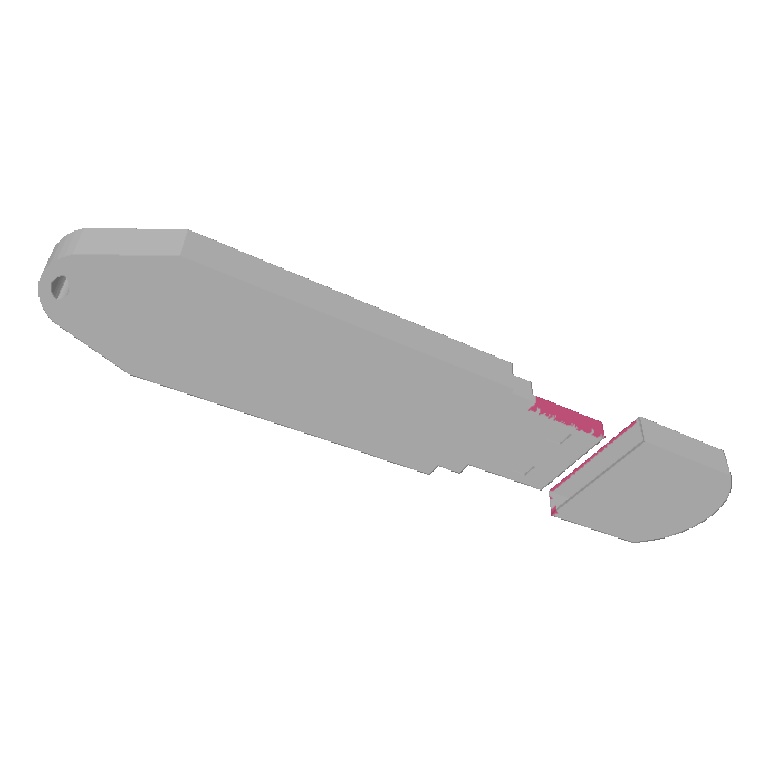} &
\includegraphics[width=0.1\textwidth]{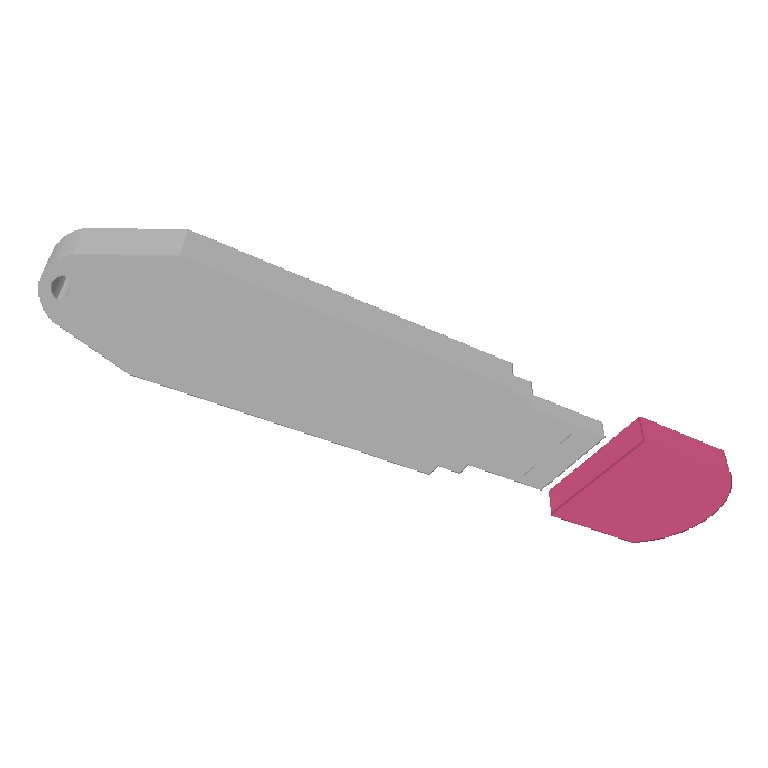} \\

\addlinespace[-2pt]
\arrayrulecolor{gray}\cmidrule(lr){1-5}
\arrayrulecolor{black}

\includegraphics[width=0.1\textwidth]{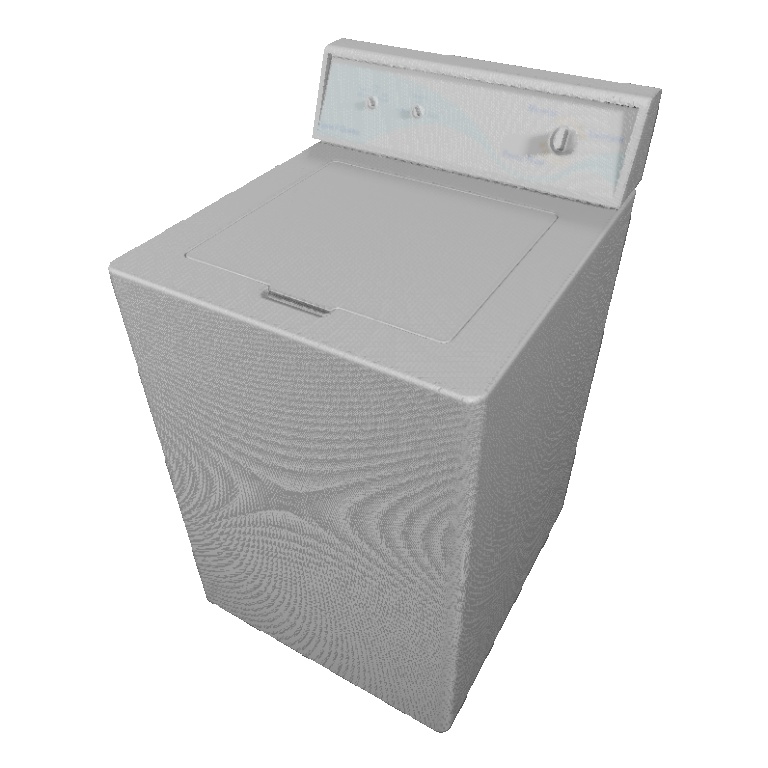} &
\includegraphics[width=0.1\textwidth]{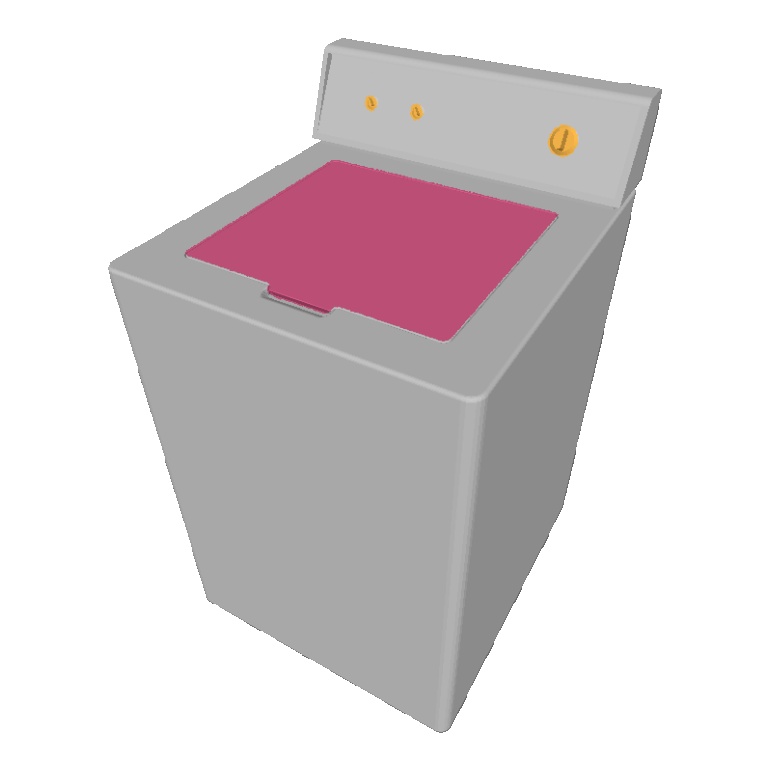} &
\includegraphics[width=0.1\textwidth]{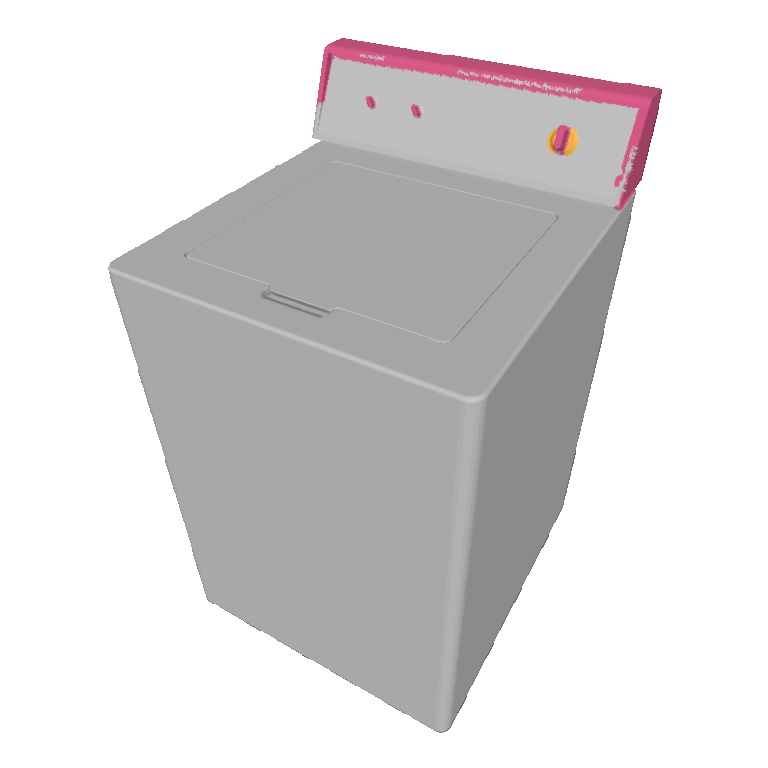} &
\includegraphics[width=0.1\textwidth]{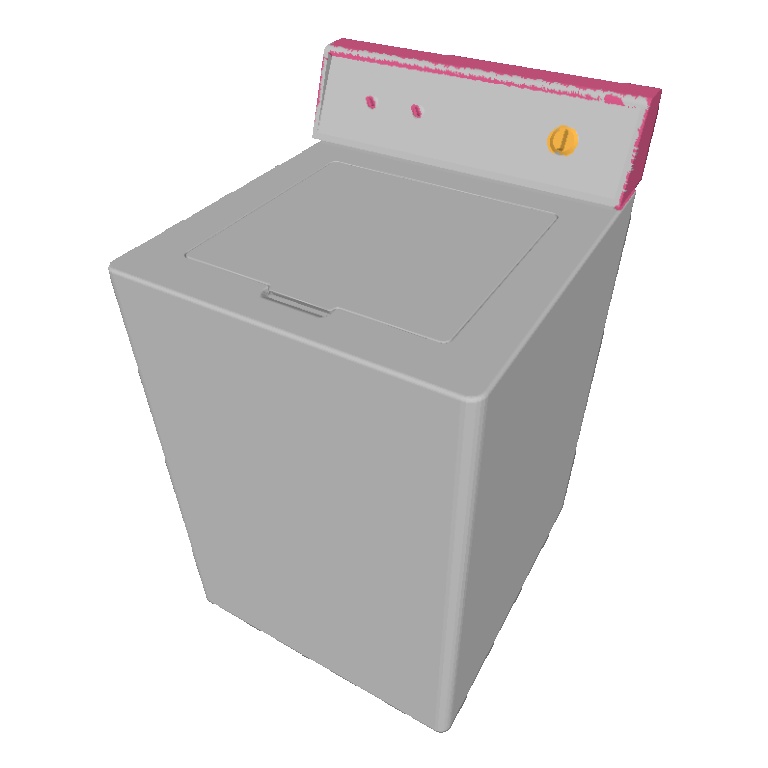} &
\includegraphics[width=0.1\textwidth]{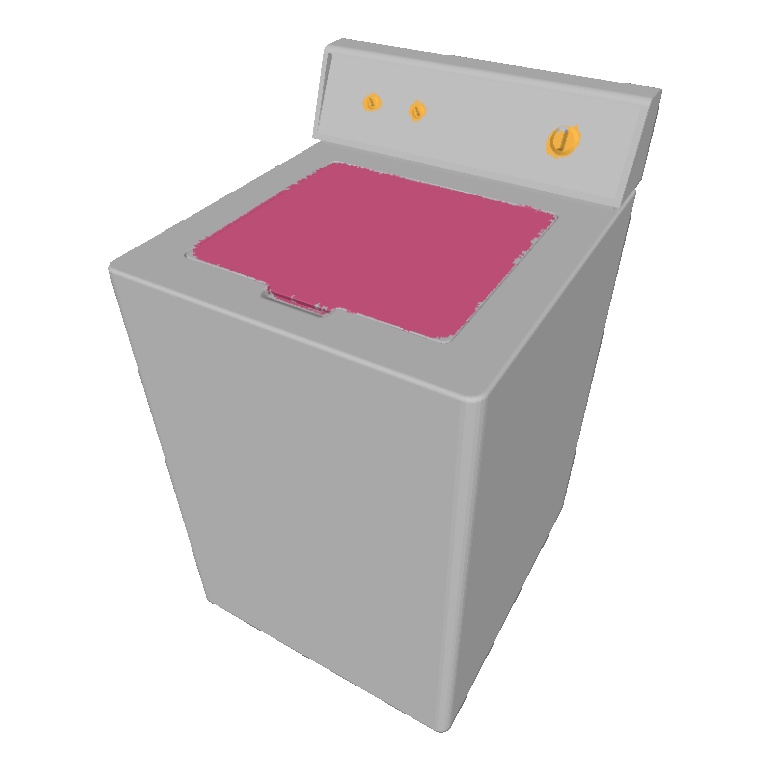} \\

\addlinespace[-2pt]
\arrayrulecolor{gray}\cmidrule(lr){1-5}
\arrayrulecolor{black}

\includegraphics[width=0.1\textwidth]{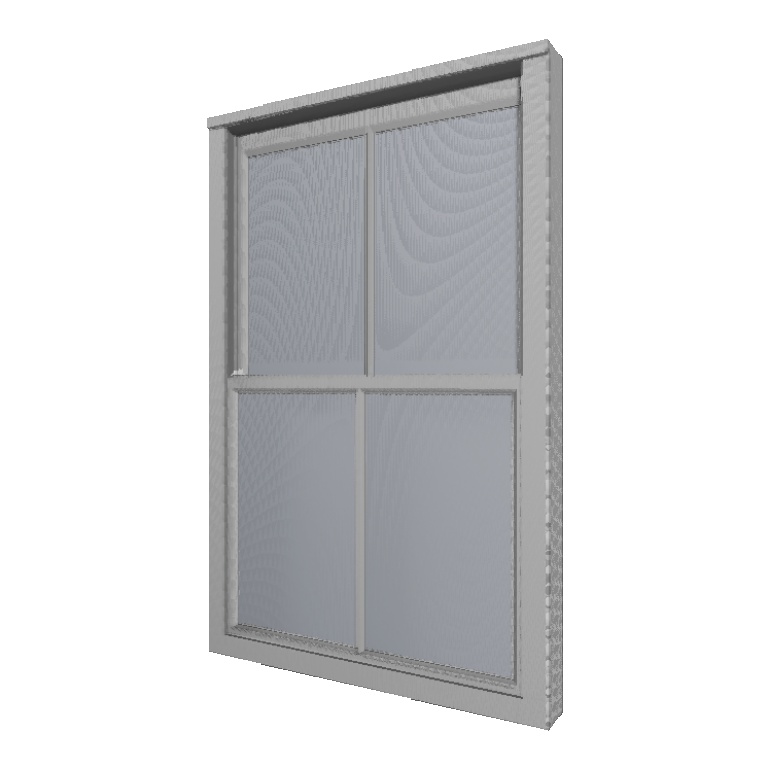} &
\includegraphics[width=0.1\textwidth]{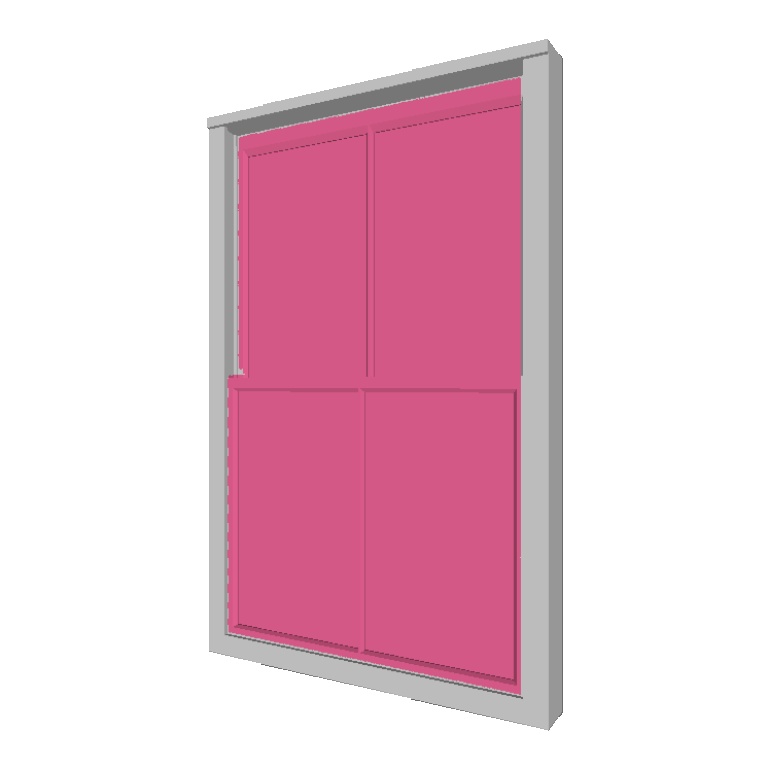} &
\includegraphics[width=0.1\textwidth]{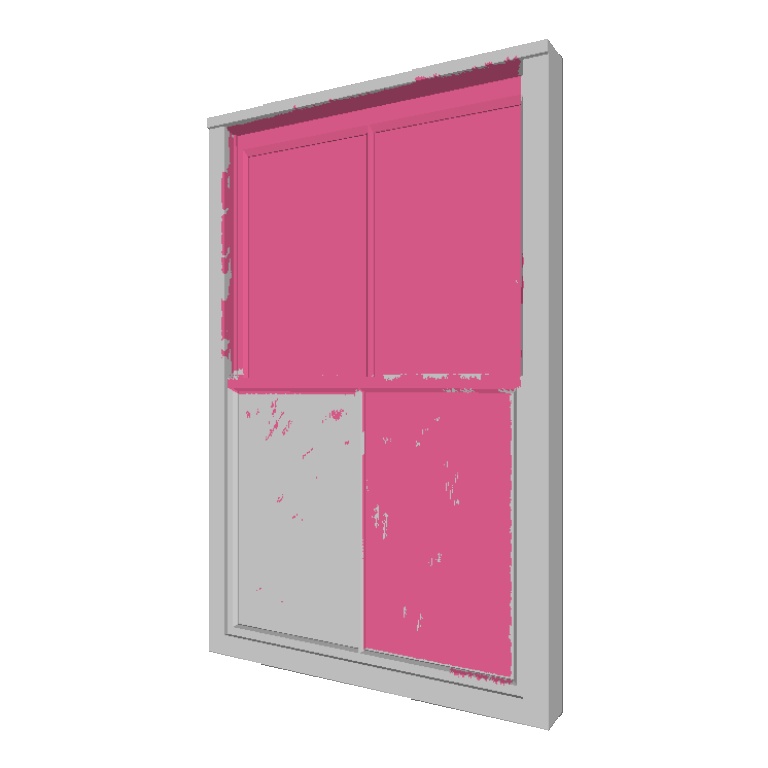} &
\includegraphics[width=0.1\textwidth]{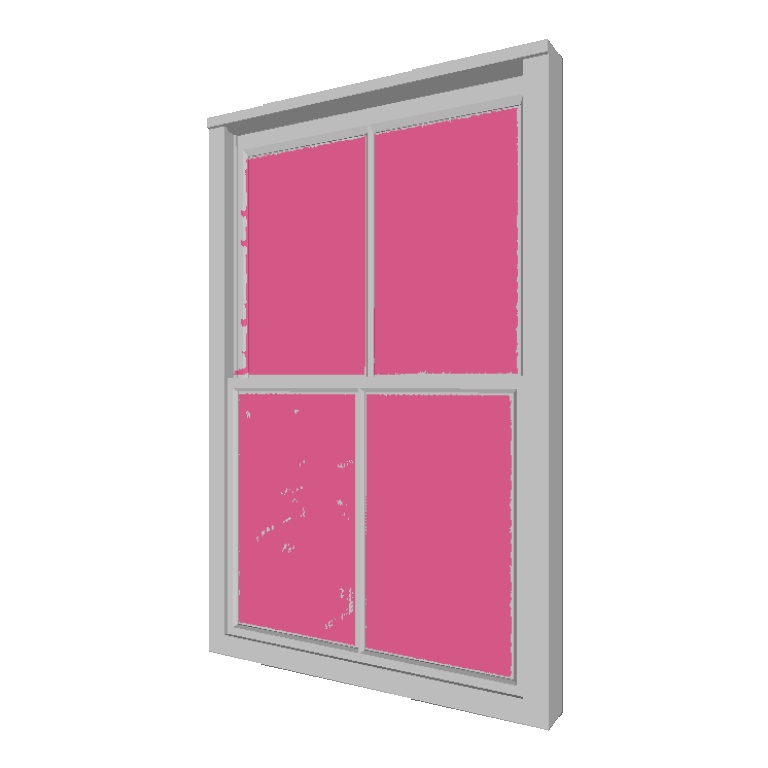} &
\includegraphics[width=0.1\textwidth]{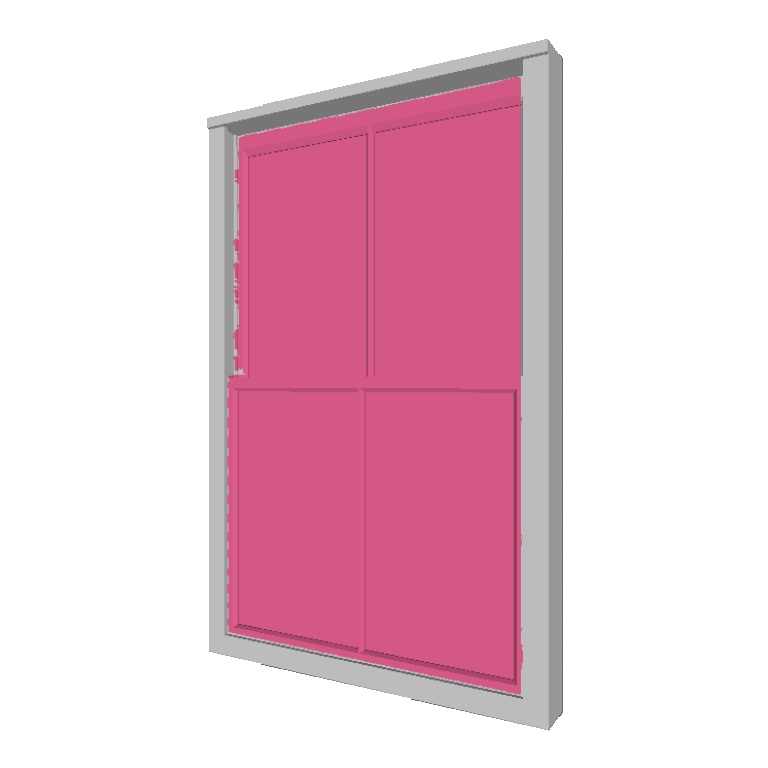} \\

\addlinespace[-2pt]
\arrayrulecolor{gray}\cmidrule(lr){1-5}
\arrayrulecolor{black}

\includegraphics[width=0.1\textwidth]{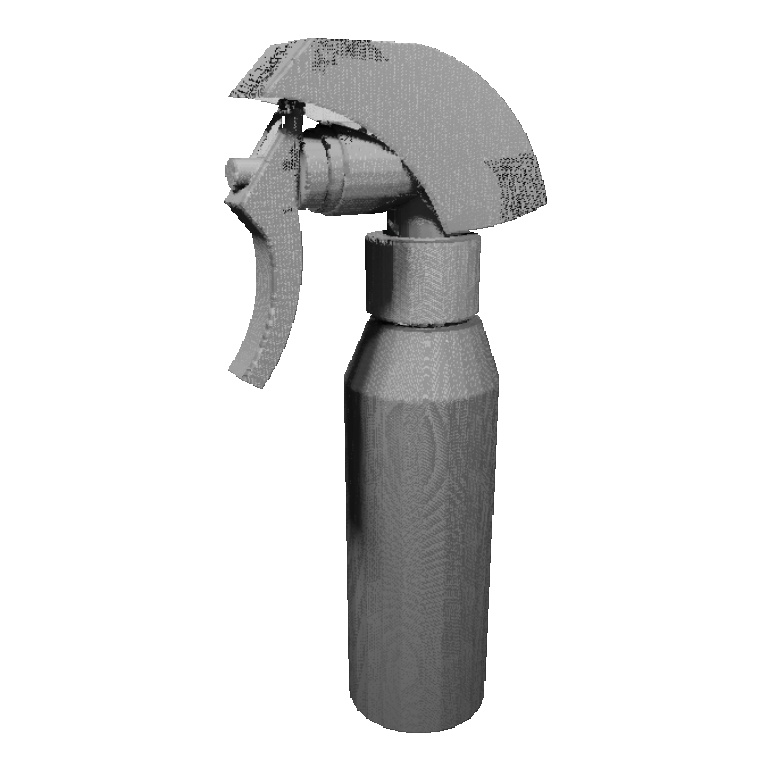} &
\includegraphics[width=0.1\textwidth]{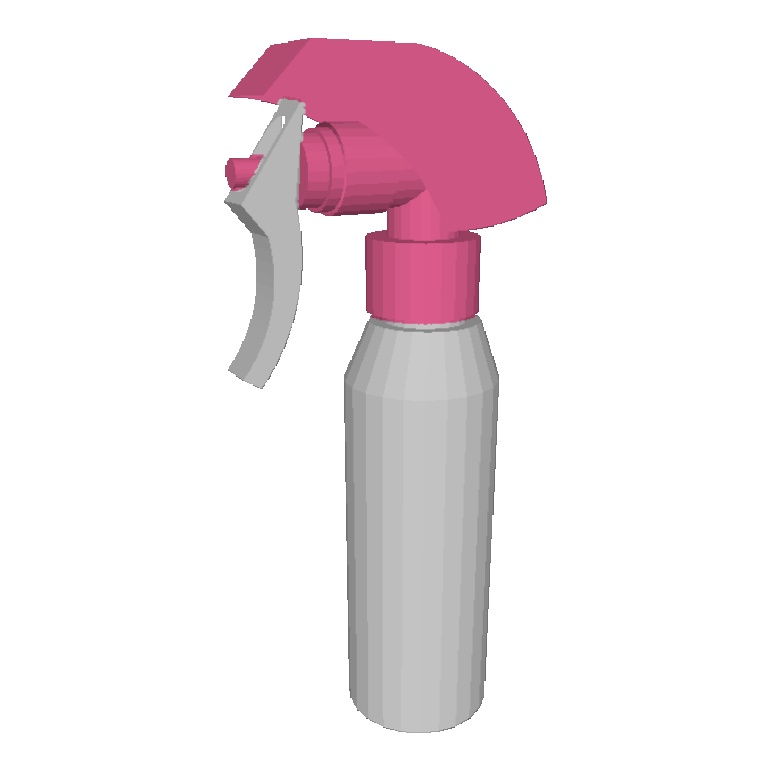} &
\includegraphics[width=0.1\textwidth]{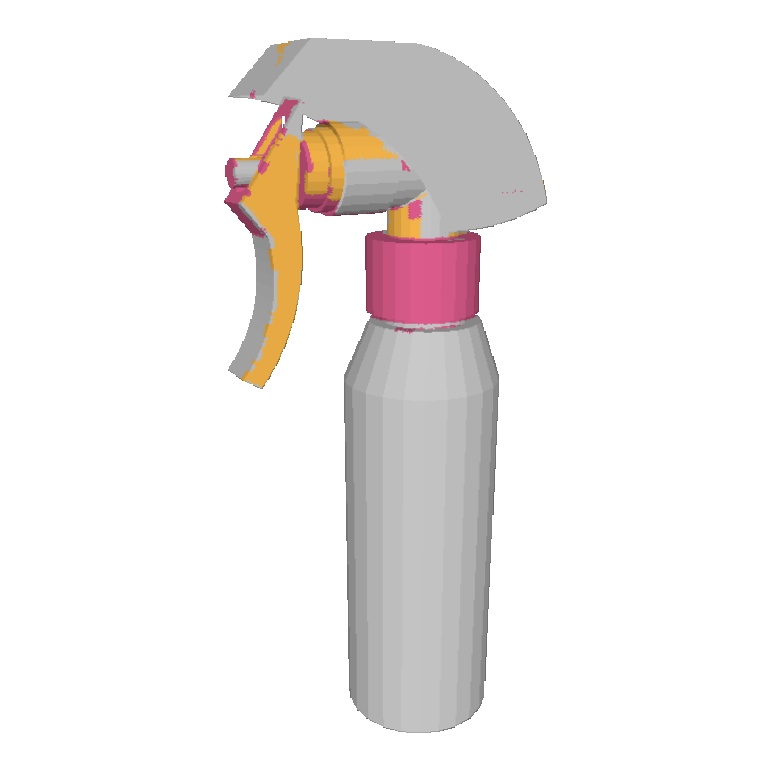} &
\includegraphics[width=0.1\textwidth]{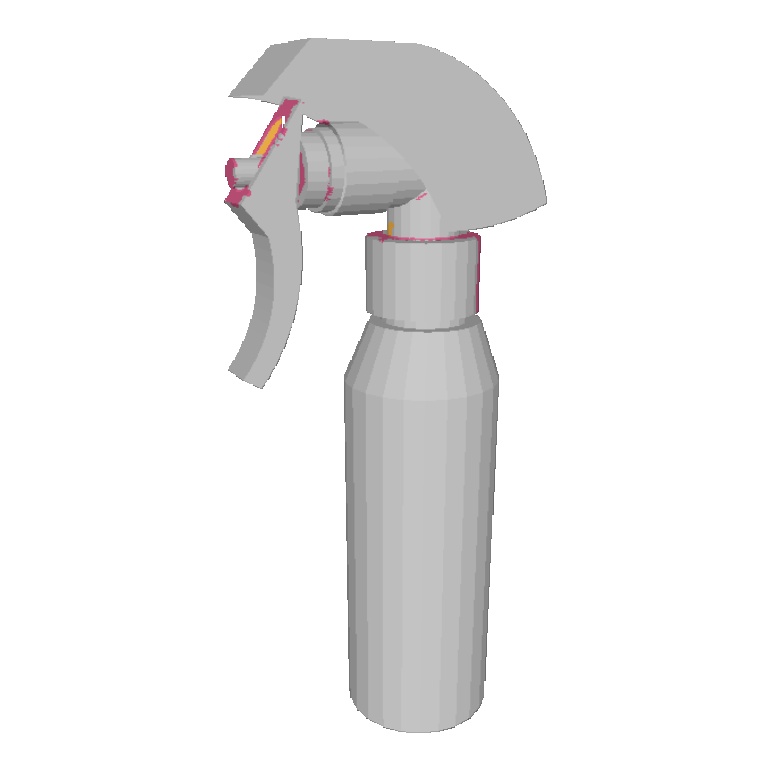} &
\includegraphics[width=0.1\textwidth]{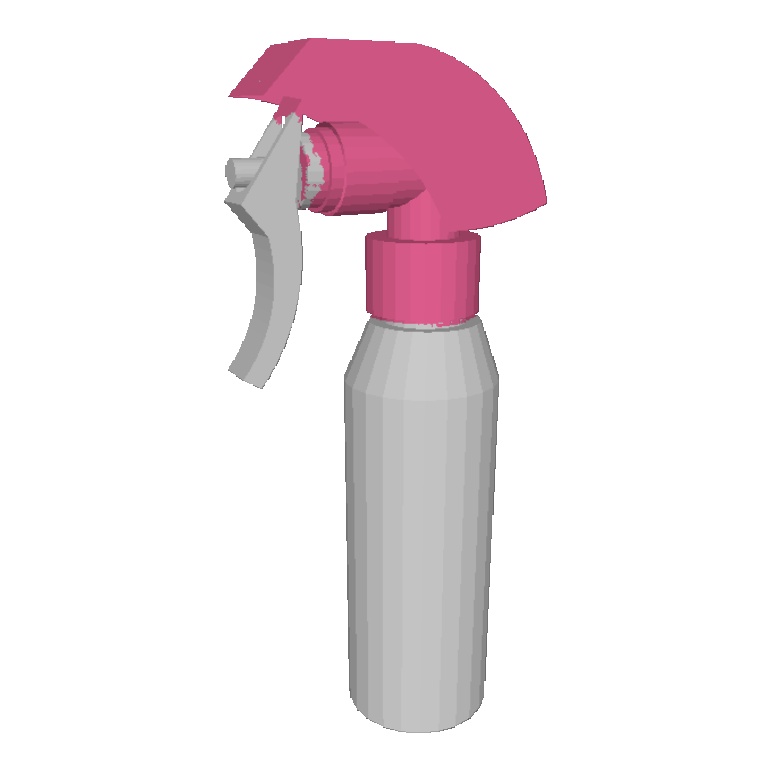} \\

\addlinespace[-2pt]
\arrayrulecolor{gray}\cmidrule(lr){1-5}
\arrayrulecolor{black}

\includegraphics[width=0.1\textwidth]{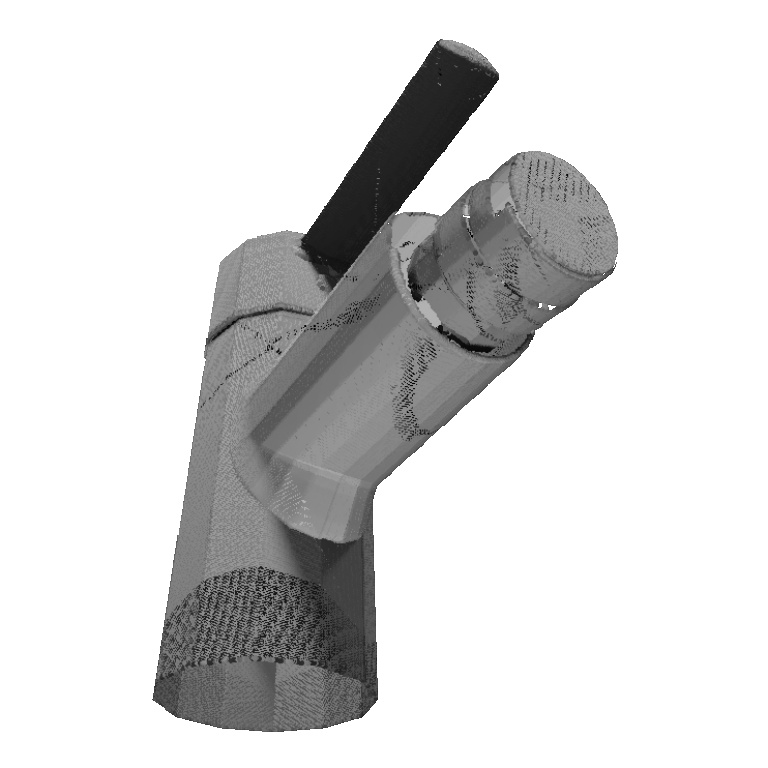} &
\includegraphics[width=0.1\textwidth]{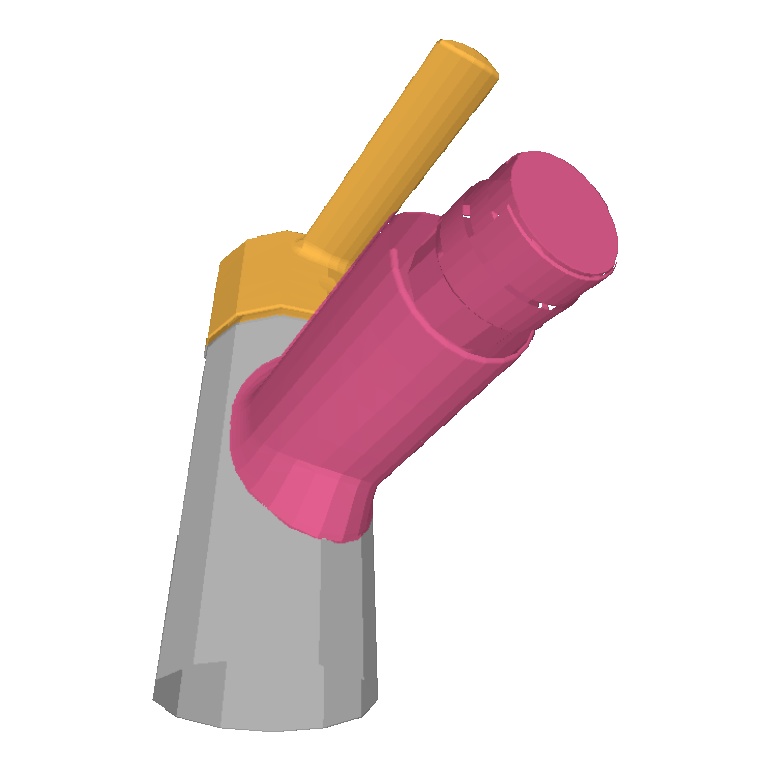} &
\includegraphics[width=0.1\textwidth]{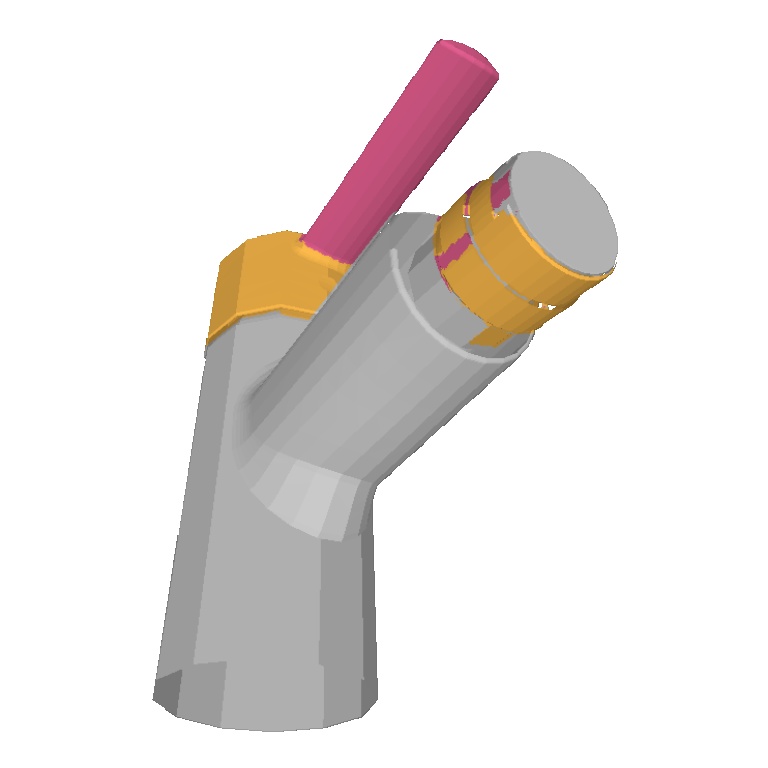} &
\includegraphics[width=0.1\textwidth]{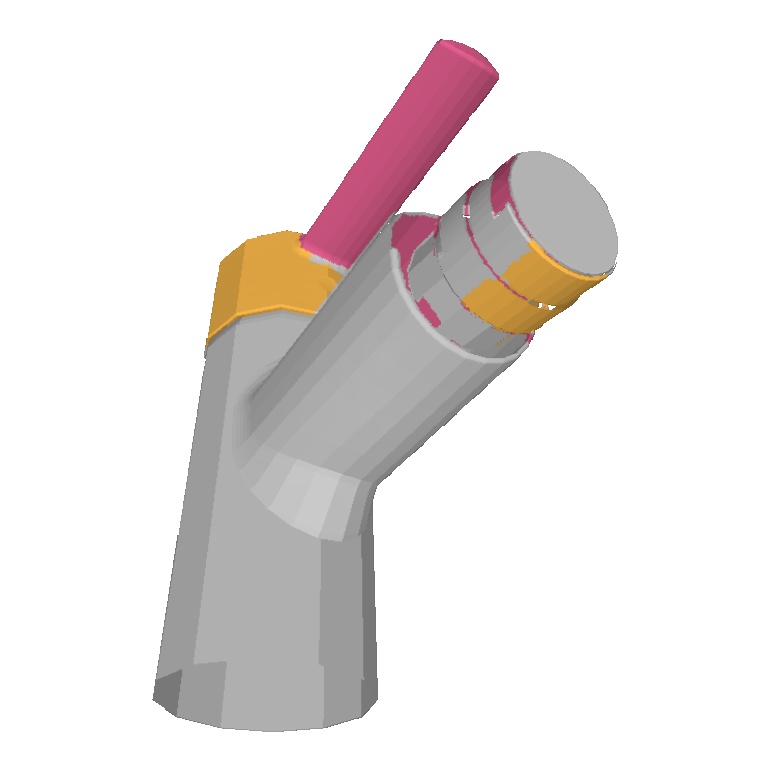} &
\includegraphics[width=0.1\textwidth]{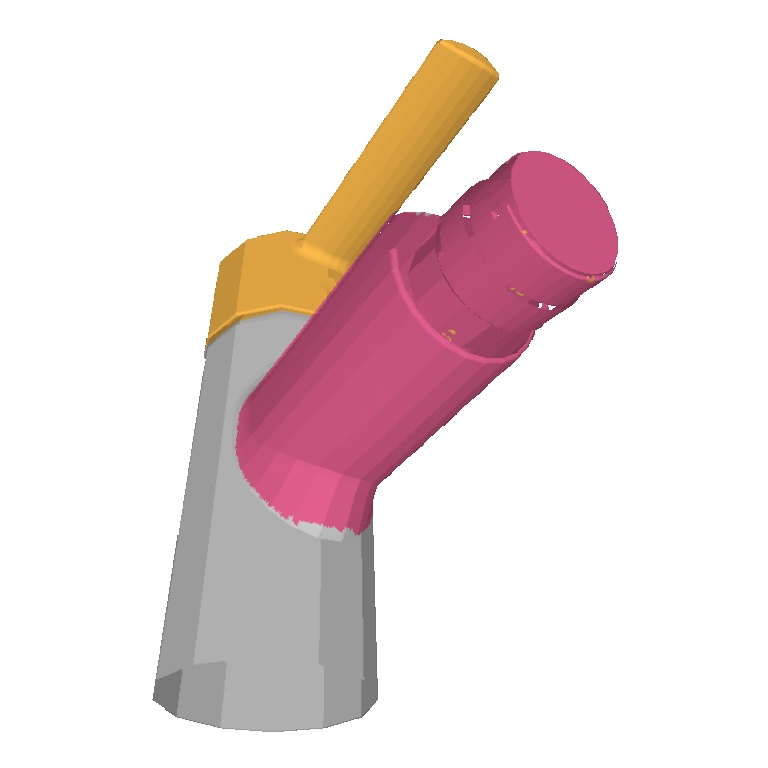} \\

\addlinespace[-2pt]
\bottomrule
\end{tabular}
}}
&
\vtop{\vskip0pt
\resizebox{0.5\textwidth}{!}{
\begin{tabular}{@{}c@{}c@{}c@{}c@{}c@{}}
\toprule
Input & GT  & PartSLIP & PartSTAD & Ours \\ \midrule
\includegraphics[width=0.1\textwidth]{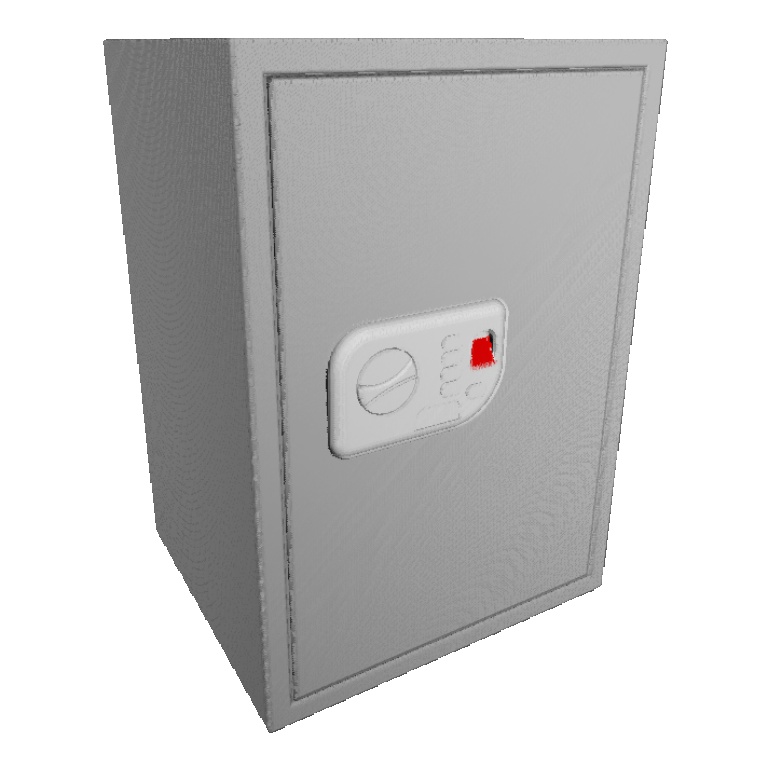} &
\includegraphics[width=0.1\textwidth]{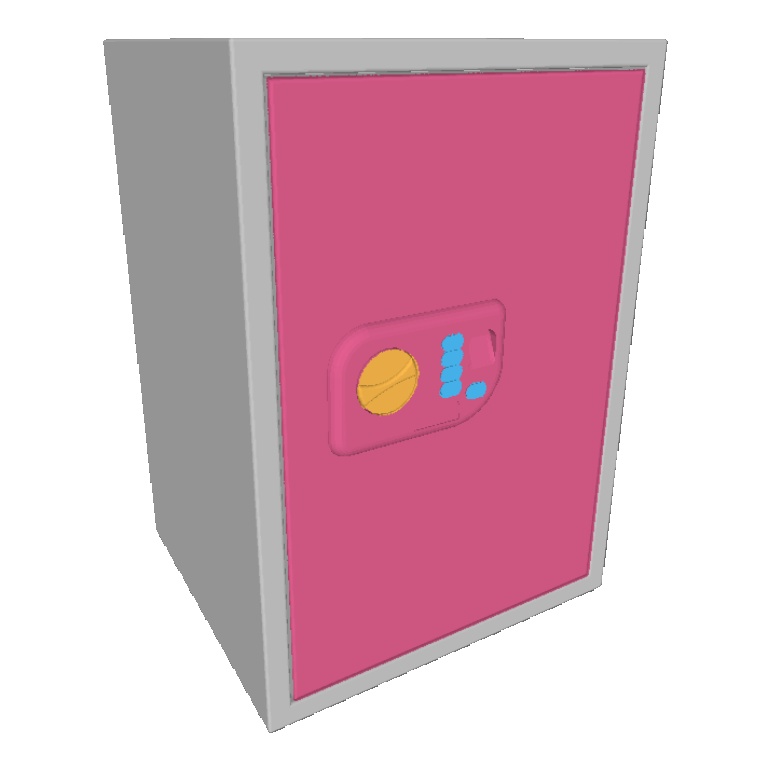} &
\includegraphics[width=0.1\textwidth]{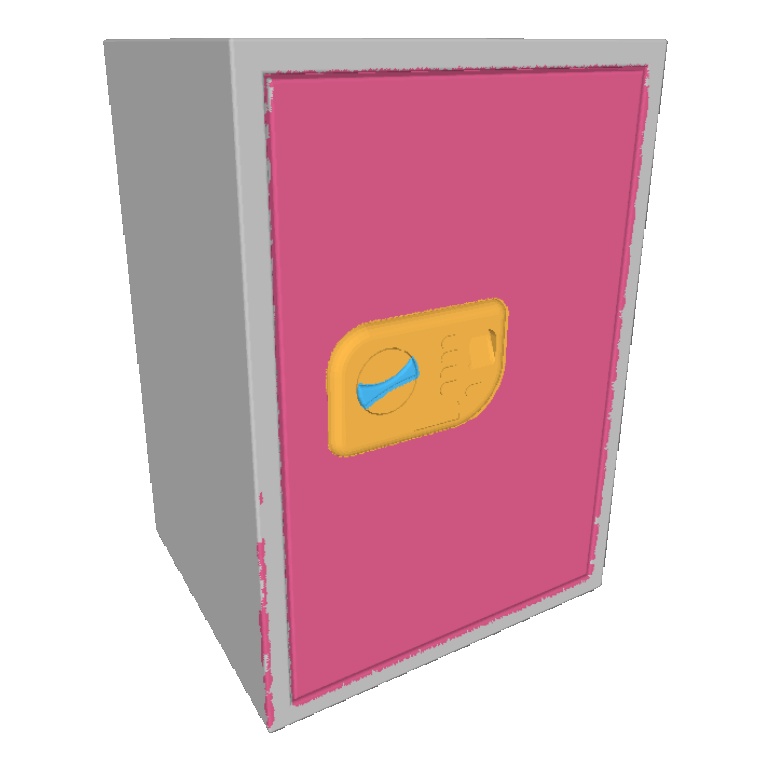} &
\includegraphics[width=0.1\textwidth]{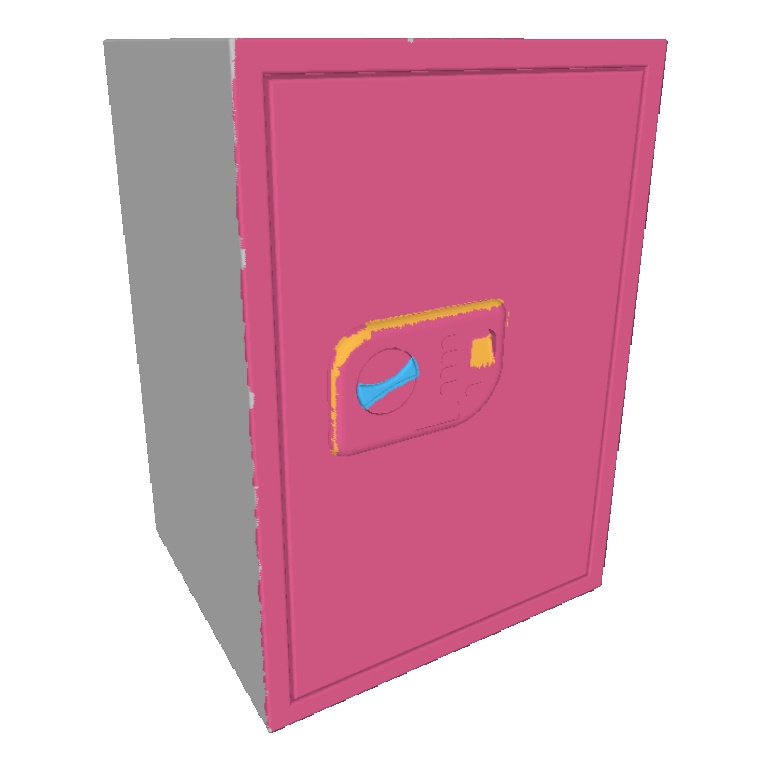} &
\includegraphics[width=0.1\textwidth]{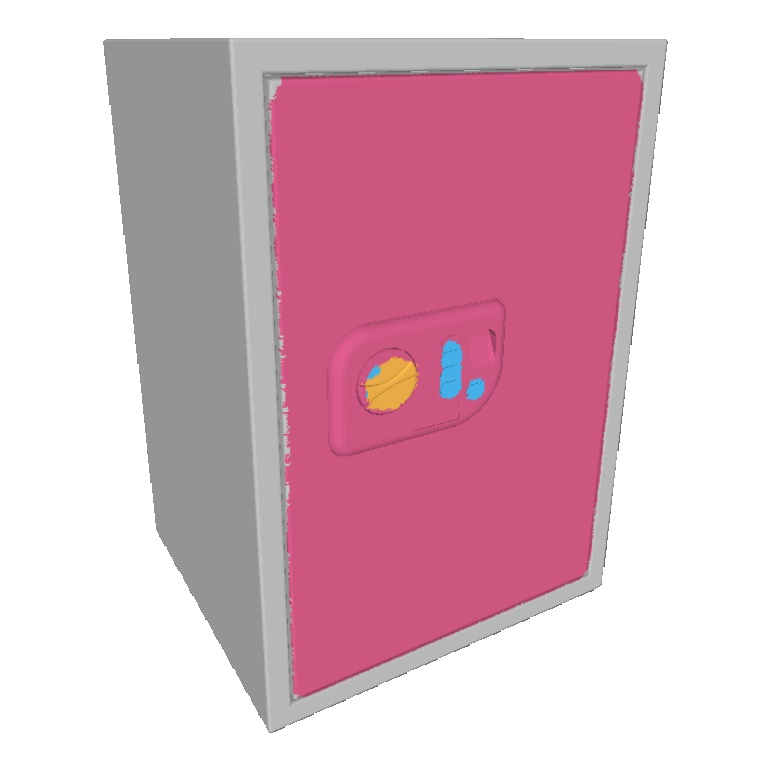} \\

\addlinespace[-2pt]
\arrayrulecolor{gray}\cmidrule(lr){1-5}
\arrayrulecolor{black}

\includegraphics[width=0.1\textwidth]{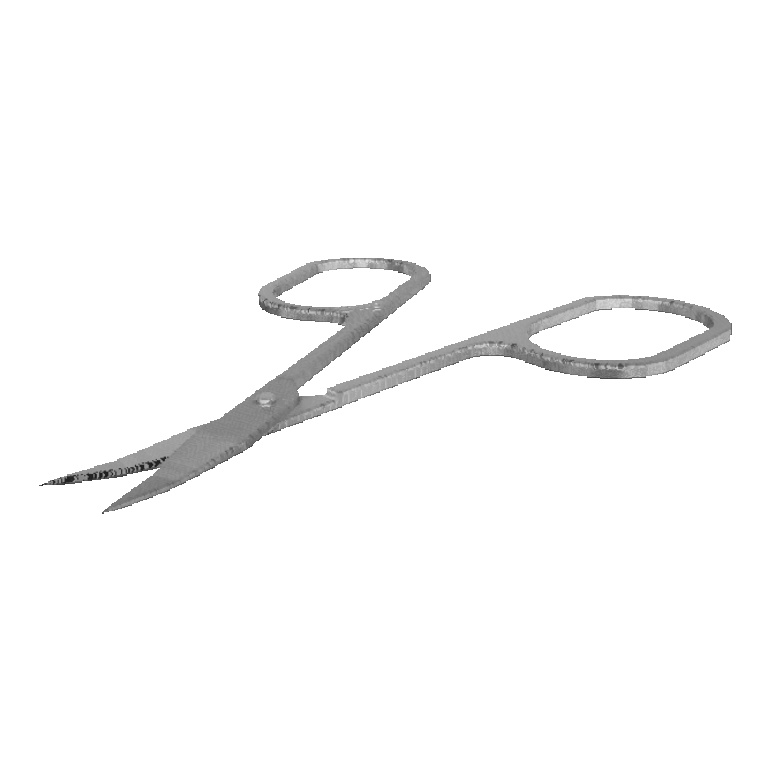} &
\includegraphics[width=0.1\textwidth]{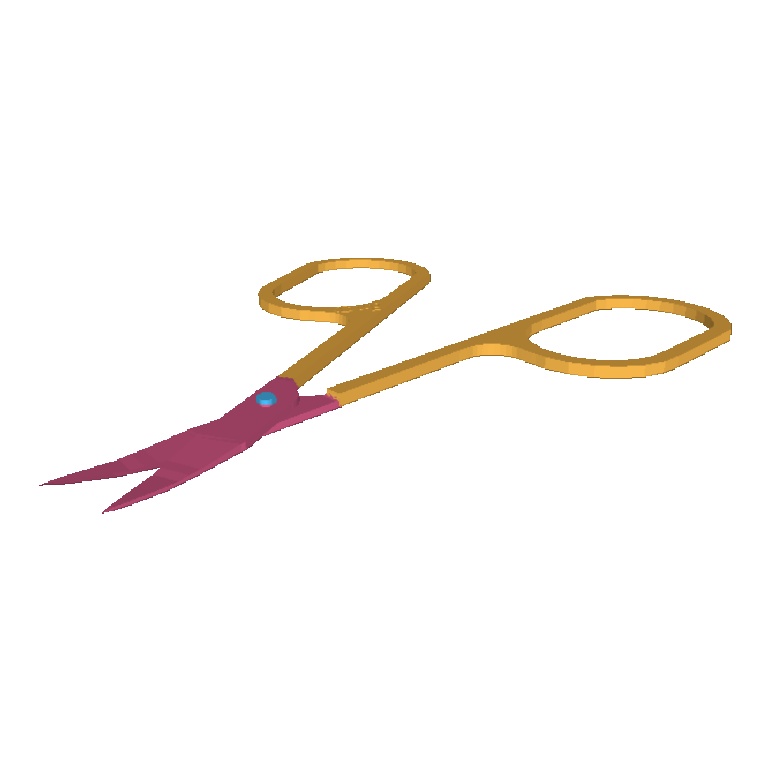} &
\includegraphics[width=0.1\textwidth]{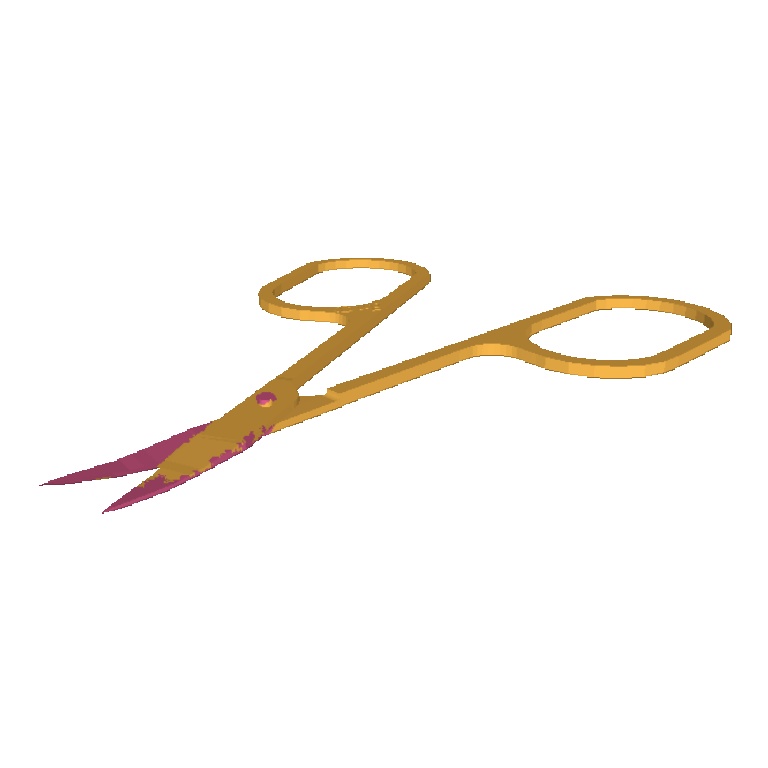} &
\includegraphics[width=0.1\textwidth]{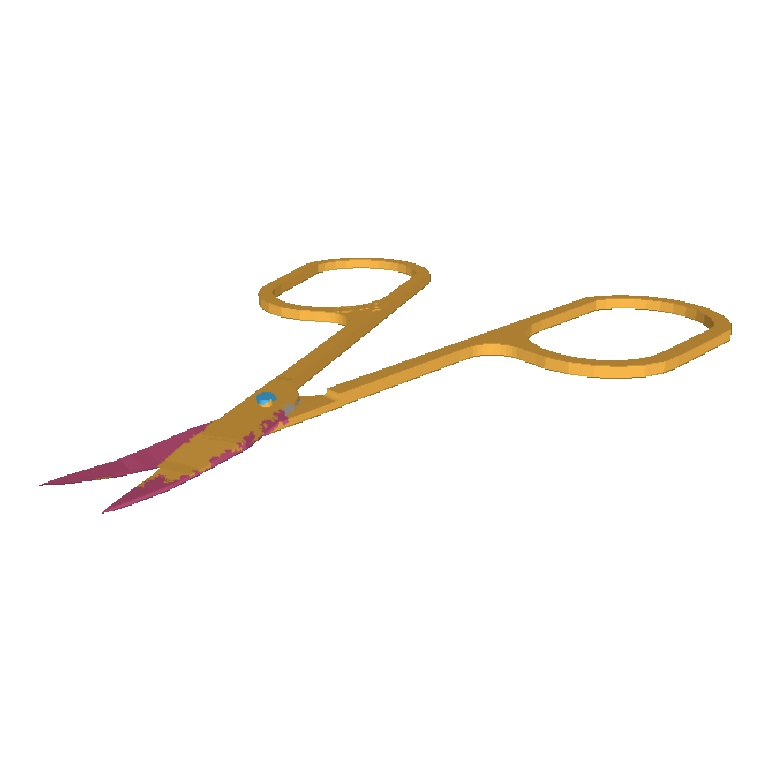} &
\includegraphics[width=0.1\textwidth]{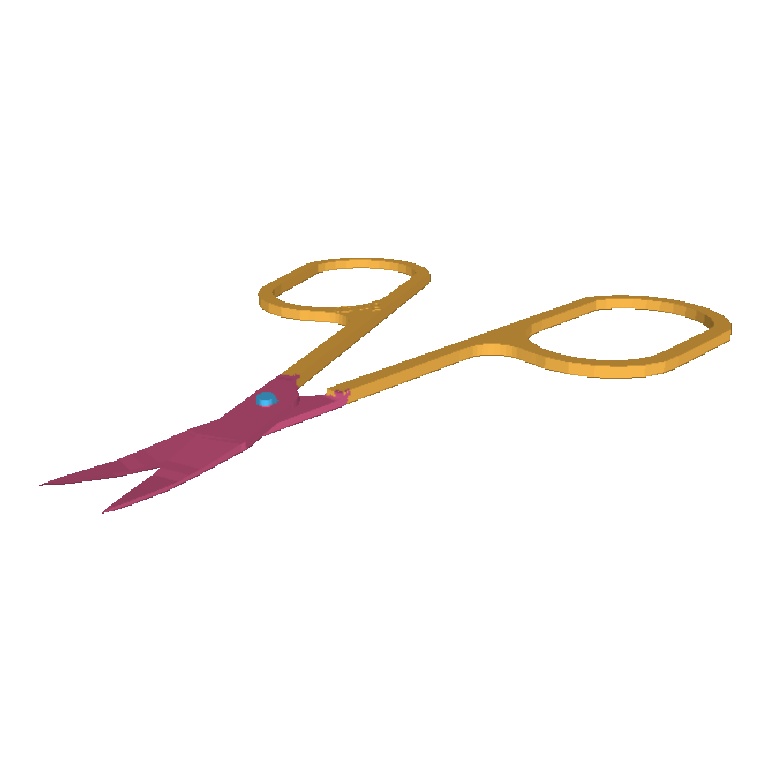} \\

\addlinespace[-2pt]
\arrayrulecolor{gray}\cmidrule(lr){1-5}
\arrayrulecolor{black}

\includegraphics[width=0.1\textwidth]{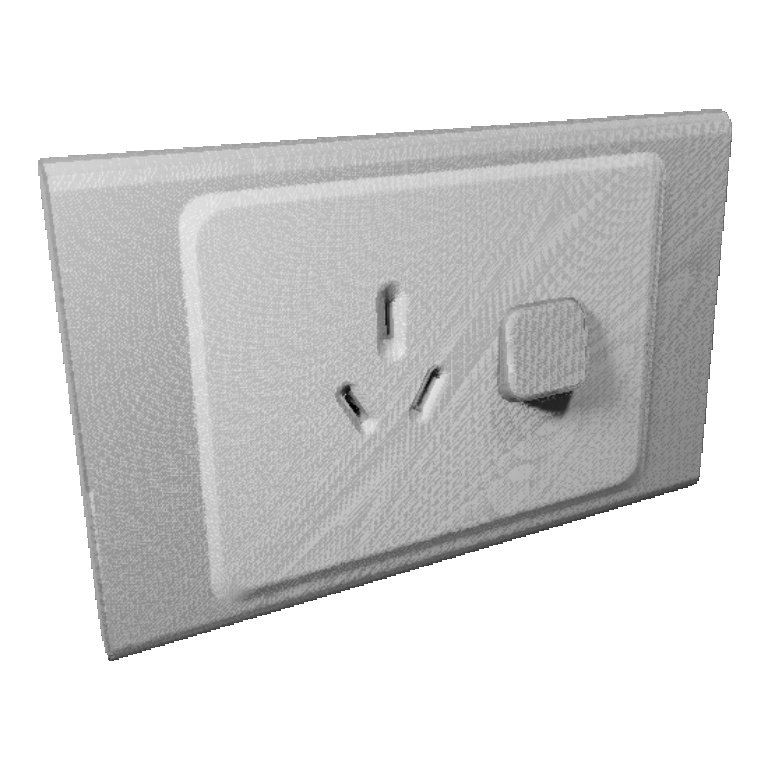} &
\includegraphics[width=0.1\textwidth]{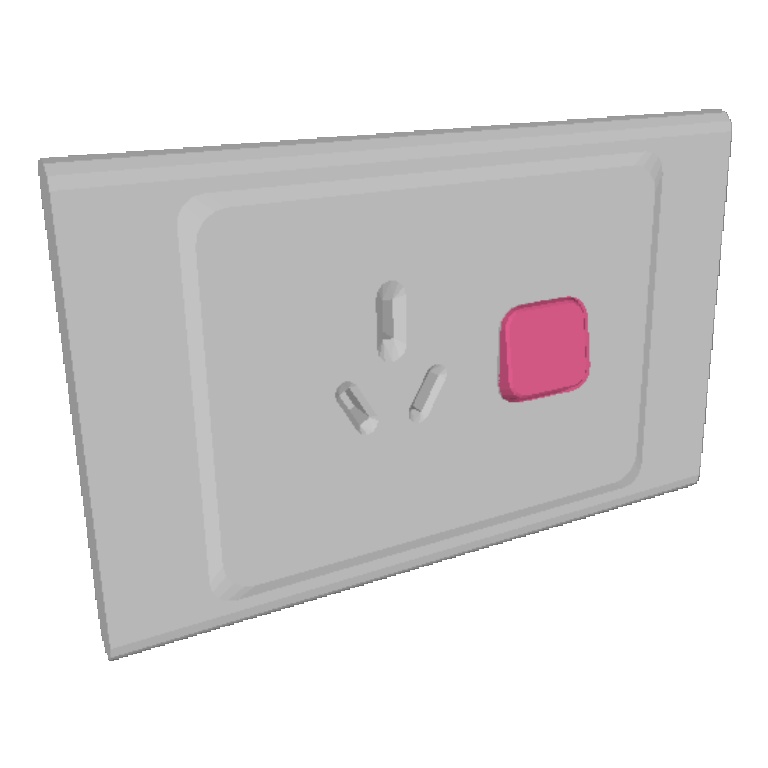} &
\includegraphics[width=0.1\textwidth]{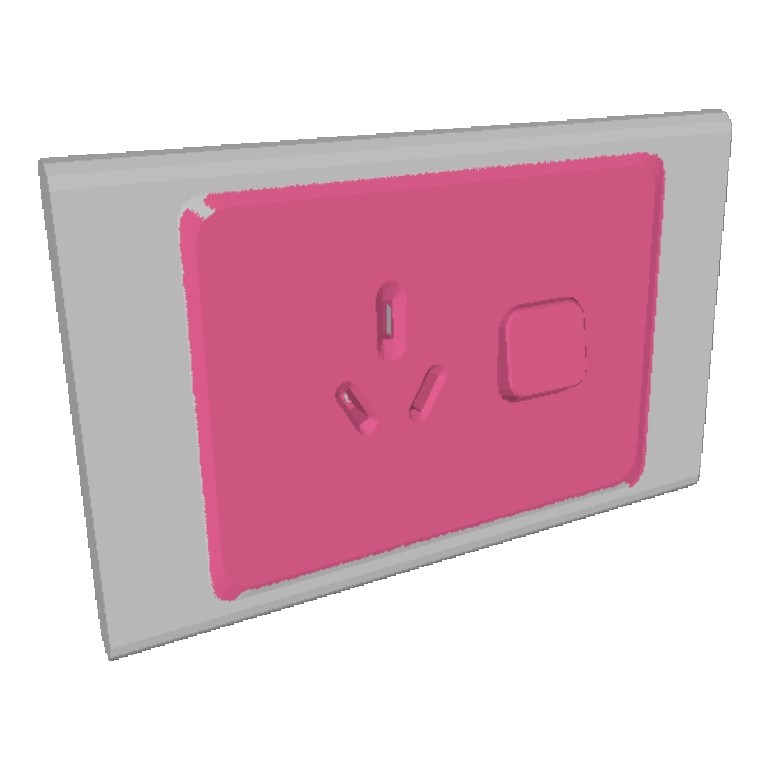} &
\includegraphics[width=0.1\textwidth]{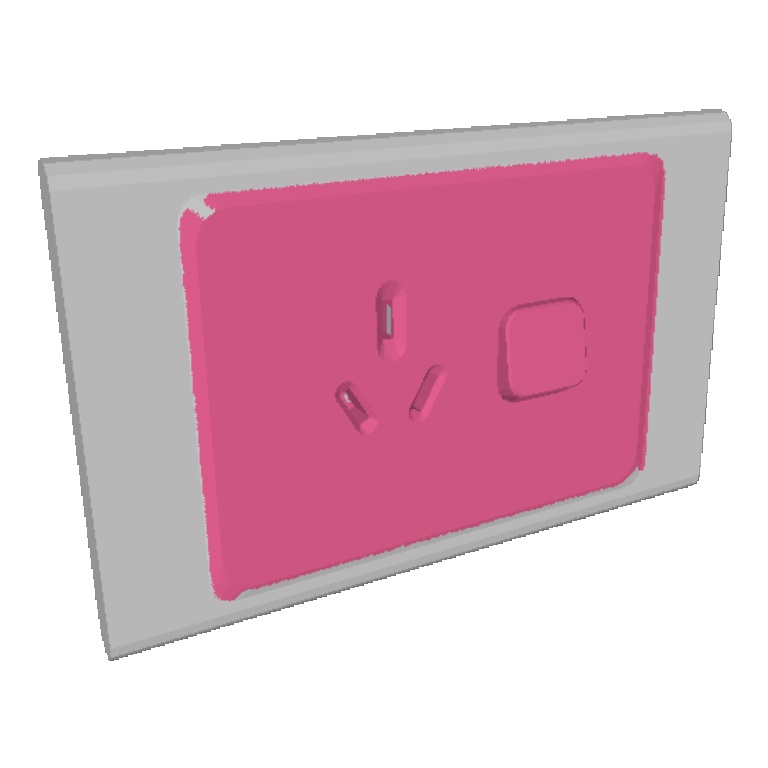} &
\includegraphics[width=0.1\textwidth]{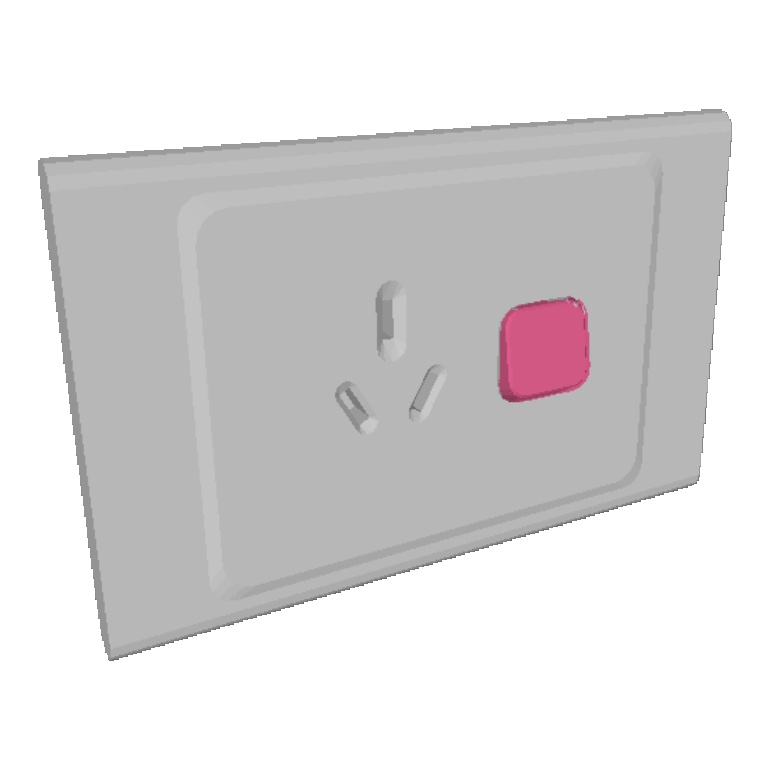} \\

\addlinespace[-2pt]
\arrayrulecolor{gray}\cmidrule(lr){1-5}
\arrayrulecolor{black}

\includegraphics[width=0.1\textwidth]{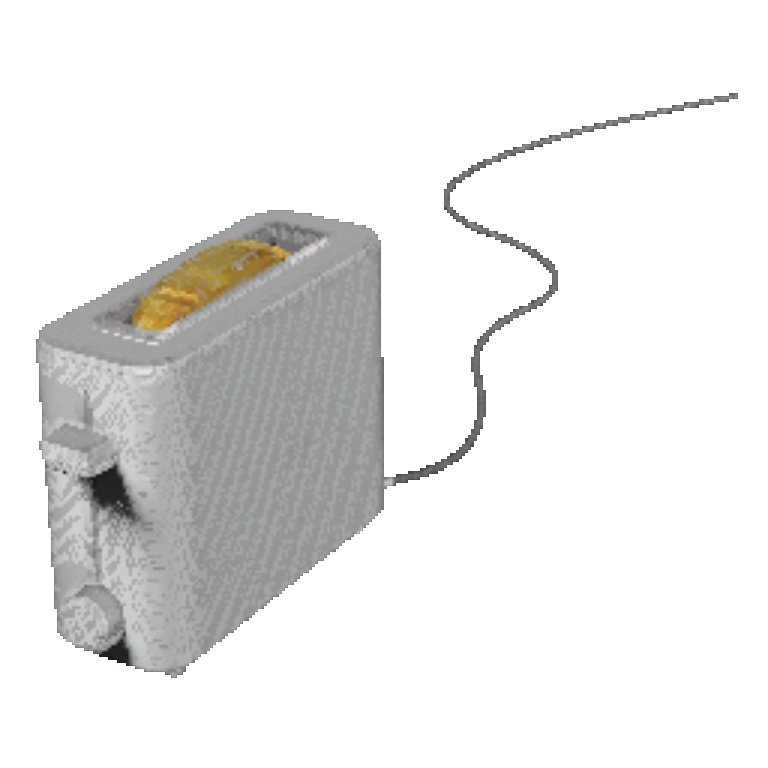} &
\includegraphics[width=0.1\textwidth]{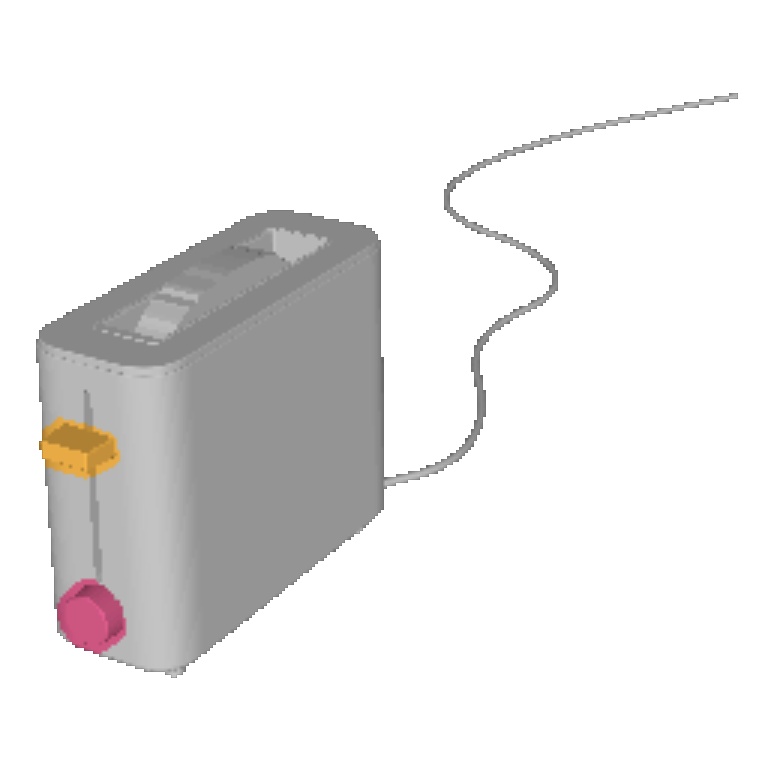} &
\includegraphics[width=0.1\textwidth]{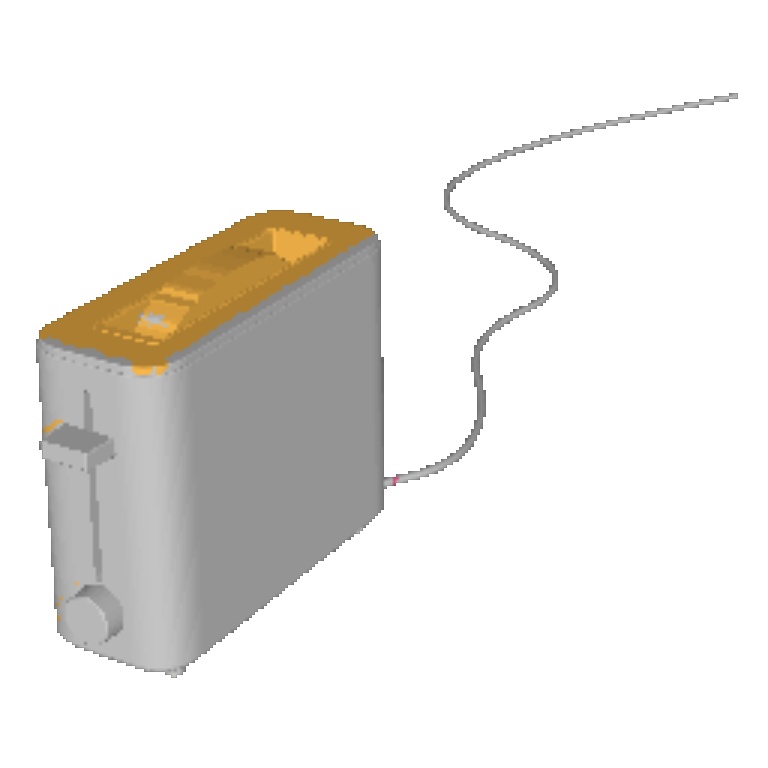} &
\includegraphics[width=0.1\textwidth]{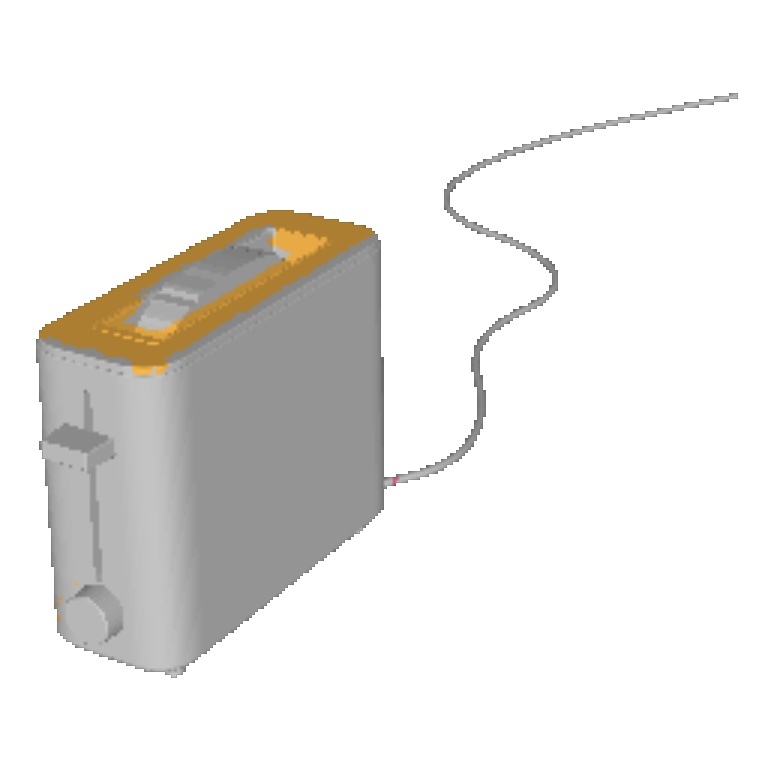} &
\includegraphics[width=0.1\textwidth]{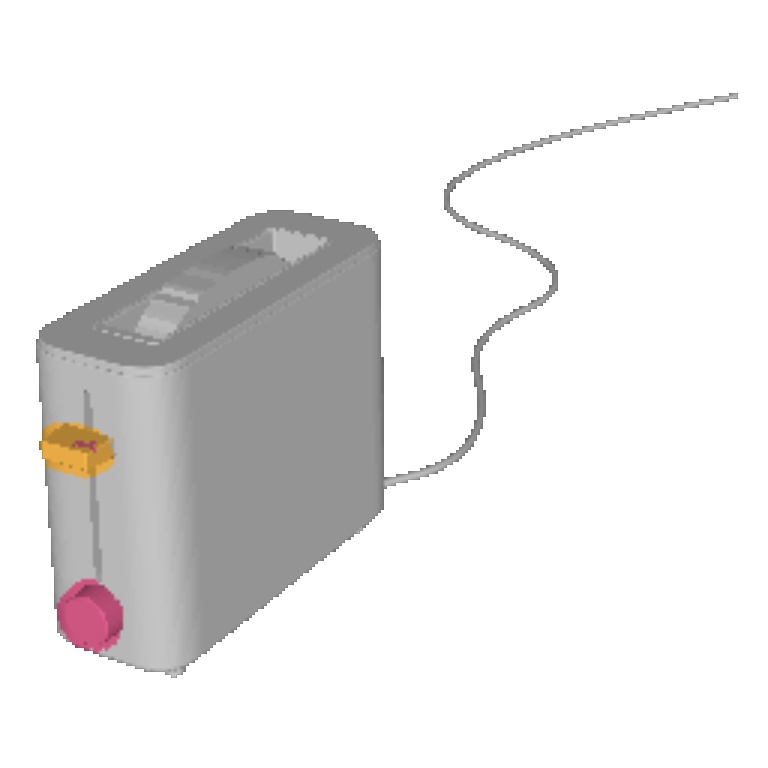} \\

\addlinespace[-2pt]
\arrayrulecolor{gray}\cmidrule(lr){1-5}
\arrayrulecolor{black}

\includegraphics[width=0.1\textwidth]{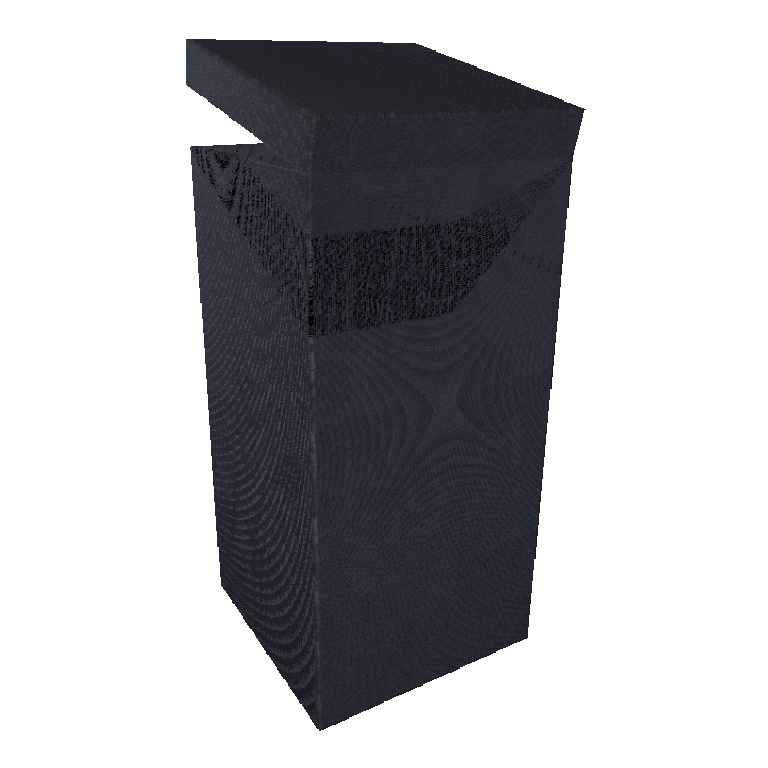} &
\includegraphics[width=0.1\textwidth]{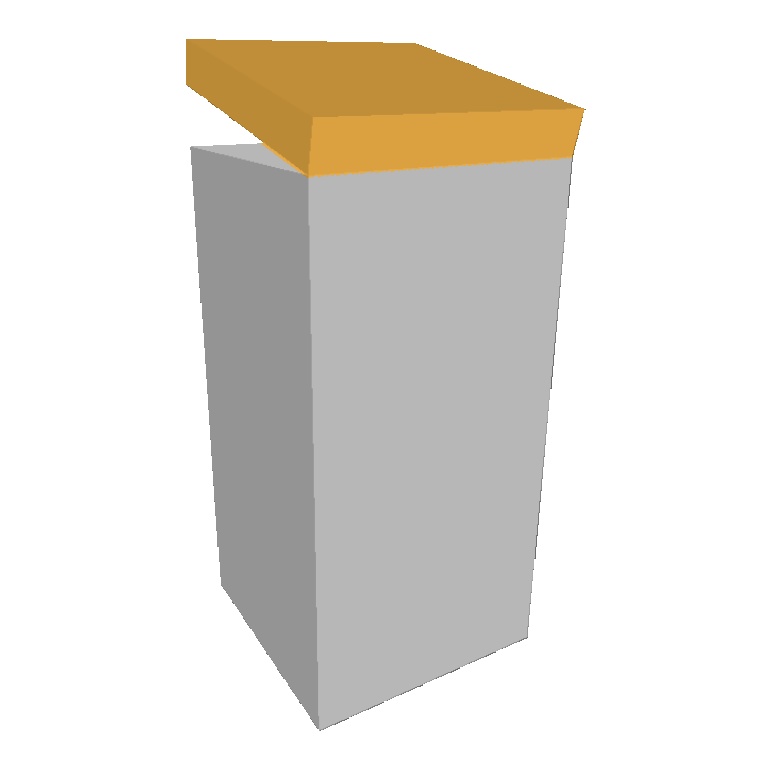} &
\includegraphics[width=0.1\textwidth]{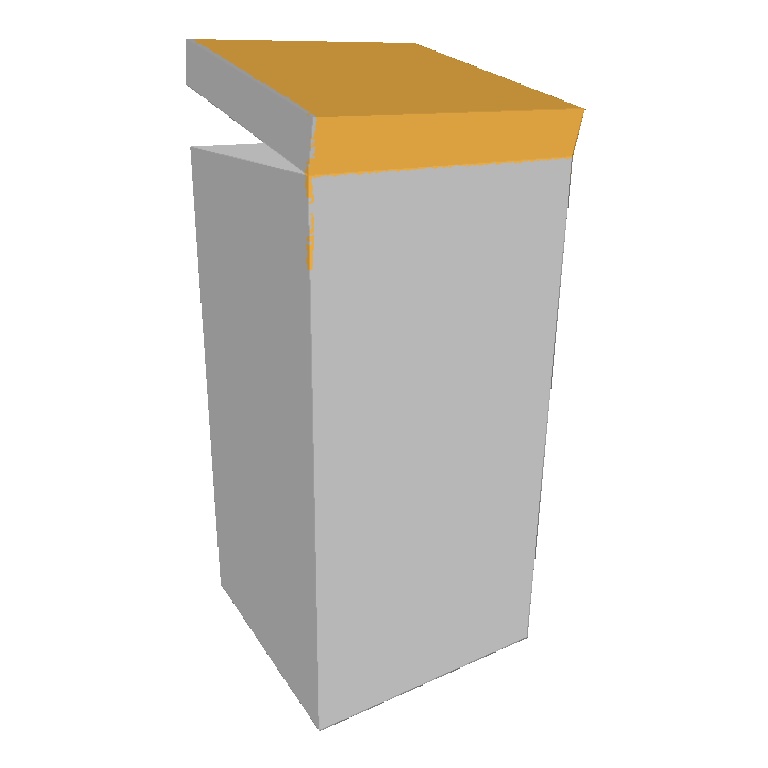} &
\includegraphics[width=0.1\textwidth]{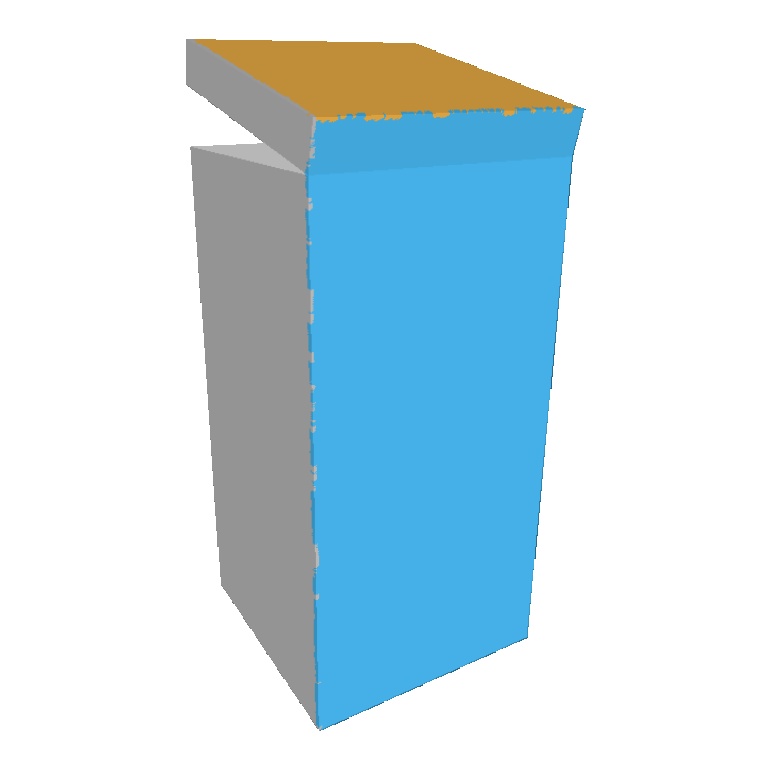} &
\includegraphics[width=0.1\textwidth]{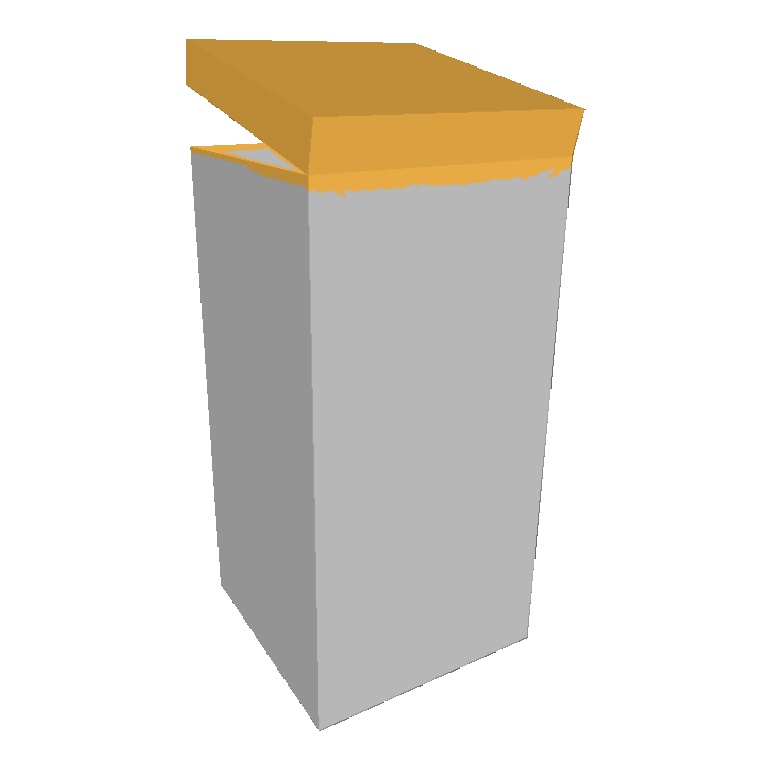} \\

\addlinespace[-2pt]
\arrayrulecolor{gray}\cmidrule(lr){1-5}
\arrayrulecolor{black}

\includegraphics[width=0.1\textwidth]{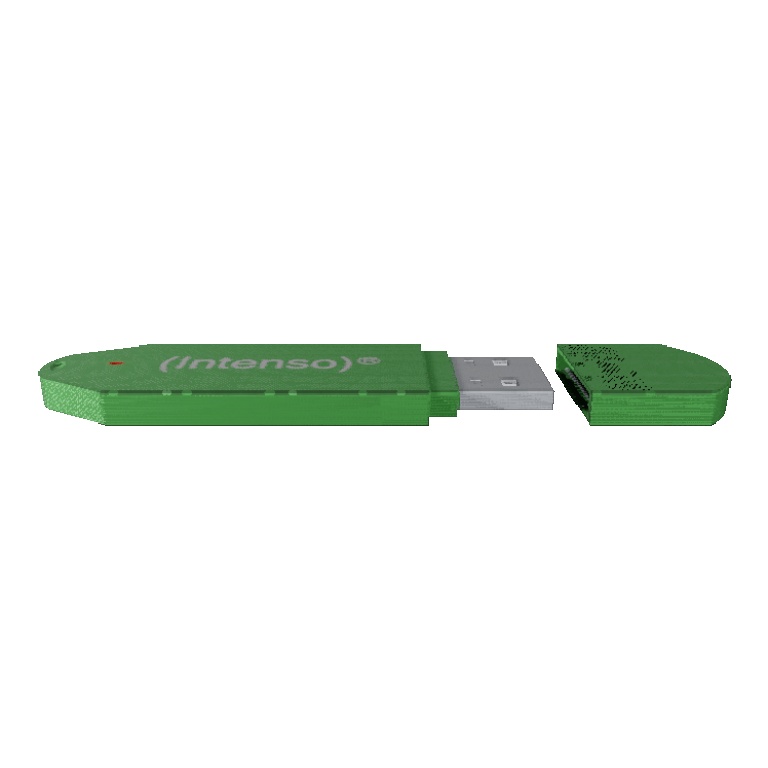} &
\includegraphics[width=0.1\textwidth]{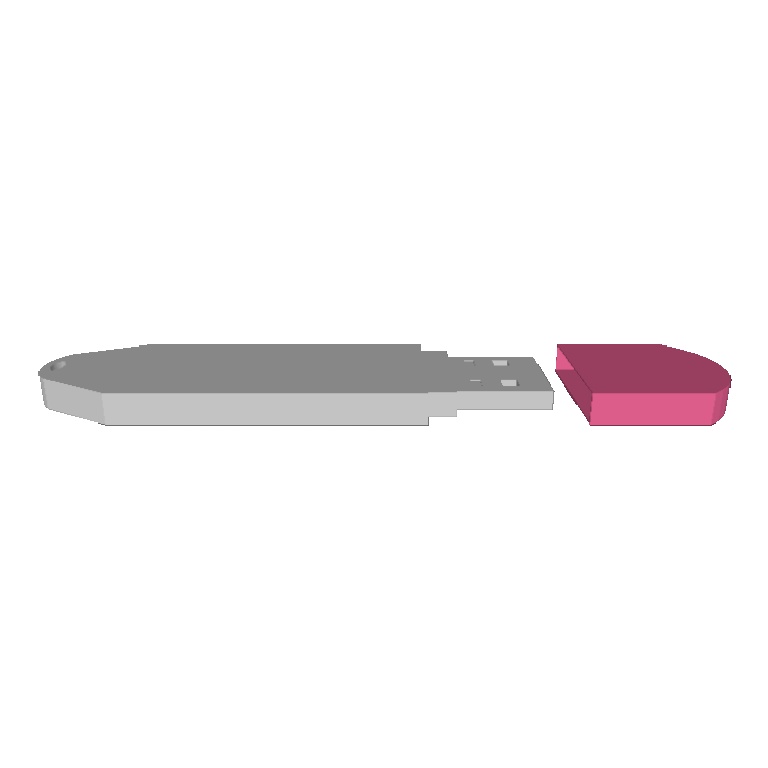} &
\includegraphics[width=0.1\textwidth]{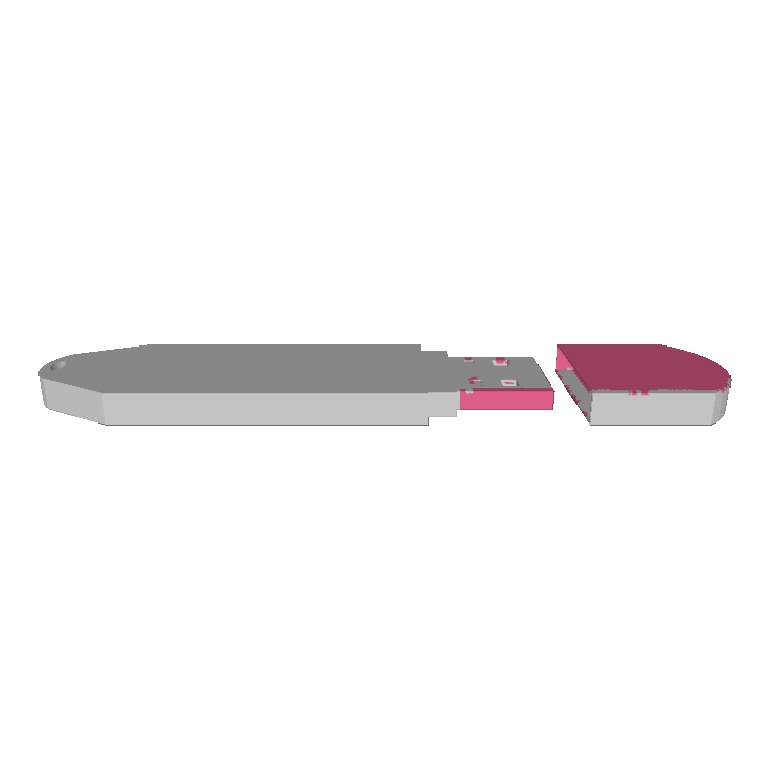} &
\includegraphics[width=0.1\textwidth]{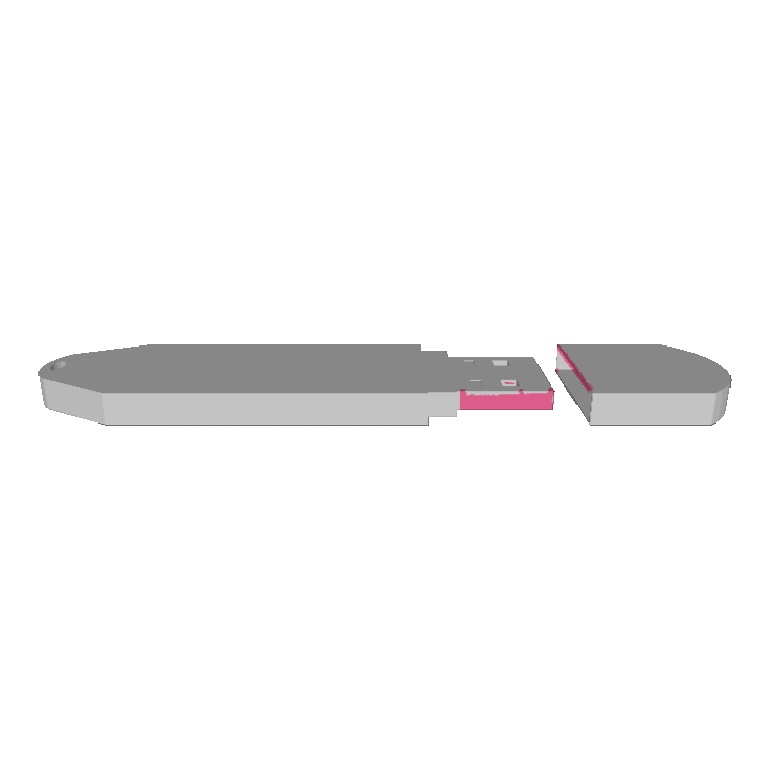} &
\includegraphics[width=0.1\textwidth]{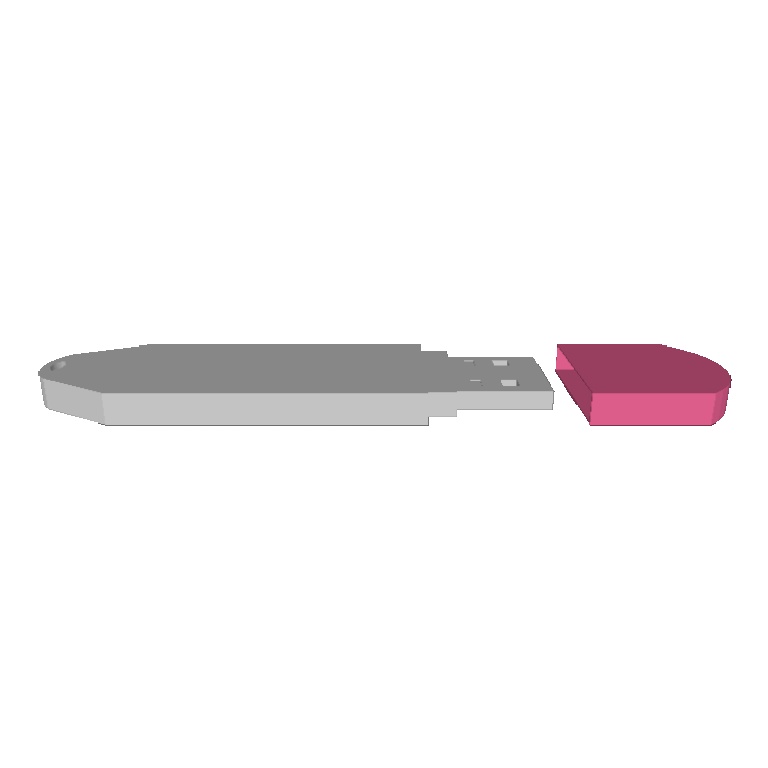} \\

\addlinespace[-2pt]
\arrayrulecolor{gray}\cmidrule(lr){1-5}
\arrayrulecolor{black}

\includegraphics[width=0.1\textwidth]{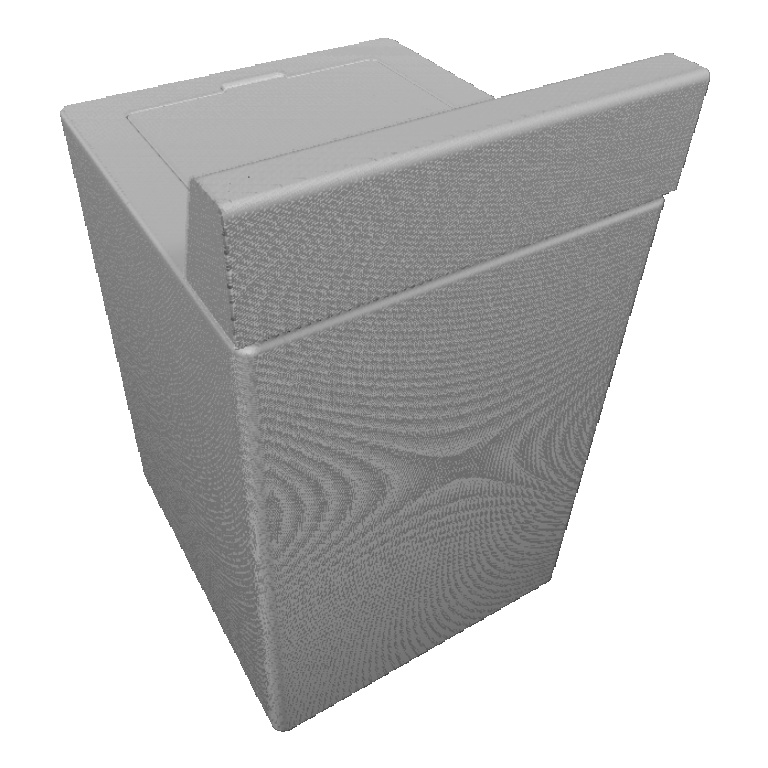} &
\includegraphics[width=0.1\textwidth]{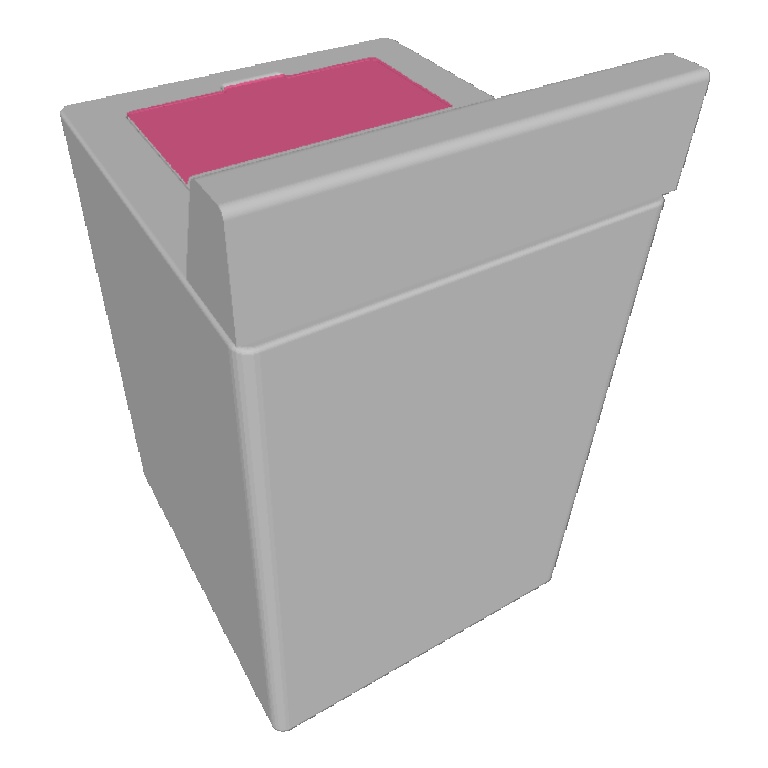} &
\includegraphics[width=0.1\textwidth]{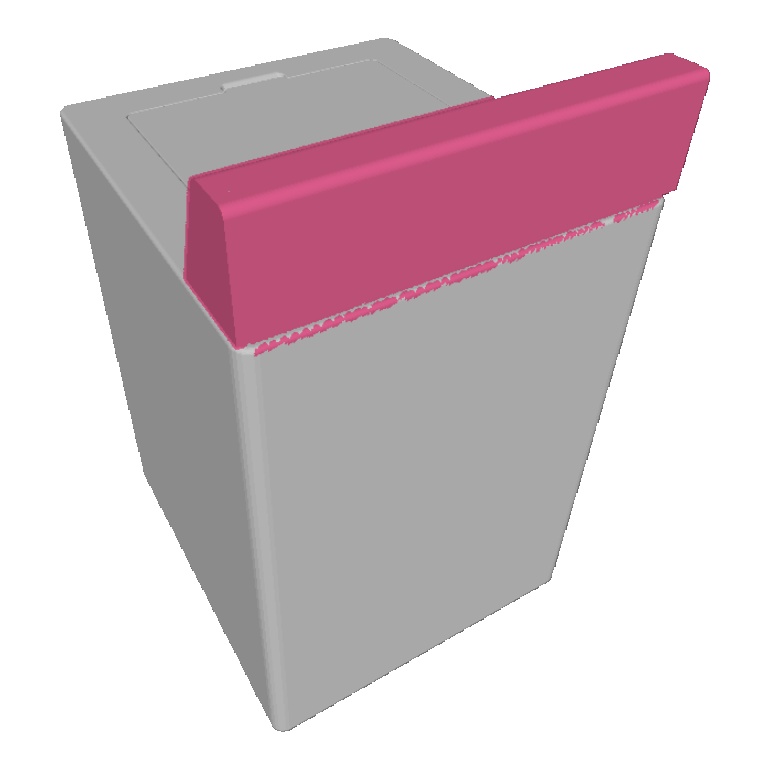} &
\includegraphics[width=0.1\textwidth]{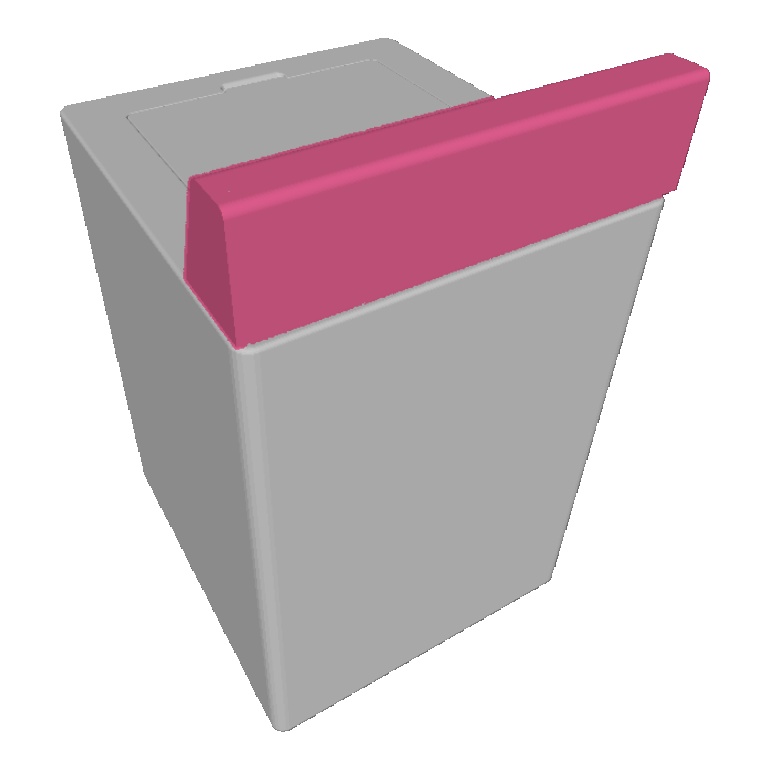} &
\includegraphics[width=0.1\textwidth]{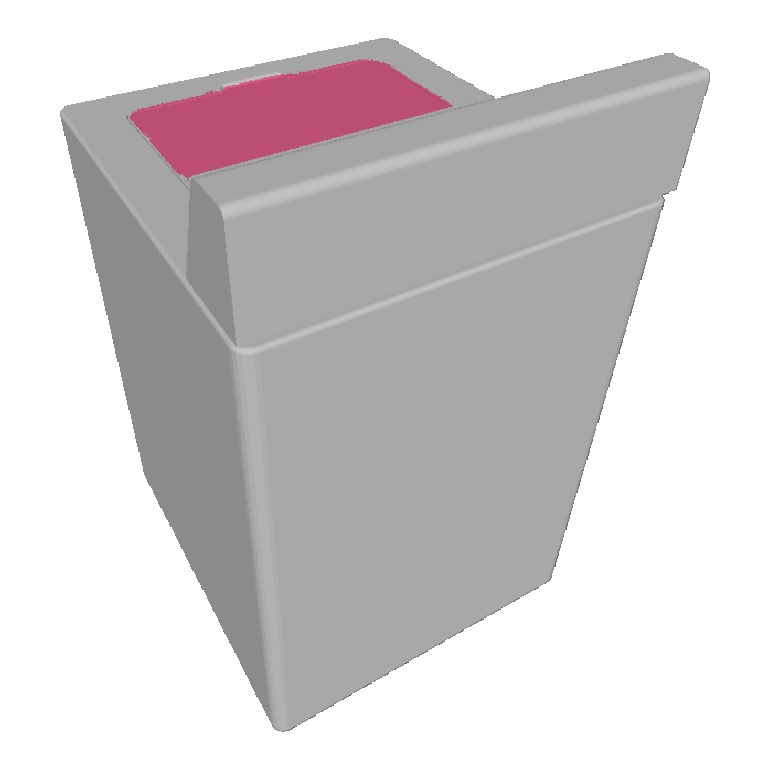} \\

\addlinespace[-2pt]
\arrayrulecolor{gray}\cmidrule(lr){1-5}
\arrayrulecolor{black}

\includegraphics[width=0.1\textwidth]{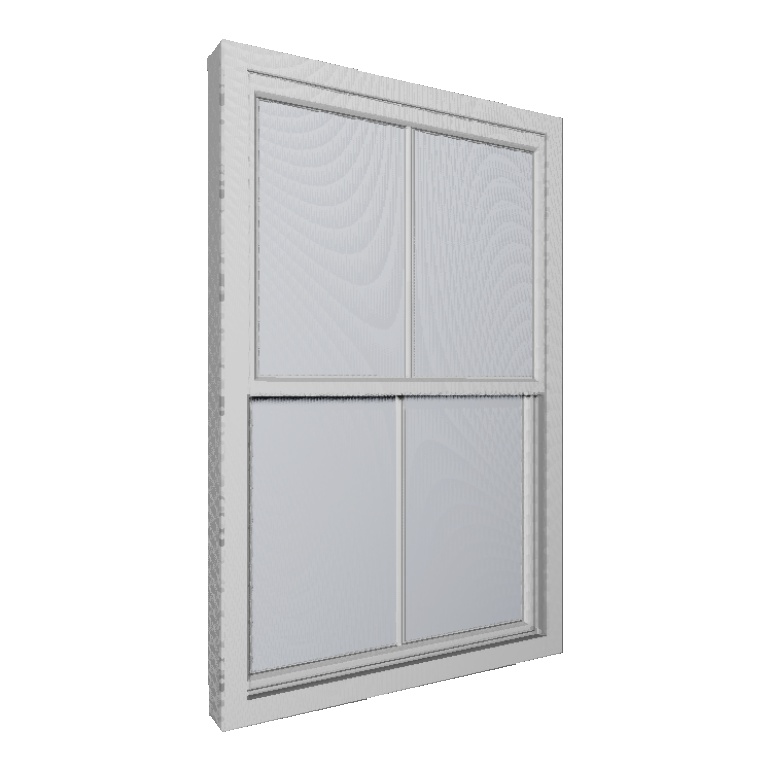} &
\includegraphics[width=0.1\textwidth]{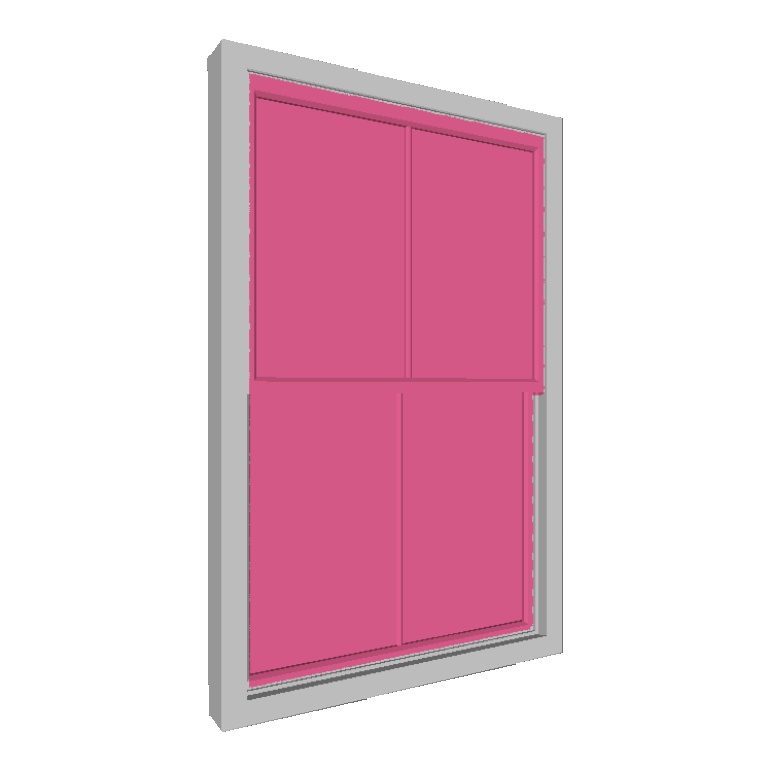} &
\includegraphics[width=0.1\textwidth]{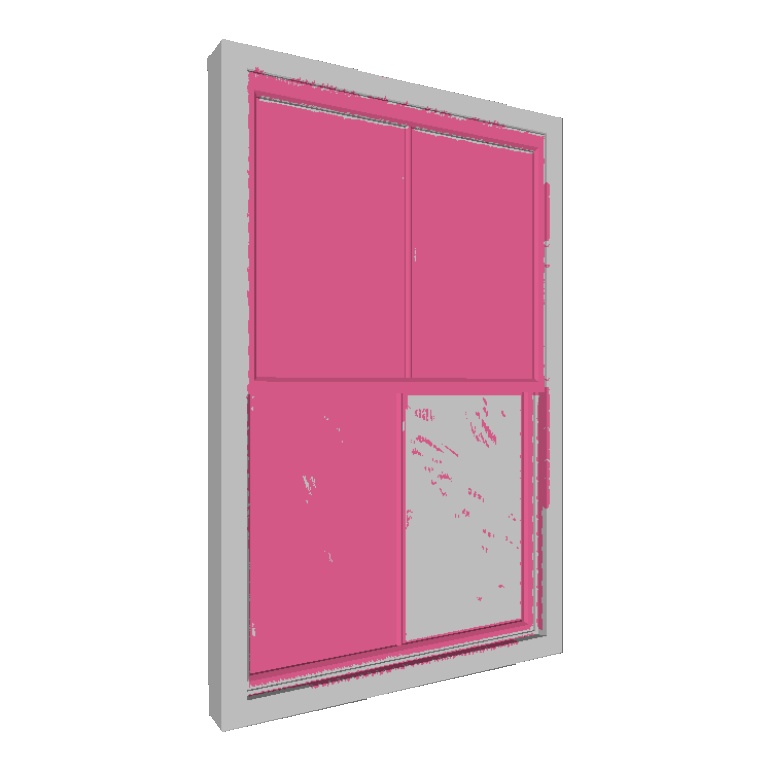} &
\includegraphics[width=0.1\textwidth]{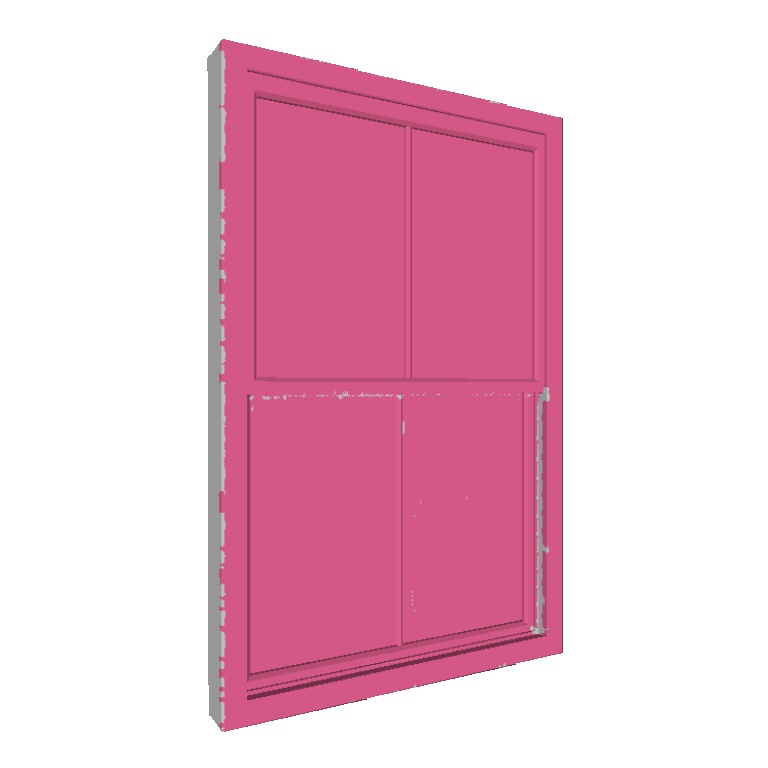} &
\includegraphics[width=0.1\textwidth]{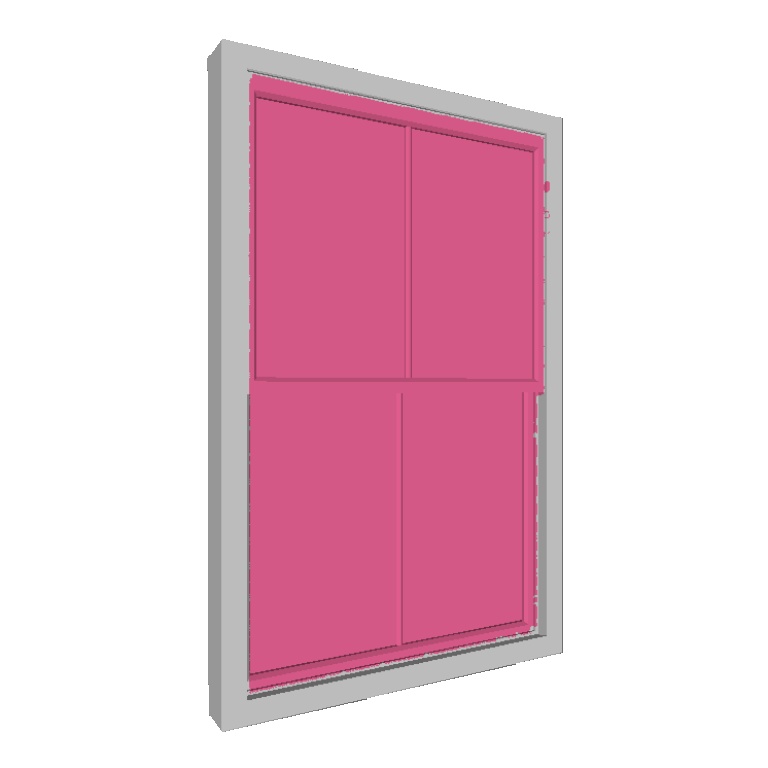} \\

\addlinespace[-2pt]
\arrayrulecolor{gray}\cmidrule(lr){1-5}
\arrayrulecolor{black}

\includegraphics[width=0.1\textwidth]{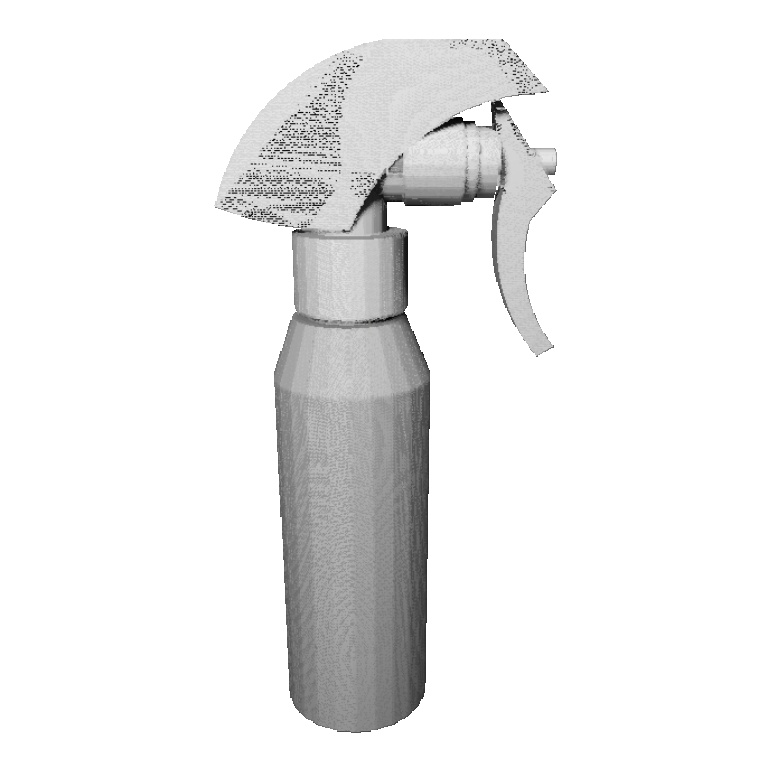} &
\includegraphics[width=0.1\textwidth]{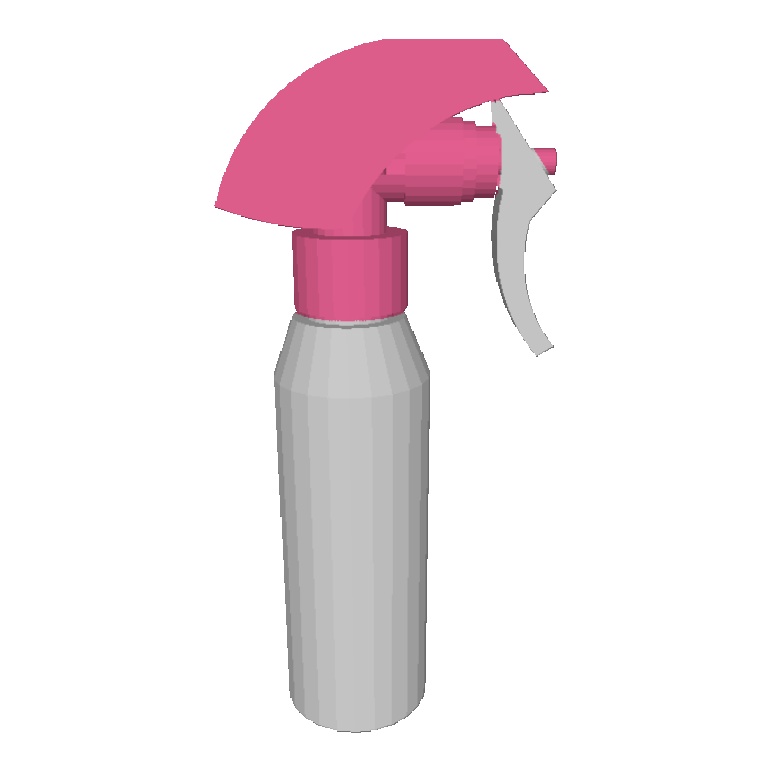} &
\includegraphics[width=0.1\textwidth]{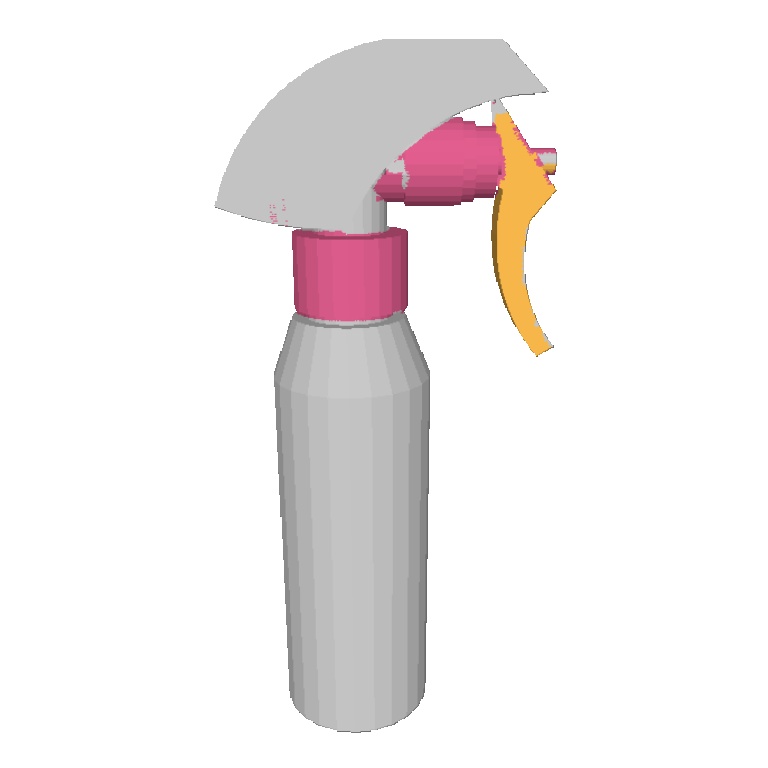} &
\includegraphics[width=0.1\textwidth]{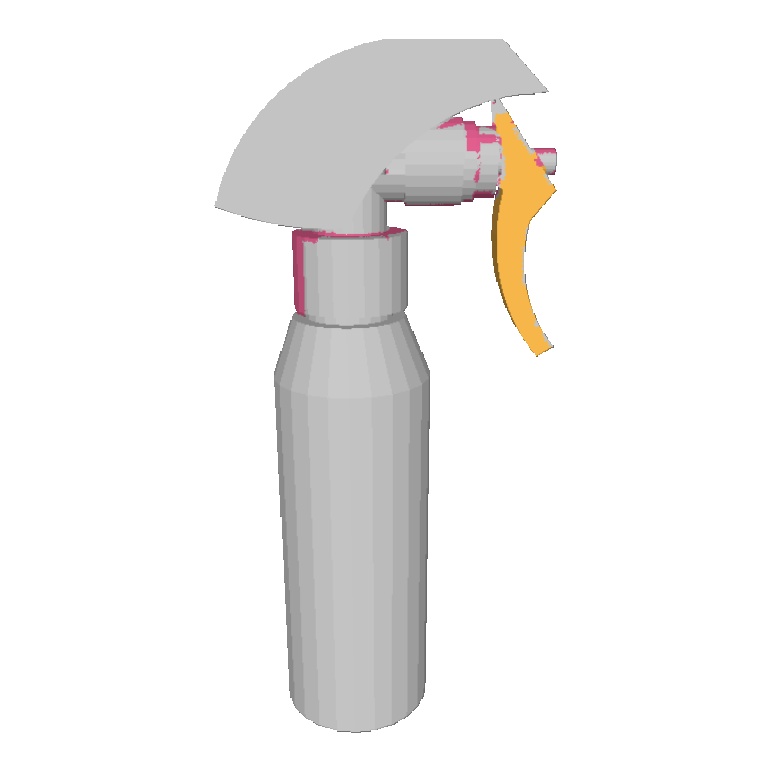} &
\includegraphics[width=0.1\textwidth]{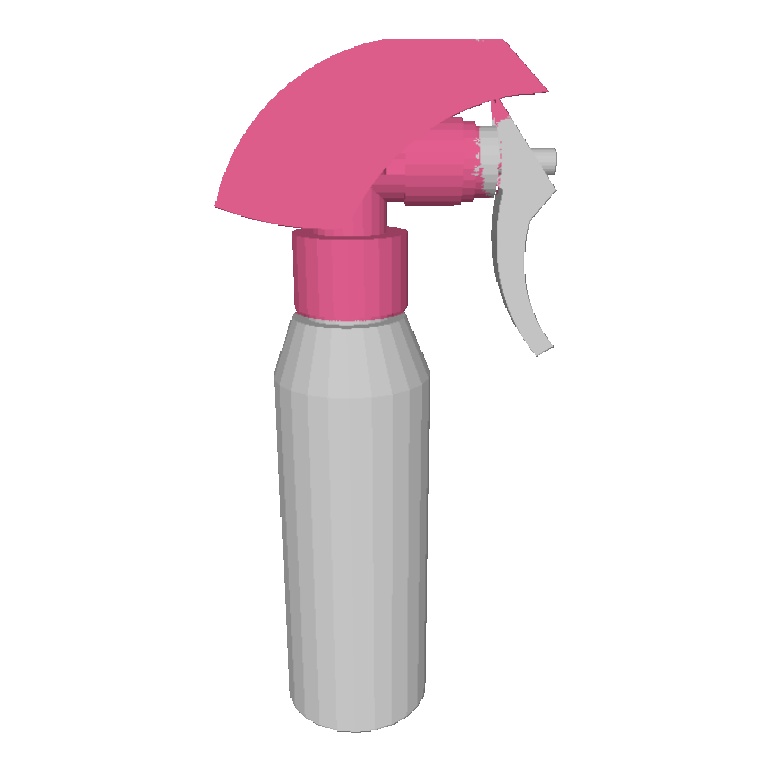} \\

\addlinespace[-2pt]
\arrayrulecolor{gray}\cmidrule(lr){1-5}
\arrayrulecolor{black}

\includegraphics[width=0.1\textwidth]{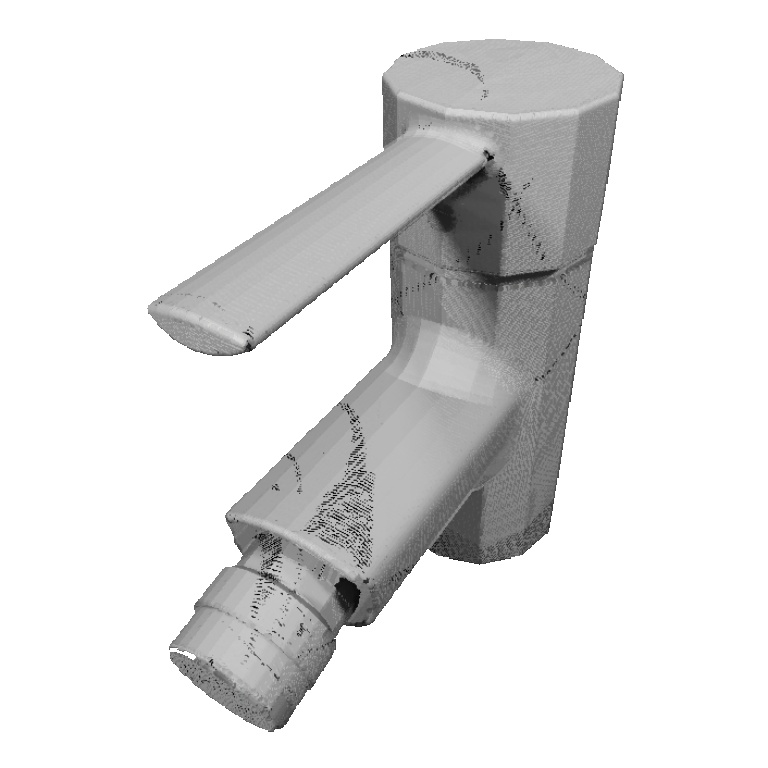} &
\includegraphics[width=0.1\textwidth]{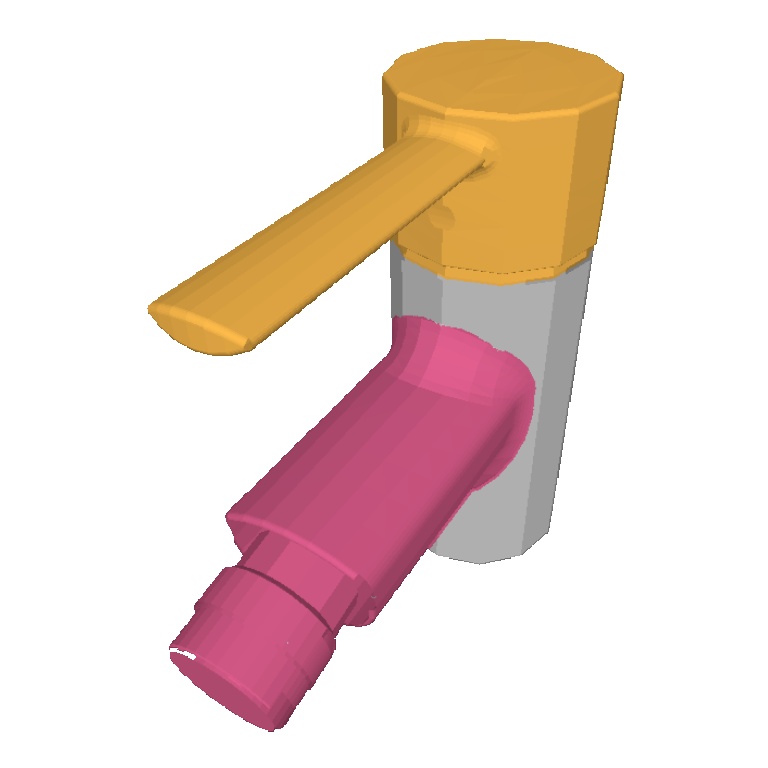} &
\includegraphics[width=0.1\textwidth]{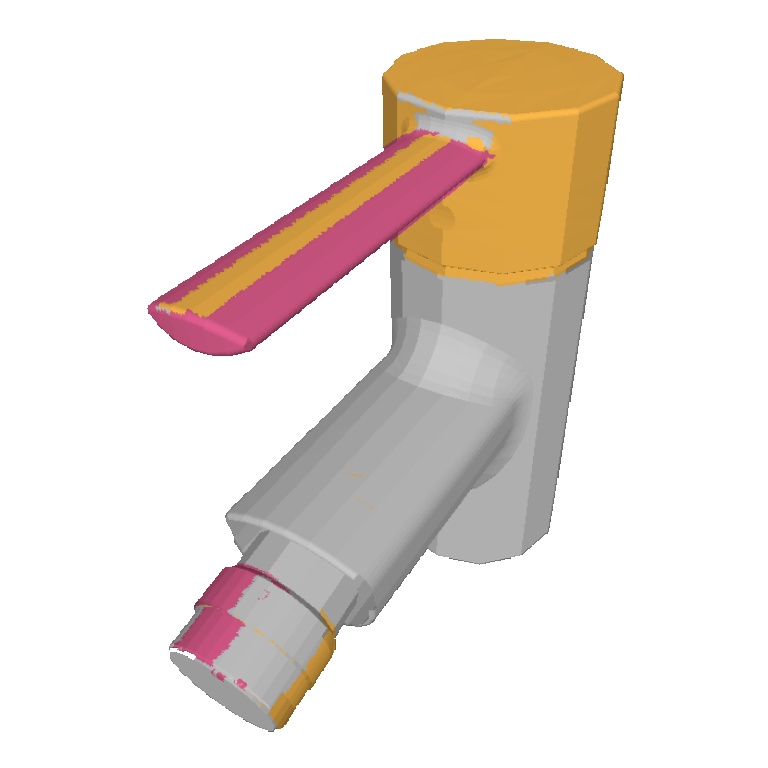} &
\includegraphics[width=0.1\textwidth]{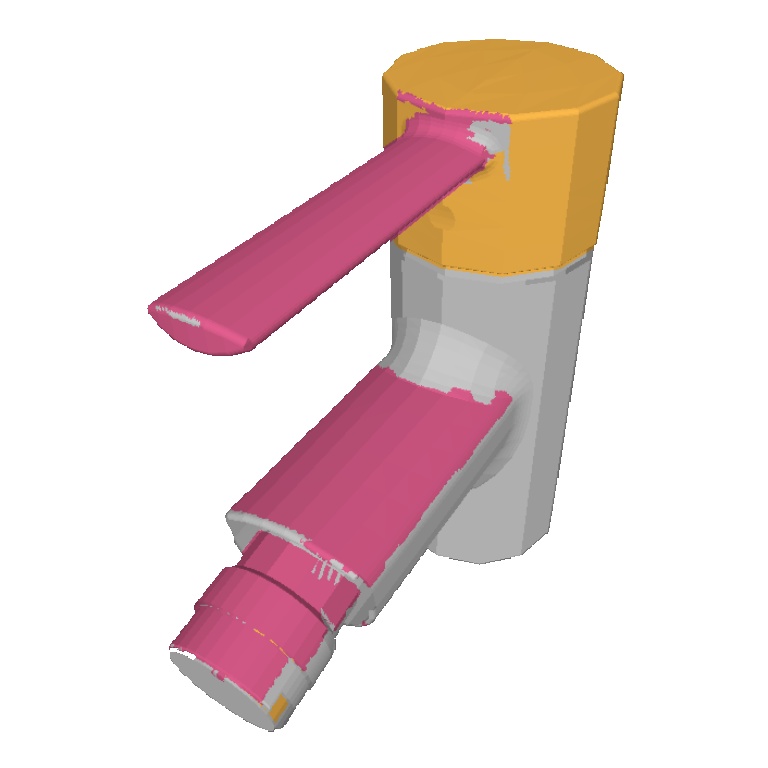} &
\includegraphics[width=0.1\textwidth]{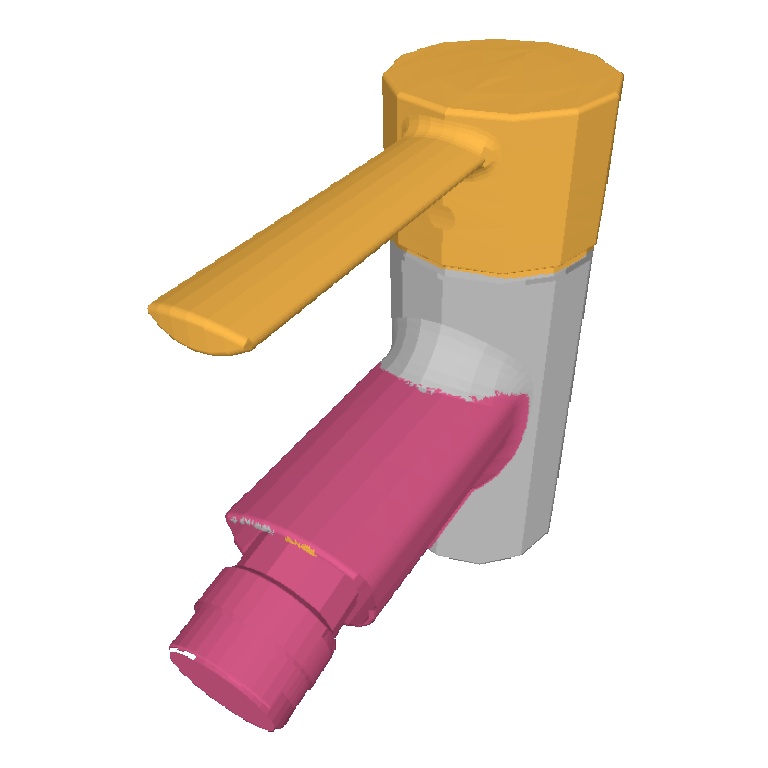} \\

\addlinespace[-2pt]
\bottomrule
\end{tabular}
}}
\end{tabular}
\end{table*}

\begin{table*}[ht]


\centering
\caption{
Qualitative comparison of results from PartSLIP~\cite{liu2023partslip}, PartSTAD~\cite{kim2024partstad}, and Ours for semantic segmentation on the PartNetE~\cite{liu2023partslip} dataset. Each row shows the same object from two different views.
}
\label{tab:PN_full_comp_3}

\begin{tabular}{@{}p{0.5\textwidth}@{} | @{}p{0.5\textwidth}@{}}

\multicolumn{1}{c}{\textbf{View 1}} & \multicolumn{1}{c}{\textbf{View 2}} \\

\vtop{\vskip0pt
\resizebox{0.5\textwidth}{!}{
\begin{tabular}{@{}c@{}c@{}c@{}c@{}c@{}}
\toprule
Input & GT  & PartSLIP & PartSTAD & Ours \\ \midrule
\includegraphics[width=0.1\textwidth]{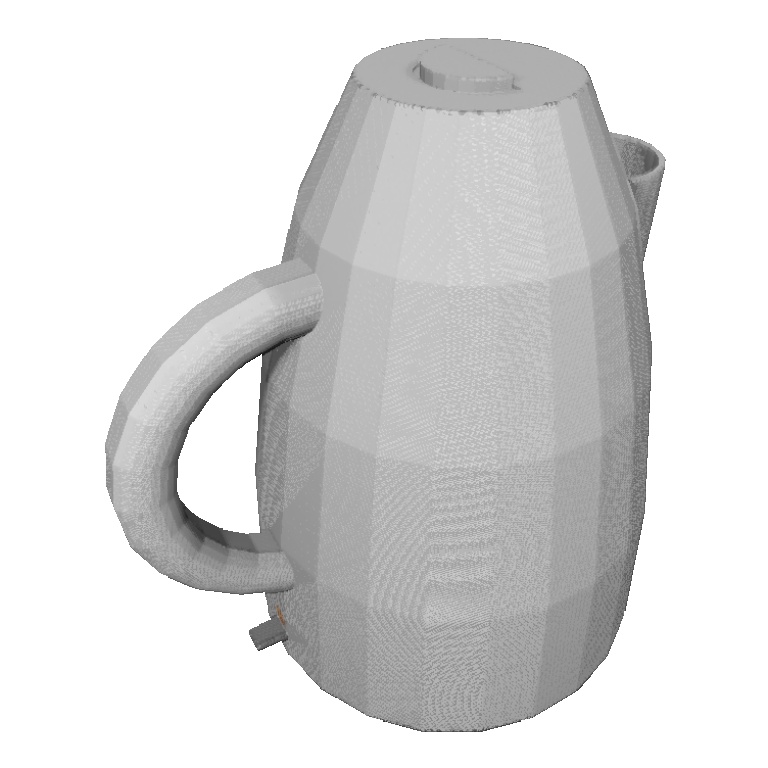} &
\includegraphics[width=0.1\textwidth]{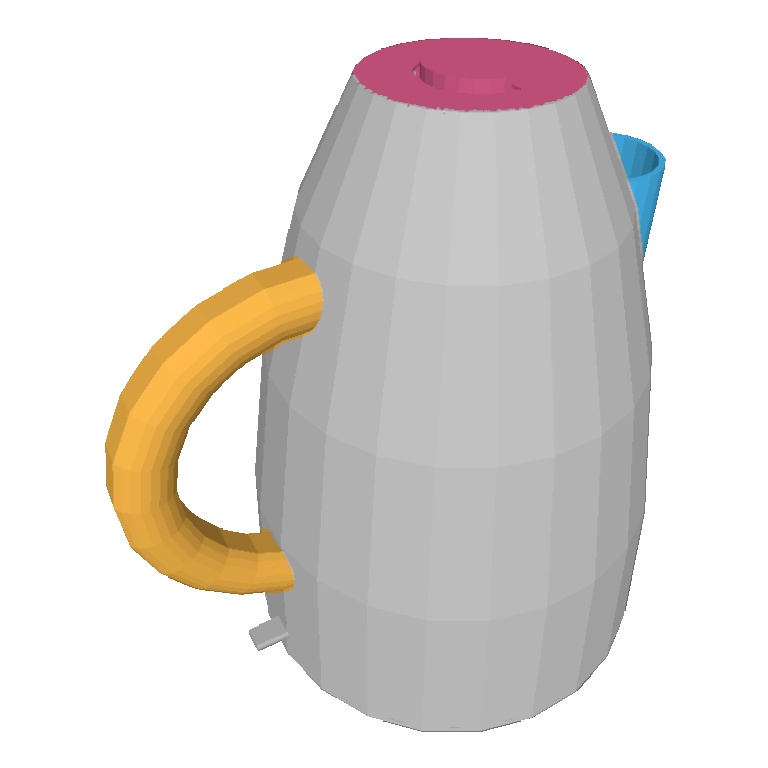} &
\includegraphics[width=0.1\textwidth]{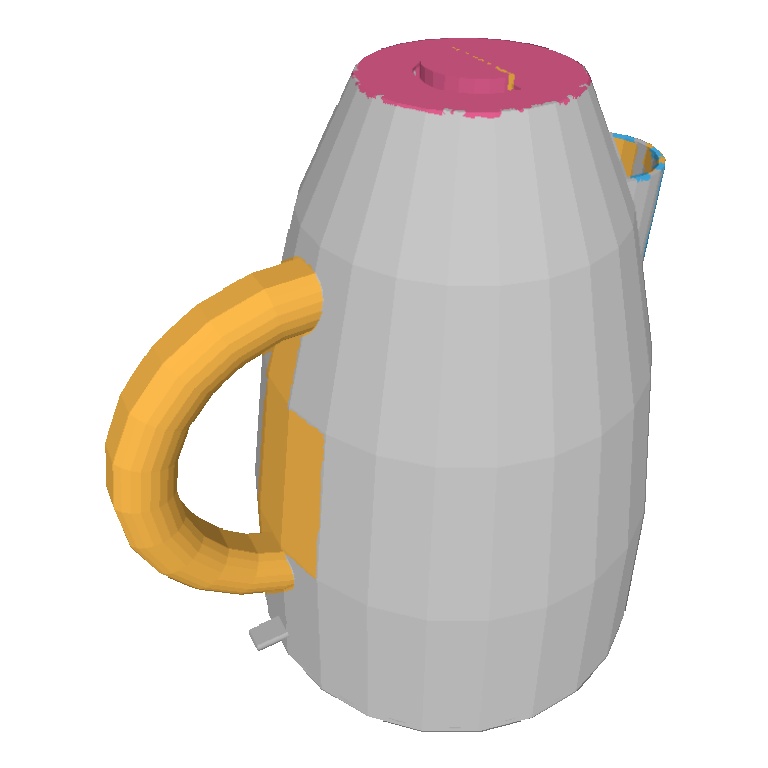} &
\includegraphics[width=0.1\textwidth]{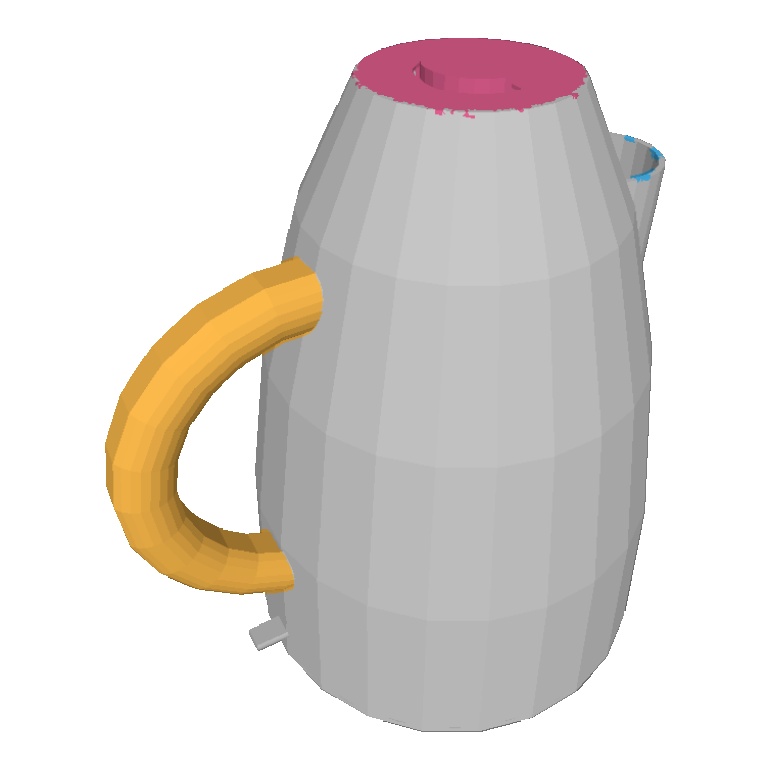} &
\includegraphics[width=0.1\textwidth]{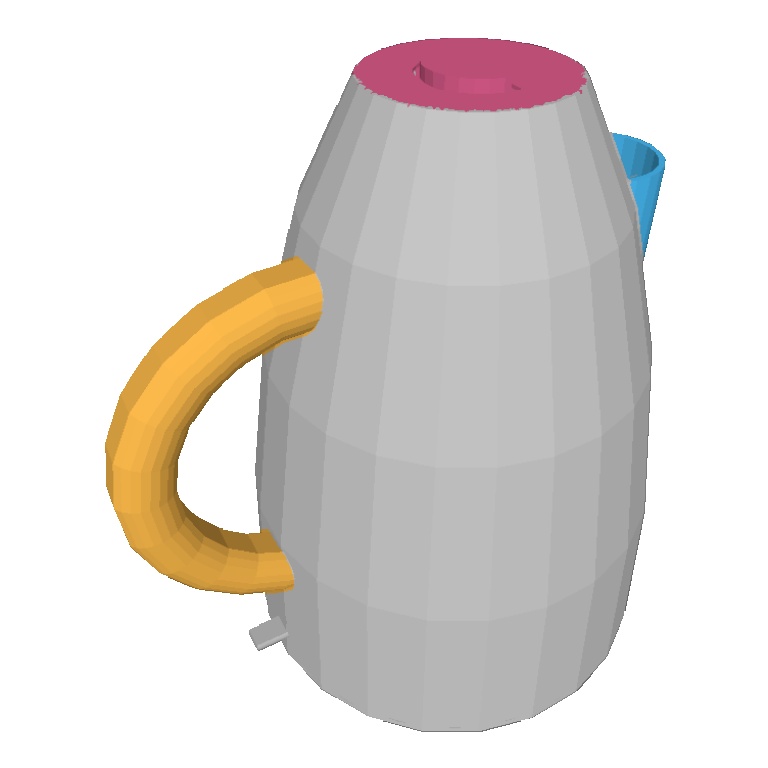} \\

\addlinespace[-2pt]
\arrayrulecolor{gray}\cmidrule(lr){1-5}
\arrayrulecolor{black}

\includegraphics[width=0.1\textwidth]{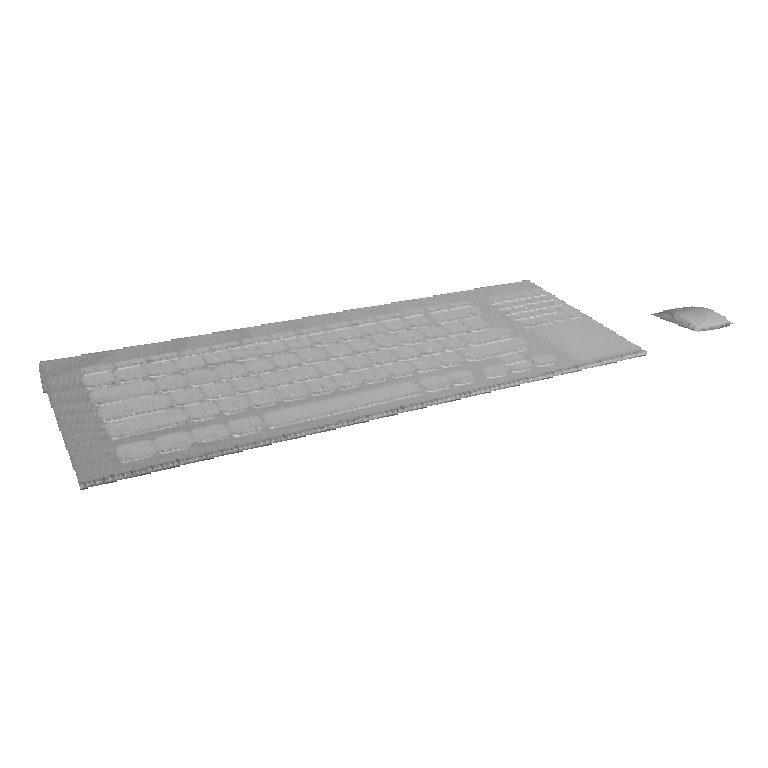} &
\includegraphics[width=0.1\textwidth]{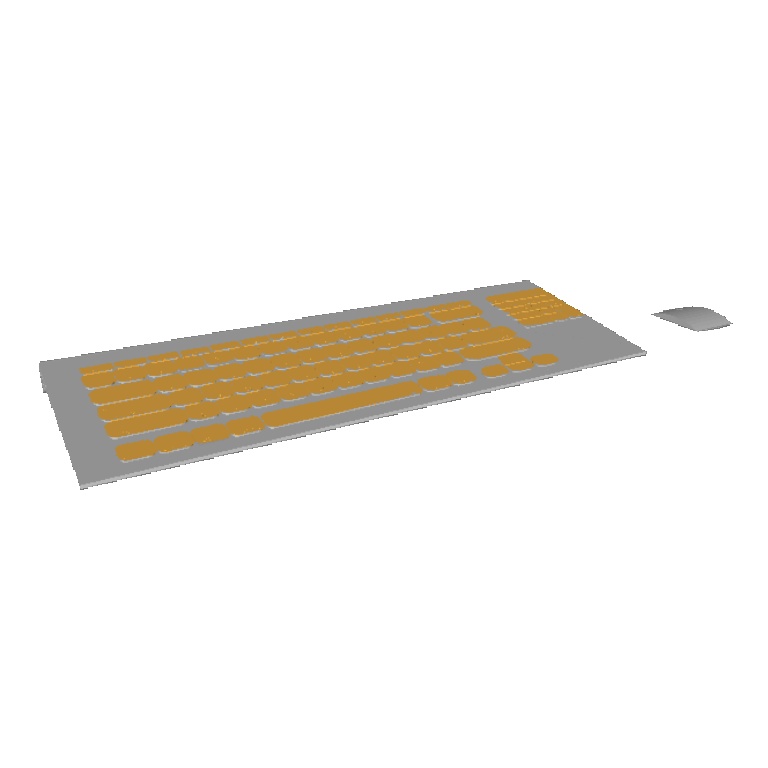} &
\includegraphics[width=0.1\textwidth]{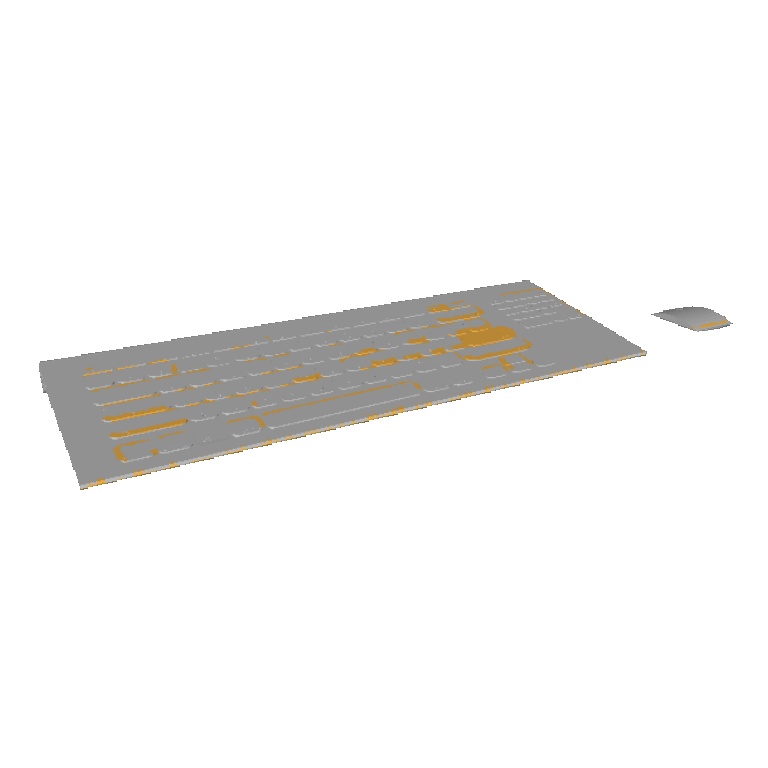} &
\includegraphics[width=0.1\textwidth]{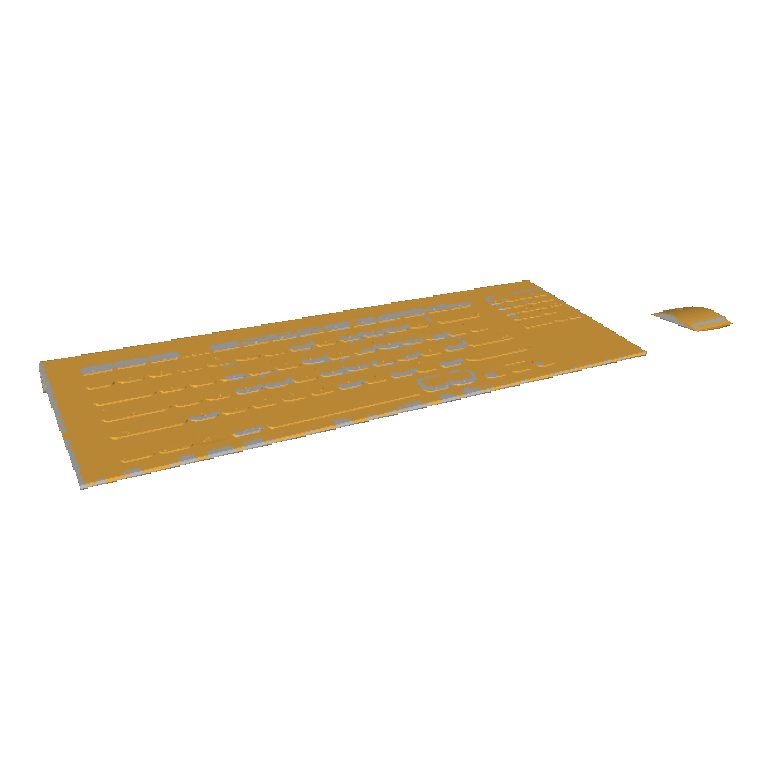} &
\includegraphics[width=0.1\textwidth]{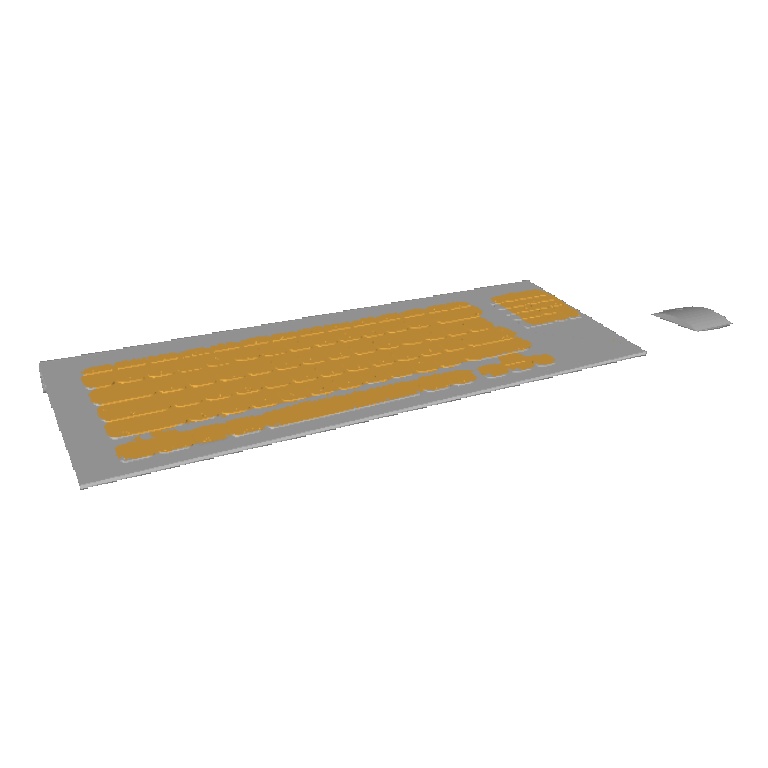} \\

\addlinespace[-2pt]
\arrayrulecolor{gray}\cmidrule(lr){1-5}
\arrayrulecolor{black}

\includegraphics[width=0.1\textwidth]{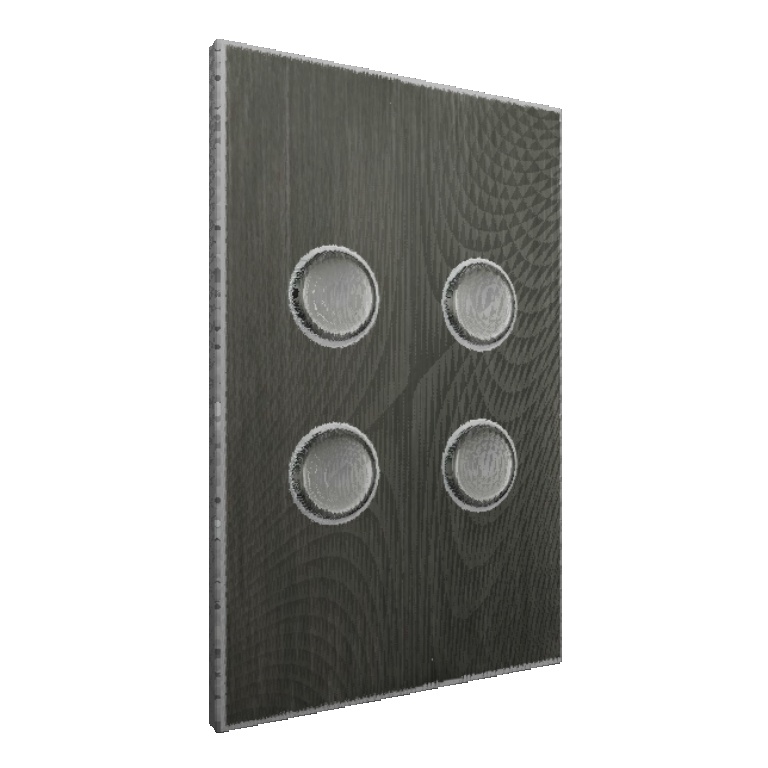} &
\includegraphics[width=0.1\textwidth]{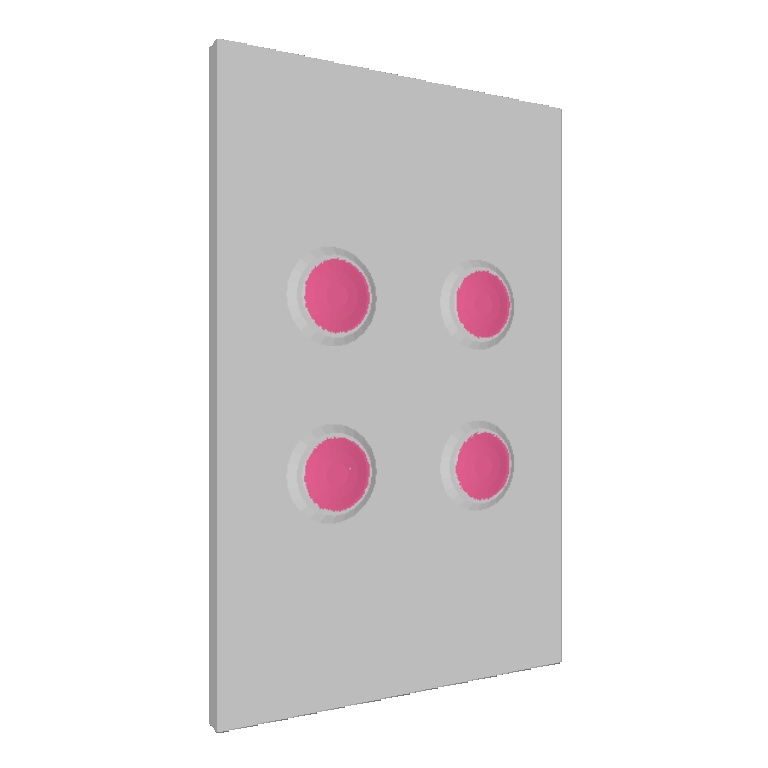} &
\includegraphics[width=0.1\textwidth]{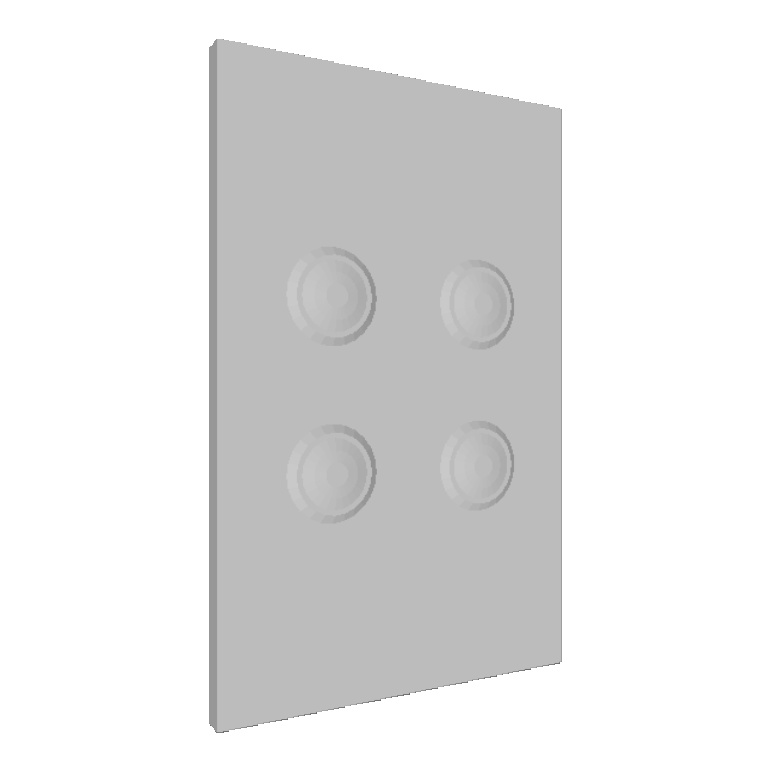} &
\includegraphics[width=0.1\textwidth]{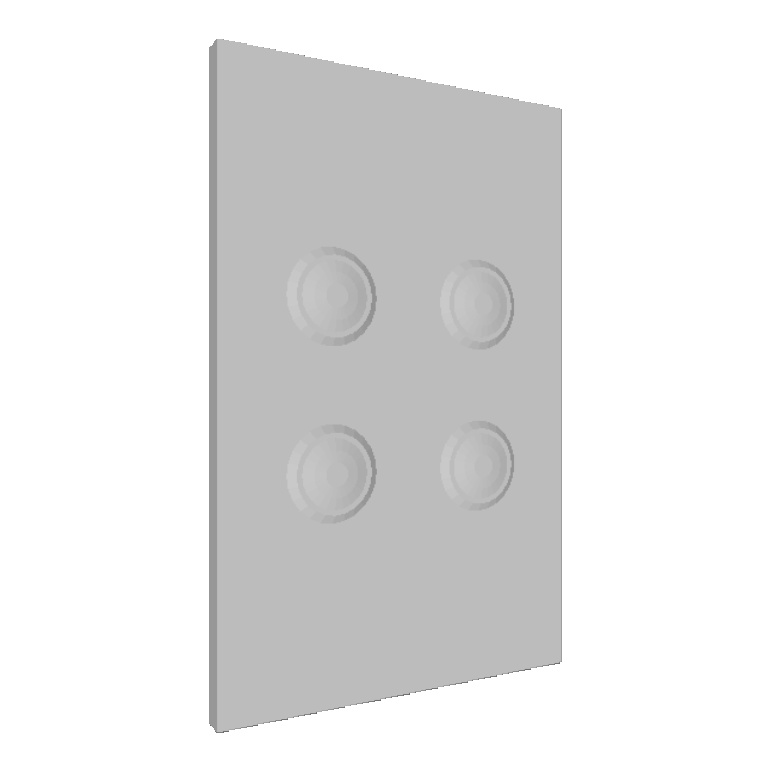} &
\includegraphics[width=0.1\textwidth]{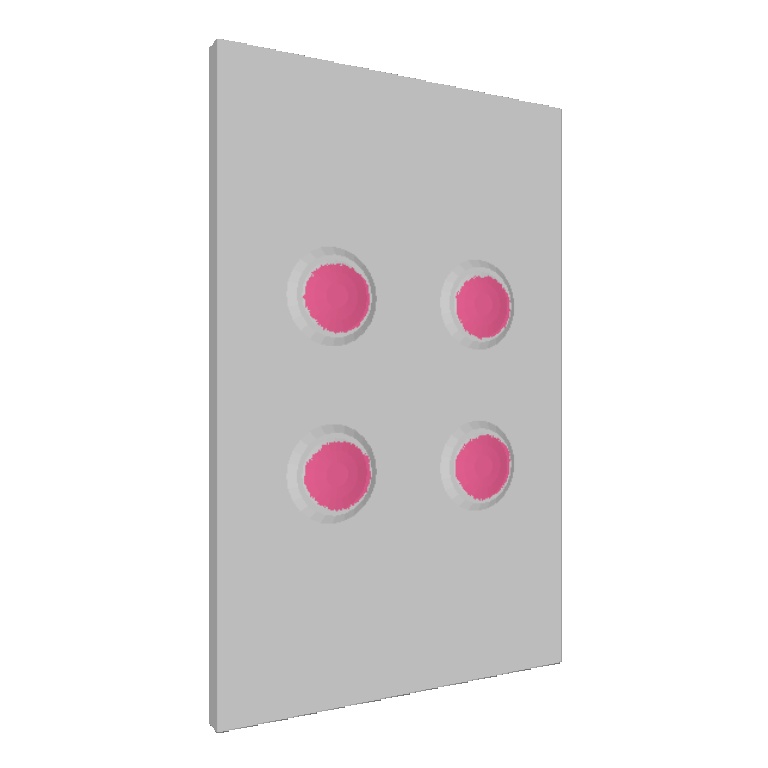} \\

\addlinespace[-2pt]
\arrayrulecolor{gray}\cmidrule(lr){1-5}
\arrayrulecolor{black}

\includegraphics[width=0.1\textwidth]{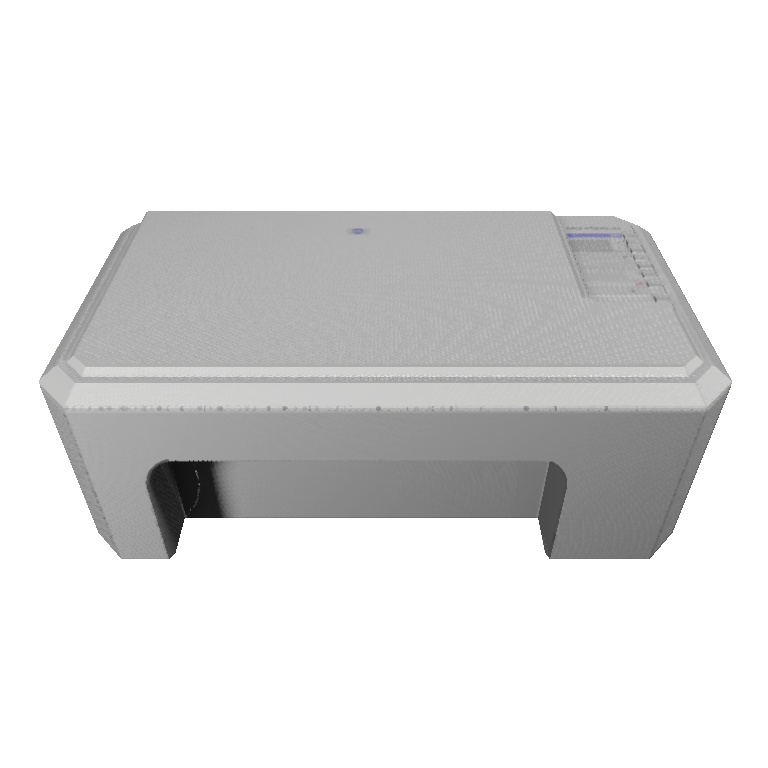} &
\includegraphics[width=0.1\textwidth]{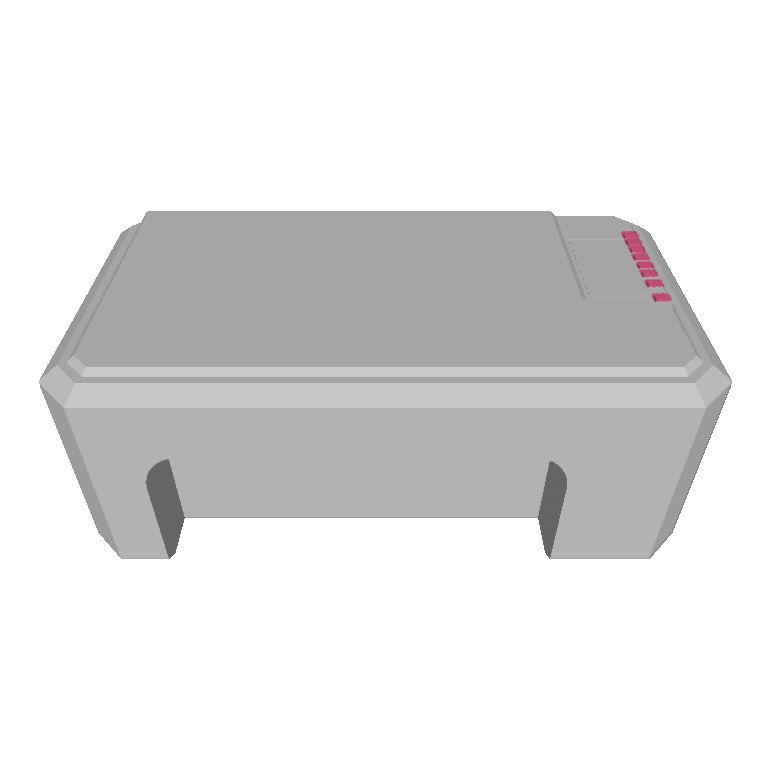} &
\includegraphics[width=0.1\textwidth]{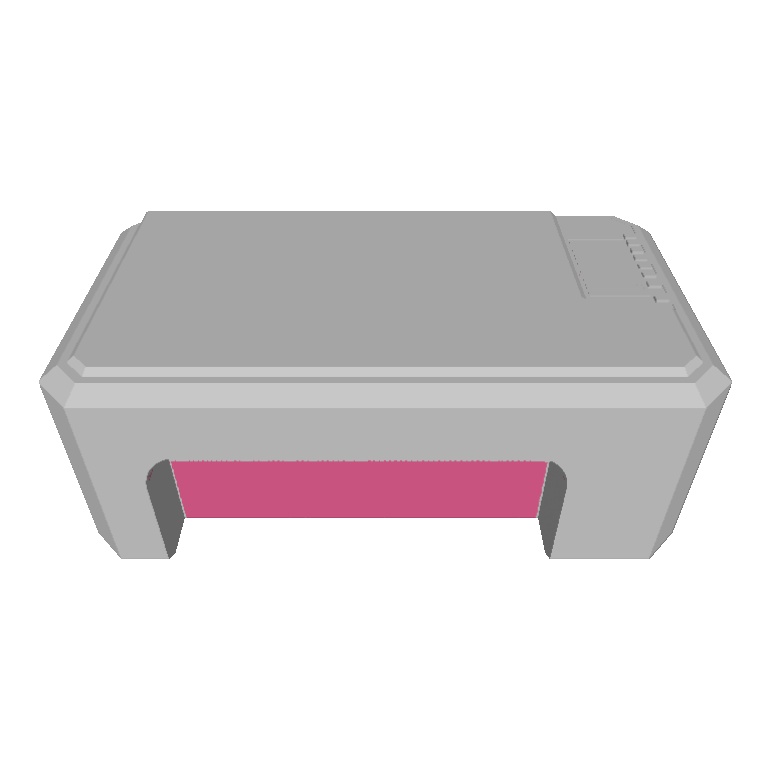} &
\includegraphics[width=0.1\textwidth]{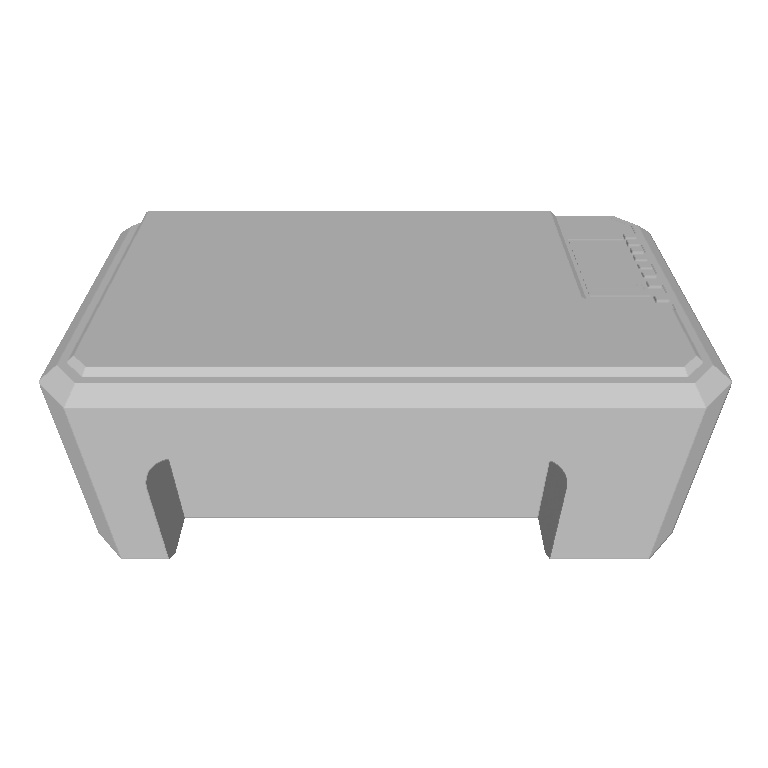} &
\includegraphics[width=0.1\textwidth]{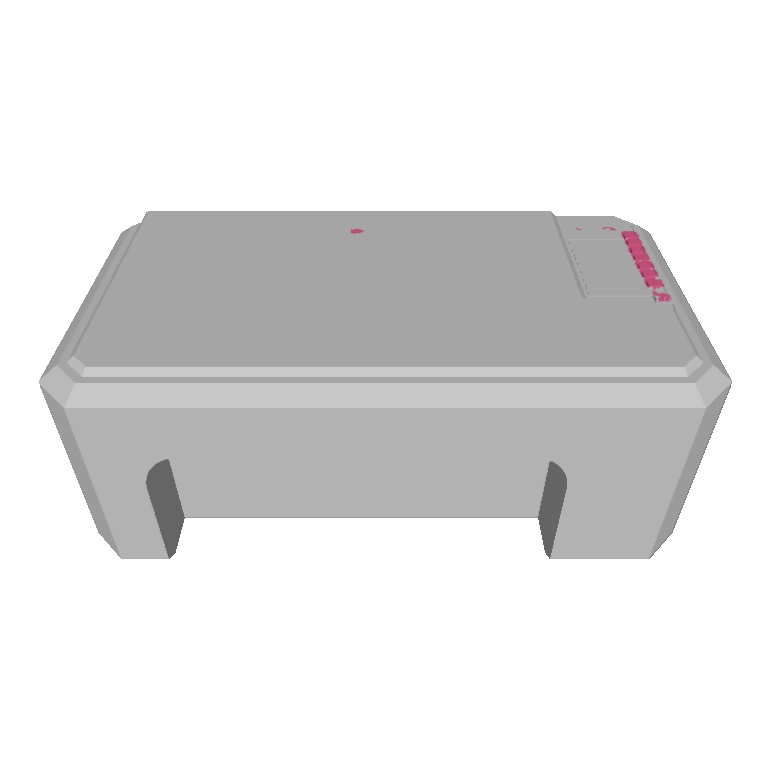} \\

\addlinespace[-2pt]
\arrayrulecolor{gray}\cmidrule(lr){1-5}
\arrayrulecolor{black}

\includegraphics[width=0.1\textwidth]{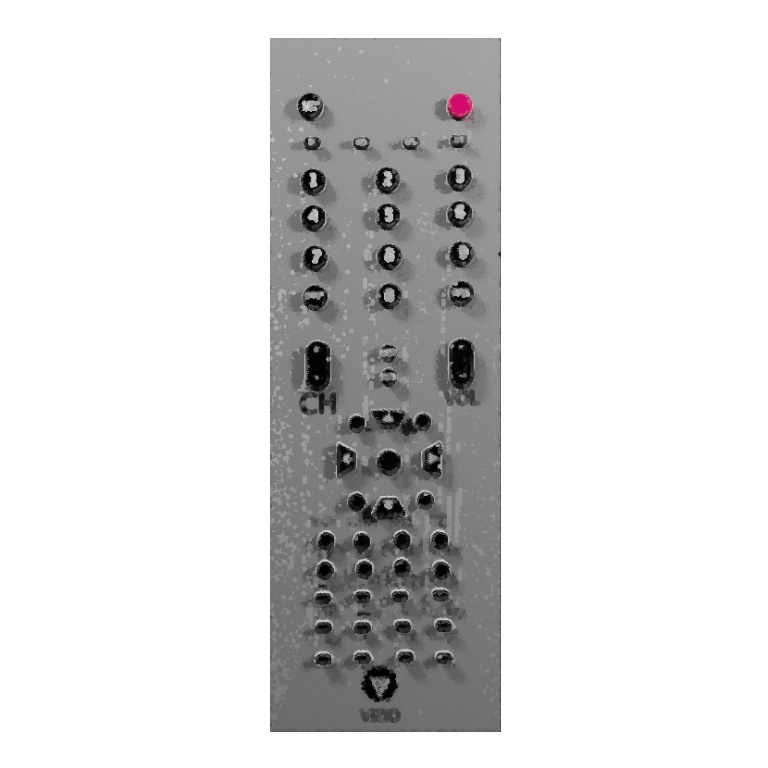} &
\includegraphics[width=0.1\textwidth]{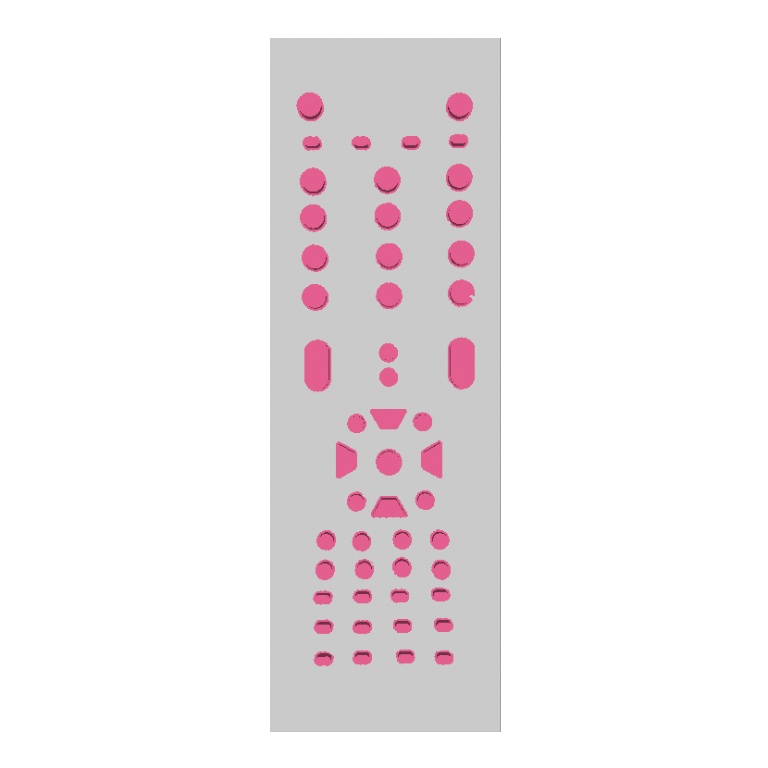} &
\includegraphics[width=0.1\textwidth]{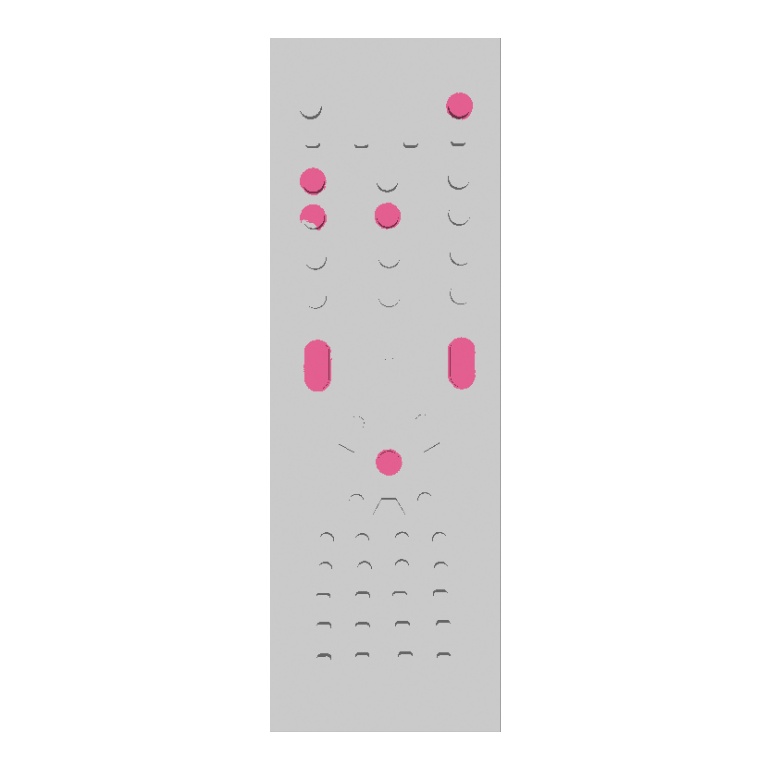} &
\includegraphics[width=0.1\textwidth]{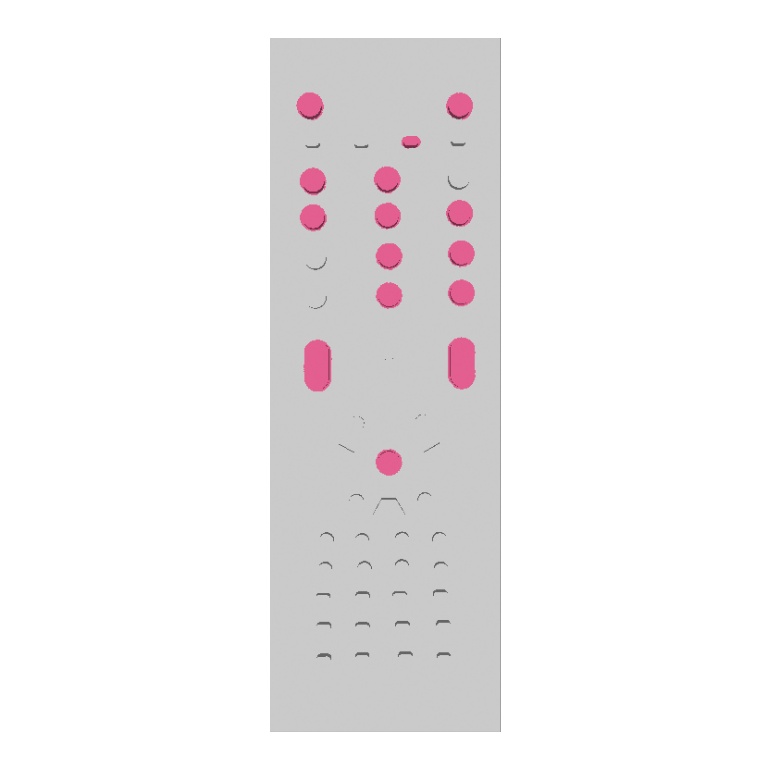} &
\includegraphics[width=0.1\textwidth]{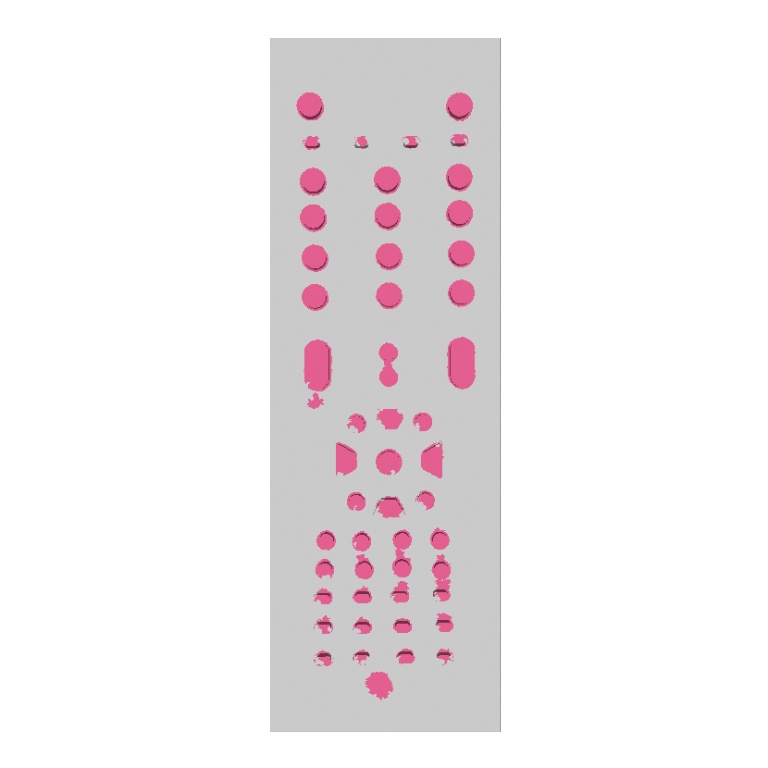} \\

\addlinespace[-2pt]
\arrayrulecolor{gray}\cmidrule(lr){1-5}
\arrayrulecolor{black}

\includegraphics[width=0.1\textwidth]{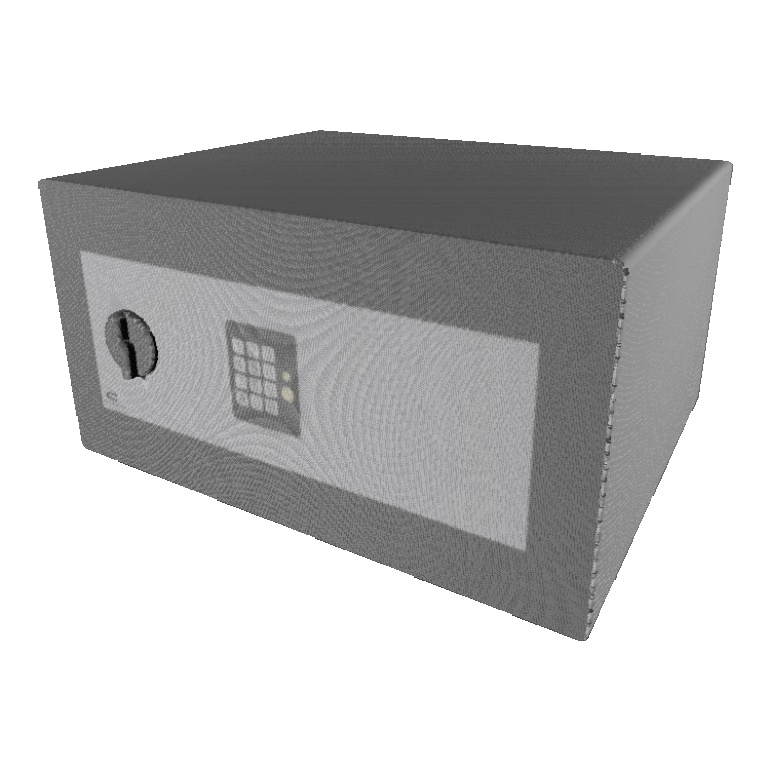} &
\includegraphics[width=0.1\textwidth]{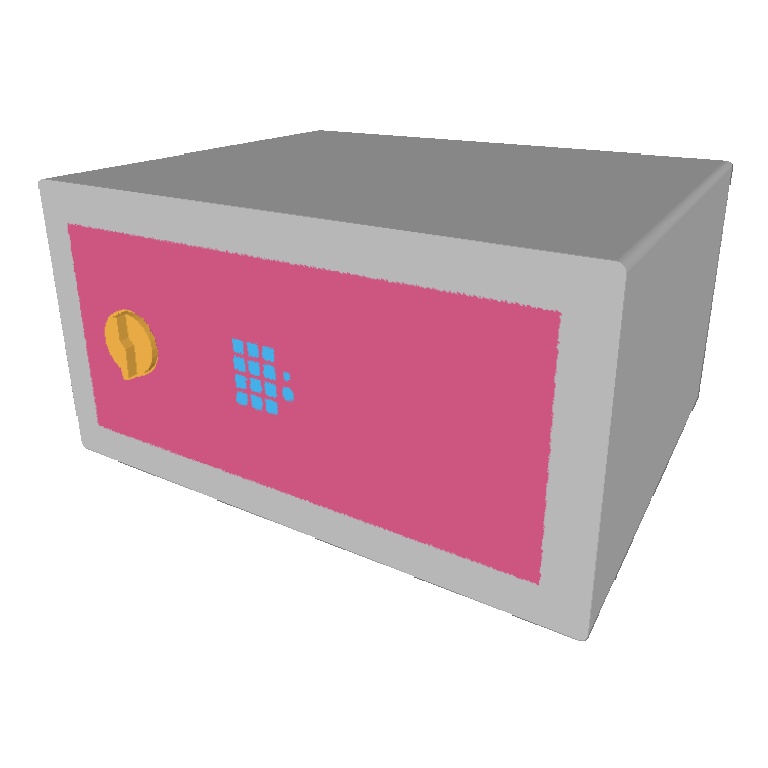} &
\includegraphics[width=0.1\textwidth]{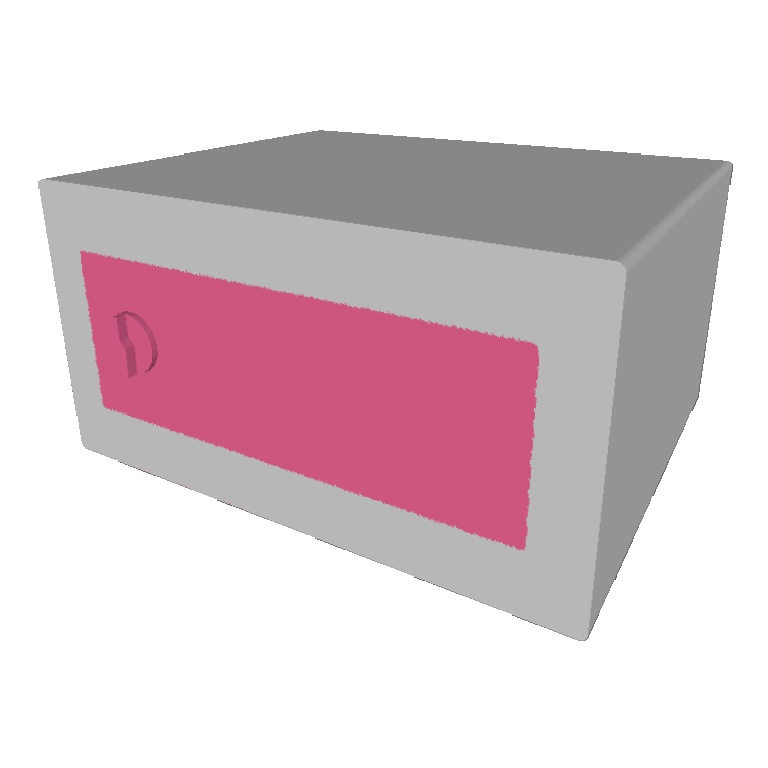} &
\includegraphics[width=0.1\textwidth]{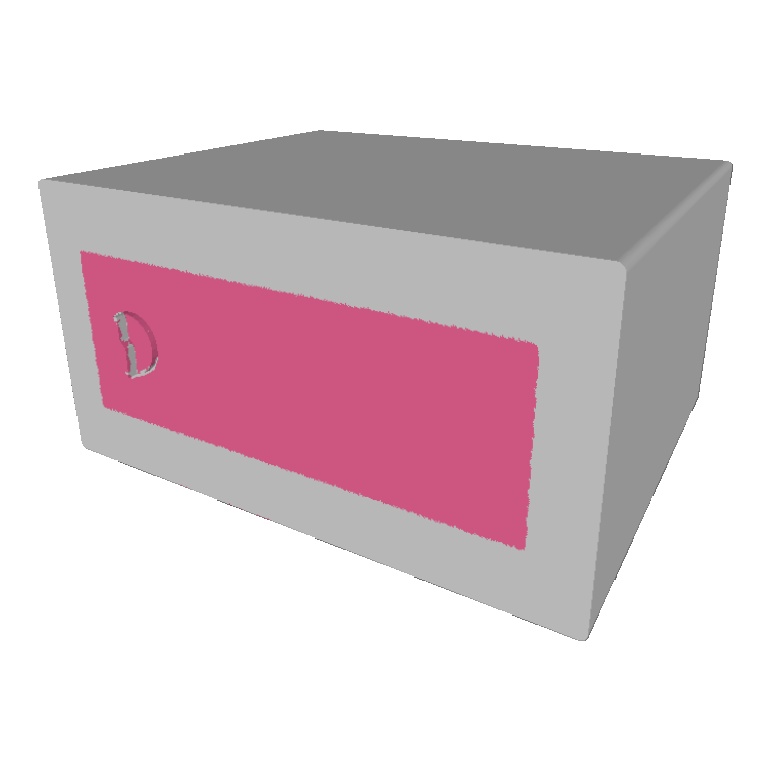} &
\includegraphics[width=0.1\textwidth]{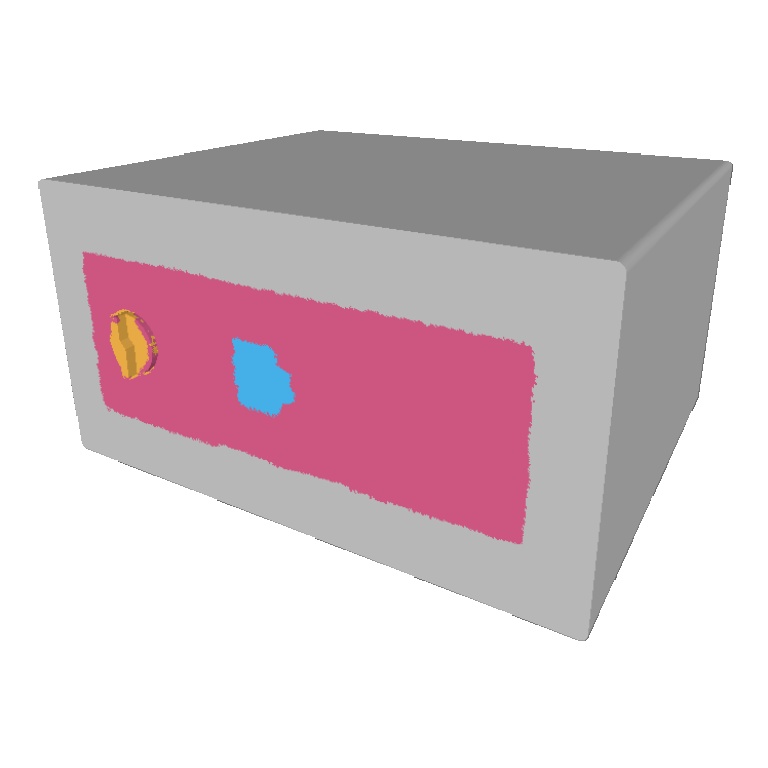} \\

\addlinespace[-2pt]
\arrayrulecolor{gray}\cmidrule(lr){1-5}
\arrayrulecolor{black}

\includegraphics[width=0.1\textwidth]{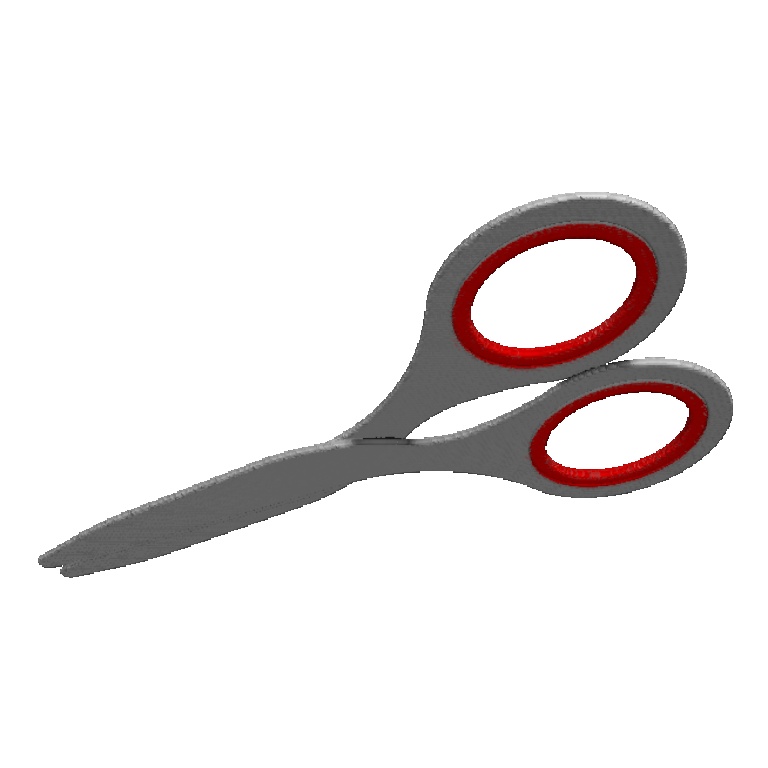} &
\includegraphics[width=0.1\textwidth]{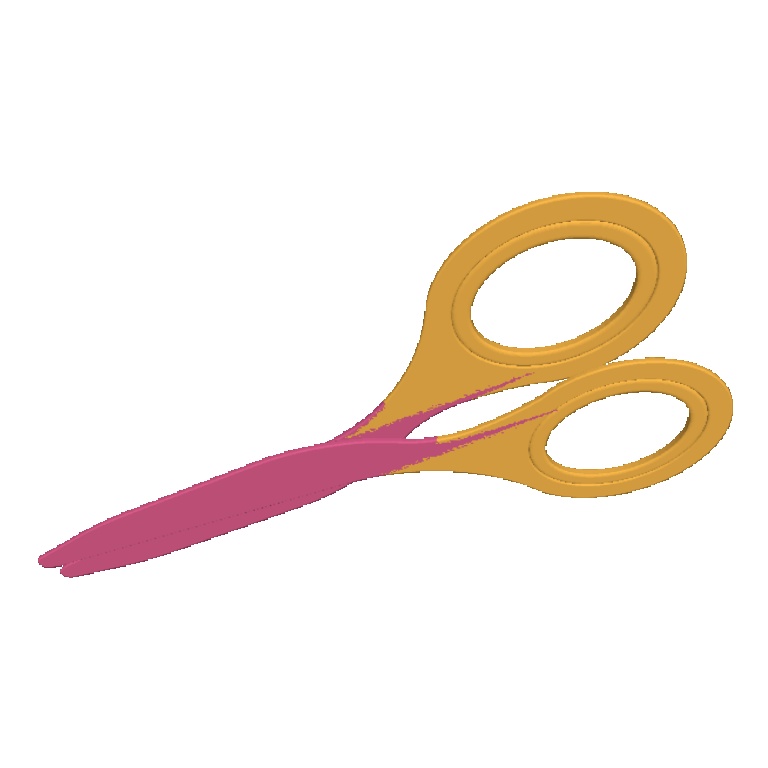} &
\includegraphics[width=0.1\textwidth]{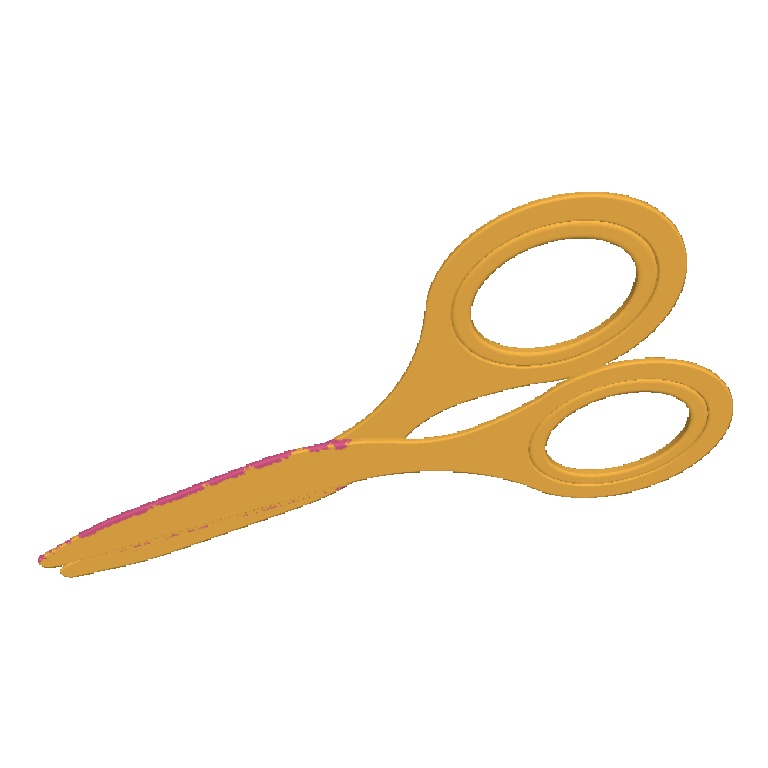} &
\includegraphics[width=0.1\textwidth]{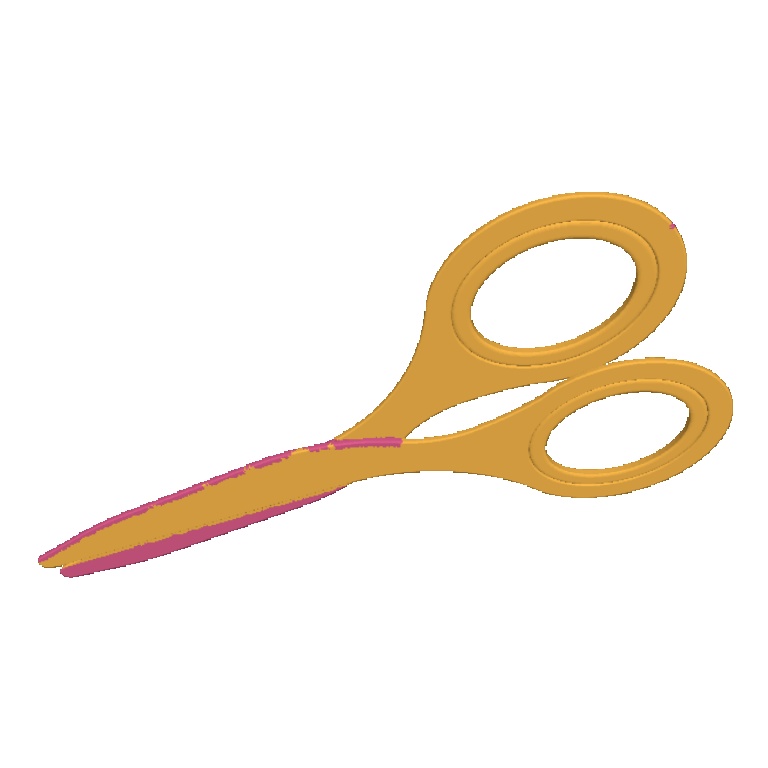} &
\includegraphics[width=0.1\textwidth]{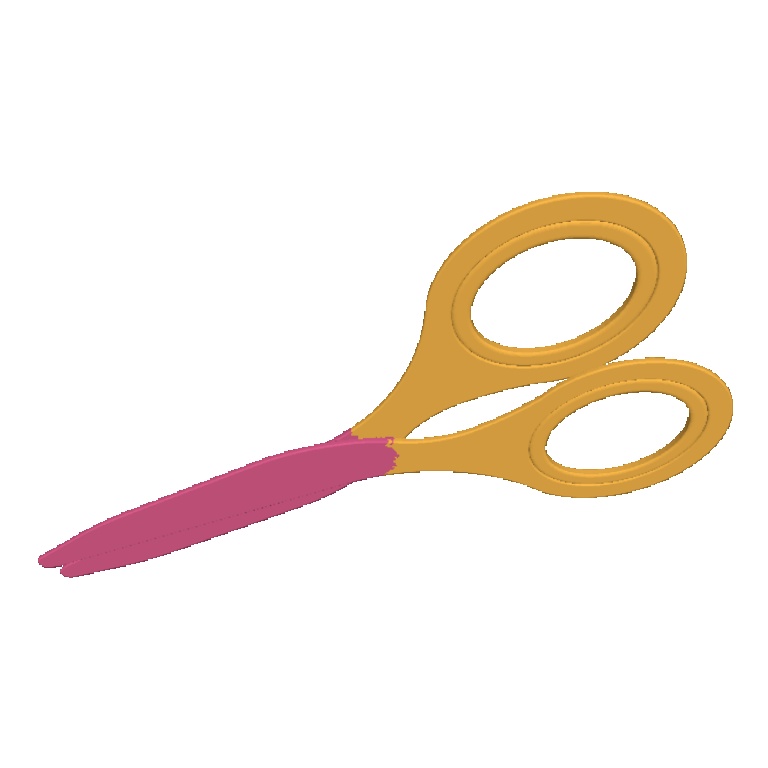} \\

\addlinespace[-2pt]
\arrayrulecolor{gray}\cmidrule(lr){1-5}
\arrayrulecolor{black}

\includegraphics[width=0.1\textwidth]{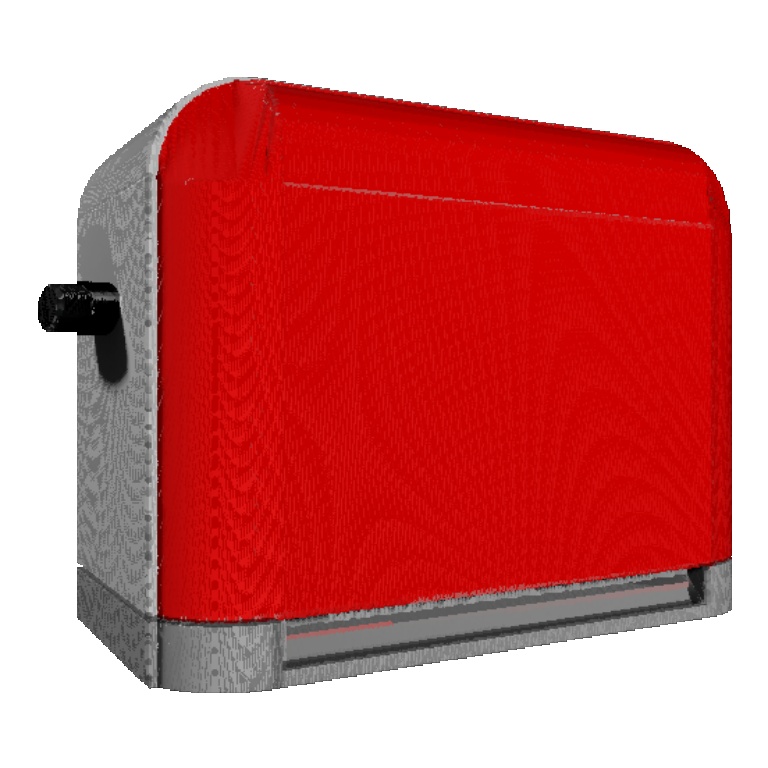} &
\includegraphics[width=0.1\textwidth]{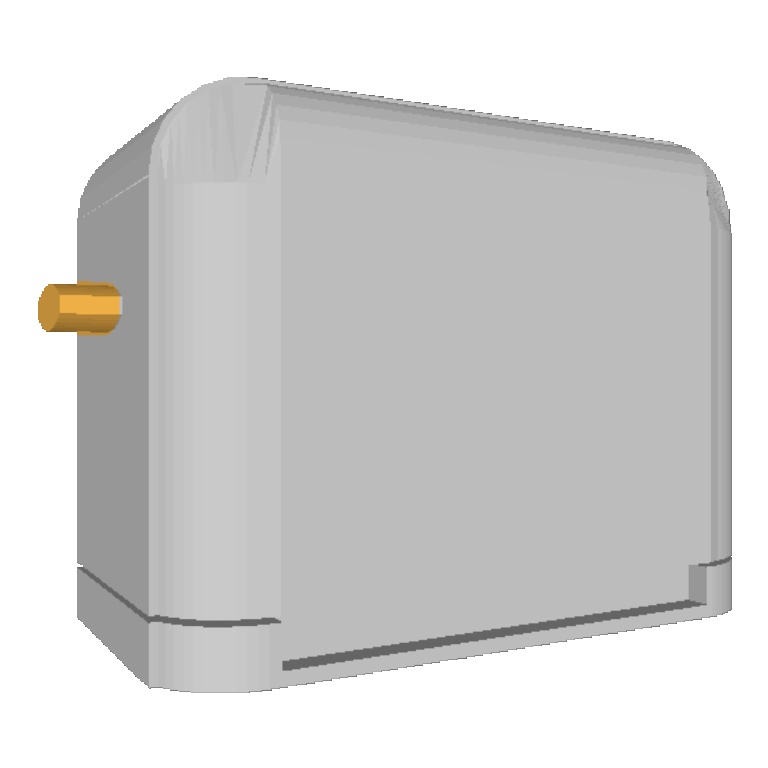} &
\includegraphics[width=0.1\textwidth]{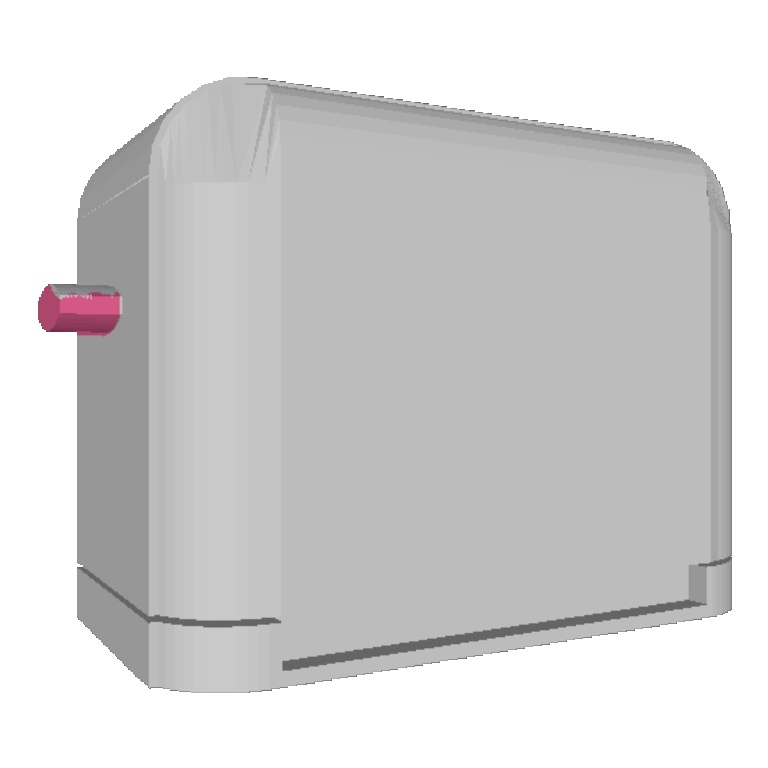} &
\includegraphics[width=0.1\textwidth]{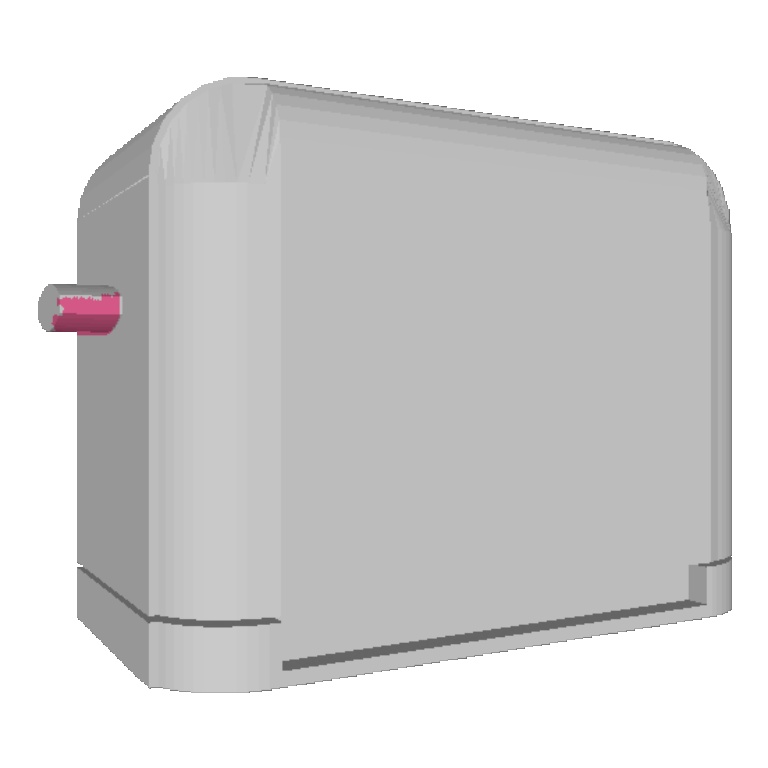} &
\includegraphics[width=0.1\textwidth]{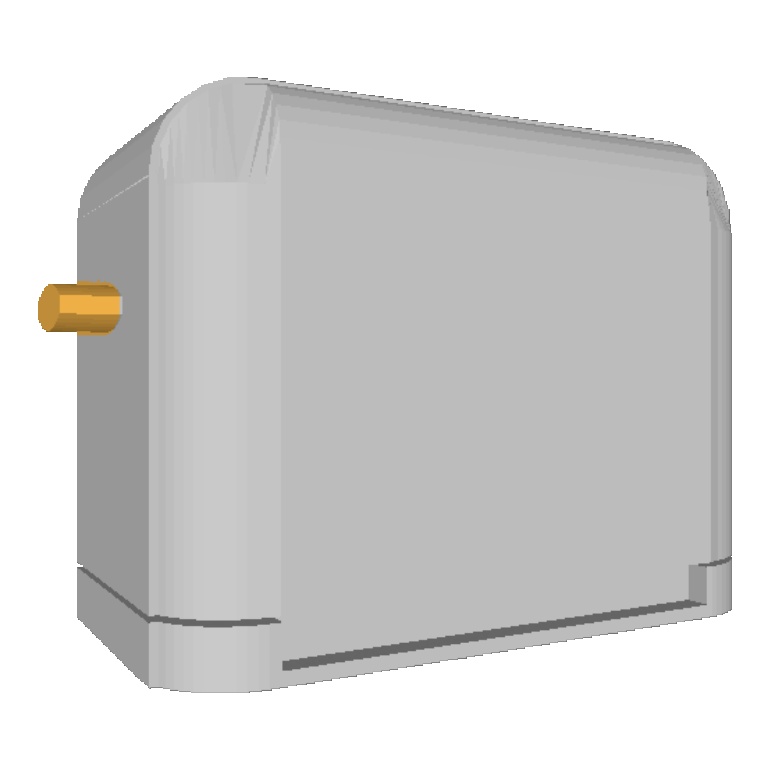} \\

\addlinespace[-2pt]
\arrayrulecolor{gray}\cmidrule(lr){1-5}
\arrayrulecolor{black}

\includegraphics[width=0.1\textwidth]{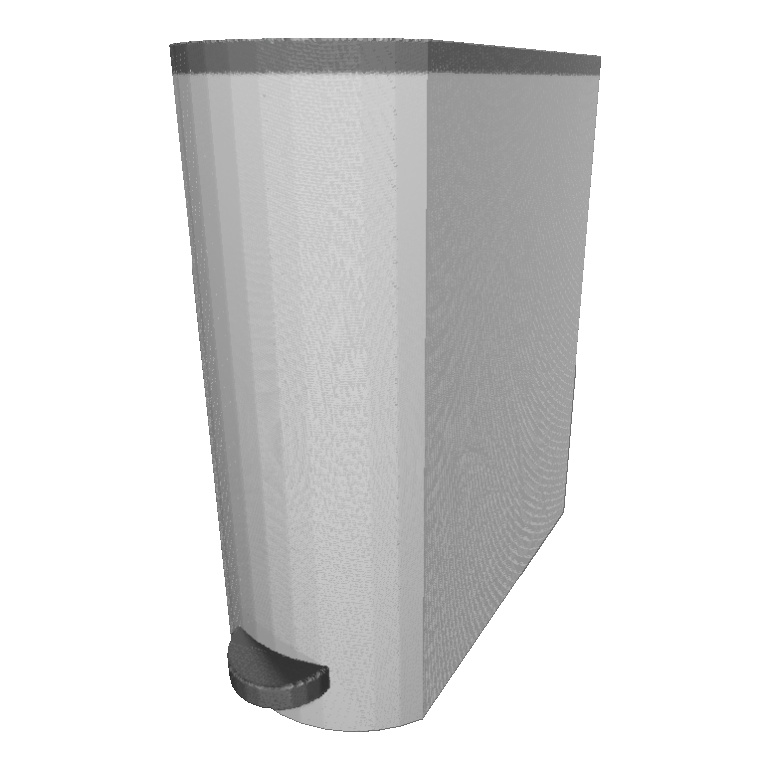} &
\includegraphics[width=0.1\textwidth]{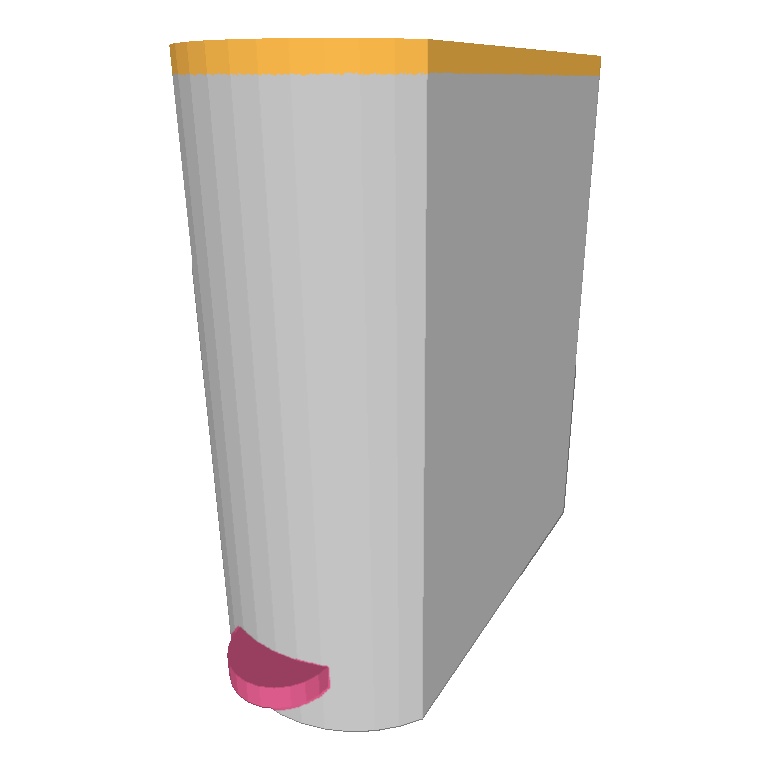} &
\includegraphics[width=0.1\textwidth]{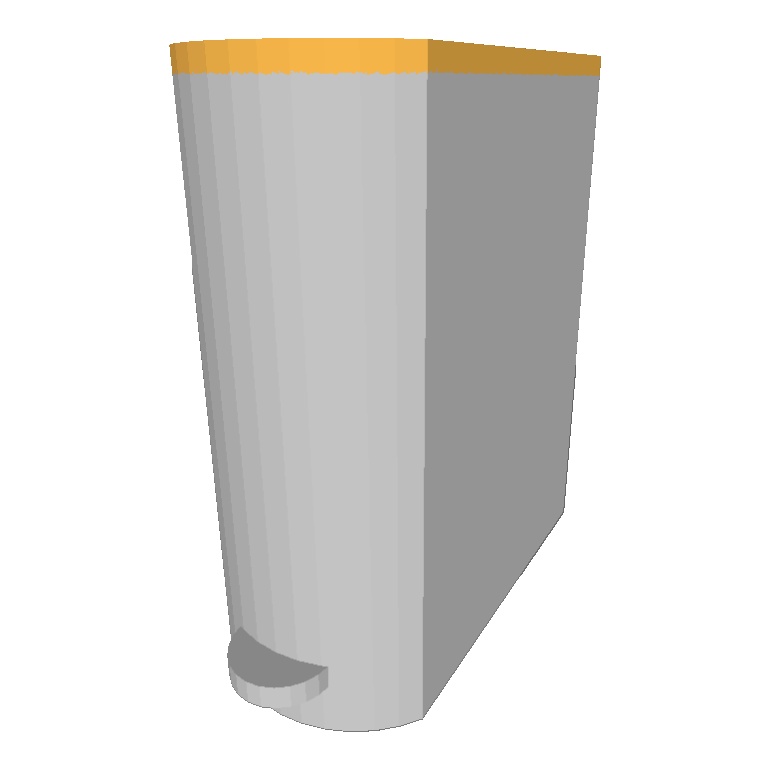} &
\includegraphics[width=0.1\textwidth]{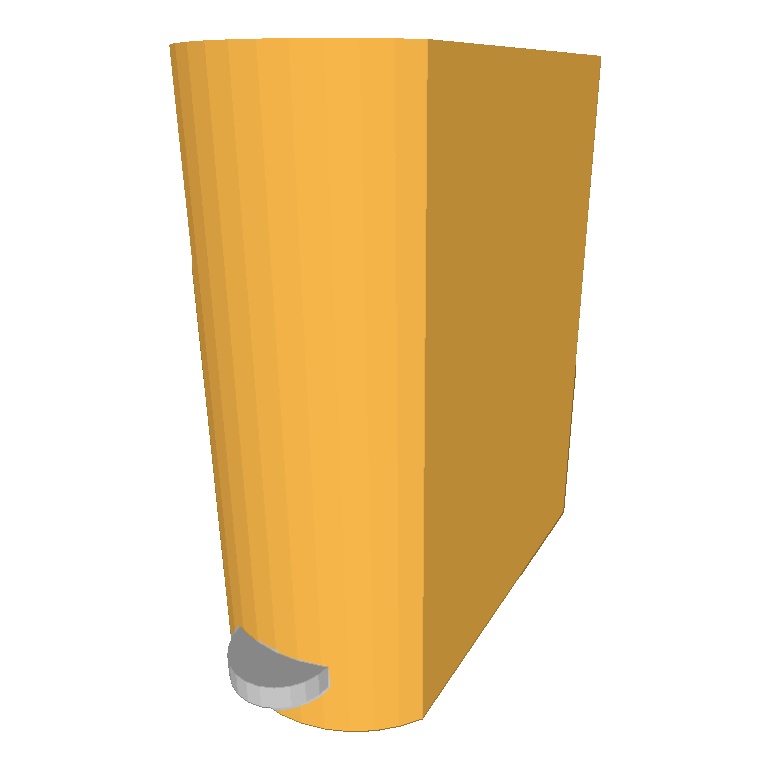} &
\includegraphics[width=0.1\textwidth]{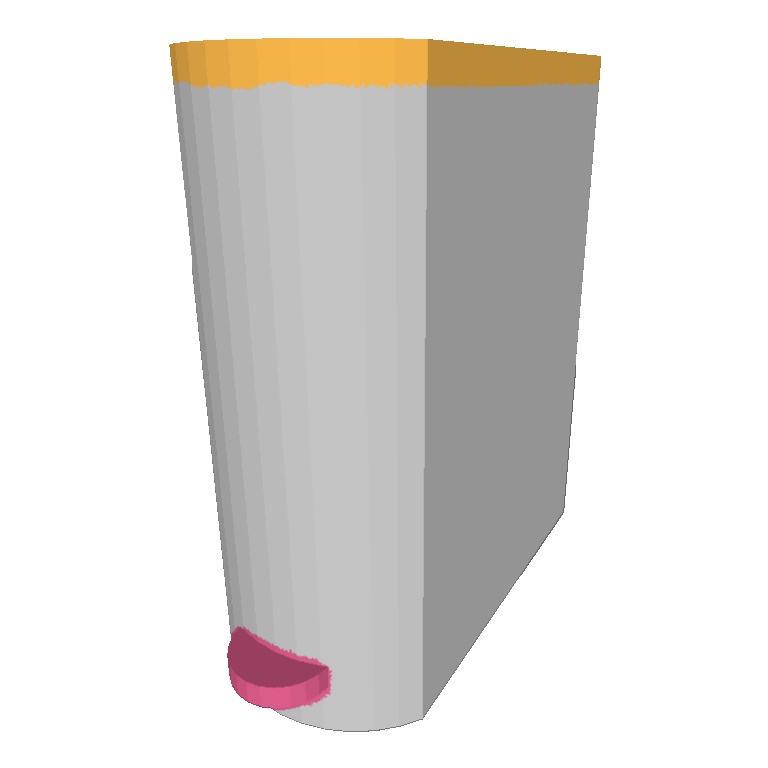} \\

\addlinespace[-2pt]
\arrayrulecolor{gray}\cmidrule(lr){1-5}
\arrayrulecolor{black}

\includegraphics[width=0.1\textwidth]{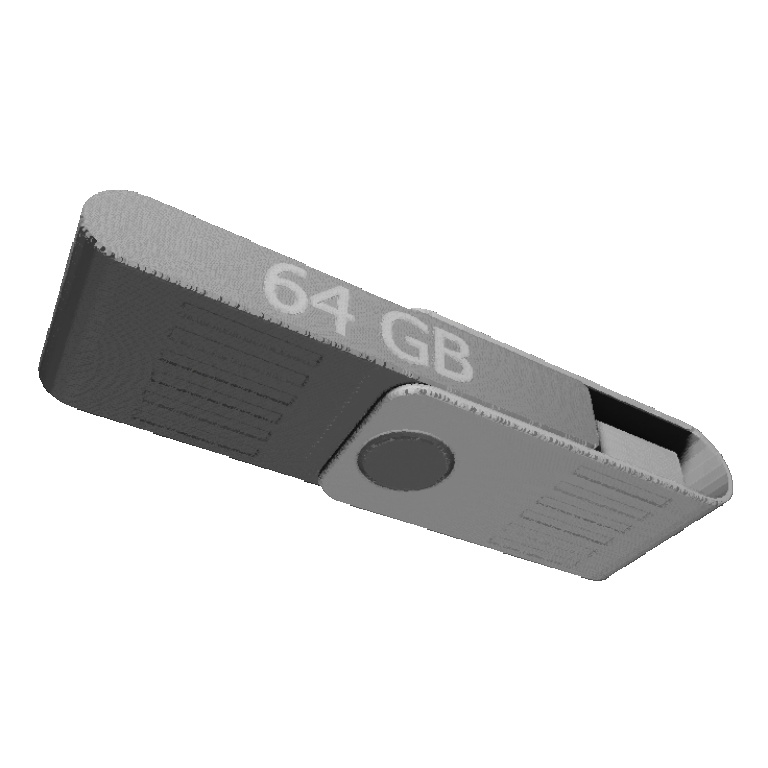} &
\includegraphics[width=0.1\textwidth]{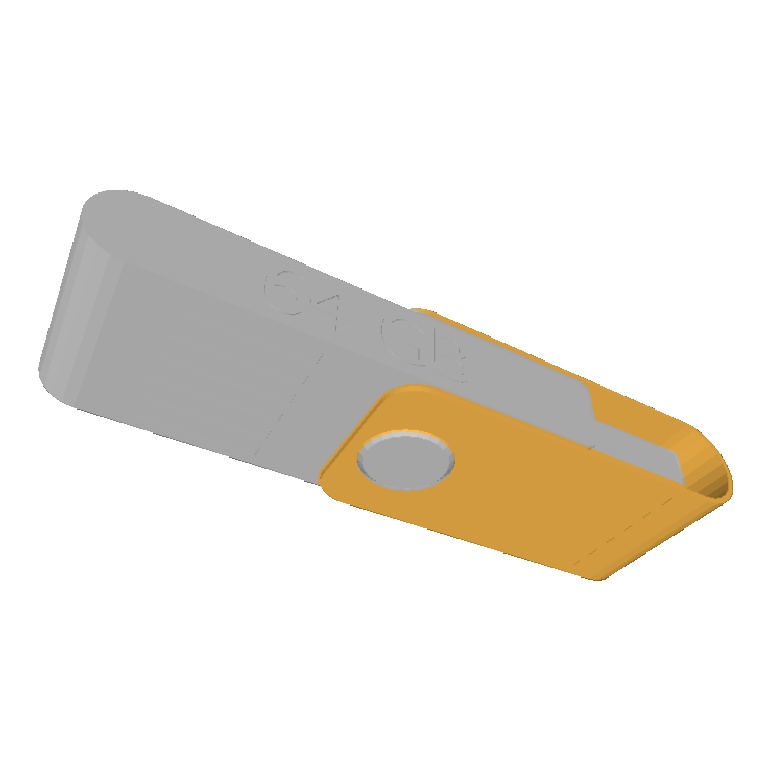} &
\includegraphics[width=0.1\textwidth]{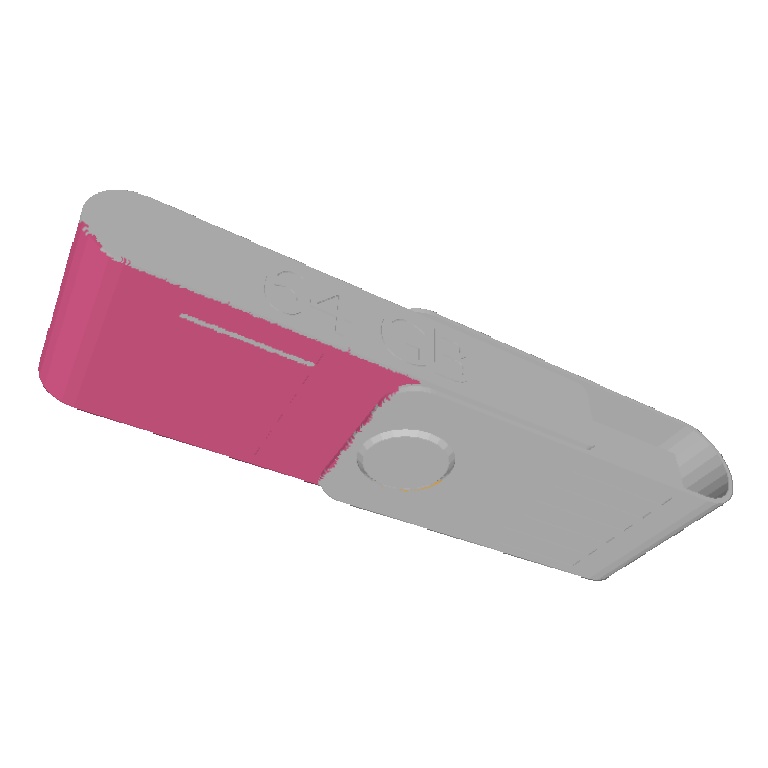} &
\includegraphics[width=0.1\textwidth]{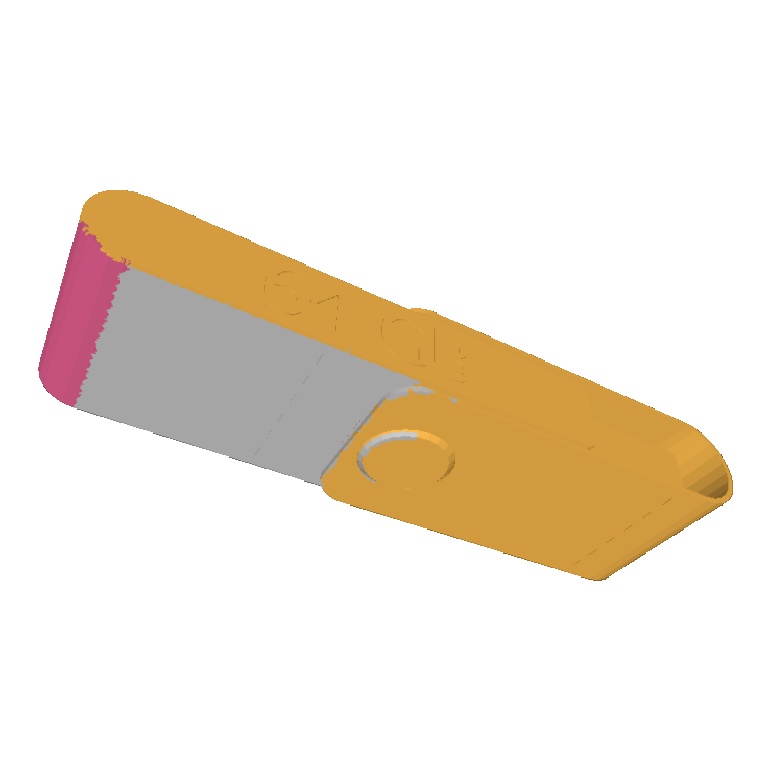} &
\includegraphics[width=0.1\textwidth]{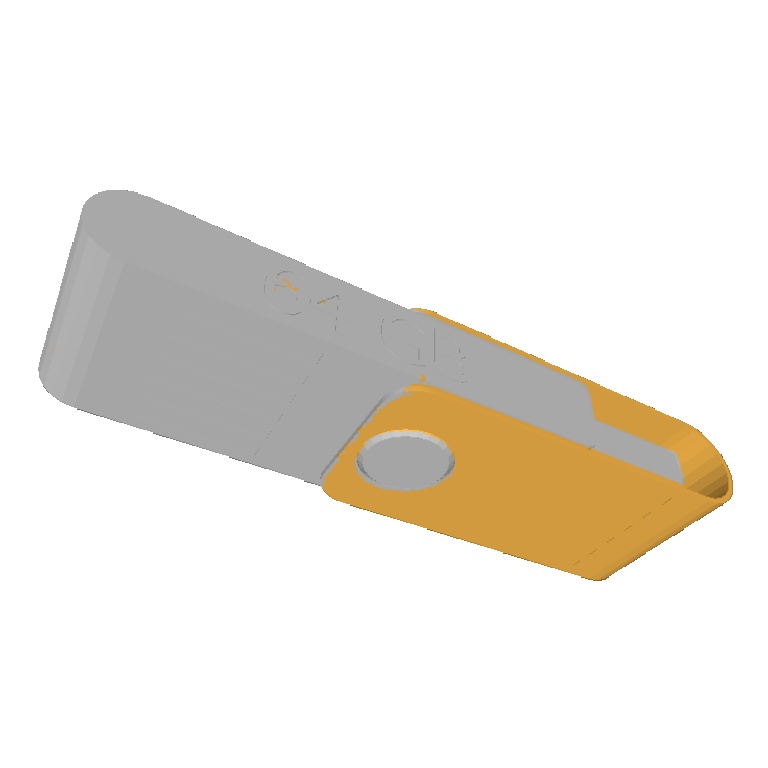} \\

\addlinespace[-2pt]
\bottomrule
\end{tabular}
}}
&
\vtop{\vskip0pt
\resizebox{0.5\textwidth}{!}{
\begin{tabular}{@{}c@{}c@{}c@{}c@{}c@{}}
\toprule
Input & GT  & PartSLIP & PartSTAD & Ours \\ \midrule
\includegraphics[width=0.1\textwidth]{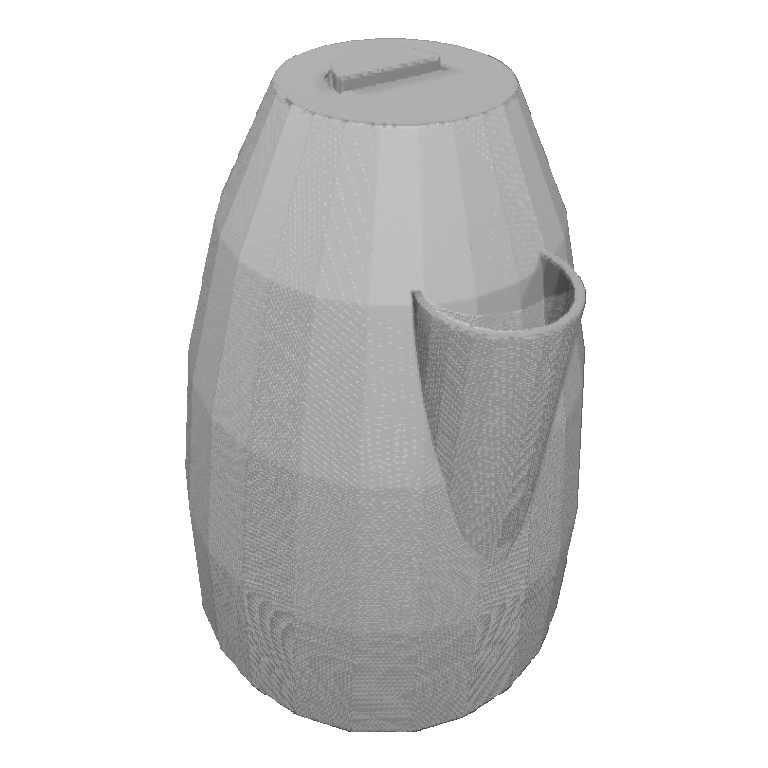} &
\includegraphics[width=0.1\textwidth]{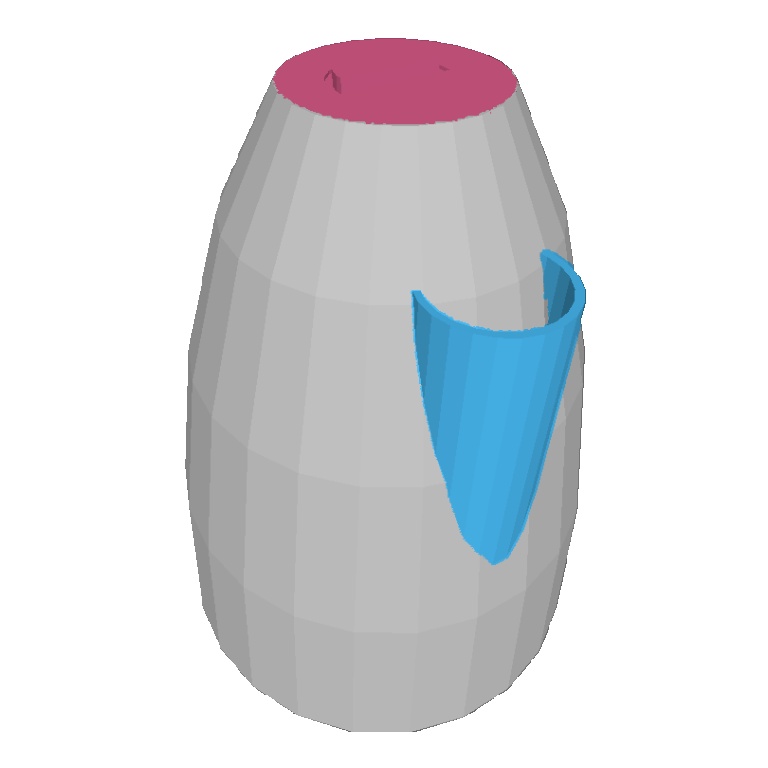} &
\includegraphics[width=0.1\textwidth]{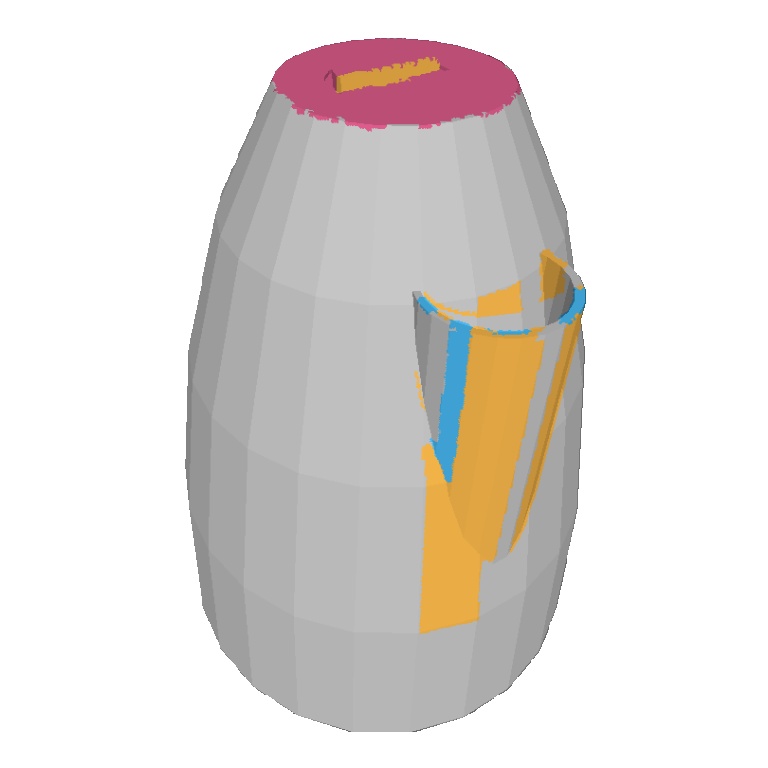} &
\includegraphics[width=0.1\textwidth]{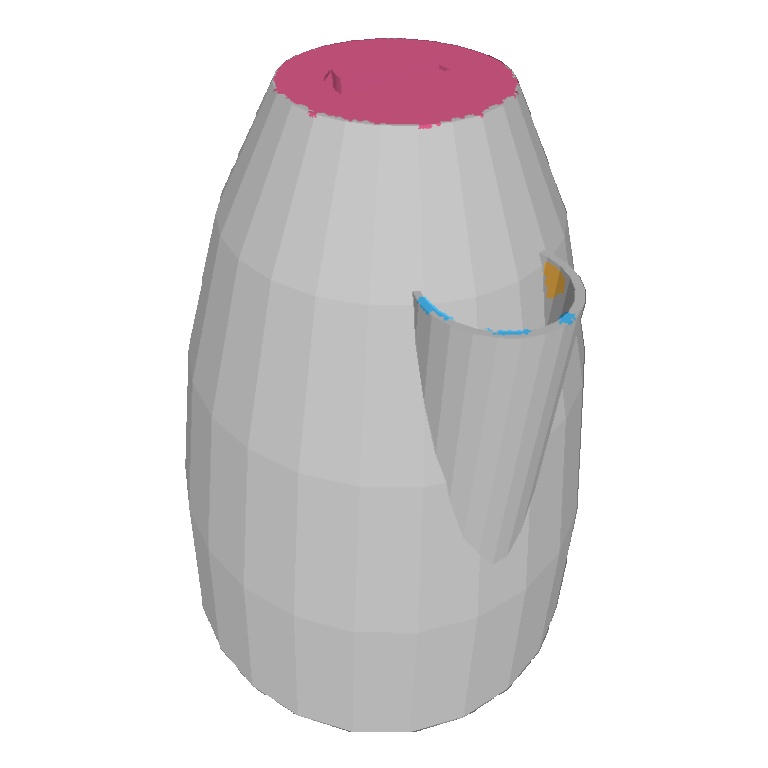} &
\includegraphics[width=0.1\textwidth]{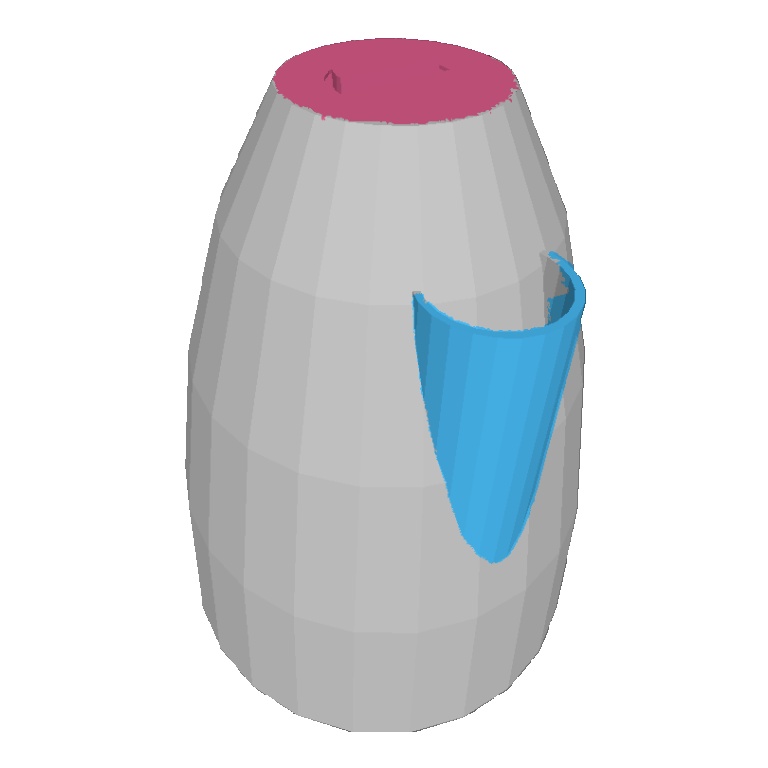} \\

\addlinespace[-2pt]
\arrayrulecolor{gray}\cmidrule(lr){1-5}
\arrayrulecolor{black}

\includegraphics[width=0.1\textwidth]{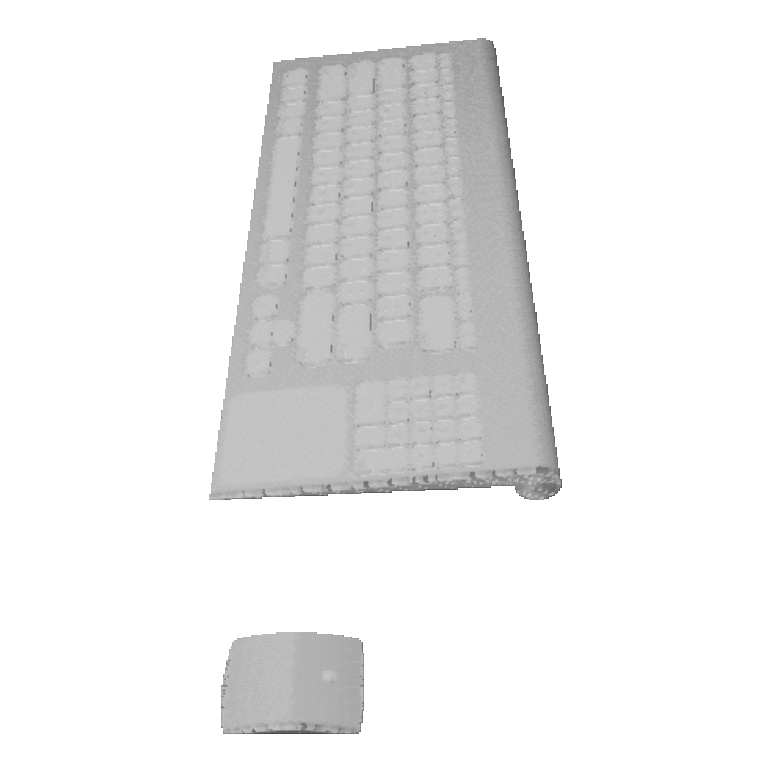} &
\includegraphics[width=0.1\textwidth]{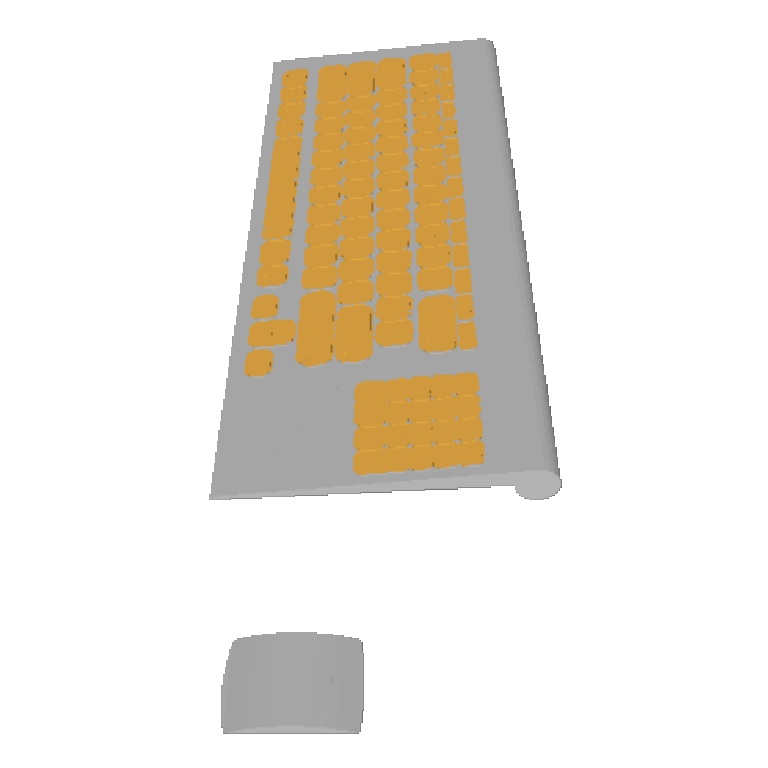} &
\includegraphics[width=0.1\textwidth]{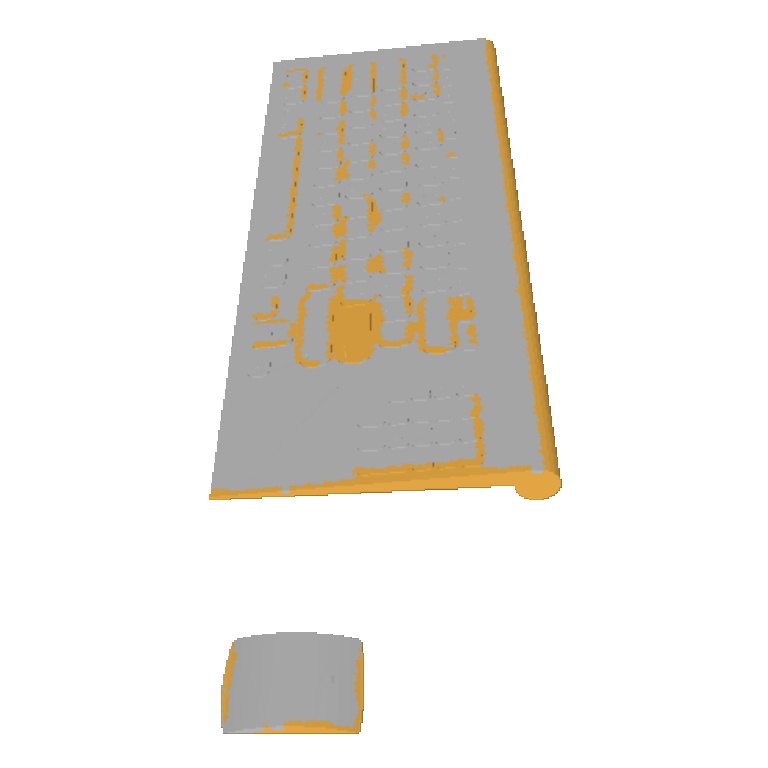} &
\includegraphics[width=0.1\textwidth]{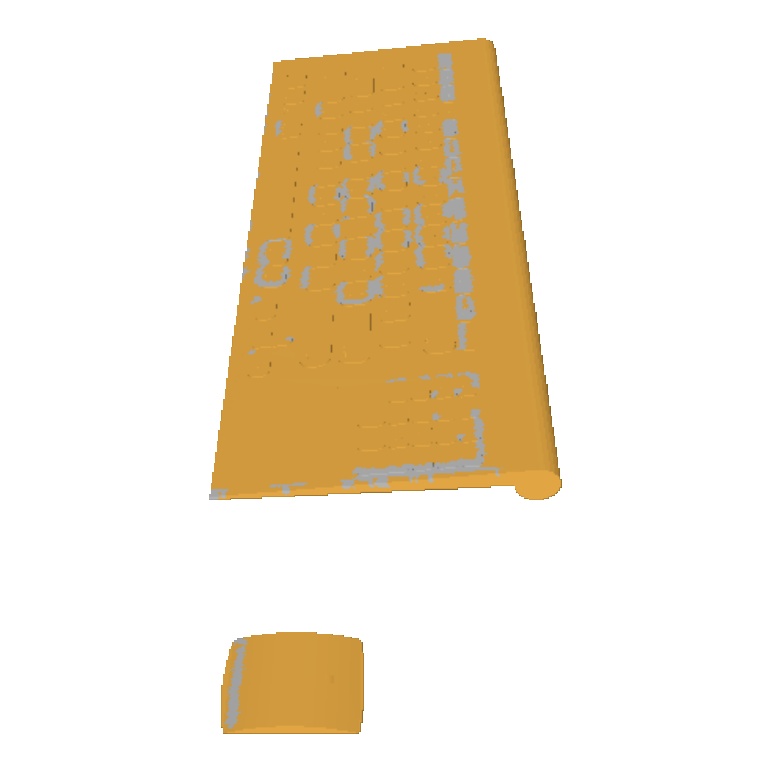} &
\includegraphics[width=0.1\textwidth]{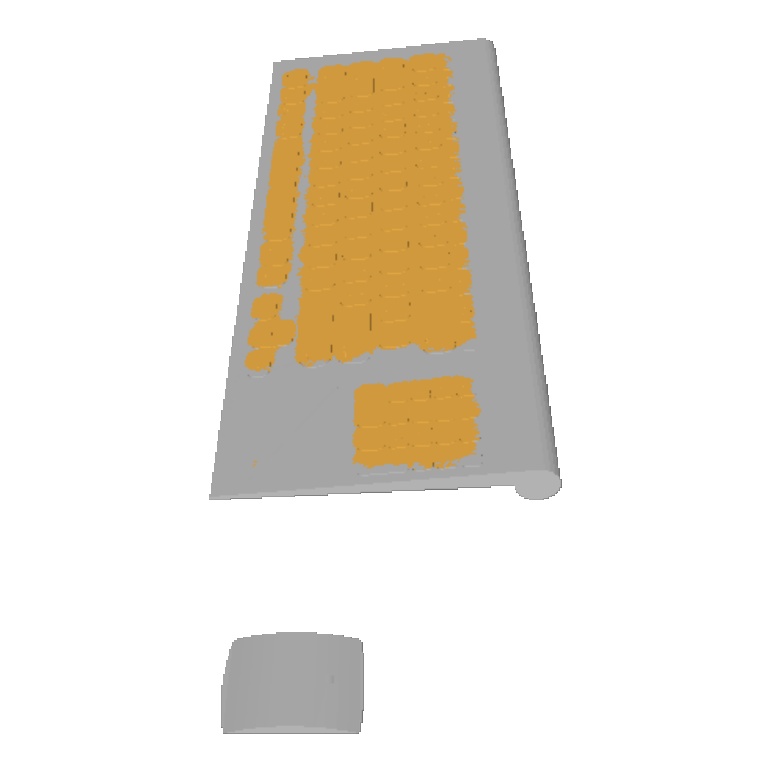} \\

\addlinespace[-2pt]
\arrayrulecolor{gray}\cmidrule(lr){1-5}
\arrayrulecolor{black}

\includegraphics[width=0.1\textwidth]{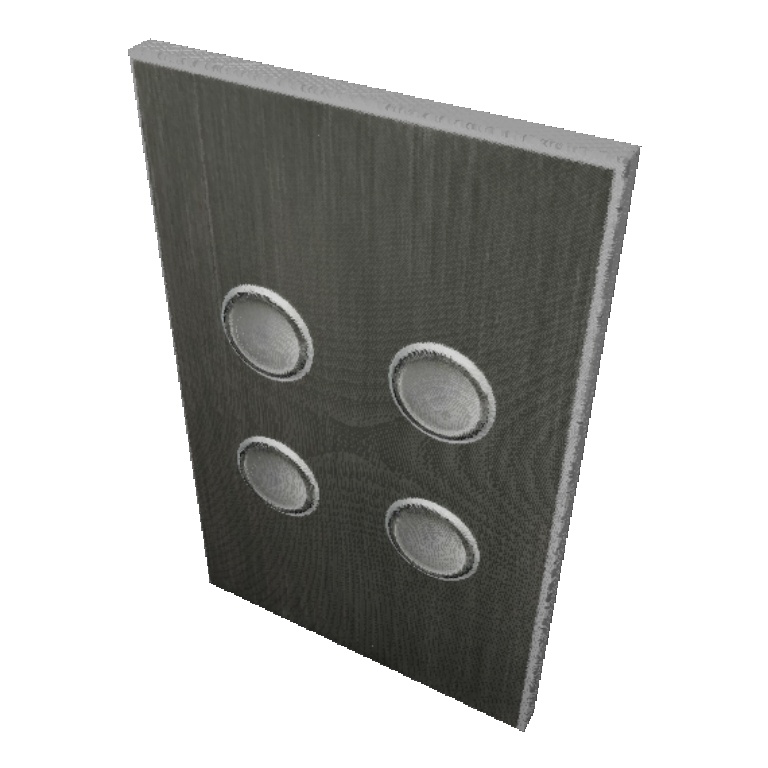} &
\includegraphics[width=0.1\textwidth]{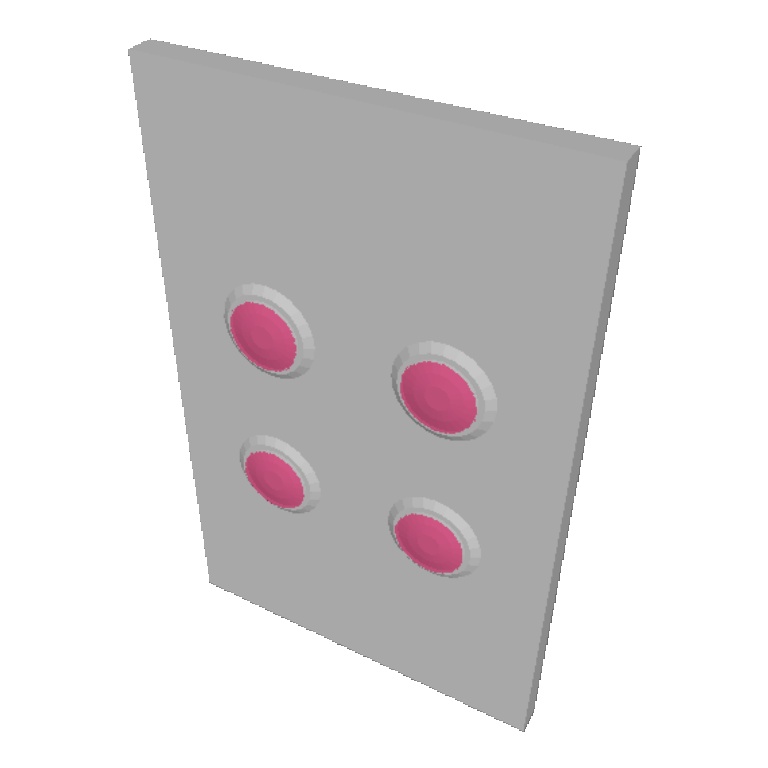} &
\includegraphics[width=0.1\textwidth]{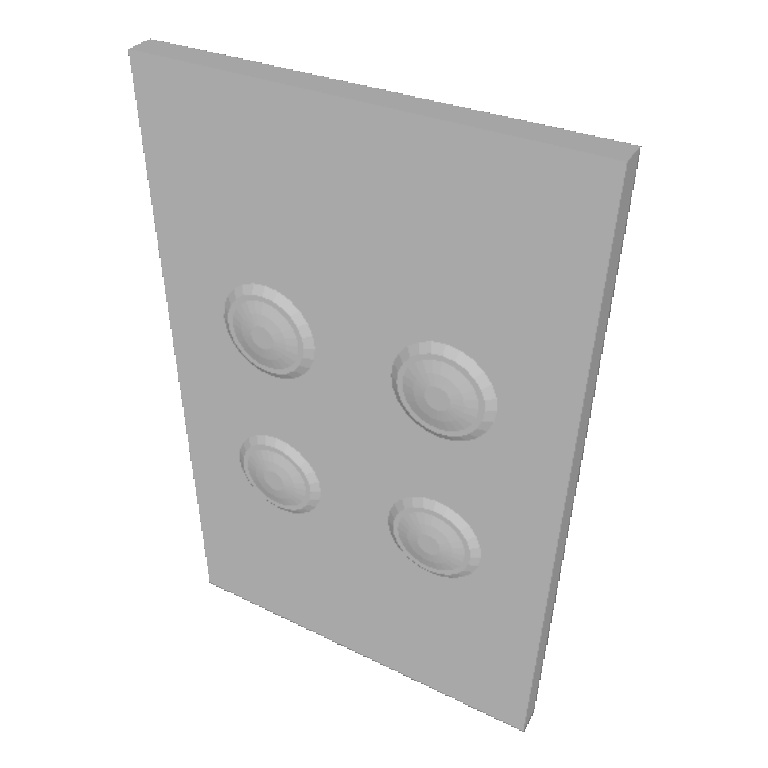} &
\includegraphics[width=0.1\textwidth]{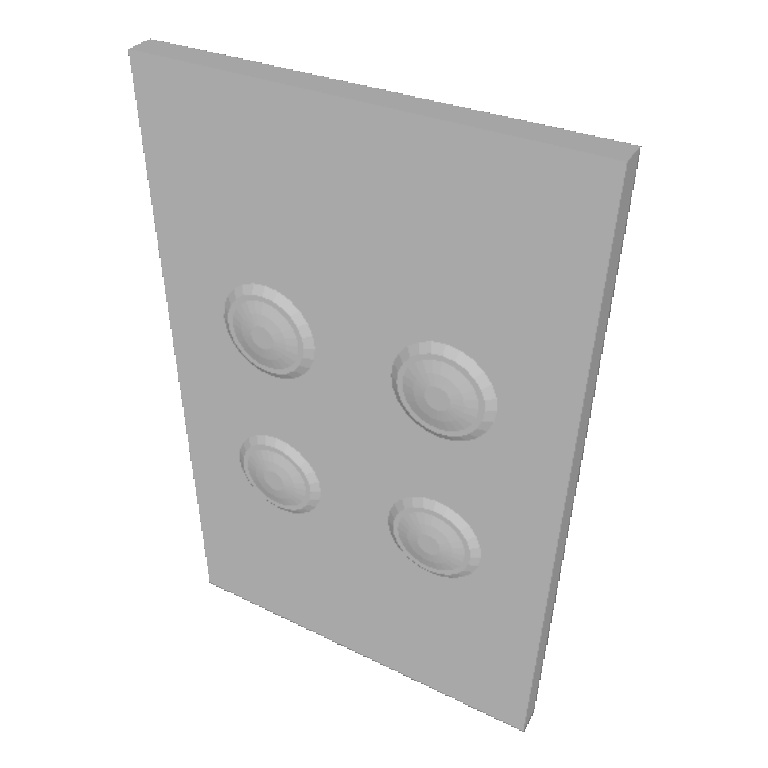} &
\includegraphics[width=0.1\textwidth]{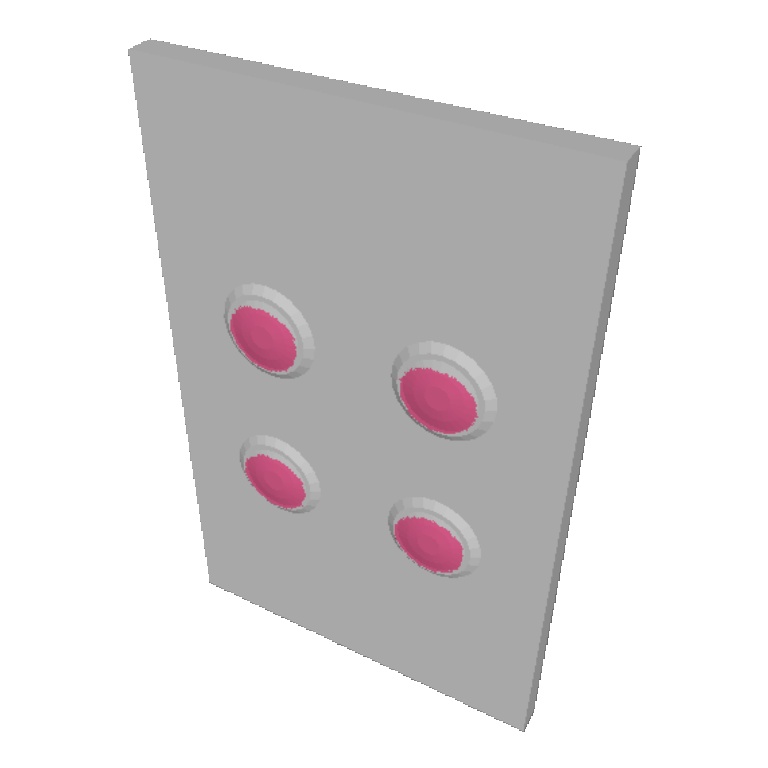} \\

\addlinespace[-2pt]
\arrayrulecolor{gray}\cmidrule(lr){1-5}
\arrayrulecolor{black}

\includegraphics[width=0.1\textwidth]{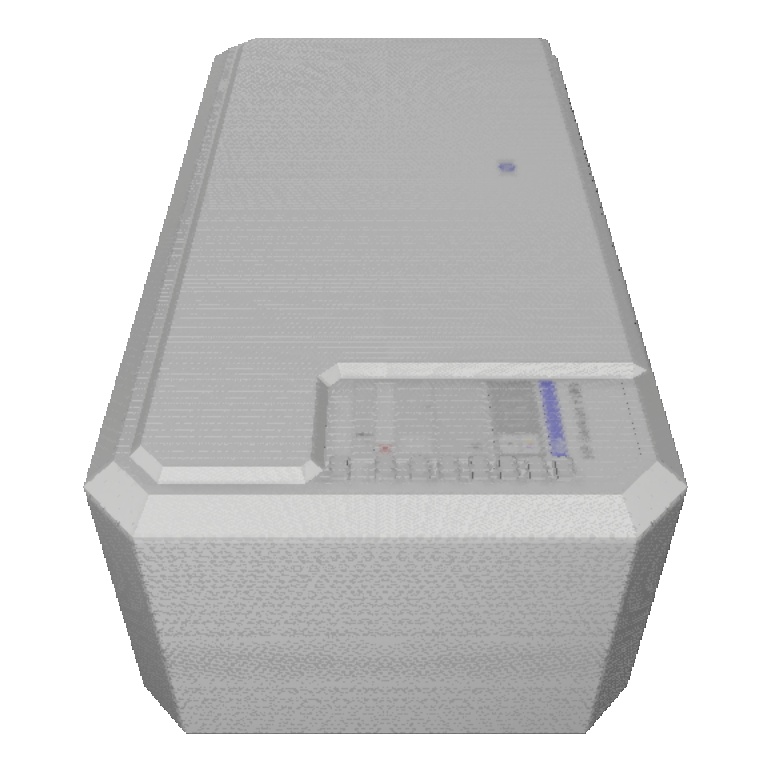} &
\includegraphics[width=0.1\textwidth]{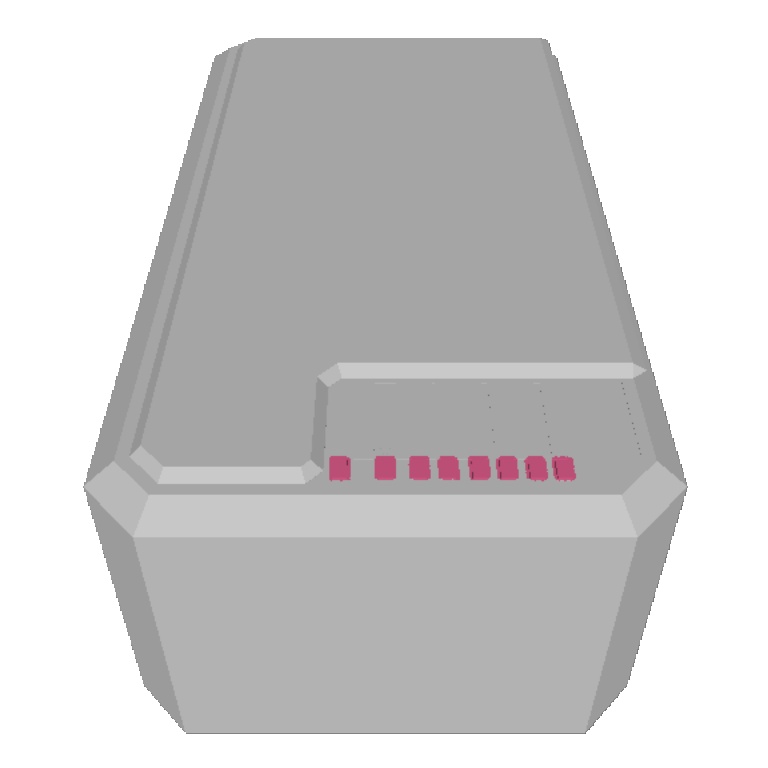} &
\includegraphics[width=0.1\textwidth]{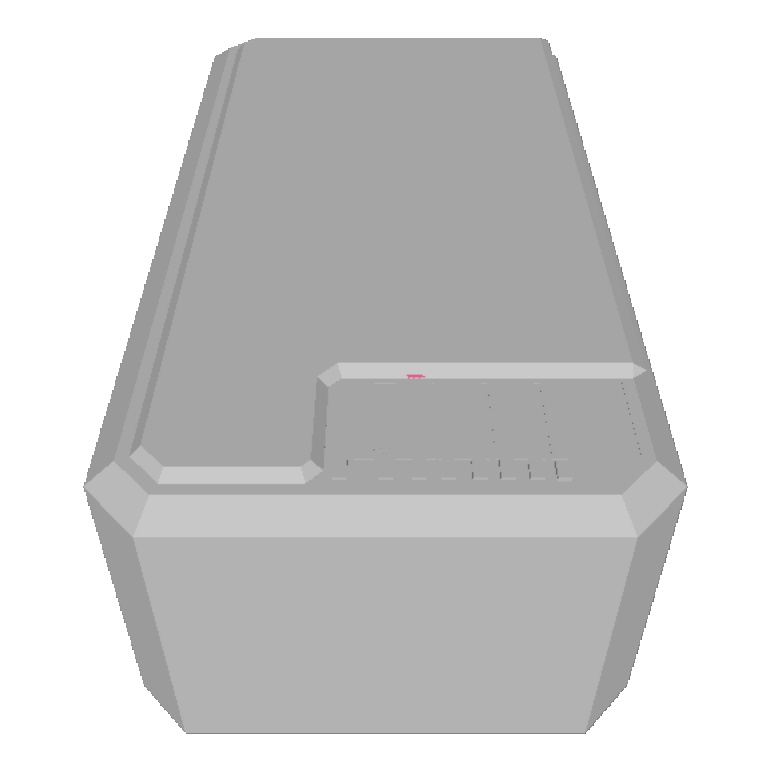} &
\includegraphics[width=0.1\textwidth]{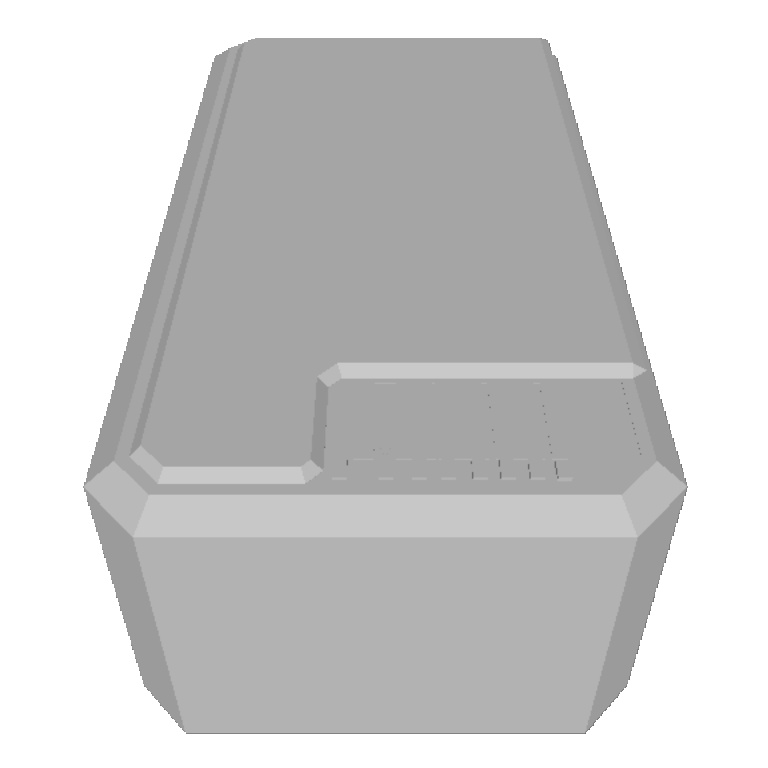} &
\includegraphics[width=0.1\textwidth]{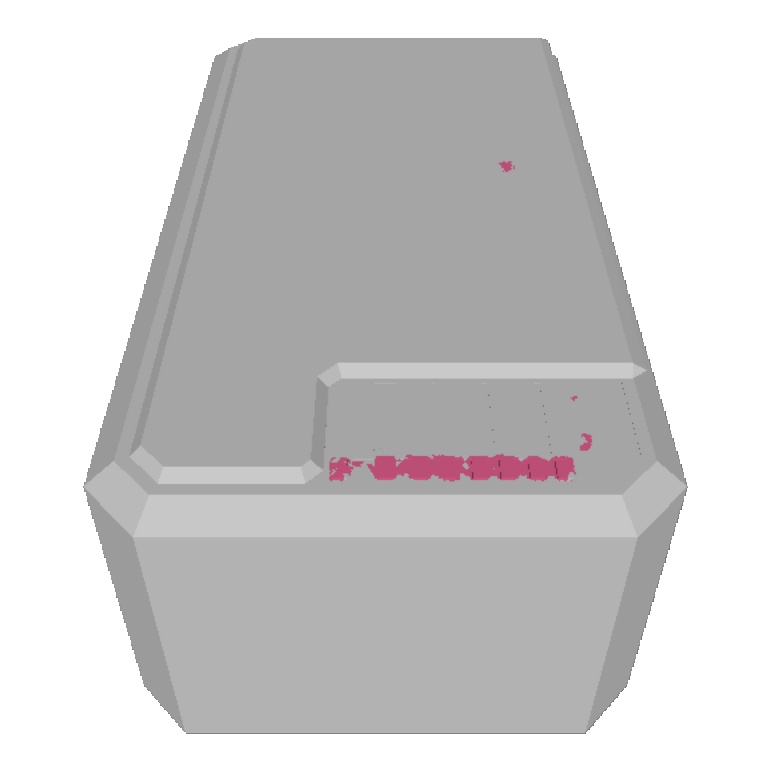} \\

\addlinespace[-2pt]
\arrayrulecolor{gray}\cmidrule(lr){1-5}
\arrayrulecolor{black}

\includegraphics[width=0.1\textwidth]{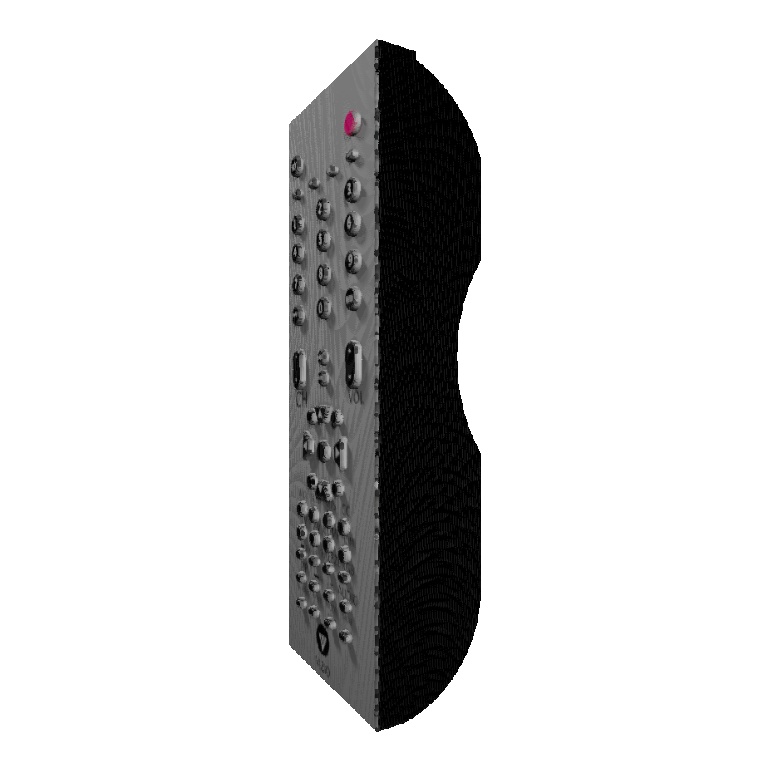} &
\includegraphics[width=0.1\textwidth]{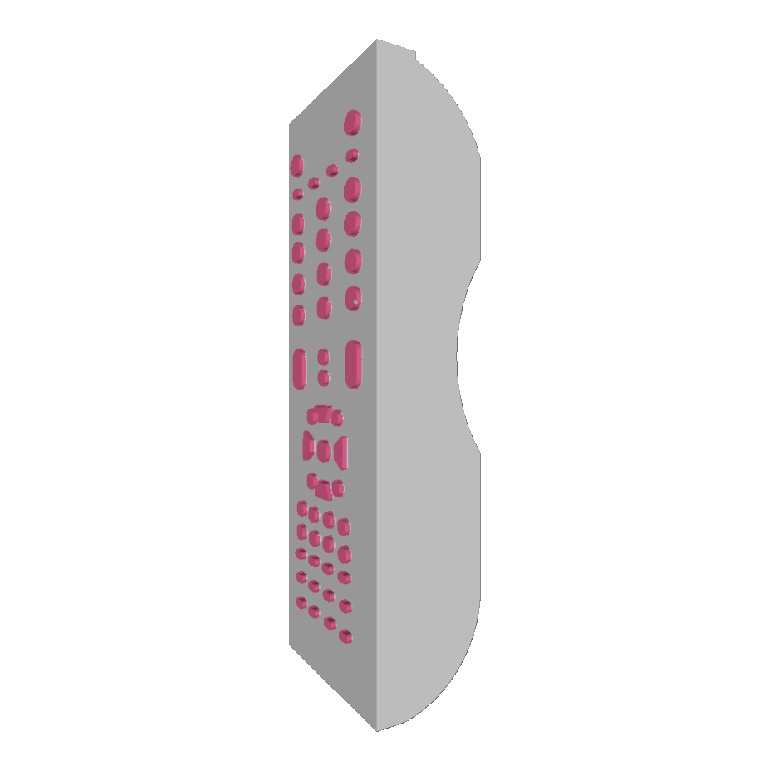} &
\includegraphics[width=0.1\textwidth]{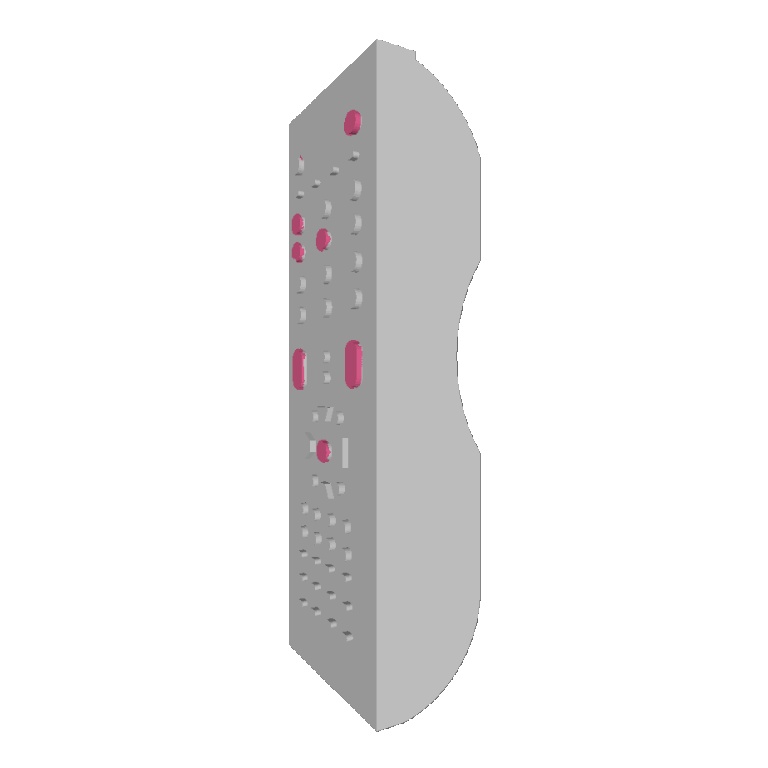} &
\includegraphics[width=0.1\textwidth]{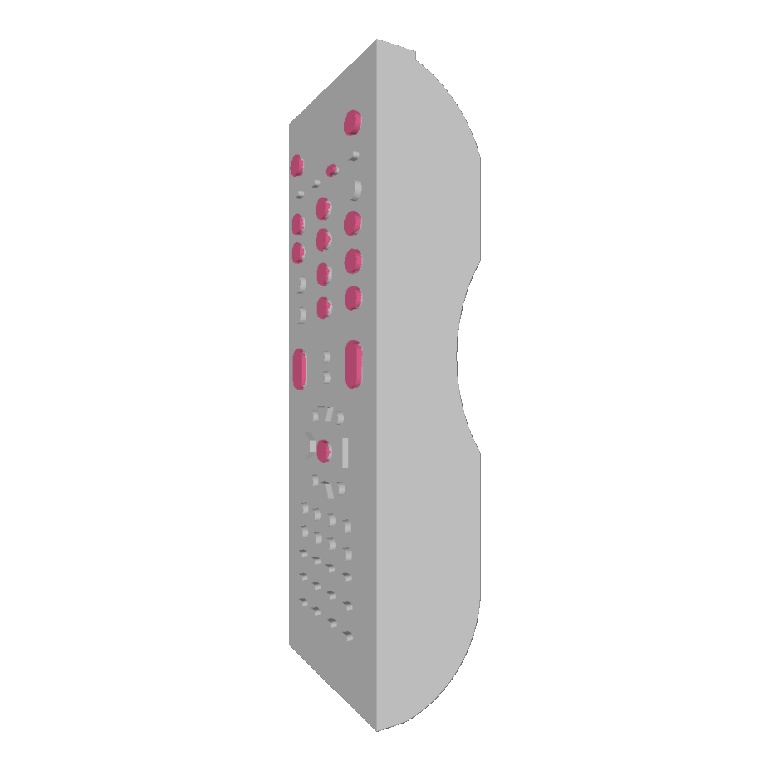} &
\includegraphics[width=0.1\textwidth]{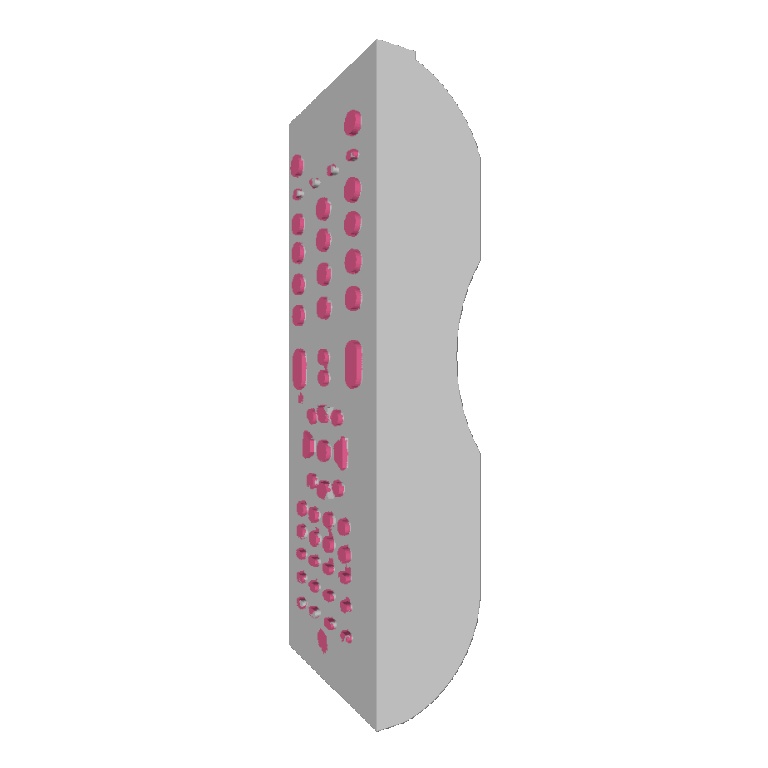} \\

\addlinespace[-2pt]
\arrayrulecolor{gray}\cmidrule(lr){1-5}
\arrayrulecolor{black}

\includegraphics[width=0.1\textwidth]{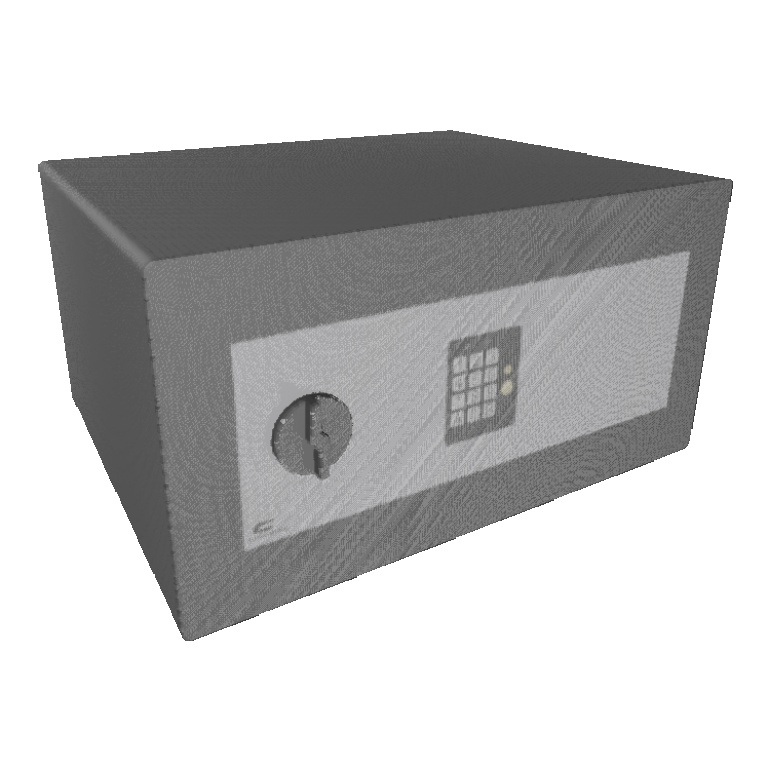} &
\includegraphics[width=0.1\textwidth]{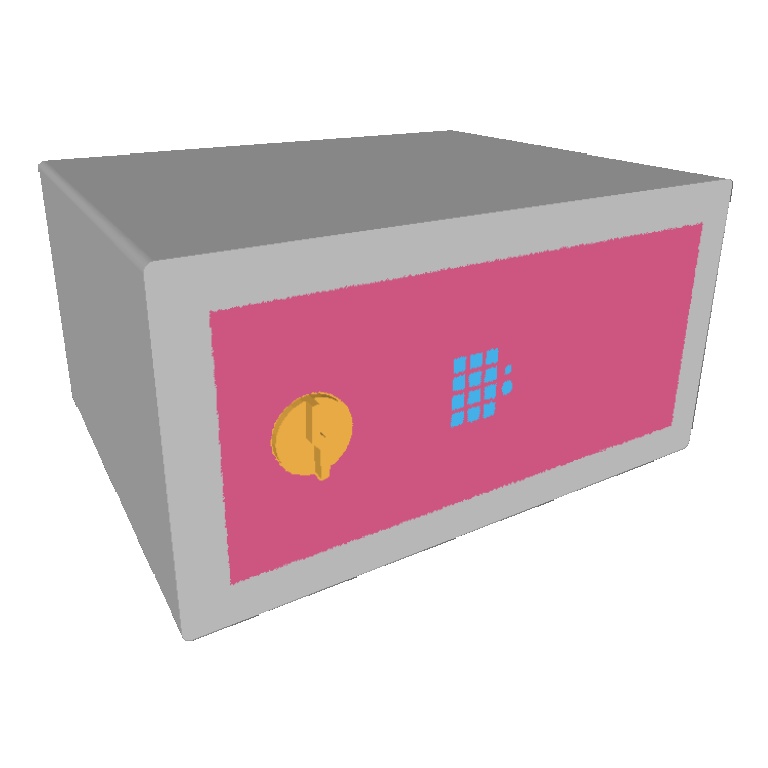} &
\includegraphics[width=0.1\textwidth]{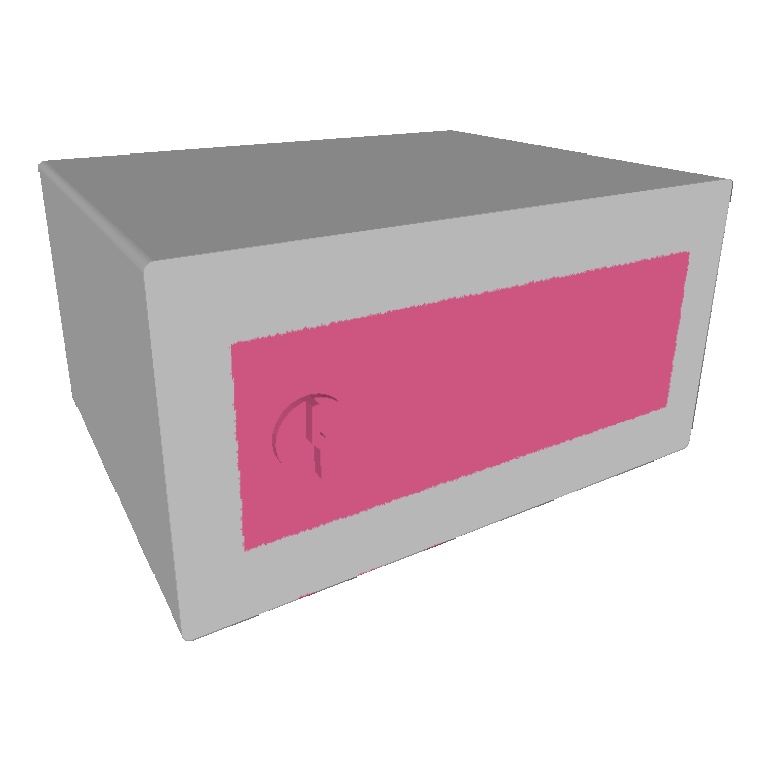} &
\includegraphics[width=0.1\textwidth]{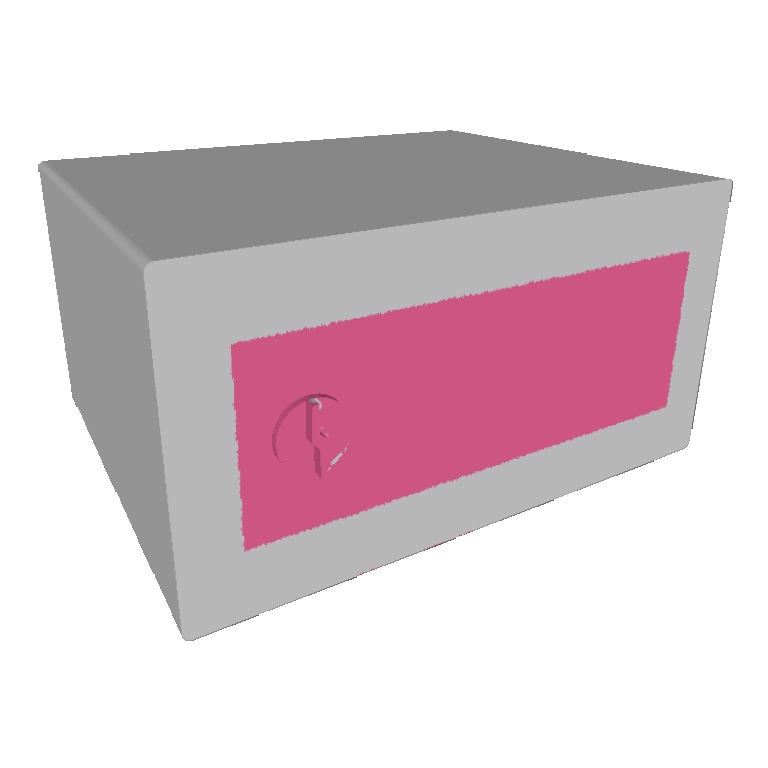} &
\includegraphics[width=0.1\textwidth]{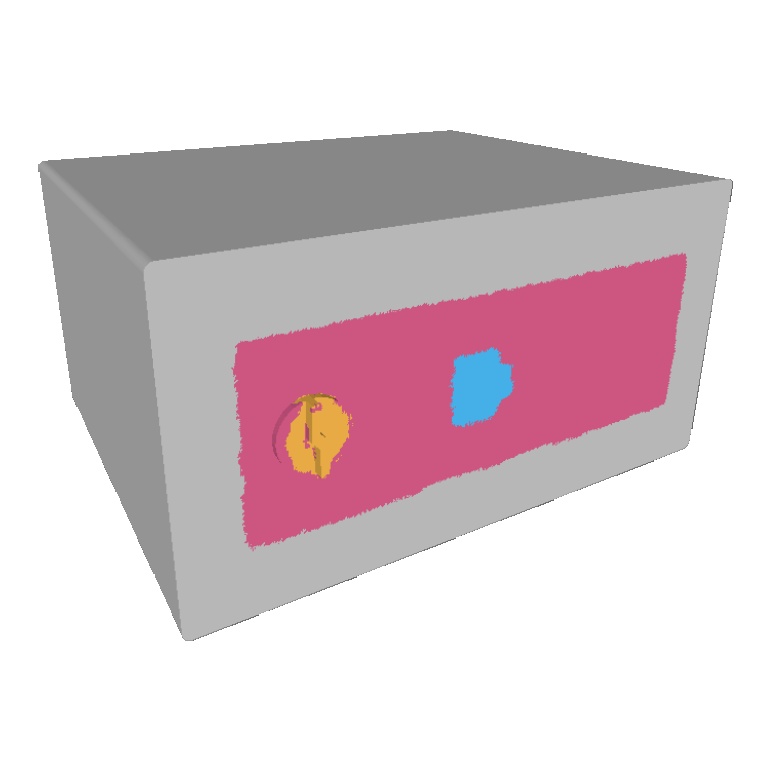} \\

\addlinespace[-2pt]
\arrayrulecolor{gray}\cmidrule(lr){1-5}
\arrayrulecolor{black}

\includegraphics[width=0.1\textwidth]{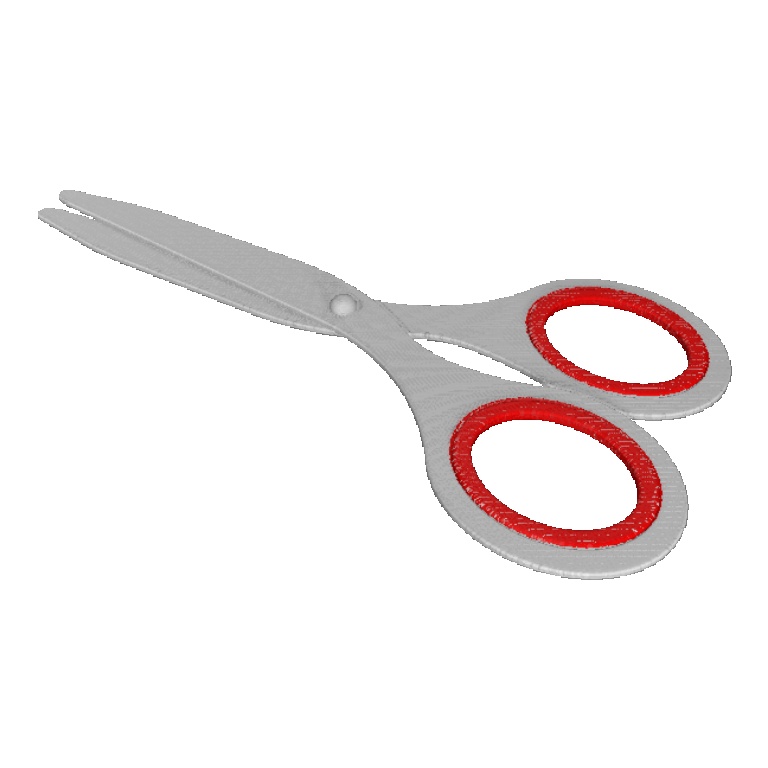} &
\includegraphics[width=0.1\textwidth]{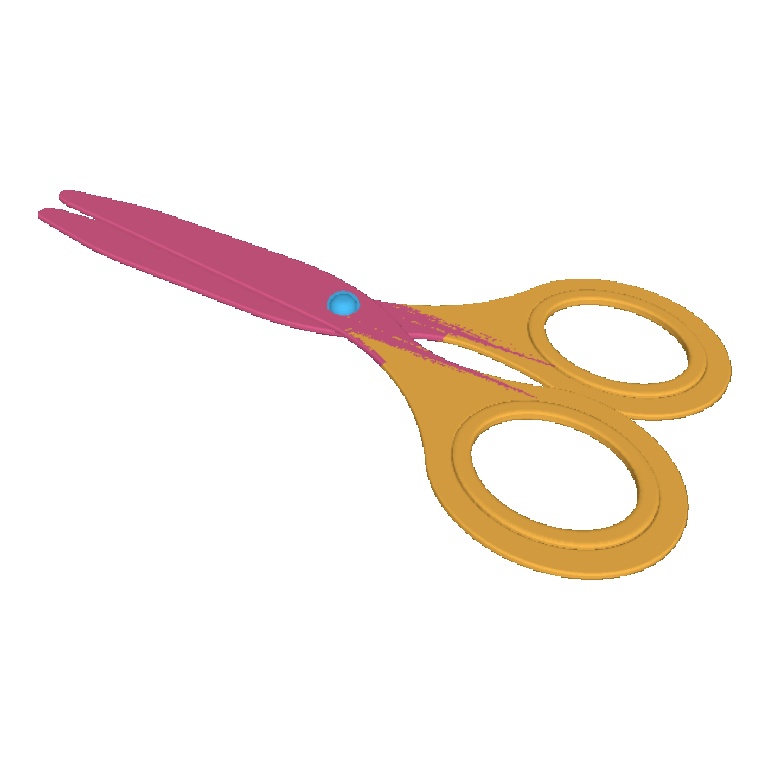} &
\includegraphics[width=0.1\textwidth]{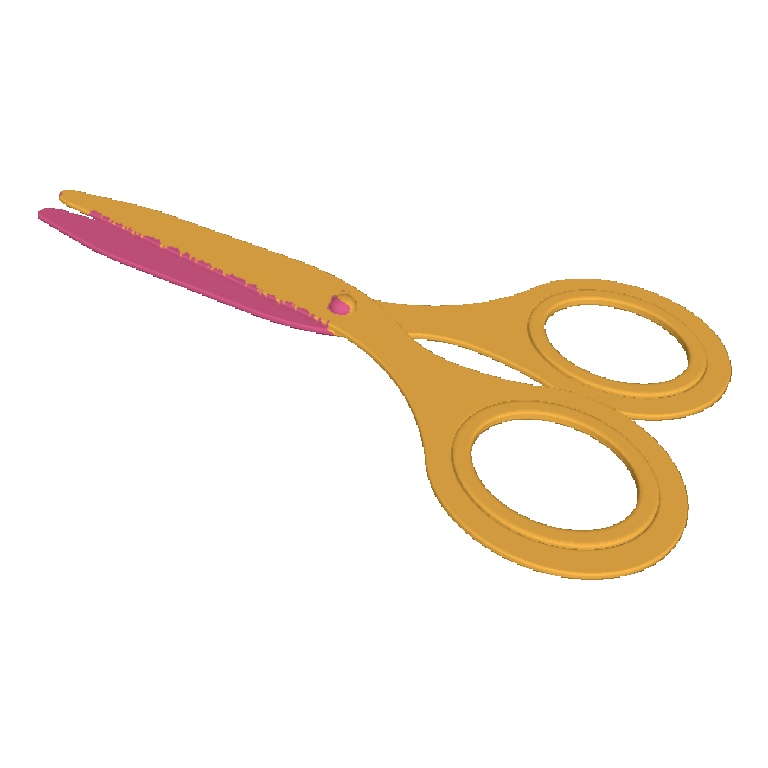} &
\includegraphics[width=0.1\textwidth]{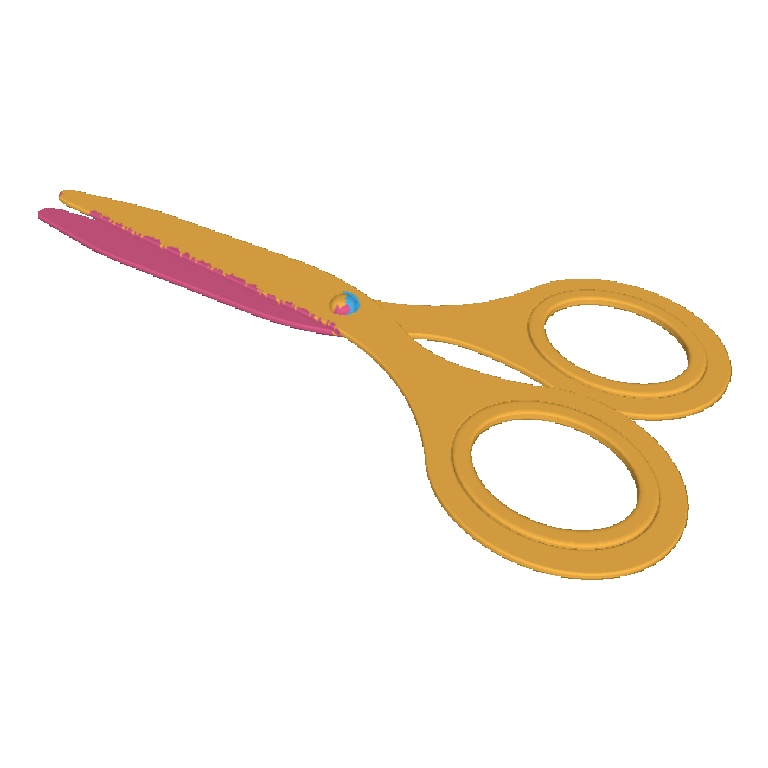} &
\includegraphics[width=0.1\textwidth]{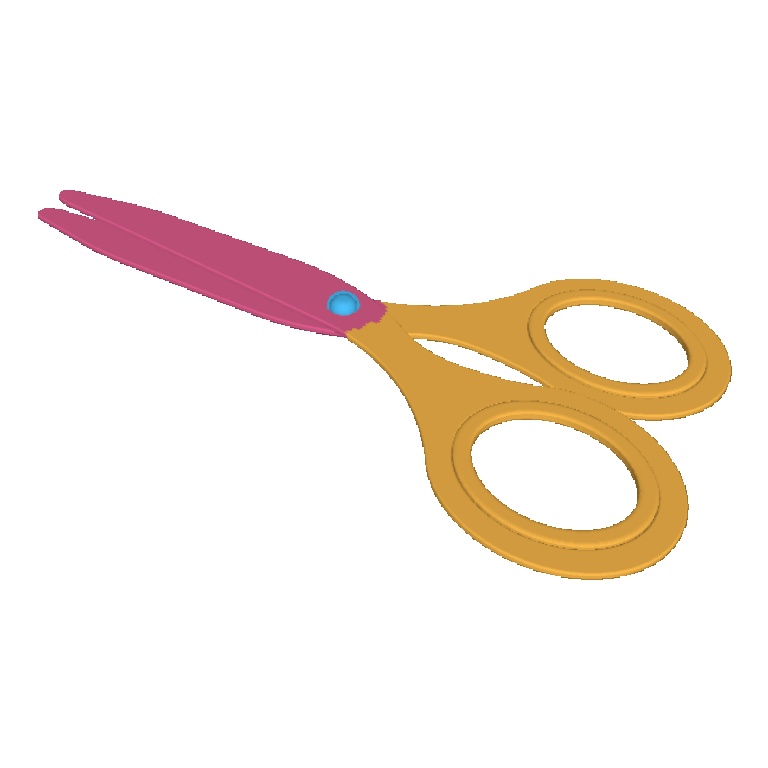} \\

\addlinespace[-2pt]
\arrayrulecolor{gray}\cmidrule(lr){1-5}
\arrayrulecolor{black}

\includegraphics[width=0.1\textwidth]{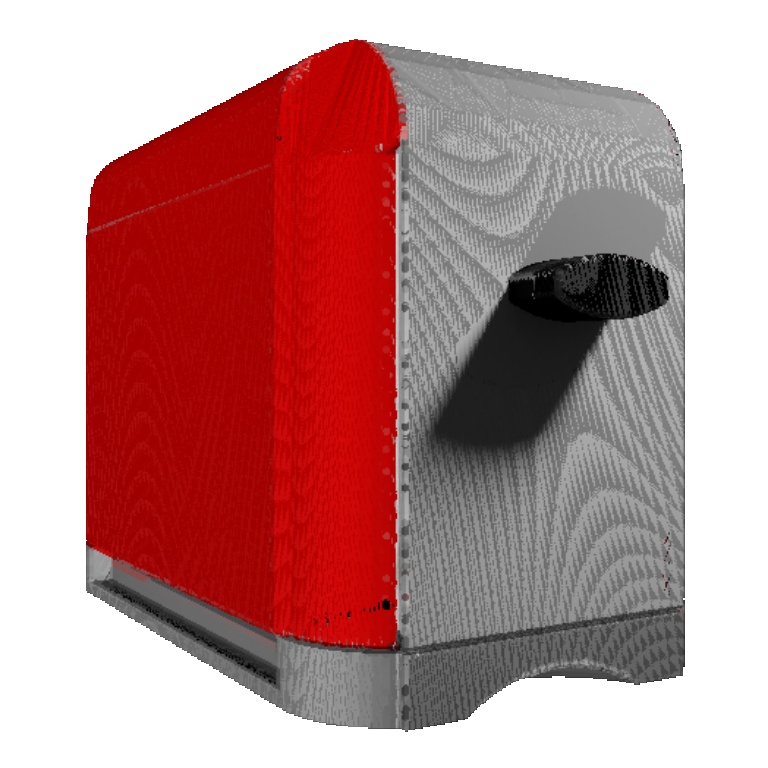} &
\includegraphics[width=0.1\textwidth]{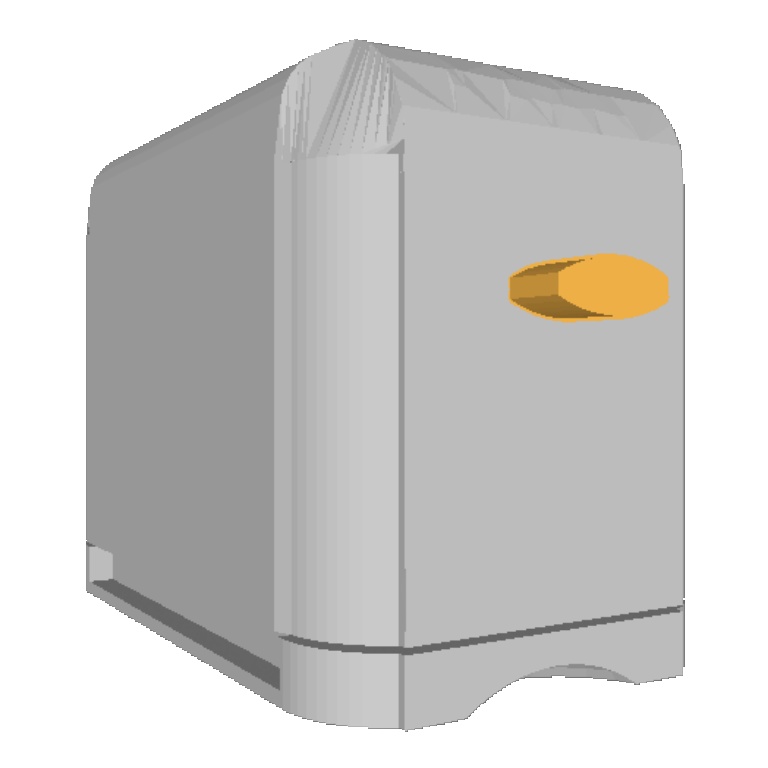} &
\includegraphics[width=0.1\textwidth]{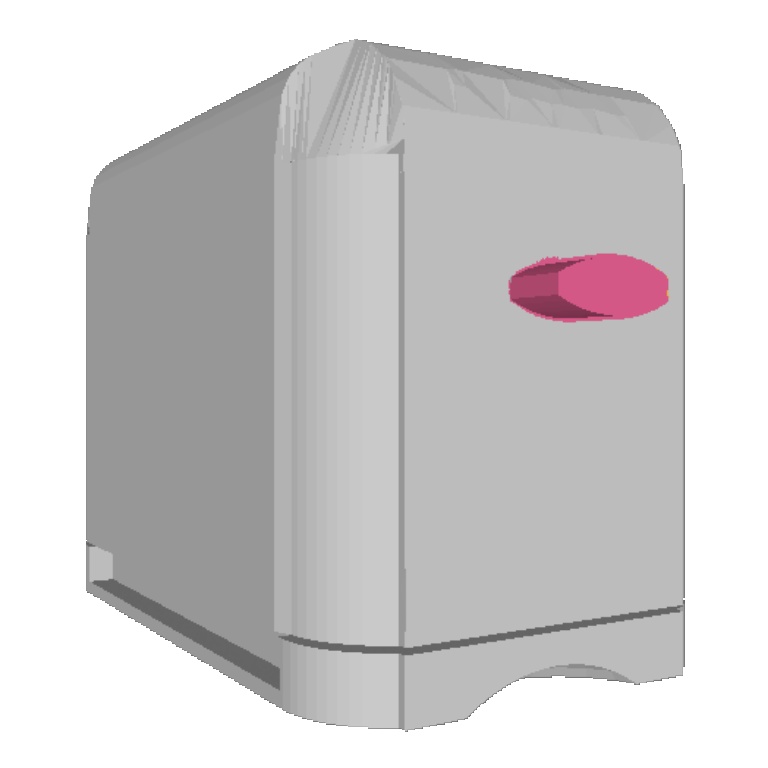} &
\includegraphics[width=0.1\textwidth]{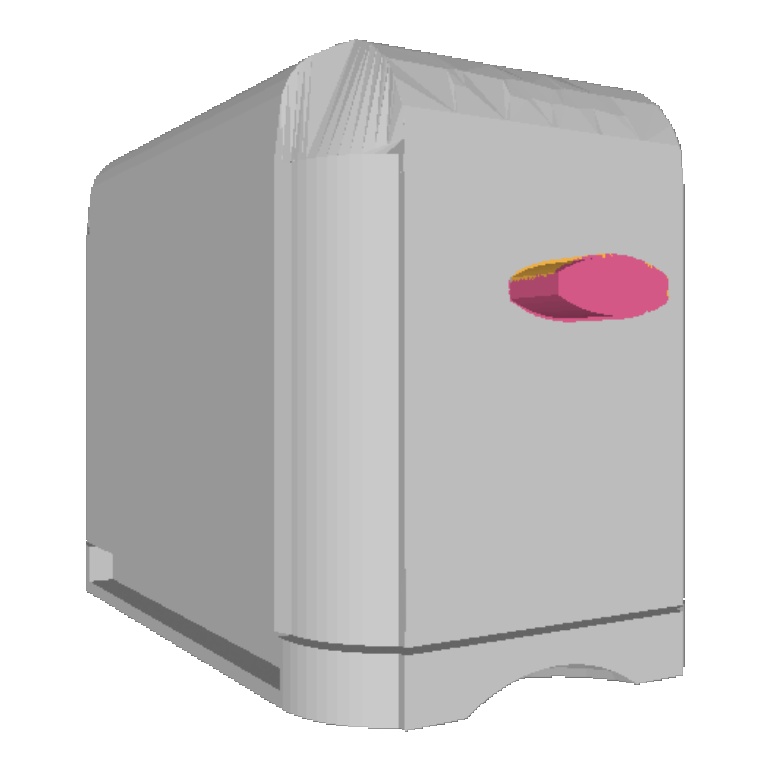} &
\includegraphics[width=0.1\textwidth]{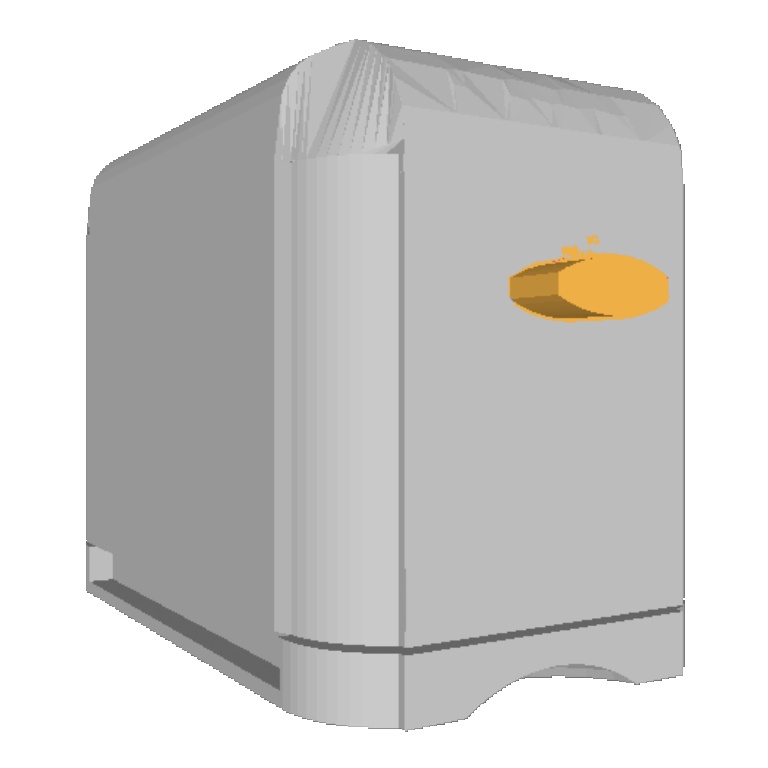} \\

\addlinespace[-2pt]
\arrayrulecolor{gray}\cmidrule(lr){1-5}
\arrayrulecolor{black}

\includegraphics[width=0.1\textwidth]{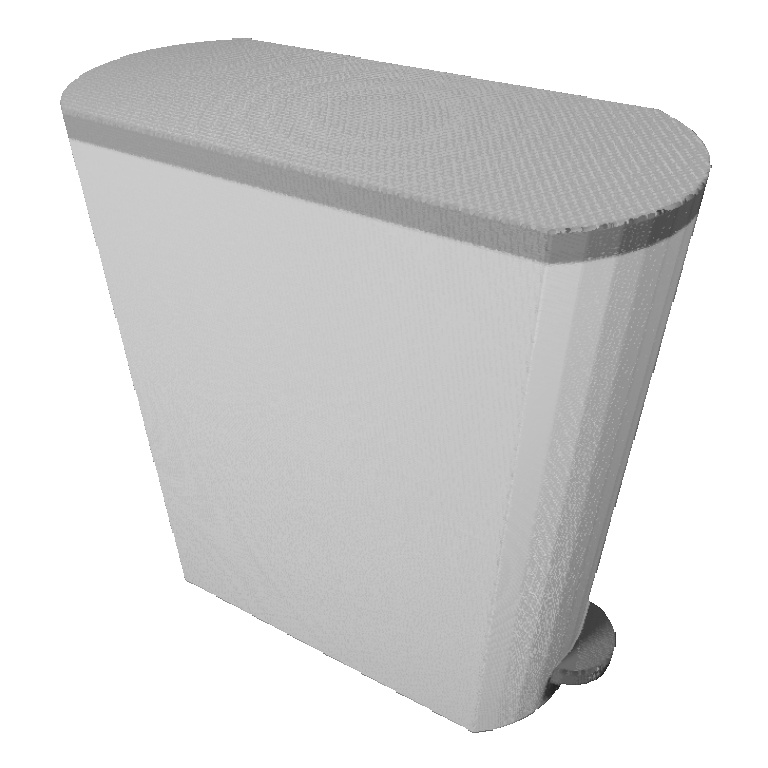} &
\includegraphics[width=0.1\textwidth]{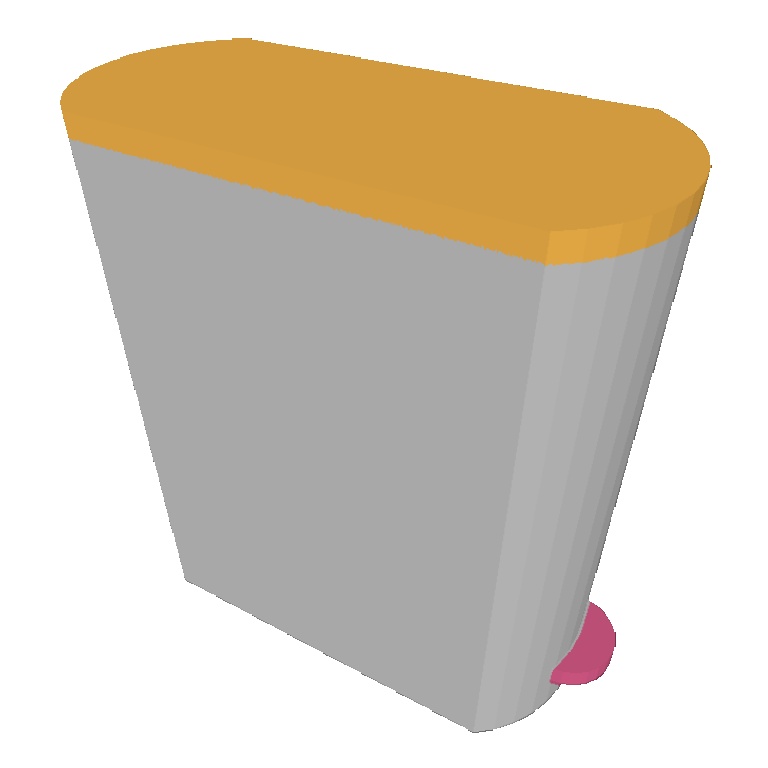} &
\includegraphics[width=0.1\textwidth]{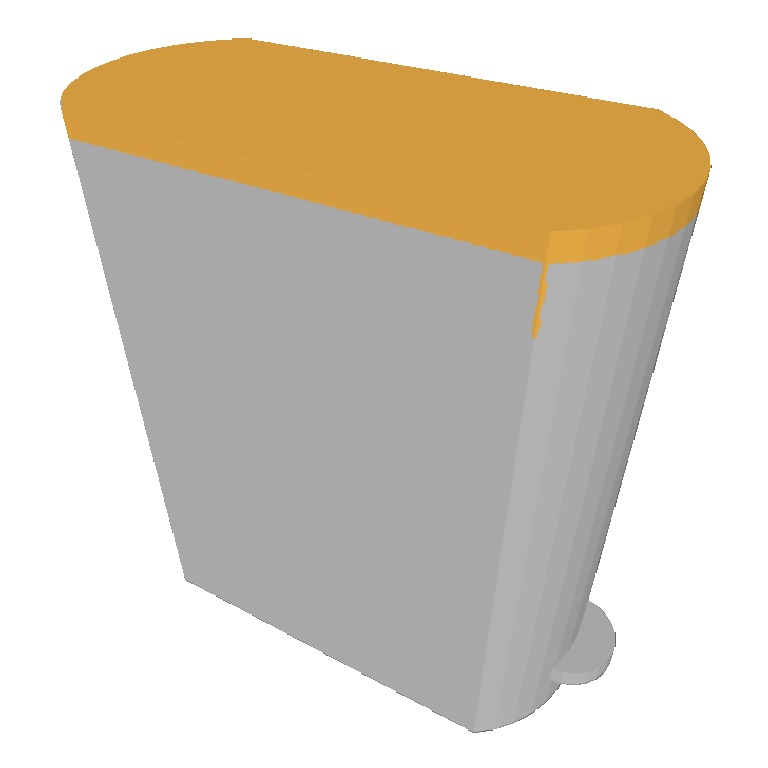} &
\includegraphics[width=0.1\textwidth]{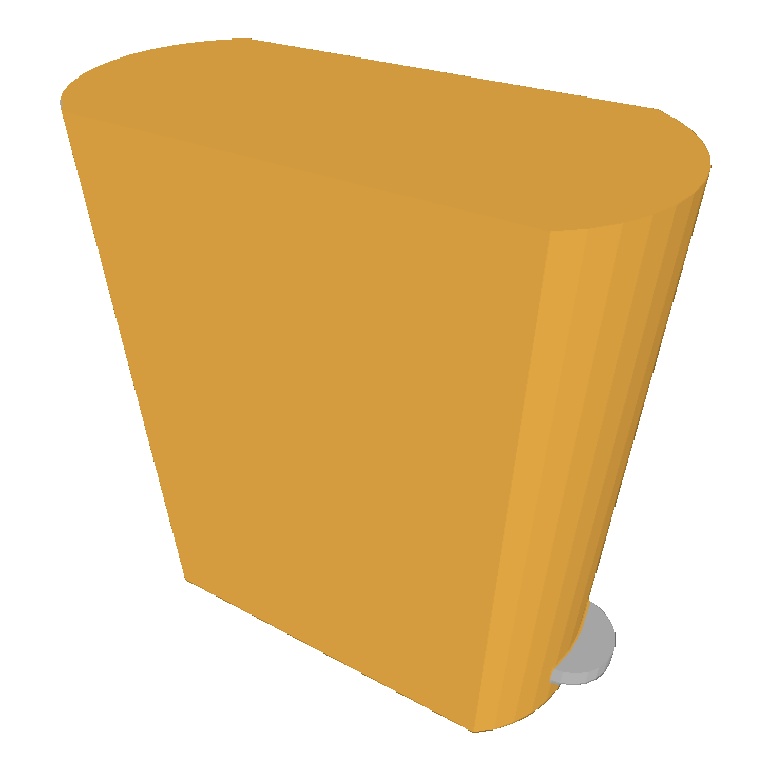} &
\includegraphics[width=0.1\textwidth]{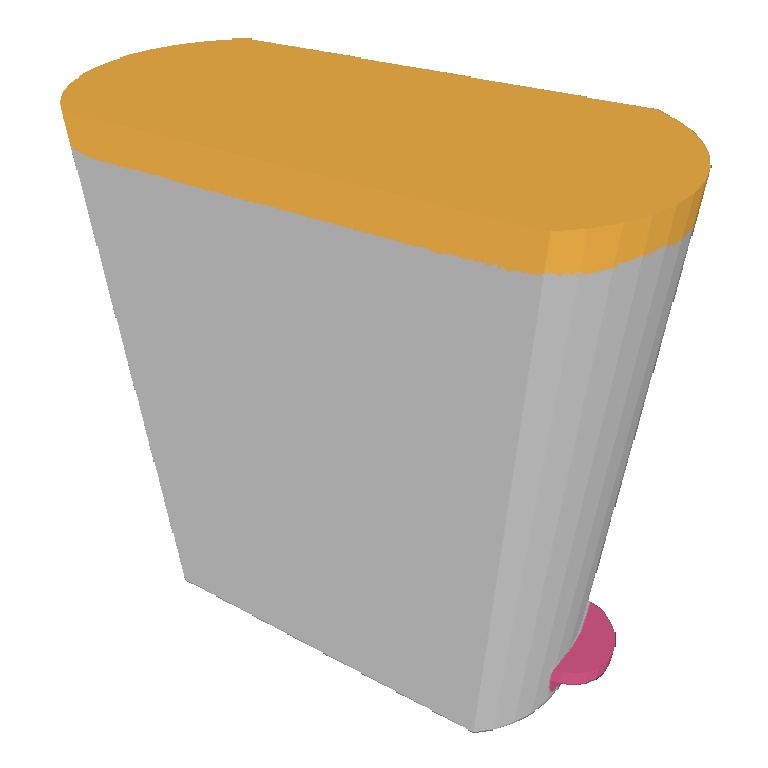} \\

\addlinespace[-2pt]
\arrayrulecolor{gray}\cmidrule(lr){1-5}
\arrayrulecolor{black}

\includegraphics[width=0.1\textwidth]{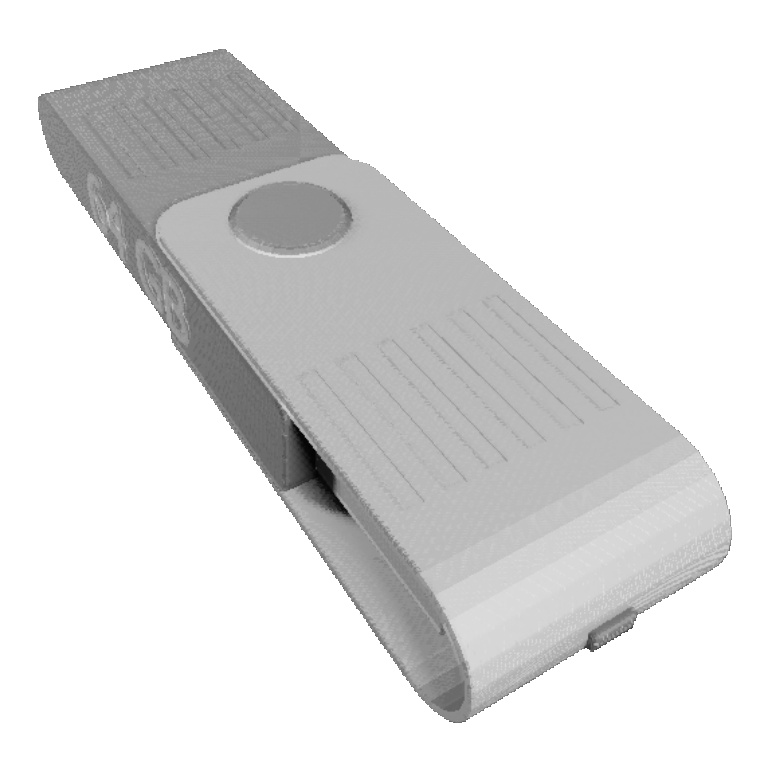} &
\includegraphics[width=0.1\textwidth]{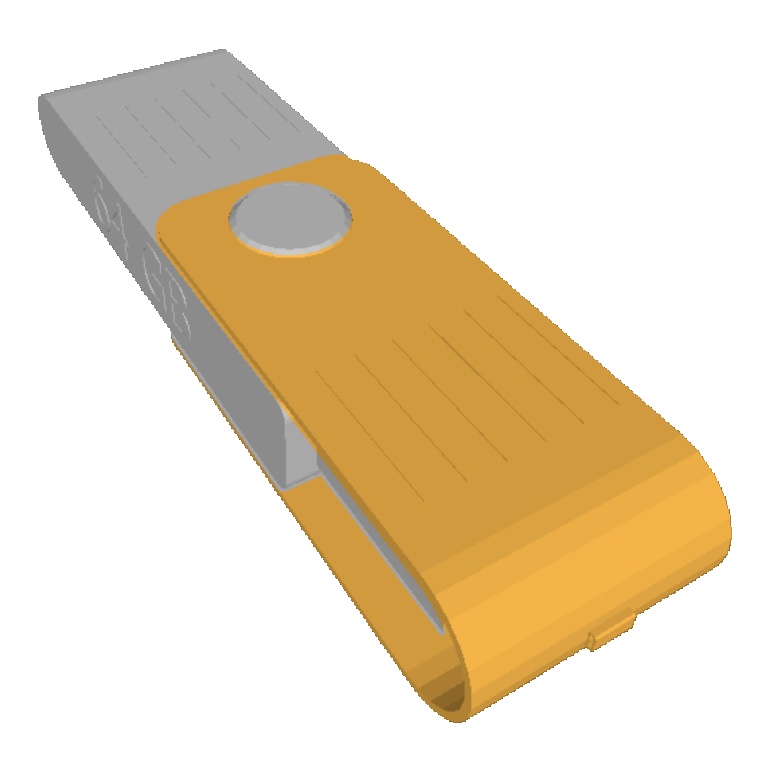} &
\includegraphics[width=0.1\textwidth]{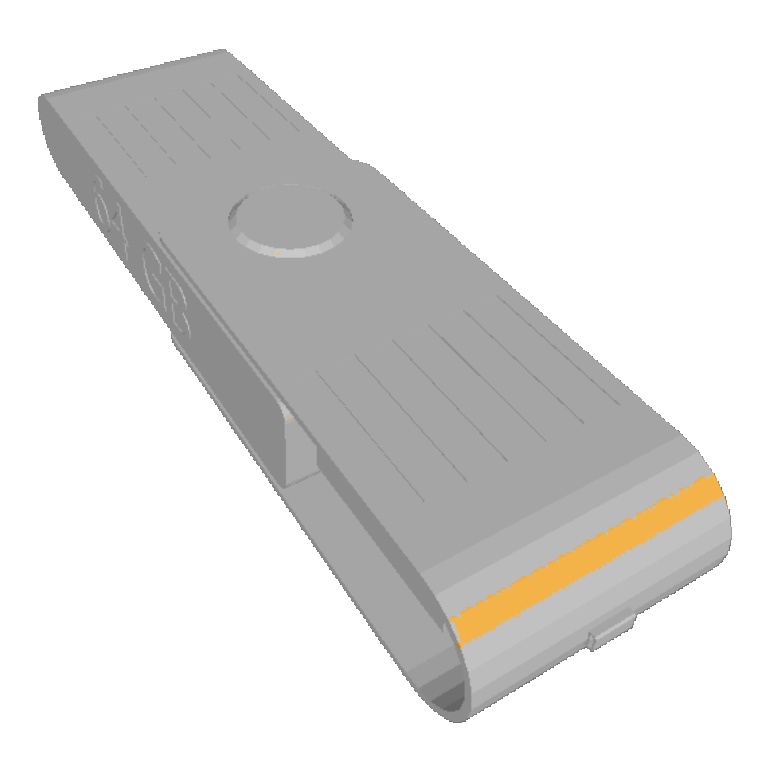} &
\includegraphics[width=0.1\textwidth]{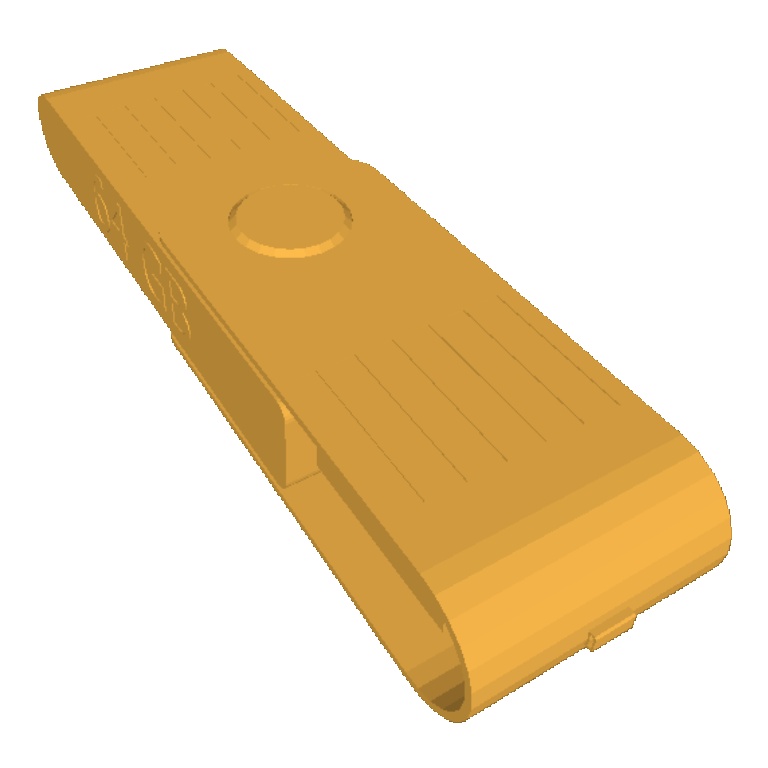} &
\includegraphics[width=0.1\textwidth]{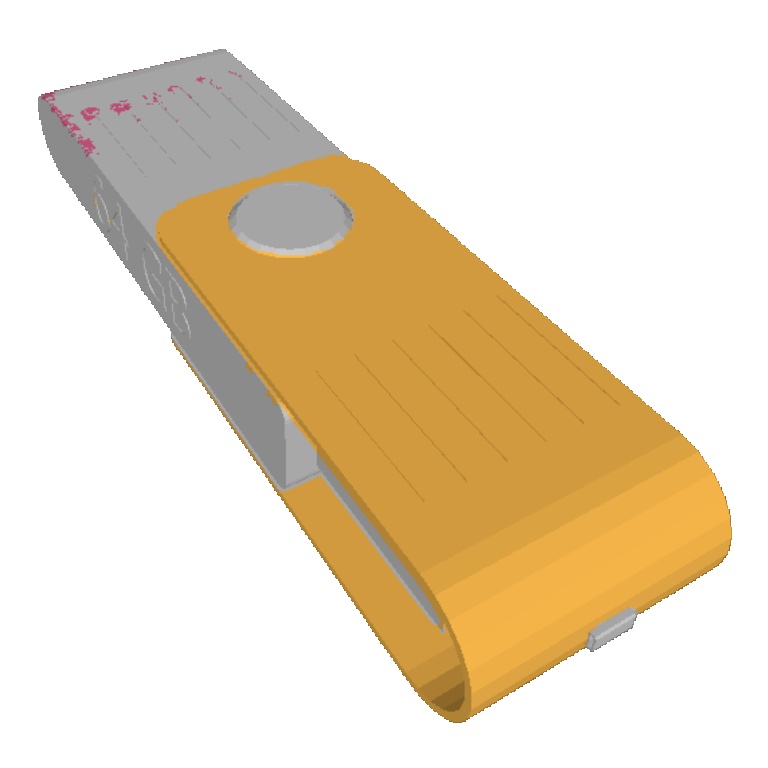} \\

\addlinespace[-2pt]
\bottomrule
\end{tabular}
}}
\end{tabular}
\end{table*}


\clearpage
\newpage
\bibliographystyle{ACM-Reference-Format}
\bibliography{main}


\begin{thebibliography}{55}


\ifx \showCODEN    \undefined \def \showCODEN     #1{\unskip}     \fi
\ifx \showISBNx    \undefined \def \showISBNx     #1{\unskip}     \fi
\ifx \showISBNxiii \undefined \def \showISBNxiii  #1{\unskip}     \fi
\ifx \showISSN     \undefined \def \showISSN      #1{\unskip}     \fi
\ifx \showLCCN     \undefined \def \showLCCN      #1{\unskip}     \fi
\ifx \shownote     \undefined \def \shownote      #1{#1}          \fi
\ifx \showarticletitle \undefined \def \showarticletitle #1{#1}   \fi
\ifx \showURL      \undefined \def \showURL       {\relax}        \fi
\providecommand\bibfield[2]{#2}
\providecommand\bibinfo[2]{#2}
\providecommand\natexlab[1]{#1}
\providecommand\showeprint[2][]{arXiv:#2}

\bibitem[Abdelreheem et~al\mbox{.}(2023)]%
        {abdelreheem2023satr}
\bibfield{author}{\bibinfo{person}{Ahmed Abdelreheem}, \bibinfo{person}{Ivan Skorokhodov}, \bibinfo{person}{Maks Ovsjanikov}, {and} \bibinfo{person}{Peter Wonka}.} \bibinfo{year}{2023}\natexlab{}.
\newblock \showarticletitle{SATR: Zero-Shot Semantic Segmentation of 3D Shapes}. In \bibinfo{booktitle}{\emph{Proceedings of the IEEE/CVF International Conference on Computer Vision}}. \bibinfo{pages}{15120--15133}.
\newblock
\href{https://doi.org/10.1109/ICCV51070.2023.01392}{doi:\nolinkurl{10.1109/ICCV51070.2023.01392}}


\bibitem[Alimohammadi et~al\mbox{.}(2025)]%
        {alimohammadi2024smitesegmenttime}
\bibfield{author}{\bibinfo{person}{Amirhossein Alimohammadi}, \bibinfo{person}{Sauradip Nag}, \bibinfo{person}{Saeid Asgari}, \bibinfo{person}{Andrea Tagliasacchi}, \bibinfo{person}{Ghassan Hamarneh}, {and} \bibinfo{person}{Ali~Mahdavi Amiri}.} \bibinfo{year}{2025}\natexlab{}.
\newblock \showarticletitle{SMITE: Segment Me In TimE}. In \bibinfo{booktitle}{\emph{International Conference on Learning Representations}}.
\newblock


\bibitem[Alliegro et~al\mbox{.}(2023)]%
        {alliegro2023polydiff}
\bibfield{author}{\bibinfo{person}{Antonio Alliegro}, \bibinfo{person}{Yawar Siddiqui}, \bibinfo{person}{Tatiana Tommasi}, {and} \bibinfo{person}{Matthias Nie{\ss}ner}.} \bibinfo{year}{2023}\natexlab{}.
\newblock \bibinfo{title}{PolyDiff: Generating 3D Polygonal Meshes with Diffusion Models}.
\newblock
\showeprint[arxiv]{2312.11417}~[cs.CV]


\bibitem[Avrahami et~al\mbox{.}(2023)]%
        {avrahami2023bas}
\bibfield{author}{\bibinfo{person}{Omri Avrahami}, \bibinfo{person}{Kfir Aberman}, \bibinfo{person}{Ohad Fried}, \bibinfo{person}{Daniel Cohen-Or}, {and} \bibinfo{person}{Dani Lischinski}.} \bibinfo{year}{2023}\natexlab{}.
\newblock \showarticletitle{Break-A-Scene: Extracting Multiple Concepts from a Single Image}. In \bibinfo{booktitle}{\emph{ACM SIGGRAPH Asia Conference Proceedings}}. \bibinfo{publisher}{ACM, New York, NY, USA}, Article \bibinfo{articleno}{96}.
\newblock
\href{https://doi.org/10.1145/3610548.3618154}{doi:\nolinkurl{10.1145/3610548.3618154}}


\bibitem[Banani et~al\mbox{.}(2024)]%
        {el2024probing}
\bibfield{author}{\bibinfo{person}{Mohamed~El Banani}, \bibinfo{person}{Amit Raj}, \bibinfo{person}{Kevis-Kokitsi Maninis}, \bibinfo{person}{Abhishek Kar}, \bibinfo{person}{Yuanzhen Li}, \bibinfo{person}{Michael Rubinstein}, \bibinfo{person}{Deqing Sun}, \bibinfo{person}{Leonidas Guibas}, \bibinfo{person}{Justin Johnson}, {and} \bibinfo{person}{Varun Jampani}.} \bibinfo{year}{2024}\natexlab{}.
\newblock \showarticletitle{Probing the 3D Awareness of Visual Foundation Models}. In \bibinfo{booktitle}{\emph{Proceedings of the IEEE/CVF Conference on Computer Vision and Pattern Recognition}}. \bibinfo{pages}{21795--21806}.
\newblock
\href{https://doi.org/10.1109/CVPR52733.2024.02059}{doi:\nolinkurl{10.1109/CVPR52733.2024.02059}}


\bibitem[Blattmann et~al\mbox{.}(2023)]%
        {blattmann2023stable}
\bibfield{author}{\bibinfo{person}{Andreas Blattmann}, \bibinfo{person}{Tim Dockhorn}, \bibinfo{person}{Sumith Kulal}, \bibinfo{person}{Daniel Mendelevitch}, \bibinfo{person}{Maciej Kilian}, \bibinfo{person}{Dominik Lorenz}, \bibinfo{person}{Yam Levi}, \bibinfo{person}{Zion English}, \bibinfo{person}{Vikram Voleti}, \bibinfo{person}{Adam Letts}, \bibinfo{person}{Varun Jampani}, {and} \bibinfo{person}{Robin Rombach}.} \bibinfo{year}{2023}\natexlab{}.
\newblock \bibinfo{title}{Stable Video Diffusion: Scaling Latent Video Diffusion Models to Large Datasets}.
\newblock
\showeprint[arxiv]{2311.15127}~[cs.CV]


\bibitem[Chen et~al\mbox{.}(2024)]%
        {chen2024training}
\bibfield{author}{\bibinfo{person}{Minghao Chen}, \bibinfo{person}{Iro Laina}, {and} \bibinfo{person}{Andrea Vedaldi}.} \bibinfo{year}{2024}\natexlab{}.
\newblock \showarticletitle{Training-Free Layout Control with Cross-Attention Guidance}. In \bibinfo{booktitle}{\emph{Proceedings of the IEEE/CVF Winter Conference on Applications of Computer Vision}}. \bibinfo{pages}{5331--5341}.
\newblock
\href{https://doi.org/10.1109/WACV57701.2024.00526}{doi:\nolinkurl{10.1109/WACV57701.2024.00526}}


\bibitem[Decatur et~al\mbox{.}(2024)]%
        {decatur2024paintbrush}
\bibfield{author}{\bibinfo{person}{Dale Decatur}, \bibinfo{person}{Itai Lang}, \bibinfo{person}{Kfir Aberman}, {and} \bibinfo{person}{Rana Hanocka}.} \bibinfo{year}{2024}\natexlab{}.
\newblock \showarticletitle{3D Paintbrush: Local Stylization of 3D Shapes with Cascaded Score Distillation}. In \bibinfo{booktitle}{\emph{Proceedings of the IEEE/CVF Conference on Computer Vision and Pattern Recognition}}. \bibinfo{pages}{4473--4483}.
\newblock


\bibitem[Decatur et~al\mbox{.}(2023)]%
        {decatur20233d}
\bibfield{author}{\bibinfo{person}{Dale Decatur}, \bibinfo{person}{Itai Lang}, {and} \bibinfo{person}{Rana Hanocka}.} \bibinfo{year}{2023}\natexlab{}.
\newblock \showarticletitle{3D Highlighter: Localizing Regions on 3D Shapes via Text Descriptions}. In \bibinfo{booktitle}{\emph{Proceedings of the IEEE/CVF Conference on Computer Vision and Pattern Recognition}}. \bibinfo{pages}{20930--20939}.
\newblock
\href{https://doi.org/10.1109/CVPR52729.2023.02005}{doi:\nolinkurl{10.1109/CVPR52729.2023.02005}}


\bibitem[Deitke et~al\mbox{.}(2023a)]%
        {deitke2023objaverse_xl}
\bibfield{author}{\bibinfo{person}{Matt Deitke}, \bibinfo{person}{Ruoshi Liu}, \bibinfo{person}{Matthew Wallingford}, \bibinfo{person}{Huong Ngo}, \bibinfo{person}{Oscar Michel}, \bibinfo{person}{Aditya Kusupati}, \bibinfo{person}{Alan Fan}, \bibinfo{person}{Christian Laforte}, \bibinfo{person}{Vikram Voleti}, \bibinfo{person}{Samir~Yitzhak Gadre}, \bibinfo{person}{Eli VanderBilt}, \bibinfo{person}{Aniruddha Kembhavi}, \bibinfo{person}{Carl Vondrick}, \bibinfo{person}{Georgia Gkioxari}, \bibinfo{person}{Kiana Ehsani}, \bibinfo{person}{Ludwig Schmidt}, {and} \bibinfo{person}{Ali Farhadi}.} \bibinfo{year}{2023}\natexlab{a}.
\newblock \showarticletitle{Objaverse-XL: A Universe of 10M+ 3D Objects}.
\newblock \bibinfo{journal}{\emph{Advances in Neural Information Processing Systems}}  \bibinfo{volume}{36} (\bibinfo{year}{2023}), \bibinfo{pages}{35799--35813}.
\newblock


\bibitem[Deitke et~al\mbox{.}(2023b)]%
        {deitke2023objaverse}
\bibfield{author}{\bibinfo{person}{Matt Deitke}, \bibinfo{person}{Dustin Schwenk}, \bibinfo{person}{Jordi Salvador}, \bibinfo{person}{Luca Weihs}, \bibinfo{person}{Oscar Michel}, \bibinfo{person}{Eli VanderBilt}, \bibinfo{person}{Ludwig Schmidt}, \bibinfo{person}{Kiana Ehsanit}, \bibinfo{person}{Aniruddha Kembhavi}, {and} \bibinfo{person}{Ali Farhadi}.} \bibinfo{year}{2023}\natexlab{b}.
\newblock \showarticletitle{Objaverse: A Universe of Annotated 3D Objects}. In \bibinfo{booktitle}{\emph{Proceedings of the IEEE/CVF Conference on Computer Vision and Pattern Recognition}}. \bibinfo{pages}{13142--13153}.
\newblock
\href{https://doi.org/10.1109/CVPR52729.2023.01263}{doi:\nolinkurl{10.1109/CVPR52729.2023.01263}}


\bibitem[Gal et~al\mbox{.}(2023)]%
        {gal2022image}
\bibfield{author}{\bibinfo{person}{Rinon Gal}, \bibinfo{person}{Yuval Alaluf}, \bibinfo{person}{Yuval Atzmon}, \bibinfo{person}{Or Patashnik}, \bibinfo{person}{Amit~Haim Bermano}, \bibinfo{person}{Gal Chechik}, {and} \bibinfo{person}{Daniel Cohen-or}.} \bibinfo{year}{2023}\natexlab{}.
\newblock \showarticletitle{An Image is Worth One Word: Personalizing Text-to-Image Generation using Textual Inversion}. In \bibinfo{booktitle}{\emph{International Conference on Learning Representations}}.
\newblock


\bibitem[Hanocka et~al\mbox{.}(2019)]%
        {hanocka2019meshcnn}
\bibfield{author}{\bibinfo{person}{Rana Hanocka}, \bibinfo{person}{Amir Hertz}, \bibinfo{person}{Noa Fish}, \bibinfo{person}{Raja Giryes}, \bibinfo{person}{Shachar Fleishman}, {and} \bibinfo{person}{Daniel Cohen-Or}.} \bibinfo{year}{2019}\natexlab{}.
\newblock \showarticletitle{MeshCNN: A Network with an Edge}.
\newblock \bibinfo{journal}{\emph{ACM Transactions on Graphics}} \bibinfo{volume}{38}, \bibinfo{number}{4}, Article \bibinfo{articleno}{90} (\bibinfo{year}{2019}).
\newblock
\href{https://doi.org/10.1145/3306346.3322959}{doi:\nolinkurl{10.1145/3306346.3322959}}


\bibitem[He et~al\mbox{.}(2025)]%
        {he2024deeplearningbased3d}
\bibfield{author}{\bibinfo{person}{Yong He}, \bibinfo{person}{Hongshan Yu}, \bibinfo{person}{Xiaoyan Liu}, \bibinfo{person}{Zhengeng Yang}, \bibinfo{person}{Wei Sun}, \bibinfo{person}{Saeed Anwar}, {and} \bibinfo{person}{Ajmal Mian}.} \bibinfo{year}{2025}\natexlab{}.
\newblock \bibinfo{title}{Deep Learning based 3D Segmentation in Computer Vision: A Survey}.
\newblock
\href{https://doi.org/10.1016/j.inffus.2024.102722}{doi:\nolinkurl{10.1016/j.inffus.2024.102722}}


\bibitem[Hedlin et~al\mbox{.}(2024)]%
        {hedlin2024unsupervised}
\bibfield{author}{\bibinfo{person}{Eric Hedlin}, \bibinfo{person}{Gopal Sharma}, \bibinfo{person}{Shweta Mahajan}, \bibinfo{person}{Hossam Isack}, \bibinfo{person}{Abhishek Kar}, \bibinfo{person}{Andrea Tagliasacchi}, {and} \bibinfo{person}{Kwang~Moo Yi}.} \bibinfo{year}{2024}\natexlab{}.
\newblock \showarticletitle{Unsupervised Semantic Correspondence Using Stable Diffusion}.
\newblock \bibinfo{journal}{\emph{Advances in Neural Information Processing Systems}}  \bibinfo{volume}{36} (\bibinfo{year}{2024}), \bibinfo{pages}{8266--8279}.
\newblock


\bibitem[Ho et~al\mbox{.}(2020)]%
        {ho2020denoising}
\bibfield{author}{\bibinfo{person}{Jonathan Ho}, \bibinfo{person}{Ajay Jain}, {and} \bibinfo{person}{Pieter Abbeel}.} \bibinfo{year}{2020}\natexlab{}.
\newblock \showarticletitle{Denoising Diffusion Probabilistic Models}.
\newblock \bibinfo{journal}{\emph{Advances in Neural Information Processing Systems}}  \bibinfo{volume}{33} (\bibinfo{year}{2020}), \bibinfo{pages}{6840--6851}.
\newblock


\bibitem[Hu et~al\mbox{.}(2022)]%
        {hu2021lora}
\bibfield{author}{\bibinfo{person}{Edward~J. Hu}, \bibinfo{person}{Yelong Shen}, \bibinfo{person}{Phillip Wallis}, \bibinfo{person}{Zeyuan Allen-Zhu}, \bibinfo{person}{Yuanzhi Li}, \bibinfo{person}{Shean Wang}, \bibinfo{person}{Lu Wang}, {and} \bibinfo{person}{Weizhu Chen}.} \bibinfo{year}{2022}\natexlab{}.
\newblock \showarticletitle{Lo{RA}: Low-Rank Adaptation of Large Language Models}. In \bibinfo{booktitle}{\emph{International Conference on Learning Representations}}.
\newblock


\bibitem[Imagen-Team-Google et~al\mbox{.}(2024)]%
        {baldridge2024imagen}
\bibfield{author}{\bibinfo{person}{Imagen-Team-Google}, \bibinfo{person}{:}, \bibinfo{person}{Jason Baldridge}, \bibinfo{person}{Jakob Bauer}, \bibinfo{person}{Mukul Bhutani}, \bibinfo{person}{Nicole Brichtova}, \bibinfo{person}{Andrew Bunner}, \bibinfo{person}{Lluis Castrejon}, \bibinfo{person}{Kelvin Chan}, \bibinfo{person}{Yichang Chen}, \bibinfo{person}{Sander Dieleman}, \bibinfo{person}{Yuqing Du}, \bibinfo{person}{Zach Eaton-Rosen}, {et~al\mbox{.}}} \bibinfo{year}{2024}\natexlab{}.
\newblock \bibinfo{title}{Imagen 3}.
\newblock
\showeprint[arxiv]{2408.07009}~[cs.CV]


\bibitem[Jones et~al\mbox{.}(2022)]%
        {jones2022SHRED}
\bibfield{author}{\bibinfo{person}{R.~Kenny Jones}, \bibinfo{person}{Aalia Habib}, {and} \bibinfo{person}{Daniel Ritchie}.} \bibinfo{year}{2022}\natexlab{}.
\newblock \showarticletitle{SHRED: 3D Shape Region Decomposition with Learned Local Operations}.
\newblock \bibinfo{journal}{\emph{ACM Transactions on Graphics}} \bibinfo{volume}{41}, \bibinfo{number}{6}, Article \bibinfo{articleno}{186} (\bibinfo{year}{2022}).
\newblock
\href{https://doi.org/10.1145/3550454.3555440}{doi:\nolinkurl{10.1145/3550454.3555440}}


\bibitem[Jun and Nichol(2023)]%
        {jun2023shap}
\bibfield{author}{\bibinfo{person}{Heewoo Jun} {and} \bibinfo{person}{Alex Nichol}.} \bibinfo{year}{2023}\natexlab{}.
\newblock \bibinfo{title}{Shap-E: Generating Conditional 3D Implicit Functions}.
\newblock
\showeprint[arxiv]{2305.02463}~[cs.CV]


\bibitem[Katz and Tal(2003)]%
        {katz2003hierarchical}
\bibfield{author}{\bibinfo{person}{Sagi Katz} {and} \bibinfo{person}{Ayellet Tal}.} \bibinfo{year}{2003}\natexlab{}.
\newblock \showarticletitle{Hierarchical Mesh Decomposition using Fuzzy Clustering and Cuts}.
\newblock \bibinfo{journal}{\emph{ACM Transactions on Graphics}} \bibinfo{volume}{22}, \bibinfo{number}{3} (\bibinfo{year}{2003}), \bibinfo{pages}{954--961}.
\newblock
\href{https://doi.org/10.1145/882262.882369}{doi:\nolinkurl{10.1145/882262.882369}}


\bibitem[Khani et~al\mbox{.}(2024)]%
        {khani2023slime}
\bibfield{author}{\bibinfo{person}{Aliasghar Khani}, \bibinfo{person}{Saeid Asgari}, \bibinfo{person}{Aditya Sanghi}, \bibinfo{person}{Ali Mahdavi-Amiri}, {and} \bibinfo{person}{Ghassan Hamarneh}.} \bibinfo{year}{2024}\natexlab{}.
\newblock \showarticletitle{SLiMe: Segment Like Me}. In \bibinfo{booktitle}{\emph{International Conference on Learning Representations}}.
\newblock


\bibitem[Kim and Sung(2024)]%
        {kim2024partstad}
\bibfield{author}{\bibinfo{person}{Hyunjin Kim} {and} \bibinfo{person}{Minhyuk Sung}.} \bibinfo{year}{2024}\natexlab{}.
\newblock \showarticletitle{PartSTAD: 2D-to-3D Part Segmentation Task Adaptation}. In \bibinfo{booktitle}{\emph{European Conference on Computer Vision}}. \bibinfo{publisher}{Springer}, \bibinfo{pages}{422--439}.
\newblock
\href{https://doi.org/10.1007/978-3-031-72652-1_25}{doi:\nolinkurl{10.1007/978-3-031-72652-1_25}}


\bibitem[Kingma and Welling(2014)]%
        {kingma2014vae}
\bibfield{author}{\bibinfo{person}{Diederik~P. Kingma} {and} \bibinfo{person}{Max Welling}.} \bibinfo{year}{2014}\natexlab{}.
\newblock \showarticletitle{Auto-Encoding Variational Bayes}. In \bibinfo{booktitle}{\emph{International Conference on Learning Representations}}.
\newblock


\bibitem[Kirillov et~al\mbox{.}(2023)]%
        {SAM_SegmentAnything}
\bibfield{author}{\bibinfo{person}{Alexander Kirillov}, \bibinfo{person}{Eric Mintun}, \bibinfo{person}{Nikhila Ravi}, \bibinfo{person}{Hanzi Mao}, \bibinfo{person}{Chloe Rolland}, \bibinfo{person}{Laura Gustafson}, \bibinfo{person}{Tete Xiao}, \bibinfo{person}{Spencer Whitehead}, \bibinfo{person}{Alexander~C. Berg}, \bibinfo{person}{Wan-Yen Lo}, \bibinfo{person}{Piotr Dollár}, {and} \bibinfo{person}{Ross Girshick}.} \bibinfo{year}{2023}\natexlab{}.
\newblock \showarticletitle{Segment Anything}. In \bibinfo{booktitle}{\emph{Proceedings of the IEEE/CVF International Conference on Computer Vision}}. \bibinfo{pages}{3992--4003}.
\newblock
\href{https://doi.org/10.1109/ICCV51070.2023.00371}{doi:\nolinkurl{10.1109/ICCV51070.2023.00371}}


\bibitem[Landrieu and Simonovsky(2018)]%
        {landrieu2018large}
\bibfield{author}{\bibinfo{person}{Loic Landrieu} {and} \bibinfo{person}{Martin Simonovsky}.} \bibinfo{year}{2018}\natexlab{}.
\newblock \showarticletitle{Large-Scale Point Cloud Semantic Segmentation with Superpoint Graphs}. In \bibinfo{booktitle}{\emph{Proceedings of the IEEE/CVF Conference on Computer Vision and Pattern Recognition}}. \bibinfo{pages}{4558--4567}.
\newblock
\href{https://doi.org/10.1109/CVPR.2018.00479}{doi:\nolinkurl{10.1109/CVPR.2018.00479}}


\bibitem[Lang et~al\mbox{.}(2024)]%
        {lang2024iseg}
\bibfield{author}{\bibinfo{person}{Itai Lang}, \bibinfo{person}{Fei Xu}, \bibinfo{person}{Dale Decatur}, \bibinfo{person}{Sudarshan Babu}, {and} \bibinfo{person}{Rana Hanocka}.} \bibinfo{year}{2024}\natexlab{}.
\newblock \showarticletitle{{iSeg: Interactive 3D Segmentation via Interactive Attention}}. In \bibinfo{booktitle}{\emph{ACM SIGGRAPH Asia Conference Papers}}. \bibinfo{publisher}{ACM, New York, NY, USA}.
\newblock
\href{https://doi.org/10.1145/3680528.3687605}{doi:\nolinkurl{10.1145/3680528.3687605}}


\bibitem[Li et~al\mbox{.}(2022)]%
        {li2022grounded}
\bibfield{author}{\bibinfo{person}{Liunian~Harold Li}, \bibinfo{person}{Pengchuan Zhang}, \bibinfo{person}{Haotian Zhang}, \bibinfo{person}{Jianwei Yang}, \bibinfo{person}{Chunyuan Li}, \bibinfo{person}{Yiwu Zhong}, \bibinfo{person}{Lijuan Wang}, \bibinfo{person}{Lu Yuan}, \bibinfo{person}{Lei Zhang}, \bibinfo{person}{Jenq-Neng Hwang}, \bibinfo{person}{Kai-Wei Chang}, {and} \bibinfo{person}{Jianfeng Gao}.} \bibinfo{year}{2022}\natexlab{}.
\newblock \showarticletitle{Grounded Language-Image Pre-training}. In \bibinfo{booktitle}{\emph{Proceedings of the IEEE/CVF Conference on Computer Vision and Pattern Recognition}}. \bibinfo{pages}{10955--10965}.
\newblock
\href{https://doi.org/10.1109/CVPR52688.2022.01069}{doi:\nolinkurl{10.1109/CVPR52688.2022.01069}}


\bibitem[Liu et~al\mbox{.}(2023)]%
        {liu2023partslip}
\bibfield{author}{\bibinfo{person}{Minghua Liu}, \bibinfo{person}{Yinhao Zhu}, \bibinfo{person}{Hong Cai}, \bibinfo{person}{Shizhong Han}, \bibinfo{person}{Zhan Ling}, \bibinfo{person}{Fatih Porikli}, {and} \bibinfo{person}{Hao Su}.} \bibinfo{year}{2023}\natexlab{}.
\newblock \showarticletitle{PartSLIP: Low-Shot Part Segmentation for 3D Point Clouds via Pretrained Image-Language Models}. In \bibinfo{booktitle}{\emph{Proceedings of the IEEE/CVF Conference on Computer Vision and Pattern Recognition}}. \bibinfo{pages}{21736--21746}.
\newblock
\href{https://doi.org/10.1109/CVPR52729.2023.02082}{doi:\nolinkurl{10.1109/CVPR52729.2023.02082}}


\bibitem[Mo et~al\mbox{.}(2019)]%
        {mo2019partnet}
\bibfield{author}{\bibinfo{person}{Kaichun Mo}, \bibinfo{person}{Shilin Zhu}, \bibinfo{person}{Angel~X. Chang}, \bibinfo{person}{Li Yi}, \bibinfo{person}{Subarna Tripathi}, \bibinfo{person}{Leonidas~J. Guibas}, {and} \bibinfo{person}{Hao Su}.} \bibinfo{year}{2019}\natexlab{}.
\newblock \showarticletitle{PartNet: A Large-Scale Benchmark for Fine-Grained and Hierarchical Part-Level 3D Object Understanding}. In \bibinfo{booktitle}{\emph{Proceedings of the IEEE/CVF Conference on Computer Vision and Pattern Recognition}}. \bibinfo{pages}{909--918}.
\newblock
\href{https://doi.org/10.1109/CVPR.2019.00100}{doi:\nolinkurl{10.1109/CVPR.2019.00100}}


\bibitem[Namekata et~al\mbox{.}(2024)]%
        {namekata2024emerdiff}
\bibfield{author}{\bibinfo{person}{Koichi Namekata}, \bibinfo{person}{Amirmojtaba Sabour}, \bibinfo{person}{Sanja Fidler}, {and} \bibinfo{person}{Seung~Wook Kim}.} \bibinfo{year}{2024}\natexlab{}.
\newblock \showarticletitle{EmerDiff: Emerging Pixel-level Semantic Knowledge in Diffusion Models}. In \bibinfo{booktitle}{\emph{International Conference on Learning Representations}}.
\newblock


\bibitem[Perla et~al\mbox{.}(2024)]%
        {perla2024easitex}
\bibfield{author}{\bibinfo{person}{Sai Raj~Kishore Perla}, \bibinfo{person}{Yizhi Wang}, \bibinfo{person}{Ali Mahdavi-Amiri}, {and} \bibinfo{person}{Hao Zhang}.} \bibinfo{year}{2024}\natexlab{}.
\newblock \showarticletitle{EASI-Tex: Edge-Aware Mesh Texturing from Single Image}.
\newblock \bibinfo{journal}{\emph{ACM Transactions on Graphics (Proceedings of SIGGRAPH)}}  \bibinfo{volume}{43}, Article \bibinfo{articleno}{40} (\bibinfo{year}{2024}).
\newblock
\href{https://doi.org/10.1145/3658222}{doi:\nolinkurl{10.1145/3658222}}


\bibitem[Podolak et~al\mbox{.}(2006)]%
        {podolak2006planar}
\bibfield{author}{\bibinfo{person}{Joshua Podolak}, \bibinfo{person}{Philip Shilane}, \bibinfo{person}{Aleksey Golovinskiy}, \bibinfo{person}{Szymon Rusinkiewicz}, {and} \bibinfo{person}{Thomas Funkhouser}.} \bibinfo{year}{2006}\natexlab{}.
\newblock \showarticletitle{A Planar-Reflective Symmetry Transform for 3D Shapes}.
\newblock \bibinfo{journal}{\emph{ACM Transactions on Graphics}} \bibinfo{volume}{25}, \bibinfo{number}{3} (\bibinfo{year}{2006}), \bibinfo{pages}{549–559}.
\newblock
\href{https://doi.org/10.1145/1141911.1141923}{doi:\nolinkurl{10.1145/1141911.1141923}}


\bibitem[Qi et~al\mbox{.}(2017a)]%
        {qi2017pointnet}
\bibfield{author}{\bibinfo{person}{Charles~R. Qi}, \bibinfo{person}{Hao Su}, \bibinfo{person}{Mo Kaichun}, {and} \bibinfo{person}{Leonidas~J. Guibas}.} \bibinfo{year}{2017}\natexlab{a}.
\newblock \showarticletitle{PointNet: Deep Learning on Point Sets for 3D Classification and Segmentation}. In \bibinfo{booktitle}{\emph{Proceedings of the IEEE/CVF Conference on Computer Vision and Pattern Recognition}}. \bibinfo{pages}{77--85}.
\newblock
\href{https://doi.org/10.1109/CVPR.2017.16}{doi:\nolinkurl{10.1109/CVPR.2017.16}}


\bibitem[Qi et~al\mbox{.}(2017b)]%
        {qi2017pointnet++}
\bibfield{author}{\bibinfo{person}{Charles~Ruizhongtai Qi}, \bibinfo{person}{Li Yi}, \bibinfo{person}{Hao Su}, {and} \bibinfo{person}{Leonidas~J. Guibas}.} \bibinfo{year}{2017}\natexlab{b}.
\newblock \showarticletitle{PointNet++: Deep Hierarchical Feature Learning on Point Sets in a Metric Space}.
\newblock \bibinfo{journal}{\emph{Advances in Neural Information Processing Systems}}  \bibinfo{volume}{30} (\bibinfo{year}{2017}).
\newblock


\bibitem[Radford et~al\mbox{.}(2021)]%
        {radford2021learning}
\bibfield{author}{\bibinfo{person}{Alec Radford}, \bibinfo{person}{Jong~Wook Kim}, \bibinfo{person}{Chris Hallacy}, \bibinfo{person}{Aditya Ramesh}, \bibinfo{person}{Gabriel Goh}, \bibinfo{person}{Sandhini Agarwal}, \bibinfo{person}{Girish Sastry}, \bibinfo{person}{Amanda Askell}, \bibinfo{person}{Pamela Mishkin}, \bibinfo{person}{Jack Clark}, \bibinfo{person}{Gretchen Krueger}, {and} \bibinfo{person}{Ilya Sutskever}.} \bibinfo{year}{2021}\natexlab{}.
\newblock \showarticletitle{Learning Transferable Visual Models From Natural Language Supervision}. In \bibinfo{booktitle}{\emph{Proceedings of the International Conference on Machine Learning}}, Vol.~\bibinfo{volume}{139}. \bibinfo{publisher}{PMLR}, \bibinfo{pages}{8748--8763}.
\newblock


\bibitem[Ravi et~al\mbox{.}(2024)]%
        {ravi2024sam2}
\bibfield{author}{\bibinfo{person}{Nikhila Ravi}, \bibinfo{person}{Valentin Gabeur}, \bibinfo{person}{Yuan-Ting Hu}, \bibinfo{person}{Ronghang Hu}, \bibinfo{person}{Chaitanya Ryali}, \bibinfo{person}{Tengyu Ma}, \bibinfo{person}{Haitham Khedr}, \bibinfo{person}{Roman R{\"a}dle}, \bibinfo{person}{Chloe Rolland}, \bibinfo{person}{Laura Gustafson}, \bibinfo{person}{Eric Mintun}, \bibinfo{person}{Junting Pan}, \bibinfo{person}{Kalyan~Vasudev Alwala}, \bibinfo{person}{Nicolas Carion}, \bibinfo{person}{Chao-Yuan Wu}, \bibinfo{person}{Ross Girshick}, \bibinfo{person}{Piotr Doll{\'a}r}, {and} \bibinfo{person}{Christoph Feichtenhofer}.} \bibinfo{year}{2024}\natexlab{}.
\newblock \bibinfo{title}{SAM 2: Segment Anything in Images and Videos}.
\newblock
\showeprint[arxiv]{2408.00714}~[cs.CV]


\bibitem[Ravi et~al\mbox{.}(2020)]%
        {ravi2020pytorch3d}
\bibfield{author}{\bibinfo{person}{Nikhila Ravi}, \bibinfo{person}{Jeremy Reizenstein}, \bibinfo{person}{David Novotny}, \bibinfo{person}{Taylor Gordon}, \bibinfo{person}{Wan-Yen Lo}, \bibinfo{person}{Justin Johnson}, {and} \bibinfo{person}{Georgia Gkioxari}.} \bibinfo{year}{2020}\natexlab{}.
\newblock \bibinfo{title}{Accelerating 3D Deep Learning with PyTorch3D}.
\newblock \bibinfo{howpublished}{arXiv:2007.08501}.
\newblock
\href{https://doi.org/10.48550/arXiv.2007.08501}{doi:\nolinkurl{10.48550/arXiv.2007.08501}}


\bibitem[Richardson et~al\mbox{.}(2023)]%
        {richardson2023texture}
\bibfield{author}{\bibinfo{person}{Elad Richardson}, \bibinfo{person}{Gal Metzer}, \bibinfo{person}{Yuval Alaluf}, \bibinfo{person}{Raja Giryes}, {and} \bibinfo{person}{Daniel Cohen-Or}.} \bibinfo{year}{2023}\natexlab{}.
\newblock \showarticletitle{TEXTure: Text-Guided Texturing of 3D Shapes}. In \bibinfo{booktitle}{\emph{ACM SIGGRAPH Conference Proceedings}}. \bibinfo{publisher}{ACM, New York, NY, USA}, Article \bibinfo{articleno}{54}.
\newblock
\href{https://doi.org/10.1145/3588432.3591503}{doi:\nolinkurl{10.1145/3588432.3591503}}


\bibitem[Rombach et~al\mbox{.}(2022)]%
        {rombach2021highresolution}
\bibfield{author}{\bibinfo{person}{Robin Rombach}, \bibinfo{person}{Andreas Blattmann}, \bibinfo{person}{Dominik Lorenz}, \bibinfo{person}{Patrick Esser}, {and} \bibinfo{person}{Björn Ommer}.} \bibinfo{year}{2022}\natexlab{}.
\newblock \bibinfo{title}{High-Resolution Image Synthesis with Latent Diffusion Models}.
\newblock \bibinfo{numpages}{10674-10685}~pages.
\newblock
\href{https://doi.org/10.1109/CVPR52688.2022.01042}{doi:\nolinkurl{10.1109/CVPR52688.2022.01042}}


\bibitem[Ronneberger et~al\mbox{.}(2015)]%
        {ronneberger2015u}
\bibfield{author}{\bibinfo{person}{Olaf Ronneberger}, \bibinfo{person}{Philipp Fischer}, {and} \bibinfo{person}{Thomas Brox}.} \bibinfo{year}{2015}\natexlab{}.
\newblock \showarticletitle{U-Net: Convolutional Networks for Biomedical Image Segmentation}. In \bibinfo{booktitle}{\emph{Medical Image Computing and Computer-Assisted Intervention}}. Springer, \bibinfo{pages}{234--241}.
\newblock
\href{https://doi.org/10.1007/978-3-319-24574-4_28}{doi:\nolinkurl{10.1007/978-3-319-24574-4_28}}


\bibitem[Shamir(2006)]%
        {shamir2006segmentation}
\bibfield{author}{\bibinfo{person}{Ariel Shamir}.} \bibinfo{year}{2006}\natexlab{}.
\newblock \showarticletitle{Segmentation and Shape Extraction of 3D Boundary Meshes}. In \bibinfo{booktitle}{\emph{Eurographics - State of the Art Reports}}. \bibinfo{pages}{137--149}.
\newblock
\href{https://doi.org/10.2312/egst.20061056}{doi:\nolinkurl{10.2312/egst.20061056}}


\bibitem[Sohl-Dickstein et~al\mbox{.}(2015)]%
        {sohl2015deep}
\bibfield{author}{\bibinfo{person}{Jascha Sohl-Dickstein}, \bibinfo{person}{Eric Weiss}, \bibinfo{person}{Niru Maheswaranathan}, {and} \bibinfo{person}{Surya Ganguli}.} \bibinfo{year}{2015}\natexlab{}.
\newblock \showarticletitle{Deep Unsupervised Learning using Nonequilibrium Thermodynamics}. In \bibinfo{booktitle}{\emph{Proceedings of the International Conference on Machine Learning}}, Vol.~\bibinfo{volume}{37}. \bibinfo{publisher}{PMLR}, \bibinfo{pages}{2256--2265}.
\newblock


\bibitem[Song and Ermon(2019)]%
        {song2019generative}
\bibfield{author}{\bibinfo{person}{Yang Song} {and} \bibinfo{person}{Stefano Ermon}.} \bibinfo{year}{2019}\natexlab{}.
\newblock \showarticletitle{Generative Modeling by Estimating Gradients of the Data Distribution}.
\newblock \bibinfo{journal}{\emph{Advances in Neural Information Processing Systems}}  \bibinfo{volume}{32} (\bibinfo{year}{2019}).
\newblock


\bibitem[Tang et~al\mbox{.}(2023)]%
        {tang2023emergent}
\bibfield{author}{\bibinfo{person}{Luming Tang}, \bibinfo{person}{Menglin Jia}, \bibinfo{person}{Qianqian Wang}, \bibinfo{person}{Cheng~Perng Phoo}, {and} \bibinfo{person}{Bharath Hariharan}.} \bibinfo{year}{2023}\natexlab{}.
\newblock \showarticletitle{Emergent Correspondence from Image Diffusion}.
\newblock \bibinfo{journal}{\emph{Advances in Neural Information Processing Systems}}  \bibinfo{volume}{36} (\bibinfo{year}{2023}), \bibinfo{pages}{1363--1389}.
\newblock


\bibitem[Thai et~al\mbox{.}(2024)]%
        {3by2}
\bibfield{author}{\bibinfo{person}{Anh Thai}, \bibinfo{person}{Weiyao Wang}, \bibinfo{person}{Hao Tang}, \bibinfo{person}{Stefan Stojanov}, \bibinfo{person}{James~M. Rehg}, {and} \bibinfo{person}{Matt Feiszli}.} \bibinfo{year}{2024}\natexlab{}.
\newblock \showarticletitle{$3 \times 2$: 3D Object Part Segmentation by 2D Semantic Correspondences}. In \bibinfo{booktitle}{\emph{European Conference on Computer Vision}}. \bibinfo{publisher}{Springer}, \bibinfo{pages}{149--166}.
\newblock
\href{https://doi.org/10.1007/978-3-031-72920-1_9}{doi:\nolinkurl{10.1007/978-3-031-72920-1_9}}


\bibitem[Umam et~al\mbox{.}(2024)]%
        {umam2024partdistill}
\bibfield{author}{\bibinfo{person}{Ardian Umam}, \bibinfo{person}{Cheng-Kun Yang}, \bibinfo{person}{Min-Hung Chen}, \bibinfo{person}{Jen-Hui Chuang}, {and} \bibinfo{person}{Yen-Yu Lin}.} \bibinfo{year}{2024}\natexlab{}.
\newblock \showarticletitle{PartDistill: 3D Shape Part Segmentation by Vision-Language Model Distillation}. In \bibinfo{booktitle}{\emph{Proceedings of the IEEE/CVF Conference on Computer Vision and Pattern Recognition}}. \bibinfo{pages}{3470--3479}.
\newblock
\href{https://doi.org/10.1109/CVPR52733.2024.00333}{doi:\nolinkurl{10.1109/CVPR52733.2024.00333}}


\bibitem[Wang et~al\mbox{.}(2013)]%
        {wang_projective}
\bibfield{author}{\bibinfo{person}{Yunhai Wang}, \bibinfo{person}{Minglun Gong}, \bibinfo{person}{Tianhua Wang}, \bibinfo{person}{Daniel Cohen-Or}, \bibinfo{person}{Hao Zhang}, {and} \bibinfo{person}{Baoquan Chen}.} \bibinfo{year}{2013}\natexlab{}.
\newblock \showarticletitle{Projective Analysis for 3D Shape Segmentation}.
\newblock \bibinfo{journal}{\emph{ACM Transactions on Graphics}} \bibinfo{volume}{32}, \bibinfo{number}{6}, Article \bibinfo{articleno}{192} (\bibinfo{year}{2013}).
\newblock
\href{https://doi.org/10.1145/2508363.2508393}{doi:\nolinkurl{10.1145/2508363.2508393}}


\bibitem[Xiang et~al\mbox{.}(2020)]%
        {Xiang_2020_SAPIEN}
\bibfield{author}{\bibinfo{person}{Fanbo Xiang}, \bibinfo{person}{Yuzhe Qin}, \bibinfo{person}{Kaichun Mo}, \bibinfo{person}{Yikuan Xia}, \bibinfo{person}{Hao Zhu}, \bibinfo{person}{Fangchen Liu}, \bibinfo{person}{Minghua Liu}, \bibinfo{person}{Hanxiao Jiang}, \bibinfo{person}{Yifu Yuan}, \bibinfo{person}{He Wang}, \bibinfo{person}{Li Yi}, \bibinfo{person}{Angel~X. Chang}, \bibinfo{person}{Leonidas~J. Guibas}, {and} \bibinfo{person}{Hao Su}.} \bibinfo{year}{2020}\natexlab{}.
\newblock \showarticletitle{SAPIEN: A SimulAted Part-Based Interactive ENvironment}. In \bibinfo{booktitle}{\emph{Proceedings of the IEEE/CVF Conference on Computer Vision and Pattern Recognition}}. \bibinfo{pages}{11094--11104}.
\newblock
\href{https://doi.org/10.1109/CVPR42600.2020.01111}{doi:\nolinkurl{10.1109/CVPR42600.2020.01111}}


\bibitem[Yang et~al\mbox{.}(2024)]%
        {yang2024sampart3d}
\bibfield{author}{\bibinfo{person}{Yunhan Yang}, \bibinfo{person}{Yukun Huang}, \bibinfo{person}{Yuan-Chen Guo}, \bibinfo{person}{Liangjun Lu}, \bibinfo{person}{Xiaoyang Wu}, \bibinfo{person}{Edmund~Y. Lam}, \bibinfo{person}{Yan-Pei Cao}, {and} \bibinfo{person}{Xihui Liu}.} \bibinfo{year}{2024}\natexlab{}.
\newblock \bibinfo{title}{SAMPart3D: Segment Any Part in 3D Objects}.
\newblock
\showeprint[arxiv]{2411.07184}~[cs.CV]


\bibitem[Yu et~al\mbox{.}(2023)]%
        {yu2023hal3d}
\bibfield{author}{\bibinfo{person}{Fenggen Yu}, \bibinfo{person}{Yiming Qian}, \bibinfo{person}{Francisca Gil-Ureta}, \bibinfo{person}{Brian Jackson}, \bibinfo{person}{Eric Bennett}, {and} \bibinfo{person}{Hao Zhang}.} \bibinfo{year}{2023}\natexlab{}.
\newblock \showarticletitle{HAL3D: Hierarchical Active Learning for Fine-Grained 3D Part Labeling}. In \bibinfo{booktitle}{\emph{Proceedings of the IEEE/CVF International Conference on Computer Vision}}. \bibinfo{pages}{865--875}.
\newblock
\href{https://doi.org/10.1109/ICCV51070.2023.00086}{doi:\nolinkurl{10.1109/ICCV51070.2023.00086}}


\bibitem[Zhang et~al\mbox{.}(2023)]%
        {zhang2023adding}
\bibfield{author}{\bibinfo{person}{Lvmin Zhang}, \bibinfo{person}{Anyi Rao}, {and} \bibinfo{person}{Maneesh Agrawala}.} \bibinfo{year}{2023}\natexlab{}.
\newblock \showarticletitle{Adding Conditional Control to Text-to-Image Diffusion Models}. In \bibinfo{booktitle}{\emph{Proceedings of the IEEE/CVF International Conference on Computer Vision}}. \bibinfo{pages}{3813--3824}.
\newblock
\href{https://doi.org/10.1109/ICCV51070.2023.00355}{doi:\nolinkurl{10.1109/ICCV51070.2023.00355}}


\bibitem[Zhang et~al\mbox{.}(2022)]%
        {zhang2022point}
\bibfield{author}{\bibinfo{person}{Renrui Zhang}, \bibinfo{person}{Ziyu Guo}, \bibinfo{person}{Rongyao Fang}, \bibinfo{person}{Bin Zhao}, \bibinfo{person}{Dong Wang}, \bibinfo{person}{Yu Qiao}, \bibinfo{person}{Hongsheng Li}, {and} \bibinfo{person}{Peng Gao}.} \bibinfo{year}{2022}\natexlab{}.
\newblock \showarticletitle{Point-M2AE: Multi-scale Masked Autoencoders for Hierarchical Point Cloud Pre-training}.
\newblock \bibinfo{journal}{\emph{Advances in Neural Information Processing Systems}}  \bibinfo{volume}{35} (\bibinfo{year}{2022}), \bibinfo{pages}{27061--27074}.
\newblock


\bibitem[Zhong et~al\mbox{.}(2024)]%
        {zhong2024meshsegmenter}
\bibfield{author}{\bibinfo{person}{Ziming Zhong}, \bibinfo{person}{Yanyu Xu}, \bibinfo{person}{Jing Li}, \bibinfo{person}{Jiale Xu}, \bibinfo{person}{Zhengxin Li}, \bibinfo{person}{Chaohui Yu}, {and} \bibinfo{person}{Shenghua Gao}.} \bibinfo{year}{2024}\natexlab{}.
\newblock \showarticletitle{MeshSegmenter: Zero-Shot Mesh Semantic Segmentation via Texture Synthesis}. In \bibinfo{booktitle}{\emph{European Conference on Computer Vision}}. Springer, \bibinfo{pages}{182--199}.
\newblock
\href{https://doi.org/10.1007/978-3-031-72980-5_11}{doi:\nolinkurl{10.1007/978-3-031-72980-5_11}}


\bibitem[Zhou et~al\mbox{.}(2024)]%
        {zhou2024image}
\bibfield{author}{\bibinfo{person}{Tianfei Zhou}, \bibinfo{person}{Wang Xia}, \bibinfo{person}{Fei Zhang}, \bibinfo{person}{Boyu Chang}, \bibinfo{person}{Wenguan Wang}, \bibinfo{person}{Ye Yuan}, \bibinfo{person}{Ender Konukoglu}, {and} \bibinfo{person}{Daniel Cremers}.} \bibinfo{year}{2024}\natexlab{}.
\newblock \bibinfo{title}{Image Segmentation in Foundation Model Era: A Survey}.
\newblock
\showeprint[arxiv]{2408.12957}~[cs.CV]


\end{thebibliography}

\end{document}